\documentclass[preprint,times,final,nopreprintline,authoryear]{elsarticle}
\pdfoutput=1 

\usepackage{xifthen}
\newboolean{arxivpreprint}
\setboolean{arxivpreprint}{True}
\newboolean{myreview}
\setboolean{myreview}{False}
\newboolean{comments}
\setboolean{comments}{False}
\newboolean{draftimg}
\setboolean{draftimg}{False}

\ifthenelse{\boolean{draftimg}}{%
  \usepackage[allfiguresdraft]{draftfigure}
}{}
  
\usepackage{framed,multirow}

\ifthenelse{\boolean{arxivpreprint} \OR \boolean{myreview}}{%
  \setboolean{comments}{False}
}{}

\usepackage{amssymb}
\usepackage{latexsym}

\usepackage{url}
\usepackage[table,svgnames]{xcolor}

\input{prelude.part}

\usepackage{hyperref}

\definecolor{newcolor}{rgb}{.8,.349,.1}

\ifthenelse{\boolean{arxivpreprint}}{}{%
  \journal{Computerized Medical Imaging and Graphics}
}

\ifthenelse{\boolean{arxivpreprint}}{%
  \setboolean{myreview}{False} 
}{}

\ifthenelse{\boolean{myreview}}{%
  \usepackage[switch]{lineno}
}{}

\newcommand{\snm}[1]{#1}
\newcommand{\KWD}{}

\begin{document}

\newboolean{perpetualfalse}
\setboolean{perpetualfalse}{False}

\begin{frontmatter}

\title{Registration of serial sections: An~evaluation method based on distortions of the ground truths}
\author[1,2]{Oleg~\snm{Lobachev}\corref{cor1}}
\cortext[cor1]{Corresponding author
}
\ead{oleg.lobachev@leibniz-fh.de}
\ead[orcid]{https://orcid.org/0000-0002-7193-6258}
\ead[url]{https://leibniz-fh.de/mitarbeiter/oleg-lobachev/}
\author[3]{Takuya~\snm{Funatomi}}
\ifthenelse{\boolean{arxivpreprint}}{}{%
	\ead[orcid]{https://orcid.org/0000-0001-5588-5932}
}
\author[1]{Alexander~\snm{Pfaffenroth}}
\author[7]{Reinhold~\snm{Förster}}
\author[1,4]{Lars~\snm{Knudsen}}
\ifthenelse{\boolean{arxivpreprint}}{}{%
}
\author[1,4,8]{Christoph~\snm{Wrede}}
\author[14]{Michael~\snm{Guthe}}
\ifthenelse{\boolean{arxivpreprint}}{}{%
}
\author[5]{David~\snm{Haberthür}}
\ifthenelse{\boolean{arxivpreprint}}{}{%
	\ead[orcid]{https://orcid.org/0000-0003-3388-9187}
}
\author[5]{Ruslan~\snm{Hlushchuk}}
\ifthenelse{\boolean{arxivpreprint}}{}{%
	\ead[orcid]{https://orcid.org/0000-0002-6722-8996}
}
\author[9]{Thomas~\snm{Salaets}}
\author[9]{Jaan~\snm{Toelen}}
\author[16]{Simone~\snm{Gaffling}}
\ifthenelse{\boolean{arxivpreprint}}{}{%
  \ead[orcid]{https://orcid.org/0000-0002-7811-6180}
}
\author[1,4,8]{Christian~\snm{Mühlfeld}}
\ifthenelse{\boolean{arxivpreprint}}{}{%
}
\author[1,18]{Roman~\snm{Grothausmann}}
\ifthenelse{\boolean{arxivpreprint}}{}{%
  \ead[orcid]{https://orcid.org/0000-0001-5550-4239}
}

\address[1]{Hannover Medical School, Institute of Functional and Applied Anatomy, OE~4120, Carl-Neuberg-Straße~1, 30625 Hannover, Germany}
\address[7]{Hannover Medical School, Institute of Immunology, OE~5240, Carl-Neuberg-Straße~1, 30625 Hannover, Germany}
\address[4]{Biomedical Research in Endstage and Obstructive Lung Disease Hannover (BREATH), Member of the German Center for Lung Research (DZL), Hannover, Germany}
\address[8]{Hannover Medical School, Research Core Unit Electron Microscopy, OE~8840, Carl-Neuberg-Straße~1, 30625 Hannover, Germany}
\address[2]{Leibniz-Fachhochschule School of Business, Expo Plaza~11, 30539 Hannover, Germany}
\address[3]{Nara Institute of Science and Technology, 8916-5 Takayama-cho, Ikoma, Nara 630-0192, Japan}
\address[14]{University of Bayreuth, 95440 Bayreuth, Germany}
\address[5]{University of Bern, Institute of Anatomy, Baltzerstrasse~2, 3012 Bern, Switzerland}
\address[9]{KU Leuven, Herestraat~49, 3000 Leuven, Belgium}
\address[16]{Chimaera GmbH, Am Weichselgarten~7, 91058 Erlangen, Germany}
\address[18]{HAWK University of Applied Sciences and Arts, Faculty of Engineering and Health, Von-Ossietzky-Str.~99, 37085 Göttingen, Germany}

\begin{abstract}
Registration of histological serial sections is a challenging task.
Serial sections exhibit distortions and damage from sectioning.
Missing information on how the tissue looked before cutting makes a realistic validation  of 2D registrations extremely difficult.

This work proposes methods for ground-truth-based evaluation of registrations.
Firstly, we present a methodology to generate test data for registrations.
We distort an innately registered image stack in the manner similar to the cutting distortion of serial sections.
Test cases are generated from existing 3D data sets, thus the ground truth is known.
Secondly, our test case generation premises evaluation of the registrations with known ground truths.  Our methodology for such an evaluation technique distinguishes this work from other approaches. Both under- and over-registration become evident in our evaluations. We also survey existing validation efforts.

We present a full-series evaluation across six different registration
methods applied to our distorted 3D data sets of animal lungs.
Our distorted and ground truth data sets are made publicly available.
\end{abstract}

\end{frontmatter}

\ifthenelse{\boolean{arxivpreprint}}{%
  \blfootnote{\ccbyncndeu~
    Licensed under  Creative Commons Attribution-NonCommercial-NoDerivatives 4.0 International licence.}}{}


\ifthenelse{\boolean{myreview}}{%
  \linenumbers
}{}

\newcommand{\elastix}{{Elastix}\xspace}
\newcommand{\activecontour}{\textsf{activecontour}\xspace}
\newcommand{\mytimes}{\ensuremath{\mskip\medmuskip\times\mskip\medmuskip}}
\newcommand{\kroi}[2]{\textup{#1}\textit{k}$\mytimes$\textup{#2}\textit{k}~pixels}
\newcommand{\xroi}[2]{\textup{#1}$\mytimes$\textup{#2}~pixels}

\hypertarget{introduction}{%
\section{Introduction}\label{introduction}}

Microscopy has a long tradition, and microscopic imaging is still one of
the most frequently used and powerful tools in biomedical research. From
light microscopic (LM) techniques, including conventional fluorescent
stainings, to transmission and scanning electron microscopic (EM)
methods, the last two decades have witnessed substantial methodological
progress in terms of resolution, speed, and automation. Tissue clearing
and super resolution LM at the one end, and serial block-face as well as
focused ion beam scanning EM at the other end, have paved the way for a
three-dimensional visualization of biological specimens.

Still, the use of serial sections remains an essential and
cost-efficient tool to gain 3D insight into specimens for several
reasons: Despite the progress in LM techniques the penetration depth of
staining solutions, in particular fluorescent antibody staining, is
limited, thus limiting the size of the sample that can be visualized.
Genetically modified organisms, such as mice, expressing fluorescent
proteins under cell-specific promoters, are available. This method,
though, cannot be applied to human samples for obvious reasons. Thus,
the use of serial sections in LM is often the method of choice for
obtaining 3D information both from conventional and fluorescent
microscopic imaging, in particular when human samples are investigated.

However, microscopic sections are inherently two-dimensional (2D) and
their 3D information has to be regained from 2D images. The manual or
automated cutting of thin sections for microscopy, however,
induces---even in perfectly trained and experienced hands---varying
distortions and deformations, such as stretching or compression.
Positioning the sections on the glass slides further contributes to
spatial distortion. This problem is especially evident in large
sections. Further processing such as antigen retrieval and staining may
further damage the section. Even digitization of the sections can be
faulty and create partially corrupt representations.

The alignment or registration is an important method in medical image
processing. An absence of the ground truth is a major problem during the
development of new and fine-tuning of existing registration methods.
While some simple synthetic data or phantoms can be generated, those
would not adequately represent the problem.
\olcomment{this might be too harsh, but I also found no papers of mocking up serial sectioning!}
Firstly, some registration methods, for example, those based on feature
detection, thrive from complexity of the input data. Such
\enquote{sparse} methods might perform very well especially in the lung
tissue, where the fraction of empty space is very high. Secondly, the
distortions in phantom data might not truly represent the distortions in
serial sections. Thirdly, in most real cases, no ground truths from
other modalities exist in the typical acquisition resolution of the
serial sections. Most \enquote{real} 3D methods either do not reach the
resolution of conventional LM (\eg micro-CT) or have typically much
smaller spatial dimensions of the probe (\eg nano-CT, EM).
Aforementioned LS microscopy and tissue clearing are possible
palliatives in model animals, but all those methods are still too
complex, too expensive, or require a radically different biological
processing pipeline that makes it impossible to apply both modalities to
the same specimen. It is also much harder to apply aforementioned
advanced methods in humans.
\dhcomment{Why? Because of the issues mentioned above? Maybe we could quickly reiterate.}

\hypertarget{contributions}{%
\subsection{Contributions}\label{contributions}}

\begin{figure}[tb]
\centering
\subfloat[][Input]{\label{fig:teaser-a}\includegraphics[width=0.249\linewidth]{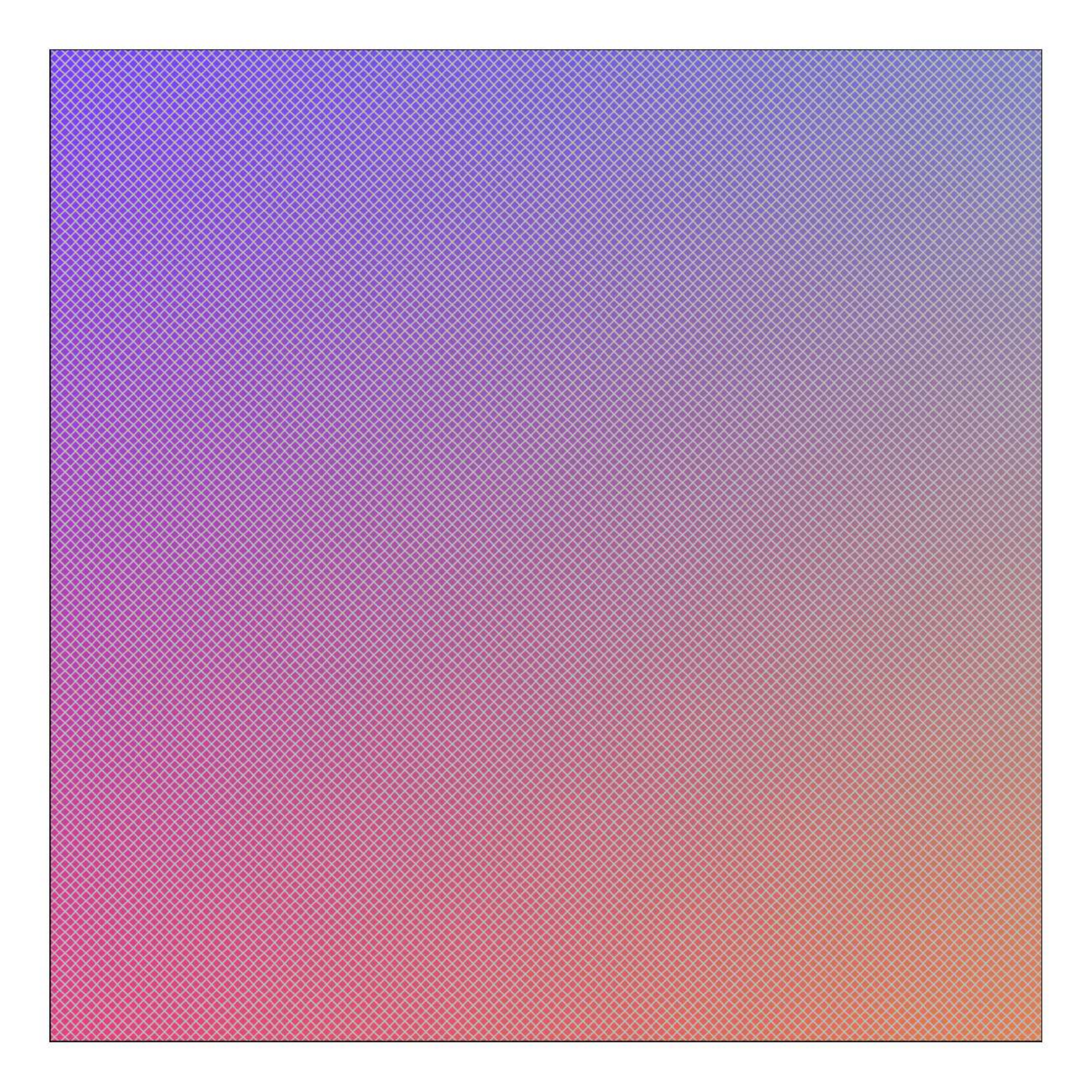}}\hfill%
\subfloat[][Local distortions]{\label{fig:teaser-b}\includegraphics[width=0.249\linewidth]{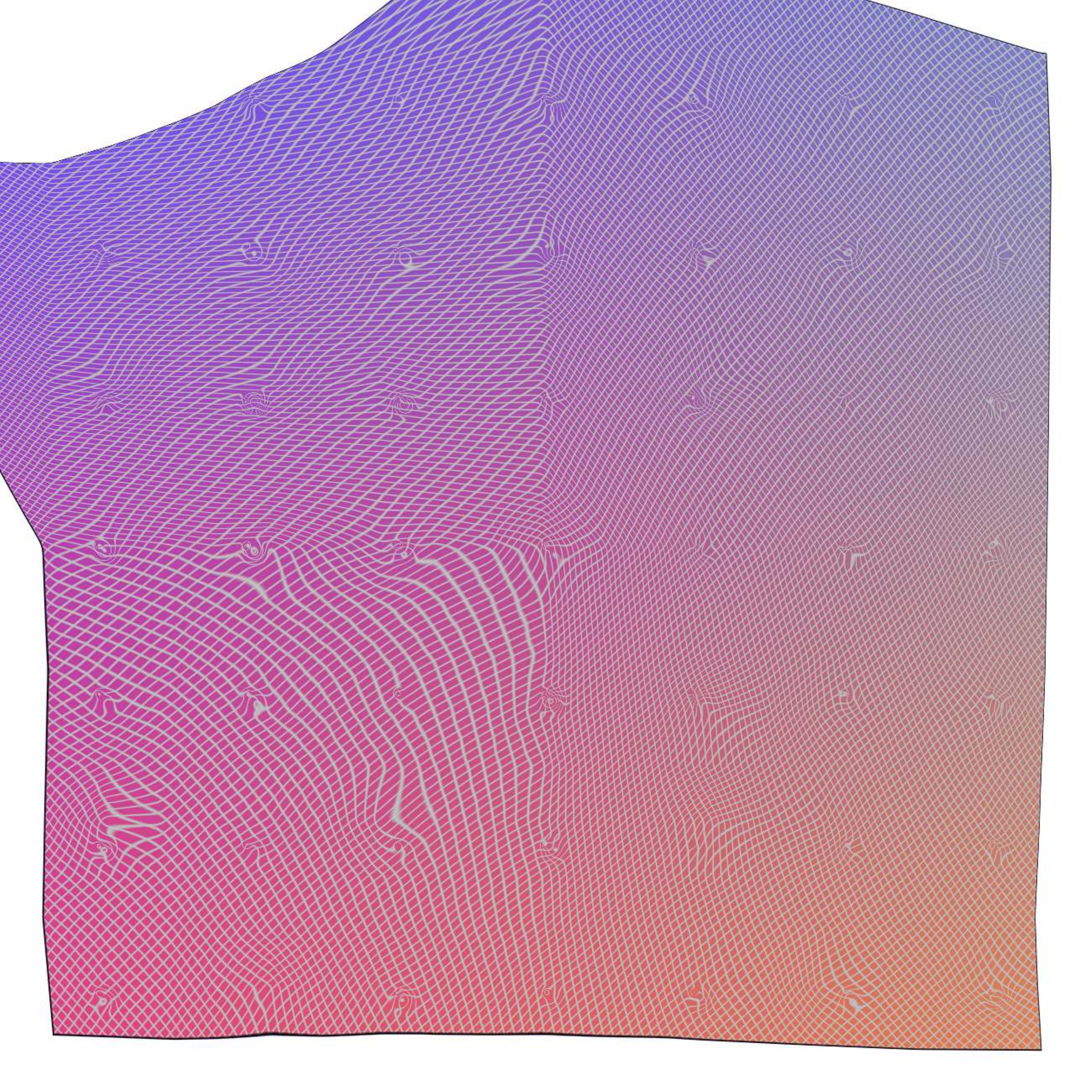}}\hfill%
\subfloat[][Distortion map]{\label{fig:teaser-c}\includegraphics[width=0.249\linewidth]{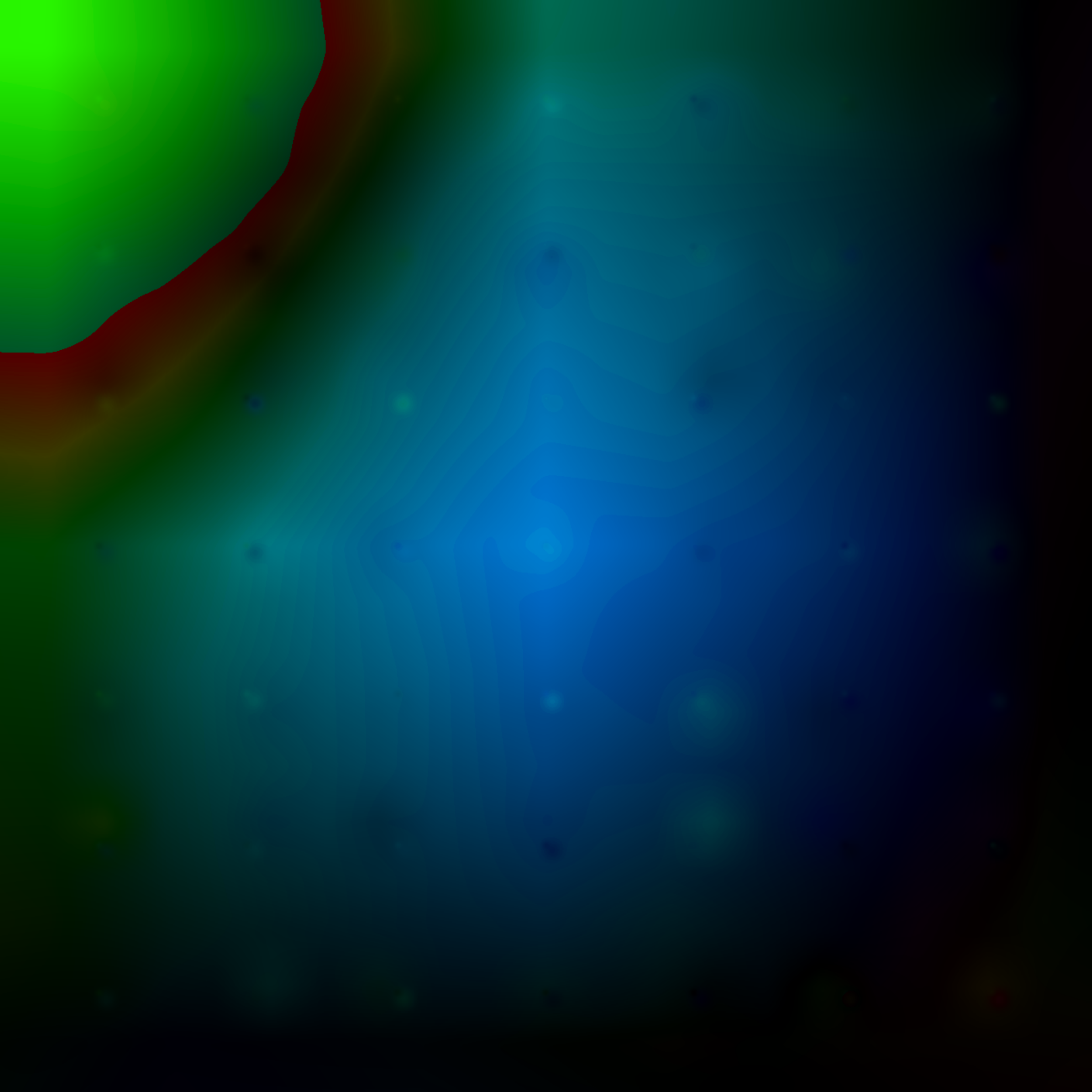}}\hfill%
\subfloat[][Rigid transform]{\label{fig:teaser-d}\includegraphics[width=0.249\linewidth]{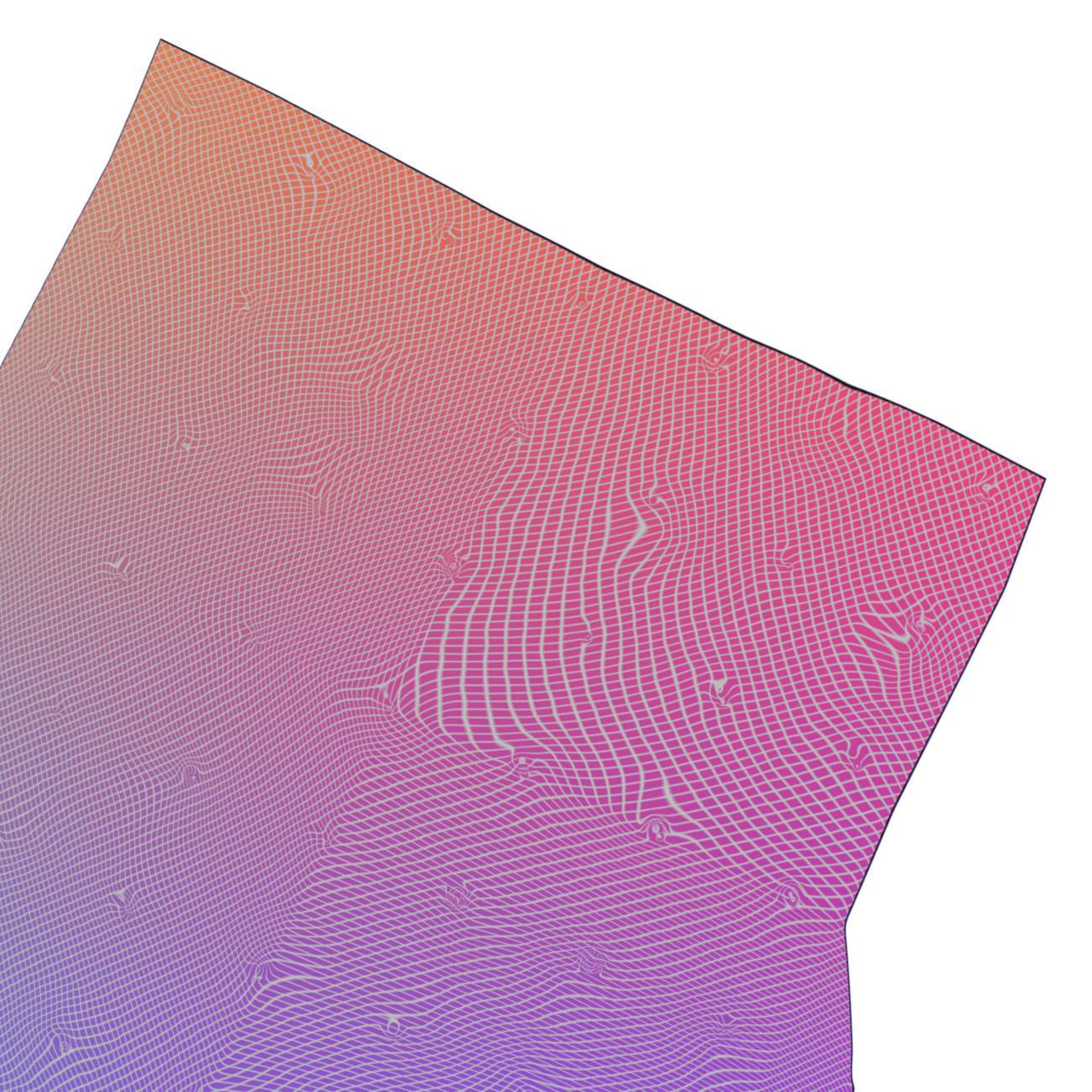}}
\caption{Showcasing our method on a synthetic image of a color gradient. The distortion magnitude is increased tenfold for demonstration. The images are slightly cropped for presentation.}
\label{fig:teaser}
\end{figure}

\begin{figure}[!t]
\centering
\subfloat[][Input image]{\label{fig:ship-a}\includegraphics[width=0.33\linewidth]{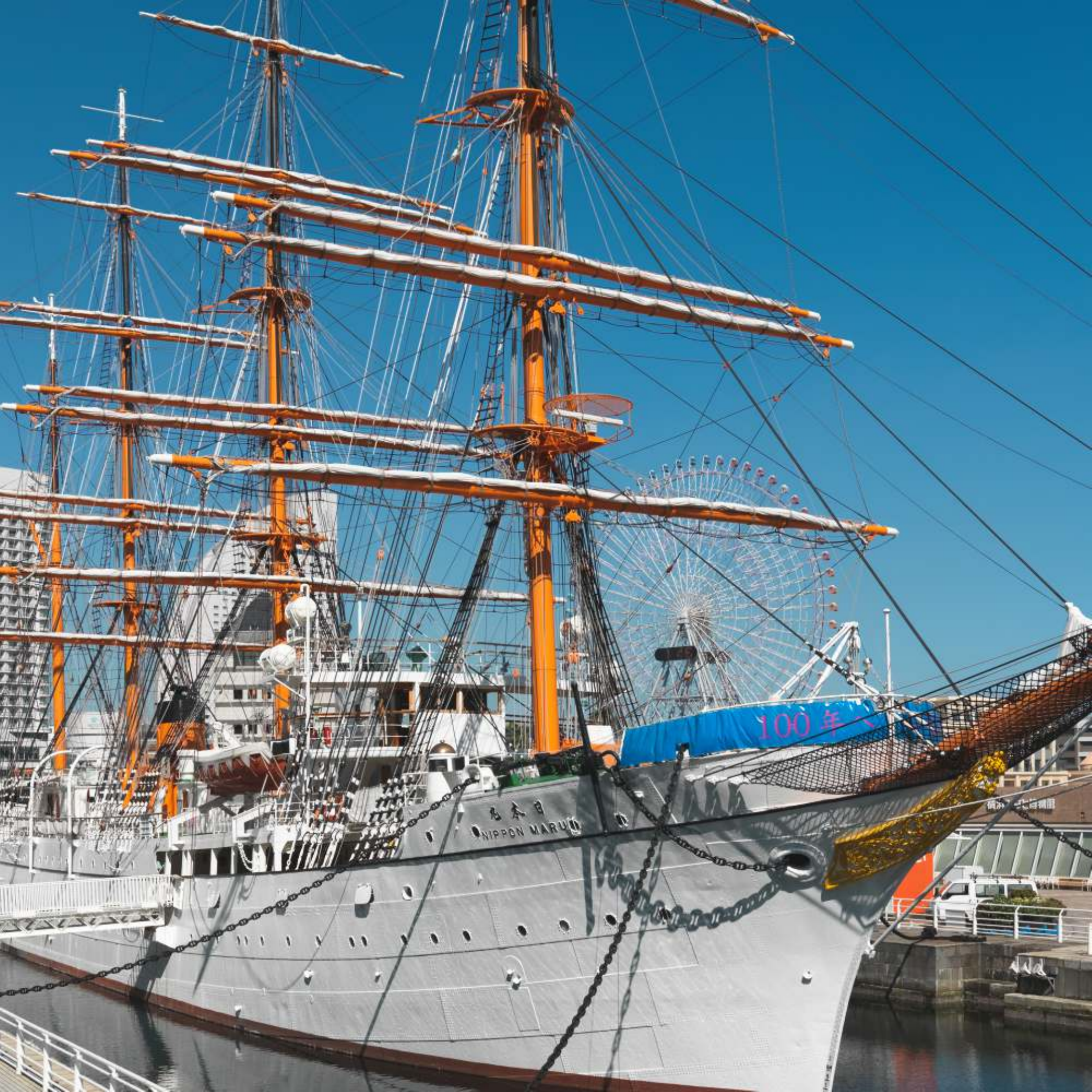}}\hfill
\subfloat[][Extended borders]{\label{fig:ship-b}\includegraphics[width=0.33\linewidth]{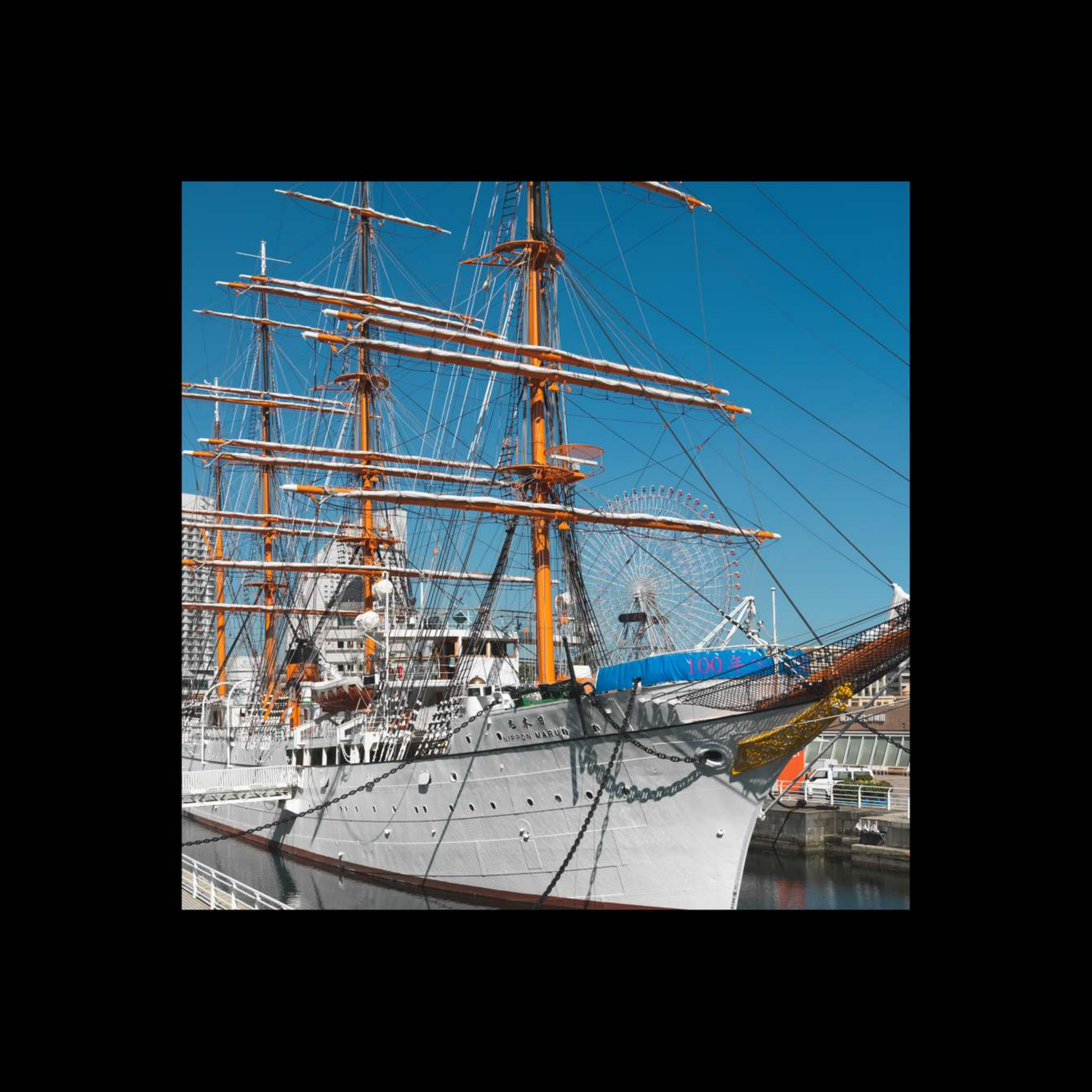}}\hfill
\subfloat[][Local distortions]{\label{fig:ship-c}\includegraphics[width=0.33\linewidth]{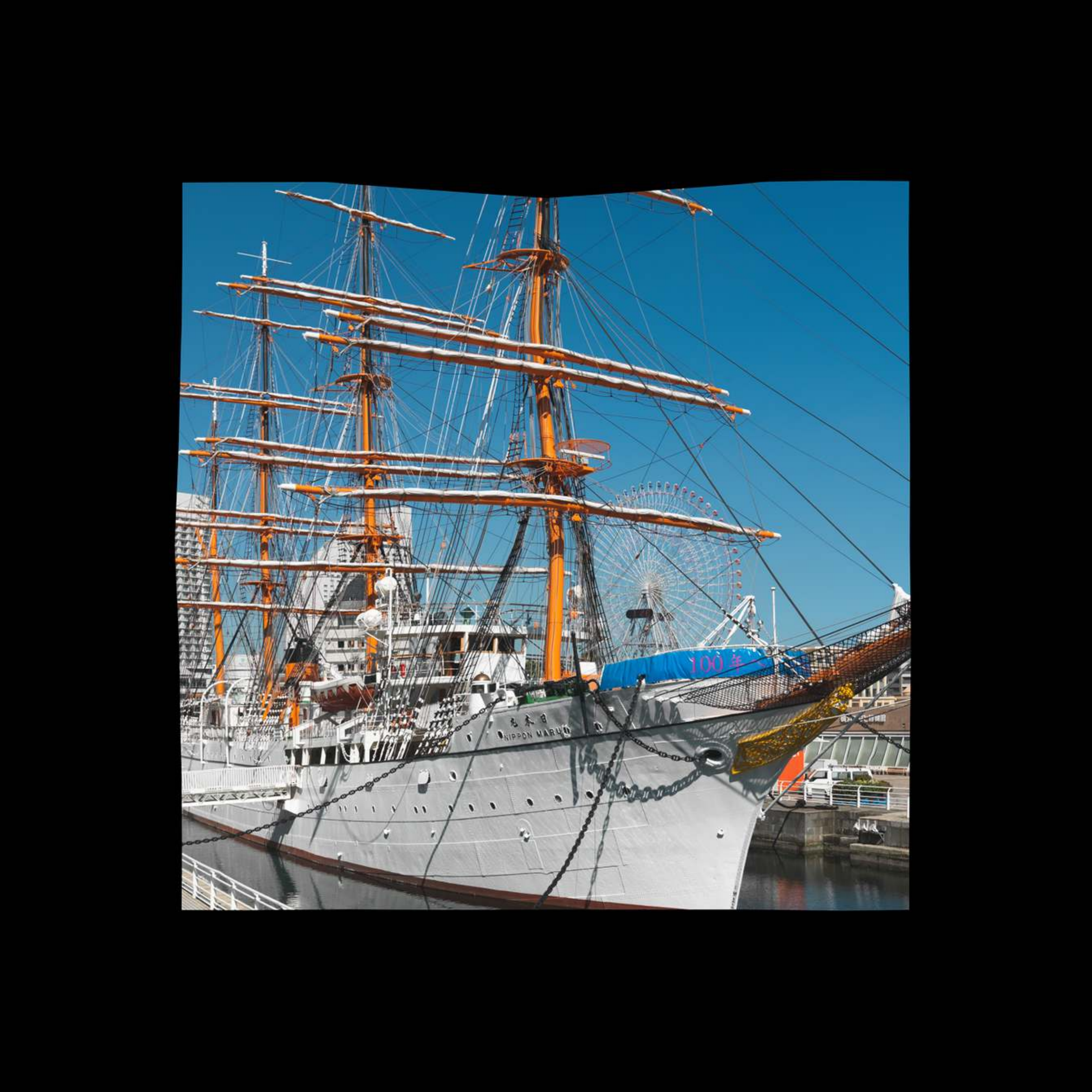}}\hfill

\subfloat[][Visualization of the distortion map for \protect\subref{fig:ship-c}]{\label{fig:ship-d}\includegraphics[width=0.33\linewidth]{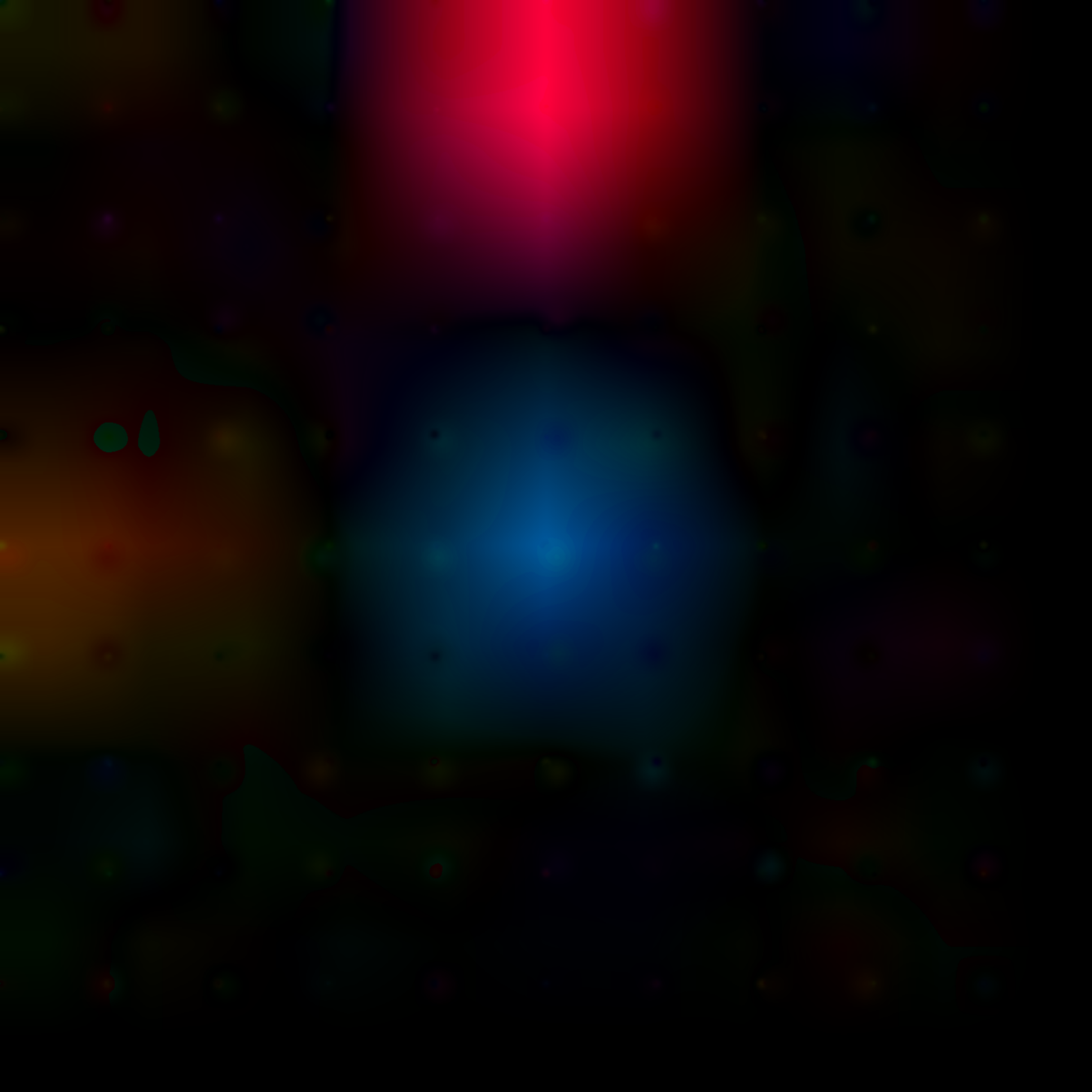}}\hfill
\subfloat[][Center from~\subref{fig:ship-b}]{\label{fig:ship-b2}\includegraphics[width=0.33\linewidth]{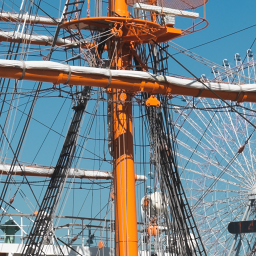}}\hfill
\subfloat[][Center from~\subref{fig:ship-c}]{\label{fig:ship-c2}\includegraphics[width=0.33\linewidth]{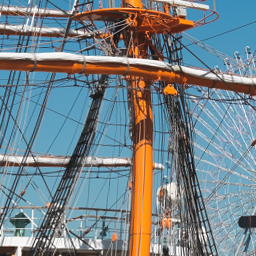}}\hfill

\subfloat[][Center from~\subref{fig:ship-d}]{\label{fig:ship-d2}\includegraphics[width=0.33\linewidth]{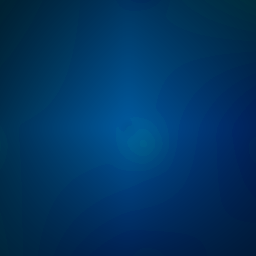}}\hfill
\subfloat[][Optional damage]{\label{fig:ship-e}\includegraphics[width=0.33\linewidth]{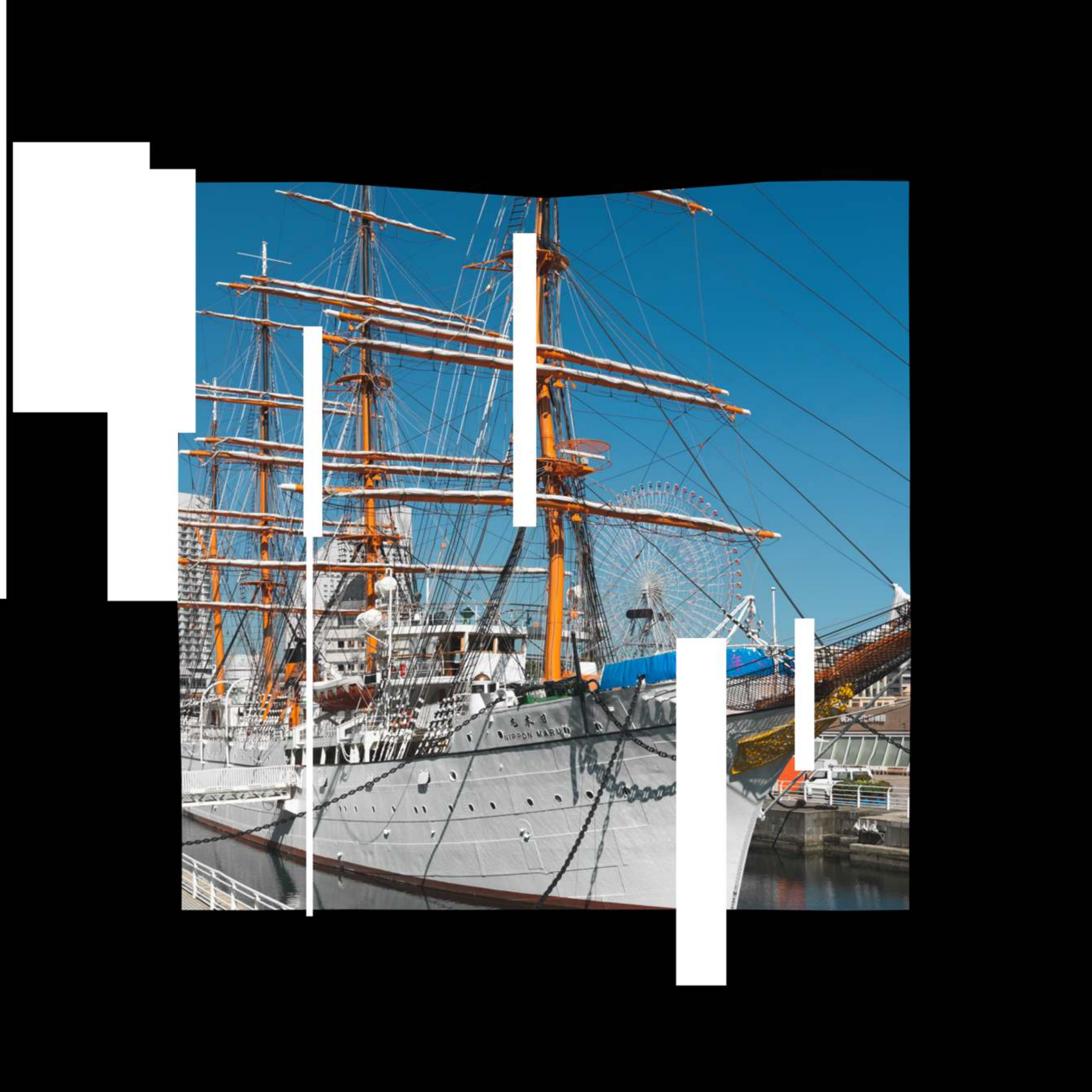}}\hfill
\subfloat[][Highlighted differences]{\label{fig:ship-f}\includegraphics[width=0.33\linewidth]{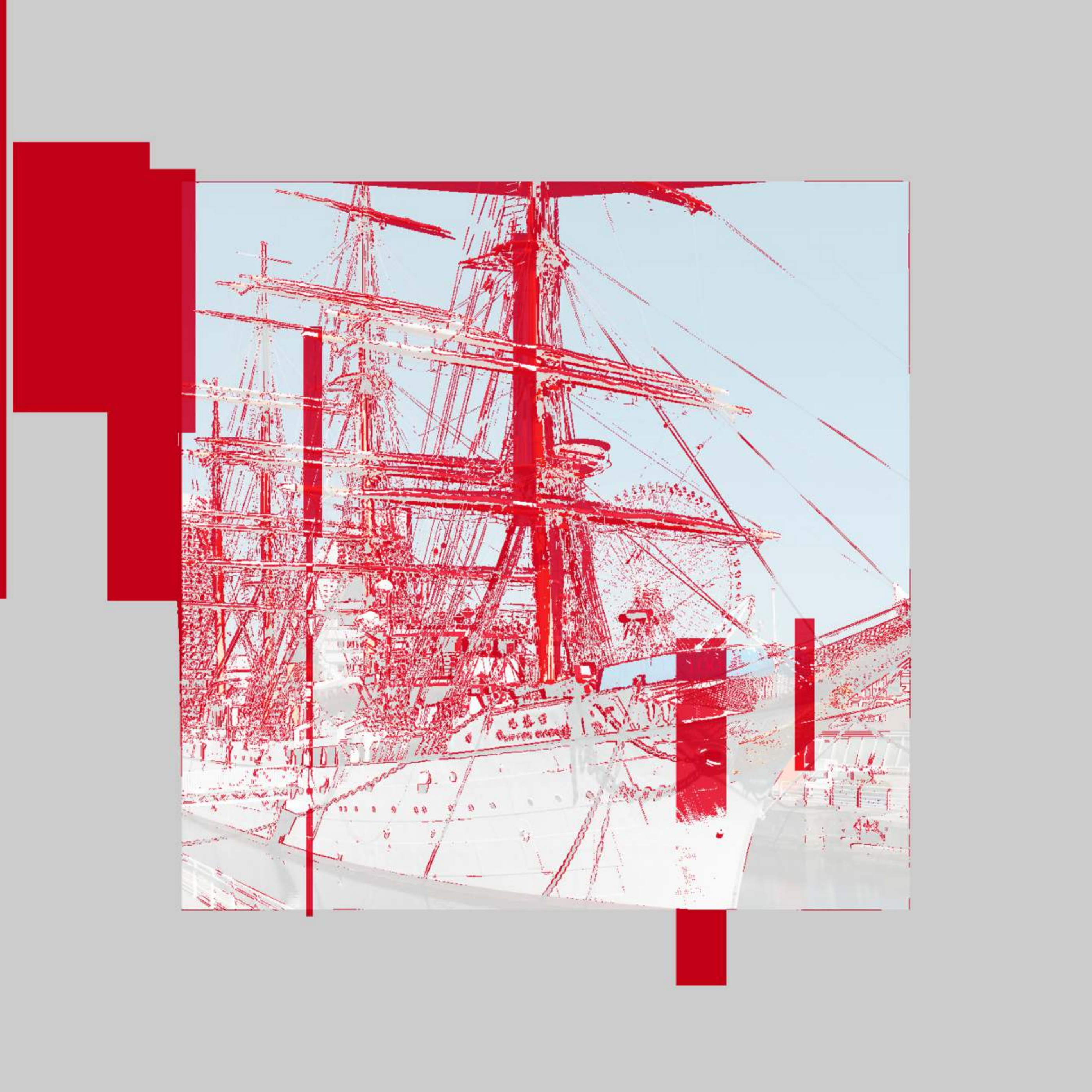}}
\caption{Details of local distortion and damage generation. Our method works also on general-purpose images. The test image is from the Japanese ITE data set of UHD images, we took \kroi{1}{1} crop from the center of the "Ship" image (U10, 2K version). A lot of straight lines allows for good identification of distortions. Global rigid transformation is omitted for its simplicity. The distortion map is in line with our usual settings. The visualization in \protect\subref{fig:ship-f} compares PSNR between \protect\subref{fig:ship-b} and \protect\subref{fig:ship-e}; it is thresholded at 20\%.}
\label{fig:ship}
\end{figure}

In this paper we present a methodology to apply typical sectioning
distortions to real data sets from other modalities. We digitally
\enquote{mock up} the distortions from sectioning on real biological
data. Arbitrary general-purpose images can be distorted
(Fig.~\ref{fig:teaser}, Fig.~\ref{fig:ship}) and registration methods
can be applied to distorted images. The results of the registrations can
be immediately compared with ground truth data. Our benchmark is open to
further registrations, new quality measures, and new images. As we
present the method and not only the data, further data sets, even from
additional modalities, can be produced by others. We focus our current
presentation on animal lung images. However, our method is generic; it
should be applicable to virtually any kind of innately 3D data of any
organ from any species. Our goal is to enable the evaluation of
registration methods for serial sections with a ground truth from real
biological data. To fulfill it, we mimic sectioning distortions in an
artificial, but statistically meaningful and reproducible manner. We
then proceed to evaluate some existing registrations with our method.
Among other approaches we present a full-series evaluation.

In this paper, we consider possible distortions during sectioning and
apply those to 2D series from the innately 3D data. Our data sets
originate from further modalities in bioimaging. The data sets aim to
come close to LM sections of the lung in their scale---on both sides. We
use both CT and LS as coarser scale and EM as a finer scale. As
original, non-distorted data fit perfectly, those serve as a ground
truth.

The contributions of this paper are threefold. Firstly, we provide an
overview over the field with the emphasis on validations of
registration. In most such validations, the problem of an absent ground
truth motivates the search for further methods. Exactly our approach has
been not suggested before. Secondly, we suggest a technique to generate
a benchmark input from existing inherently 3D data. This way, we are,
thirdly, able to compare registration methods on a common foundation by
comparing the registered data with the ground truth. We perform an
extensive image-based statistical evaluation of the full series.

The source code for this paper is available under
https://github.com/olegl/distort, the distorted and ground truth data
sets can be found under https://zenodo.org/record/4282448.

\hypertarget{paper-structure}{%
\subsection{Paper structure}\label{paper-structure}}

The remaining part of the paper is organized as follows: In
Section~\ref{related-work} we survey existing registration methods and
discuss the approaches towards validation of registration.
Section~\ref{methods} elaborates on our approach for generating
distortions. The same section also presents the registration methods we
used in our benchmark. In Section~\ref{results} we present the data sets
and evaluate the results of the registration benchmark. We compare the
registered images with the ground truth in this section. We present
there both image-based evaluations and statistical gauges of the
results. Section~\ref{discussion} discusses possible limitations and
further developments of our method. Section~\ref{conclusions} concludes
the manuscript.

\hypertarget{related-work}{%
\section{Related work}\label{related-work}}

\hypertarget{registration-in-general}{%
\subsection{Registration in general}\label{registration-in-general}}

There is a lot of research on registration, esp.~in the context of
medical imaging. \citet{brown_survey_1992}, \citet{zitova_image_2003}
provide early surveys; \citet{maintz_survey_1998},
\citet{pluim_mutual-information-based_2003},
\citet{el-baz_state-art_2011}, and \citet{oliveira_medical_2014} are
more focused on the medical topics. \citet{viergever_survey_2016} and
\citet{pichat_survey_2018} are recent overviews of the field.
\citet{zitova_mathematical_2019} gives a mathematical overview of the
methods. Although some manual alignment
\citep[\egX ][]{van_krieken_splenic_1985} has been performed in the
past, we ultimately focus on computational methods.

\citet{lester_survey_1999}, \citet{crum_non-rigid_2004},
\citet{sotiras_deformable_2013} survey non-rigid registration methods.
In the context of the registration of serial sections, non-rigid
registration is definitely required, as distortions during sectioning
are non-linear. Even if we currently do not represent tearings or
foldings of the tissue, cutting-induced local distortions are still
present even in the most perfectly prepared sections. (Instead of
tearings and foldings we can easily encompass missing parts of the
images. Salvaging damaged or missing sections is a separate problem in
our eyes.) Non-rigid registration methods include
\citet{rueckert_nonrigid_1999}; \citet{schnabel_validation_2003};
\citet{chui_unified_2003}; \citet{homke_multigrid_2006};
\citet{zhang_non-rigid_2015}. \citet{saalfeld_as-rigid-as-possible_2010}
focused on as-rigid-as-possible registration for EM.

\citet{punithakumar_gpu-accelerated_2017} is a recent example of a
GPU-accelerated registration. \citeauthor{crum_zen_2003}
\citetext{\citeyear{crum_zen_2003}; \citeyear{crum_non-rigid_2004}}
provide an overview of medical image registration, they highlight both
the importance of validation and its difficulty. One of the popular
software packages for registration is \elastix{}
\citep{klein_elastix:_2010, shamonin_fast_2014} and further developments
around it \citep{berendsen_design_2016, marstal_simpleelastix_2016}.
Another popular package is ANTs
\citep{avants_reproducible_2011, avants_insight_2014}.

One of the somewhat frequent ideas is to work with images on multiple
levels \citep[see,
\eg ][]{haber_multilevel_2006, bagci_hierarchical_2012, lorenz_multi-resolution_2012, lobachev_feature-based_2017}.

A~kind of \enquote{sparse} methods involves feature detection and
description. \citet{ma_image_2020} presents a recent survey of the
field. The actual detectors and descriptors include SIFT
\citep{lowe_object_1999, lowe_distinctive_2004}, SURF
\citep{bay_surf:_2006, bay_speeded-up_2008}, AKAZE
\citep{fitzgibbon_kaze_2012, alcantarilla_fast_2013}. A~basic
\enquote{sparse} registration identifies distinctive regions of both
input images and then computes a correspondence between them based on
the correspondence of the regions alone. In such a rigid registration
RANSAC \citep{fischler_random_1981} is used. \enquote{Sparse} methods
have been used to register medical images
\citep{can_feature-based_2002, urschler_sift_2006, ruiz_non-rigid_2009, sargent_feature_2009, han_feature-constrained_2010, wu_feature-based_2012, ulrich_imaging_2014, shojaii_optimized_2016, jamil_image_2017, lobachev_feature-based_2017, wang_visual_2020}.
\citet{argandacarreras_3d_2010} is the origin of ImageJ's
\enquote{Register virtual stack slices} implementation. (ImageJ
\citep{schneider_nih_2012} and Fiji \citep{schindelin_fiji:_2012} have
served as a basis for many registration and analysis approaches.) They
focus strongly on various rigid approaches, although an elastic
extension exists. \citeauthor{argandacarreras_3d_2010} use micro-CT for
quality control, they utilized Hausdorff distance as a measure. The
ImageJ plugin \enquote{TrakEM2} \citep{cardona_trakem2_2012} also
utilizes feature detection, but it not only performs registration, but
also includes tools for 3D modeling, editing, and annotation.
\citet{ma_locality_2019} and \citet{zhang_adaptive_2020}, for example,
improve the correspondence of features (\enquote{matching}).
\citet{cieslewski_matching_2019} is an example of an alternative to
feature descriptors. Registration of whole sections
\citep[\egX ][]{mueller_real-time_2011} motivated the usage of feature
detection in \citet{ulrich_imaging_2014}.

Optical flow \citep{lucas_iterative_1981, horn_determining_1981} is a
yet another method to find \enquote{moving parts} in images
\citep{farnebaeck_two-frame_2003, brox_high_2004, brox_large_2009, sun_secrets_2010, revaud_epicflow:_2015}.
One of the most recent methods involves machine learning
\citep{liu_selflow_2019}. Applications of optical flow in medical images
include
\citep{dougherty_alignment_2003, ehrhardt_optical_2007, carata_improving_2013, lobachev_compensating_2017}.
Feature descriptors have been used on dense, optical-flow-like data
\citep{ce_liu_sift_2011, sharma_briefmatch:_2017}.

A~diffusion model based on thermodynamics \citep{thirion_image_1998} is
widely used
\citep{peyrat_registration_2010, cunliffe_lung_2012, cifor_hybrid_2013, lan_non-rigid_2019, zhang_tissue-specific_2021}.
Further registration approaches include graph-cut-based methods
\citep{lombaert_landmark-based_2007}, smoothness assumption
\citep{cifor_smoothness-guided_2011}, higher-order derivatives
\citep{wirtz_robust_2005}, chamfer matching
\citep{becker_automated_2015}, particle swarm optimization
\citep{tang_automatic_2011}, Gauss-Seidel optimization
\citep{gaffling_gauss-seidel_2015}, Markov random fields
\citep{glocker_dense_2008, glocker_deformable_2011, feuerstein_reconstruction_2011},
structural probability maps \citep{muller_deformable_2014},
over-segmentation regularization \citep{navab_liver_2015}, Bayesian
inference and regularization \citep{simpson_probabilistic_2015}, elastic
triangulation of a spring model \citep{saalfeld_elastic_2012}, blending
rigid transforms \citep{kajihara_non-rigid_2019}, empirical mode
decomposition \citep{guryanov_fast_2017}, and remote sensing
\citep{chang_remote_2019}.

Some methods register a complete stack of images at once, this approach
was used, \eg by
\citep{likar_hierarchical_2001, nikou_robust_2003, saalfeld_as-rigid-as-possible_2010, lobachev_feature-based_2017}.

\begin{figure}[tb]
\centering
\subfloat[][Before deformation]{\label{fig:real-effect-a}%
\scalebarme{%
\includegraphics[width=0.249\linewidth]{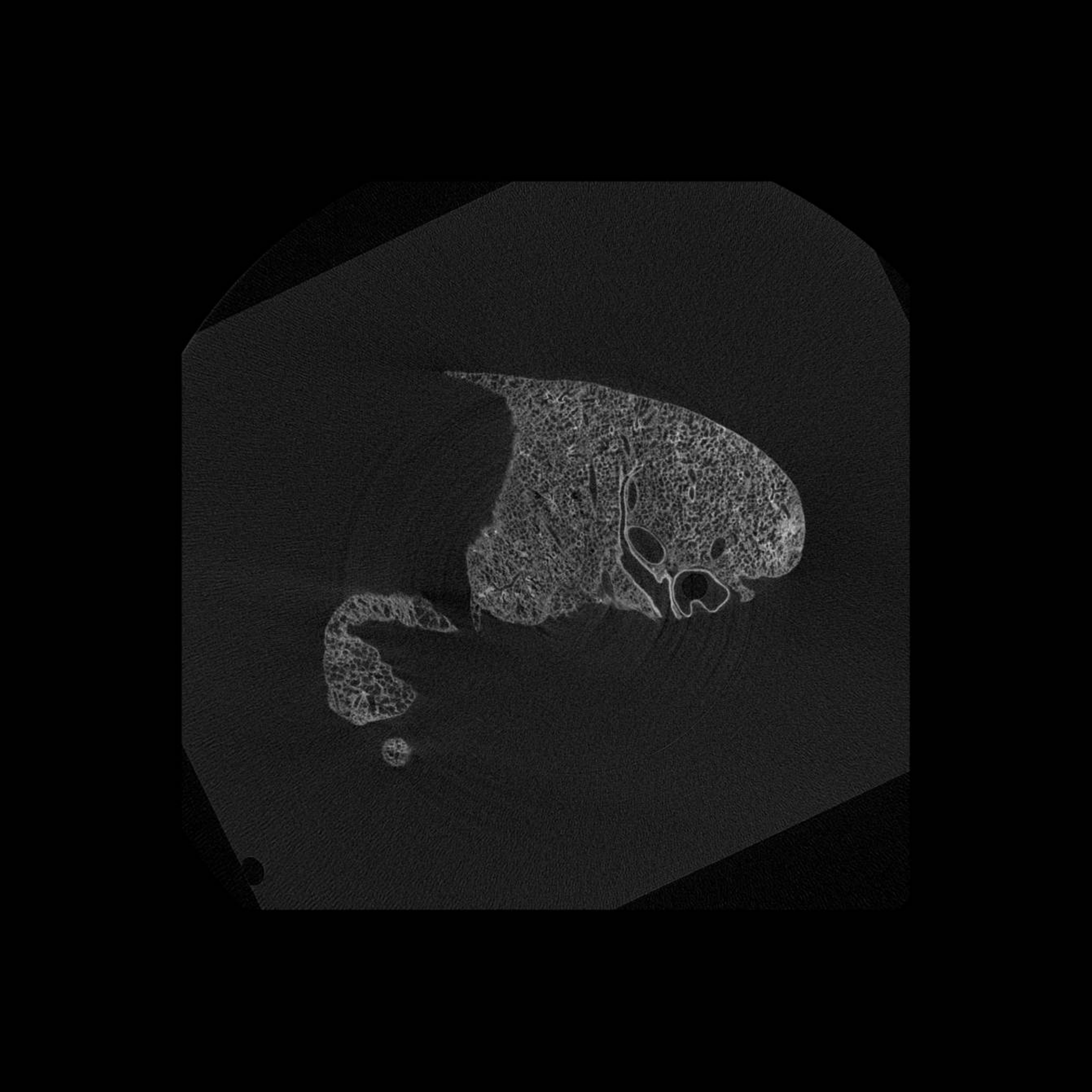}}{%
\includegraphics[width=0.249\linewidth]{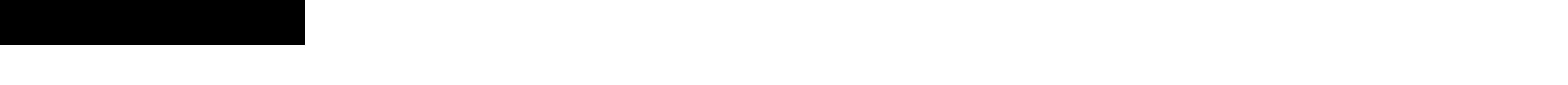} 
}}\hfill%
\subfloat[][Locally deformed]{\label{fig:real-effect-b}%
\scalebarme{%
\includegraphics[width=0.249\linewidth]{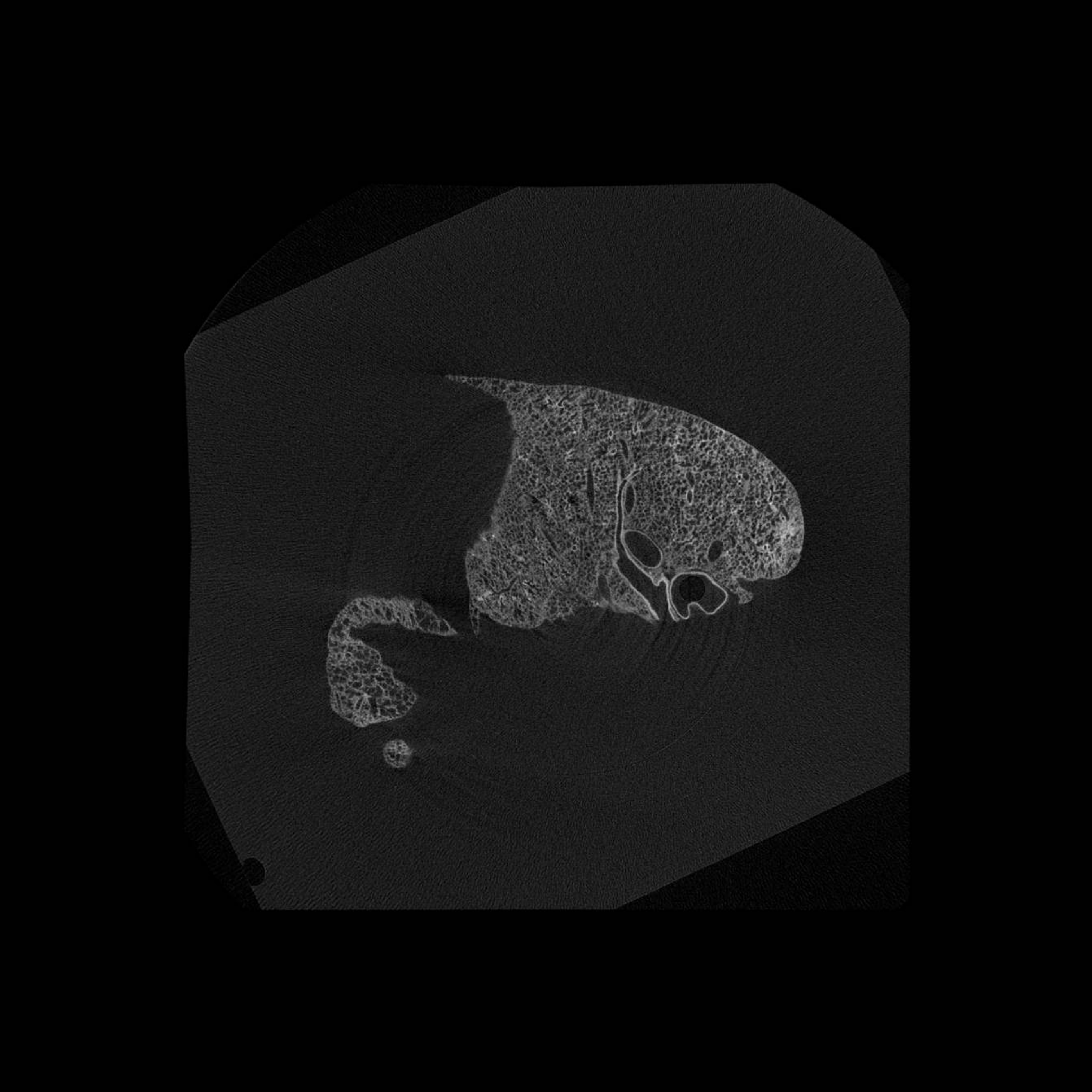}}{%
\includegraphics[width=0.249\linewidth]{image/scalebars/scale_ct-3a.png} 
}}\hfill%
\subfloat[][Differences, PSNR, 5\%]{\label{fig:real-effect-c}%
\scalebarme{%
\includegraphics[width=0.249\linewidth]{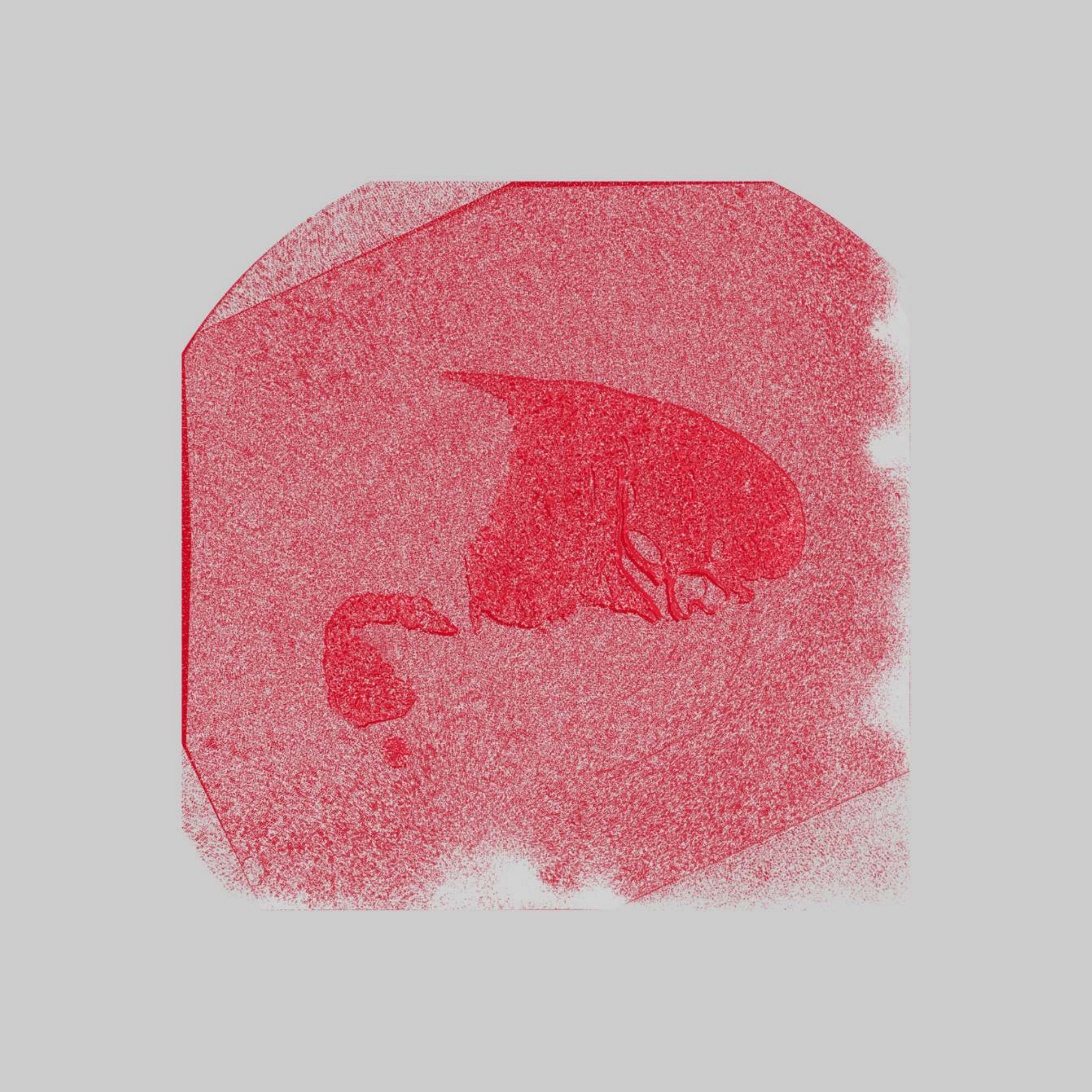}}{%
\includegraphics[width=0.249\linewidth]{image/scalebars/scale_ct-3a.png} 
}}\hfill%
\subfloat[][Differences, PSNR, 20\%]{\label{fig:real-effect-d}\scalebarme{%
\includegraphics[width=0.249\linewidth]{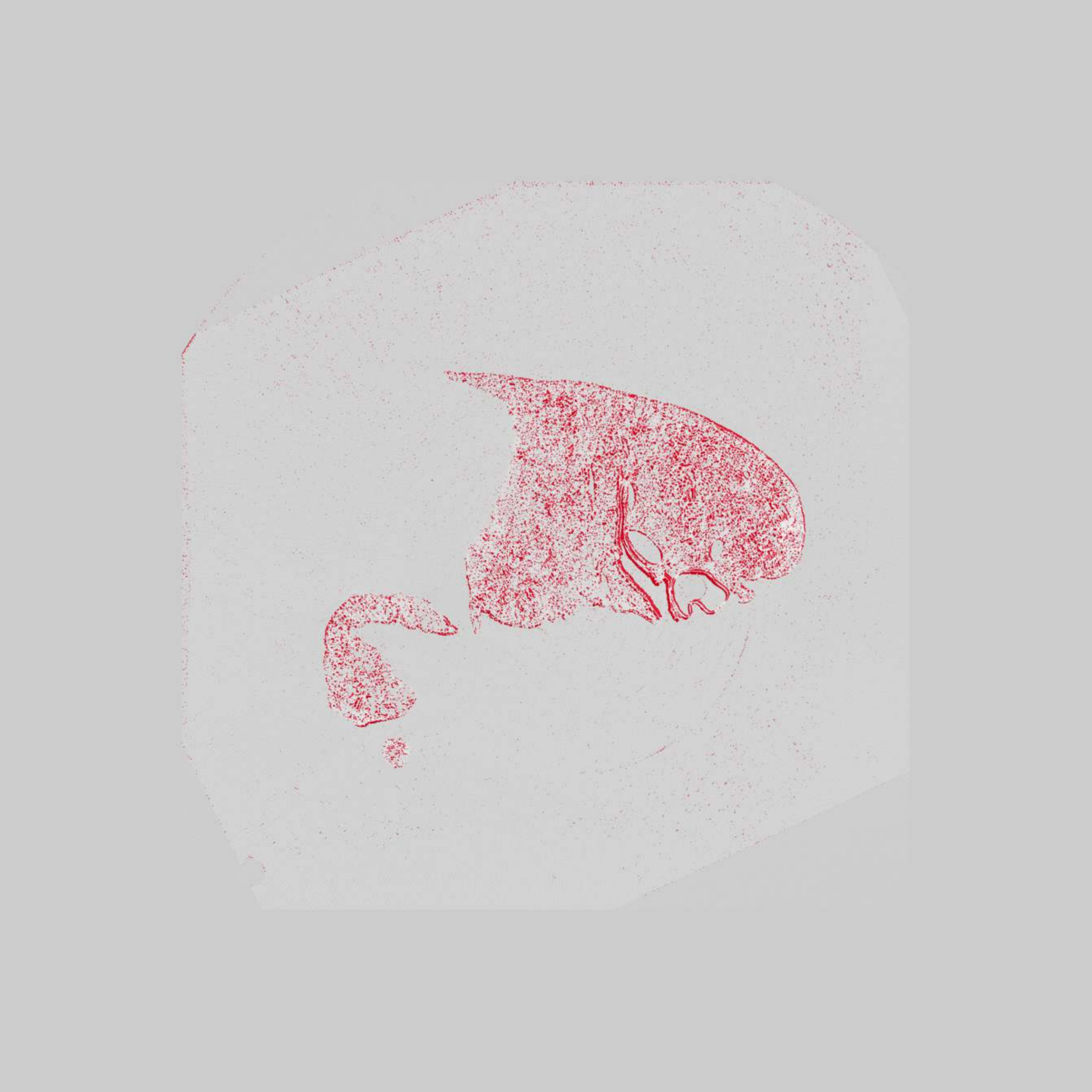}}{%
\includegraphics[width=0.249\linewidth]{image/scalebars/scale_ct-3a.png} 
}}

\subfloat[][Crop from~\protect\subref{fig:real-effect-a}]{\label{fig:real-effect-a2}%
\scalebarme{%
\includegraphics[width=0.249\linewidth]{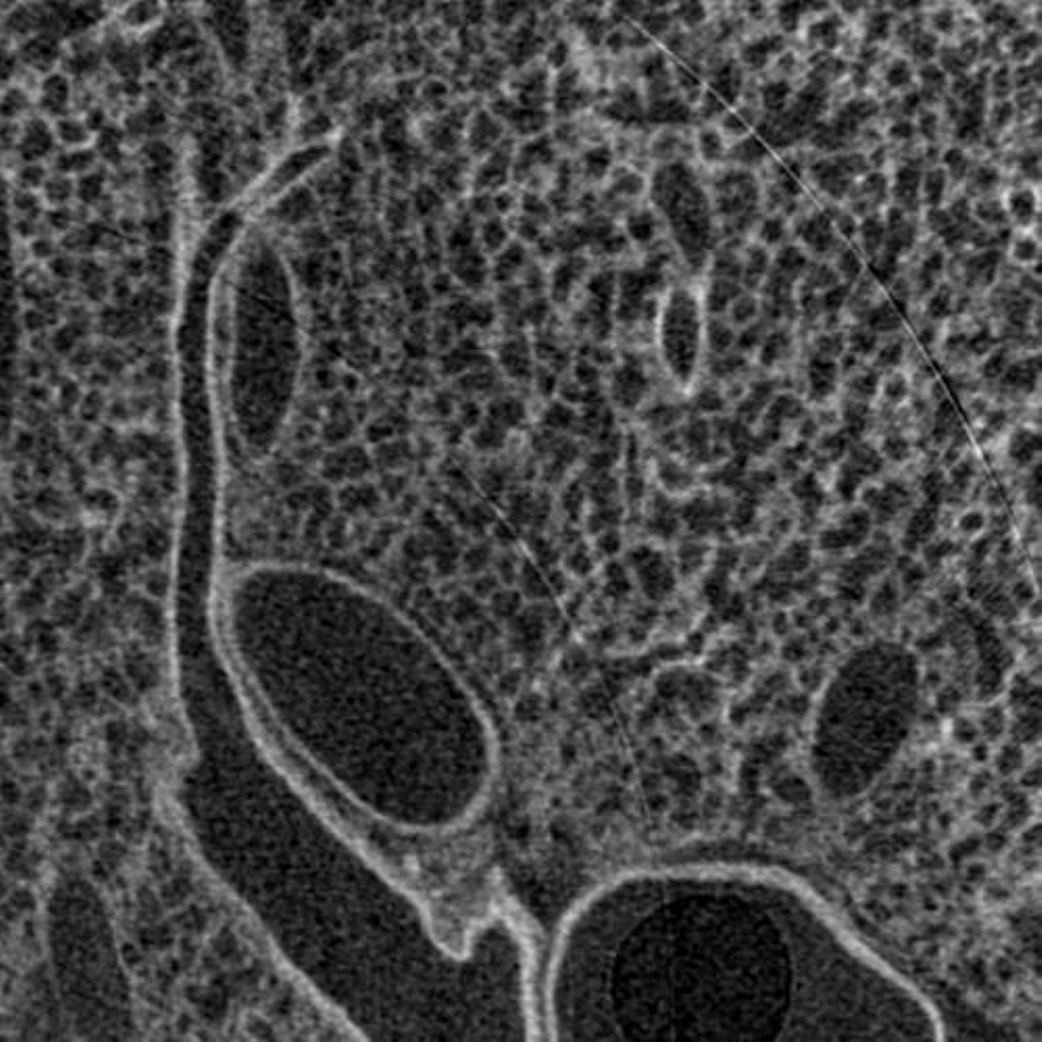}}{%
\includegraphics[width=0.249\linewidth]{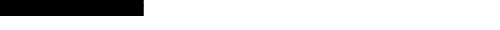} 
}}\hfill%
\subfloat[][Crop from~\protect\subref{fig:real-effect-b}]{\label{fig:real-effect-b2}%
\scalebarme{%
\includegraphics[width=0.249\linewidth]{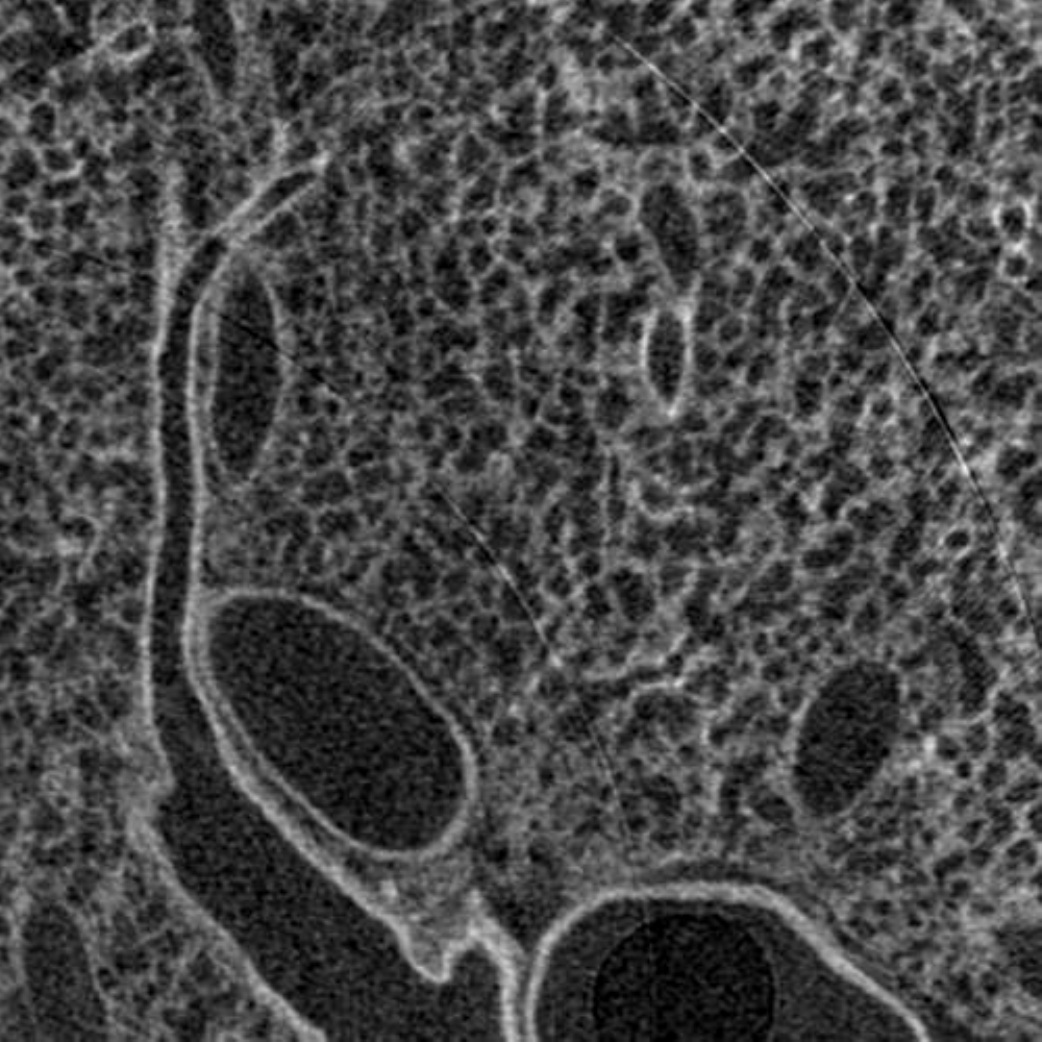}}{%
\includegraphics[width=0.249\linewidth]{image/scalebars/scale_ct-500.png} 
}}\hfill%
\subfloat[][Optical flow, crop]{\label{fig:real-effect-flow}%
\scalebarme{%
\includegraphics[width=0.249\linewidth]{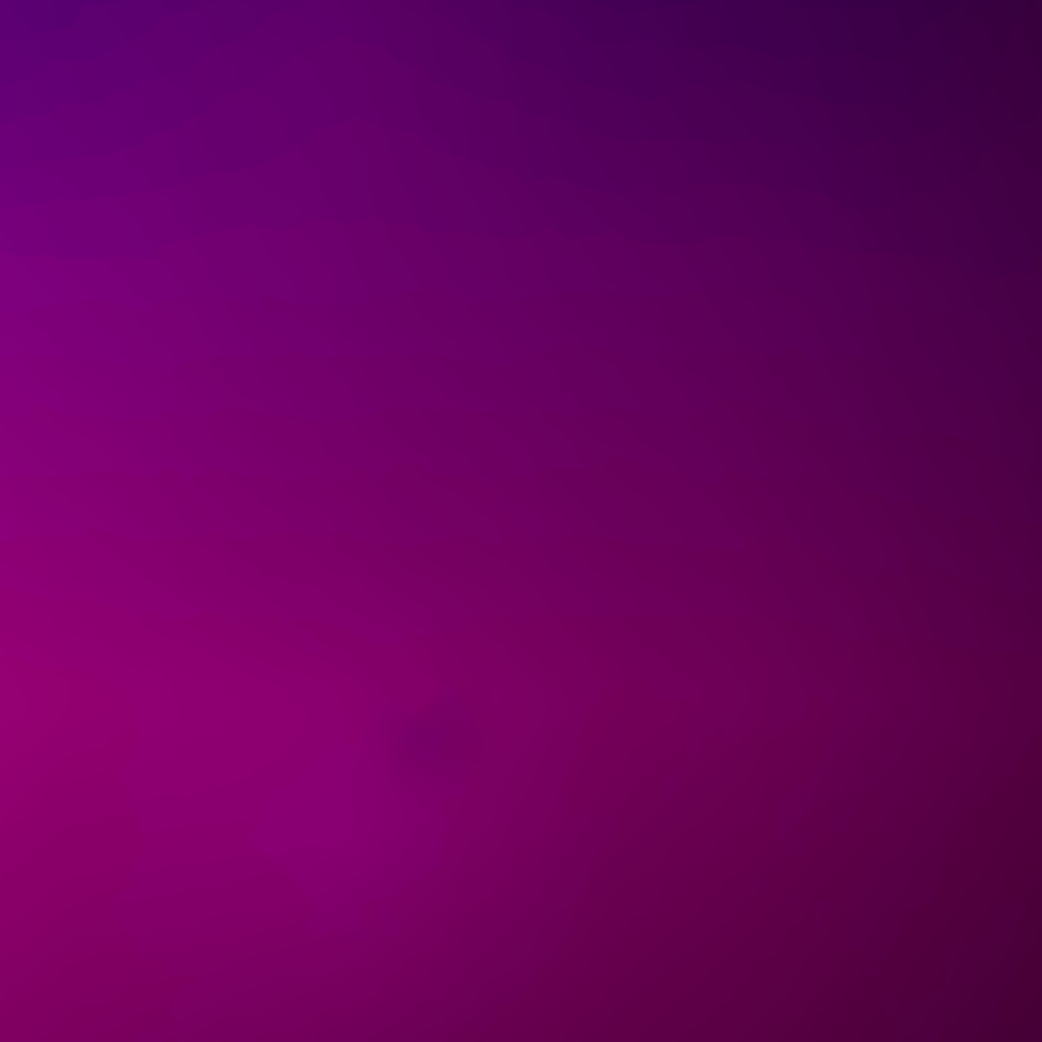}}{%
\includegraphics[width=0.249\linewidth]{image/scalebars/scale_ct-500.png} 
}}\hfill%
\subfloat[][Crop from~\protect\subref{fig:real-effect-d}]{\label{fig:real-effect-d2}%
\scalebarme{%
\includegraphics[width=0.249\linewidth]{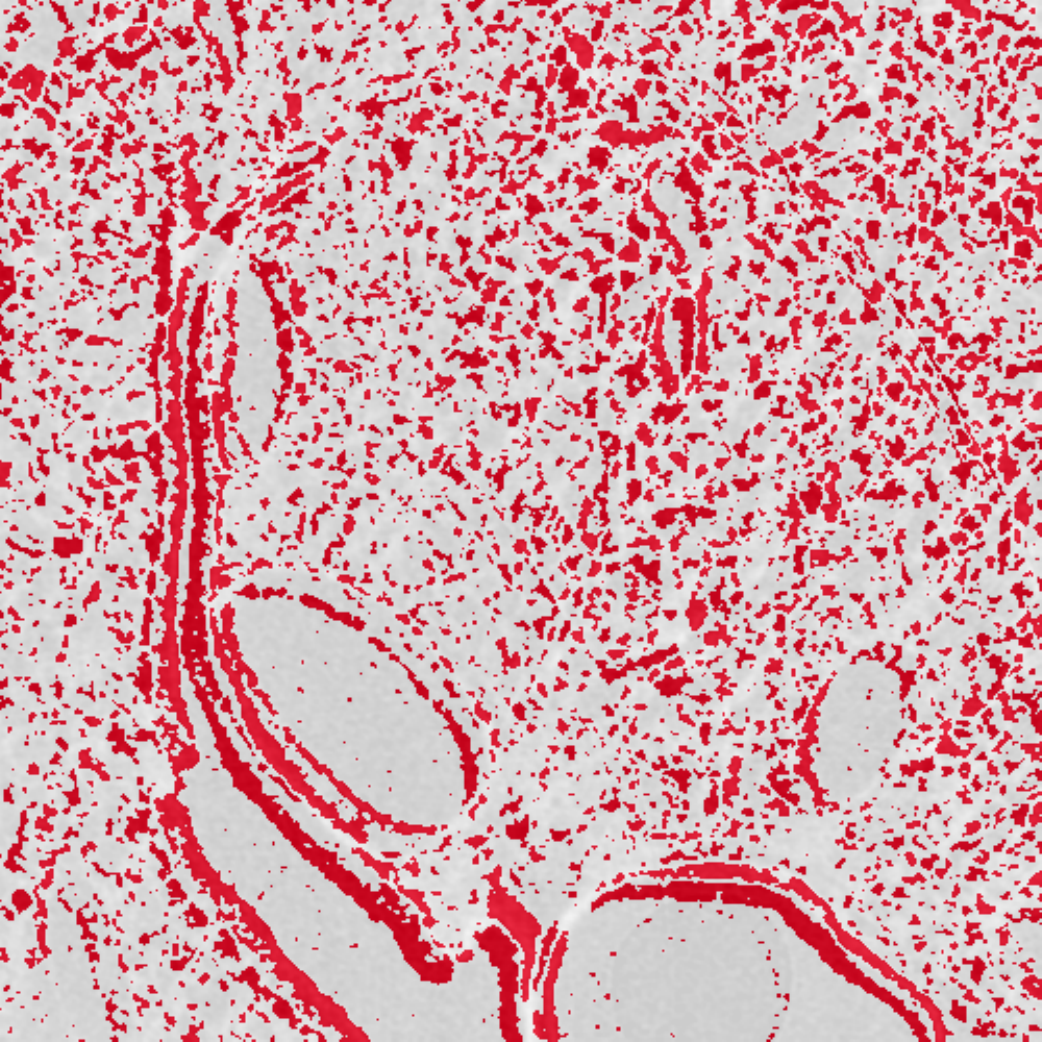}}{%
\includegraphics[width=0.249\linewidth]{image/scalebars/scale_ct-500.png} 
}}

\caption{The effect of the local distortions on real data: an
unfiltered micro-CT image of a rabbit lung. Fig.~\protect\subref{fig:real-effect-a}
features the extended (for later rigid transformation), but not
distorted image. Fig.~\protect\subref{fig:real-effect-b} is the result
of non-linear distortions. The distortions may be hardly noticeable by a human, but they are enough to confuse computational methods. Such distortions are clearly visible in the next panels. Figs.~\protect\subref{fig:real-effect-c}
and~\protect\subref{fig:real-effect-d} show the PSNR visualization
between \protect\subref{fig:real-effect-a} and
\protect\subref{fig:real-effect-b} with 5\% and 20\% threshold for
the red color.
Figs.~\protect\subref{fig:real-effect-a2}--\protect\subref{fig:real-effect-d2}
show crops, Fig.~\protect\subref{fig:real-effect-flow} shows an
optical flow visualization computed from full images, but then cropped
in the same manner as others.
The optical flow shows the \enquote{movements} between two input images.%
\newline%
Scale bars in \protect\subref{fig:real-effect-a}--\protect\subref{fig:real-effect-d} are \SI{5}{\milli\meter}, scale bars in \protect\subref{fig:real-effect-a2}--\protect\subref{fig:real-effect-d2} are \SI{1}{\milli\meter}.}
\label{fig:real-effect}
\end{figure}

\hypertarget{validation-image-generation-and-benchmarks}{%
\subsection{Validation, image generation, and
benchmarks}\label{validation-image-generation-and-benchmarks}}

In a sense, this paper is dual to \citet{cifor_smoothness-guided_2011}.
They thought explicitly about possible distortions during sectioning and
displaced the real section images in a way that would counter this
distortion---a similar idea is behind most registration methods. Our
distorted ground-truth data are the input of a registration benchmark.
We validate multiple existing registration methods by comparing
(previously distorted and) registered images to the distorted, but not
registered, and to the (not distorted, perfectly aligned) original data.
Sections~\ref{sec:eval} and~\ref{sec:full-method} detail on our
evaluation methodology.

\citet{pluim_truth_2016} provide an overview on validations of medical
registrations. \citet{cunliffe_lung_2012} are concerned with changes to
medical images effected by registration.

\citet{van_sint_jan_registration_2002} showcase a very special kind of a
registration that was validated with kinematics. The method by
\citet{delaby_3d_2010} is a more typical case, where the 3D
reconstructions were validated by a different modality.
\citet{schnabel_validation_2003} discuss physically plausible
distortions in breast MR data. \citet{shojaii_validation_2011} use
block-face images and fluidical markers for validation of the
registration. \citet{kybic_fast_2008} undertakes special efforts for the
evaluation of registration accuracy in absence of ground truth. In
contrast to all those approaches, we suggest to use multiple image-based
quality measures for the evaluation of the registrations. The essence of
the present manuscript is the availability of ground truth, so no
further modalities, manual interventions or external markers are
required.

Generation of further images is by far not new in medical image
processing, to name a few, \citet{xue_motion_2012} generate synthetic
images for better T1 MRI; \citet{duchateau_model-based_2018} generate
pathological cardiac images; and \citet{grova_methodology_2001} use
computer-generated SPECT data to validate their MRI-SPECT registration.
Image generation is connected to validation, because as long as we are
able to generate images with given properties, these can be used to
validate other image-based methods. \citet{hamarneh_simulation_2008} use
all kinds of statistical and physically-based distortions, noise,
artifacts but their method is focused on MRI and CT data. Even though
distortions of 2D images are also possible with their framework, the
method is focused on other modalities.

\citet{vlachopoulos_selecting_2015} generate distorted CT images using
landmarks and thin-plate splines. Their warping method was specially
chosen to imitate aspiration. The images were used to evaluate
registration methods in normal lungs and organs with interstitial lung
disease. The idea of the evaluation is similar to ours, however, we
focus on histological serial sections and provide an elaborate
methodology to generate such distorted images. Both the nature of the
deformation and the method of its implementation are different in this
work. We are concerned with sectioning and processing artifacts in
removed tissue. We do not use thin-plate splines and landmarks to
compute distortions.

\citet{zhang_all-pass_2020} both generate synthetic images and use real
data to compare their global registration method to others with
promising results. Unlike the present work here, they distort the images
using quadric 2D polynomials and focus on either homography or
distortions commonly found in digital cameras. These kinds of
distortions differ a lot from our approach, since they are induced by
the optical pathway in the camera and not by physical sectioning of the
specimen.

Related to above methods are registration benchmarks and challenges. We
would like to specially mention the EMPIRE10 challenge
\citep{murphy_evaluation_2011}. \citet{borovec_benchmarking_2018}
benchmark registrations of differently stained serial sections \citep[a
rather distinctive kind of
registration:][]{likar_hierarchical_2001, mueller_real-time_2011, song_unsupervised_2014, trahearn_hyper-stain_2017, lobachev_tempest_2020, wodzinski_deephistreg_2021}.
Basically, registration of differently stained serial sections is about
transferring the distortions from one kind of staining to another one.
Also co-registrations across modalities are a tangent topic to our work,
\eg CT to MRI
\citep{bardinet_co-registration_2002, tang_locally-adaptive_2012, gehrung_co-registration_2020},
serial sections to MRI
\citep{jacobs_registration_1999, du_bois_daische_efficient_2005, foster_mri_2017},
or MRI to ultrasound \citep{guo_deep_2020}. We \emph{distort} images
from other modalities (similarly to the distortions in serial sections)
in order to obtain challenges for testing registration methods with a
known ground truth. In this work, during each of the registration
attempts we remain within a single selected modality.

A~recent ANHIR challenge \citep{borovec_anhir_2020} uses multiple data
sets, where the ground truth is obtained by external markers on the
histological series. Their landmarks were placed by multiple human
annotators. We use automatic image-based metrics in this work and do not
use external markers, however our approach is open to further measures.
(It would be easiest to integrate further image-based markers, though.)
We eschew external markers, as we \emph{have} a ground truth, which
contrasts our work from histology-based challenges, where no direct
ground truth is available. Further, the ability to fully automatically
compute the \enquote{score} of a registration method from ground truths
and registration output allows our approach to be used in automatic
tests of registrations, such as continuous integration
(Section~\ref{sec:ci}).

The NIREP project \citep{christensen_introduction_2006} evaluates
specifically non-rigid registrations, with absent ground truth. This
paper is about distorting a known ground truth for the evaluation of
rigid and, mostly, non-rigid registrations, we circumvent the main
problem of non-available ground truth.

To name further related papers, \citet{pontre_open_2017} presents a
cardiac perfusion MRI registration challenge focused on motion
correction; \citet{brock_results_2010} compare accuracy of different
deformable registration methods on MRI and CT data;
\citet{west_comparison_1997} and \citet{hellier_retrospective_2003} are
examples of the evaluations of inter-subject registrations.
\citet{klein_evaluation_2009} and \citet{ou_comparative_2014} evaluate
registration methods for inter-patient brain MRI.

\hypertarget{quality-measures-for-registration}{%
\subsection{Quality measures for
registration}\label{quality-measures-for-registration}}

Image registration can be seen as an optimization problem. Similarity
measures are key to both good registrations and their evaluation, as a
similarity measure is basically the objective of optimization. Mutual
information is often used as such a measure
\citep{viola_alignment_1997, pluim_interpolation_2000, pluim_mutual-information-based_2003, deniz_multi-stained_2015, polfliet_intrasubject_2018}.
A~related problem is the selection of the reference in a series of
histological sections \citep{bagci_automatic_2010}.
\citet{nanayakkara_surface-based_2009} introduce a metric for
registration errors. \citet{luo_are_2020} discuss the relation between
registration errors and uncertainty.

In this work, we choose image-based measures as an arbiter in quality of
the registration. Such approach allows not only for automatic generation
of the inputs and for automatic execution of the registration, but also
for an automatic evaluation of the results. In our evaluation we use the
standard measures by \citet{jaccard_distribution_1912} and
\citet{wang_image_2004}, as well as the dense optical flow
\citep{farnebaeck_two-frame_2003}. We also use the
\citet{dice_measures_1945} measure as a visualization of the Jaccard
measure---as the formulation of Dice can be converted to a formulation
of Jaccard. (Details on thresholding methods are in the supplementary
material.) We use a PSNR-based visualization as well.

\citet{crum_generalized_2006} discuss generalizations of overlap-based
measures, but in this work we opted for the well-known measures.
\citet{rohlfing_image_2012} criticizes the usual image-based measures,
but our method can use any measures for evaluation. Our method is not
imbued with the measures we use, hence any extensions or further
measures are possible. The core idea is to use (now-distorted)
inherently 3D images to benchmark 2D registrations.

\hypertarget{machine-learning}{%
\subsection{Machine learning}\label{machine-learning}}

With modern deep learning methods, the measure can be implicitly
learned, as \citet{descoteaux_robust_2017} mention.
\citet{maier_gentle_2019} provide an introduction to deep learning in
medical imaging. \citet{cheplygina_not-so-supervised:_2018} survey
semi-supervised, multiple instance, and transfer learning techniques in
medical imaging. Machine learning functions best with a lot of ground
truth data from which the relations can be learned. In this context, our
work might provide a way of generating the so much needed input--output
data pairs. \citet{klaser_deep_2018} generate synthetic CT images from
MRI. \citet{li_random_2020} use style transfer to generate images from
different vendors, such generated images enable better machine learning.
For conventional registration the style is less relevant.

Machine learning techniques for registration include
\citep{yang_fast_2016, cardoso_ssemnet:_2017, dalca_unsupervised_2018, balakrishnan_voxelmorph:_2019, de_vos_deep_2019, zhao_unsupervised_2019, gu_consistent_2021, qian_cascade-network_2021, zhang_tissue-specific_2021}.
\citet{fu_deep_2020} and \citet{haskins_deep_2020} provide an overview.

We stress that our method generates distorted images without any use of
machine learning. Thus, our method can be used to generate additional
data sets or to augment machine learning input.

\hypertarget{lung-in-3d}{%
\subsection{Lung in 3D}\label{lung-in-3d}}

\olcomment{--- -> The part about lungs needs review from medical side:}

The methods, options, and research outcomes in 3D reconstructions of the
lung using any modalities from corrosion cast and up to 3D EM methods
\citep[\egX ][]{schneider_volume_2021} are reviewed by
\citet{muhlfeld_recent_2018}; practical applications include
\citep{mayhew_review_2009, grothausmann_digital_2016}. Although, EM
studies of the lung are popular and important
\citep[\egX ][]{ochs_closer_2010, ochs_using_2016, schneider_topological_2019, muhlfeld_plate_2021},
serial sections for LM have their place in the investigation repertoire
\citep[\egX ][]{woodward_study_2008, muhlfeld_visualization_2017, pentinga_3d_2018}.
Putting stereology aside, a proper registration is paramount in any
investigation of serial sections as a 3D data set.

\hypertarget{methods}{%
\section{Methods}\label{methods}}

The typical sectioning-induced distortion was found by
\citet{schormann_statistics_1995} to be Rayleigh distributed. This
basically means a normal distribution in each of the image axis. We
showcase our method on a synthetic image (Fig.~\ref{fig:teaser}), on a
standard test image (Fig.~\ref{fig:ship}), and on real data, a 2D image
from a micro-CT scan of an animal lung (Fig.~\ref{fig:real-effect}).
Notice that Fig.~\ref{fig:teaser} overemphasizes the effect of our
method: we use there 10~times larger distortions than usual.

Now, Figs.~\ref{fig:teaser-a}, \ref{fig:ship-a} show the original
specimen. In real applications we extend the border
(Figs.~\ref{fig:ship-b}, \ref{fig:real-effect-a}).
Figs.~\ref{fig:teaser-b}, \ref{fig:ship-c}, \ref{fig:real-effect-b}
demonstrate the distorted images, Figs.~\ref{fig:teaser-c},
\ref{fig:ship-d} show the color-coded distortion map. The detailed crops
in Figs.~\ref{fig:ship}, \subref{fig:ship-b2}--\subref{fig:ship-d2}, and
Figs.~\ref{fig:real-effect},
\subref{fig:real-effect-a2}--\subref{fig:real-effect-d2}, demonstrate
the local movements induced by the distortion. The goal of the
registration is to precisely eliminate such movements.

To simulate lost or damaged parts of the sections \citep[which we
recently learned how to repair:][]{lobachev_tempest_2020}, additional
arbitrarily placed \enquote{damage} can be added to the image
(Fig.~\ref{fig:ship-e}). Finally, a global rigid transformation is
applied (Fig.~\ref{fig:teaser-d}). The rationale behind this step is
that only in very rare cases the sections can be placed on the glass
slide while maintaining the exact orientation. We randomly apply a
rotation and translation to the images to simulate the uneven
positioning. Such images serve then as inputs for the benchmark of
registrations.

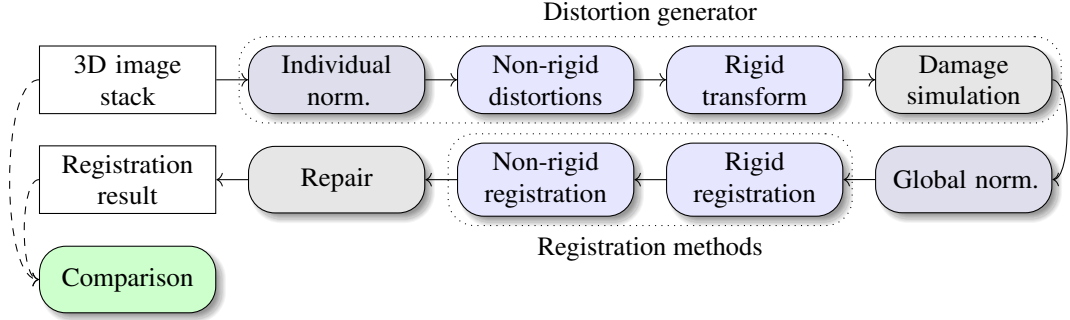
\begin{figure}[tb]
\centering
\begin{tikzpicture}[align=center,node distance = 1.2em and 1.2em,  
    rounded corners=10, blur shadow={shadow blur steps=50}]
  \tikzstyle{block} = [rectangle, fill=blue!10, draw, text width=6em,
  text centered, minimum height=2.1\baselineskip,
  blur shadow={shadow blur steps=50}
  ]
  \tikzstyle{edge_block} = [rectangle, draw, text width=6em,
  text centered, minimum height=2.1\baselineskip,
  rounded corners=0,
  ]
  \tikzstyle{less_block} = [rectangle, fill=blue!30!gray!20, draw, text width=6em,
  text centered, minimum height=2.1\baselineskip,
  blur shadow={shadow blur steps=50}
  ]
  \tikzstyle{special_block} = [rectangle, fill=green!20, draw, text width=6em,
  text centered, minimum height=2.1\baselineskip,
  blur shadow={shadow blur steps=50}
  ]

  \tikzstyle{meh_block} = [rectangle, fill=gray!20, draw, text width=6em,
  text centered, minimum height=2.1\baselineskip,
  blur shadow={shadow blur steps=50}
  ]
  \tikzstyle{disabled_block} = [rectangle, fill=gray!10, draw, text
  width=7em, text centered, minimum height=1.5\baselineskip,
  blur shadow={shadow blur steps=50}
  ]
  \tikzstyle{blob} = [circle, fill=blue!10, draw,
  text centered, minimum height=1ex]
  \tikzstyle{ghost} = []
  \tikzstyle{quest} = [diamond, fill=red!10, draw, text
  width=3em, text centered, minimum height=0.8\baselineskip]
  \tikzstyle{filler_block} = [rectangle, text width=6em,
  text centered, minimum height=2.1\baselineskip,
  ]

  \node (norm1) [less_block] {Individual norm.};
  \node (rep) [meh_block,below=of norm1] {Repair};
  \node (nr1) [block,right=of norm1] {Non-rigid distortions};
  \node (rig1) [block,right=of nr1] {Rigid transform};
  \node (dmg) [meh_block,right=of rig1] {Damage simulation};

\node (nr2) [block,right=of rep] {Non-rigid registration};
\node (rig2) [block,right=of nr2] {Rigid registration};
  \node (norm2) [less_block,below=of dmg] {Global norm.};

  \node (inp-pos) [edge_block,left=of norm1]  {3D image stack};
   \node (res-pos) [edge_block,below=of inp-pos,left=of rep] {Registration result};

   \node (cmp) [special_block,below=of inp-pos, below=of res-pos] {Comparison};

\draw [->] (inp-pos) -- (norm1);
\draw [->] (norm1) -- (nr1);
\draw [->] (nr1) -- (rig1);
\draw [->] (rig1) -- (dmg);
\begin{pgfinterruptboundingbox}
\draw [->,looseness=0.5,in=0,out=0] (dmg.east) to (norm2.east);
\end{pgfinterruptboundingbox}
\draw [->] (norm2) -- (rig2);
\draw [->] (rig2) -- (nr2);
\draw [->] (nr2) -- (rep);
\draw [->] (rep) -- (res-pos);

\begin{pgfinterruptboundingbox}
\draw [dashed,->,looseness=0.5,in=180,out=180]  (inp-pos.west) to (cmp.west);
\draw [dashed,->,looseness=0.5,in=180,out=180]  (res-pos.west) to (cmp.west);
\end{pgfinterruptboundingbox}


\node (fit-dist) [fit=(norm1) (nr1) (rig1) (dmg), shape=rectangle, draw, dotted]{};
\node (fit-regs) [fit=(nr2) (rig2), shape=rectangle, draw, dotted]{};
\node [above=0.05em of fit-dist] {Distortion generator};
\node [below=0.05em of fit-regs] {Registration methods};
\end{tikzpicture}

\caption{Illustrating the steps in this paper. Individual normalization of benchmark inputs and the application of global normalization are not mandatory. Currently, we do not use damage generation and damage repair in our benchmark images, but we could also test the repair methods in this manner. The evaluation ensues from comparison of the image series. The dotted boxes show two larger conceptual components, the distortion generator and the registration.}
\label{fig:dia}
\end{figure}

Fig.~\ref{fig:dia} visualizes the core approach and the complete
pipeline of this paper. We present a distortion generator that is in a
sense dual to a registration. We derive an evaluation of a registration
method from original 3D stack and registration results.

\hypertarget{local-distortions}{%
\subsection{Local distortions}\label{local-distortions}}

Generation of local, non-rigid distortions is of high importance for our
method. At the heart of the local distortion lies the generation of
normally distributed values. Two independent normally distributed random
values form the~\(x\) and \(y\) coordinates of a displacement, making
the displacements Rayleigh-distributed
\citep{schormann_statistics_1995}. The coordinates of locations,
\emph{where} the distortions are placed, lie on a rectilinear grid.

We generate multiple distortion \enquote{levels} using a classical
multiscale approach. The distortions are stored as coordinates of
\enquote{new} points in a matrix holding both~\(x\) and \(y\)
coordinates as an element---this is a typical \texttt{remap} matrix of
OpenCV. The distortions are blurred with a Gaussian kernel in each
multiscale level to make the displacements smoother. This way we avoid
undesirable and unrealistic foldings, as real sections folds look
differently.

In the implementation, we used Mersenne Twister pseudorandom generator
\citep[a standard one in Python]{matsumoto_mersenne_1998}. The
distortion maps are applied to the input images with OpenCV
\texttt{remap} function using bicubic interpolation. The distortion maps
are saved for further analysis. We can generate those maps in a fully
deterministic manner, if desired. This determinism contributes to
reproducibility.

In our application, the final distortion map is visualized
(Fig.~\ref{fig:teaser-c}, \ref{fig:ship-d}) using HSV colorspace. The
Cartesian coordinates of the displacements are mapped to a polar angle
(associated with hue); vector magnitude basically codes the intensity.

Our distortion maps are a simple, reproducible, and well-defined way to
add sectioning-inspired distortions to arbitrarily registered data. We
aimed to define a stable and reproducible way to model such distortions
using the statistical properties of the real-world distortions.

The reproducibility is given through multiple efforts. We have the
initial, \enquote{ideal} state, the ground truth. We save the exact
rigid transform, the non-rigid distortion field, and the distorted
result. Through the use of pre-defined, deterministic states of the
pseudorandom generator, we can basically save all the transformations in
form of the seed value (plus original images, of course).
\dhcomment{Are 'transforms' and 'transformations' the same? If so, they could be homogenized throughout the manuscript}\olcomment{errr... yes? But I mulled over some instances of thosese names and I like them the way they are.}
Above issues would be useful to ensure reproducibility, \eg for
automatic regression testing of registration methods.

\hypertarget{adding-rigid-transform}{%
\subsection{Adding rigid transform}\label{adding-rigid-transform}}

\label{sec:apply-rigid}

The locally distorted images are further processed. In our application
we add an image-wide rigid transform: a random rotation and a
translation (Fig.~\ref{fig:teaser-d}). The rotation angle is uniformly
distributed between \(-180\) and \(+180\) degrees. The translations are
also uniformly distributed, but they are chosen in a range
\([-d/4, d/4]\), where \(d\) is maximal image dimension, in order to not
truncate too much of the image content. The rigid transform is recorded,
as it is the ground truth for the first, rigid step of the registration.

\hypertarget{the-use-of-the-distorted-images}{%
\subsection{The use of the distorted
images}\label{the-use-of-the-distorted-images}}

The distorted images serve as a starting point for the evaluation of
registration methods. The registration should, basically, \enquote{undo}
the distortions and transforms we applied to the original images. For
the benchmark and evaluation purposes we suggest using innately
registered data, i.e., original 3D images. The benefit of doing so is
the available ground truth, the original undistorted 3D stack.
Summarizing, we circumvent the problem of missing the real ground truth
when comparing registrations of serial sections.

\hypertarget{damaging-the-images-optional}{%
\subsection{Damaging the images
(optional)}\label{damaging-the-images-optional}}

An optional extension is to mimic the damage to which the sections are
subjected to during the processing (Fig.~\ref{fig:ship-e}). Using
normally-distributed pseudorandom values for dimensions and placement we
iteratively generate an ImageMagick \citep{imagemagick} script that
deletes selected parts of an image. This is a quite crude approximation
to the variety of possible kinds of damage to a section
\citep{lobachev_tempest_2020}---from an unsharp region (due to focus
error) to a teared section. However, we would like to model missing
parts of a section in an understandable and straightforward way. Our
test subject, a registration method, would not necessarily care about
the reason why some image regions are not matchable to other images. It
is not our current goal to model the damaged sections realistically.

\hypertarget{registration-methods}{%
\subsection{Registration methods}\label{registration-methods}}

In order to demonstrate our methodology of the evaluation of
registration, we apply the following registration methods to our
distorted data sets:

\begin{enumerate}
\def\labelenumi{\arabic{enumi}.}
\tightlist
\item
  \enquote{Rigid-SURF}: Feature-based rigid-only registration based on
  weighted RANSAC
  \citep{fischler_random_1981, lobachev_feature-based_2017} and SURF
  feature detector \citep{bay_surf:_2006};
\item
  \enquote{Rigid-SIFT}: same as above, but with SIFT feature detector
  \citep{lowe_object_1999, lowe_distinctive_2004};
\item
  \enquote{Deform-SURF}: Feature-based deformable registration
  \citep{lobachev_feature-based_2017}, first stage based on
  \enquote{Rigid-SURF}, followed by multiple non-rigid stages using
  B-splines;
\item
  \enquote{Elastix}: a generic \elastix{}
  \citep{klein_elastix:_2010, shamonin_fast_2014} configuration, we used
  rigidly registered result from \enquote{Rigid-SURF} as input. The
  parameter file is made available in the supplementary material. The
  non-rigid stage is \emph{not} a feature-based method;
\item
  \enquote{GS}: Registration method based on Gauss-Seidel optimization
  \citep{gaffling_gauss-seidel_2015}, we used rigidly registered result
  from \enquote{Rigid-SURF} as input. However, the actual non-rigid
  deformation is based on different principles, among others, on the
  gray-level co-occurrence matrices;
\item
  \enquote{Blending}: Registration method based on blending rigid
  transforms in image regions \citep{kajihara_non-rigid_2019};
\end{enumerate}

\olcomment{the comment wanted by SG:}

The rigid methods work pair-wise on the images. We operated
\elastix pair-wise on the images, hence the possible accumulation of
\enquote{drift} with the progress of the series. \enquote{Deform-SURF}
optimizes the whole stack at once in the non-rigid phase, \enquote{GS}
does the same. The \enquote{Blending} method works backward and forward
from a reference frame. In this case, the inputs were used \enquote{back
and forth}, for a series of \(1,\dots, n\) images, the input was
\(n,\dots,1,1,\dots, n\), essentially doubling the length of the series.
The border interpolation was less of our concern, the images were padded
before processing.

Why the rigid transformations? The rigid-only methods we used clearly
cannot undo the non-rigid distortions. But the non-rigid distortions
also make harder the search for correspondences for the rigid
transformations between image features. Further, a rigid-only
registration is also not necessarily perfect in what it does, in other
words, a rigidly transformed (Section~\ref{sec:apply-rigid}) and then
rigid-only registered series is different from the series before such
transformations.

\hypertarget{result-evaluation-with-quality-measures}{%
\subsection{Result evaluation with quality
measures}\label{result-evaluation-with-quality-measures}}

\label{sec:eval}

We propose the use of established image quality measures: structural
similarity \citep[SSIM,][]{wang_image_2004}, Jaccard measure
\citep{jaccard_distribution_1912}, often visualized here with slight
implementation differences as a Dice measure \citep{dice_measures_1945},
a visualization of dense optical flow \citep{farnebaeck_two-frame_2003}.
In the latter, color stands for a direction and intensity for the
magnitude of the movement. We also use PSNR visualization, as
implemented in ImageMagick. There, red highlights a non-correspondence.
For Jaccard measure we use global thresholding. For Dice measure
visualization we use Otsu's method \citep{otsu_threshold_1979} on
blurred images, as implemented in OpenCV
\citep{bradski_opencv_2000, kaehler_learning_2014}. Further details are
in the supplementary material. The latter material also details on the
manner in which we crop the images for evaluation in order to eliminate
border effects. In our visualizations, black is \enquote{neither},
magenta and green means \enquote{in one image, but not in another},
white is in both.

\begin{figure}[!t]
\centering
\sisetup{round-mode = figures, round-precision = 3}
\subfloat[][Original crop~1]{\label{fig:eval-ex-a}%
\scalebarme{%
\includegraphics[width=0.165\linewidth]{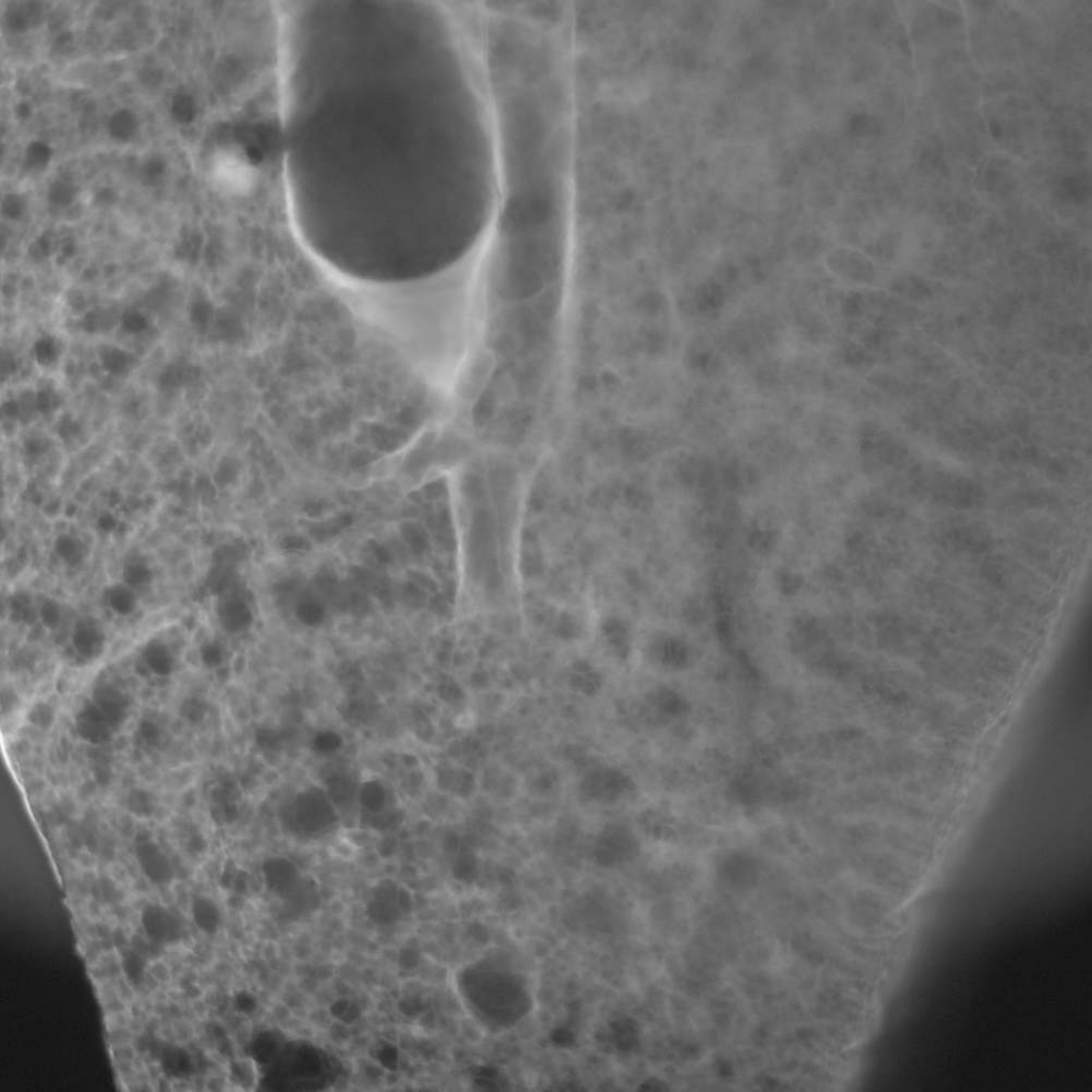}}{%
\includegraphics[width=0.165\linewidth]{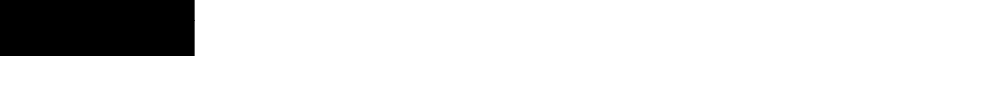}}}\hfill
\subfloat[][Distorted,\\$\textup{SSIM} = \num{0.71363}$]{\label{fig:eval-ex-b}%
\scalebarme{\includegraphics[width=0.165\linewidth]{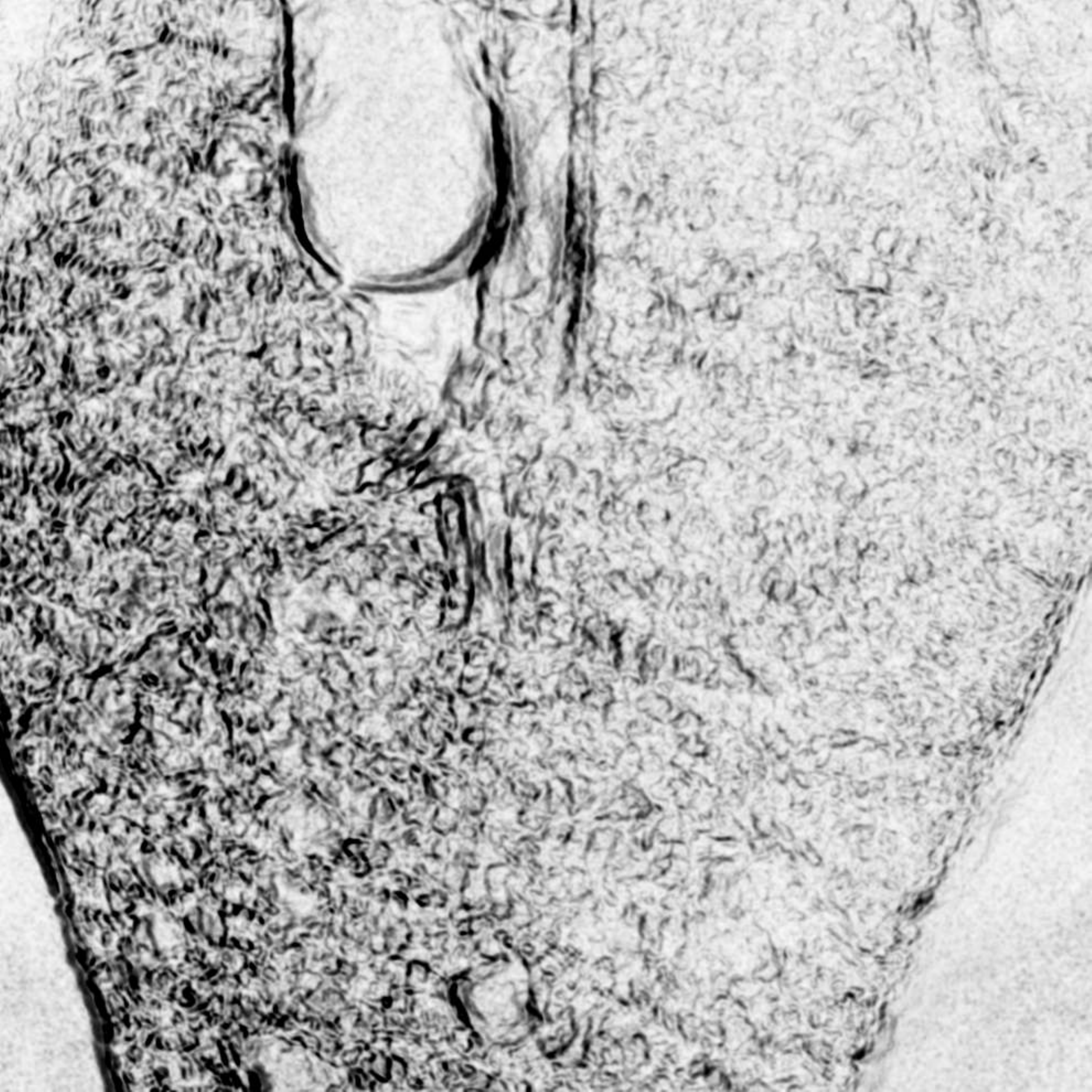}}{%
\includegraphics[width=0.165\linewidth]{image/scalebars/scale_ls-5a.png}}}\hfill
\subfloat[][Rigid-SURF,\\$\textup{SSIM} = \num{0.80171}$]{\label{fig:eval-ex-c}%
\scalebarme{\includegraphics[width=0.165\linewidth]{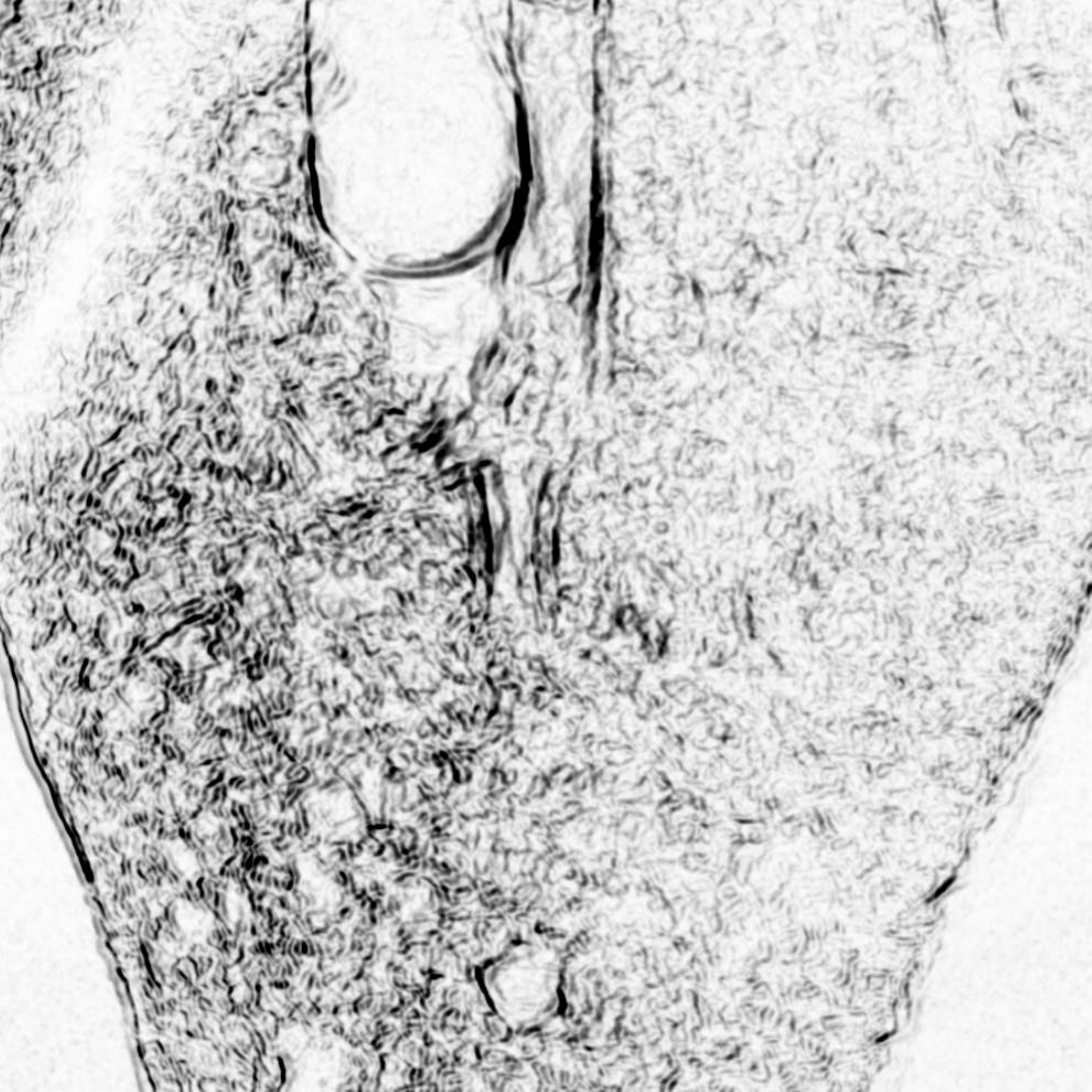}}{%
\includegraphics[width=0.165\linewidth]{image/scalebars/scale_ls-5a.png}}}\hfill
\subfloat[][Deform-SURF,\\$\textup{SSIM} = \num{0.96686}$]{\label{fig:eval-ex-d}%
\scalebarme{\includegraphics[width=0.165\linewidth]{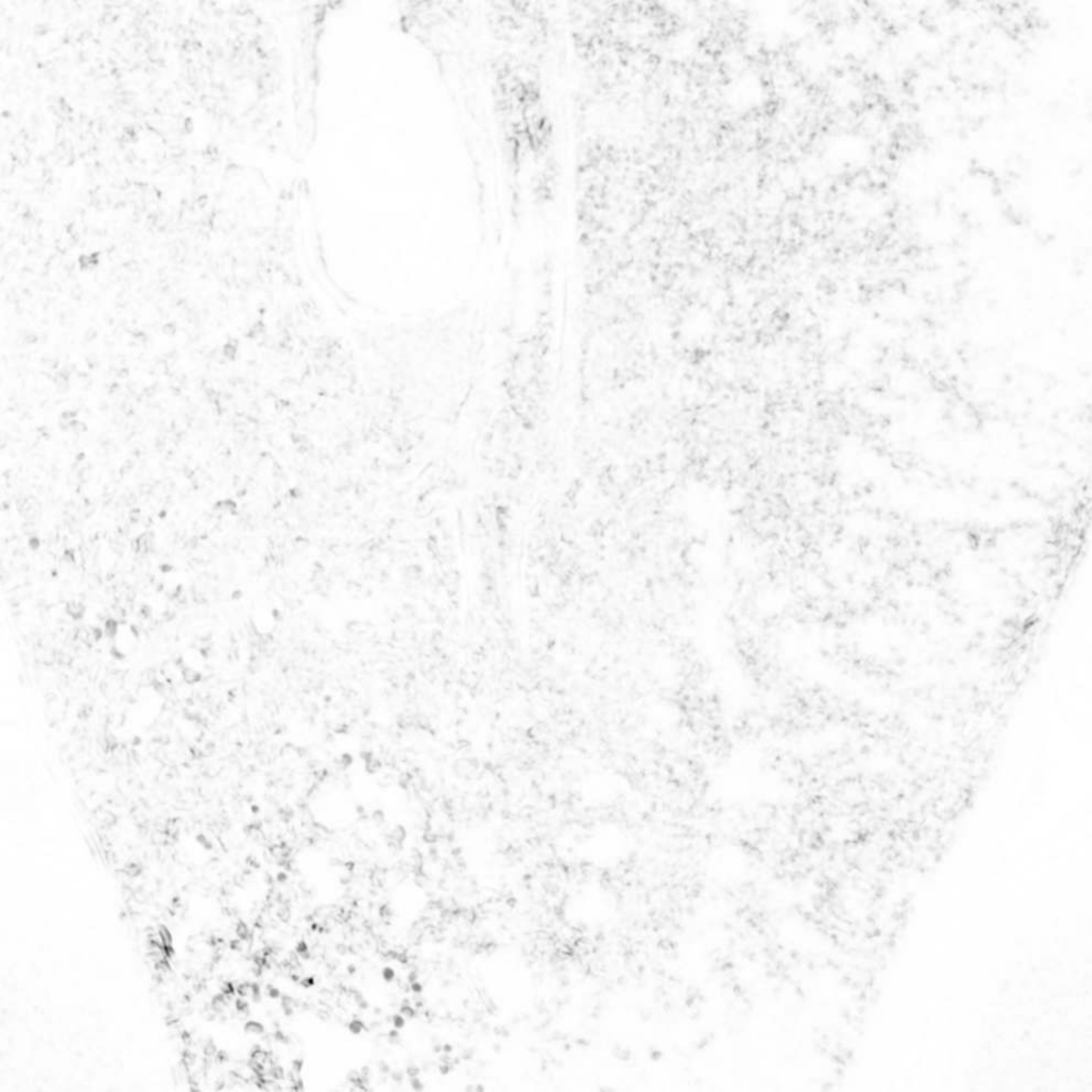}}{%
\includegraphics[width=0.165\linewidth]{image/scalebars/scale_ls-5a.png}}}\hfill
\subfloat[][GS, $\textup{SSIM} = \num{0.97065}$]{\label{fig:eval-ex-e}%
\scalebarme{\includegraphics[width=0.165\linewidth]{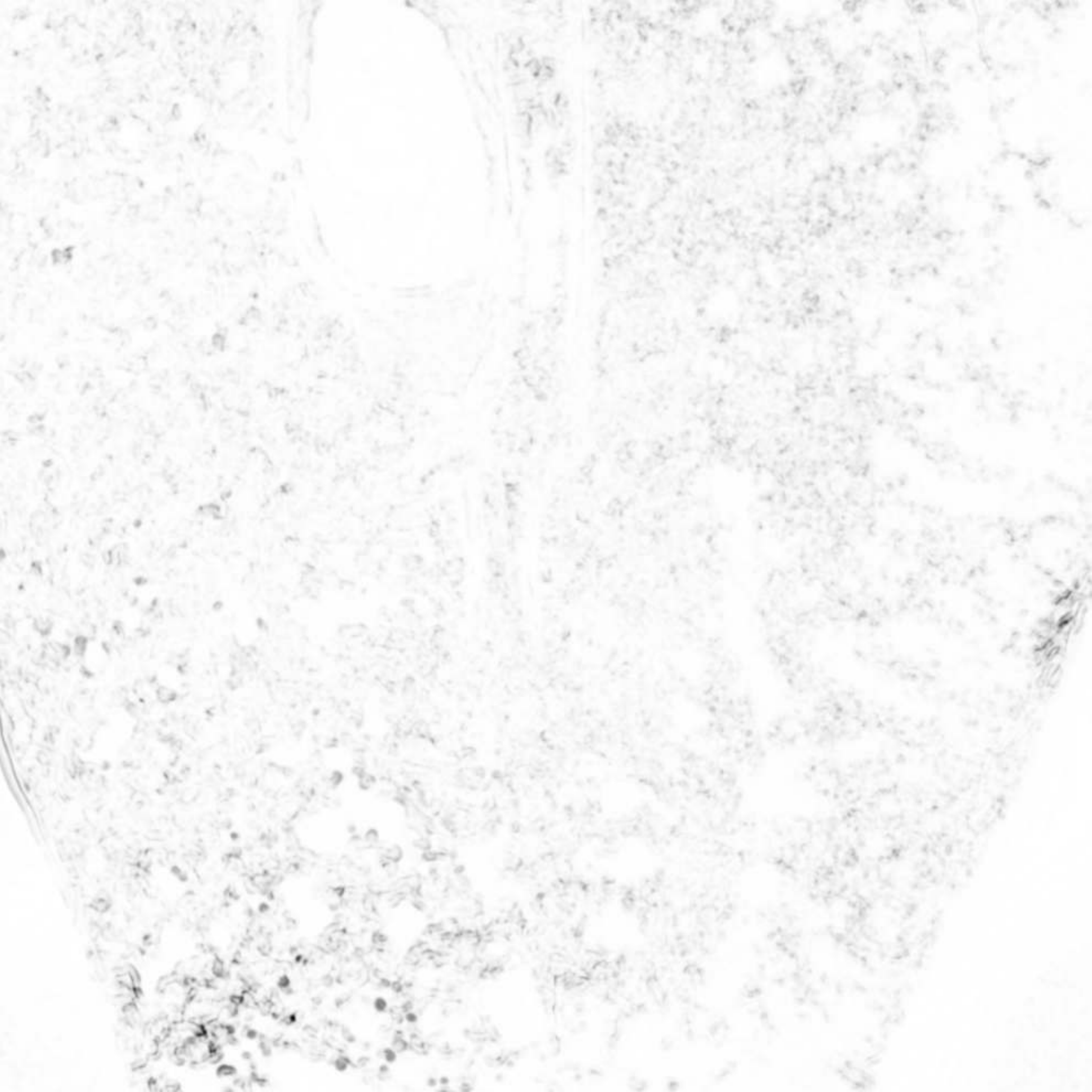}}{%
\includegraphics[width=0.165\linewidth]{image/scalebars/scale_ls-5a.png}}}\hfill
\subfloat[][Ground truth,\\$\textup{SSIM} = \num{0.93832}$]{\label{fig:eval-ex-f}%
\scalebarme{\includegraphics[width=0.165\linewidth]{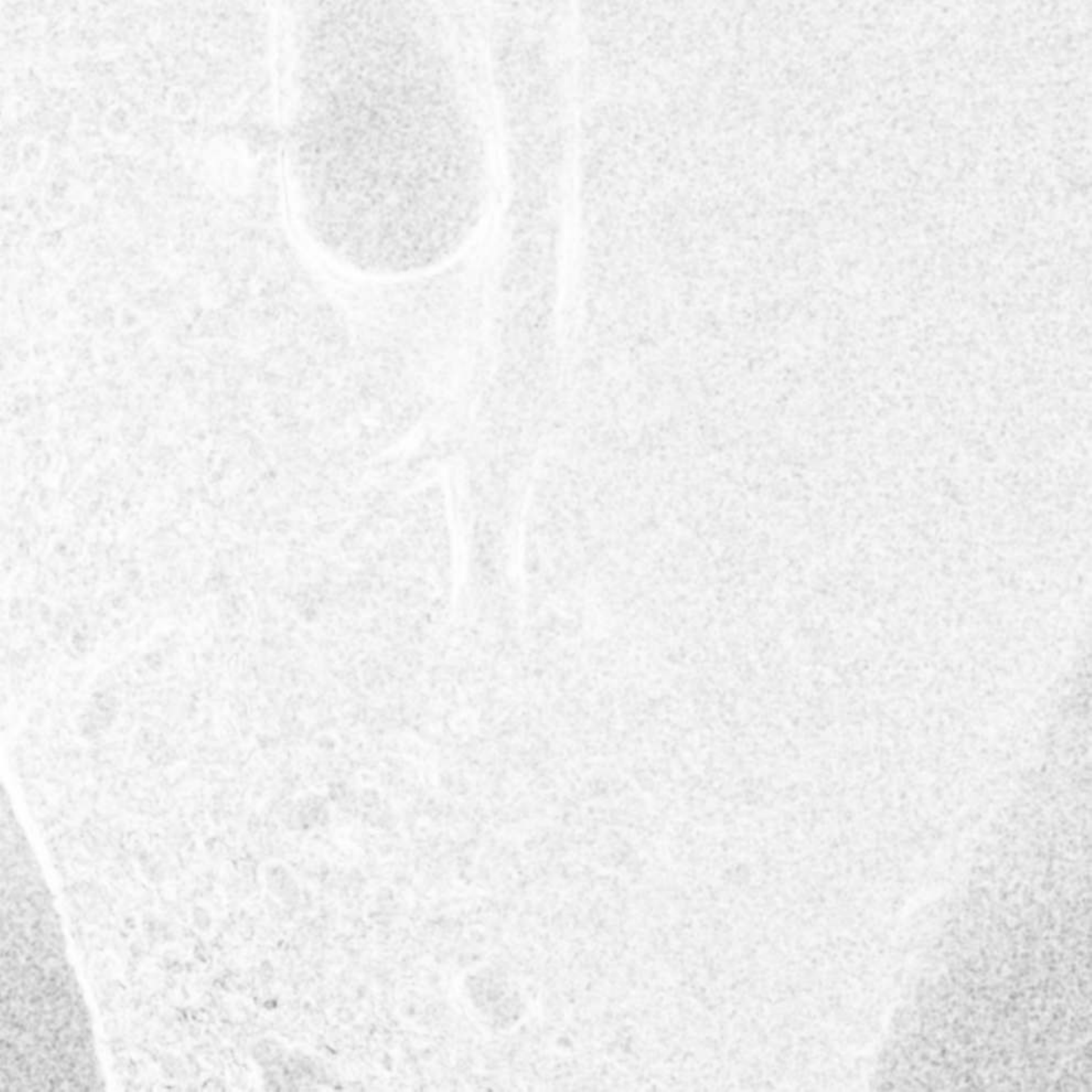}}{%
\includegraphics[width=0.165\linewidth]{image/scalebars/scale_ls-5a.png}}}

\subfloat[][Original crop~2]{\label{fig:eval-ex-a2}%
\scalebarme{\includegraphics[width=0.165\linewidth]{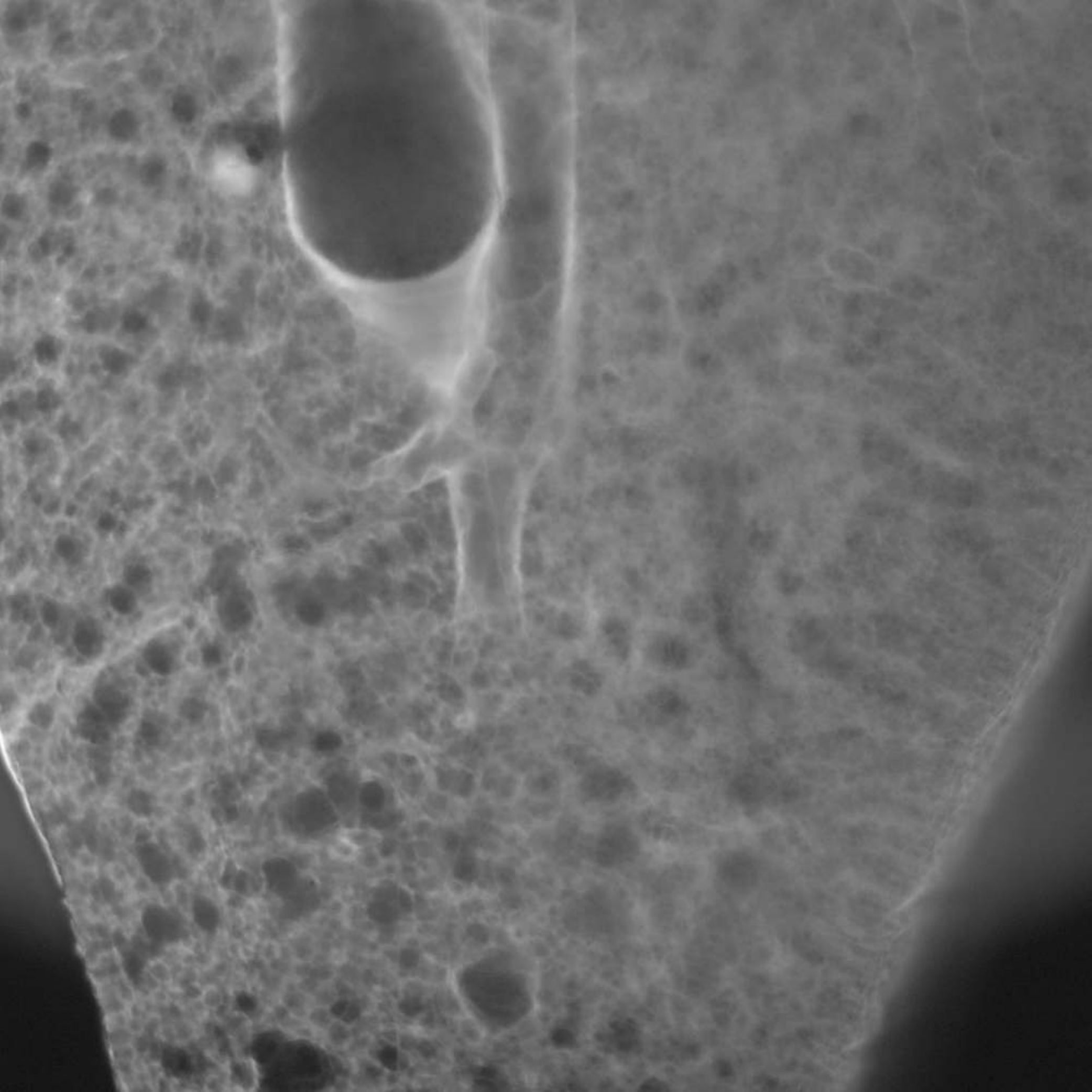}}{%
\includegraphics[width=0.165\linewidth]{image/scalebars/scale_ls-5a.png}}}\hfill
\subfloat[][Distorted, flow]{\label{fig:eval-ex-b2}%
\scalebarme{\includegraphics[width=0.165\linewidth]{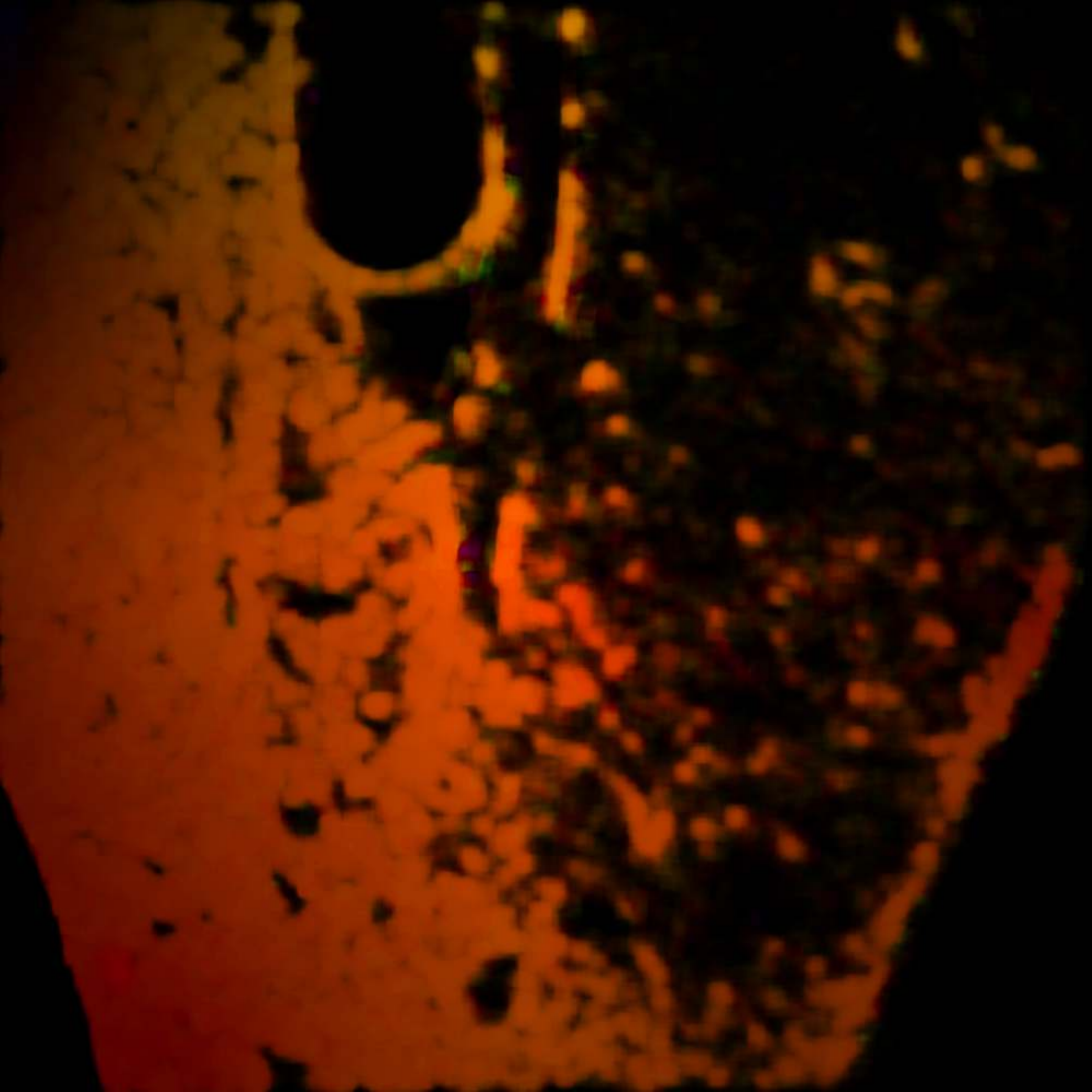}}{%
\includegraphics[width=0.165\linewidth]{image/scalebars/scale_ls-5a.png}}}\hfill
\subfloat[][Rigid-SIFT, flow]{\label{fig:eval-ex-c2}%
\scalebarme{\includegraphics[width=0.165\linewidth]{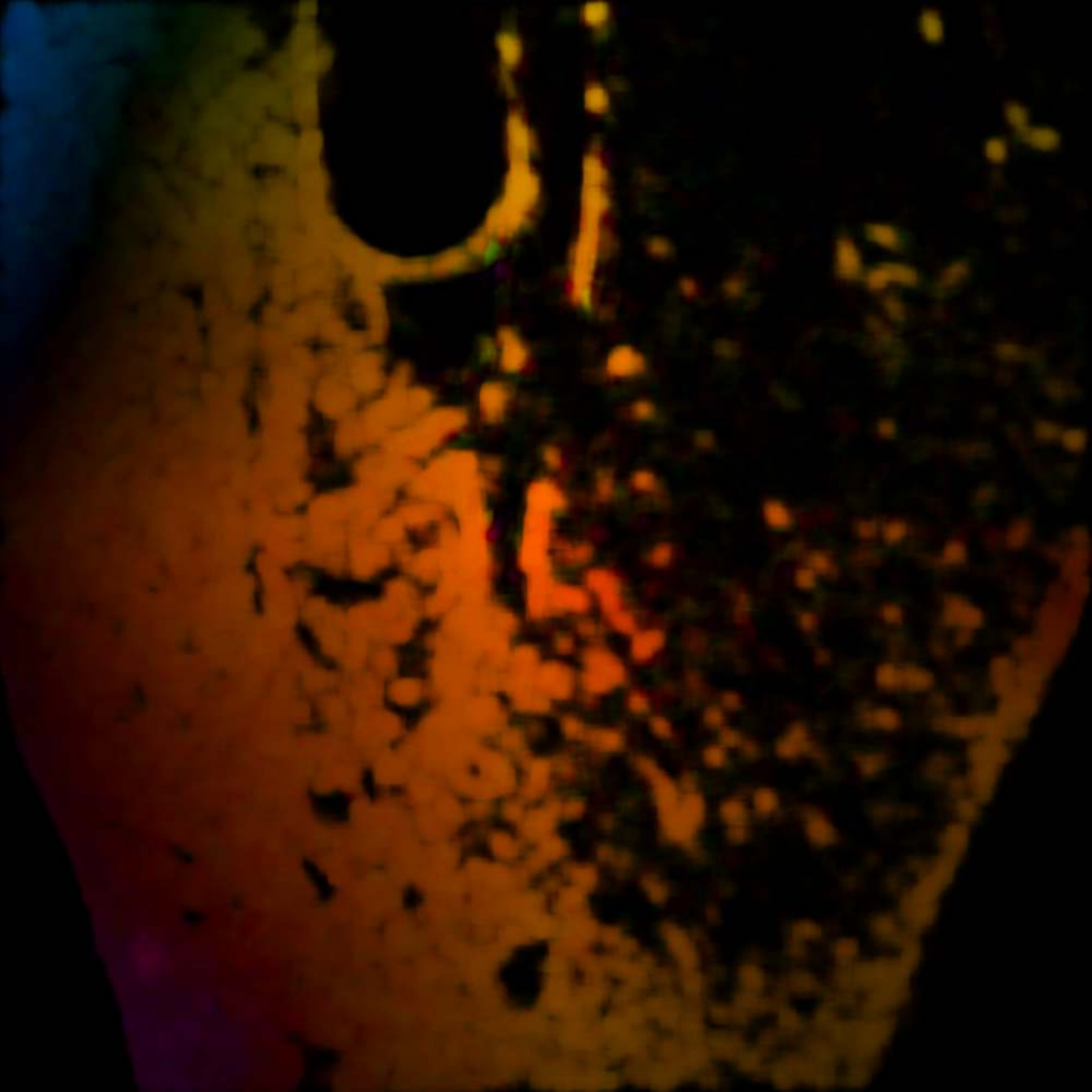}}{%
\includegraphics[width=0.165\linewidth]{image/scalebars/scale_ls-5a.png}}}\hfill
\subfloat[][Deform-SURF, flow]{\label{fig:eval-ex-d2}%
\scalebarme{\includegraphics[width=0.165\linewidth]{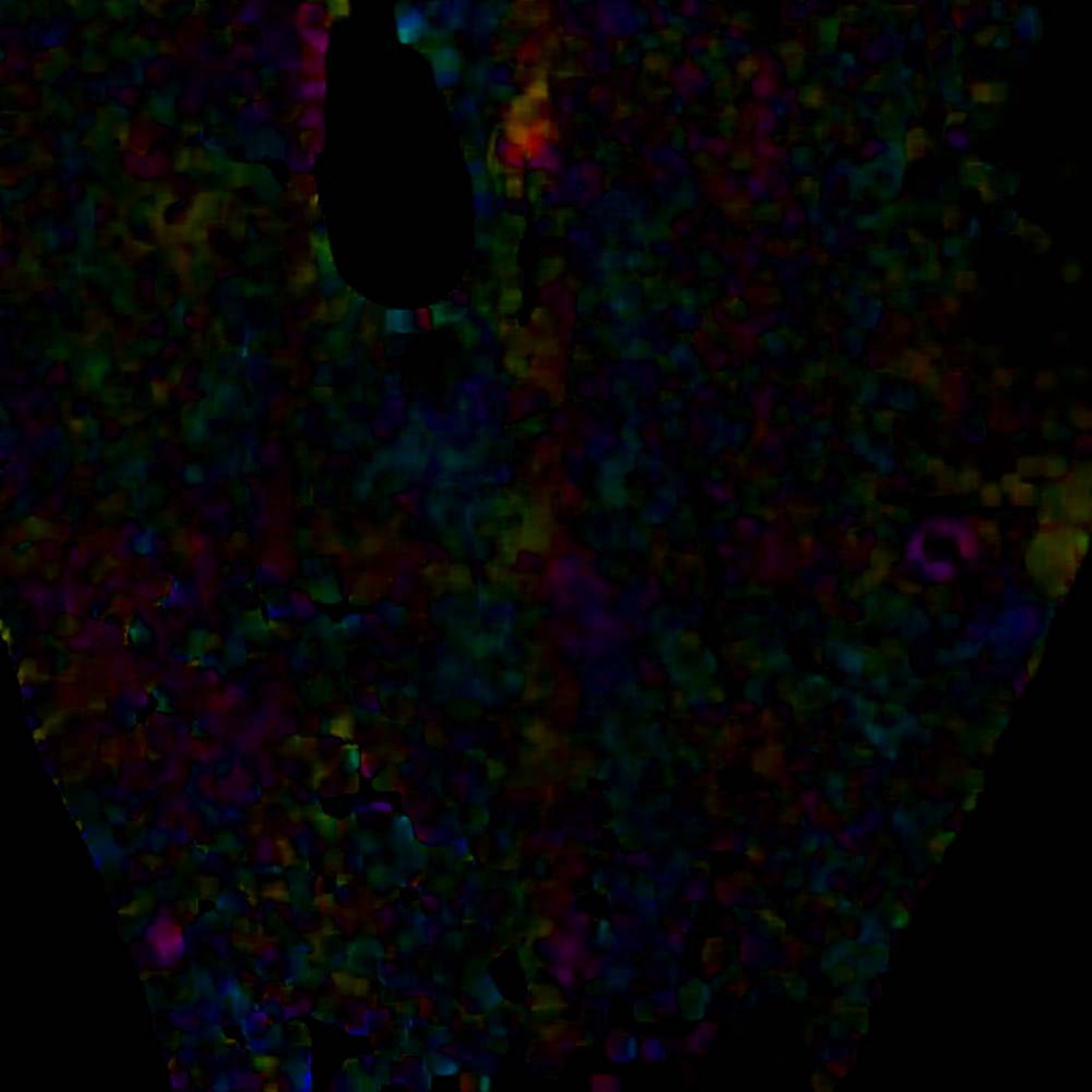}}{%
\includegraphics[width=0.165\linewidth]{image/scalebars/scale_ls-5a.png}}}\hfill
\subfloat[][GS, flow]{\label{fig:eval-ex-e2}%
\scalebarme{\includegraphics[width=0.165\linewidth]{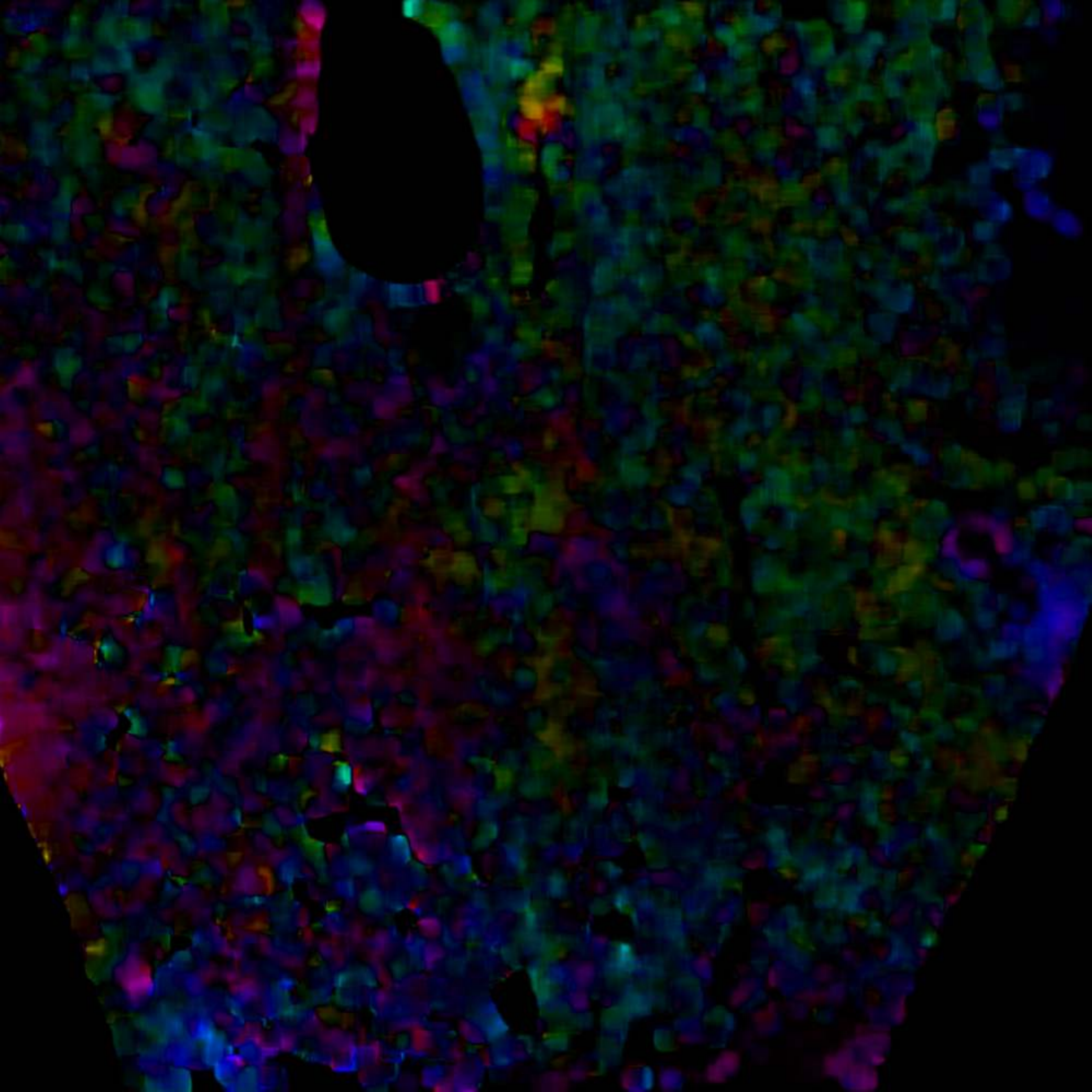}}{%
\includegraphics[width=0.165\linewidth]{image/scalebars/scale_ls-5a.png}}}\hfill
\subfloat[][Ground truth, flow]{\label{fig:eval-ex-f2}%
\scalebarme{\includegraphics[width=0.165\linewidth]{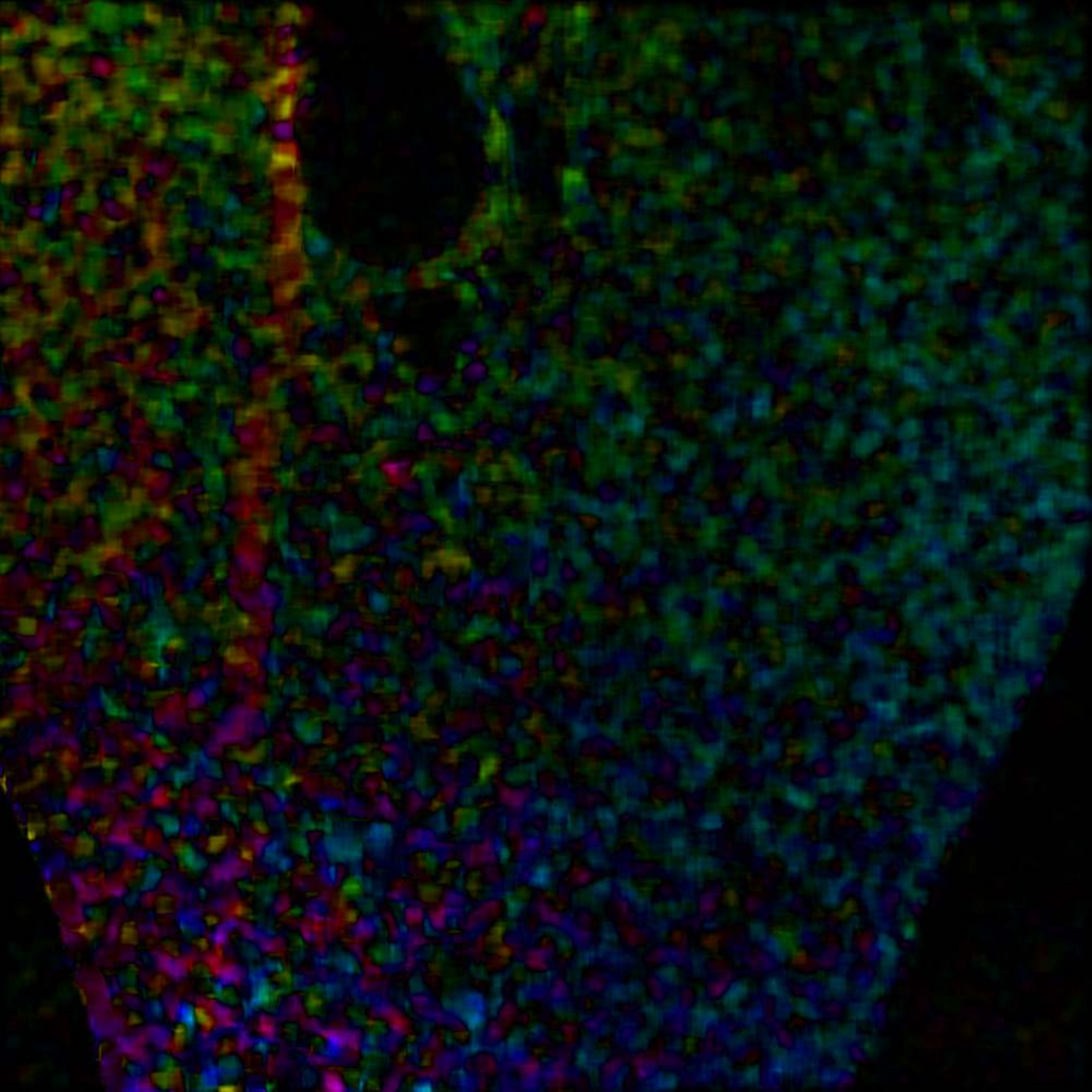}}{%
\includegraphics[width=0.165\linewidth]{image/scalebars/scale_ls-5a.png}}}

\caption{Evaluating image measures on
\kroi{1}{1} crop from full registration of LS. The \enquote{flow} images are the visualization of optical flow. The color codes the direction of the movement. Images~\protect\subref{fig:eval-ex-f}, \protect\subref{fig:eval-ex-f2} show the ground truth values: here two consecutive images of the ground truth were used to produce the quality measures.\newline%
All scale bars are \SI{1}{\milli\meter}.}
\label{fig:eval-ex}
\end{figure}

To give an example, Figure~\ref{fig:eval-ex} shows some of the quality
measures, applied to different registration of the same data. Top row
\subref{fig:eval-ex-b}--\subref{fig:eval-ex-f} shows the SSIM, bottom
row \subref{fig:eval-ex-b2}--\subref{fig:eval-ex-f2} shows the optical
flow visualization. We registered a full LS series with all methods. We
used the distorted images that were also globally transformed as the
input. Then, a \kroi{1}{1} crop from the full registration was used for
evaluation to eschew the border effects.
\dhcomment{I changed several 'Fig' instances to 'Panel'. Maybe one could replace the 'subref'-command with 'ref' instead of writing 'Panels' over and over.}
Panels \subref{fig:eval-ex-a} and \subref{fig:eval-ex-a2} from Fig.
\ref{fig:eval-ex} show regions cropped from the original images. Panels
\ref{fig:eval-ex}\subref{fig:eval-ex-b} and \subref{fig:eval-ex-b2} show
locally distorted images, before the global rigid transform. For all
registrations, the images were locally distorted and globally
transformed. Images shown in panels \subref{fig:eval-ex-c} and
\subref{fig:eval-ex-c2} were then registered with \enquote{Rigid-SIFT},
it is the rigid transformation only. Panels \subref{fig:eval-ex-d} and
\subref{fig:eval-ex-d2} show the evaluation of images, registered with
\enquote{Deform-SURF}, both rigidly and non-rigidly. Images shown in
panels \subref{fig:eval-ex-e} and \subref{fig:eval-ex-e2} were rigidly
registered with \enquote{Rigid-SURF}, then non-rigidly registered with
\enquote{GS}. Panels~\subref{fig:eval-ex-f} and \subref{fig:eval-ex-f2}
show the evaluation of original, non-distorted data, \ie the ground
truth.

\hypertarget{results}{%
\section{Results}\label{results}}

\olcomment{I have a feeling, evaluation methodology is scattered through the paper. But evaluation is not the main focus, even if it takes up so much space!}

First, we establish our specimens, discuss the data processing
(Section~\ref{specimens}), and data preparation
(Section~\ref{sec:padding}). Our main result is a full-series evaluation
with statistical means (Section~\ref{full-series-evaluation}). We also
include some special cases (Section~\ref{special-cases}). A~pair-wise
evaluation is included in the supplementary material.

Notice that full images were always used for the registration. We mostly
look at the center crops for the consistency of the evaluation, but full
images were processed beforehand.

\hypertarget{specimens}{%
\subsection{Specimens}\label{specimens}}

\begin{figure}[tb]
\centering
\subfloat[][Micro-CT, rabbit lung]{\label{fig:specimens-a}%
\scalebarme{\includegraphics[height=0.26\linewidth]{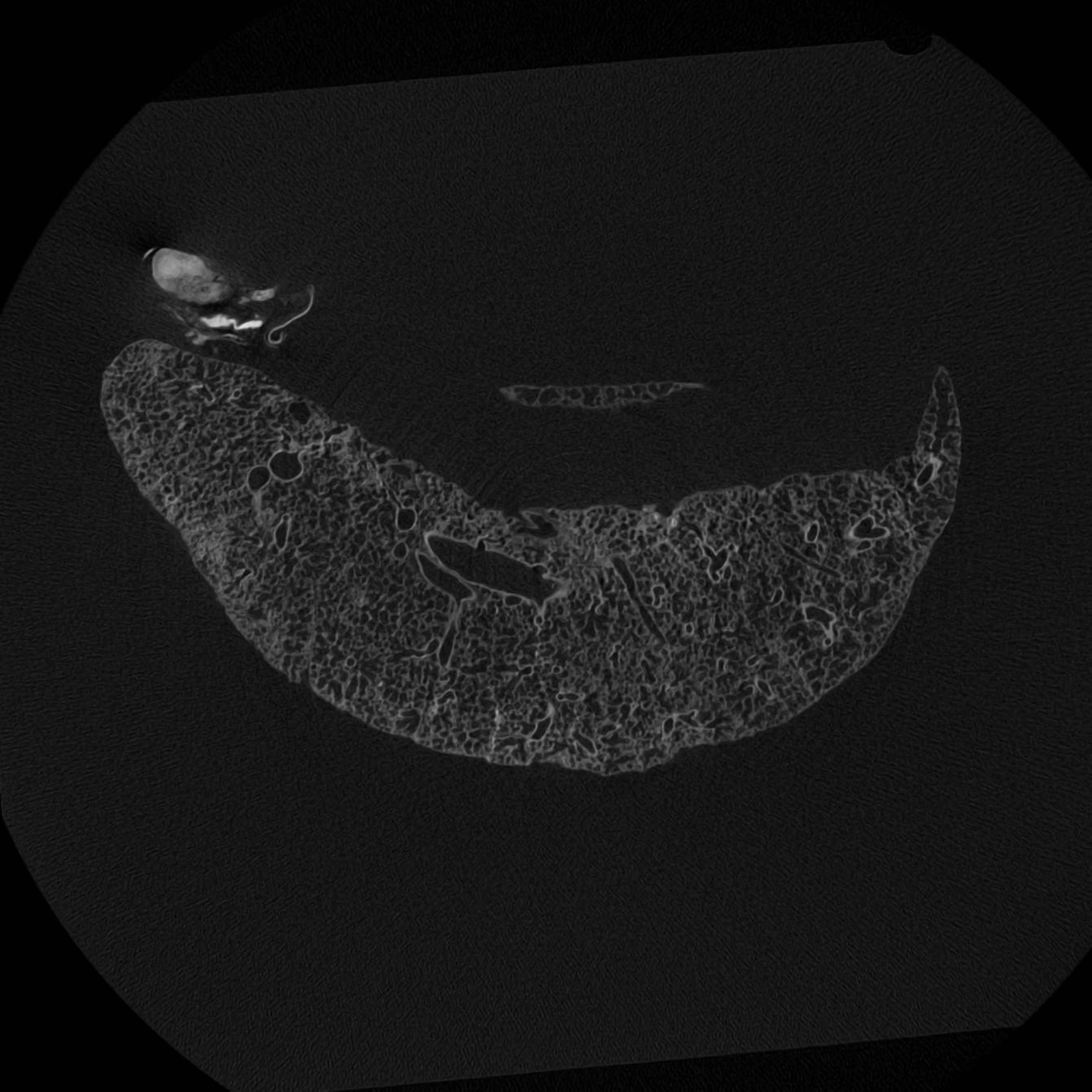}}{%
\includegraphics[width=0.26\linewidth]{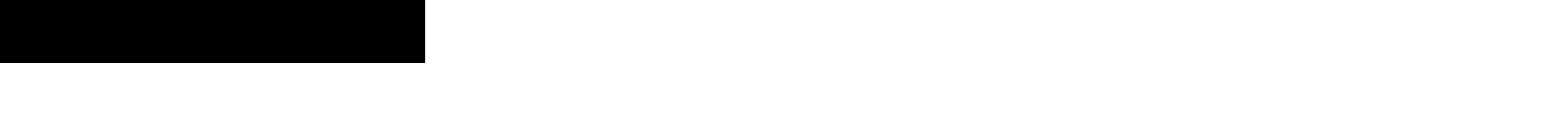}}}\hfill
\subfloat[][Electron microscopy, mouse lung]{\label{fig:specimens-c}%
\scalebarme{\includegraphics[height=0.26\linewidth]{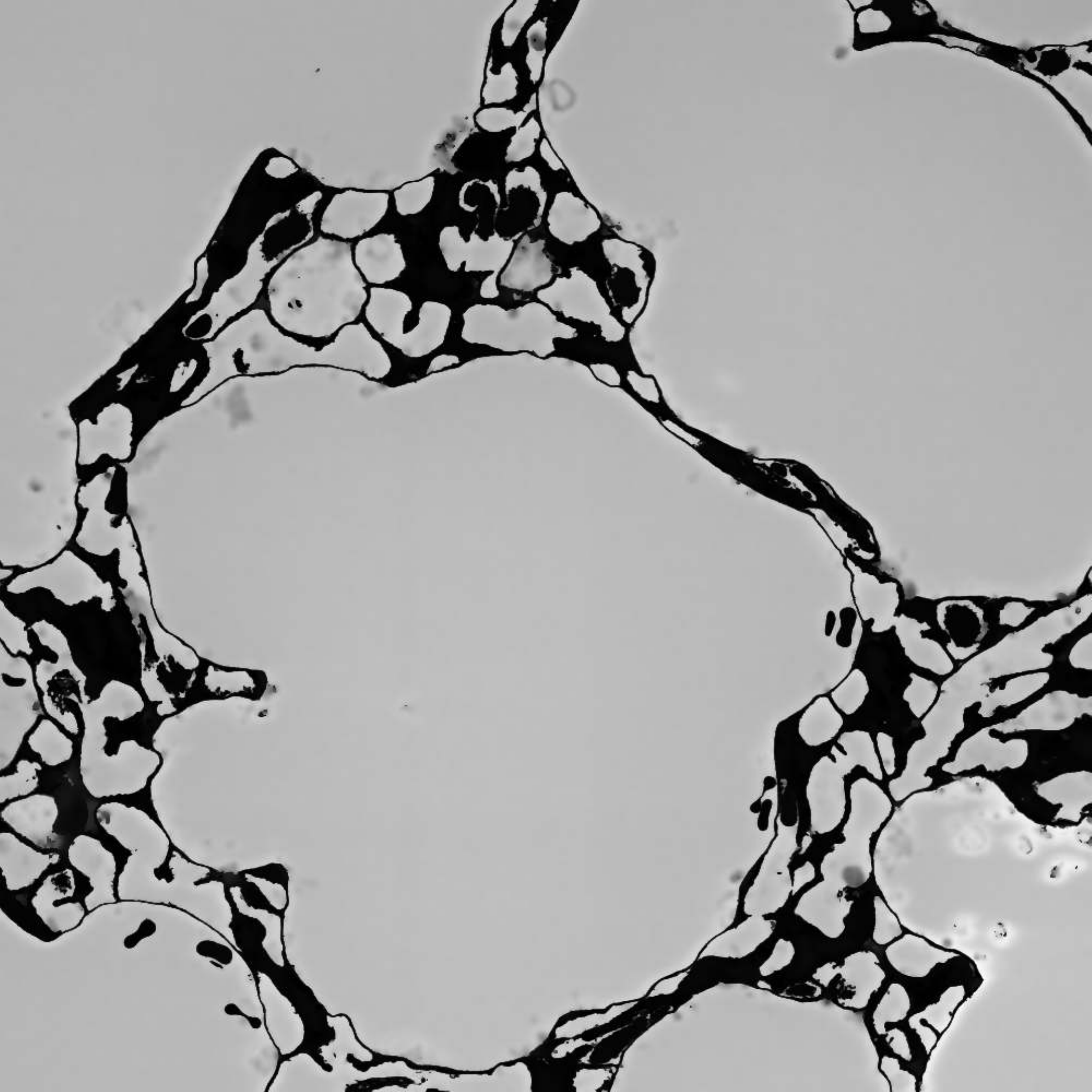}}{%
\includegraphics[width=0.26\linewidth]{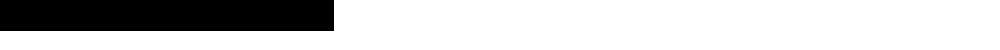}}}\hfill
\subfloat[][Light sheet microscopy, rat lung]{\label{fig:specimens-b}%
\scalebarme{\includegraphics[height=0.26\linewidth]{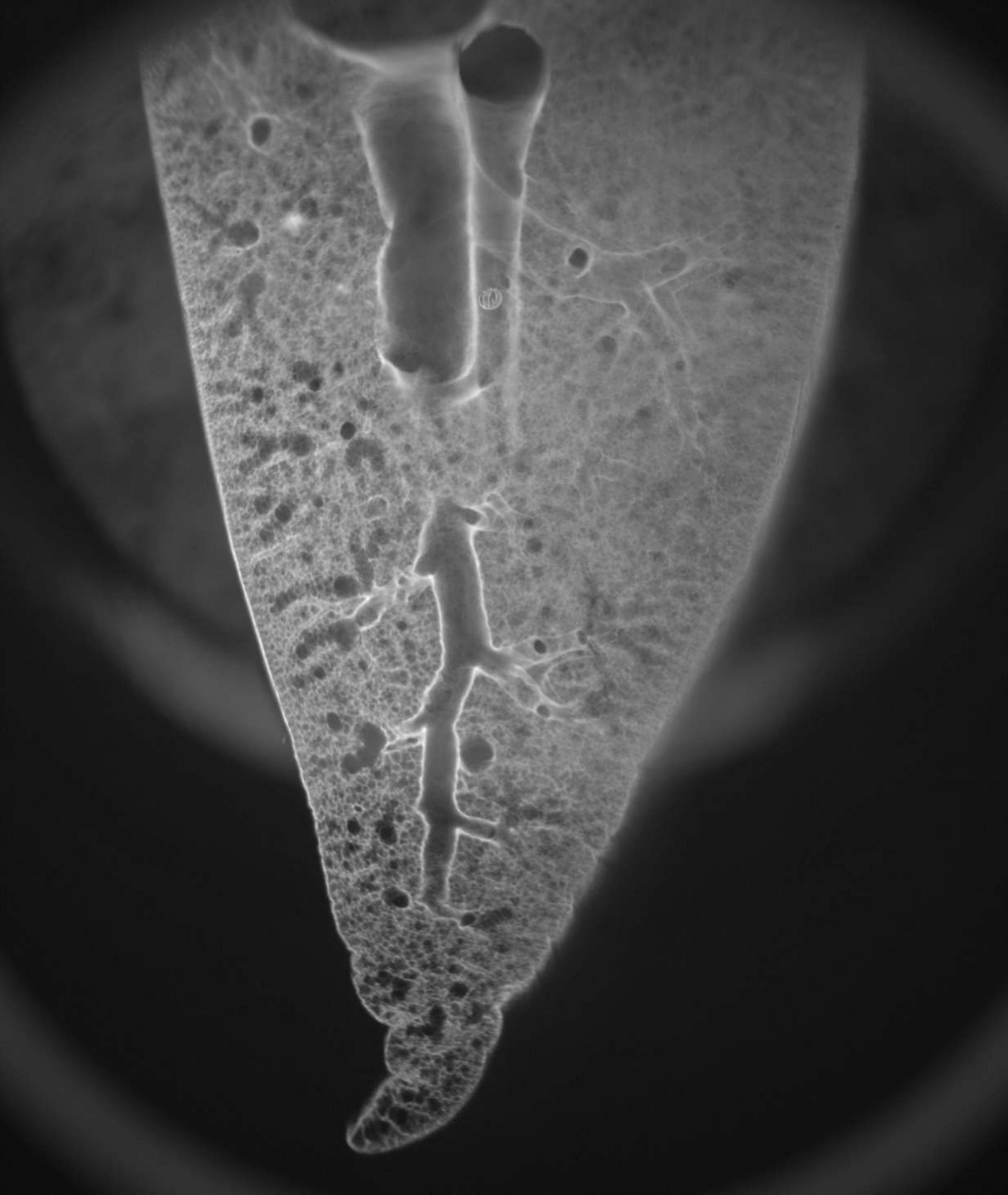}}{%
\includegraphics[width=0.219375\linewidth]{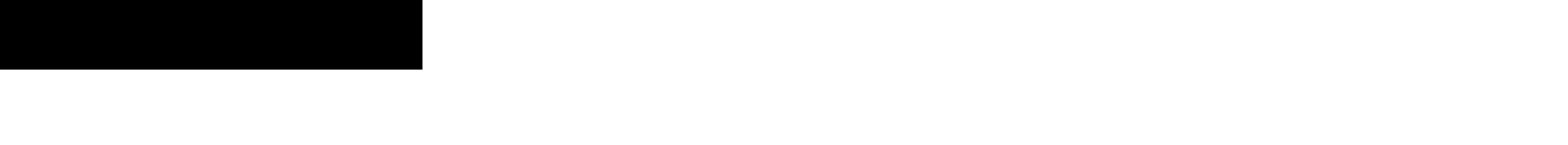}}}\hfill
\subfloat[][A fragment of a reference serial section, rabbit lung, toluidine blue]{\label{fig:specimens-d}%
\scalebarme{\includegraphics[height=0.26\linewidth]{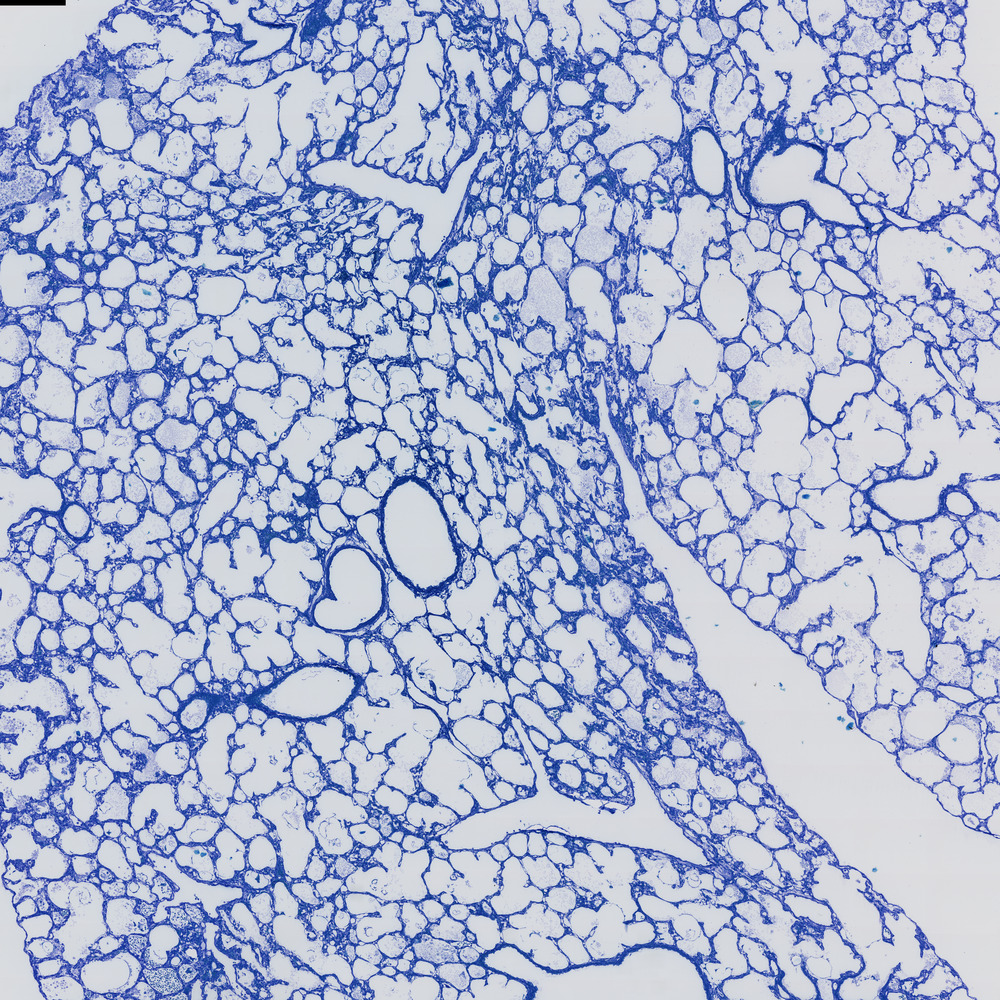}}{%
\includegraphics[width=0.26\linewidth]{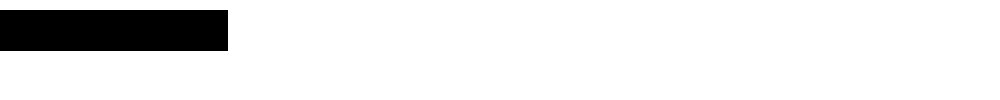}}}

\caption{The data sets used in our benchmark. The micro-CT~\protect\subref{fig:specimens-a} is filtered with anisotropic diffusion \citep{fehrenbach_sparse_2014}; the EM~\protect\subref{fig:specimens-c} and the light sheet data~\protect\subref{fig:specimens-b} are normalized individually; two consecutive non-registered serial sections~\protect\subref{fig:specimens-d} are provided as a reference. Further data sets can be created from 3D data using the methodology we present here.\newline%
Scale bars are: \protect\subref{fig:specimens-a}:~\SI{5}{\milli\meter}, \protect\subref{fig:specimens-c}:~\SI{50}{\micro\meter}, \protect\subref{fig:specimens-b}:~\SI{3}{\milli\meter}, \protect\subref{fig:specimens-d}:~\SI{1}{\milli\meter}.}
\label{fig:specimens}
\end{figure}

We applied our method to micro-CT, light sheet, and EM images of animal
lungs (Fig.~\ref{fig:specimens}). The data was processed in the manner
standard for each modality, basically, for the processing in this paper,
we perceived the data as already processed and ready-to-use 3D images.
The images were extended to the size specified below in order to not
lose data during the global movement phase. Specifically, we used:

\begin{itemize}
\item
  A rabbit lung acquired with micro-CT (Fig.~\ref{fig:specimens-a}). The
  specimen was a New Zealand White rabbit that was artificially
  delivered 3 days early by cesarean section and that spent 7~days in
  hyperoxia (\SI{95}{\percent}), the lung was perfusion fixed.
  \dhcomment{Citation of perfusion fixation?} The sample comes from a
  project studying the bronchopulmonary dysplasia in a hyperoxia preterm
  rabbit model \citep{mhh20}, part of a larger study of bronchopulmonary
  dysplasia models \citep{appuhn_capillary_2021}. The experiments have
  been approved by the ethics committee for animal experimentation of KU
  Leuven, project number P081/2017.

  The sample was imaged on a Bruker SkyScan 1272 high-resolution
  microtomography machine (Control software version 1.1.19, Bruker
  microCT, Kontich, Belgium). The X-ray source was set to a tube voltage
  of \SI{80}{\kilo\volt} and a tube current of
  \SI{125.0}{\micro\ampere}, the X-ray spectrum was filtered by
  \SI{1}{\milli\meter} of Aluminum prior to incidence onto the sample.
  We recorded a set of 2 stacked scans overlapping the sample height,
  each stack was recorded with 488 projections of \xroi{3104}{1091}
  (2~projections stitched laterally) at every \SI{0.4}{\degree} over a
  \SI{180}{\degree} sample rotation. Every single projection was exposed
  for \SI{2247}{\milli\second}, 5 projections were averaged to greatly
  reduce image noise. This resulted in a scan time of approximately 8
  hours. The projection images were then subsequently reconstructed into
  a 3D stack of images with NRecon (Version 1.7.4.2, Bruker microCT,
  Kontich, Belgium) using a ring artifact correction of 7. The whole
  process resulted in a data set of 1135 images with an isometric voxel
  size of \SI{7.0}{\micro\meter} (see also Fig.~\ref{fig:real-effect}).
  \dhcomment{Paragraph above corrected based on \path{R:/Archiv_Tape/Lung\ Muehlfeld\191146-1\proj\191146-1.log}}
  \dhcomment{I don't like to speak of 'resolution', 'voxel size' is better.}
  The images were pre-processed with a 2D anisotropic diffusion
  denoising filter based on lattice basis reduction
  \citep{fehrenbach_sparse_2014}.

  We extracted 600~images from the middle of the filtered data set for
  our benchmark and padded them (Sec.~\ref{sec:padding}), yielding
  images at \xroi{3954}{3954};
\item
  An EM serial block-face (SBF-SEM) data set of adult mouse lung
  (Fig.~\ref{fig:specimens-c}). The specimen was a 4~weeks old C57BL/6
  mouse, the lung was perfusion-fixed \citep{buchacker_assessment_2019}.
  The experiments were approved by Regierungspräsidium Karlsruhe.
  Overall, 5246 sections with \SI{80}{\nano\meter} thickness were cut in
  a Zeiss Merlin VP Compact SEM (Carl Zeiss Microscopy GmbH, Jena,
  Germany), using a Gatan 3View2XP system (Gatan Inc., Pleasanton, CA,
  USA). The block-face was captured with the view port of
  525\(\mytimes\)\SI{525}{\micro\meter}, yielding \kroi{15}{15} with
  \SI{0.5}{\micro\second} dwell time, \SI{3.0}{\kilo\volt} acceleration
  voltage and variable pressure mode at \SI{30}{\pascal}.

  The benchmark uses a crop from the full data set with 1000 images at
  \kroi{1.5}{1.5}. The final resolution is
  \SI{0.15}{\micro\meter /voxel}. The data set was denoised with
  gradient anisotropic diffusion using ITK before usage. We did not
  apply any registration in post-processing, but we individually
  normalized the images---as detailed below;
\item
  A~lung for the light sheet (LS) data set was obtained from a male 24
  week-old Fisher~344 rat with a body weight of \SI{320}{\gram}, which
  was part of a ventilation study approved by the LAVES in Oldenburg,
  the number of animal experiment proposal is~17/2608. The lung was
  fixed in an inflated state with an airway pressure corresponding to
  \SI{20}{\centi\meter} of \ce{H2O} and perfusion fixed, compare
  \citet{krischer_mechanical_2021}.
  \dhcomment{Again, maybe a citation to the perfusion fixation, as it's given for the EM?}
  By means of a \enquote{tissue slicer} the lung was cut in slices of
  \SI{2}{\milli\meter} thickness.

  The image data was acquired with the UltraMicroscope~II (LaVision
  BioTec GmbH, Bielefeld, Germany). The lung slices were pinned up to a
  mandrel in the ethyl cinnamate-filled detection chamber and
  illuminated unidirectionally with 6 light sheets. An~sCMOS camera
  detected the fluorescence light with a wavelength of
  \SI{490}{\nano\meter}, which matches the tissues autofluorescence,
  perpendicular to the illumination plane. Due to the large dimensions
  of the rat lung and the intention to depict the complete lung the only
  zoom factor to choose was~0.63, corresponding a 1.26-fold
  magnification. The series was acquired as 336~images with
  \textup{\num[round-mode = figures,
  round-precision = 3 ]{5.15875}$\mytimes$\num[round-mode = figures,
  round-precision = 3 ]{5.15875}$\mytimes$\SI{15}{\micro\meter /voxel}}
  (Fig.~\ref{fig:specimens-b}).

  For the benchmark we use 300 images at \kroi{3,9}{3,9} that were
  individually normalized, see below;
\item
  As a reference, we also provide two serial sections of the same rabbit
  lung as in micro-CT, stained with toluidine blue.
  Fig.~\ref{fig:specimens-d} shows one of those sections. The images
  were acquired in transmitted LM with a Zeiss AxioScan.Z1 scanning
  microscope (Carl Zeiss Microscopy GmbH, Jena, Germany) at
  \SI{0.22}{\micro\meter /pixel} (20\(\times\) lens). The section
  thickness was~\SI{2}{\micro\meter}.
\end{itemize}

\hypertarget{data-preparation-normalization-and-availability}{%
\subsection{Data preparation, normalization, and
availability}\label{data-preparation-normalization-and-availability}}

\label{sec:padding} All benchmark images were extended to a larger
square to reduce the loss of information during rotation and
translation. Their bit depth was reduced to 8~bit, LS and EM images were
normalized \emph{individually} using ImageMagick and GNU parallel
\citep{tange_2020_4118697}. The normalization of individual images is
introduced to mimic a slightly varying image intensity
\citep{bagci_automatic_2010, khan_nonlinear_2014, nadeem_multimarginal_2020}
due to varying thickness of slices
\citep{hanslovsky_post-acquisition_2015} or varying penetration of the
fixation and the staining solutions
\citep[\egX ][]{bulmer_observations_1962, lanir_histological_1984, dullin_ct_2017}.

\olcomment{CHECKLIST SUPPLEMENTARY, also add: elastix config}

We provide as supplementary data the original, undistorted images, the
locally distorted images, the locally distorted and rigidly transformed
images, the local distortions, and the values for the rigid
transformations.

\hypertarget{special-cases}{%
\subsection{Special cases}\label{special-cases}}

\begin{figure}[tbh]
\subfloat[][CT, local dist.]{\label{fig:special-a}%
\scalebarme{%
\includegraphics[width=0.199\linewidth]{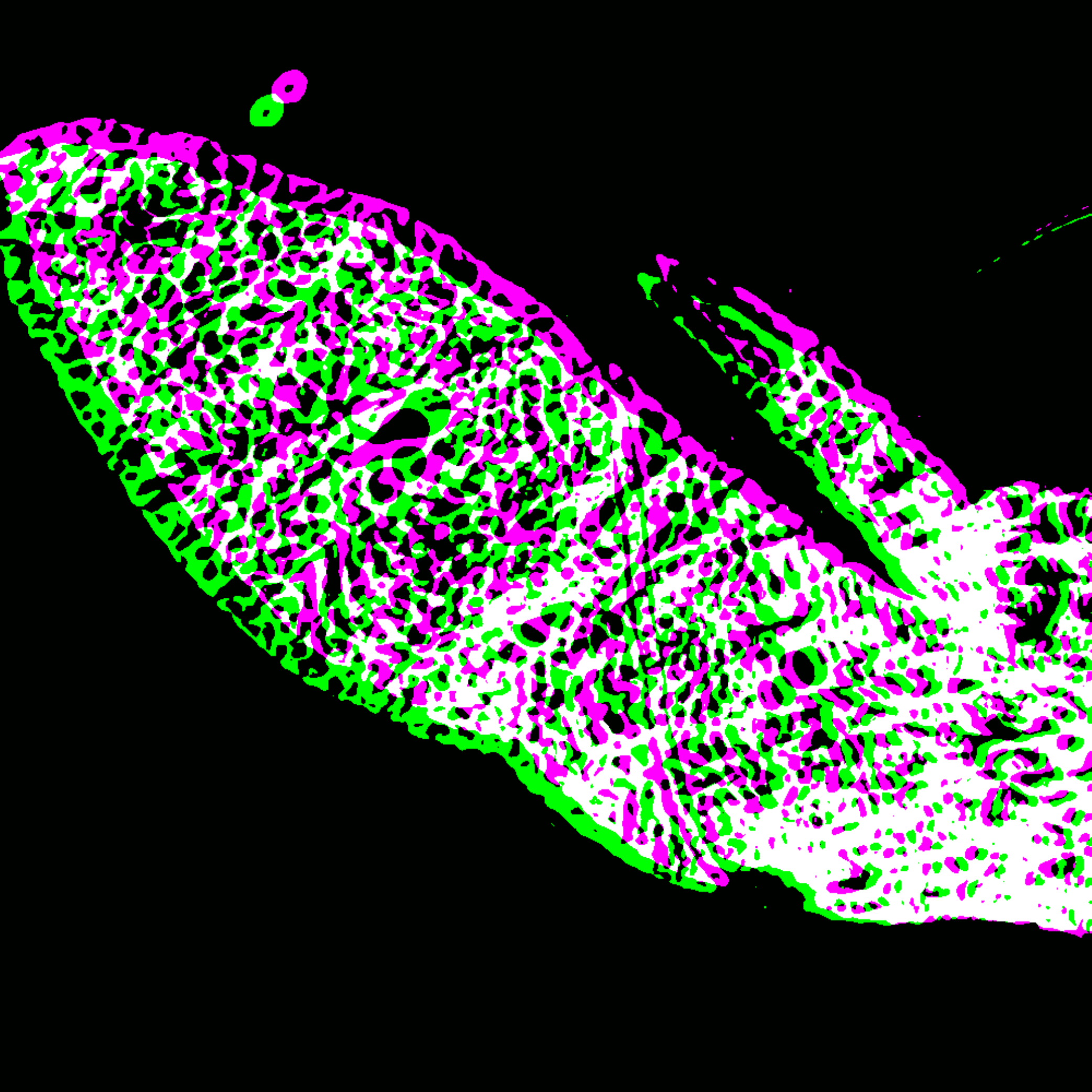}}{%
\includegraphics[width=0.199\linewidth]{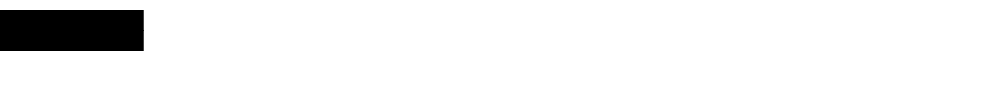}}}\hfill%
\subfloat[][CT, \elastix]{\label{fig:special-b}%
\scalebarme{%
\includegraphics[width=0.199\linewidth]{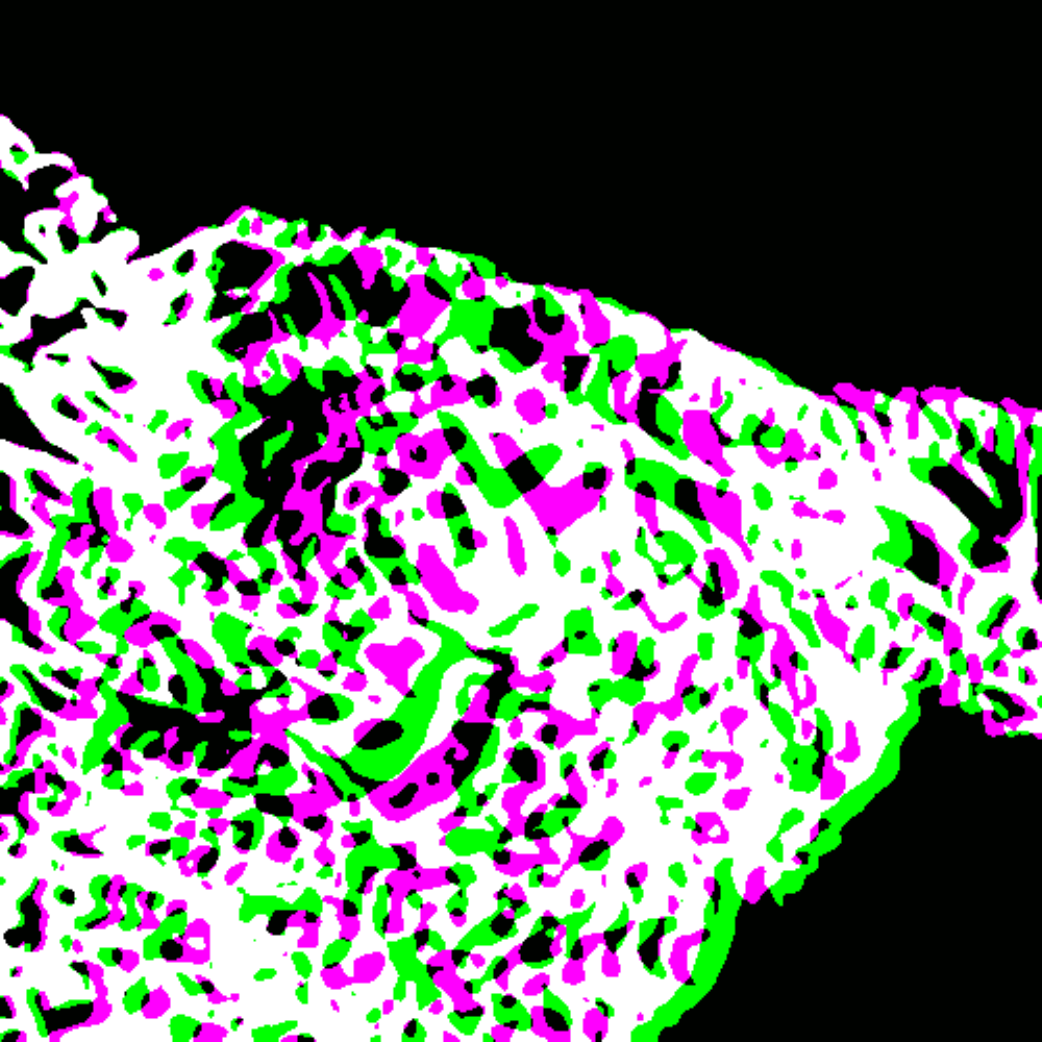}}{%
\includegraphics[width=0.199\linewidth]{image/scalebars/scale_ct-500.png}}}\hfill%
\subfloat[][CT, \elastix]{\label{fig:special-c}%
\scalebarme{\includegraphics[width=0.199\linewidth]{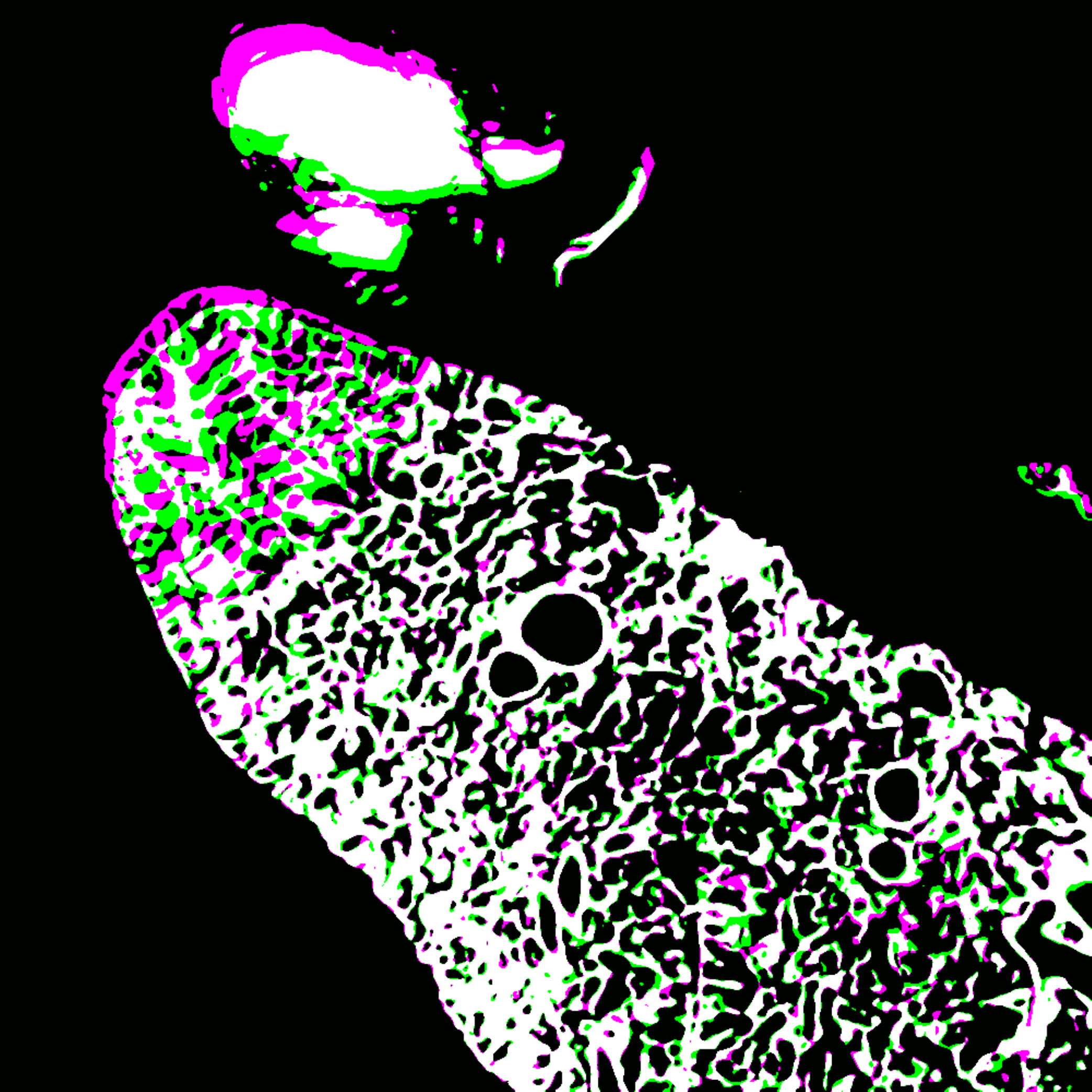}}{%
\includegraphics[width=0.199\linewidth]{image/scalebars/scale_ct-13a.png}}}\hfill%
\subfloat[][CT, \enquote{Deform-SURF}]{\label{fig:special-surf-good-1}%
\scalebarme{\includegraphics[width=0.199\linewidth]{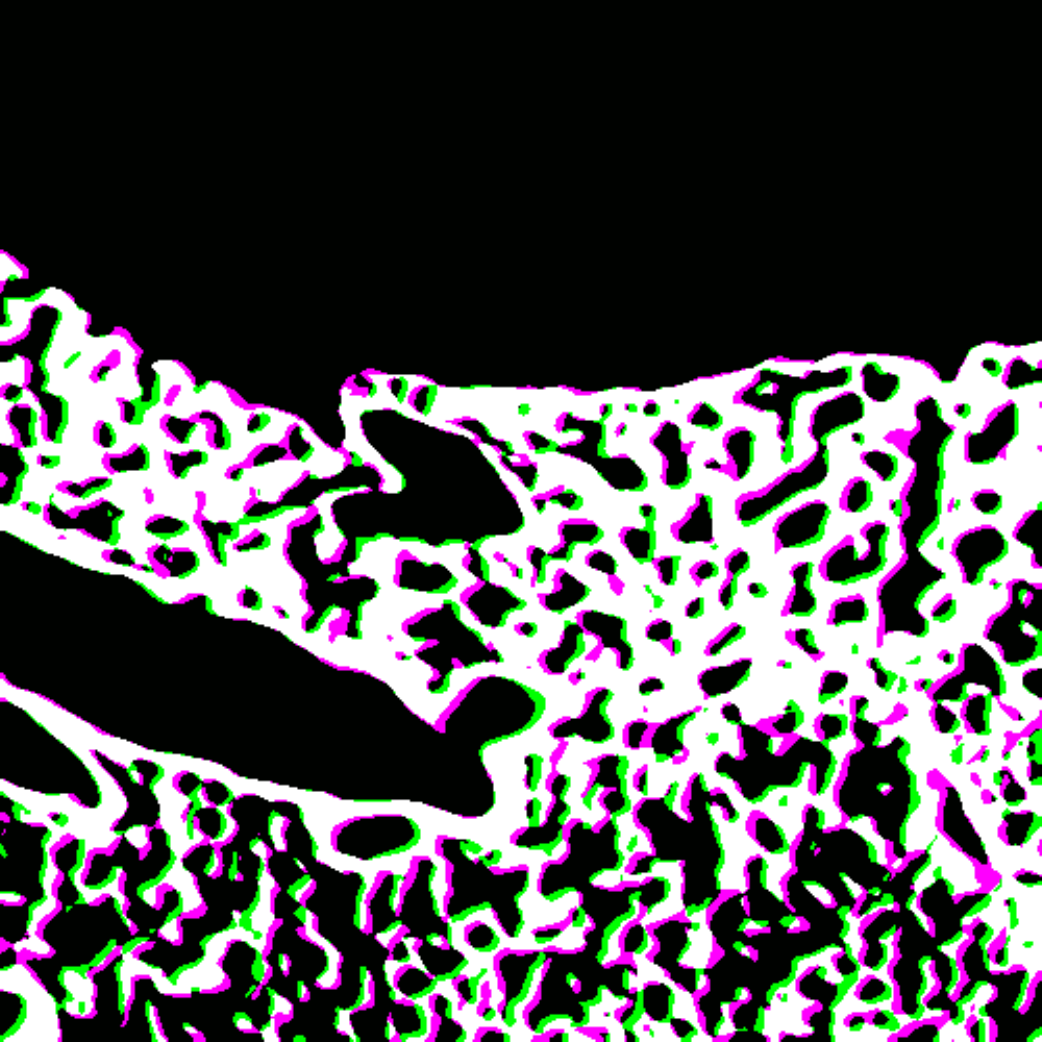}}{%
\includegraphics[width=0.199\linewidth]{image/scalebars/scale_ct-500.png}}}\hfill%
\subfloat[][CT, \enquote{Deform-SURF}]{\label{fig:special-surf-good-2}%
\scalebarme{\includegraphics[width=0.199\linewidth]{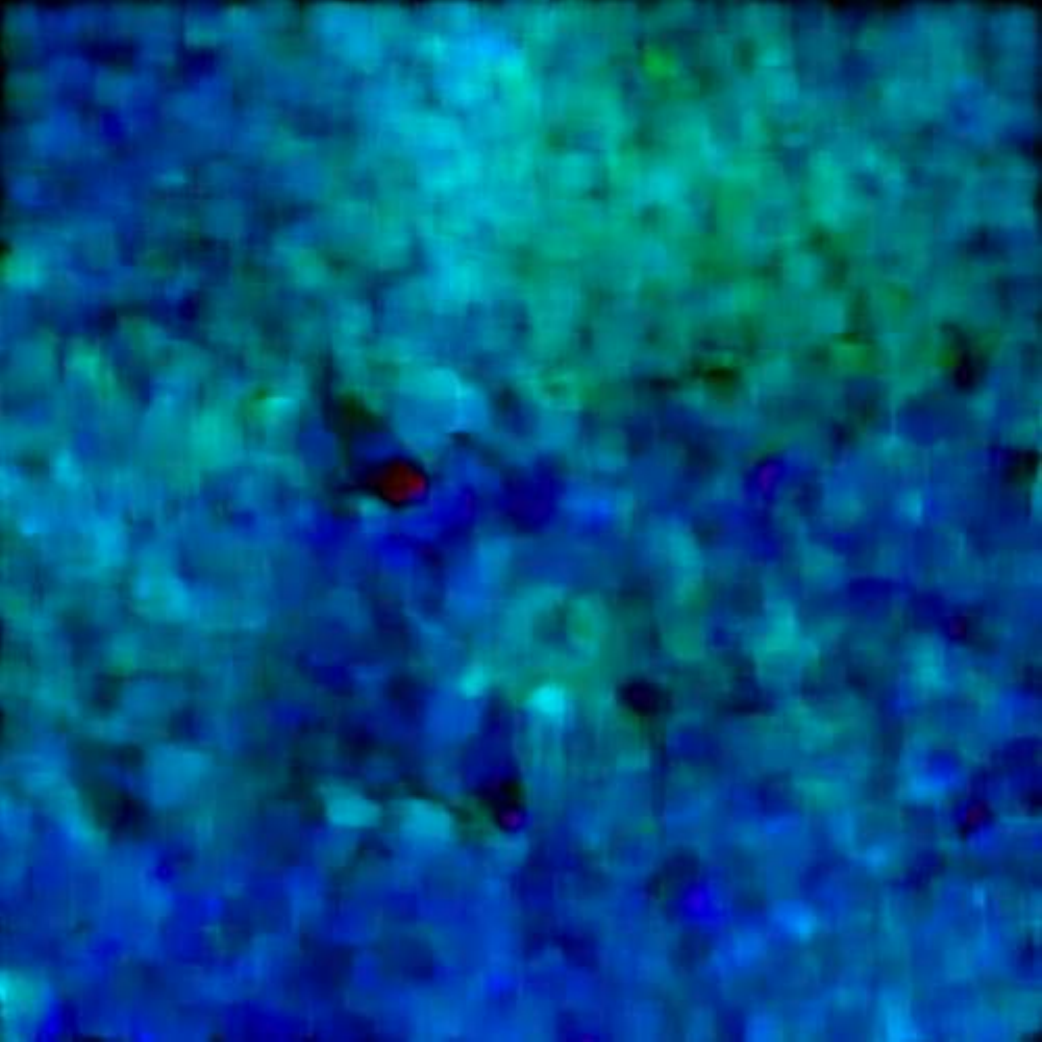}}{%
\includegraphics[width=0.199\linewidth]{image/scalebars/scale_ct-500.png}}}

\subfloat[][EM, \enquote{Rigid-SIFT}]{\label{fig:special-em-fail1}%
\scalebarme{%
\includegraphics[width=0.199\linewidth]{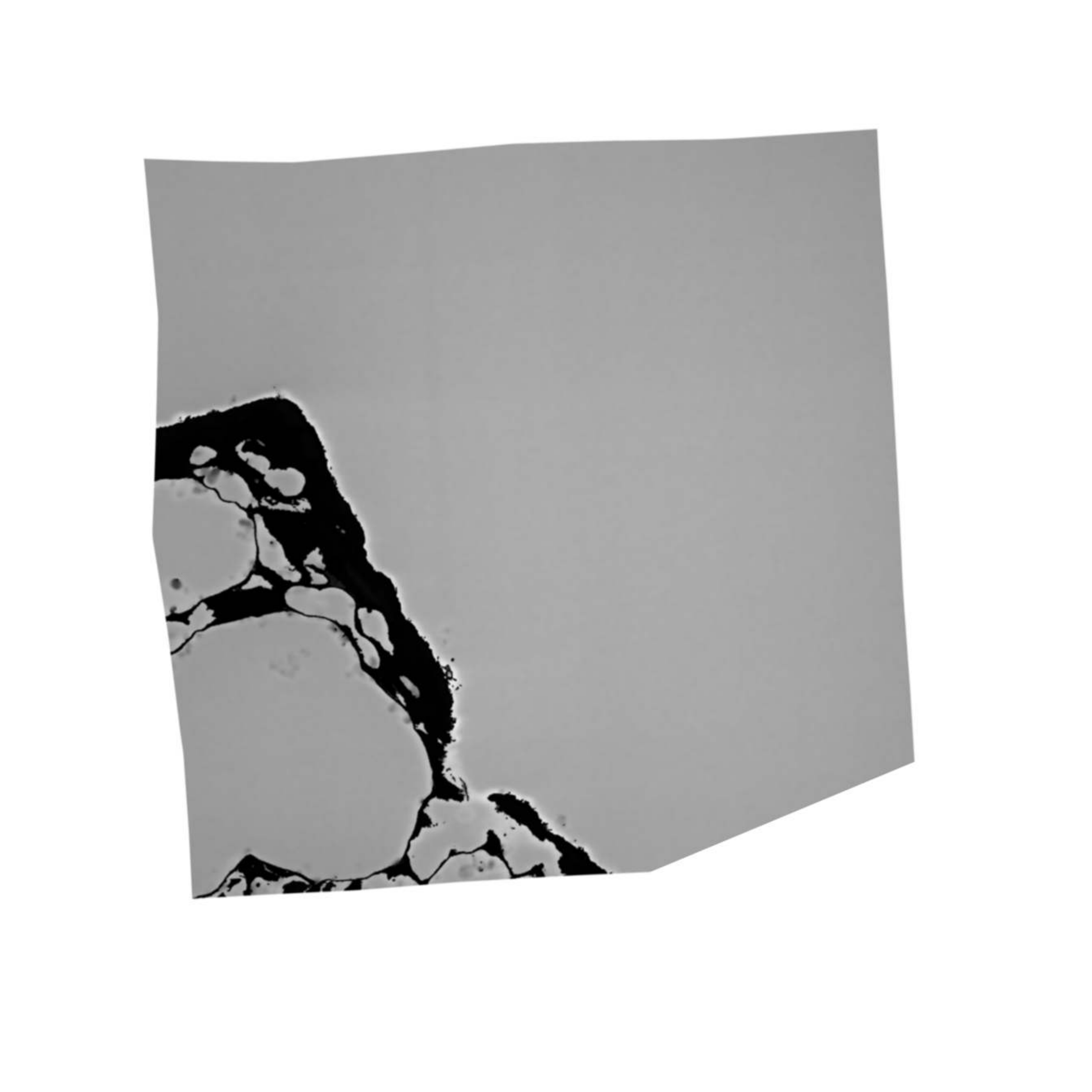}}{%
\includegraphics[width=0.199\linewidth]{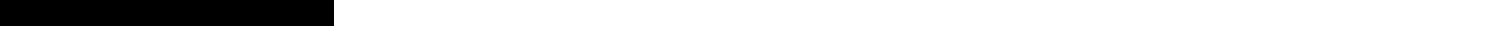}}}\hfill%
\subfloat[][EM, \enquote{Rigid-SIFT}]{\label{fig:special-em-fail2}%
\scalebarme{%
\includegraphics[width=0.199\linewidth]{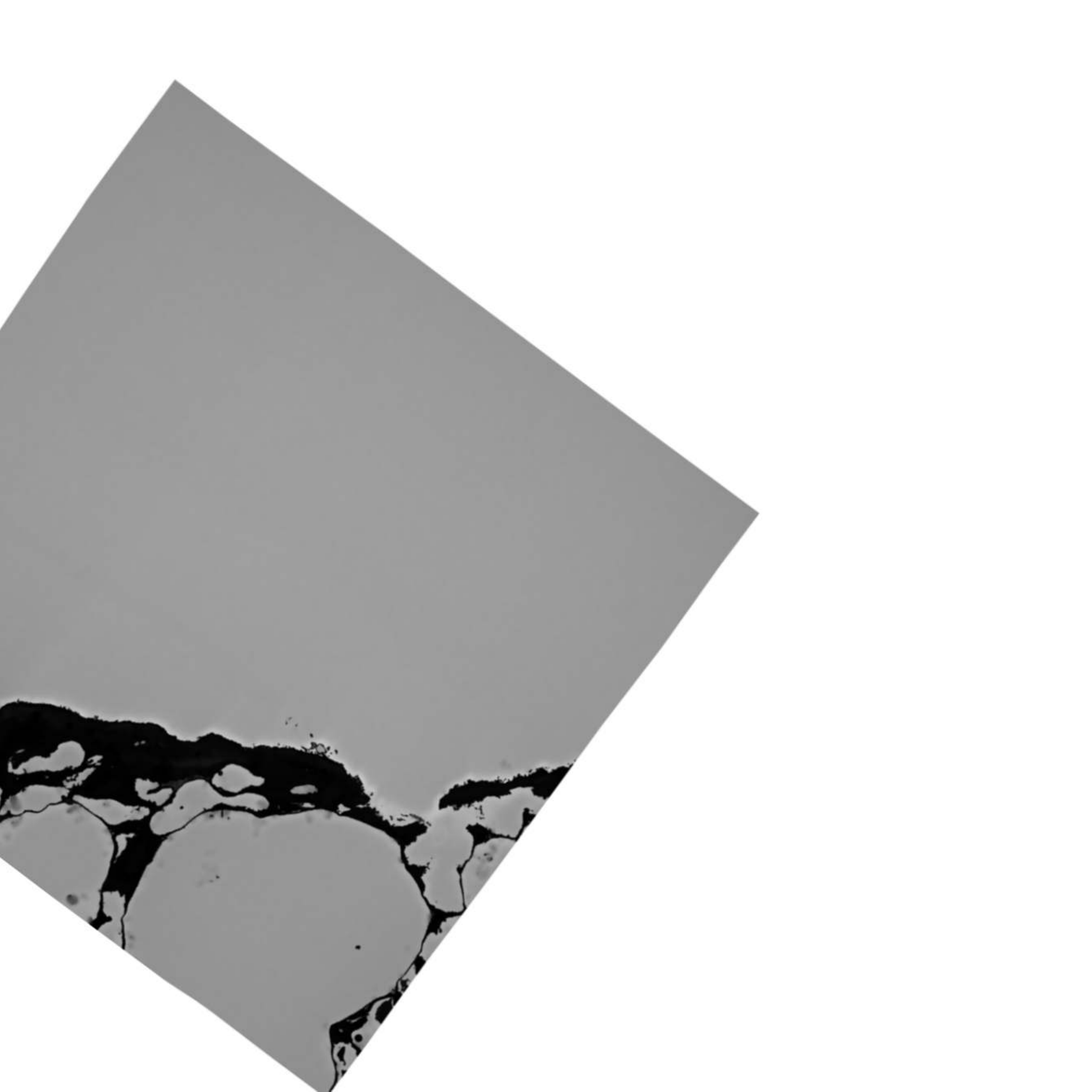}}{%
\includegraphics[width=0.199\linewidth]{image/scalebars/scale_em-13f.png}}}\hfill%
\subfloat[][LS, \enquote{GS}]{\label{fig:special-sg-good-1}%
\scalebarme{%
\includegraphics[width=0.199\linewidth]{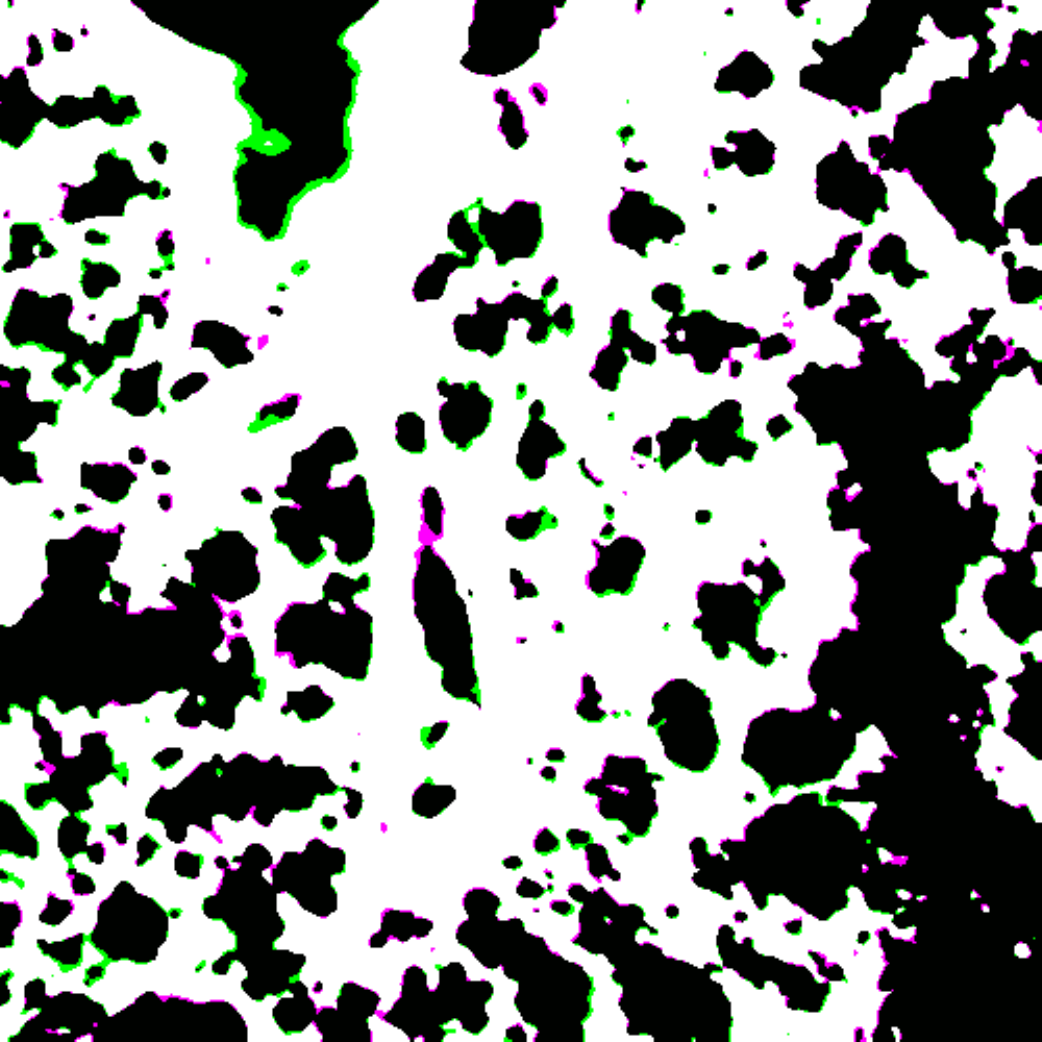}}{%
\includegraphics[width=0.199\linewidth]{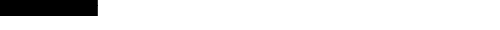}}}\hfill%
\subfloat[][LS, \enquote{GS}]{\label{fig:special-sg-good-2}%
\scalebarme{%
\includegraphics[width=0.199\linewidth]{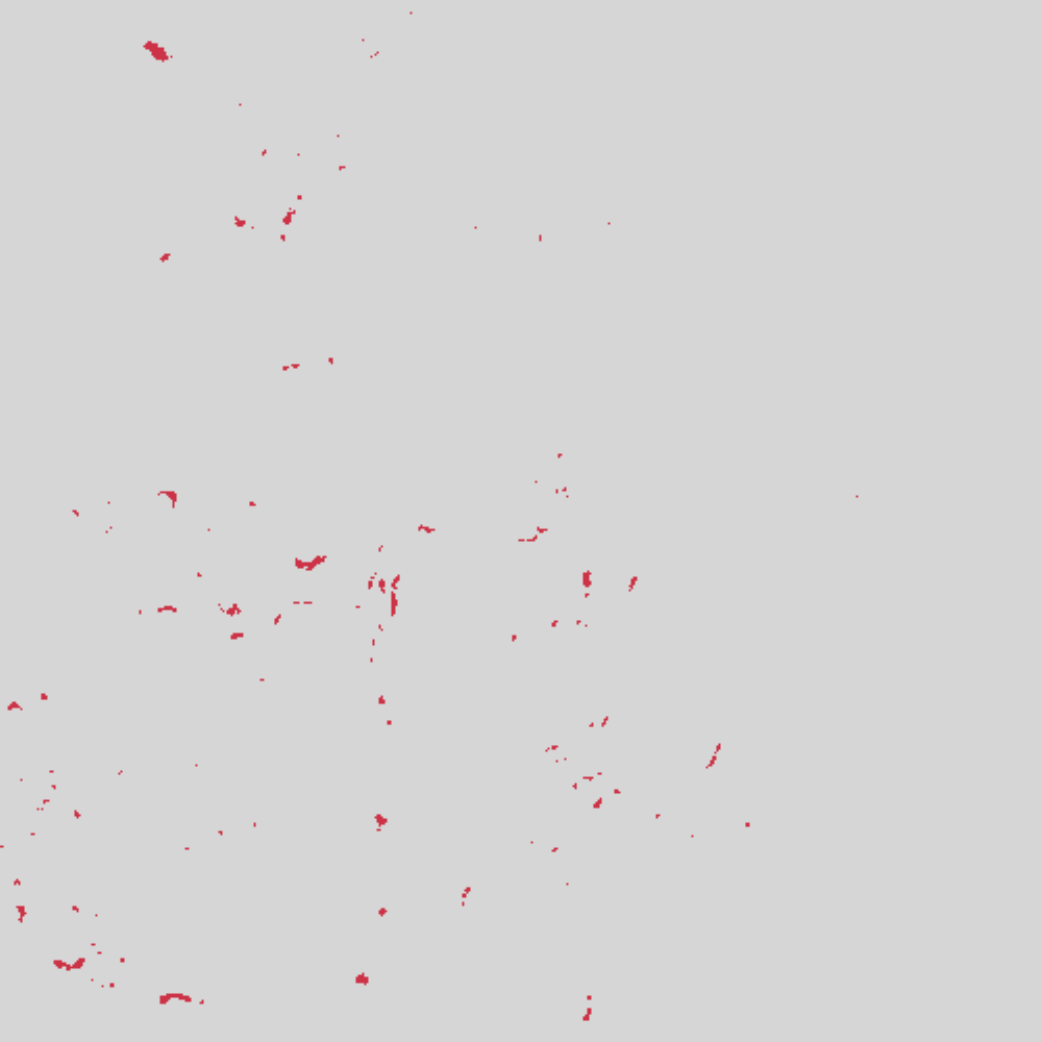}}{%
\includegraphics[width=0.199\linewidth]{image/scalebars/scale_ls-500.png}}}\hfill%
\subfloat[][LS, \enquote{GS}]{\label{fig:special-sg-good-3}%
\scalebarme{%
\includegraphics[width=0.199\linewidth]{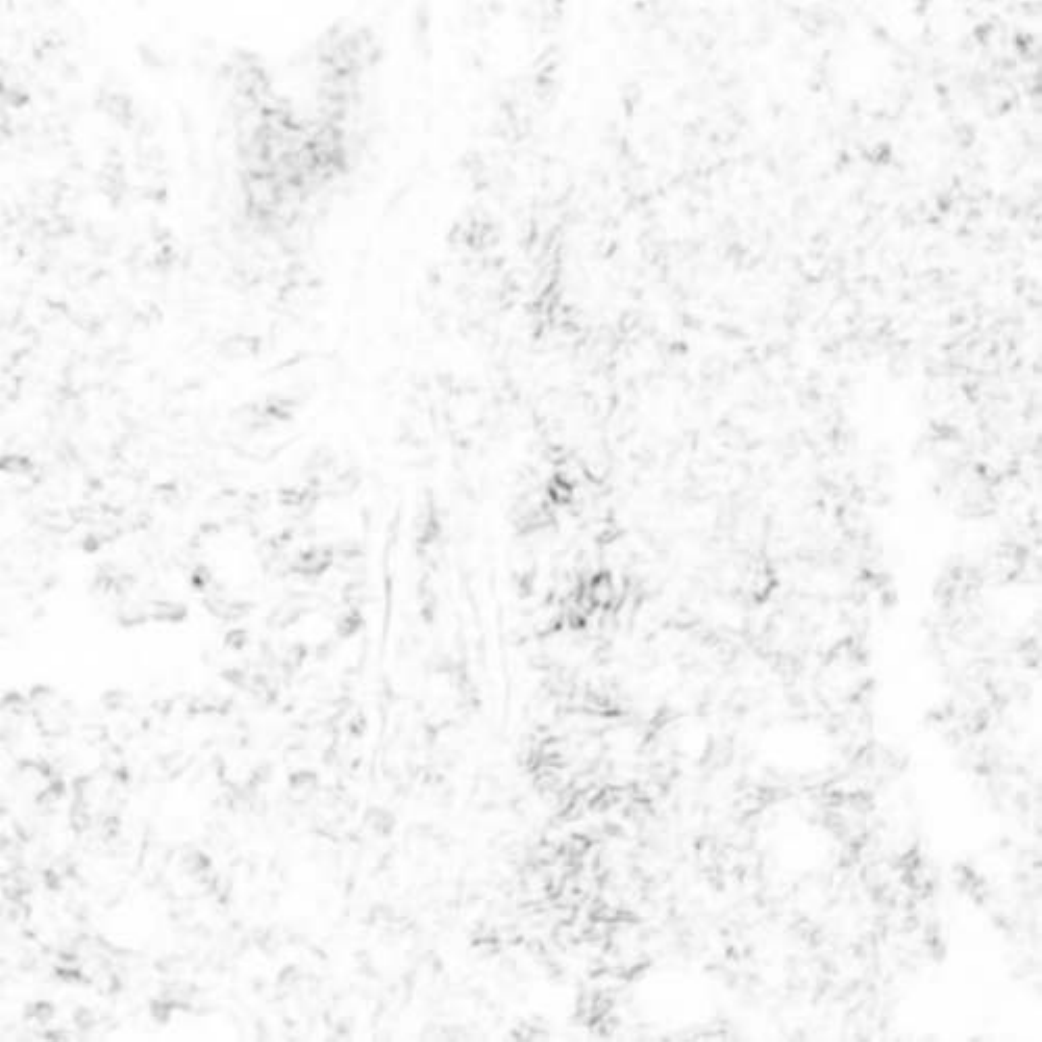}}{%
\includegraphics[width=0.199\linewidth]{image/scalebars/scale_ls-500.png}}}
\caption{Special cases, especially good or bad image pairs.\newline%
Scale bars are: \protect\subref{fig:special-a}$=$\SI{1}{\milli\meter},
\protect\subref{fig:special-b}$=$\SI{500}{\micro\meter},
\protect\subref{fig:special-c}$=$\SI{1}{\milli\meter},
\protect\subref{fig:special-surf-good-1}$=$\SI{500}{\micro\meter},
\protect\subref{fig:special-surf-good-2}$=$\SI{500}{\micro\meter},
\protect\subref{fig:special-em-fail1}$=$\SI{50}{\milli\meter},
\protect\subref{fig:special-em-fail2}$=$\SI{50}{\milli\meter},
\protect\subref{fig:special-sg-good-1}--\protect\subref{fig:special-sg-good-3}$=$\SI{500}{\micro\meter}.}
\label{fig:special}
\end{figure}

For the pair-wise evaluation in supplementary material we intentionally
picked the center of a series and a clearly defined region. Here we
would like to separately highlight some especially good of bad
consecutive image pairs in Fig.~\ref{fig:special}. This is a subjective
selection of image pairs, as contrasted with the next section.

In panel~\subref{fig:special-a} a~larger global shift in non-linear
distortion phase of the registrations' input is shown. It is CT data
set, sections 95--96, Dice visualization. Fig.~\subref{fig:special-b}
shows uncorrected global movement in \elastix-based registration, same
data set, sections 99--100, Dice visualization. There is
a~\enquote{floppy end}, a movement, in \elastix result
\subref{fig:special-c}, from same data set, sections 305--306, Dice
visualization. Mostly the registration is good, but in the depicted
region the offsets are much larger.

In panels~\subref{fig:special-surf-good-1},
\subref{fig:special-surf-good-2} the measures of an~interesting image
pair from \enquote{Deform-SURF} are depicted. We show Dice
\subref{fig:special-surf-good-1} and optical flow
\subref{fig:special-surf-good-2} visualizations from the same region.
There \emph{is} some movement, but is it all explainable with the
\enquote{natural} differences in consecutive images?

The areas with little tissue (as found in our EM series near its end)
are a greater challenge for the feature-based methods, as
Figs.~\subref{fig:special-em-fail1}, \subref{fig:special-em-fail2} show.
We see there a failure of the feature-based SIFT method to find a
correct rigid alignment. Full images from EM data set, sections
702--703, are shown. The probable reason is the low number of viable key
points found by the feature detection.

Subfigures~\subref{fig:special-sg-good-1}--\subref{fig:special-sg-good-3}
show a small evaluation of a particularly good GS-registered image pair,
LS data set, images 161--162. We show there Dice visualization, PSNR,
and SSIM, respectively, for the same region. Notice the low values and
very few differences.

\hypertarget{full-series-evaluation}{%
\subsection{Full-series evaluation}\label{full-series-evaluation}}

After we have looked into the relations of two consecutive images from
the middle of the series\olcomment{REWRITE!!!}, a question arises
naturally, how the quality measures look throughout the series. For this
sake we computed the three numerical measures SSIM, Jaccard, PSNR over
the whole series and evaluate these values statistically. All Jaccard
values below are computed with threshold~100.

\hypertarget{methodology}{%
\subsubsection{Methodology}\label{methodology}}

\label{sec:full-method}

Our concept of the evaluation focuses on comparing consecutive sections
from the registered series to the same consecutive sections from the
ground truth. We decided against comparing the images from the
registered series directly to the ground truth images: Accumulated
errors from the rigid transformations and non-rigid distortions impact
such comparisons. We would be more interested in how the now-registered
series fits to \emph{itself} in comparison to how it \emph{should have
fit,} that in a direct comparison that would find a lot of mismatch that
would be of less interest in practice. To give a simple example, many
methods might over-fit the registrations of their inputs for the better
numerical quality values (\enquote{over-registration}, \enquote{banana
problem}). With our ground truth series we would be able to find such
cases.

The measures were computed on the \xroi{500}{500} crop from the middle
of the images, throughout the series. Lower values in the beginning and
in the end of the series can be explained with less tissue in the
region. However, we advocate the use of the center crop as a simple to
define and \enquote{fair} way to define a region. As detailed in the
supplementary material, it is impractical to use the full image for the
evaluation. Also, some methods used additional padding, making the
uniform and comparable use of general cropping offsets harder. The
middle of the image should arguably have meaningful tissue contents in
most cases. The \enquote{drifting away} tissue, \ie the case when
different registrations accumulate the errors so differently, that we
obtain fully different image regions at the same offset, is rather an
exception. This way, we have meaningful content in the most of the
series duration.

We present box plots of the appropriate values of quality measures. In
this case we decided against using violin plots. Violin plots show
outliers similarly, but the median and the shape of the inliers can be
discerned with less clarity in most of our particular cases. (We still
present some violin plots in Fig.~\ref{fig:box-ls-SP}, see also
supplementary material.)

We also present a statistical evaluation. We performed an unpaired
\(t\)-test with different means and unequal variances, a Welch two
sample \(t\)-test, to be exact. We always compared a measure of
registration results to the same measure of the ground truth.

\hypertarget{ct}{%
\subsubsection{CT}\label{ct}}

\begin{figure}[p]
\centering
\subfloat[][]{\label{fig:box-ct-a}\includegraphics[width=0.9\linewidth]{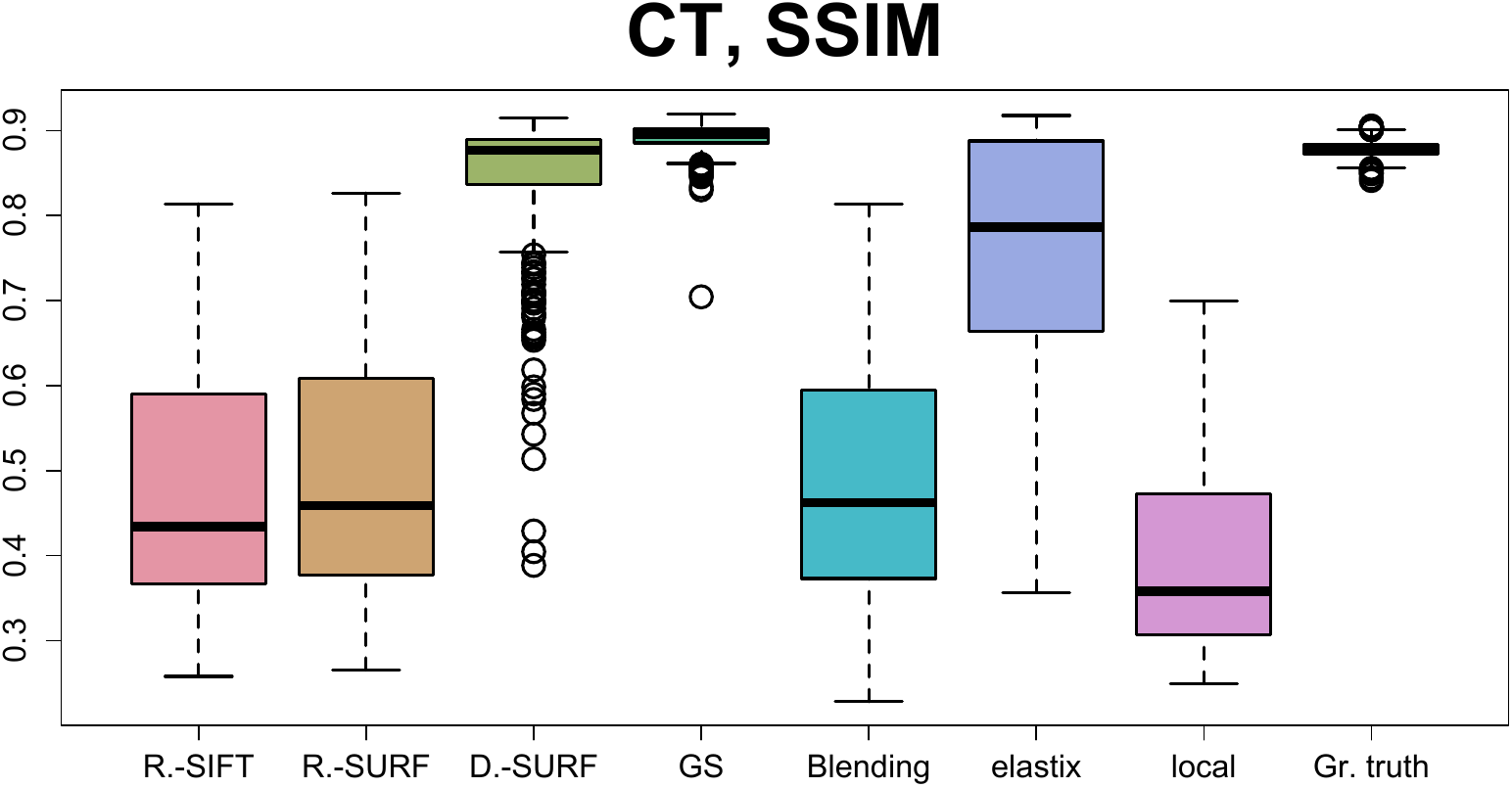}}

\subfloat[][]{\label{fig:box-ct-b}\includegraphics[width=0.9\linewidth]{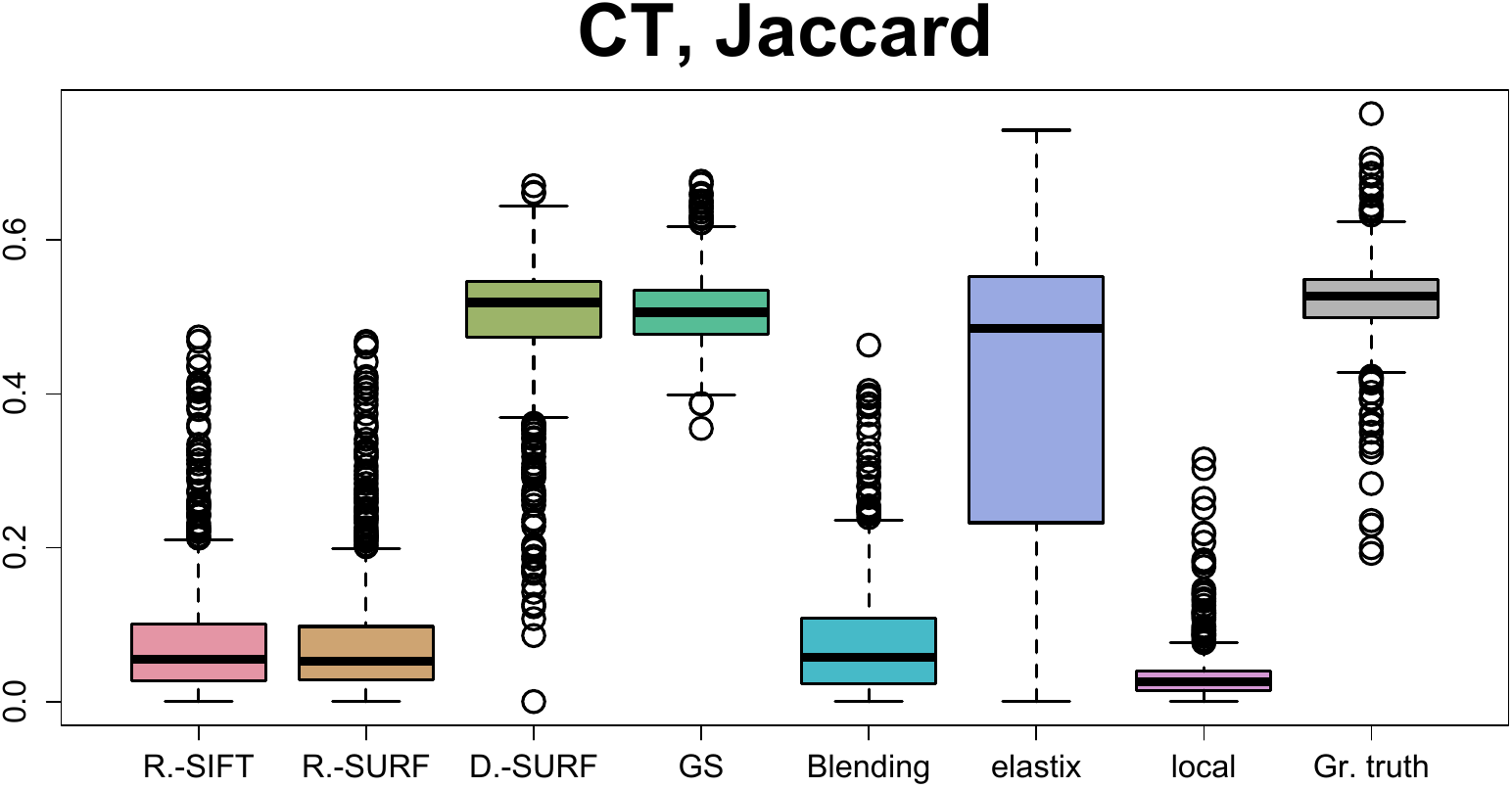}}

\subfloat[][]{\label{fig:box-ct-c}\includegraphics[width=0.9\linewidth]{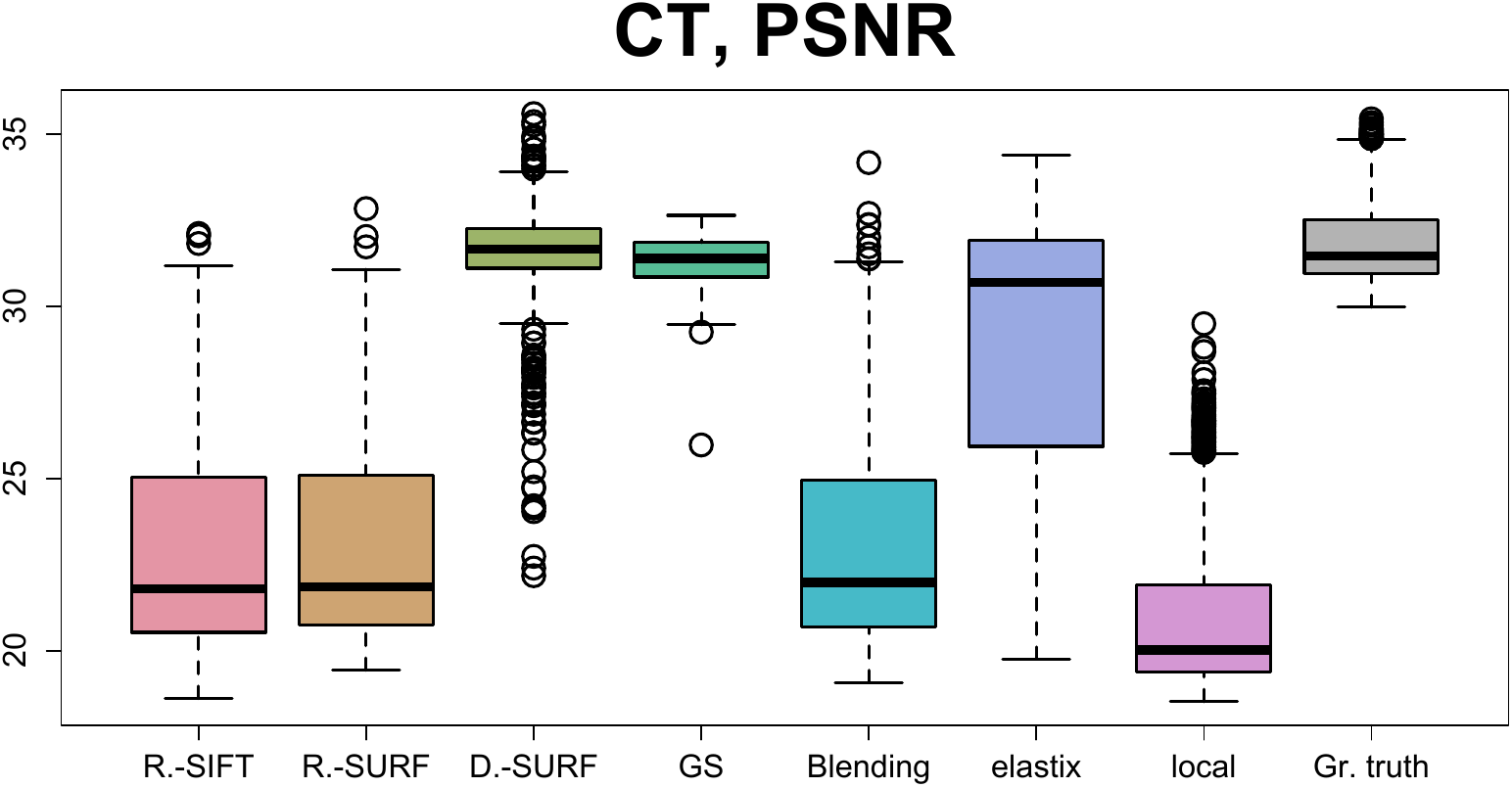}}

\caption{Box plots of quality measures for the CT data set. \enquote{R.-SIFT} stands for Rigid-SIFT, \enquote{R.-SURF} for Rigid-SURF, \enquote{D.-SURF} stands for Deform-SURF, \enquote{local} means local distortions only, without a global rigid transformation, \enquote{Gr.~truth} stands for ground truth.}
\label{fig:box-ct}
\thisfloatpagestyle{empty} 

\end{figure}

Consider Fig.~\ref{fig:box-ct}, presenting box plots of the full-series
evaluations. Overall \enquote{Deform-SURF} and \enquote{GS} stand out
for their good performance. Panel~\subref{fig:box-ct-a} shows SSIM
values. Notice, how the median in \enquote{GS} is higher than in ground
truth (\num{0.89517} vs.~\num{0.87745}). In panel~\subref{fig:box-ct-a}
\enquote{GS} and \enquote{Deform-SURF} are close to the ground truth,
while there are a lot of outliers in \enquote{Deform-SURF} and
\enquote{GS} seems to overshoot a bit, but has some lower outliers.
There are few outliers in the ground truth, too, but they are rather
symmetric. The latter also holds for the Jaccard measure. In
panel~\subref{fig:box-ct-b} there are some outliers with \emph{high}
Jaccard values in \enquote{GS}. In~\subref{fig:box-ct-c} we see, again,
similar values in \enquote{GS} and \enquote{Deform-SURF} to the ground
truth, but now there are many outliers with \emph{high} and very low
PSNR values in \enquote{Deform-SURF}. Notice also the shape of the
ground truth box plot for PSNR: there are quite many values above the
median. Overall, \elastix has good results, but quite tall box plots,
indicating high variance. A~possible reason is that \elastix operated
pair-wise on the image sequence in this case. Both \enquote{Deform-SURF}
and \enquote{GS} operate on a full image stack at once.

Table~\ref{tab:stat-ct} shows a statistical evaluation. We aim to decide
with a Welch \(t\)-test, if the quality measures of a registered series
are similar to the ground truth. This is almost never the case.
In~\enquote{GS} \wrt Jaccard measure (\subref{tab:stat-ct-jaccard}) we
see the largest similarity, according to the test, but \(p\) is still
quite low there, under~\num{4.3E-4}. Sometimes, the maximal values of
the quality measures for a registration method are higher than the
maximal ground truth value for the same measure. We attribute this to a
wider \enquote{spread} of the variance, induced by the registration. To
give an example for the SSIM measure, the variance of
\enquote{Deform-SURF} for the full series is
\num[round-mode = figures,round-precision = 3,scientific-notation = true]{0.004289982},
while the variance of the ground truth is
\num[round-mode = figures,round-precision = 3,scientific-notation = true]{0.0001052532}.

\begin{table}[p]
\caption{The Welch two sample $t$-test for the quality measures on CT data set.
Similar as above, the \enquote{locally distorted} values originate from applying only local distortions with our method. The registrations' input was also globally transformed.\newline%
Concerning the appropriate measures' values, we compare the means from each of the registration results to the mean for the ground truth. The differences are statistically significant in almost all cases, we show the $p$ value for \enquote{not equal}. The hypothesis of the equal mean was almost always refuted with a high confidence.
\newline%
We additionally show the maximal value of a measure (\enquote{Max} column). The \enquote{local} row is for the local distortions only, without global movements.\newline%
The mean value marked with \textasteriskcentered{} is the one closest to the ground per $t$-test. The maximum values marked with \textdagger{} are larger than the maximum of the ground truth. This is a \enquote{crime} many good methods commit in our evaluation.\newline%
The closer is the mean to the ground truth, the better.}
\label{tab:stat-ct}
\footnotesize
\centering
\subfloat[][]{%
\label{tab:stat-ct-ssim}
\centering
\begin{tabular}{r S[round-mode = figures, round-precision = 3, table-format = 0.3] S[round-mode = figures, round-precision = 2, table-format = 1.1, table-figures-exponent = 2] S[round-mode = figures, round-precision = 3, table-format = 0.3]}
\multicolumn{4}{c}{CT, SSIM} \\
\toprule
{Method} & {Mean} & {$p <$} & {Max}\\
\midrule
Rigid-SIFT  & 0.4765567 & 2.2e-16   & 0.81395\\
Rigid-SURF  & 0.4892105 & 2.2e-16   & 0.82664\\
Deform-SURF & 0.8517332 & 2.2e-16   & 0.91469\tabworst\\
GS          & 0.8926576 & 2.2e-16   & 0.91995\tabworst\\
Blending    & 0.4838853 & 2.2e-16   & 0.81356\\
\elastix    & 0.7545392 & 2.2e-16   & 0.91774\tabworst\\
Locally distorted
            & 0.3907646 & 2.2e-16   & 0.69903\\
Ground truth& 0.8784045 & {--}      & 0.9053\\
\bottomrule
\end{tabular}
}

\subfloat[][]{%
\label{tab:stat-ct-jaccard}
\centering
\begin{tabular}{r S[round-mode = figures, round-precision = 3, table-format = 0.3] S[round-mode = figures, round-precision = 2, table-format = 1.1, table-figures-exponent = 2] S[round-mode = figures, round-precision = 3, table-format = 0.3]}
\multicolumn{4}{c}{CT, Jaccard} \\
\toprule
{Method} & {Mean} & {$p <$} & {Max} \\
\midrule
Rigid-SIFT  & 0.08174676 & 2.2e-16   & 0.4738806\\
Rigid-SURF  & 0.08148026 & 2.2e-16   & 0.4680851\\
Deform-SURF & 0.4930395  & 5.716e-10 & 0.670412\\
GS          & 0.5097182\tabbest
                         & 0.0004338
                         & 0.6762564\\
Blending    & 0.07927393 & 2.2e-16   & 0.4632372\\
\elastix    & 0.3905055  & 2.2e-16   & 0.742515\\
Locally distorted
        & 0.03399986 & 2.2e-16   & 0.3152302\\
Ground truth& 0.52018005 & {--}      & 0.7641026\\
\bottomrule
\end{tabular}
}

\subfloat[][]{%
\label{tab:stat-ct-psnr}
\centering
\begin{tabular}{r S[round-mode = figures, round-precision = 3, table-format = 2.1] S[round-mode = figures, round-precision = 2, table-format = 1.1, table-figures-exponent = 2] S[round-mode = figures, round-precision = 3, table-format = 2.1]}
\multicolumn{4}{c}{CT, PSNR} \\
\toprule
{Method} & {Mean} & {$p <$} & {Max}\\
\midrule
Rigid-SIFT  & 23.10125 & 2.2e-16   & 32.1167\\
Rigid-SURF  & 22.97632 & 2.2e-16   & 32.8383\\
Deform-SURF & 31.49589\tabbest
                       & 3.121e-07 & 35.5956\tabworst\\
GS          & 31.32603 & 2.2e-16   & 32.6414\\
Blending    & 23.42809 & 2.2e-16   & 34.1742\\
\elastix    & 28.88296 & 2.2e-16   & 34.3893\\
locally distorted
            & 21.04715 & 2.2e-16   & 29.4969\\
Ground truth& 31.94023 & {--}      & 35.445\\
\bottomrule
\end{tabular}
}
\thisfloatpagestyle{empty} 
\end{table}

\hypertarget{em}{%
\subsubsection{EM}\label{em}}

\begin{figure}[p]
\centering
\subfloat[][]{\label{fig:box-em-a}\includegraphics[width=0.9\linewidth]{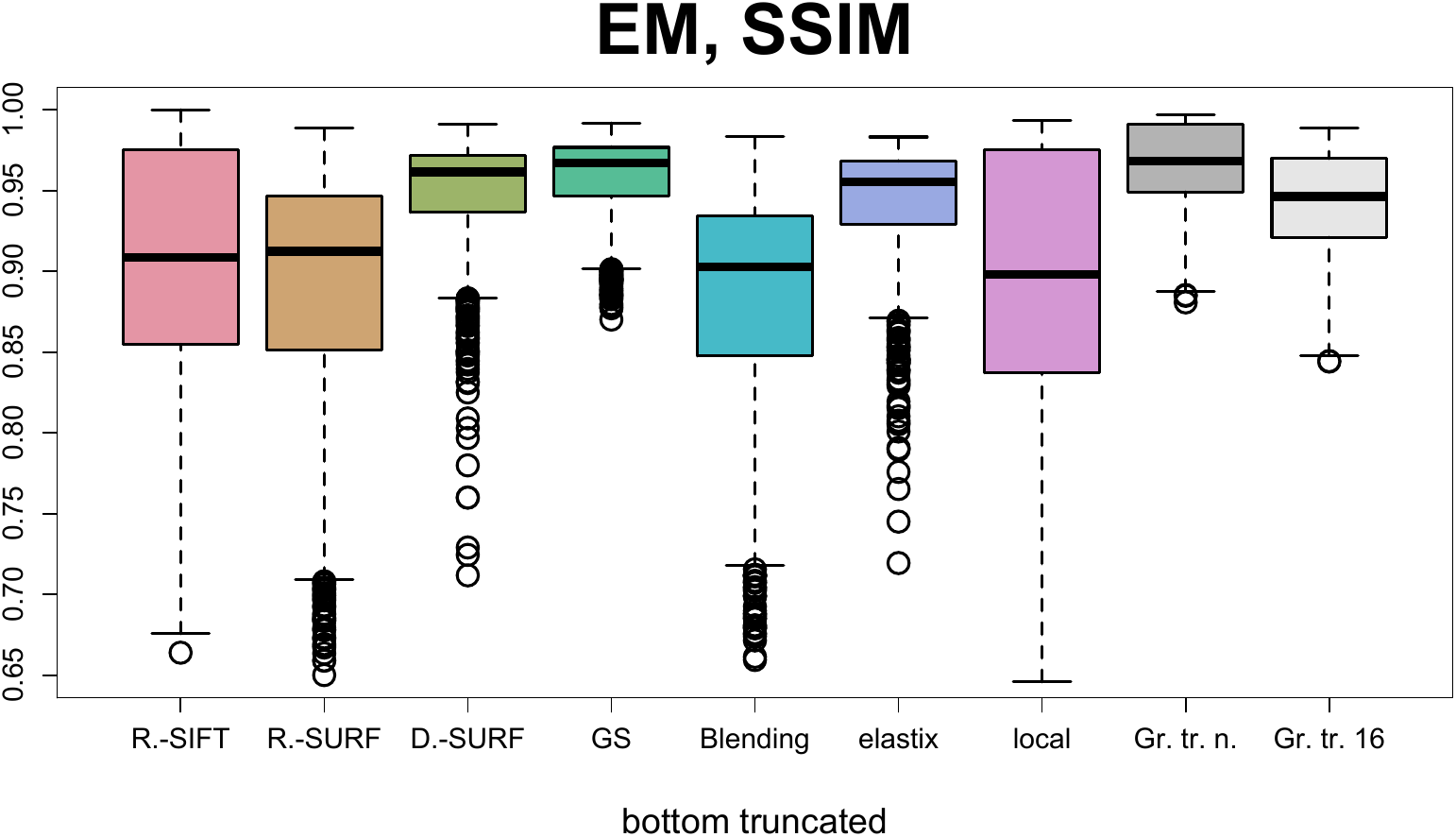}}

\subfloat[][]{\label{fig:box-em-b}\includegraphics[width=0.9\linewidth]{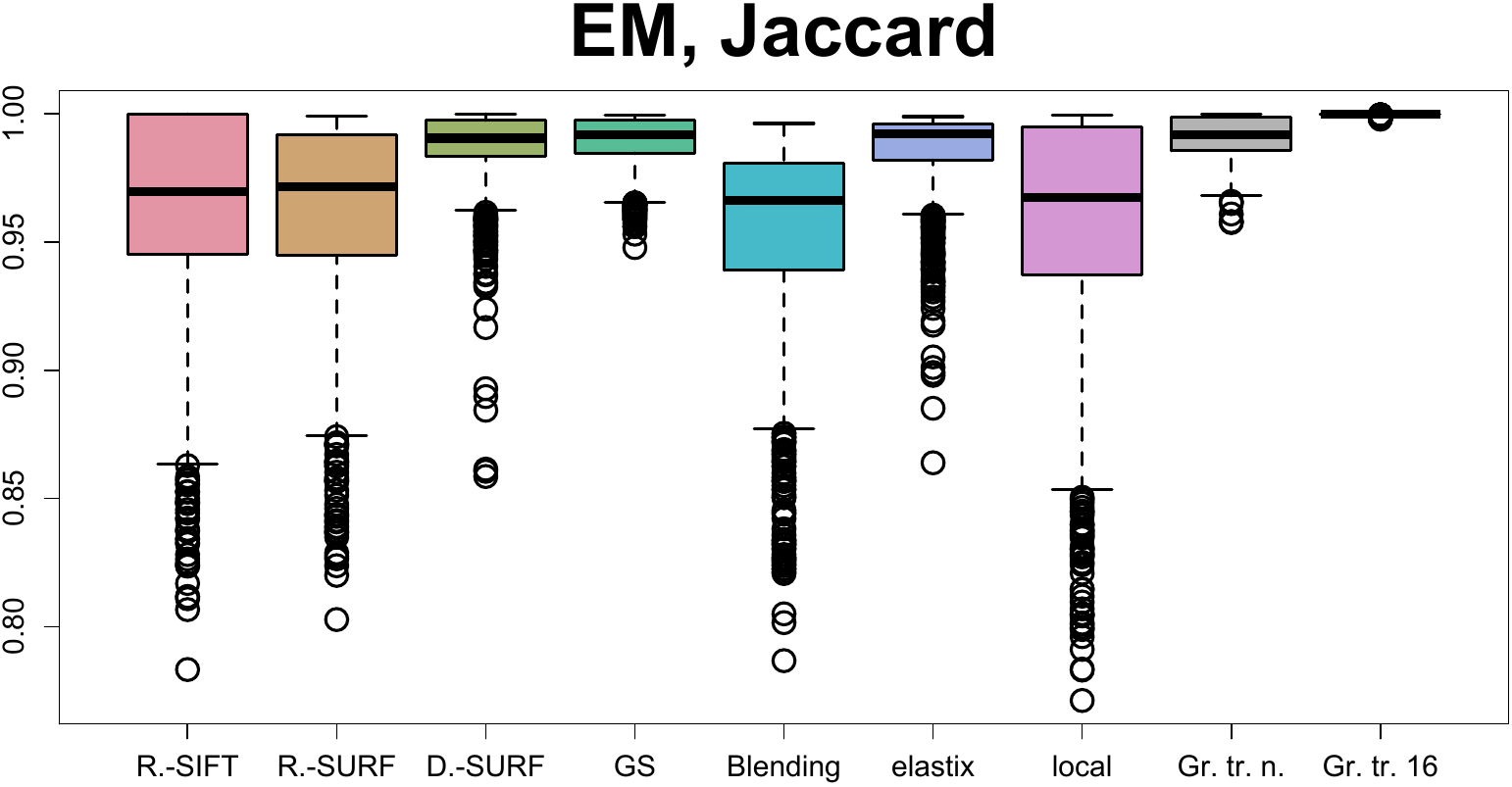}}

\subfloat[][]{\label{fig:box-em-c}\includegraphics[width=0.9\linewidth]{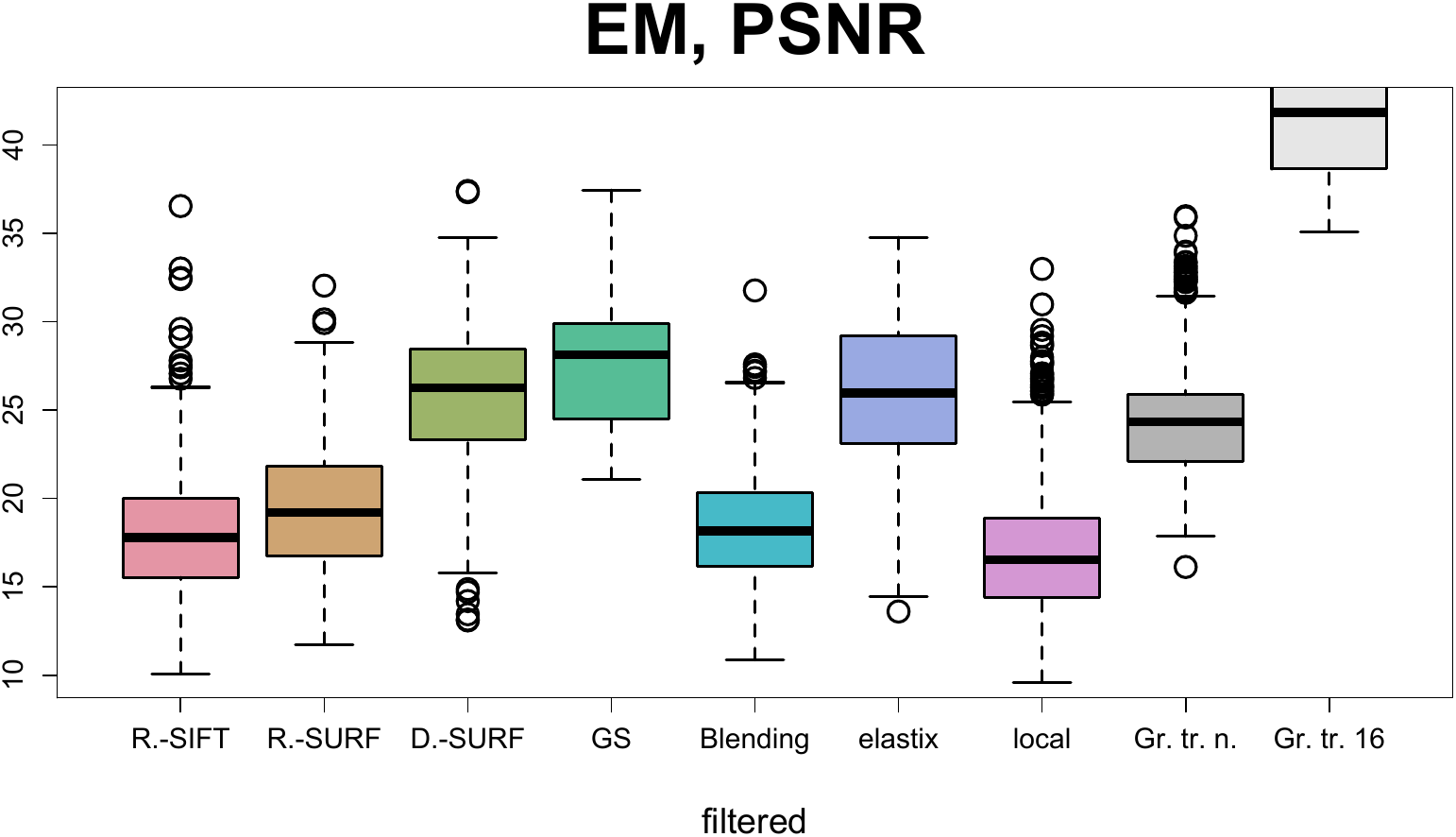}}

\caption{Box plots of quality measures for the EM data set. \enquote{R.-SIFT} stands for Rigid-SIFT, \enquote{R.-SURF} for Rigid-SURF, \enquote{D.-SURF} for Deform-SURF, \enquote{local} for only local distortions, \enquote{Gr.~tr.~n.} for \enquote{ground truth, normalized}, \enquote{Gr.~tr.~16} for \enquote{ground truth, 16~bit}.}
\label{fig:box-em}
\thisfloatpagestyle{empty} 
\end{figure}

Figure~\ref{fig:box-em} presents the box plots of the quality measures
for the full EM series. The normalized, 8~bit ground truth was the
actual input for our distortion method. Its output was the input of the
registrations. The 16~bit ground truth is the original data and is
provided as a reference. Fig.~\ref{fig:box-em} necessitated some
adjustments. There were some very low SSIM values. We adjusted the scale
of \(y\)-axis in panel~\subref{fig:box-em-a} to show more of the
relevant details around the median values and to remove some outliers.
Panel~\subref{fig:box-em-b} was plotted without any adjustments. We had
to filter the PSNR data to remove infinite values in
panel~\ref{fig:box-em-c}. Also, the quality values there for the ground
truth, 16~bit, were much higher than for other modalities. We adjusted
the scale of \(y\)-axis to show the results from the registrations more
detailed.

We see in panel~\subref{fig:box-em-a} that \enquote{GS} almost reaches
the level of the ground truth, normalized \wrt SSIM. The 16~bit ground
truth has lower SSIM values because of more details. As before,
\enquote{GS}, \enquote{Deform-SURF}, and \elastix look quite good in box
plots. The Jaccard values \subref{fig:box-em-b} were rather high,
though. The probable reason is the amount of background in the EM data
set. In both discussed measures there are some outliers on the lower
side. PSNR \subref{fig:box-em-c} shows very high values for 16~bit
ground truth, we disregard them as all other values are 8~bit.
Concerning PSNR, we see some over-registration in \enquote{GS}, less so
in \elastix and \enquote{Deform-SURF}: the median values and most of the
box contents (the box represents 50\% of the data around the median) are
\emph{higher} than in ground truth. The box plot for the normalized
ground truth shows some outliers for the larger PSNR values, however.

In the statistical evaluation (Table~\ref{tab:stat-em}), we see a
slightly larger \(p\)~value for \enquote{GS} \wrt SSIM, but nothing
extraordinary for this measure. The high values of~1.0 for SSIM
originate from the region at the end of the series with few changes
because of low amount of tissue. It rather indicates a failure of the
registration, as the measure is computed on a center crop. As mentioned
above, the Jaccard values are rather high, again, \enquote{GS} manages
to obtain a slightly higher~\(p\) for its mean. Quite of interest is
PNSR, where \enquote{Deform-SURF} manages a \(p\leq 0.02713\), but even
more spectacularly, \elastix has \(p \leq 0.7992\). This value basically
means that the mean of the PSNR for this data set, registered with
\elastix, matches the mean of the normalized ground truth with a high
probability. Such a match is an exception in our evaluations.

\begin{table}[p]
\caption{The Welch two sample $t$-test for the quality measures on EM data set.
The origin of the \enquote{locally distorted} values is as above.\newline%
We compare the means from each of the registration results to the mean for the normalized, 8~bit ground truth. The differences are statistically significant in almost all cases, we show the $p$ value for \enquote{not equal}. The hypothesis of the equal mean was almost always refuted with a high confidence. \newline%
Notice \elastix \wrt PSNR with $p < 0.7992$, it matches the mean of the normalized ground truth PSNR up to \SI[round-mode = figures, round-precision = 1]{-0.2000297}{\percent} relative error.\newline%
The mean value marked with~\textasteriskcentered{} is the one closest to the normalized ground per $t$-test. We also show the maximal value of a measure (the \enquote{Max} column). The maximum values marked with~\textdagger{} are larger than the maximum of the normalized ground truth. Notice that the quality measures are unusually high for this data set. In contrast to Fig.~\ref{fig:box-em}, we operate on unfiltered data for SSIM and Jaccard. We still had to remove infinite values of PSNR for a meaningful analysis. The sole method where data was individually filtered in the above manner is marked with~\textparagraph.\newline%
The closer is the mean to the ground truth, the better.}
\label{tab:stat-em}
\centering
\footnotesize
\subfloat[][]{%
\label{tab:stat-em-ssim}
\centering
\begin{tabular}{r S[round-mode = figures, round-precision = 3, table-format = 0.3] S[round-mode = figures, round-precision = 2, table-format = 1.1, table-figures-exponent = 2] S[round-mode = figures, round-precision = 3, table-format = 1.3]}
\multicolumn{4}{c}{EM, SSIM} \\
\toprule
{Method} & {Mean} & {$p <$} & {Max}\\
\midrule
Rigid-SIFT  & 0.9013784 & 2.2e-16   & 1.0\tabworst\\
Rigid-SURF  & 0.888223  & 2.2e-16   & 0.98863\\
Deform-SURF & 0.9480682 & 2.2e-16   & 0.99116\\
GS          & 0.9565344\tabbest
                        & 1.64e-09  & 0.99187\\
Blending    & 0.8825238 & 2.2e-16   & 0.98345\\
\elastix    & 0.9427868 & 2.2e-16   & 0.98321\\
Locally distorted
            & 0.8875335 & 2.2e-16   & 0.9933\\
Ground truth, 16~bit
            & 0.9369074 & 2.2e-16   & 0.98889\\
Ground truth, norm.
            & 0.9640380 & {--}      & 0.99689\\
\bottomrule
\end{tabular}
}

\subfloat[][]{%
\label{tab:stat-em-jaccard}
\centering
\begin{tabular}{r S[round-mode = figures, round-precision = 4, table-format = 0.4] S[round-mode = figures, round-precision = 2, table-format = 1.1, table-figures-exponent = 2] S[round-mode = figures, round-precision = 4, table-format = 1.4]}
\multicolumn{4}{c}{EM, Jaccard} \\
\toprule
{Method} & {Mean} & {$p <$} & {Max} \\
\midrule
Rigid-SIFT  & 0.960373  & 2.2e-16   & 1.0\tabworst\\
Rigid-SURF  & 0.9609596 & 2.2e-16   & 0.9990535\\
Deform-SURF & 0.9863537 & 1.217e-11 & 1.0\tabworst\\
GS          & 0.9883038\tabbest
                         & 3.194e-05 & 0.9994936\\
Blending    & 0.9530852  & 2.2e-16   & 0.9962882\\
\elastix    & 0.9860961  & 1.078e-12 & 0.9989478\\
Locally distorted
        & 0.9548106  & 2.2e-16   & 0.9995853\\
Ground truth, 16~bit
            & 0.9998352  & 2.2e-16   & 1.0\tabworst\\
Ground truth, norm.
            & 0.9901930  & {--}      & 0.999754\\
\bottomrule
\end{tabular}
}

\subfloat[][]{%
\label{tab:stat-em-psnr}
\centering
\begin{tabular}{r S[round-mode = figures, round-precision = 3, table-format = 2.1] S[round-mode = figures, round-precision = 2, table-format = 1.1, table-figures-exponent = 2] S[round-mode = figures, round-precision = 4, table-format = 2.2]}
\multicolumn{4}{c}{EM, PSNR} \\
\toprule
{Method} & {Mean} & {$p <$} & {Max}\\
\midrule
Rigid-SIFT\hspace{1pt}\tabfilter
            & 17.97261 & 2.2e-16   & 36.5545\\
Rigid-SURF  & 21.30513 & 2.2e-16   & 34.3512\\
Deform-SURF & 27.46469 & \sisetup{round-precision = 4} 0.02713
                                   & 40.1997\tabworst\\
GS          & 28.87038 & 2.2e-16   & 39.9594\\
Blending    & 19.36007 & 2.2e-16   & 31.7773\\
\elastix    & 26.92702\tabbest
                       & \sisetup{round-precision = 4} 0.7992
                           & 35.4183\\
Locally distorted
            & 20.03379 & 2.2e-16   & 37.7173\\
Ground truth, 16~bit
            & 44.94253 & 2.2e-16   & 58.5247\\
Ground truth, norm.
            & 26.98099 & {--}      & 40.1542\\
\bottomrule
\end{tabular}
}
\thisfloatpagestyle{empty} 
\end{table}

\hypertarget{ls}{%
\subsubsection{LS}\label{ls}}

\begin{figure}[p]
\centering
\subfloat[][]{\label{fig:box-ls-a}\includegraphics[width=0.9\linewidth]{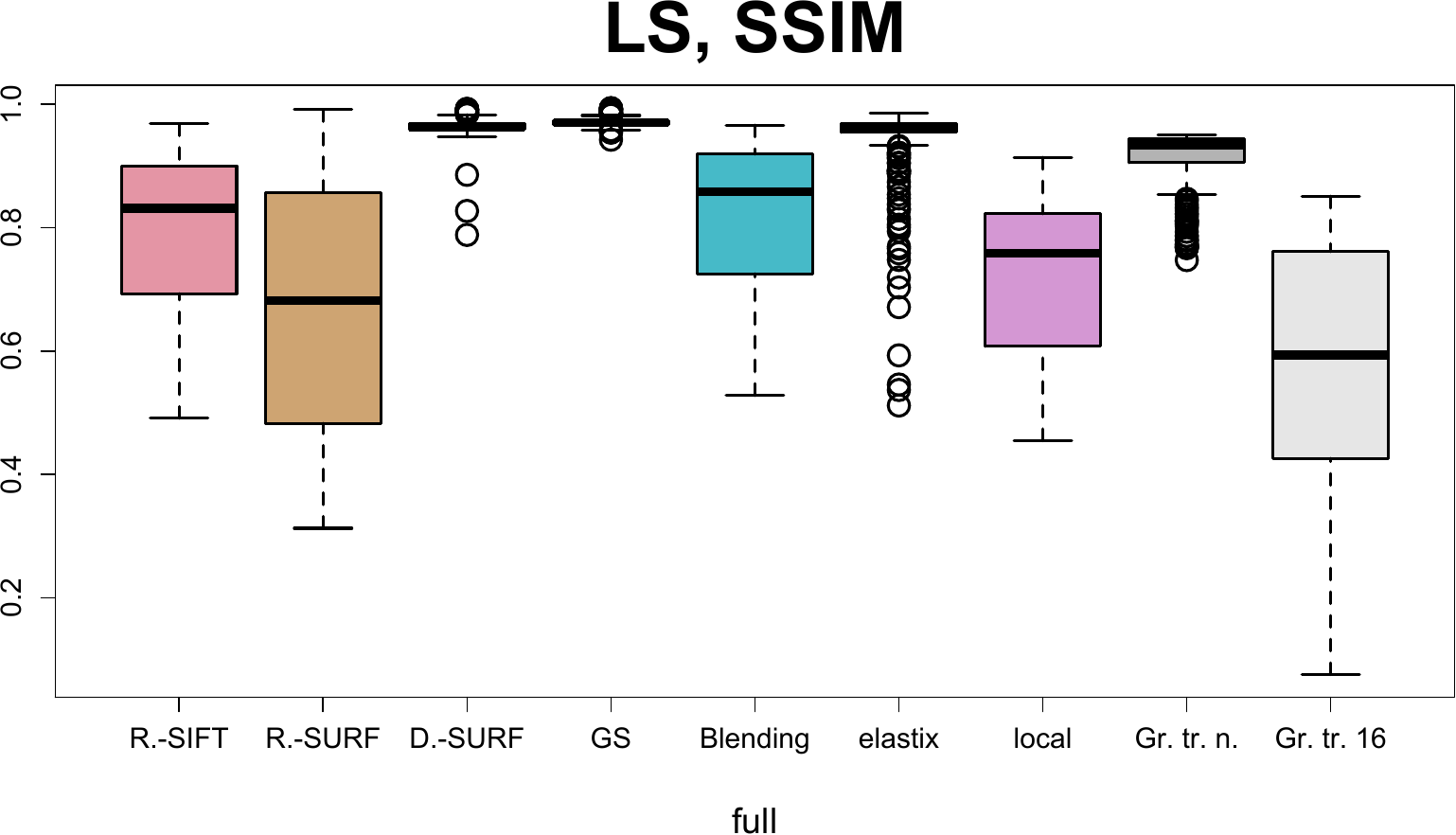}}

\subfloat[][]{\label{fig:box-ls-b}\includegraphics[width=0.9\linewidth]{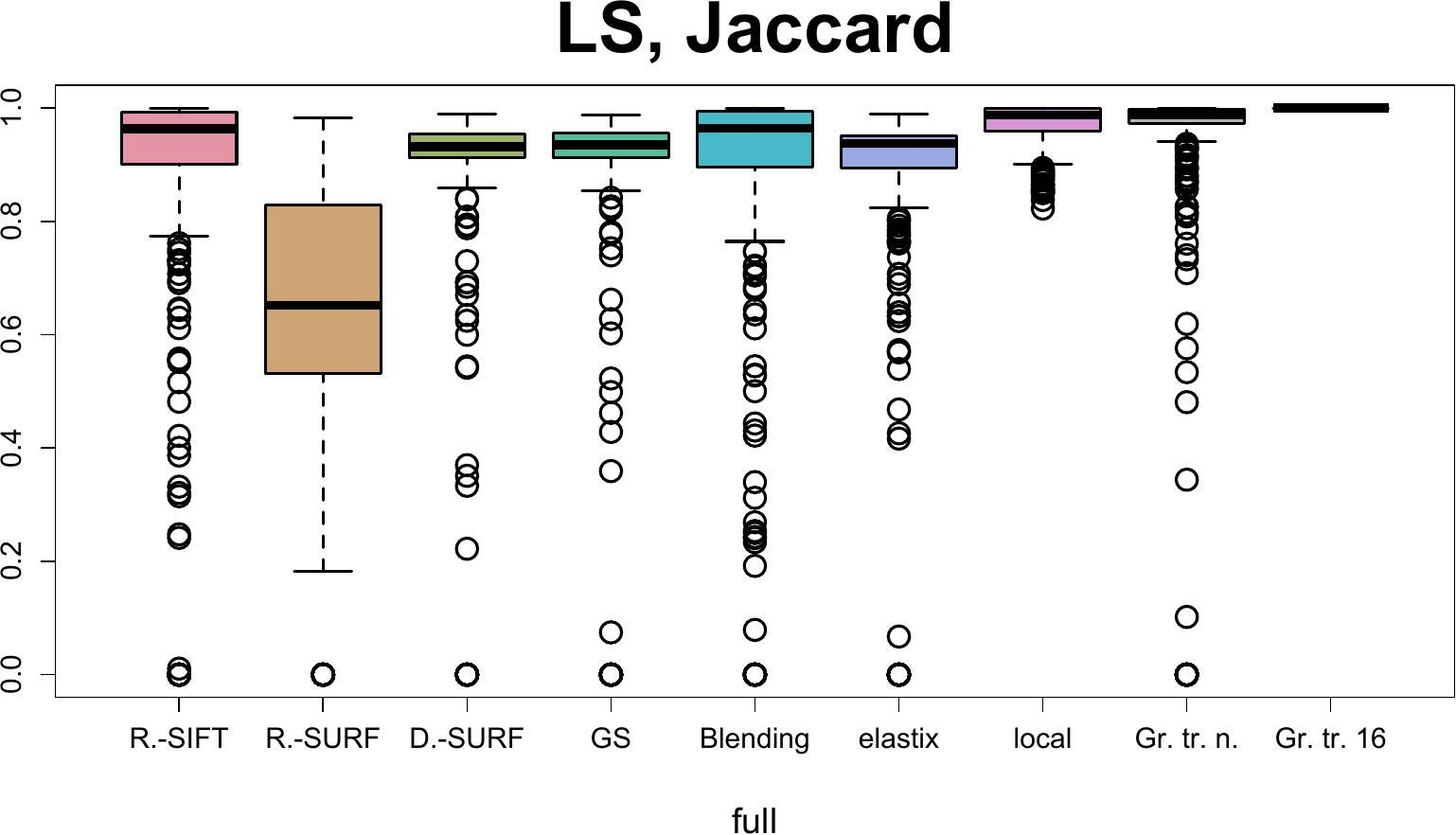}}

\subfloat[][]{\label{fig:box-ls-c}\includegraphics[width=0.9\linewidth]{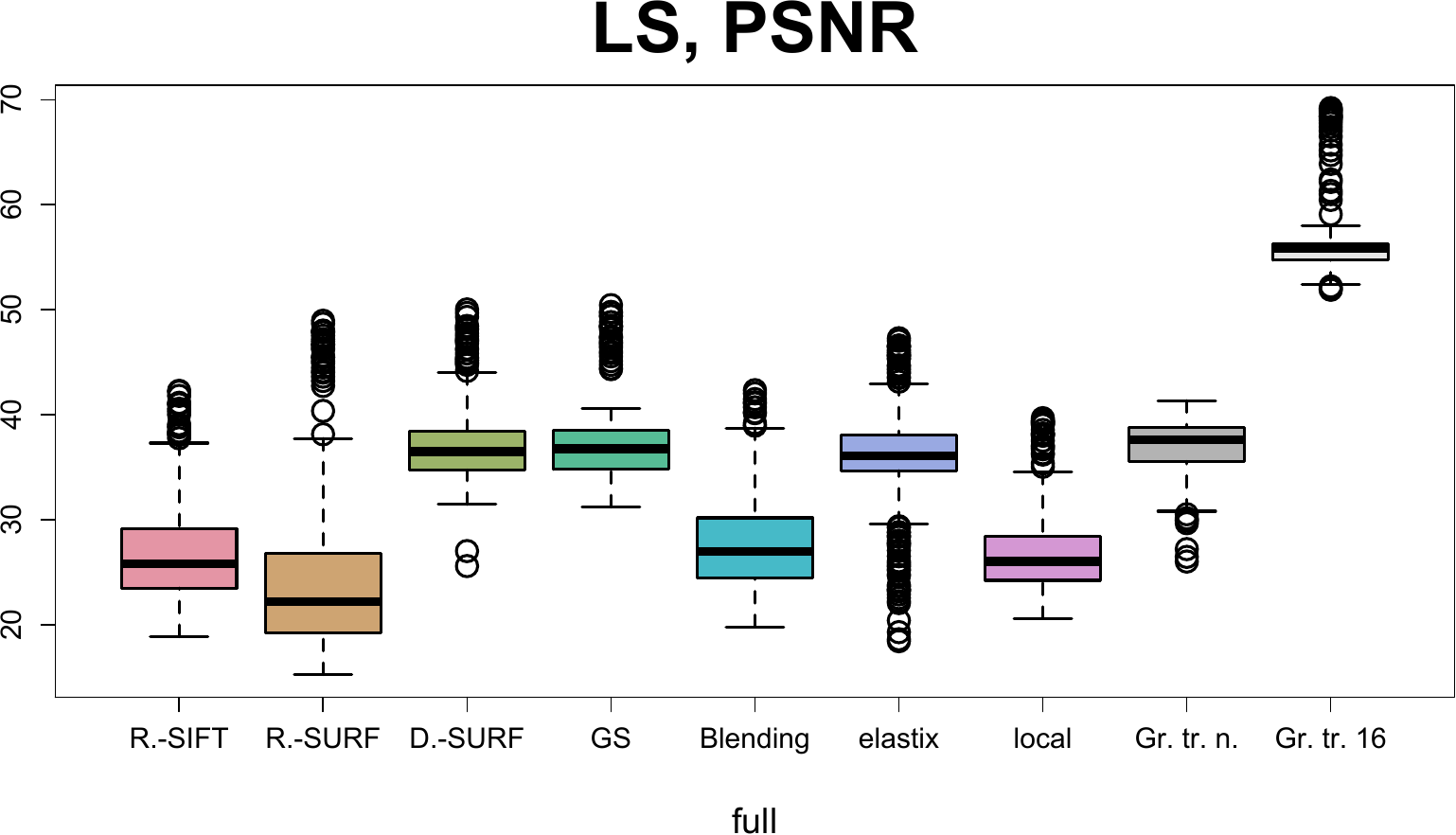}}

\caption{Box plots of quality measures for the LS data set. \enquote{R.-SIFT} stands for Rigid-SIFT, \enquote{R.-SURF} for Rigid-SURF, \enquote{D.-SURF} for Deform-SURF, \enquote{local} for only local distortions, \enquote{Gr.~tr.~n.} for \enquote{ground truth, normalized}, \enquote{Gr.~tr.~16} for \enquote{ground truth, 16~bit}. Fig.~\ref{fig:box-ls-SP} shows some further details.}
\label{fig:box-ls}
\thisfloatpagestyle{empty} 
\end{figure}

\begin{figure}[tbph]
\centering
\subfloat[][LS, SSIM, zoom in]{\label{fig:box-ls-SP-a}\includegraphics[width=0.491\linewidth]{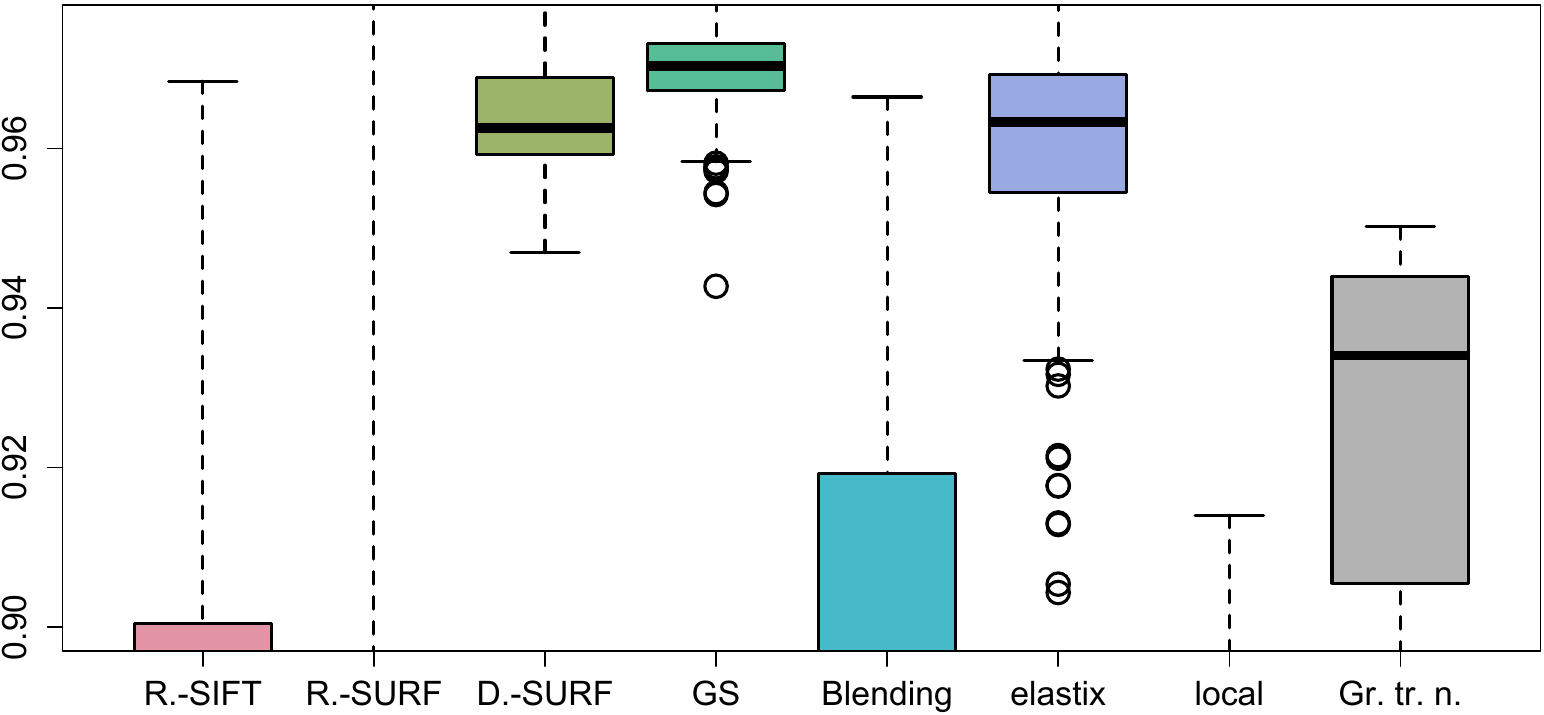}}\hfill%
\subfloat[][LS, SSIM, violin plot, bottom truncated]{\label{fig:box-ls-SP-b}\includegraphics[width=0.491\linewidth]{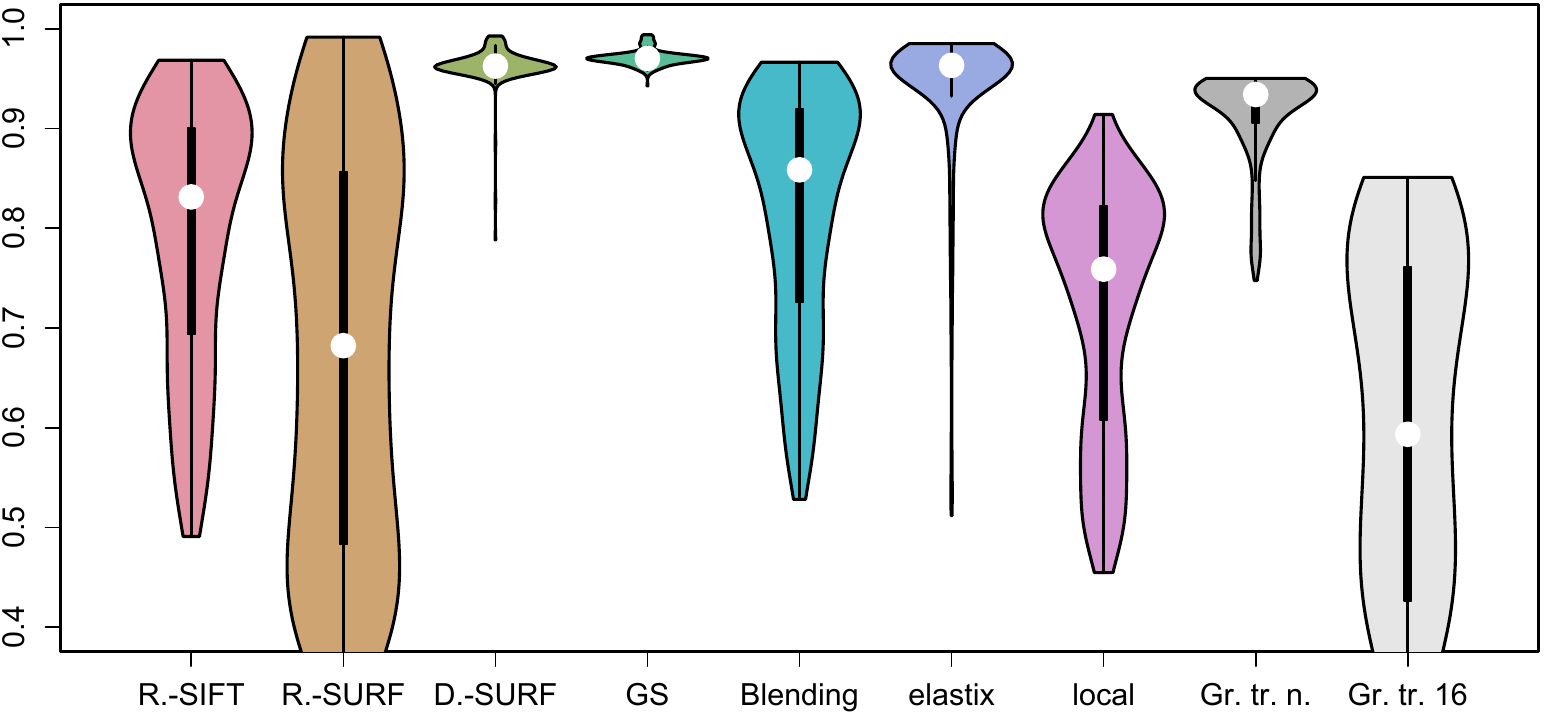}}

\subfloat[][LS, Jaccard, bottom truncated]{\label{fig:box-ls-SP-c}\includegraphics[width=0.491\linewidth]{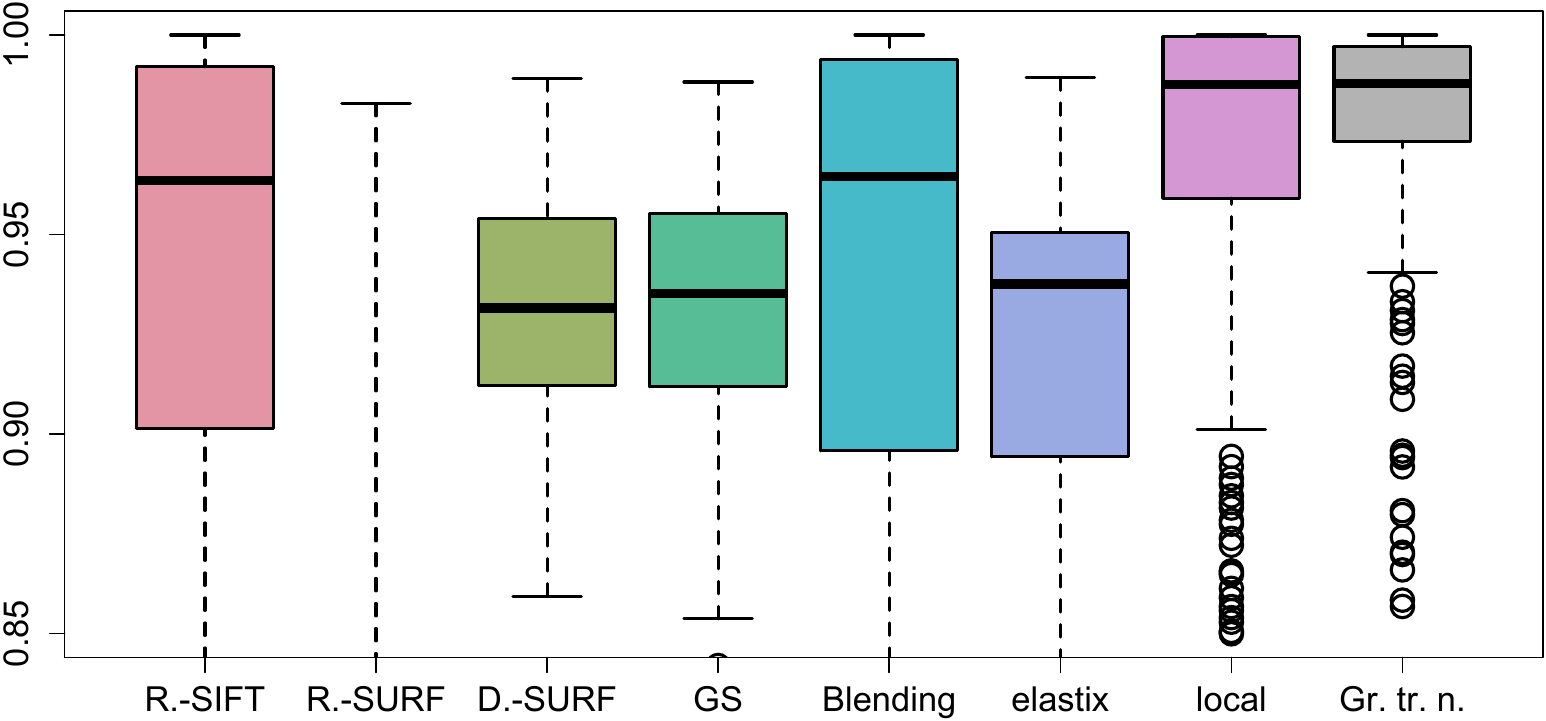}}\hfill%
\subfloat[][LS, Jaccard, filtered extreme values, bottom truncated]{\label{fig:box-ls-SP-d}\includegraphics[width=0.491\linewidth]{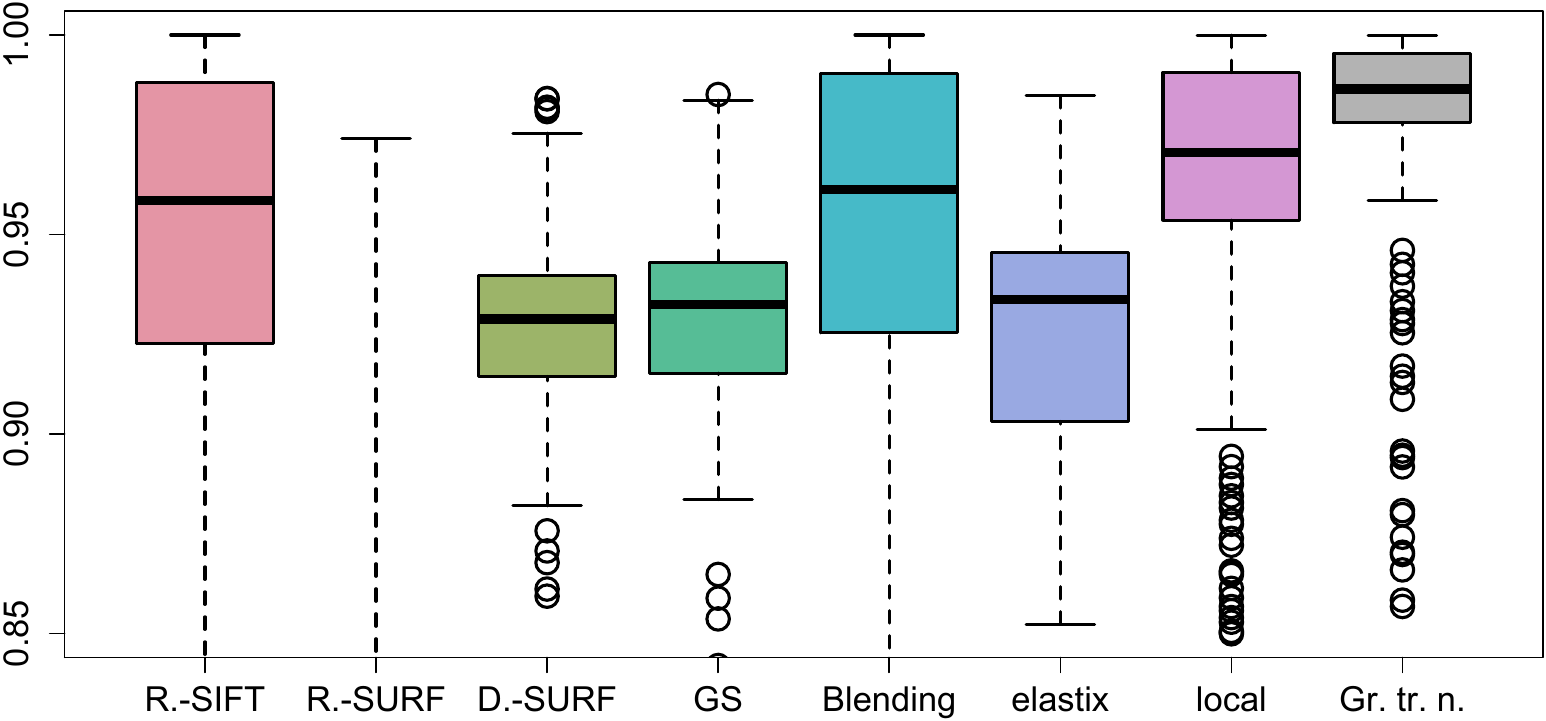}}

\subfloat[][LS, PSNR, top and bottom truncated]{\label{fig:box-ls-SP-e}\includegraphics[width=0.491\linewidth]{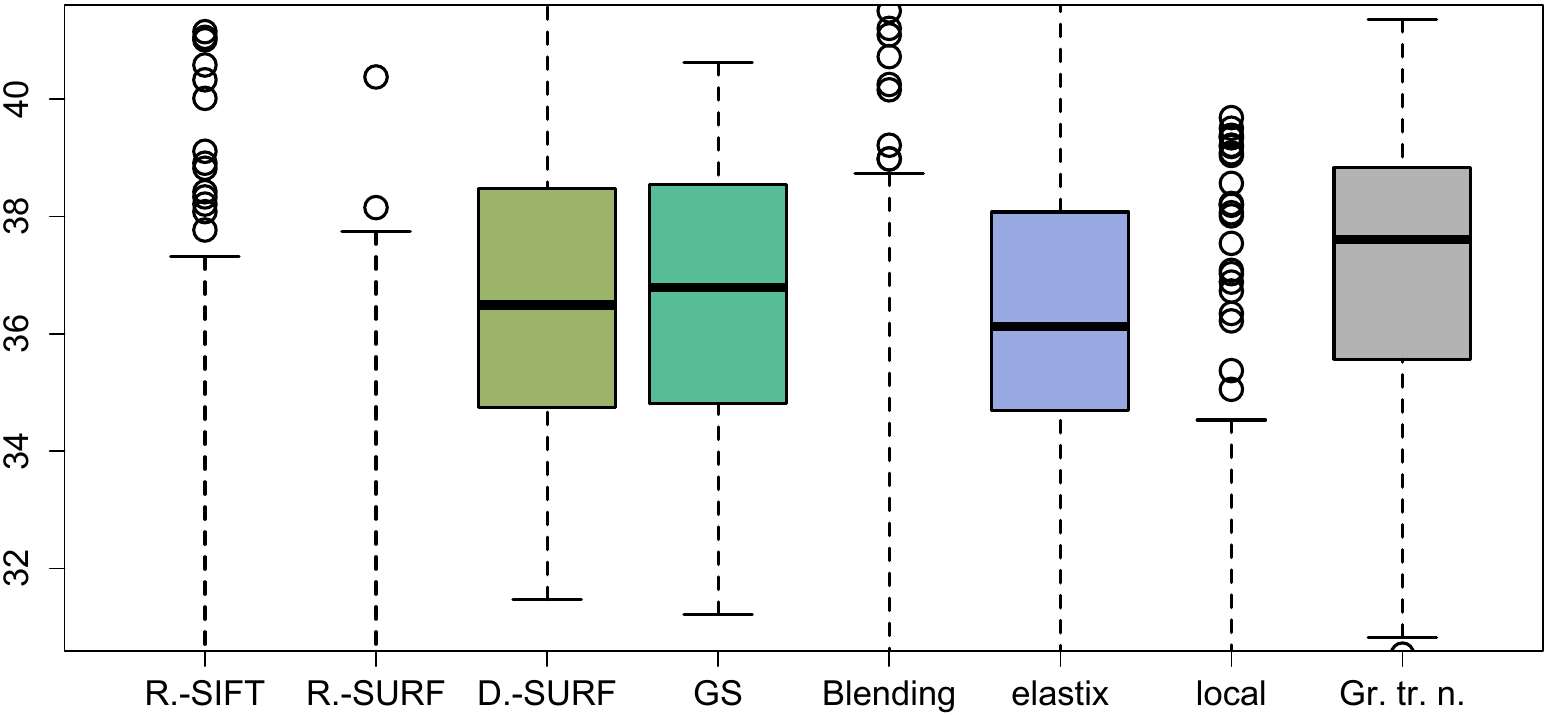}}\hfill%
\subfloat[][LS, PSNR, full violin plot]{\label{fig:box-ls-SP-f}\includegraphics[width=0.491\linewidth]{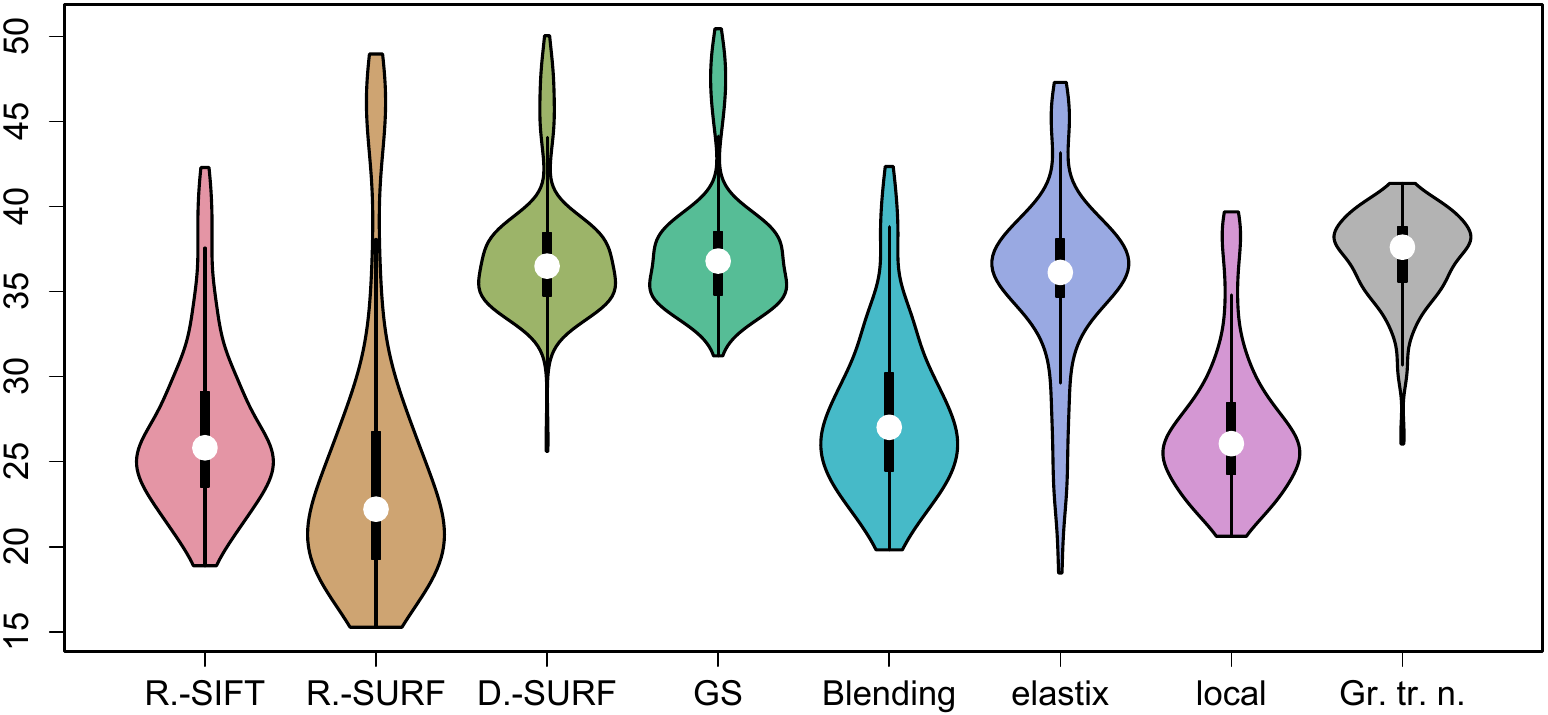}}

\caption{More detailed plots of LS quality measures. Abbreviations are same as above. \enquote{Truncated} means that not the full range of the measure was plotted. In Jaccard, \enquote{filtered extreme values} means that numerous zero and 1.0 values were removed. They can be attributed to missing tissue or completely full region, were no meaningful comparison can be made at current threshold. We also show some violin plots where those might bring additional insight through their shape.\newline%
Concerning the \enquote{Rigid-SURF} method, its quality was much lower. As this figure aims to provide a more detailed view, we scale the box plots to better visualize the differences between other methods. Fig.~\protect\ref{fig:box-ls} as well as the violin plot~\protect\subref{fig:box-ls-SP-f} show the full picture.\newline%
As before, \enquote{R.-SIFT} stands for Rigid-SIFT, \enquote{R.-SURF} for Rigid-SURF, \enquote{D.-SURF} for Deform-SURF, \enquote{local} for only local distortions, \enquote{Gr.~tr.~n.} for \enquote{ground truth, normalized}, \enquote{Gr.~tr.~16} for \enquote{ground truth, 16~bit}.}
\label{fig:box-ls-SP}
\end{figure}

\begin{table}[ptb]
\caption{The Welch two sample $t$-test for the quality measures on LS data set.
The \enquote{locally distorted} values are as above. \newline%
We compare the means from each of the registration results to the mean for the normalized, 8~bit ground truth. The differences are statistically significant in almost all cases, we show the $p$ value for \enquote{not equal}. The hypothesis of the equal mean was almost always refuted with a high confidence.\newline%
The maximum of the normalized ground truth for SSIM was rather low. For Jaccard, \enquote{Deform-SURF} reaches $p < 0.168$, and for PSNR the same method reaches $p < 0.7875$. For the latter, it matches the mean of the ground truth up to \SI{0.1865034}{\percent}, the \SI{95}{\percent} confidence interval is \numrange{-0.4345352}{0.5728342}. 
The mean value marked with~\textasteriskcentered{} is the one closest to the normalized ground per $t$-test. We also show the maximal value of a measure (the \enquote{Max} column). The maximum values marked with~\textdagger{} are larger than the maximum of the normalized ground truth. Notice that the quality measures are unusually high for this data set. Similar to Fig.~\ref{fig:box-ls}, we operate on unfiltered data for SSIM and Jaccard. The method marked with \textparagraph{} has no good values. 
The closer is the mean to the ground truth, the better.}
\label{tab:stat-ls}
\centering
\footnotesize
\subfloat[][]{%
\label{tab:stat-ls-ssim}
\centering
\begin{tabular}{r S[round-mode = figures, round-precision = 3, table-format = 0.3] S[round-mode = figures, round-precision = 2, table-format = 1.1, table-figures-exponent = 2] S[round-mode = figures, round-precision = 3, table-format = 1.3]}
\multicolumn{4}{c}{LS, SSIM} \\
\toprule
{Method} & {Mean} & {$p <$} & {Max}\\
\midrule
Rigid-SIFT  & 0.7962219 & 2.2e-16   & 0.96844\tabworst\\
Rigid-SURF  & 0.6782059 & 2.2e-16   & 0.99155\tabworst\\
Deform-SURF & 0.9640279 & 2.2e-16   & 0.99274\tabworst\\
GS          & 0.9710383 & 2.2e-16   & 0.99396\tabworst\\
Blending    & 0.8189065 & 2.2e-16   & 0.96644\tabworst\\
\elastix    & 0.9449100\tabbest
                        & 1.03e-10  & 0.98512\tabworst\\
Locally distorted
            & 0.7205307 & 2.2e-16   & 0.91403\\
Ground truth, 16~bit
            & 0.5715045 & 2.2e-16   & 0.85105\\
Ground truth, norm.
            & 0.9145253 & {--}      & 0.95019\\
\bottomrule
\end{tabular}
}

\subfloat[][]{%
\label{tab:stat-ls-jaccard}
\centering
\begin{tabular}{r S[round-mode = figures, round-precision = 4, table-format = 0.4] S[round-mode = figures, round-precision = 2, table-format = 1.1, table-figures-exponent = 2] S[round-mode = figures, round-precision = 4, table-format = 1.4]}
\multicolumn{4}{c}{LS, Jaccard} \\
\toprule
{Method} & {Mean} & {$p <$} & {Max} \\
\midrule
Rigid-SIFT  & 0.839902  & \sisetup{round-precision = 4} 0.09731
                        & 1.0\\
Rigid-SURF  & 0.6347376 & 2.2e-16   & 1.0\\
Deform-SURF & 0.848737\tabbest
                        & \sisetup{round-precision = 4} 0.168
                        & 0.9890468\\
GS          & 0.8443777 & \sisetup{round-precision = 4} 0.1245
                            & 0.9882287\\
Blending    & 0.8353499 & \sisetup{round-precision = 4} 0.06807
                            & 1.0\\
\elastix    & 0.8336558 & \sisetup{round-precision = 4} 0.0454
                           & 0.9892638\\
Locally distorted
        & 0.9699586 & 1.279e-07 & 1.0\\
Ground truth, 16~bit\hspace{1pt}\tabfilter
            & 1.000000  & 2.2e-16   & 1.0\\
Ground truth, norm.
            & 0.879492  & {--}      & 1.0\\
\bottomrule
\end{tabular}
}

\subfloat[][]{%
\label{tab:stat-ls-psnr}
\centering
\begin{tabular}{r S[round-mode = figures, round-precision = 3, table-format = 2.1] S[round-mode = figures, round-precision = 2, table-format = 1.1, table-figures-exponent = 2] S[round-mode = figures, round-precision = 4, table-format = 2.2]}
\multicolumn{4}{c}{LS, PSNR} \\
\toprule
{Method} & {Mean} & {$p <$} & {Max}\\
\midrule
Rigid-SIFT  & 26.86285 & 2.2e-16   & 42.2897\tabworst\\
Rigid-SURF  & 24.59035 & 2.2e-16   & 48.9622\tabworst\\
Deform-SURF & 37.14596 & \sisetup{round-precision = 4} 0.7875
                                   & 50.0373\tabworst\\
GS          & 37.33850 & \sisetup{round-precision = 4} 0.3101
                                   & 50.4454\tabworst\\
Blending    & 27.83801 & 2.2e-16   & 42.3521\tabworst\\
\elastix    & 35.72511 & 5.756e-05 & 47.2961\tabworst\\
Locally distorted
            & 26.98225 & 2.2e-16   & 39.6856\\
Ground truth, 16~bit
            & 56.26130 & 2.2e-16   & 69.2006\tabworst\\
Ground truth, norm.
            & 26.98099 & {--}      & 41.3583\\
\bottomrule
\end{tabular}
}
\thisfloatpagestyle{empty} 
\end{table}

Consider Fig.~\ref{fig:box-ls}. The SSIM for LS method shows quite good
values for \enquote{Deform-SURF}, \enquote{GS}, and also for \elastix,
but in this case with some outliers. Surprisingly, the plot for the
normalized ground truth is less convincing, the original, 16~bit ground
truth shows even less similarity. The variance is clearly much larger in
the 16~bit data. We can thus conclude, that above \enquote{good}
registration method over-register. Looking into
Fig.~\ref{fig:box-ls-SP-a}, we concludes that \enquote{GS},
\enquote{Deform-SURF}, and less so, \elastix, produce much higher SSIM
values than they should have in order to be similar with the normalized
ground truth. The variance in the results of those registration methods
is also lower than in the normalized ground truth.
Panel~\subref{fig:box-ls-SP-b} shows in a violin plot how the
distribution of SSIM values changed between the methods.

We see quite high values for Jaccard measure in Fig.~\ref{fig:box-ls-b},
but also a lot of outliers in almost all methods. To study those
further, we present a zoomed-in version in Fig.~\ref{fig:box-ls-SP-c}.
Even more interesting is panel~\ref{fig:box-ls-SP-d}. There we have
removed the values \num{0} and \num{1.0} from the evaluation. Basically,
those extreme Jaccard values mean that either no correspondence at all
was found or the full correspondence. The latter can be caused by too
low threshold value or too little detail in the particular region. In
those both panels we see a superior performance of the
\enquote{Blending} method compared to all other registrations. We notice
also that our local distortions do not change the Jaccard index very
much. The statistical analysis (below) does not support the superiority
of \enquote{Blending}, however.

As also in other modalities, the PSNR value of the 16~bit ground truth
is much larger, than in all other methods that utilize 8~bit images
(Fig.~\ref{fig:box-ls-c}). In a zoomed-in version in
Fig.~\ref{fig:box-ls-SP}, \subref{fig:box-ls-SP-e} we see that for PSNR
the median of no registration method exceeds the median of the ground
truth. This means that PSNR detects no over-registration in this case.
The somewhat peculiar, uneven shapes of the PSNR distributions are
visualized as violin plots in panel~\subref{fig:box-ls-SP-f}.

Now, consider Table~\ref{tab:stat-ls}. Statistically, no registration
method matches the mean of the SSIM of normalized ground truth well.
Most of the methods (namely, \enquote{Deform-SURF}, \enquote{GS},
\elastix) are well above and also all registrations overshoot the
maximum of the ground truth SSIM. In Jaccard, we have many 1.0 values
(which were not removed in this case, as we want to contrast those
statistics to the box plots). We see a match in the means with
\(p<0.168\) in \enquote{Deform-SURF}, however we would be quite cautions
in this case because of some \enquote{invalid}, too low or too high
Jaccard values. With PSNR, \enquote{Deform-SURF} manages to match the
mean with \(p<0.7875\).

Some methods (\eg \enquote{GS}, \elastix) have even larger mean values
of PSNR. We would deem those methods as better than
\enquote{Deform-SURF} in this case, if we did not have the ground truth.
We have to note, however, that \emph{all} registrations overshoot the
maximum of the PSNR in the ground truth, as evident from
Table~\ref{tab:stat-ls-psnr}. A~repeated look on
Fig.~\ref{fig:box-ls-SP}, panels~\subref{fig:box-ls-SP-e},
\subref{fig:box-ls-SP-f} shows that the distributions of the PSNR values
for \enquote{GS} and \enquote{Deform-SURF} are much more similar to each
other than to that of the normalized ground truth.

\hypertarget{discussion}{%
\section{Discussion}\label{discussion}}

\hypertarget{the-results-and-the-banana-problem}{%
\subsection{\texorpdfstring{The results and the \enquote{banana
problem}}{The results and the ``banana problem''}}\label{the-results-and-the-banana-problem}}

We have mostly discussed the results in the previous section, still
there is an issue we would like to specially highlight. We have quite
often seen that methods which produce better image-based metrics also
seem to over-register the series. It would be very hard to find such an
over-registration (a \enquote{banana problem}) without a ground truth.
CT scans of the specimens before sectioning might help, but, as
mentioned above, they lack on the resolution. Basically, our ground
truth bounds from above the amount of correspondence between consecutive
images. Such challenges as ours would help to develop better
registrations that try to reach such a bound, but not to overstep it.

Occasionally, we have found the values of our quality measures for a
particular registration method \emph{higher} than for the ground truth.
How is this possible? Our reasoning is that the ground truth does not
constitute a \emph{perfect} correspondence of the consecutive input
images. It is merely their real correspondence drawn from inherently 3D
data. This means that those registrations might create \emph{too much}
correspondence, they over-register their inputs. We have also seen an
interesting effect, where the correspondences after a registration where
more \emph{heterogeneous} than in ground truth. It appears to us that
some areas were over-registered and some areas were under-registered.
Again, without a ground truth such observations would be impossible.

In other words, to obtain good values overall or on average (while high
variance cannot be reduced), some registrations seem to attempt to shift
all of the correspondences between the images towards their maximums.
Naturally, this behavior forces the maximal values to exceed the maximum
of the ground truth while the mean is still under the mean of the ground
truth. This leads to the \enquote{banana problem}.

\hypertarget{distortion-mechanism}{%
\subsection{Distortion mechanism}\label{distortion-mechanism}}

Our distortions \enquote{stretch} and \enquote{shrink} the images, but
their positions are organized in a rectilinear grid, even if the
distortions themselves are random and Rayleigh-distributed. It would
also make sense to choose the distortion positions randomly, too.
However, we opted for a grid for a better reproducibility of the
appearance of the test images: a human would know where to look. Still,
the actual distorted images do not show that much \enquote{bad}
regularity, the above grid is not visible, so we argue that no problems
arise from such a rectilinear grid placement. If our method is used to
benchmark registrations using neural networks with a direct assignment
of neurons to either pixels or distortions, the randomization of the
distortion positions might be required.

We use a rigid transformation to model the inaccuracy in section
placement. If a fully affine or an even more generic transform is
needed, our code can be easily extended to incorporate it. Indeed, some
registration methods
\citep[\egX ][]{cardona_trakem2_2012, xu_method_2015} use affine global
transformations to model wedge-shaped sections.

To contrast phantoms to our approach: the argument on not fully
representing the distortions might also hold for our distortion
generation, but we still work with real data. Hence, the reasoning on
lacking data complexity does not hold. This issue is especially
prominent in methods based on feature detection.

\hypertarget{further-ideas-for-the-distortion-modeling}{%
\subsection{Further ideas for the distortion
modeling}\label{further-ideas-for-the-distortion-modeling}}

Our method currently does not directly account for tissue folding and
tearing. Such damaged areas can be represented with \enquote{holes} in
the images, but they currently would not correlate with larger
distortions in the connected areas. A~realistic modeling of section
damage was not our current goal. Nowadays, methods to bridge section
damage exist \citep{lobachev_tempest_2020}. Basically, any kinds of
damage can be assessed with masks for the repair, similar to the masks
we use here to simulate the damage. Some further recent works assess
cracks and discontinuities
\citep{aggrawal_image_2020, ng_unsupervised_2020}.

A~convolutional neural network, transforming \enquote{clean} images into
damaged ones with some kind of a style transfer
\citep{gatys_image_2016, shaban_staingan:_2019, liang_stain_2020, khan_generalizing_2020}
is an interesting idea. We sought for a functional and well-defined
image deformation that allowed for using transformed images as a
benchmark input. Neural-network-generated images might have some
unnoticeable for humans drawbacks that would obscure and throw off-track
some other (probably, also deep-learning-based)
registrations---detection of adversarial examples is a separate problem.
Our test images are produced through simple, robust, reproducible, and
well-understood image transformations.

\hypertarget{impact-of-a-3d-series}{%
\subsection{Impact of a 3D series}\label{impact-of-a-3d-series}}

We use full-blown, inherently 3D images as a series of 2D images. Those
2D inputs are used for our benchmark for a reason. Some registrations do
not operate on image pairs, but optimize the spatial positions of the
full image stack at one.

Next, the presence of already three-dimensional images as the starting
point enables us to state how the final image stack should look like. We
digitally simulate the distortions by sectioning and further section
handling. Then we apply a registration method (we evaluate multiple of
them in this work). The discrepancy between the distorted series is
larger, than in the registered series; this is the whole idea of
registration. But, contrary to the usual 2D registrations of serial
sections, we still have the initial starting point, the 3D images. They
are in the same resolution as the registered series. We call those
initial images the ground truth. We can compare the registered series to
the ground truth and find out, what was wrong with the registration
method in question.

\hypertarget{challenges-in-project-execution}{%
\subsection{Challenges in project
execution}\label{challenges-in-project-execution}}

It was quite hard for some methods to cope with large angles in rigid
transformations, in those cases we used SURF-based rigid registrations
as an initial phase. Many registrations are inherently trimmed for the
most used input data kinds and modalities. Adaptation to further images
is possible, but requires more or less tuning. In the best case, the
tuning can be commenced with parameter files, such as with \elastix.

The present project was quite large. A~decent automation of the workflow
(we used Python, bash, and GNU Make) was key for fast and error-free
processing. This issue was of especial importance in case of the
evaluations.

\hypertarget{evaluations}{%
\subsection{Evaluations}\label{evaluations}}

The different normalization issues, esp.~in EM data set, might also
explain the observed variations in the measurements. This is a typical
trade-off: a better normalization allows for better registration, but a
normalization also changes the data set, so a direct comparison with
non-normalized data might be harder.

The visualizations of optical flow we used as one of the measures is, on
the one hand, a valuable tool. Those visualizations show issues less
visible otherwise, the \enquote{hot spots}. On the other hand, a direct
comparison of such visualizations with each other in their present form
might be misleading because of individual magnitudes.

One of the ideas for further improvement of our work is to compare not
images, but deformation fields from various methods. However, multiple
implementation questions would arise. One of the issue is the
registration \enquote{drift} that would be different in various methods:
Currently, our challenge is complete open: the participants need to
obtain the input images and produce the result images, the registration
method itself does not need to be adapted or changed, it can remain
closed-source or even a commercial secret. If we would like to compare
the deformation fields, we would need to provide a consistent way to
output comparable distortions across all the implementations, libraries,
programming languages the participants use. The code for the actual
method needs to be changed, which means it should be available and human
resources for the change need to be allocated.

\hypertarget{continuous-integration}{%
\subsection{Continuous integration}\label{continuous-integration}}

\label{sec:ci}

Our visualizations and measures are computed automatically. No human
interaction what so ever is needed: the distortions of the ground truth,
registrations, and evaluations can happen fully automatically. Thus, our
evaluation method is a gateway to wide-scale registration challenges and
to regression testing of further developing registration methods. Our
method can be applied as a part of a continuous integration workflow
\citep{meyer_continuous_2014}. With our approach, better and more
thorough regression testing of registration becomes possible.

Basically, this paper shows a further path towards automatically testing
different regularizations in image registrations. The goal would be to
reduce the magnitude of the banana problem, while still maintaining good
registration results. Such testing can be done with image-based metrics,
as we do here, but any other metric would work too.

\hypertarget{conclusions}{%
\section{Conclusions}\label{conclusions}}

We introduce an approach to generate individual 2D distortions applied
to existing 3D medical data. Those distortions have the basic statistic
properties of the cutting-induced variations in serial sections. The
distortions are computed in a straightforward, understandable, and
reproducible manner. We also apply a global rigid transform to mimic the
inexact placement of a section on glass slide. Modeling damaged sections
is also possible. Combined, we can simulate the transition from a tissue
block to a set of serial sections. Thus, we are able to test
registration methods on such simulated data originating from real
tissues.

We provide an overview of the existing registration and evaluation
efforts. To our knowledge, the approach, we pursue in this work, has not
been undertaken previously.

The key contribution of this work is the utilization of the original,
undistorted data for the evaluation. Previously, it was impossible to
evaluate registrations of serial sections with ground truth, as the
tissue block is destroyed by sectioning. Micro-CT scans currently do not
have sufficient resolution and can serve only as a coarse guide in a
co-registration of micro-CT to real serial sections. Phantoms and
synthetic images might not have the needed complexity. We use real,
micrometer-scale lung images from other modalities in this work. By
using real images of animal lungs we affirm that the kind of the images
used is comparable to real serial sections of the same tissue. Our
method is, however, directly applicable to other organs or generic
images (Figs.~\ref{fig:teaser}, \ref{fig:ship}).

In this work we evaluate six registration methods on three distorted
data sets. In each of them, a ground truth is present. With the ground
truth, we can not only compare the registrations with each other, but
also with the inherent 3D data, in other words: with original data, with
how the 2D \enquote{slices} should have been aligned if no cutting took
place. We address the quality of registrations with four visualizations
of image metrics and three image-based quality measures. In this work,
we both look at a specific image pair (in the supplementary material)
and provide an evaluation over the full range of the series. Our method
can be applied in a continuous integration workflow.

We make the source code and the data sets publicly available. Further
contributions, both in form of additional registration results and
further data sets, are welcome.

\hypertarget{future-work}{%
\subsection{Future work}\label{future-work}}

Utilization of our method, statistical analysis of real sections
\citep[similar to][]{schormann_statistics_1995}, and an introduction of
better quality measures may lead to better registrations, both utilizing
deep learning and not. We definitely look forward to more comparisons of
registration methods.

It would be very interesting to find a way to compare the registration
result to ground truth directly. (In this work we compare consecutive
images from each of the results and evaluate the resulting measures.)
Presently, the accumulation of registration \enquote{drift} and some
global offsets make a well-founded assessment more complicated than our
present evaluation.

\hypertarget{acknowledgments}{%
\section*{Acknowledgments}\label{acknowledgments}}
\addcontentsline{toc}{section}{Acknowledgments}

We thank Adrian V.~Dalca (MIT, MA, USA), Markus Wedekind (Technische
Universität Braunschweig, Germany), and Birte S.~Steiniger
(Philipps-University of Marburg, Germany) for helpful discussions.
Susanne Kuhlmann and Susanne Fa\ss{}bender (MHH, Hannover, Germany)
provided excellent technical support. Anja Bubke (MHH, Hannover,
Germany) was involved in LS acquisition. A~GPU (Quadro~P5000) used for
this research was donated by the NVIDIA Corporation. We used the SSIM
implementation by Zhou Wang. We modified the code by Jean Francois
Pambrun for Dice measure visualizations.

This work was supported by DFG grant MU~3118/8-1. This work was
partially supported by JST, PRESTO grant number JPMJPR2025, Japan. This
research was supported by a C2 grant from KU~Leuven (C24/18/101) and a
research grant from the Research Foundation~-- Flanders (FWO~G0C4419N).
None of the funding bodies was involved in the design or execution of
the study.


{
  \footnotesize
  %
  \bibliographystyle{model2-names.bst}
  \ifthenelse{\boolean{arxivpreprint}}{}{%
    \biboptions{authoryear}
  }
  \bibliography{full}
}




\cleardoublepage
\appendix
\renewcommand\thefigure{\arabic{figure}}  

\begin{figure}[!t]
\centering
\includegraphics[width=\linewidth]{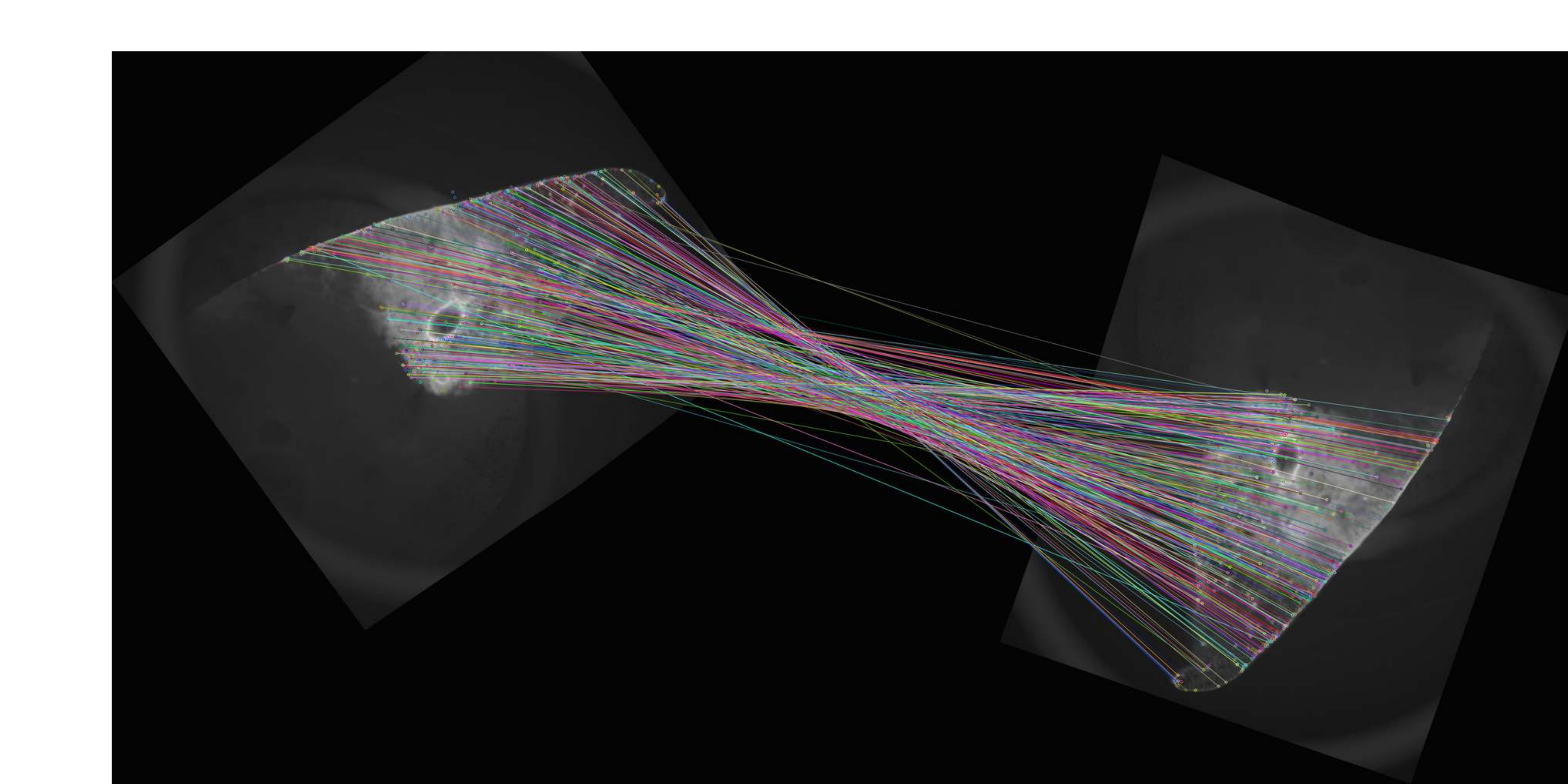}

\caption{Feature matches for affine transform estimation.}
\label{fig:ransac}
\end{figure}

\hypertarget{feature-matching}{%
\section{Feature matching}\label{feature-matching}}

Figure~\ref{fig:ransac} shows selected matches from our rigid
feature-based registration with SURF
\citep{lobachev_feature-based_2017}. The input images are globally
transformed and locally distorted consecutive images from LS data set
\citep[see also][ for details on the biological
acquisition]{krischer_mechanical_2021}.

\hypertarget{details-of-the-em-acquisition}{%
\section{Details of the EM
acquisition}\label{details-of-the-em-acquisition}}

Our EM data set was acquired with the SBF-SEM technique
\citep{buchacker_assessment_2019}. Notice that although cutting with
diamond knife might induce some non-linear distortions in the block
face, in a typical SBF-SEM acquisition in variable pressure mode at most
the translations are to be corrected in post-processing. This alignment
is often required due to movement of the whole block face in the field
of view. Non-linear distortions are minimized in SBF-SEM in contrast to,
\eg serial sectioning TEM or array tomography. In SBF-SEM no distortions
or foldings, but also no rotations are to be expected. During the
acquisition of the data, which we use as an input in this work,
\emph{no} rotation or non-rigid alignment, but also specifically
\emph{no translation} was applied.

\hypertarget{evaluation-measures}{%
\section{Evaluation measures}\label{evaluation-measures}}

\hypertarget{full-size-evaluation}{%
\subsection{Full-size evaluation}\label{full-size-evaluation}}

\begin{figure}[t!]
\centering%
\subfloat[][]{\label{fig:funatomi-full-a}\includegraphics[width=0.495\linewidth]{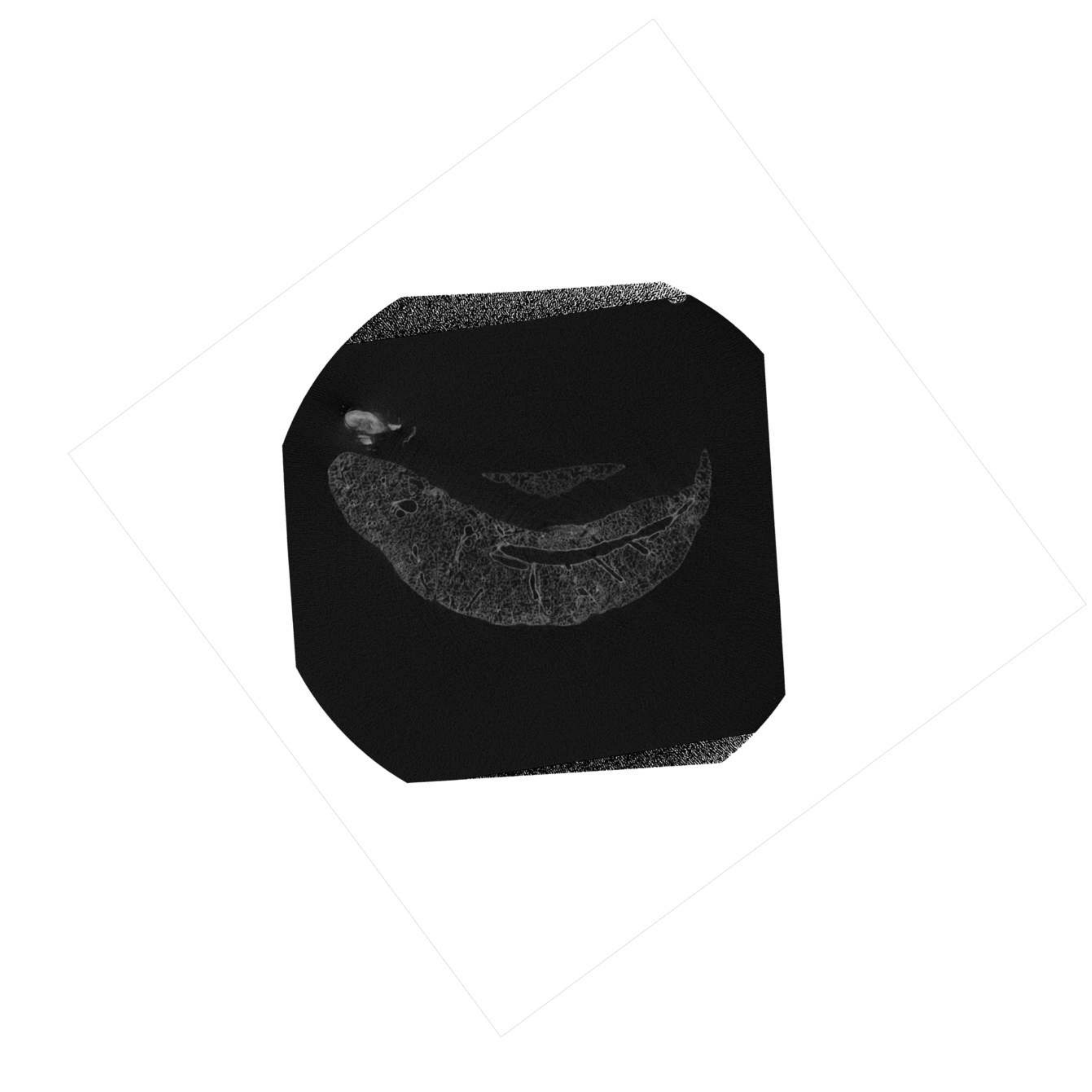}}\hfill%
\subfloat[][]{\label{fig:funatomi-full-b}\includegraphics[width=0.495\linewidth]{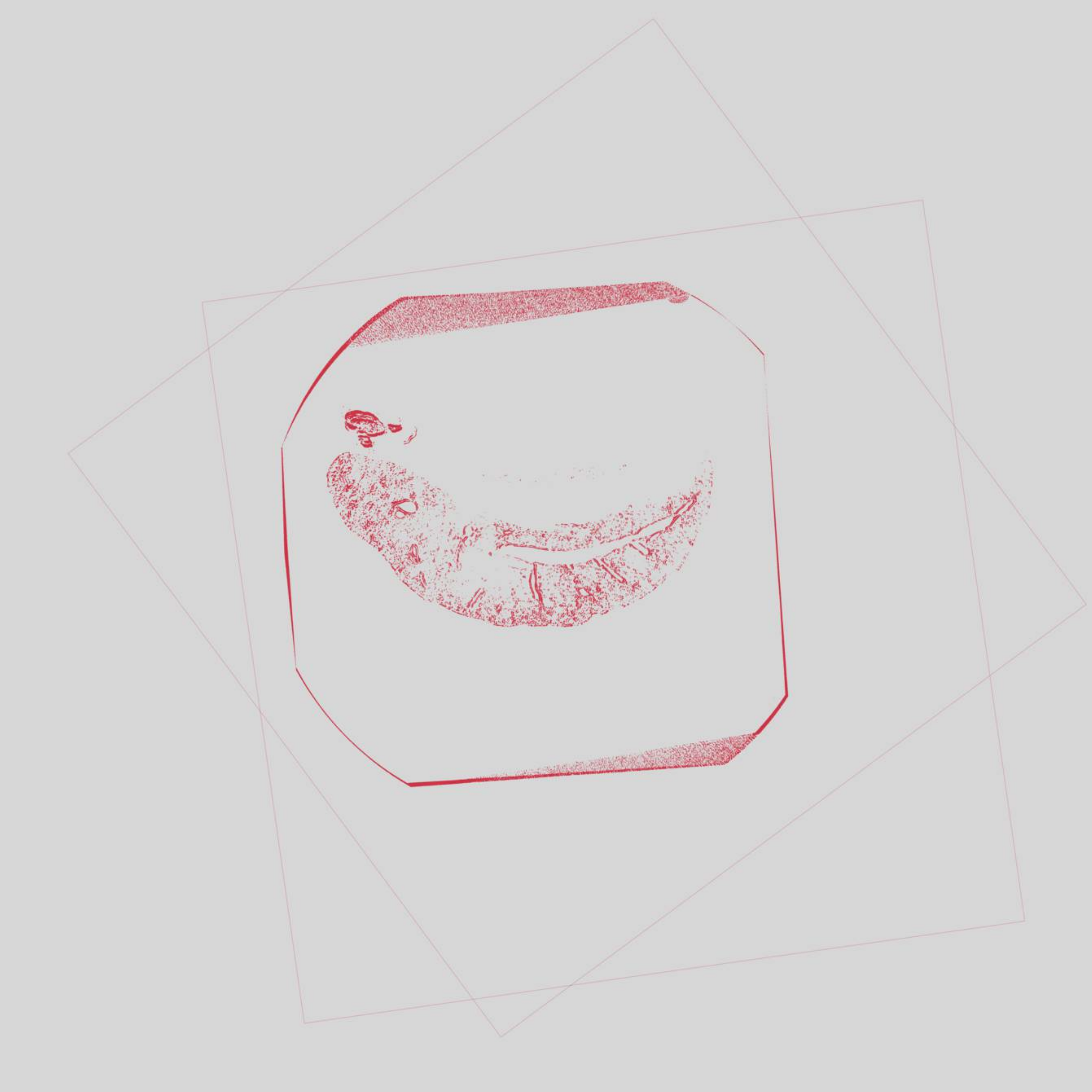}}\hfill%
\subfloat[][]{\label{fig:funatomi-full-c}\includegraphics[width=0.495\linewidth]{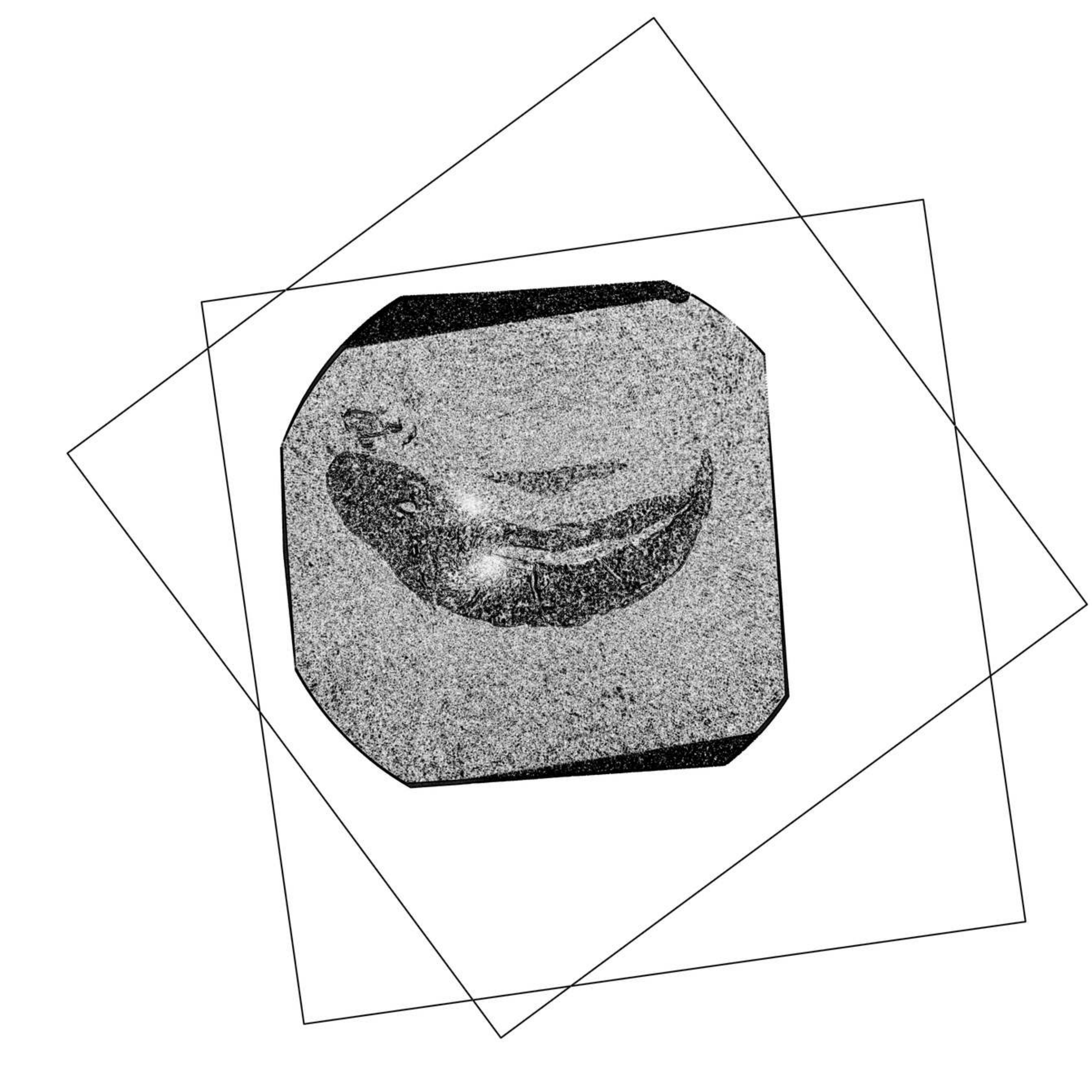}}\hfill%
\subfloat[][]{\label{fig:funatomi-full-d}\includegraphics[width=0.495\linewidth]{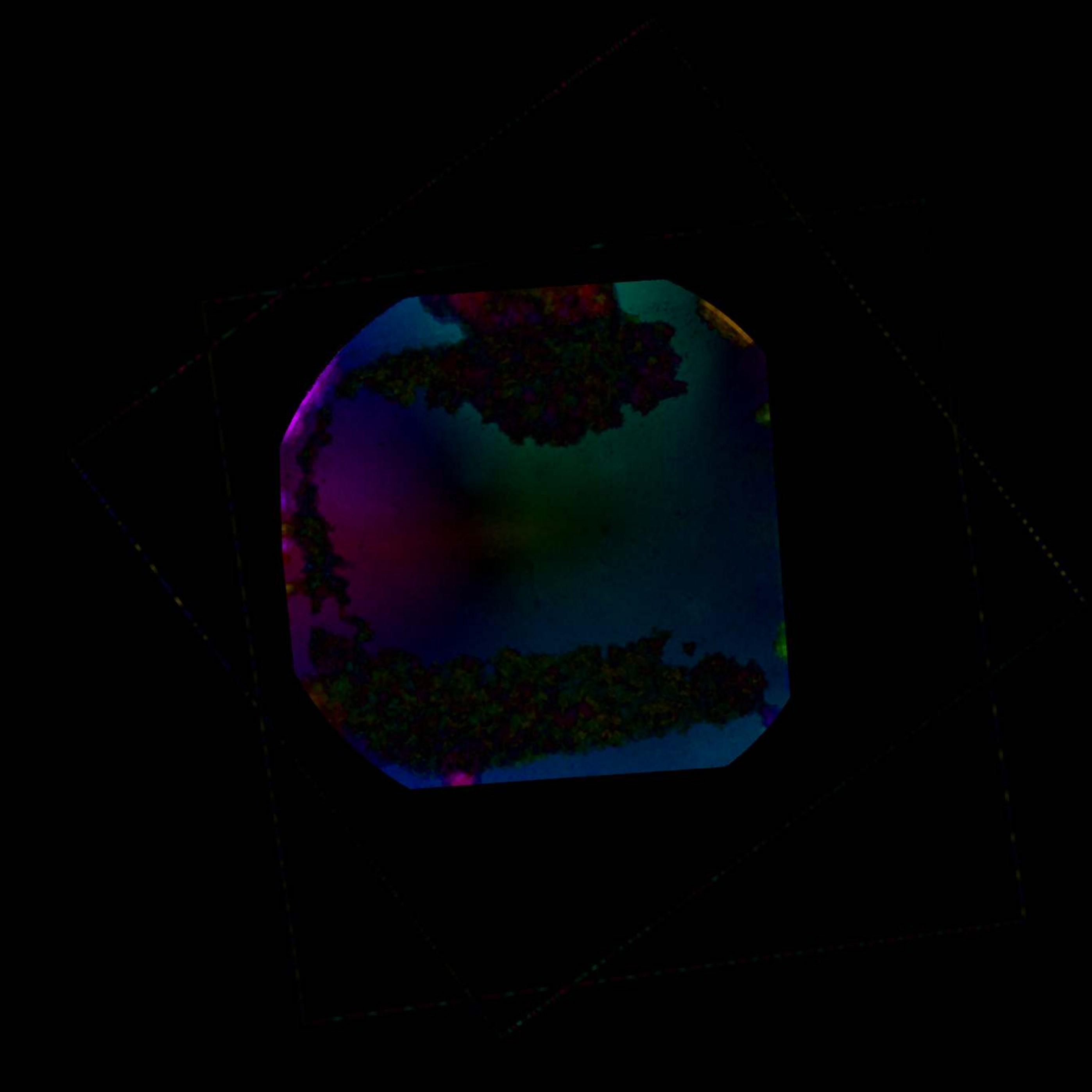}}

\caption{CT data set, \enquote{Blending} method. Here image-based evaluation methods of the full images are shown. \protect\subref{fig:funatomi-full-a}: Registered image. \protect\subref{fig:funatomi-full-b}: PSNR visualization. \protect\subref{fig:funatomi-full-c}: SSIM visualization. \protect\subref{fig:funatomi-full-d}: optical flow visualization.}
\olcomment{redo for new one?}
\label{fig:funatomi-full}
\end{figure}

\begin{figure}[t!]
\centering
\subfloat[][]{\label{fig:funatomi-flow-cmp-a}\includegraphics[width=0.495\linewidth]{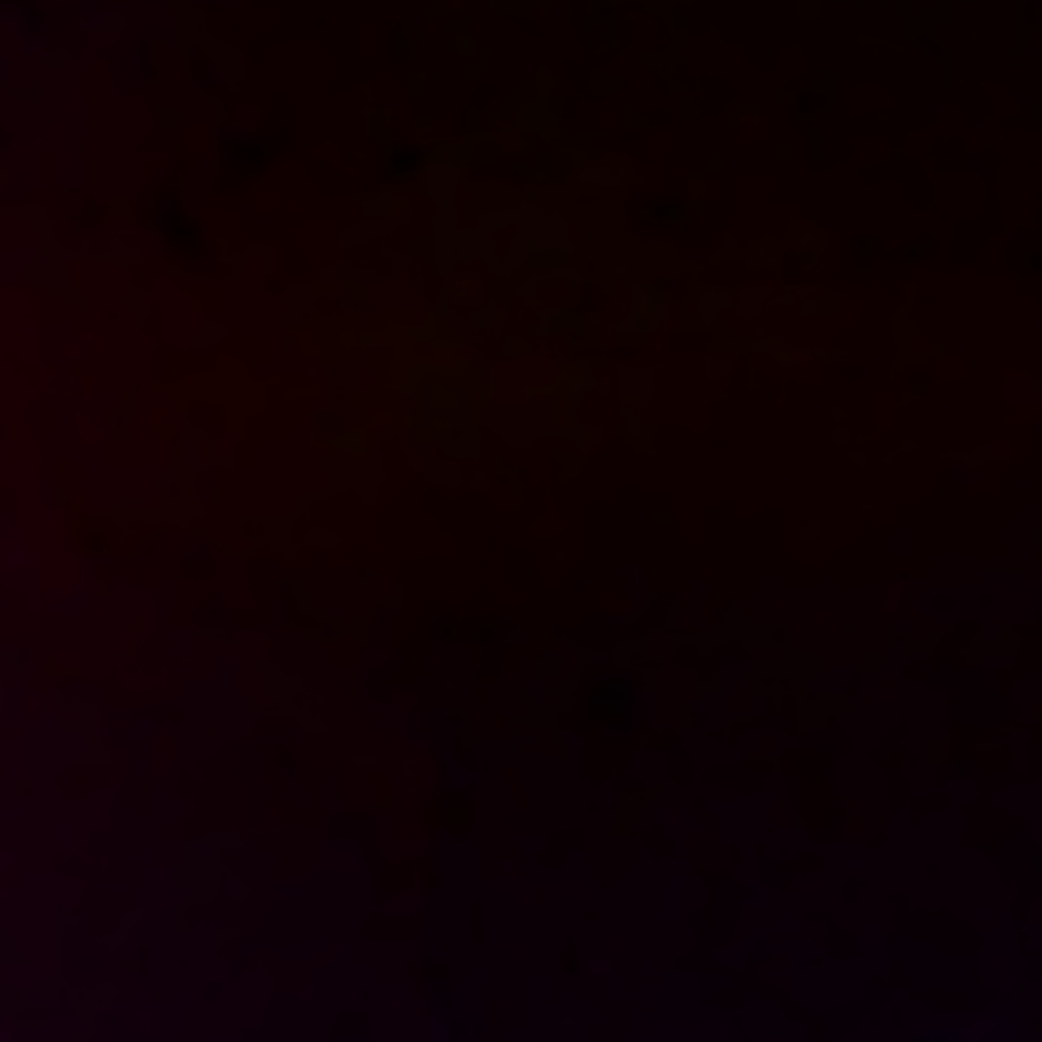}}\hfill%
\subfloat[][]{\label{fig:funatomi-flow-cmp-b}\includegraphics[width=0.495\linewidth]{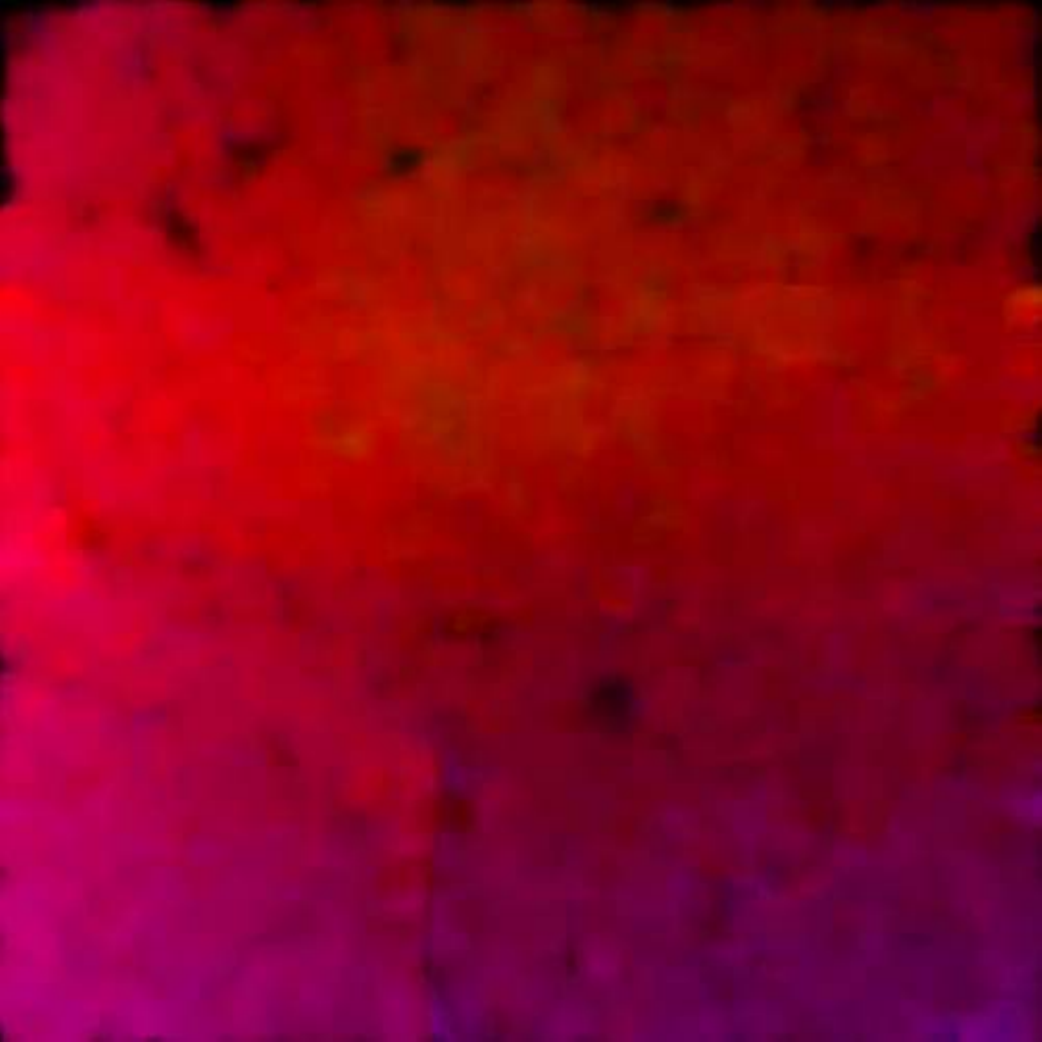}}

\caption{CT data set, an older result from the \enquote{Blending} method. \protect\subref{fig:funatomi-flow-cmp-a}: First optical flow then crop. \protect\subref{fig:funatomi-flow-cmp-b}: First crop, then optical flow. Notice the magnitude differences in the visualization.}
\label{fig:funatomi-flow-cmp}
\end{figure}

Figure~\ref{fig:funatomi-full} shows the uncropped image-based
evaluation results on CT data set
\citep[\citet{appuhn_capillary_2021}]{mhh20}, registered with
\enquote{Blending} method \citep{kajihara_non-rigid_2019}. Notice the
border effects. Fig.~\ref{fig:funatomi-flow-cmp} compares optical flow
visualization, computed on the full image (similar to
\ref{fig:funatomi-full-d}), then cropped to middle and optical flow
visualization from cropped images.

In the main text, we use SSIM \citep{wang_image_2004} values from crops.
We crop SSIM visualizations, as detailed below. We use Jaccard measures
from crops. In contrast to the main paper, the optical flow
visualization \citep{farnebaeck_two-frame_2003} in
Fig.~\ref{fig:funatomi-flow-cmp-a} is computed from the full images.

\hypertarget{computation-of-the-measures-in-general}{%
\subsection{Computation of the measures in
general}\label{computation-of-the-measures-in-general}}

A~lot of fine details impact the metrics and the visuals. In the main
paper, we used \xroi{500}{500} crops. Due to its nature, SSIM on crops
is a lesser image. We computed SSIM on the full images and cropped then.
This does not change the visual distribution of the measure. SSIM values
used in the numerical comparisons were computed on crops. This way, we
do not let the border effects from padding impact the evaluation.

Our implementation of optical flow visualization normalizes the sizes of
the displacements across the image. This leads to very boring images in
the case first the flow is computed and then the visualization is
cropped. Here, we computed the optical flow on crops. Arguably,
different images are less comparable, when computed in this manner.
However, \enquote{hot spots} and movement directions are more prominent
when visualized this way.

We computed PSNR for the visuals on color images, even though all our
result images were inherently grayscale. The reason is a slightly nicer
visualization. However, the PSNR values on such \enquote{fake} RGB
images are misleading. The numerical PSNR values were computed on
grayscale images.

The \enquote{original} ground truth images for EM are 16~bit. We
evaluate them, too, but consider the 8~bit converted (and normalized)
ground truth images to be a better reference value for the evaluation of
registrations.

\hypertarget{thresholding-and-dice-measure-visualizations}{%
\subsection{Thresholding and Dice measure
visualizations}\label{thresholding-and-dice-measure-visualizations}}

While the Dice measure can be trivially expressed in terms of the
Jaccard measure, both of them use binary images as inputs. We use a
simple threshold for Jaccard measure; we always state the threshold in
the description of the Jaccard measure. In this case we use a global
threshold. In our Dice measure visualizations below, the binary images
were produced with multiple methods:

\begin{itemize}
\tightlist
\item
  global binary threshold;
\item
  Otsu method \citep{otsu_threshold_1979};
\item
  Otsu method in blurred images;
\item
  with Matlab's \activecontour function at 300 iterations, starting from
  a binary thresholded image.
\end{itemize}

\begin{figure*}[!t]
\centering
\subfloat[][Original data]{\label{fig:SUP-eval-ex-a}\includegraphics[width=0.199\linewidth]{image/LS/add-eval/feature-rigid/crop_square_input_0150-dump.pdf}}\hfill
\subfloat[][Mask]{\label{fig:SUP-eval-ex-b}\includegraphics[width=0.199\linewidth]{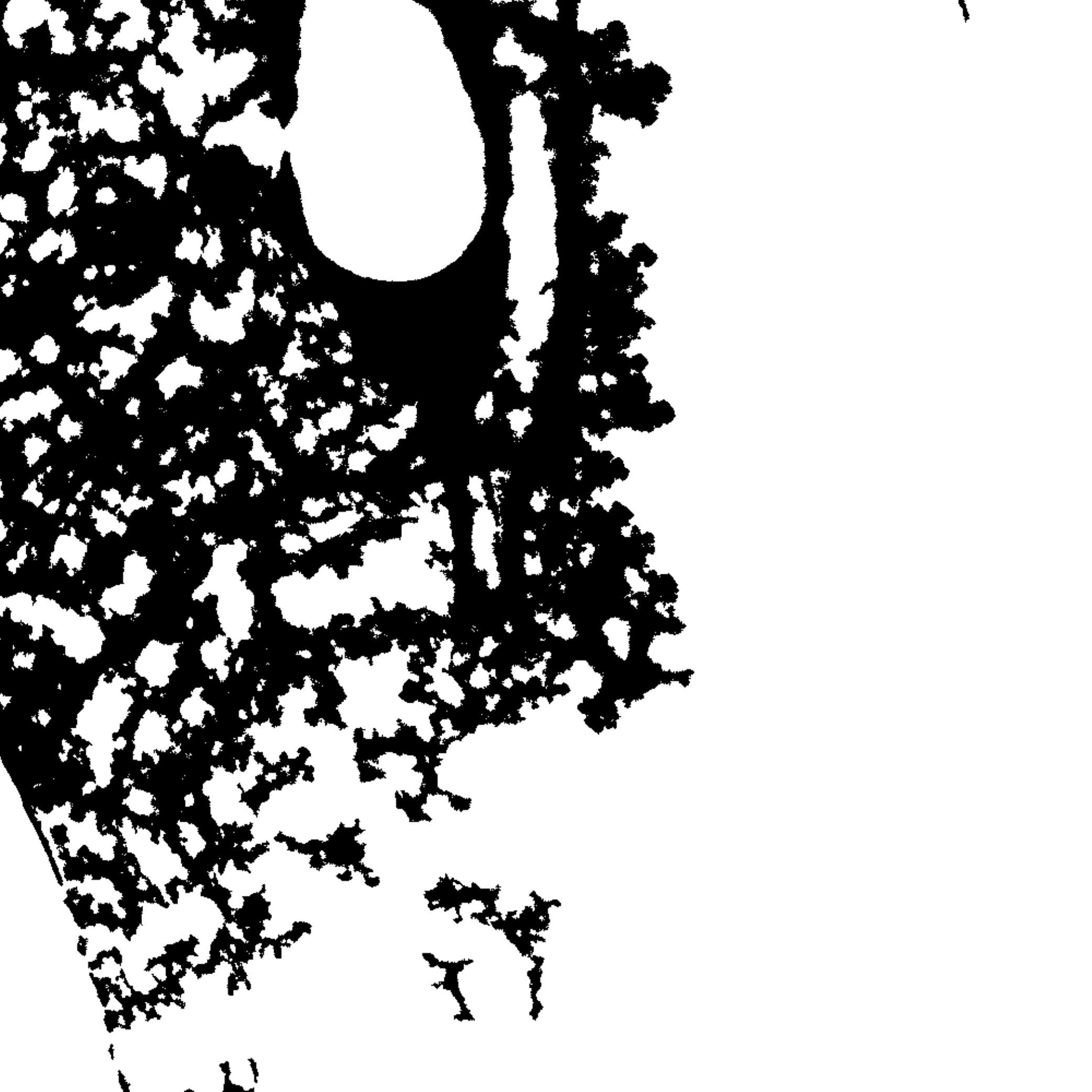}}\hfill
\subfloat[][Dice measure $=0.887$, locally distorted data, rigid transform to follow]{\label{fig:SUP-eval-ex-c}\includegraphics[width=0.199\linewidth]{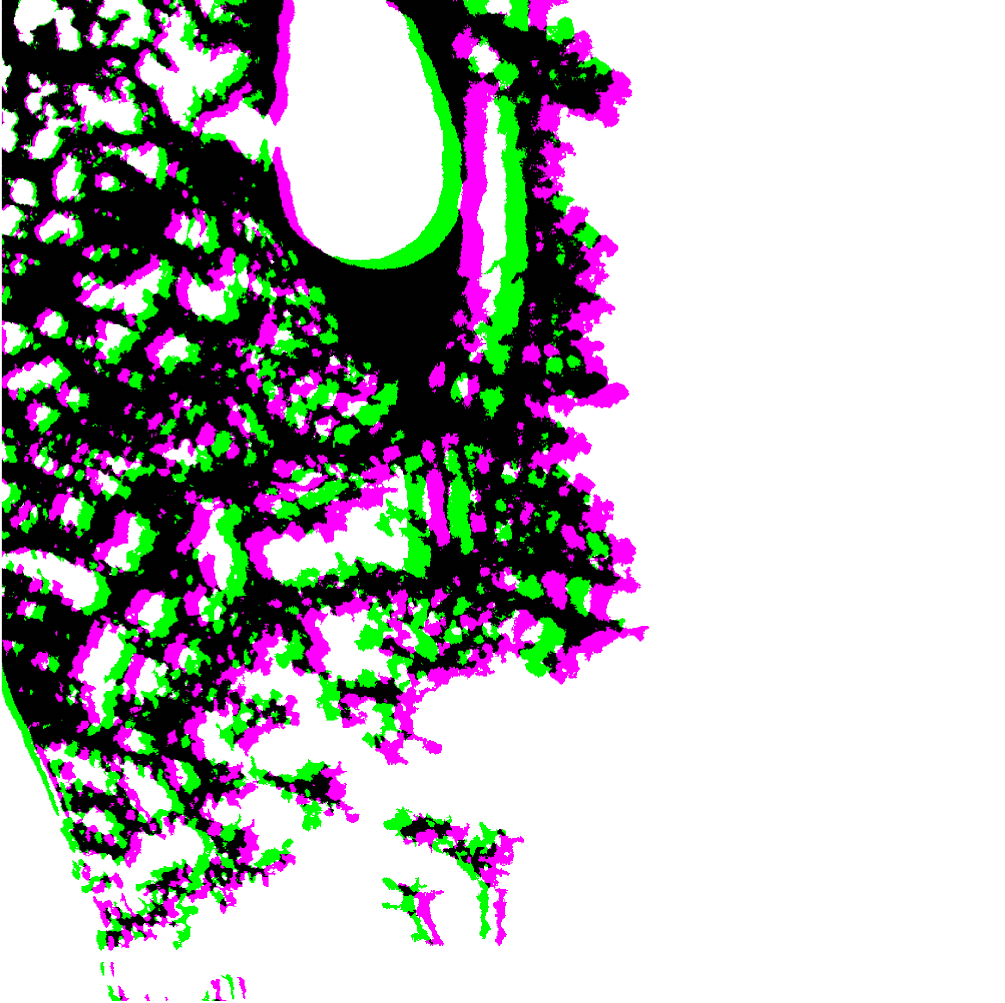}}\hfill
\subfloat[][Dice measure $=0.917$,\\after Rigid-SIFT]{\label{fig:SUP-eval-ex-d}\includegraphics[width=0.199\linewidth]{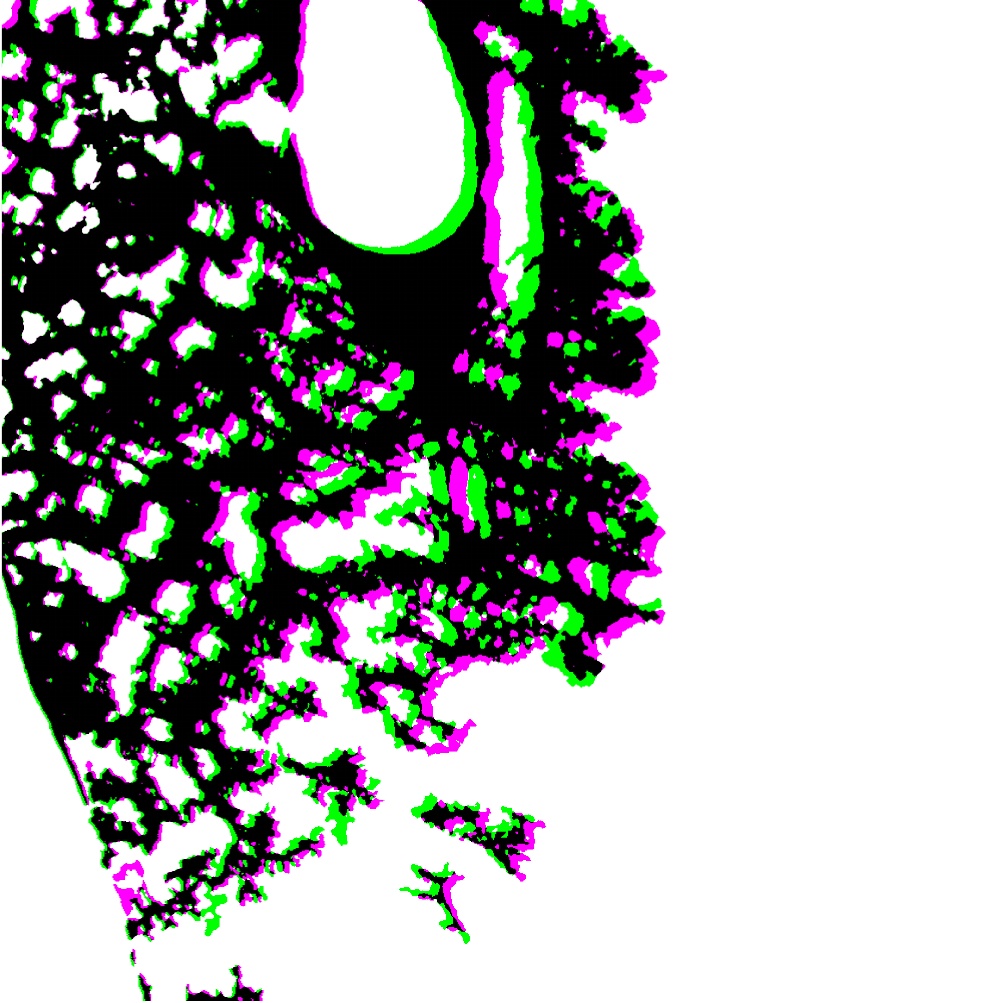}}\hfill
\subfloat[][Dice measure $=0.983$, ground truth]{\label{fig:SUP-eval-ex-e}\includegraphics[width=0.199\linewidth]{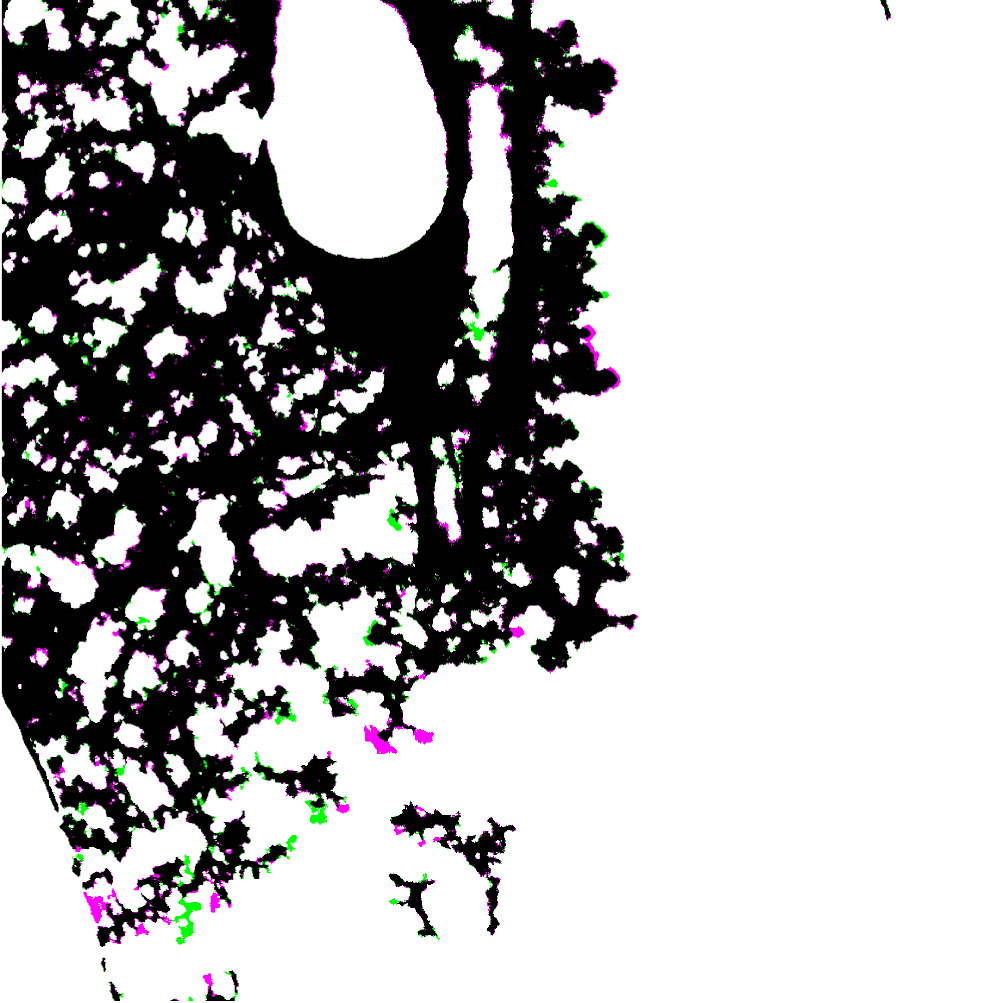}}

\caption{An example of an evaluation based on Dice
measure in Matlab. \protect\subref{fig:SUP-eval-ex-a}: The region of
interest, which we evaluate. It is a rat lung, obtained with LS microscopy.
Using Matlab's \activecontour function,
the mask~\protect\subref{fig:SUP-eval-ex-b} is computed with
threshold~150. \protect\subref{fig:SUP-eval-ex-c}: The Dice measure
after the local distortions from our method, but before rigid transform. The rigidly transformed series serves then as an input to
the registration. \protect\subref{fig:SUP-eval-ex-d}: Dice
measure on the result of Rigid-SIFT method. No non-rigid steps
are performed. The original, undistorted data is
in~\protect\subref{fig:SUP-eval-ex-e}. 
The higher the Dice measure is the better.}

\label{fig:SUP-eval-ex}
\end{figure*}

To give an example, consider Fig.~\ref{fig:SUP-eval-ex} that shows a
region from LS data set. The masks, such as~\subref{fig:SUP-eval-ex-b},
were computed using the \activecontour function in Matlab with the
threshold~150. We see that the Dice measure
in~\subref{fig:SUP-eval-ex-d} roughly midway
between~\subref{fig:SUP-eval-ex-c} and~\subref{fig:SUP-eval-ex-e}. We
cannot hope to reach values similar to~\subref{fig:SUP-eval-ex-e} with
the method used, as no non-rigid alignment happens in this case. Thus,
the result of the rigid alignment is acceptable. It manages to align the
sections rigidly, despite additional non-rigid distortions induced by
our generation of input data.

\begin{figure*}[!t]
\centering
\subfloat[][Otsu+blur threshold]{\label{fig:eval-py-b}\includegraphics[width=0.199\linewidth]{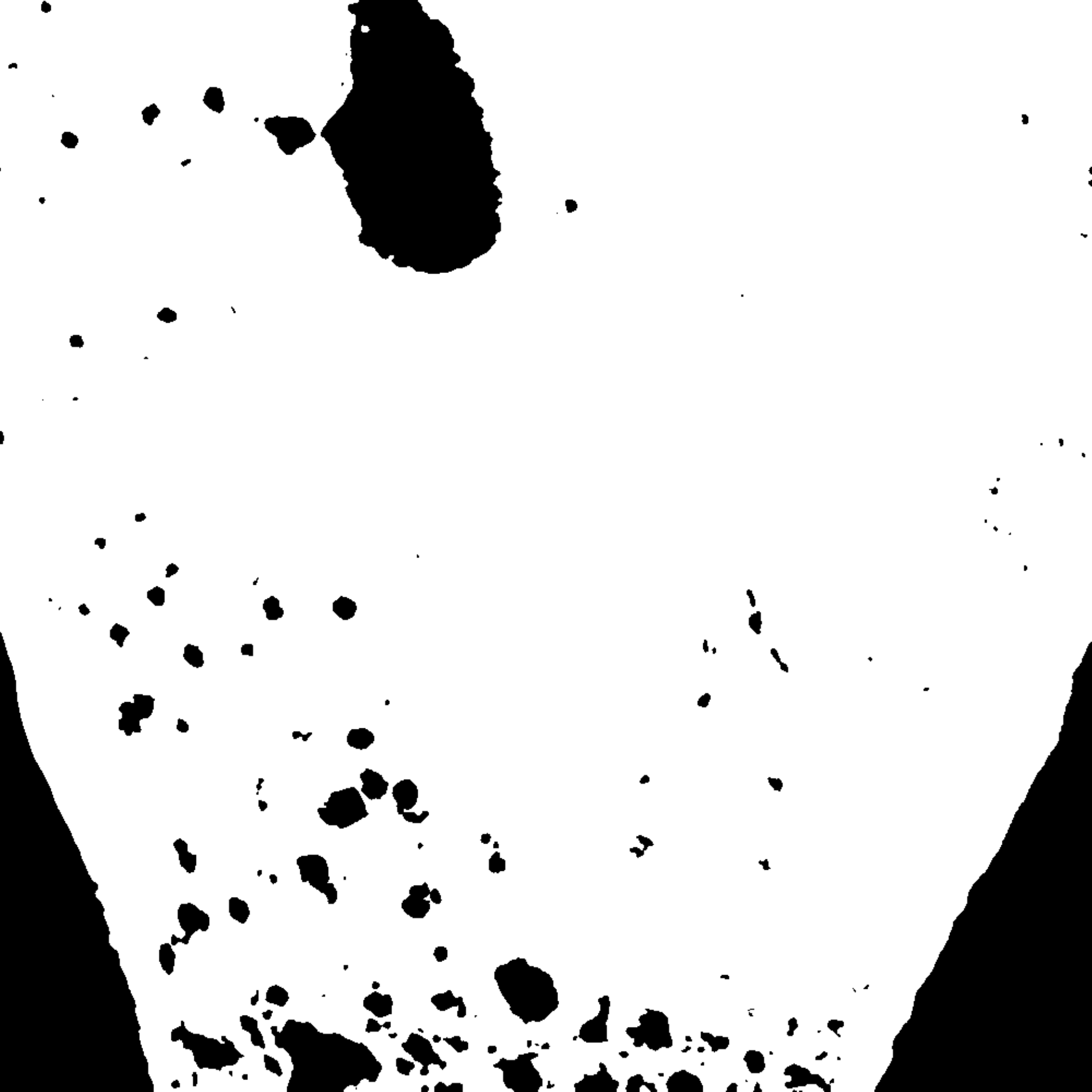}}\hfill
\subfloat[][Dice on global 100]{\label{fig:eval-py-b2}\includegraphics[width=0.199\linewidth]{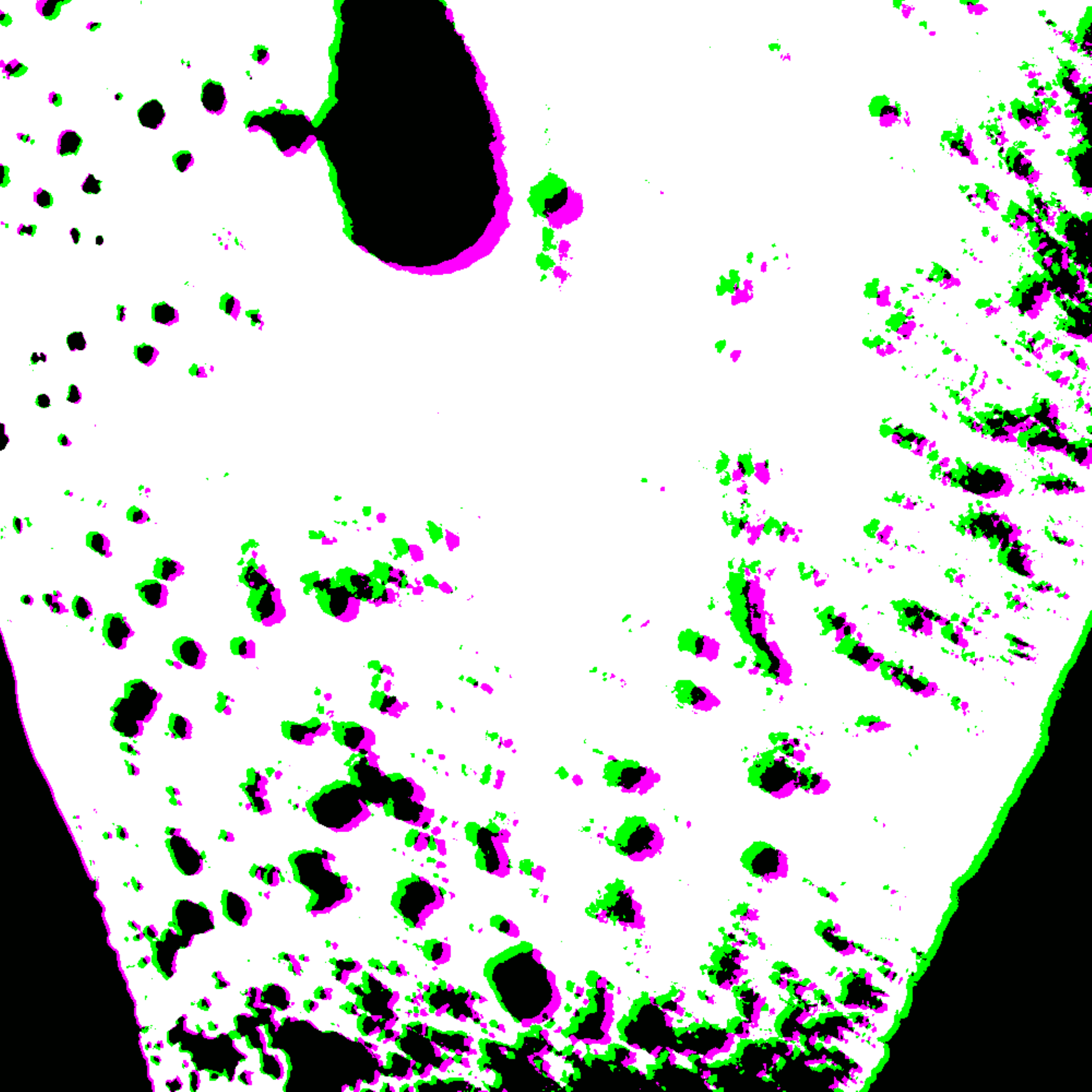}}\hfill
\subfloat[][Dice on global 150]{\label{fig:eval-py-c}\includegraphics[width=0.199\linewidth]{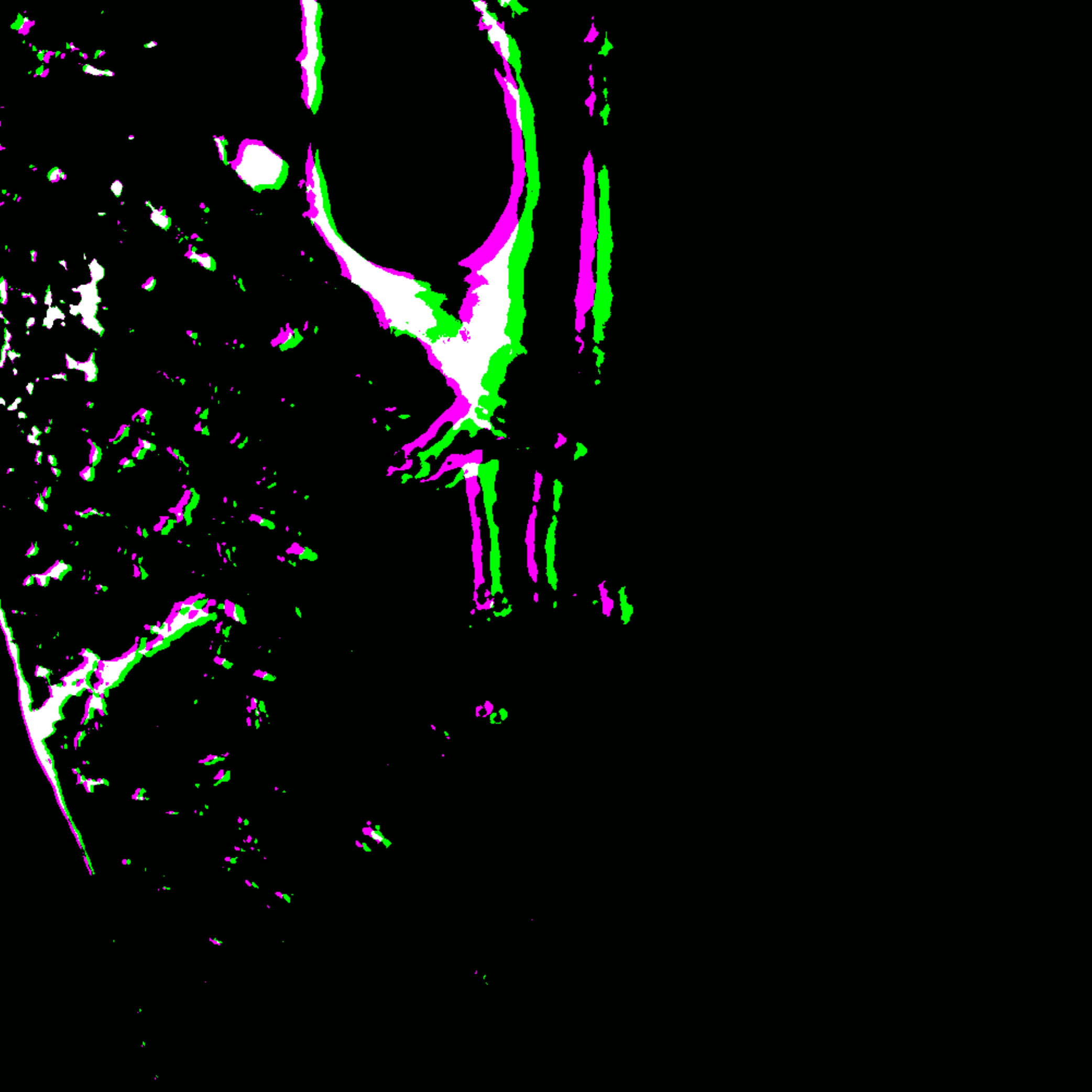}}\hfill
\subfloat[][Dice on Otsu w/o blur]{\label{fig:eval-py-d}\includegraphics[width=0.199\linewidth]{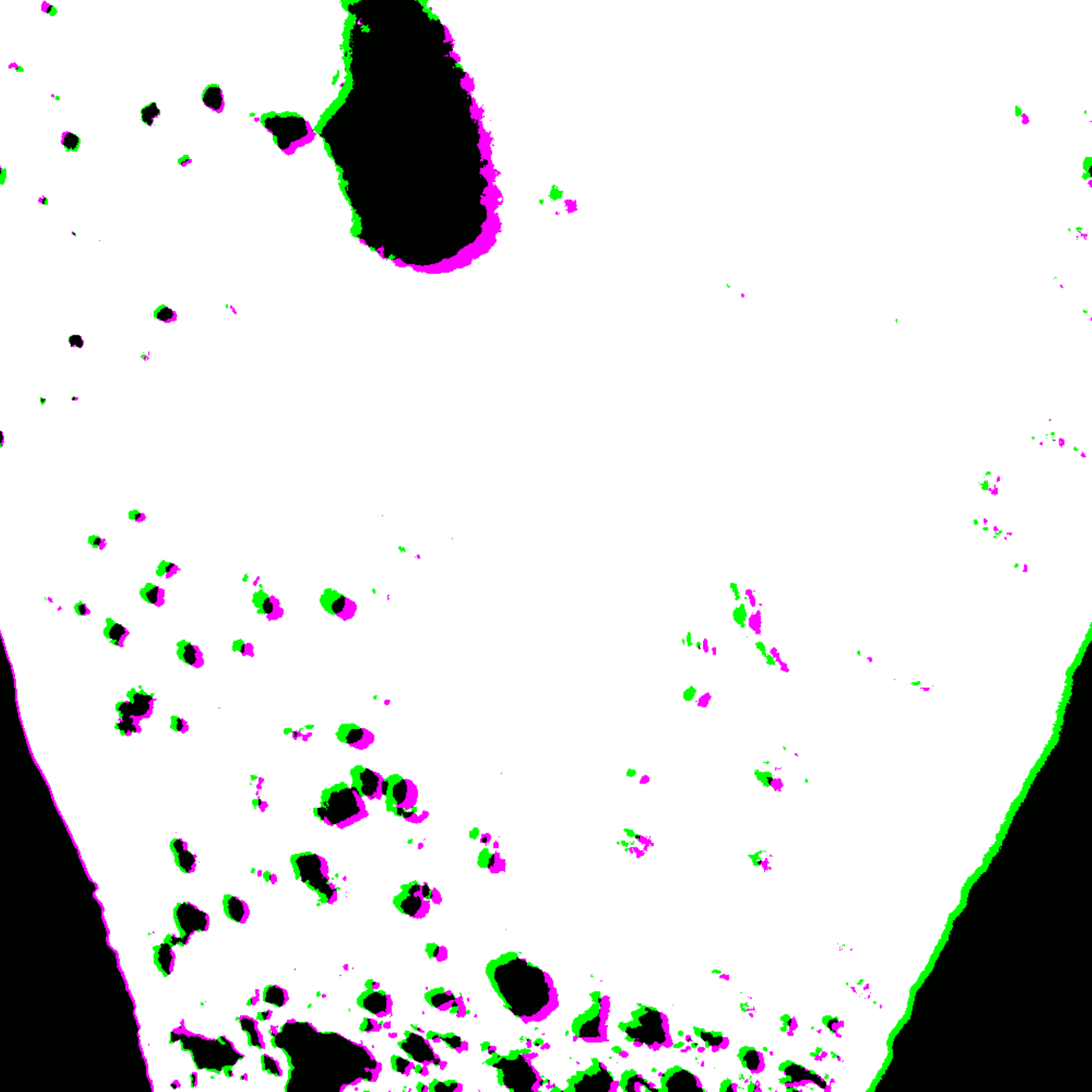}}\hfill
\subfloat[][Dice on Otsu with blur]{\label{fig:eval-py-e}\includegraphics[width=0.199\linewidth]{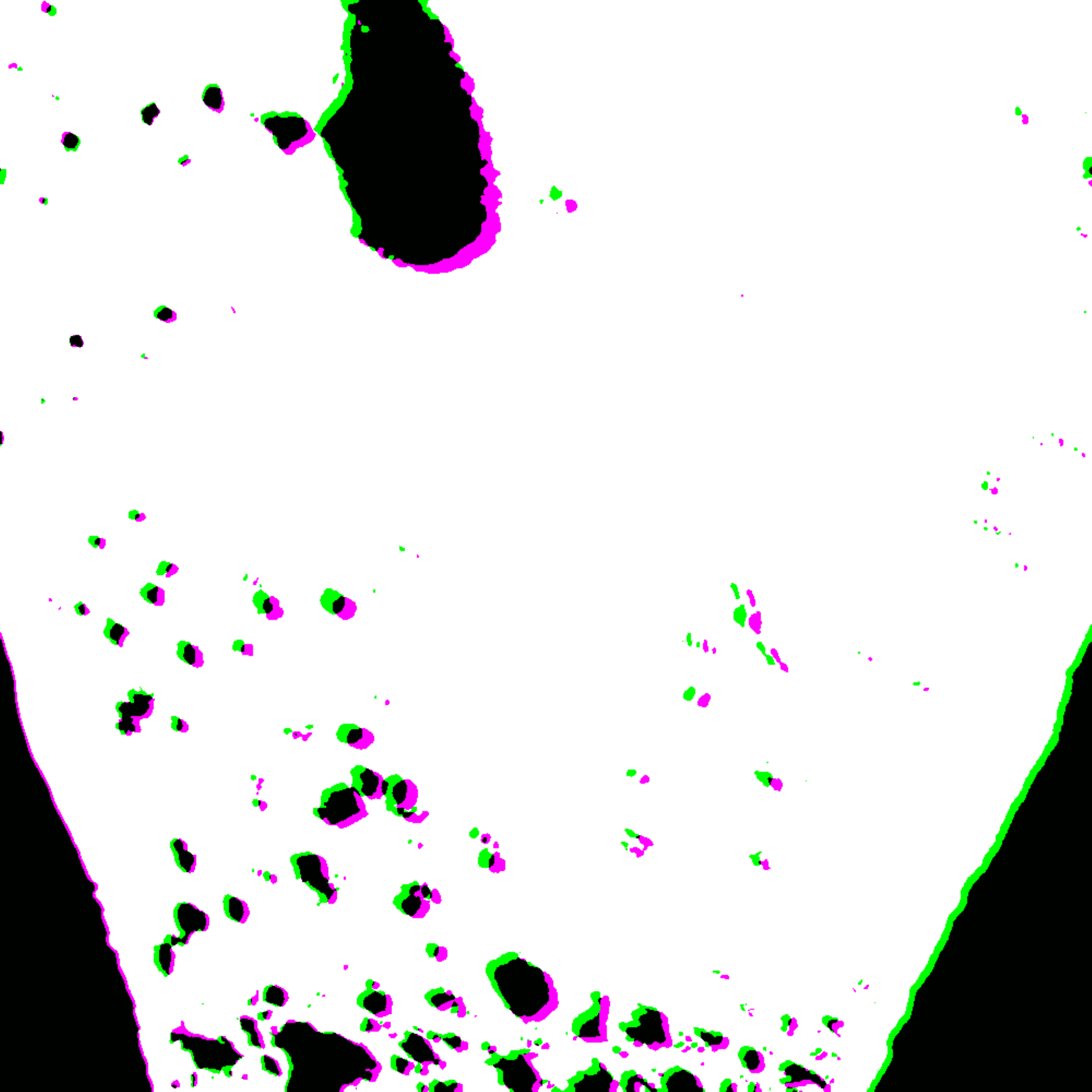}}

\caption{An example of an evaluation based on Dice
measure in Python. The region of
interest, which we evaluate, same as above.
The mask~\protect\subref{fig:eval-py-b} is computed with Otsu's method on blurred images, using OpenCV. All the measures \protect\subref{fig:eval-py-c}--\protect\subref{fig:eval-py-e} are computed for an  image pair, registered with Rigid-SIFT. In contrast to Matlab, here is white the common area in both thresholded images. \protect\subref{fig:eval-py-b2}: The Dice measure visualization on global threshold of~100. \protect\subref{fig:eval-py-c}: The Dice measure visualization on global threshold of~150. \protect\subref{fig:eval-py-d}: The Dice measure visualization on Otsu's thresholding without a blur.
\protect\subref{fig:eval-py-e}: The Dice measure visualization on Otsu's thresholding on blurred images with blur radius~5. 
The higher the overlap in Dice measure the better.}
\label{fig:eval-py}
\end{figure*}

Figure~\ref{fig:eval-py} shows the same region as
Fig.~\ref{fig:eval-ex}, the Rigid-SIFT method \subref{fig:eval-ex-d}.
Here, the same image pair is evaluated using different thresholding
methods. We conclude that Otsu's method on blurred images produces best
visualizations, although little difference is seen to Otsu without blur.
For our Dice measure visualizations in the main text we use Otsu's
method on blurred images. For Jaccard evaluation in the main text, we
use the global thresholding for its consistency. One of the reasons for
these decisions: Our input images are individually normalized. (See
Section~\ref{sec:effect-norm}.) Even if they are normalized back to a
\enquote{common denominator}, some differences may remain. We also use
the input images (with all their discrepancies) in the evaluation.

Notice, that the meaning of colors white and black in
Figures~\ref{fig:SUP-eval-ex} and~\ref{fig:eval-py} is different. We
argue that using white for the overlap is better.


\begin{table}[!t]
\centering
\caption{Numerical values for the evaluation of the registrations on the CT data set, in the middle of the series (Figs. \ref{fig:eval-ct-1}, \ref{fig:eval-ct-2}). In Jaccard measure, we used a global threshold of~100. Low quality values in blending transforms can be explained by residual shifts. Notice the differences between the rigid-only and non-rigid methods. Notice also the lower values in the ground truth, when compared to \elastix. The "Locally distorted" line shows the local distortions only, without global rigid transforms. There was some larger movement (as seen in Fig.~\ref{fig:eval-ct-2}), this fact affects the measures. The registrations' inputs, however, were also rigidly transformed beforehand.\newline%
The larger the values, the better---up to the ground truth value.}
%
%
\label{tab:eval-ct}
\begin{tabular}{r S[round-mode = figures, round-precision = 3, table-format = 1.3] S[round-mode = figures, round-precision = 3, table-format = 2.1] S[round-mode = figures, round-precision = 3, table-format = 1.3]}
\multicolumn{4}{c}{\emph{CT}} \\
\toprule
\multicolumn{1}{c}{Method}       & {Jaccard} & {PSNR} & {SSIM} \\
\midrule
Rigid-SIFT   & 0.12618114772989167  & 22.2166 & 0.47626 \\
Rigid-SURF   & 0.1350185873605948   & 22.7097 & 0.50812 \\
Deform-SURF  & 0.3753199268738574   & 28.5062 & 0.77221 \\
GS           & 0.4426400166562565   & 30.9448 & 0.87398 \\
Blending     & 0.12305350661515045  & 22.2765 & 0.46608 \\
\elastix     & 0.5668404588112618   & 32.951  & 0.89829 \\
Locally distorted  &
               0.028323699421965318 & 19.8731 & 0.39437 \\
Ground truth & 0.5169049621530698   & 31.4465 & 0.87478 \\
\bottomrule
\end{tabular}
\end{table}

\begin{table}[!t]
\centering
\caption{Numerical values for the evaluation of the registrations on the EM data set, in the middle of the series (Figs. \ref{fig:eval-em-1}, \ref{fig:eval-em-2}). In Jaccard measure, we use global thresholds of~100 and 150, as indicated with \enquote{J~100} and \enquote{J~150}. \enquote{Locally distorted} means images resulting from our method applied to normalized ground truth for \emph{local} distortions, but no global rigid transformations were used in this case. The actual registrations operate on images that were also rigidly transformed. Notice the differences between the rigid-only and non-rigid methods. Notice also the differences between both rigid versions as well as the impact of normalization and 8~bit discretization on the ground truth values.\newline%
The larger the values, the better---up to the ground truth value.}
\label{tab:eval-em}
\let\tabsepbakup\tabcolsep
\setlength{\tabcolsep}{5pt}
\begin{tabular}{r S[round-mode = figures, round-precision = 3, table-format = 1.3] S[round-mode = figures, round-precision = 3, table-format = 1.3] S[round-mode = figures, round-precision = 3, table-format = 2.1] S[round-mode = figures, round-precision = 3, table-format = 1.3]}
\multicolumn{5}{c}{\emph{EM}} \\
\toprule
\multicolumn{1}{c}{Method}     & {J~100}      & {J~150}      & {PSNR}  & {SSIM} \\
\midrule
Rigid-SIFT   & 0.9276081431119362 & 0.9225060245713806 & 15.5516 & 0.81254 \\
Rigid-SURF   & 0.924928999536801  & 0.9254346345398599 & 16.4699 & 0.77089 \\
Deform-SURF  & 0.9745534130998824 & 0.9761642655434835 & 22.3554 & 0.88543 \\ 
GS           & 0.9644736716429315 & 0.9671098323246354 & 21.843  & 0.88490 \\
Blending     & 0.9250378431658511 & 0.9192509609315263 & 16.3058 & 0.81581 \\
\elastix     & 0.978667106853811  & 0.9814930078003433 & 23.4023 & 0.90006 \\
Locally distorted
             & 0.9098030739673391 & 0.9064435973320805 & 13.8757 & 0.77175 \\
Ground truth, norm'ed
             & 0.9739544464013851 & 0.9742027371979077 & 21.8014 & 0.92483 \\
Ground truth, original
             & 0.9995468762972292 & 0.9816452794000238 & 37.0671 & 0.86913 \\
\bottomrule
\end{tabular}
\end{table}

\begin{table}[!t]
\centering
\caption{Numerical values for the evaluation of the registrations on the LS data set, in the middle of the series (Figs. \ref{fig:eval-ls-1}, \ref{fig:eval-ls-2}.). In Jaccard measure, we used a global threshold of~100. Low values in \enquote{Blending} and in rigid-only methods can be explained by residual shifts. Notice the improvements in non-rigid methods. The shorthand $s$ stands for \enquote{stretch}.\newline%
The larger the values, the better---up to the ground truth value.}
\label{tab:eval-ls}
\begin{tabular}{r S[round-mode = figures, round-precision = 3, table-format = 1.3] S[round-mode = figures, round-precision = 3, table-format = 2.1] S[round-mode = figures, round-precision = 3, table-format = 1.3]}
\multicolumn{4}{c}{\emph{LS}} \\
\toprule
\multicolumn{1}{c}{Method}       & {Jaccard} & {PSNR}   & {SSIM}   \\
\midrule
Rigid-SIFT           & 0.9293306568382256    & 24.8826  & 0.75905  \\
Rigid-SURF           & 0.5994671672065636    & 19.0251  & 0.47885  \\
Deform-SURF, $s=\num{e-3}$
             & 0.9302626964887405    & 36.0127  & 0.96236  \\
Deform-SURF, $s=\num{5e-3}$
             & 0.9325623530318801    & 36.356   & 0.96456  \\
Deform-SURF, $s=\num{e-2}$
             & 0.9326951353526598    & 36.3505  & 0.96448  \\
GS                   & 0.9404650887398789    & 37.0971  & 0.97291  \\
Blending             & 0.9281149307337095    & 25.3271  & 0.77684  \\
\elastix             & 0.9409431110127936    & 37.1795  & 0.96830  \\
Locally distorted    & 0.909083130281718     & 23.3455  & 0.70967  \\
Ground truth, norm'ed& 0.9814964512292671    & 40.3112  & 0.94642  \\
\bottomrule
\end{tabular}
\end{table}

\begin{figure*}[ptb]
\centering
\setlength{\myfig}{0.2\linewidth-0.001\linewidth-2pt-0.2em}
\setlength{\tabcolsep}{1pt}
\begin{tabular}{m{1em} m{\myfig} m{\myfig} m{\myfig} m{\myfig} m{\myfig}}
& \multicolumn{1}{c}{Image} & \multicolumn{1}{c}{Dice} & \multicolumn{1}{c}{PSNR} & \multicolumn{1}{c}{SSIM} & \multicolumn{1}{c}{Flow} \\
\rot{\myfig}{Rigid-SIFT} &
\scalebarme{%
\includegraphics[height=\myfig]{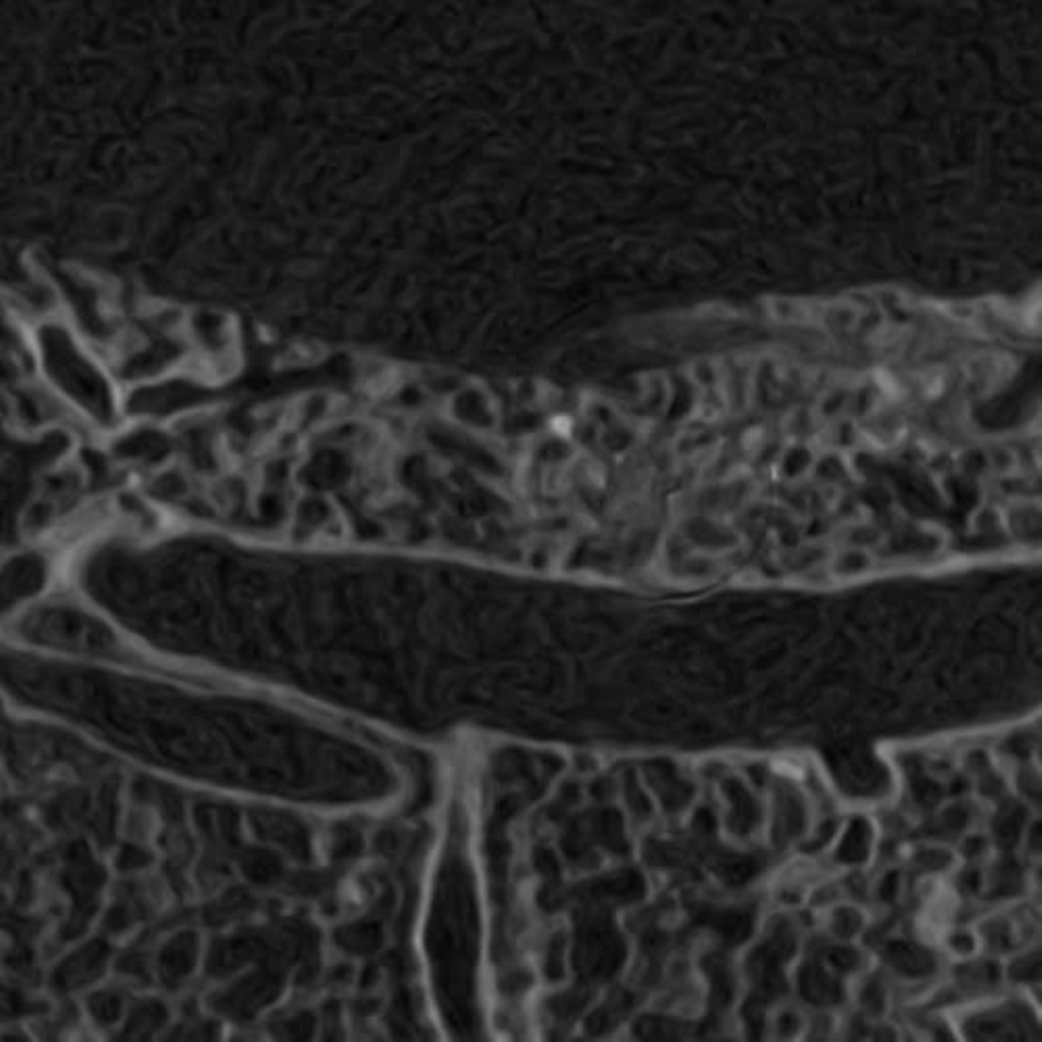}}{%
\includegraphics[width=\myfig]{image/scalebars/scale_ct-500.png}} &
\includegraphics[height=\myfig]{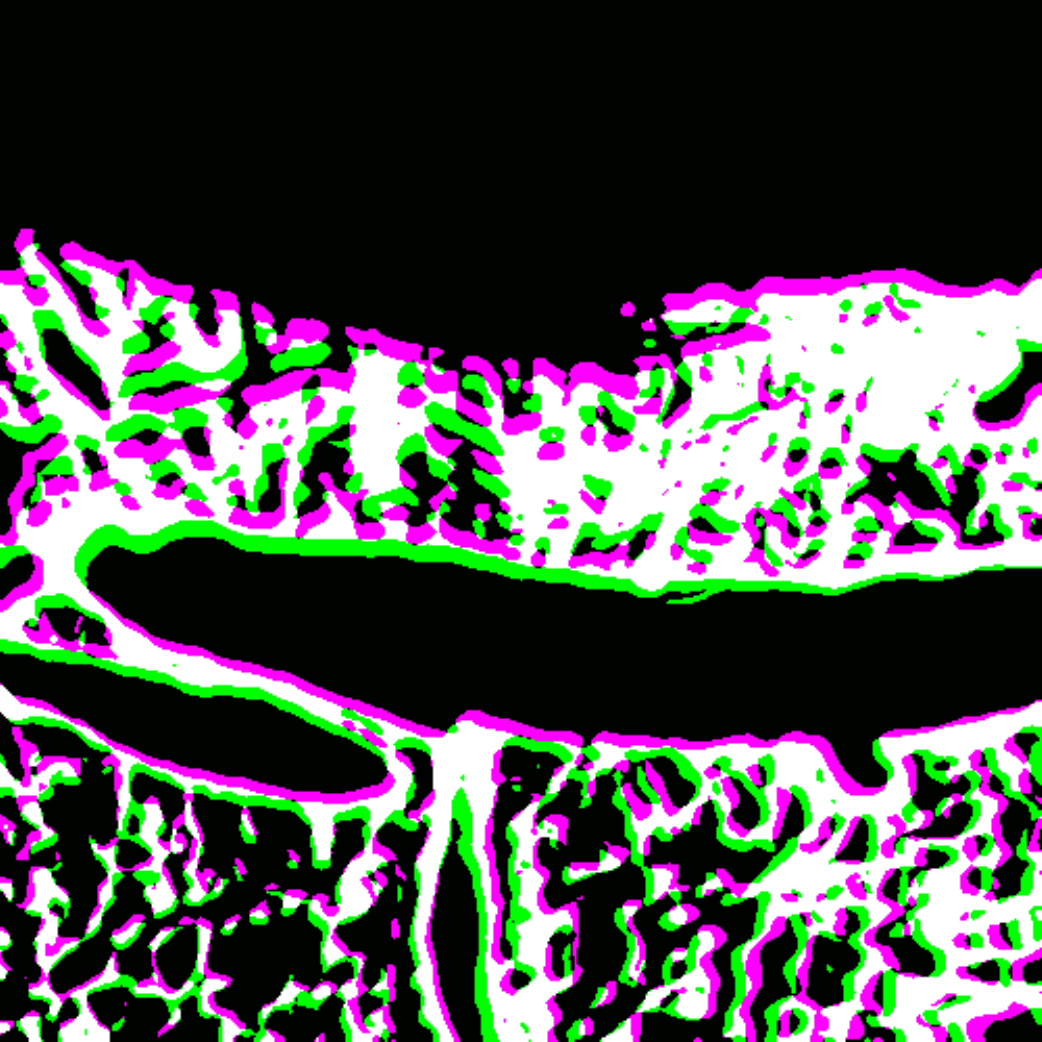} &
\includegraphics[height=\myfig]{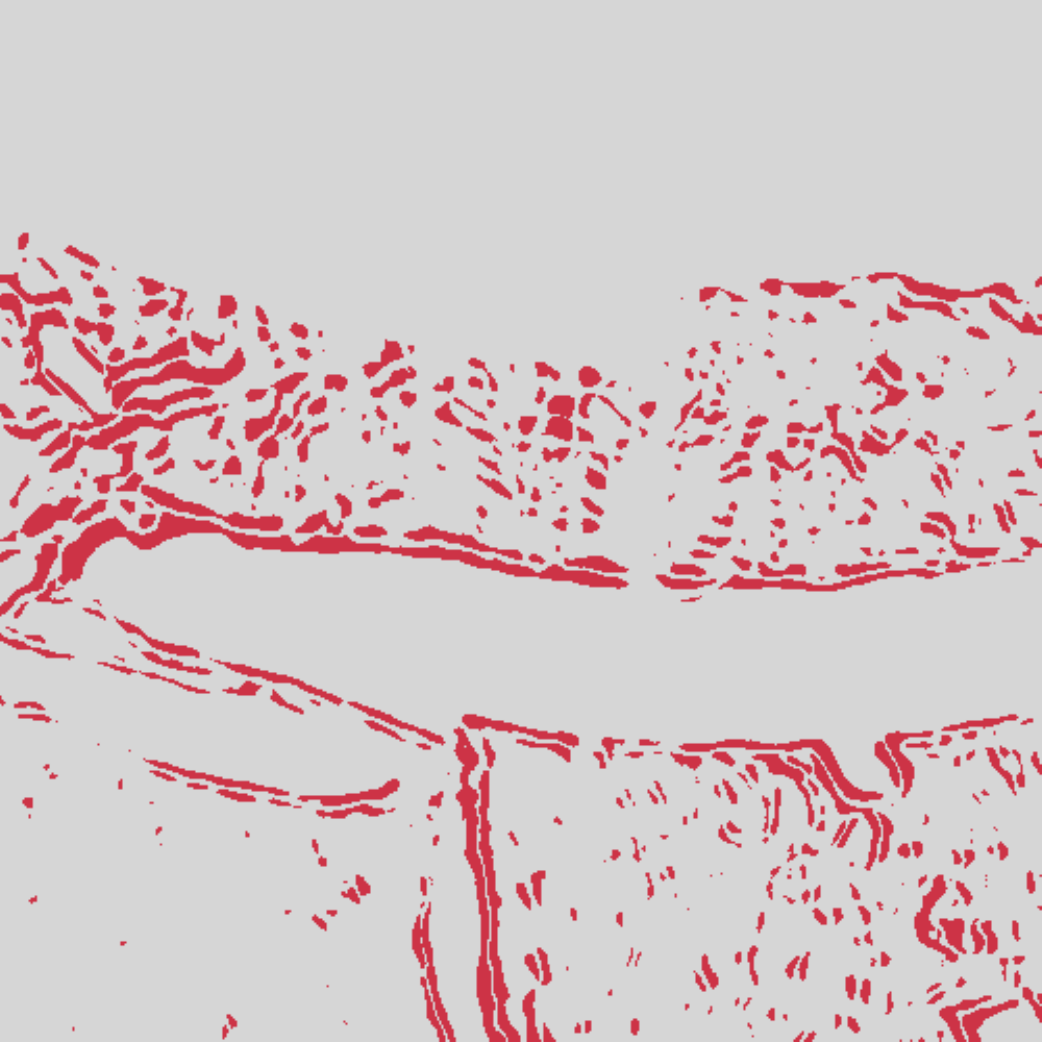} &
\includegraphics[height=\myfig]{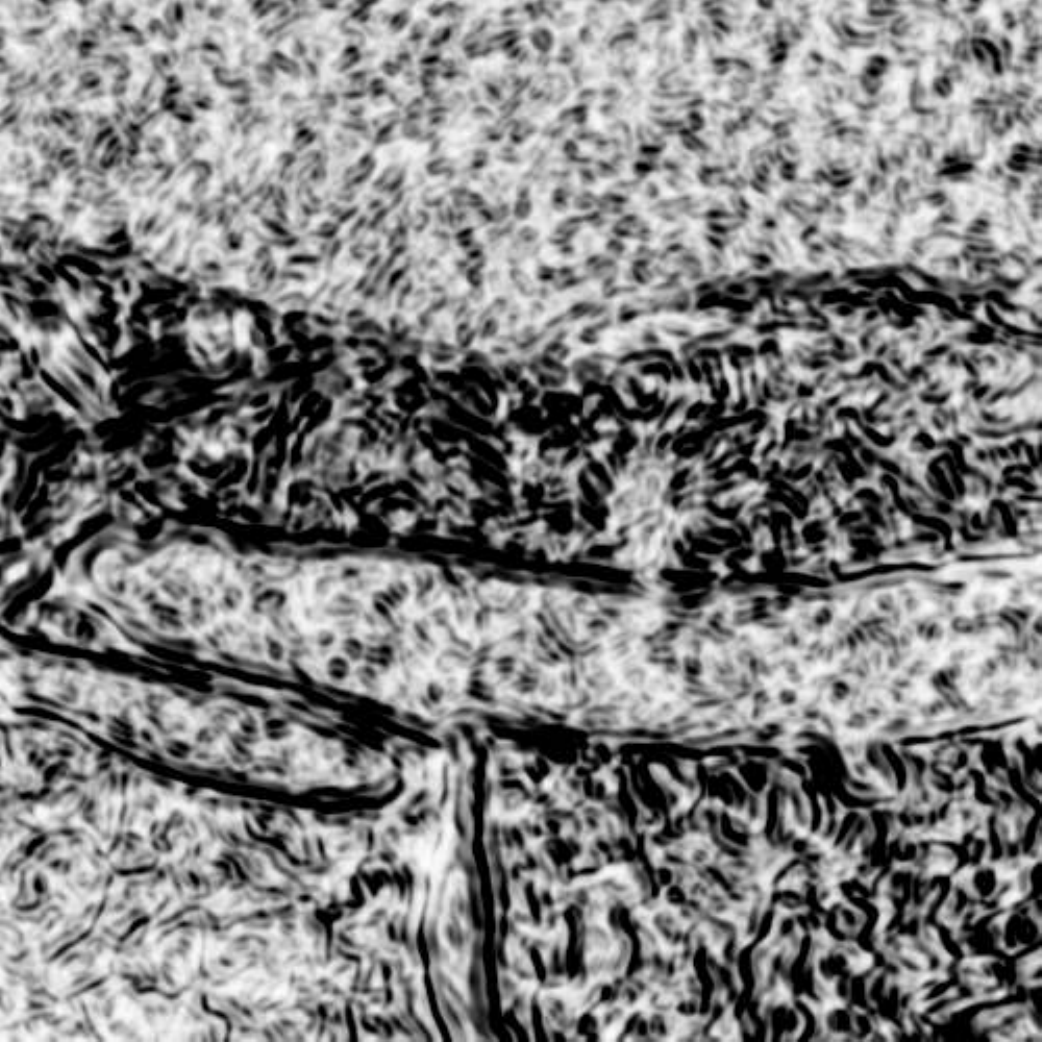} &
\includegraphics[height=\myfig]{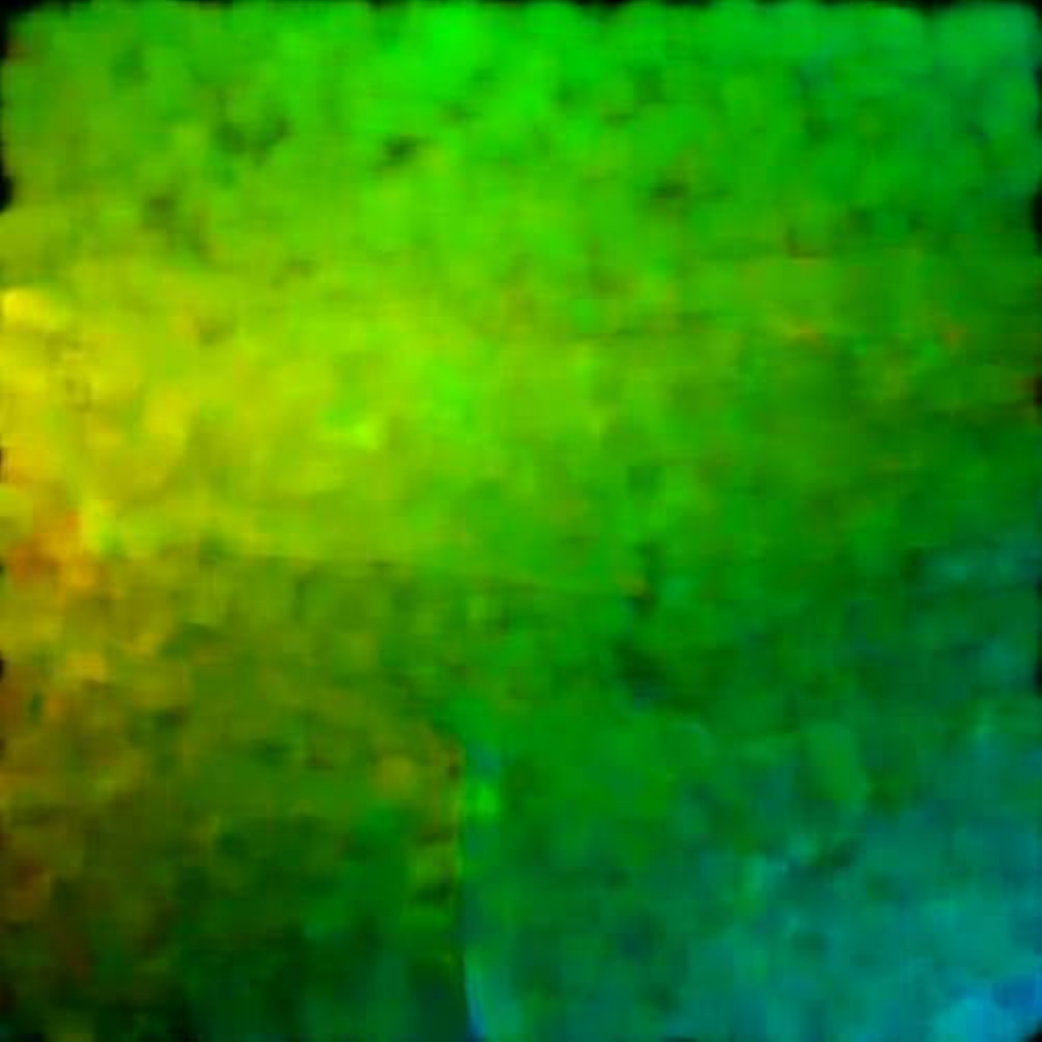} 
\\

\rot{\myfig}{Rigid-SURF} &
\scalebarme{%
\includegraphics[height=\myfig]{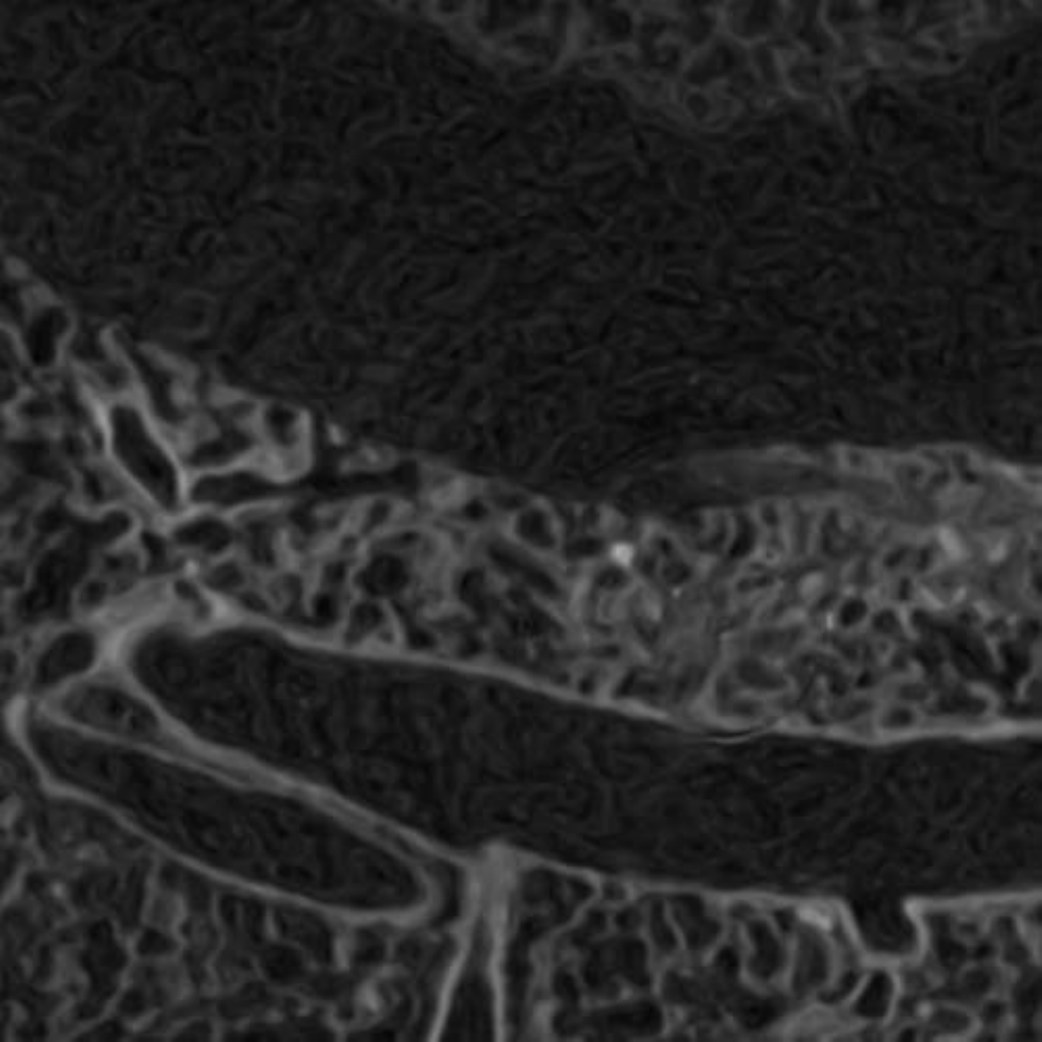}}{%
\includegraphics[width=\myfig]{image/scalebars/scale_ct-500.png}} &
\includegraphics[height=\myfig]{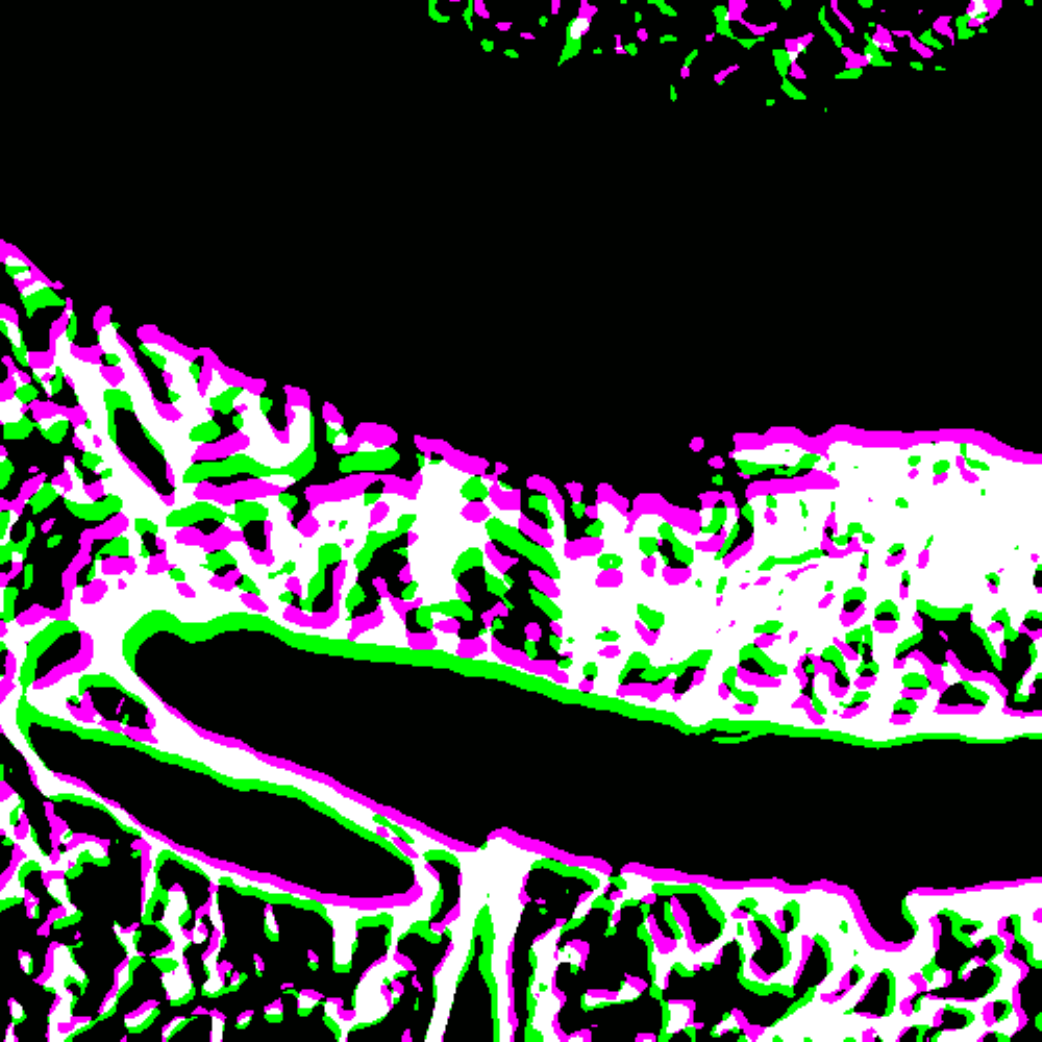} &
\includegraphics[height=\myfig]{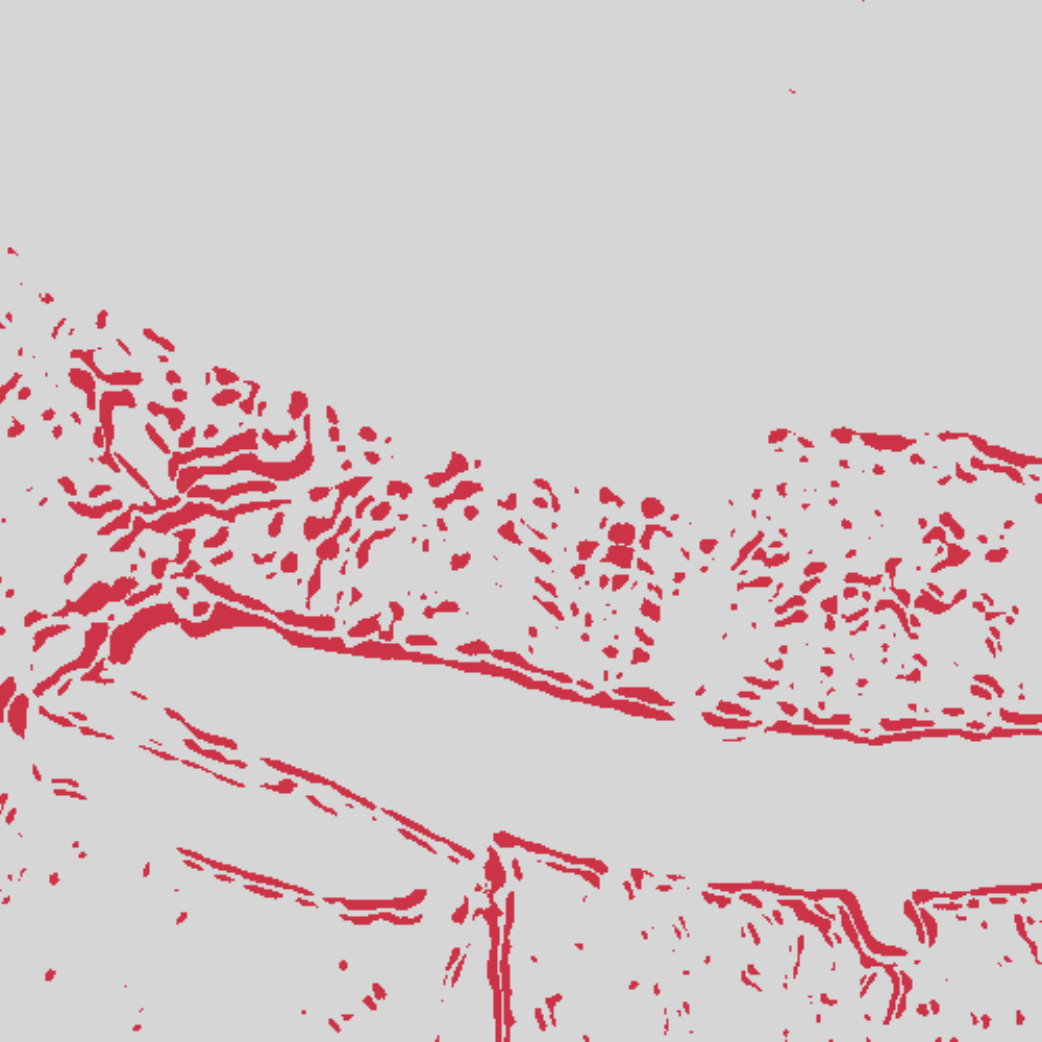} &
\includegraphics[height=\myfig]{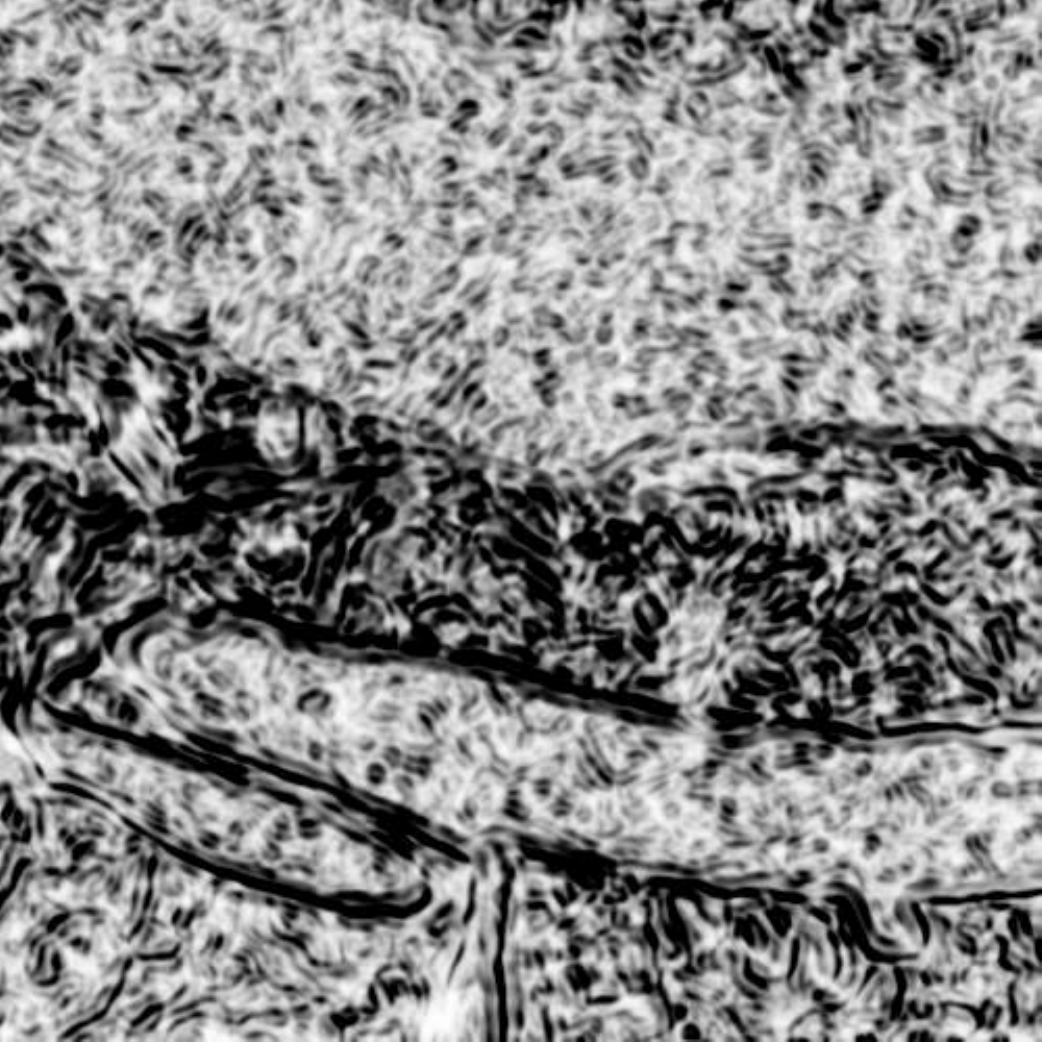} &
\includegraphics[height=\myfig]{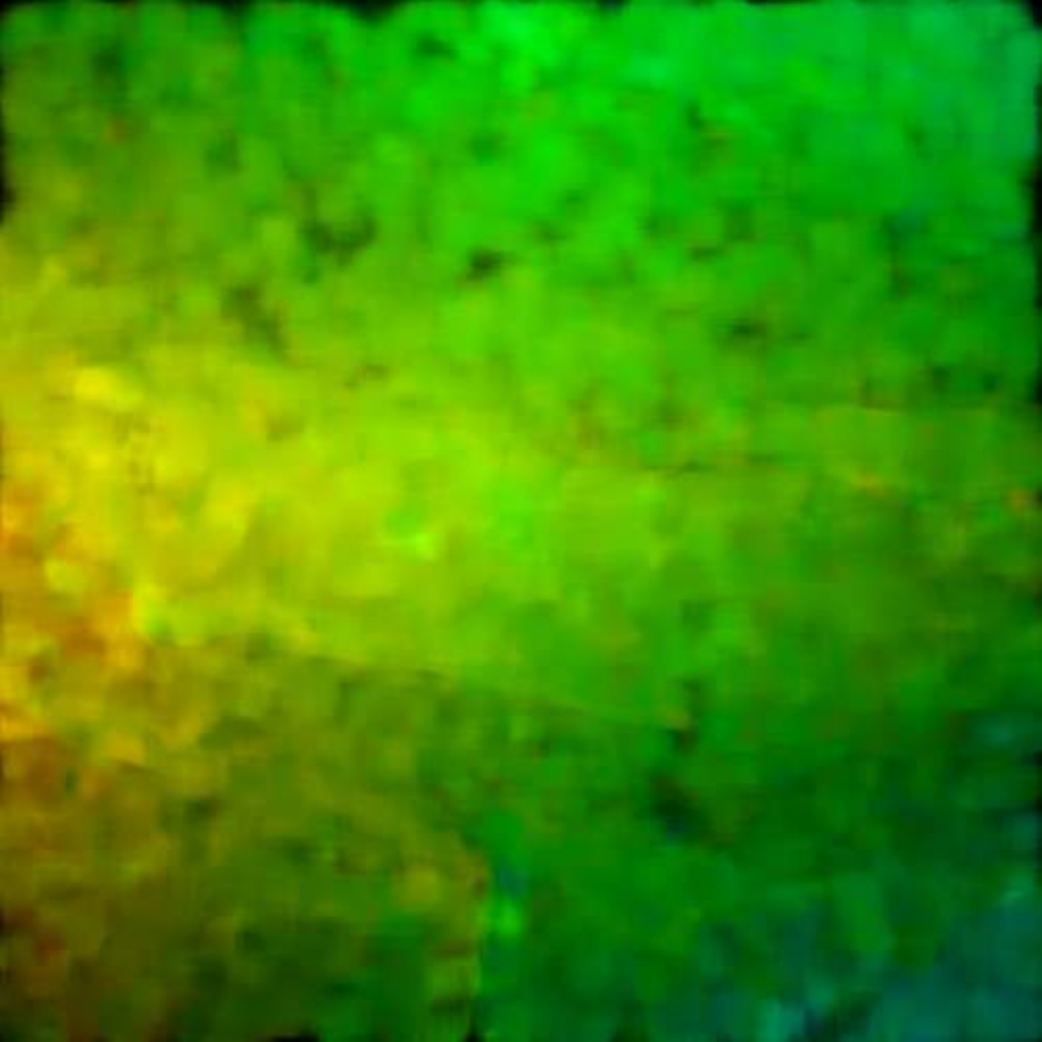}
\\

\rot{\myfig}{~~Deform-SURF} &
\scalebarme{%
\includegraphics[height=\myfig]{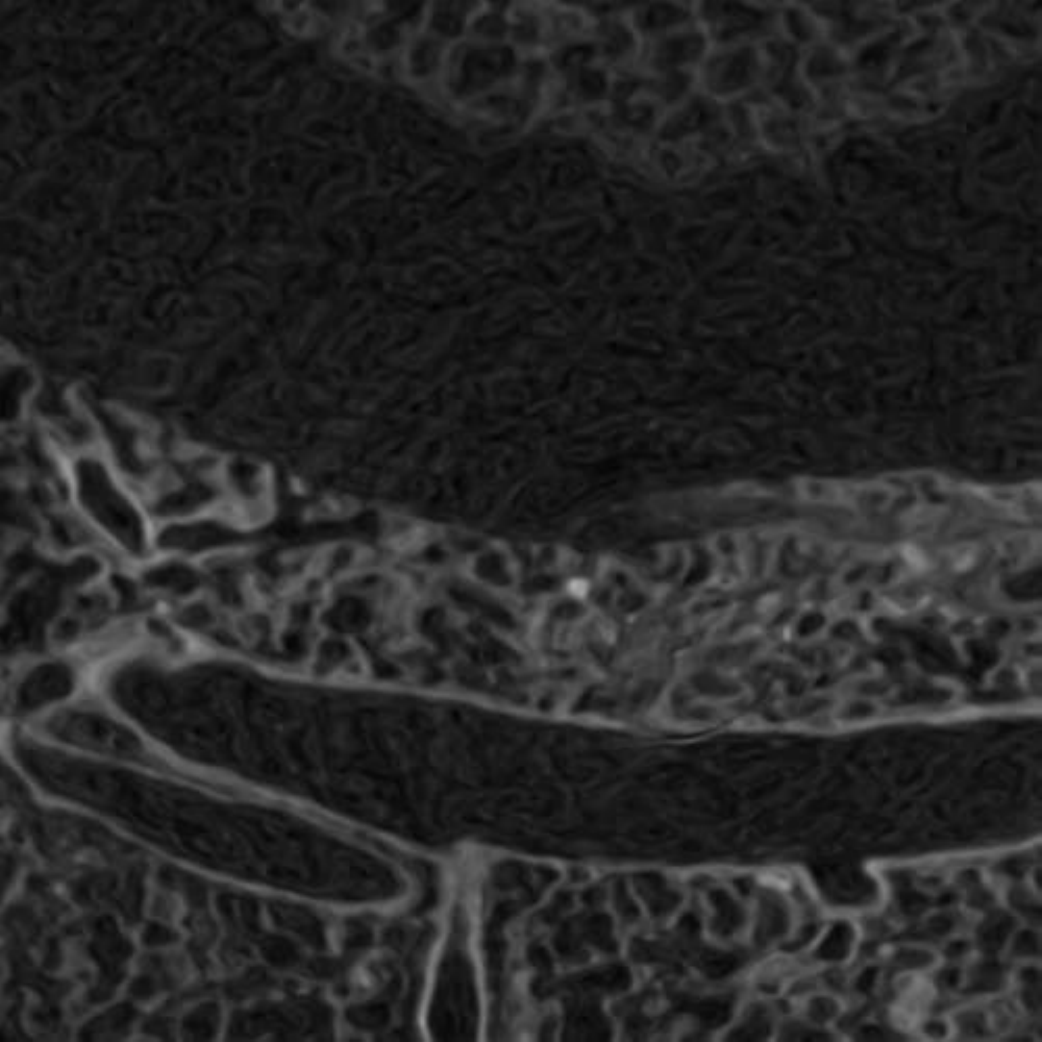}}{%
\includegraphics[width=\myfig]{image/scalebars/scale_ct-500.png}} &
\includegraphics[height=\myfig]{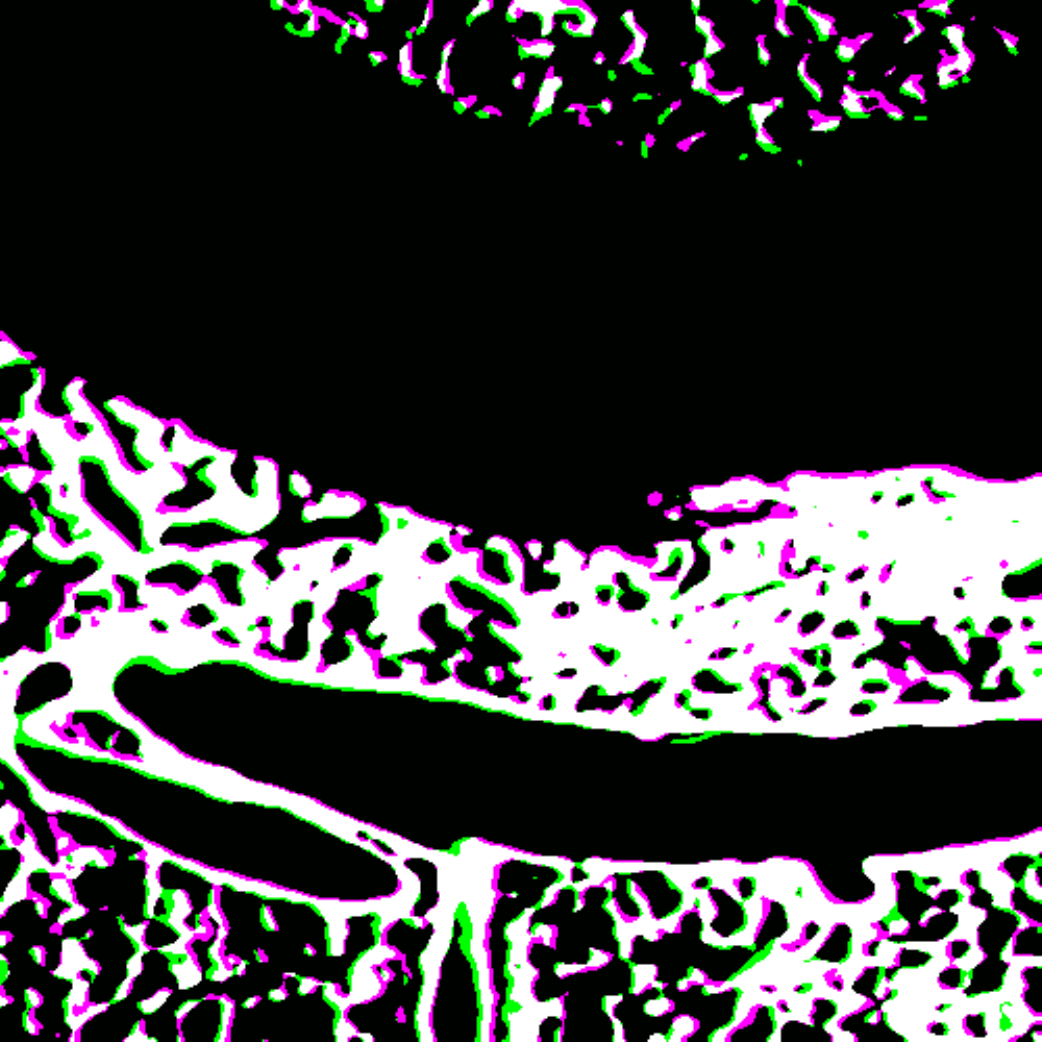} &
\includegraphics[height=\myfig]{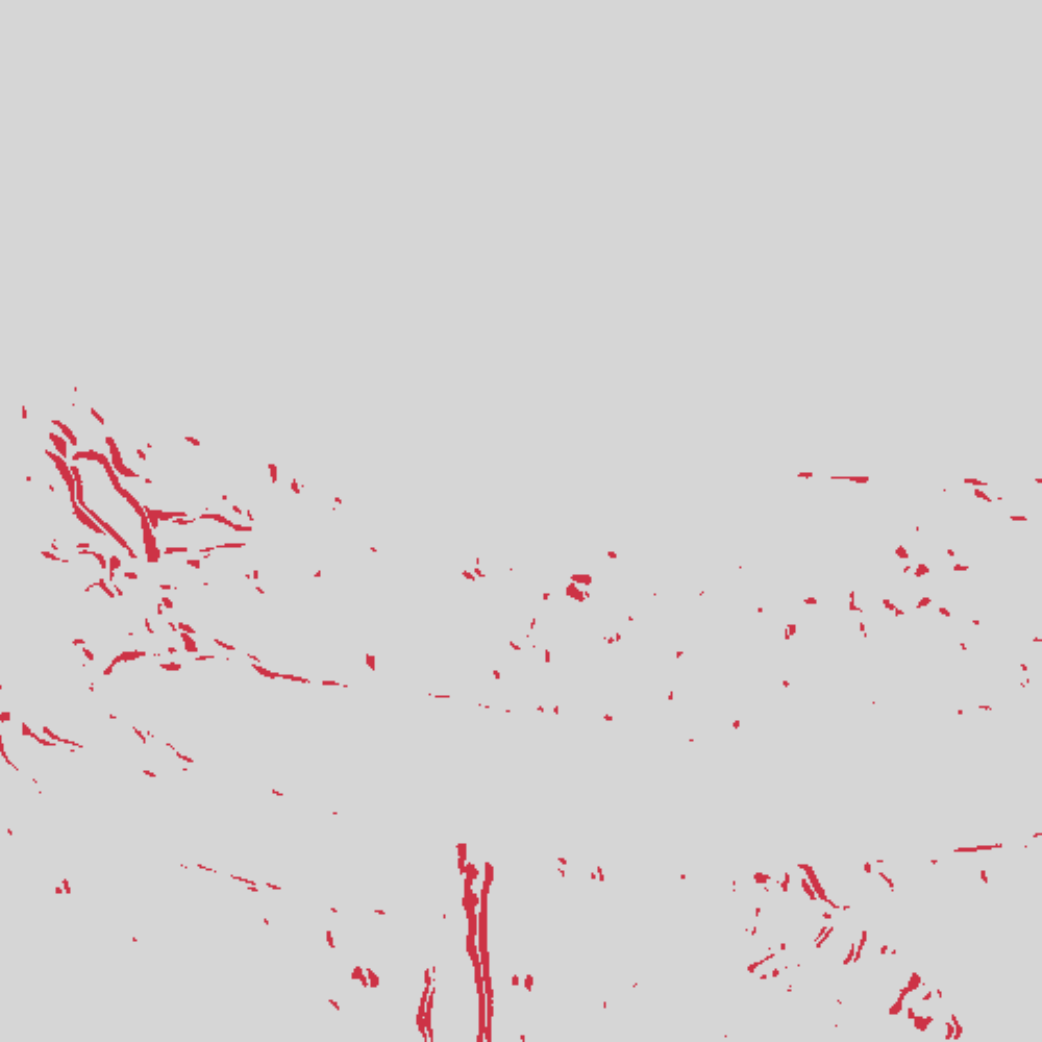} &
\includegraphics[height=\myfig]{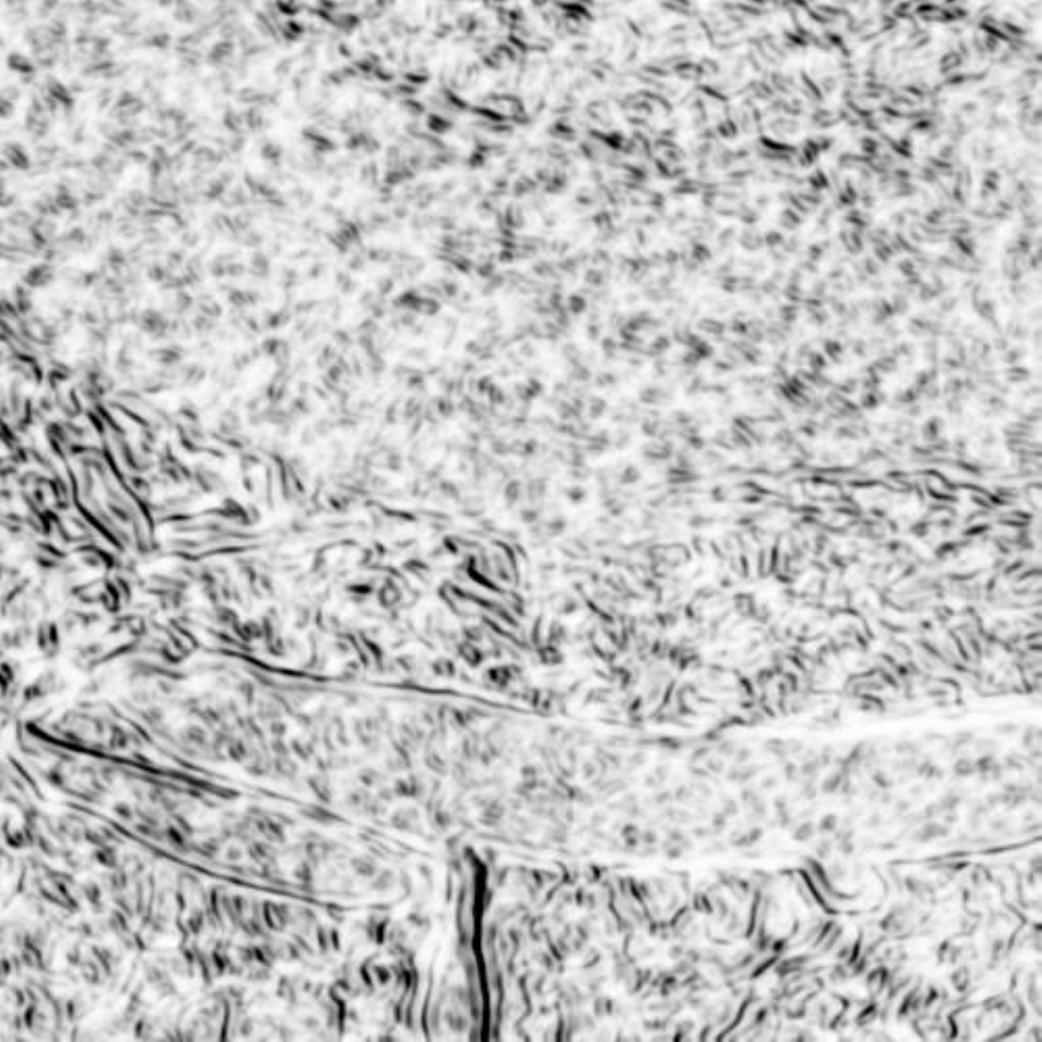} &
\includegraphics[height=\myfig]{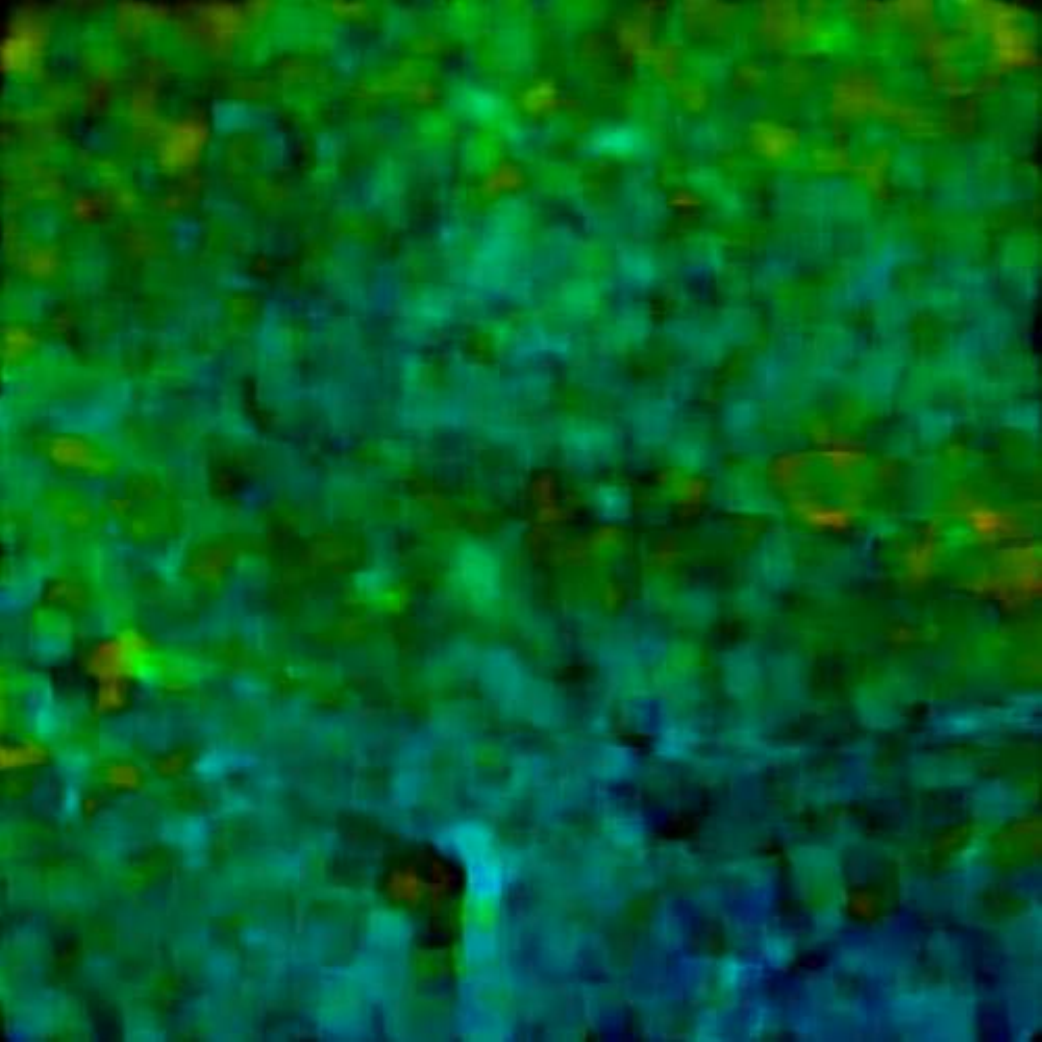}
\\

\rot{\myfig}{GS} &
\scalebarme{%
\includegraphics[height=\myfig]{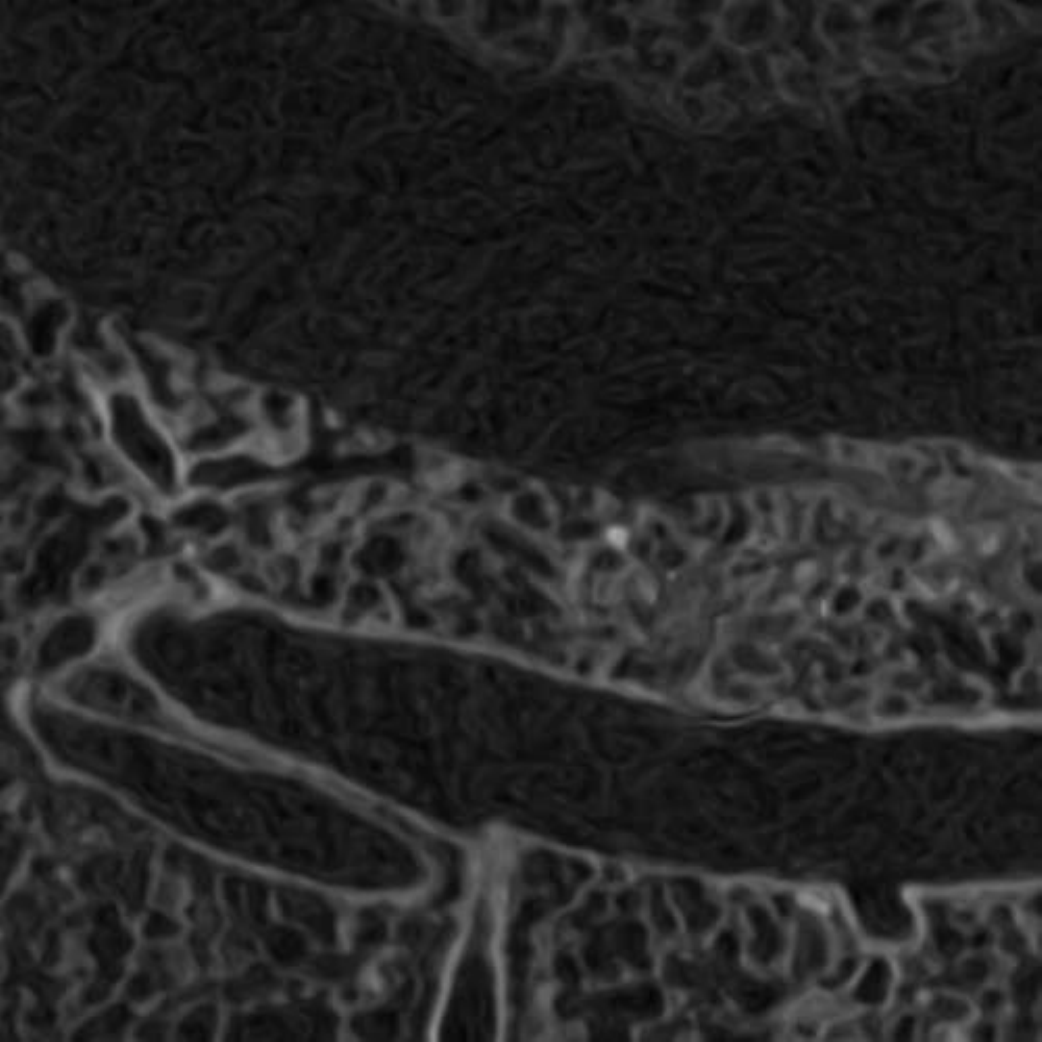}}{%
\includegraphics[width=\myfig]{image/scalebars/scale_ct-500.png}} &
\includegraphics[height=\myfig]{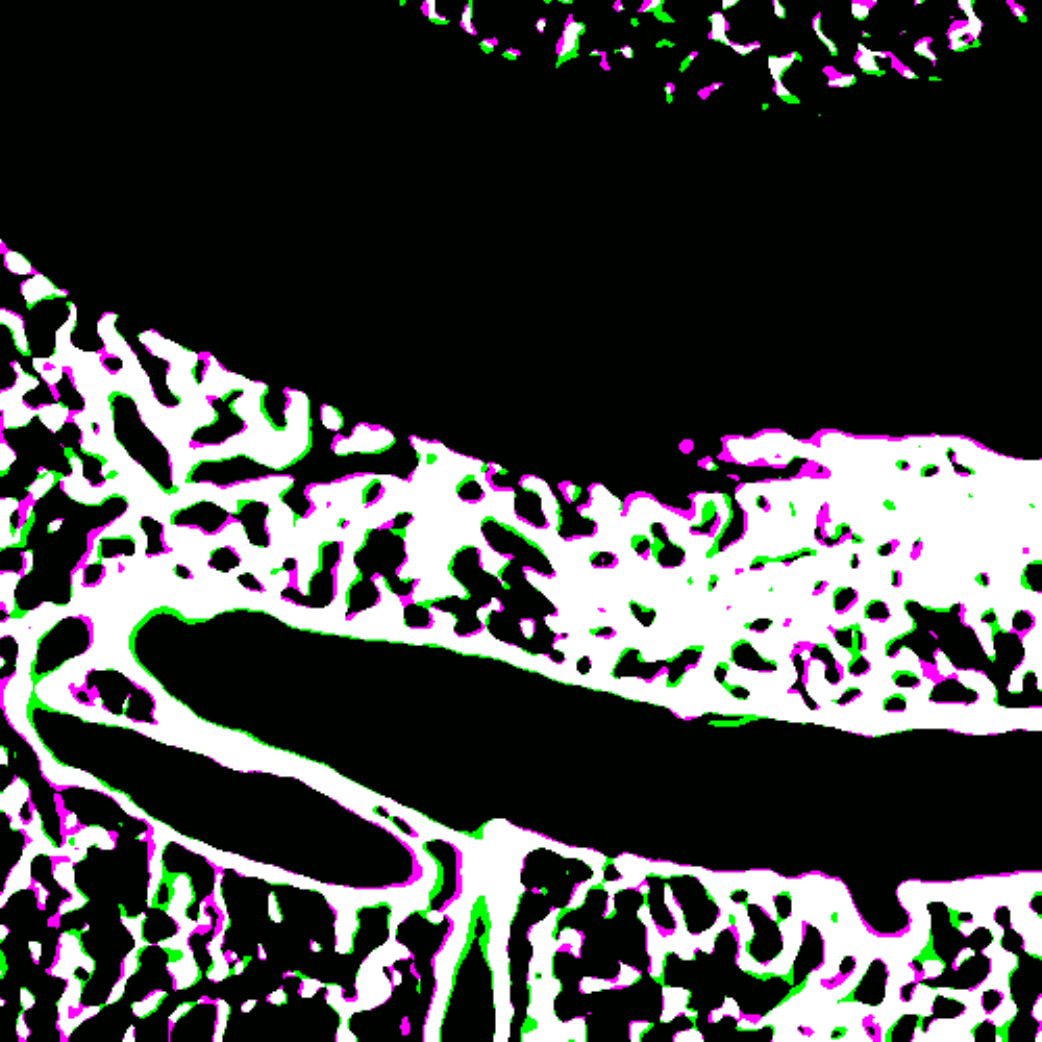} &
\includegraphics[height=\myfig]{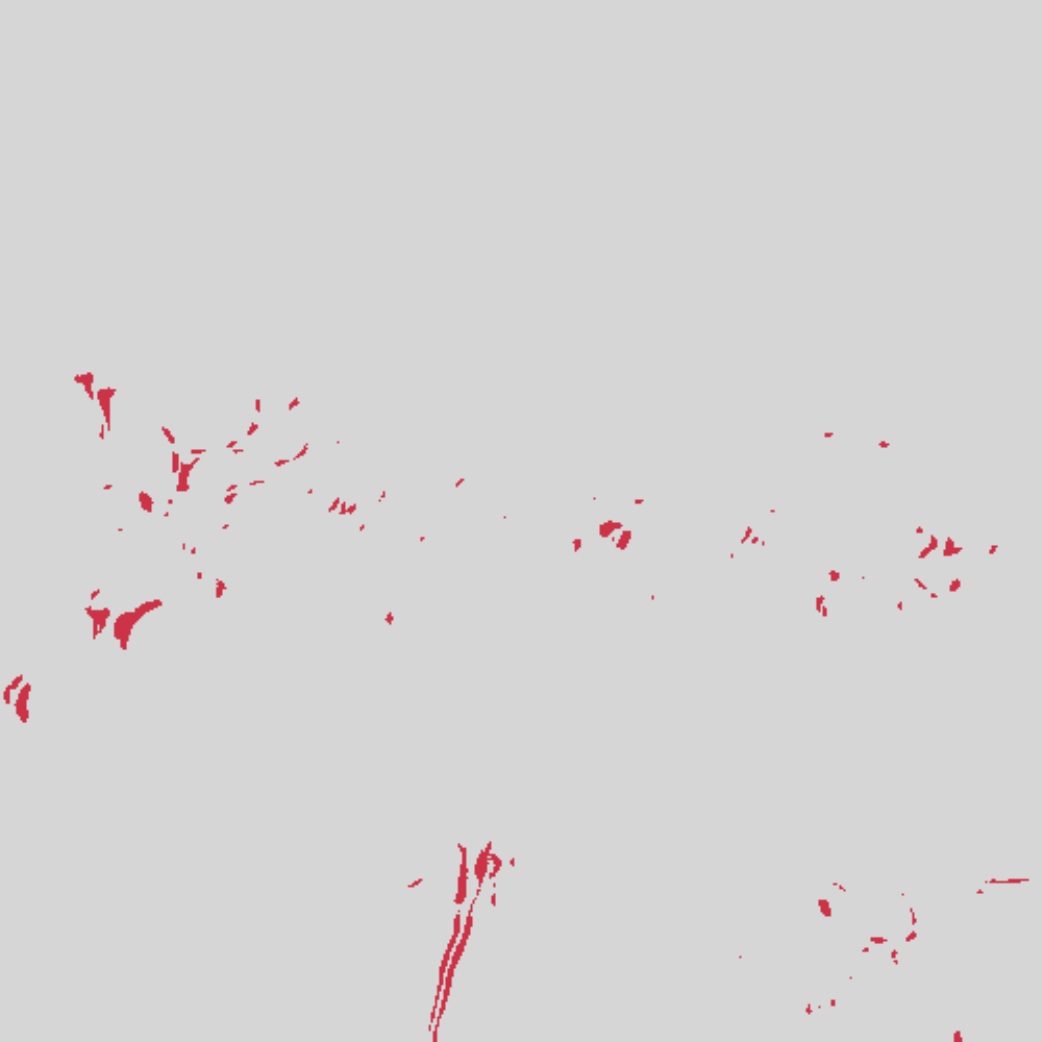} &
\includegraphics[height=\myfig]{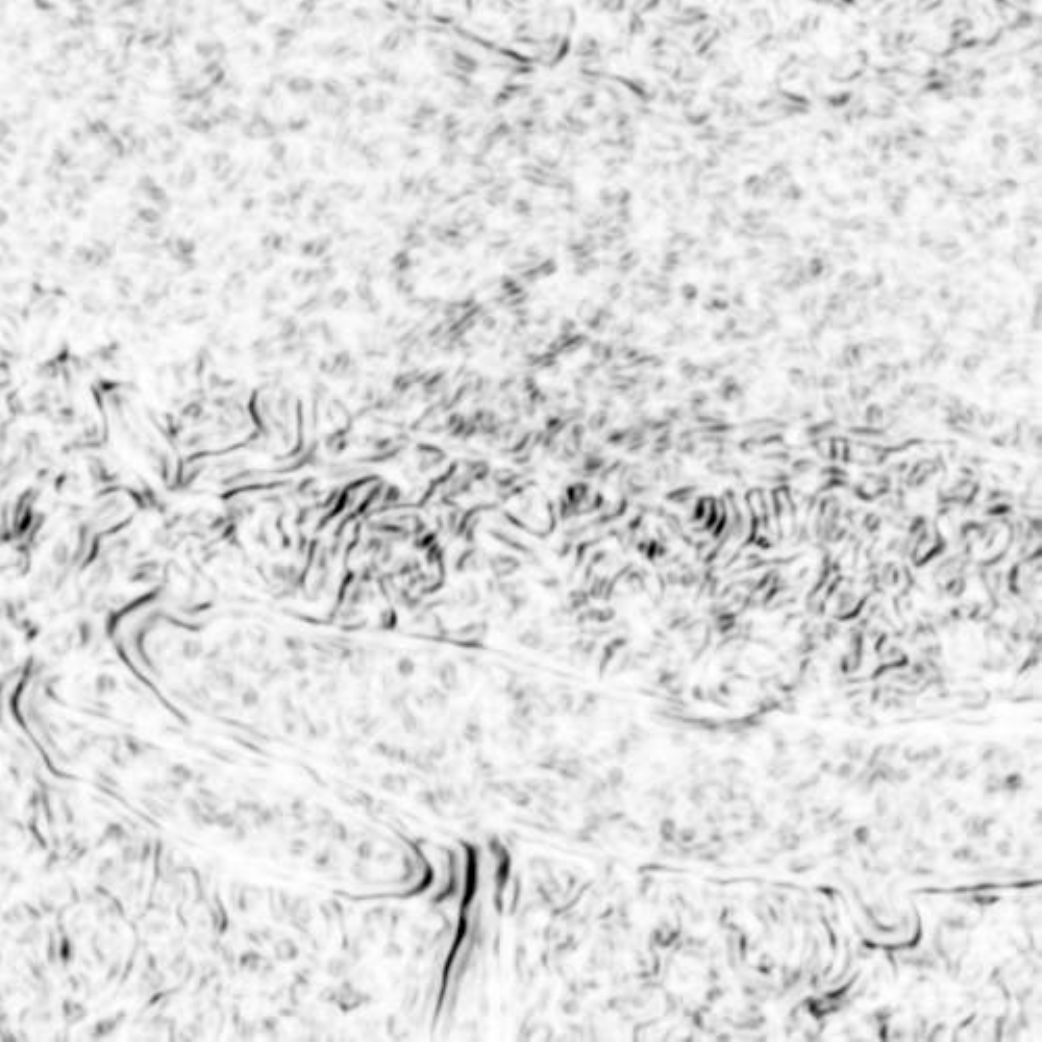} &
\includegraphics[height=\myfig]{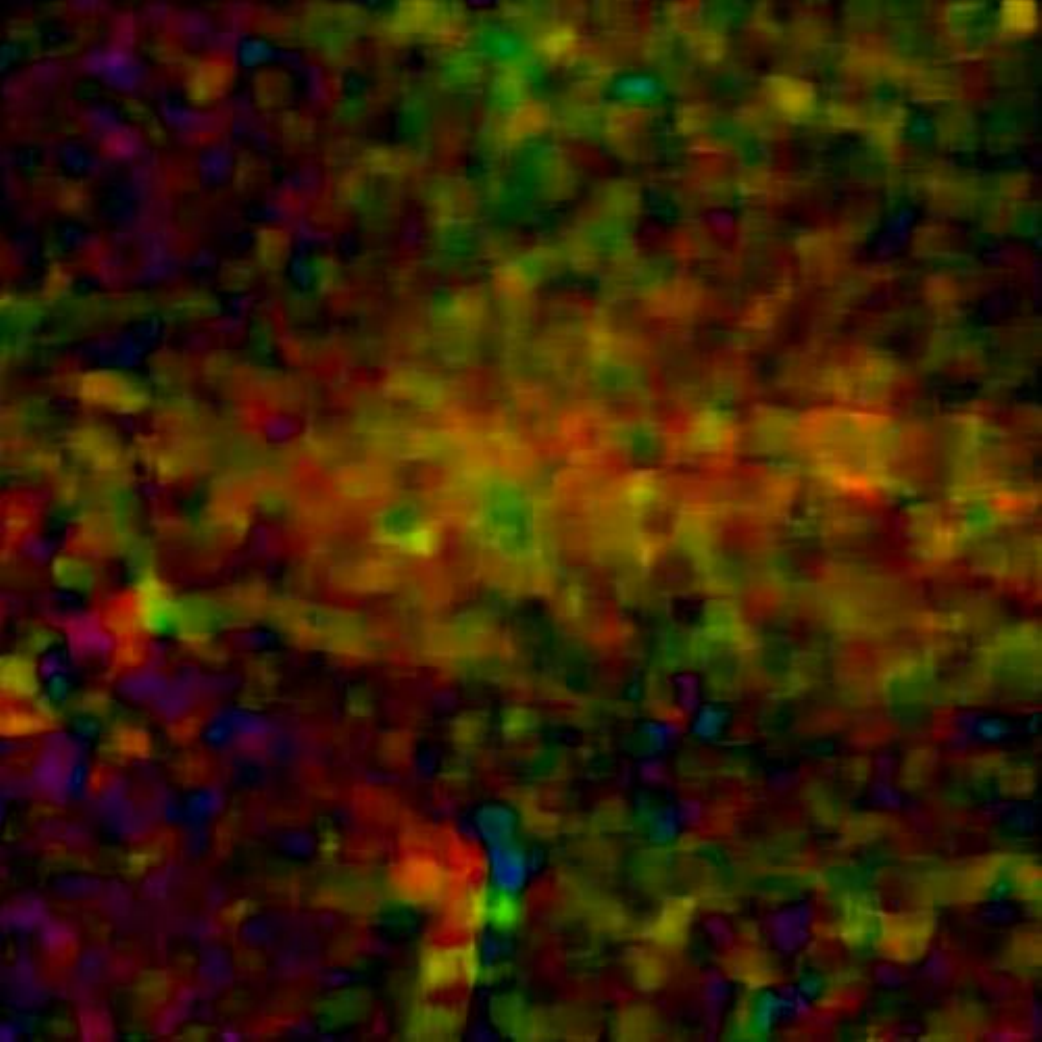} 
\\

\rot{\myfig}{Ground truth} &
\scalebarme{%
\includegraphics[height=\myfig]{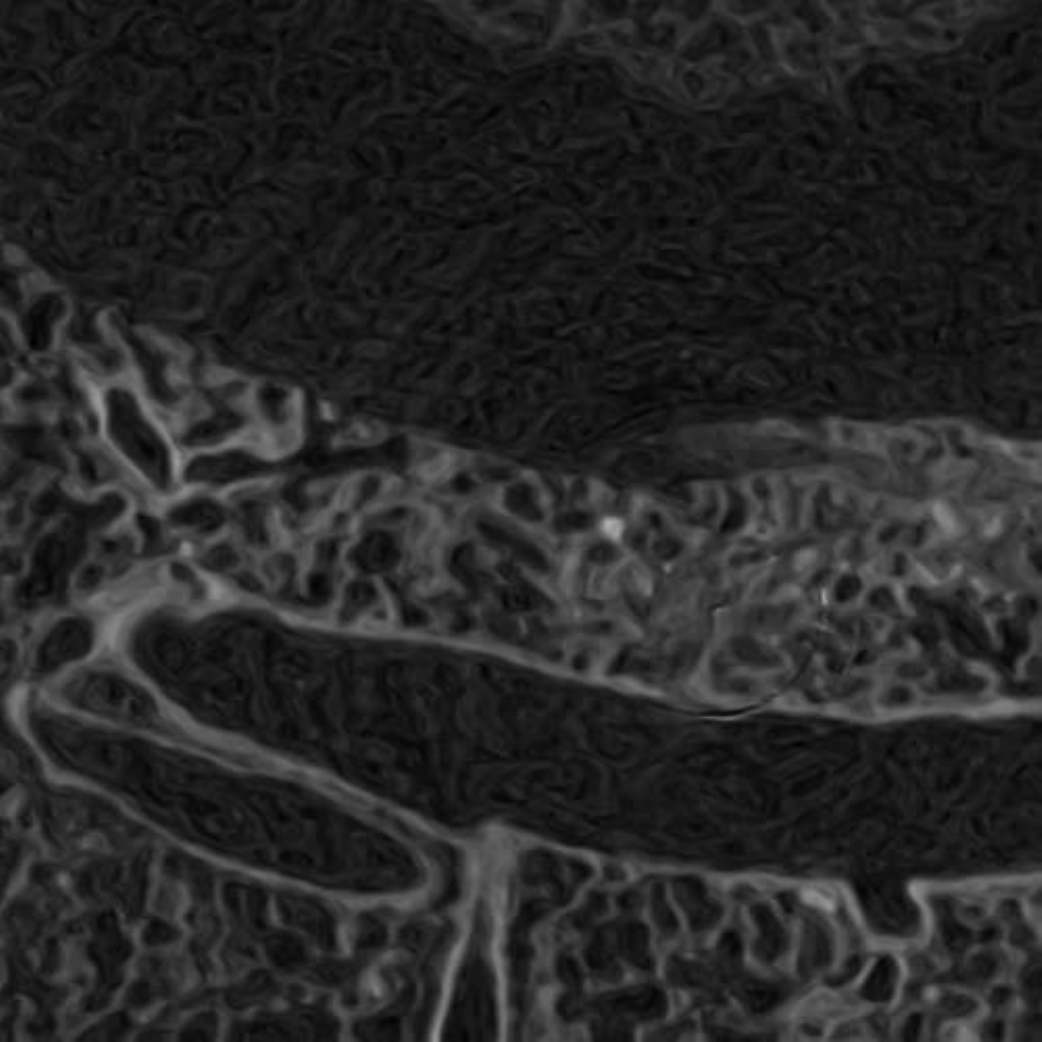}}{%
\includegraphics[width=\myfig]{image/scalebars/scale_ct-500.png}} &
\includegraphics[height=\myfig]{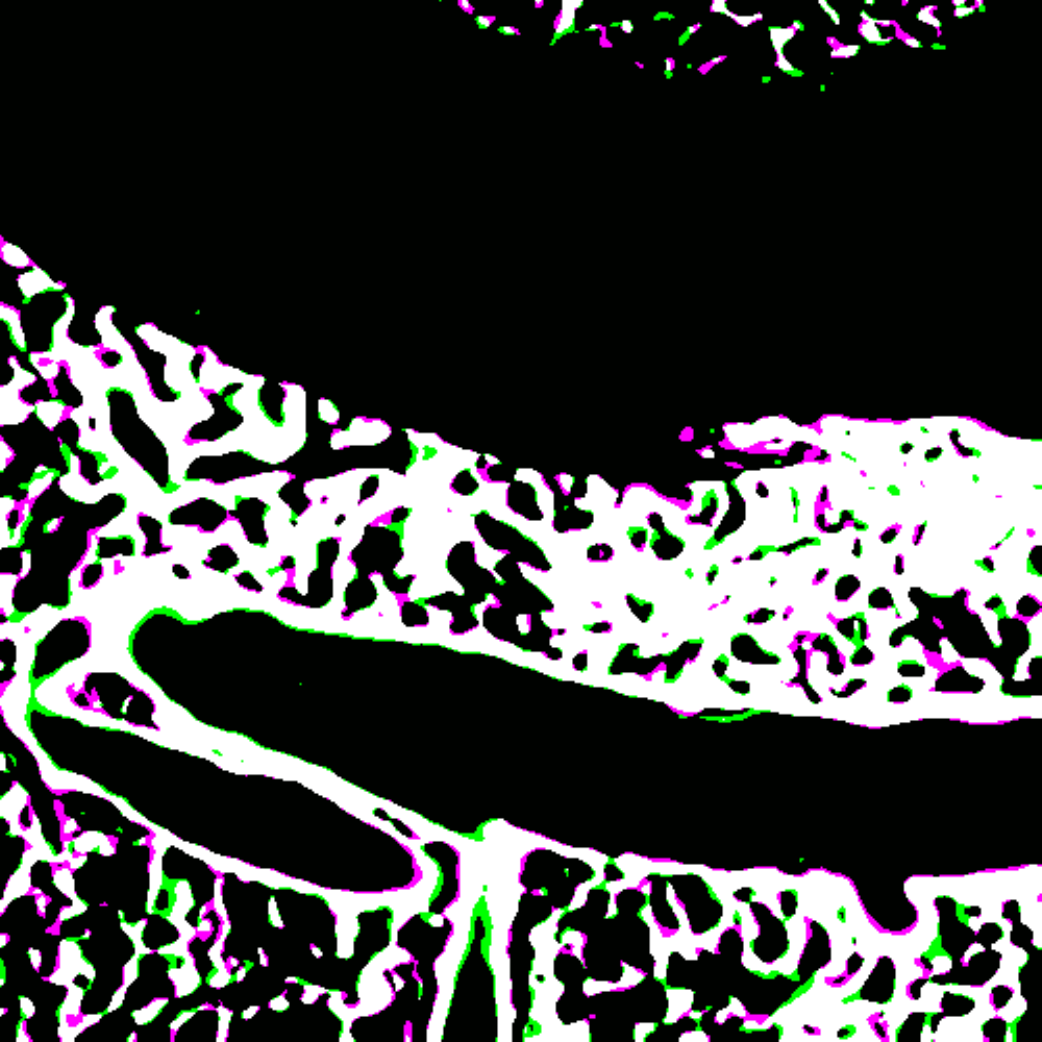} &
\includegraphics[height=\myfig]{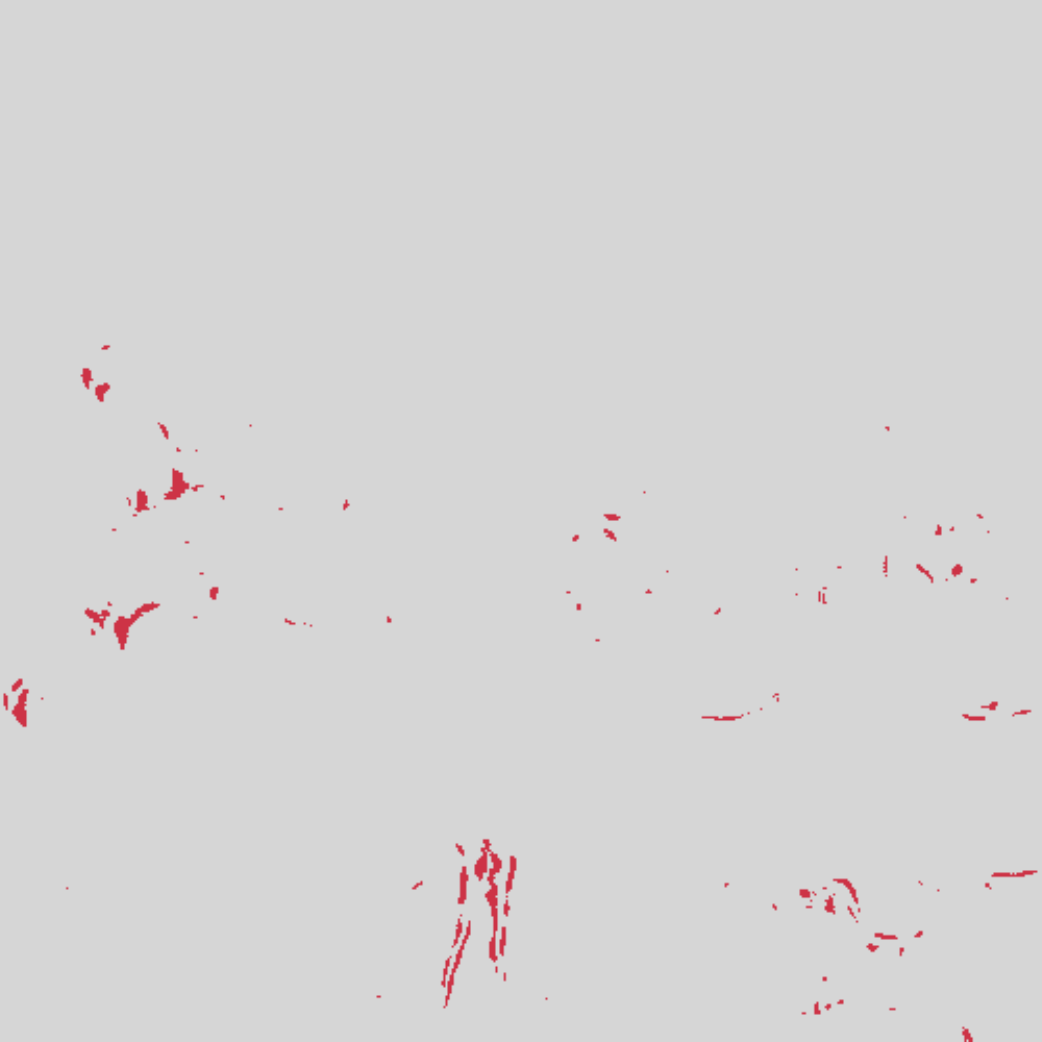} &
\includegraphics[height=\myfig]{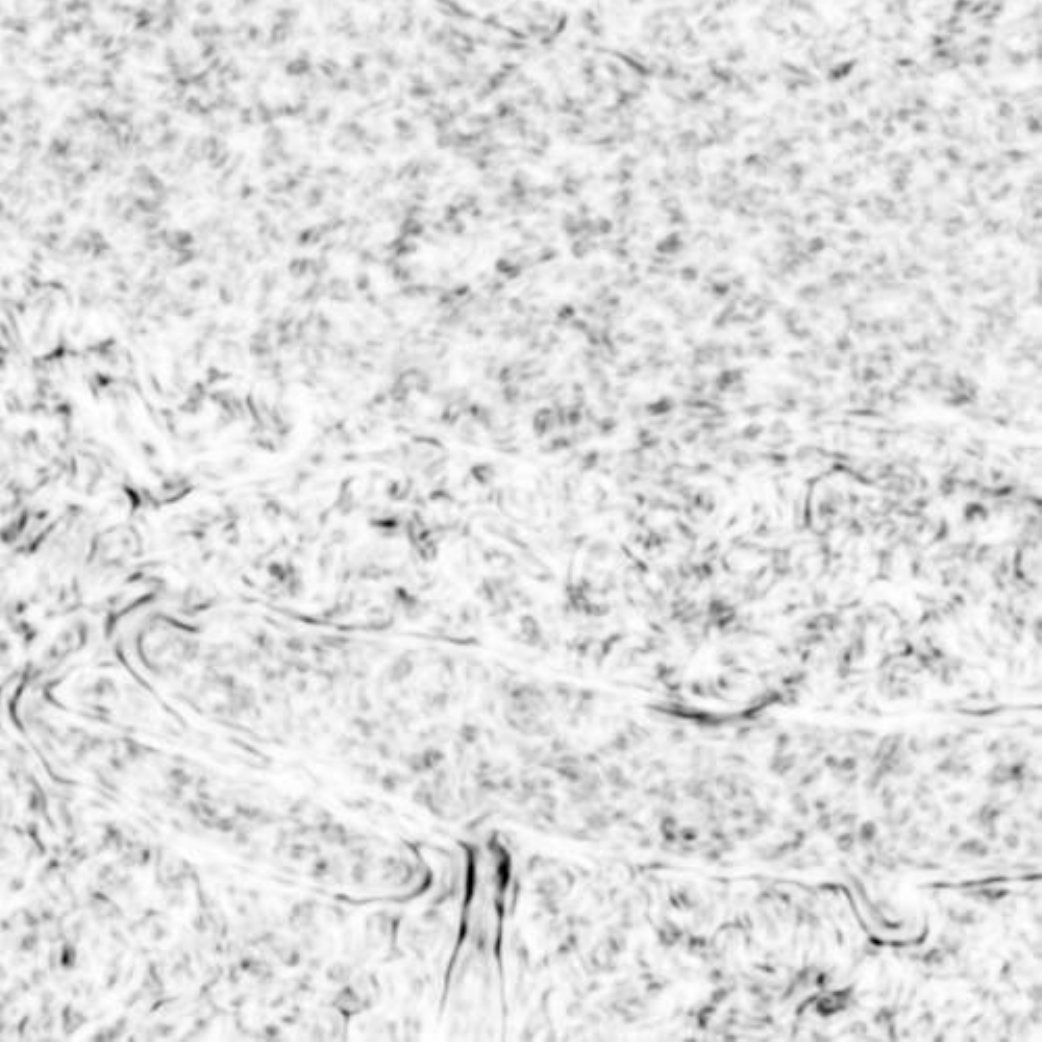} &
\includegraphics[height=\myfig]{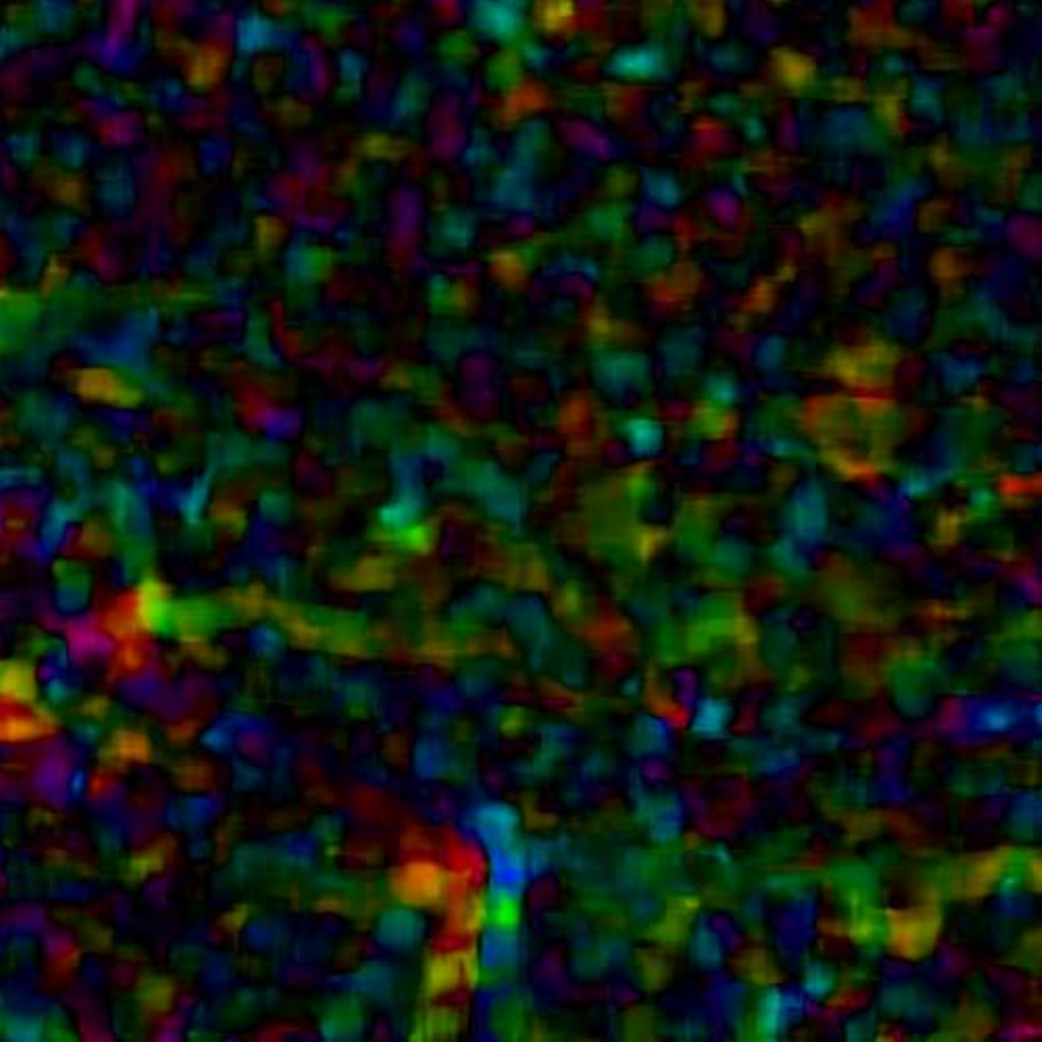}
\\

& \multicolumn{1}{c}{Image} & \multicolumn{1}{c}{Dice} & \multicolumn{1}{c}{PSNR} & \multicolumn{1}{c}{SSIM} & \multicolumn{1}{c}{Flow} \\
\end{tabular}

\caption{Evaluation of CT data set with image measures, part~1. Continued in Fig.~\ref{fig:eval-ct-2}. All scale bars are \SI{1}{\milli\meter}.}
\label{fig:eval-ct-1}
\setlength{\tabcolsep}{\tabcolbackup}
\end{figure*}
\begin{figure*}[!t]
\centering
\setlength{\myfig}{0.2\linewidth-0.001\linewidth-2pt-0.2em}
\setlength{\tabcolsep}{1pt}
\begin{tabular}{m{1em} m{\myfig} m{\myfig} m{\myfig} m{\myfig} m{\myfig}}
& \multicolumn{1}{c}{Image} & \multicolumn{1}{c}{Dice} & \multicolumn{1}{c}{PSNR} & \multicolumn{1}{c}{SSIM} & \multicolumn{1}{c}{Flow} \\

\rot{\myfig}{Blending} &
\scalebarme{%
\includegraphics[height=\myfig]{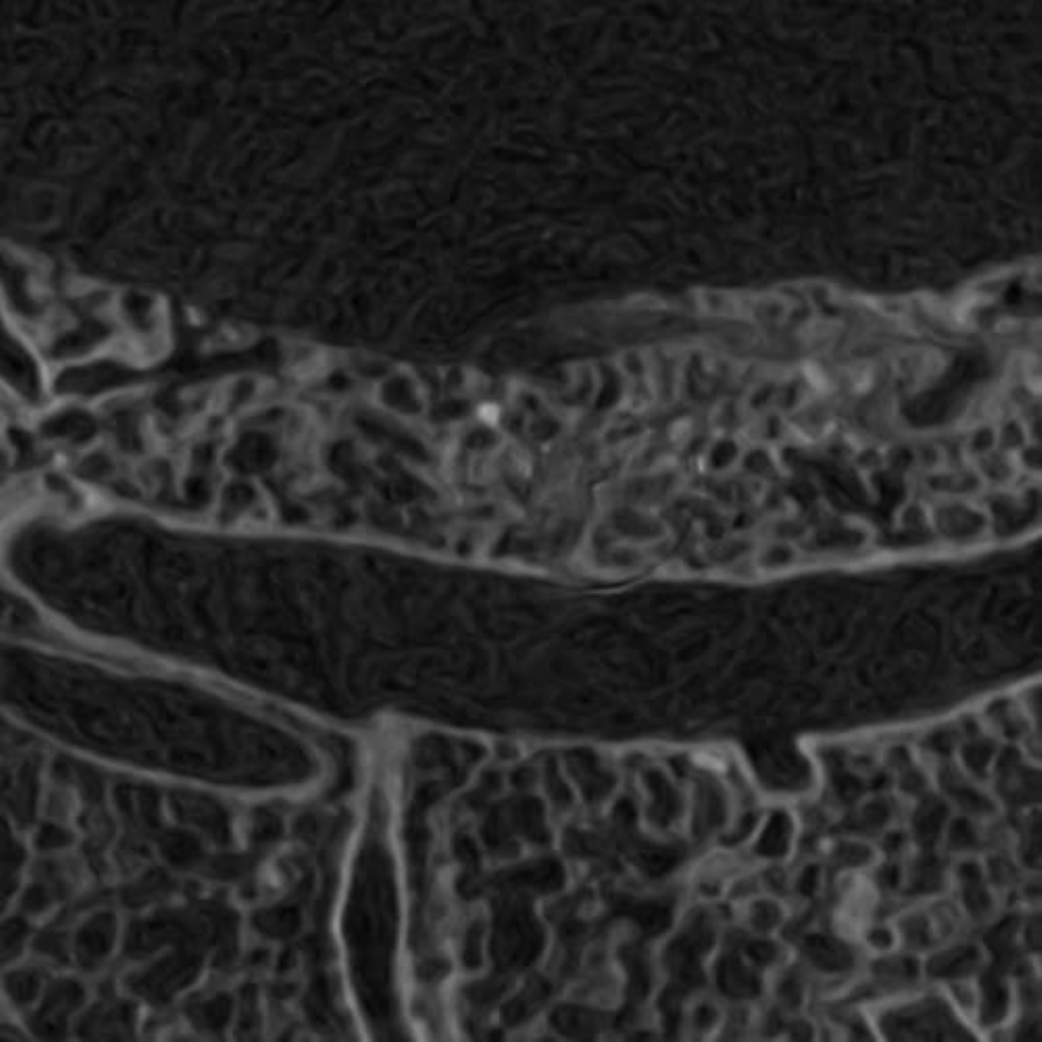}}{%
\includegraphics[width=\myfig]{image/scalebars/scale_ct-500.png}} &
\includegraphics[height=\myfig]{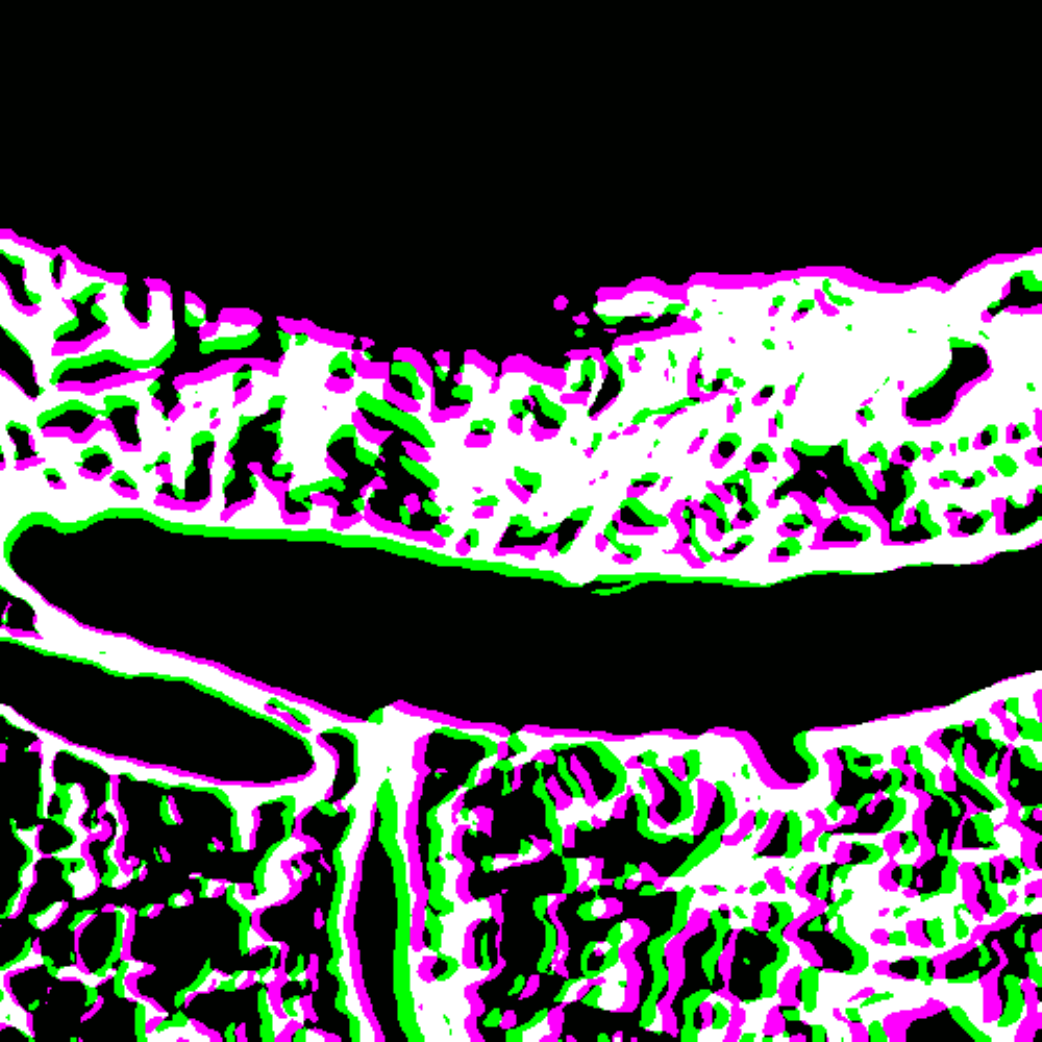} &
\includegraphics[height=\myfig]{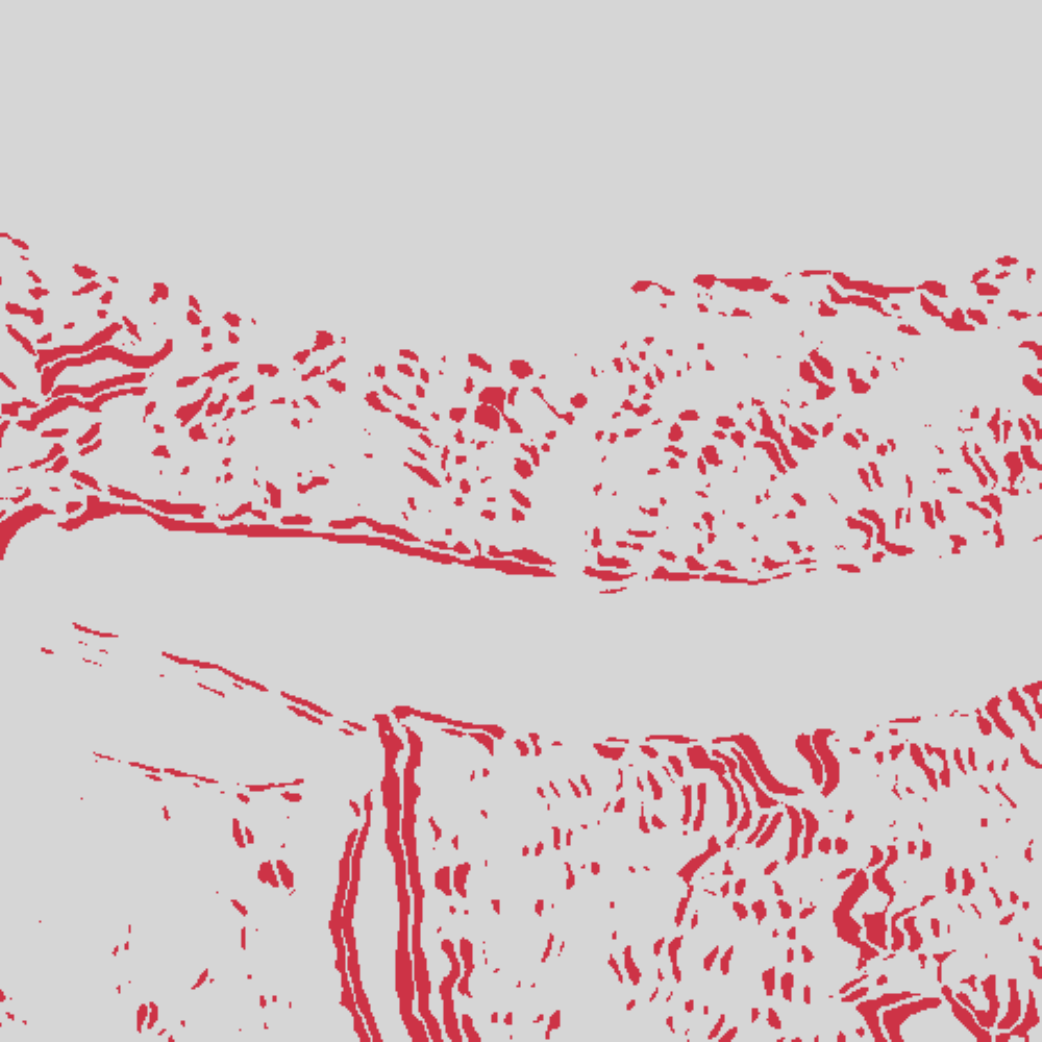} &
\includegraphics[height=\myfig]{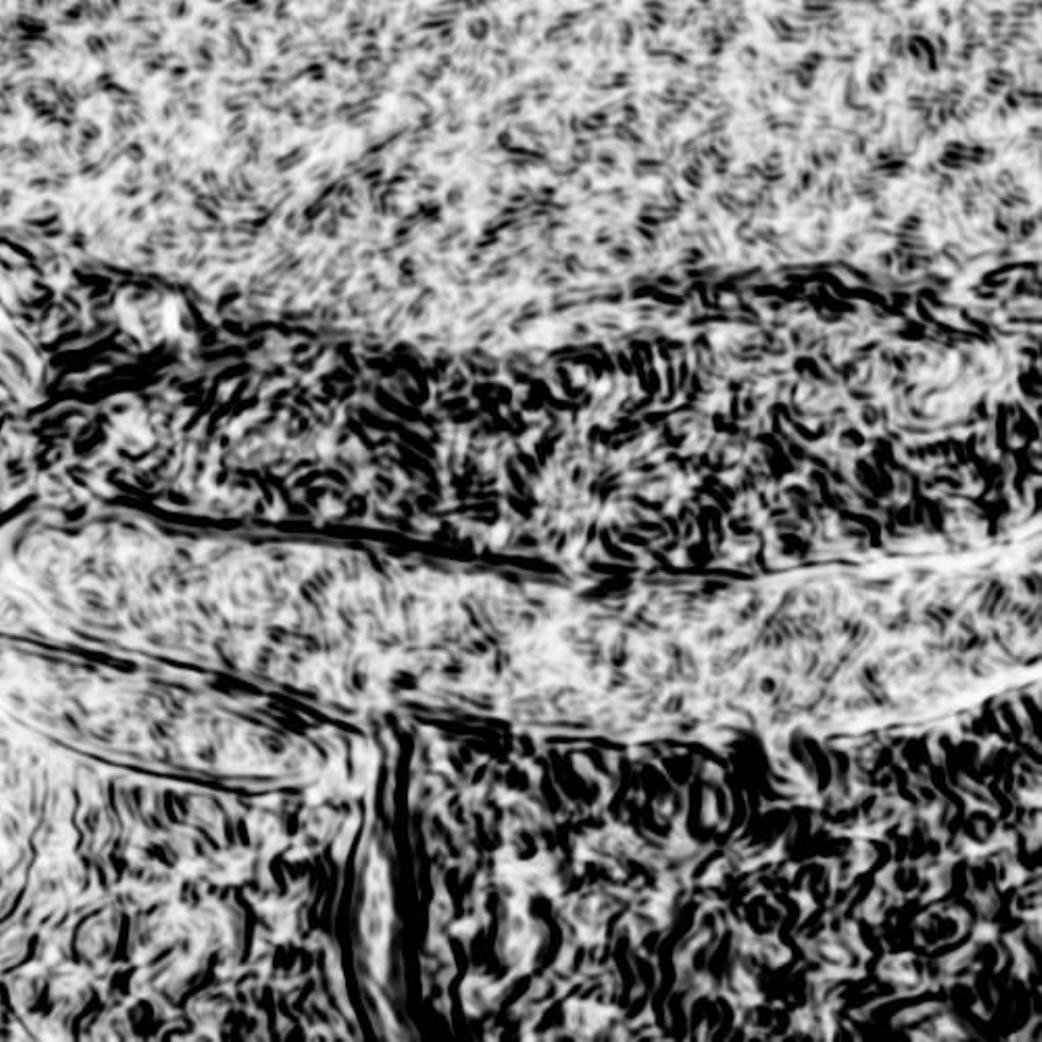} &
\includegraphics[height=\myfig]{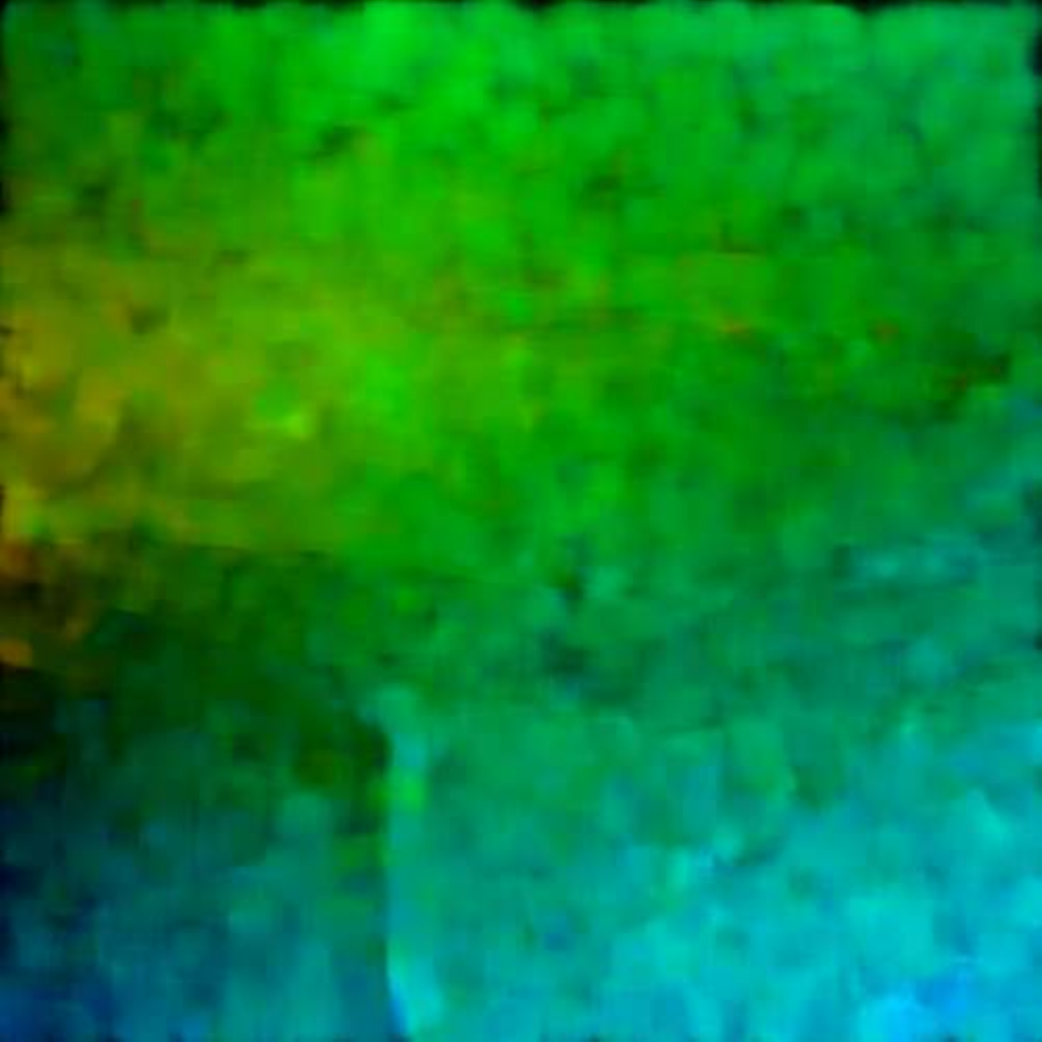}
\\

\rot{\myfig}{\elastix} &
\scalebarme{%
\includegraphics[height=\myfig]{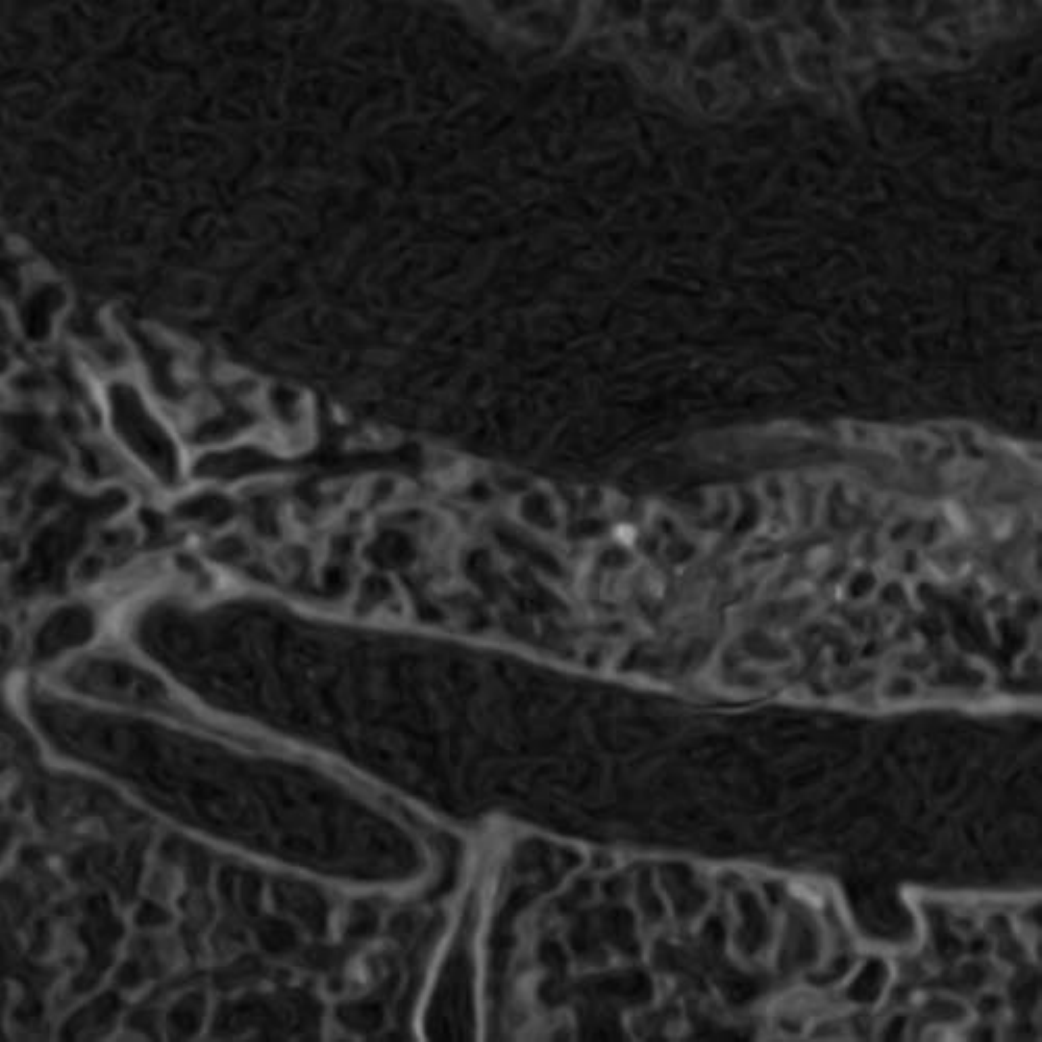}}{%
\includegraphics[width=\myfig]{image/scalebars/scale_ct-500.png}} &
\includegraphics[height=\myfig]{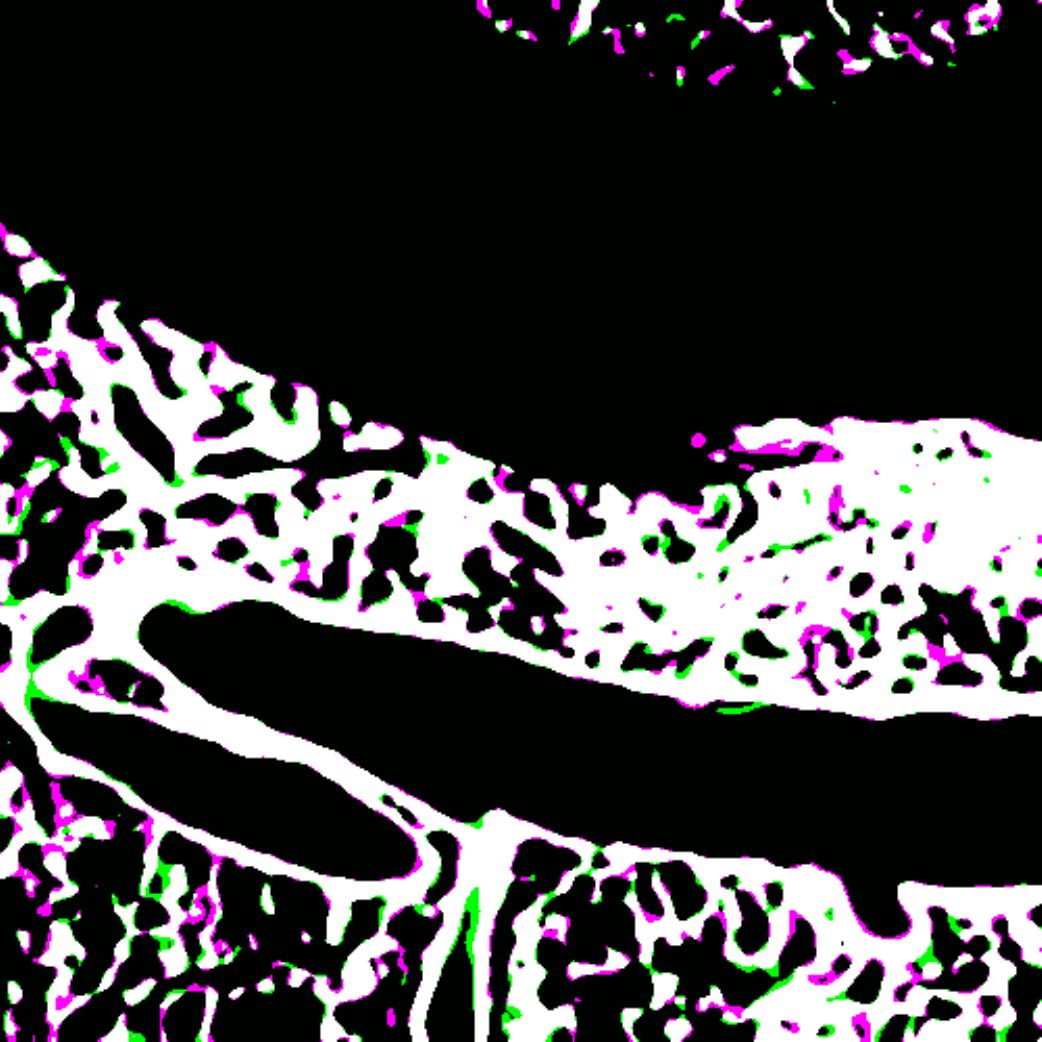} &
\includegraphics[height=\myfig]{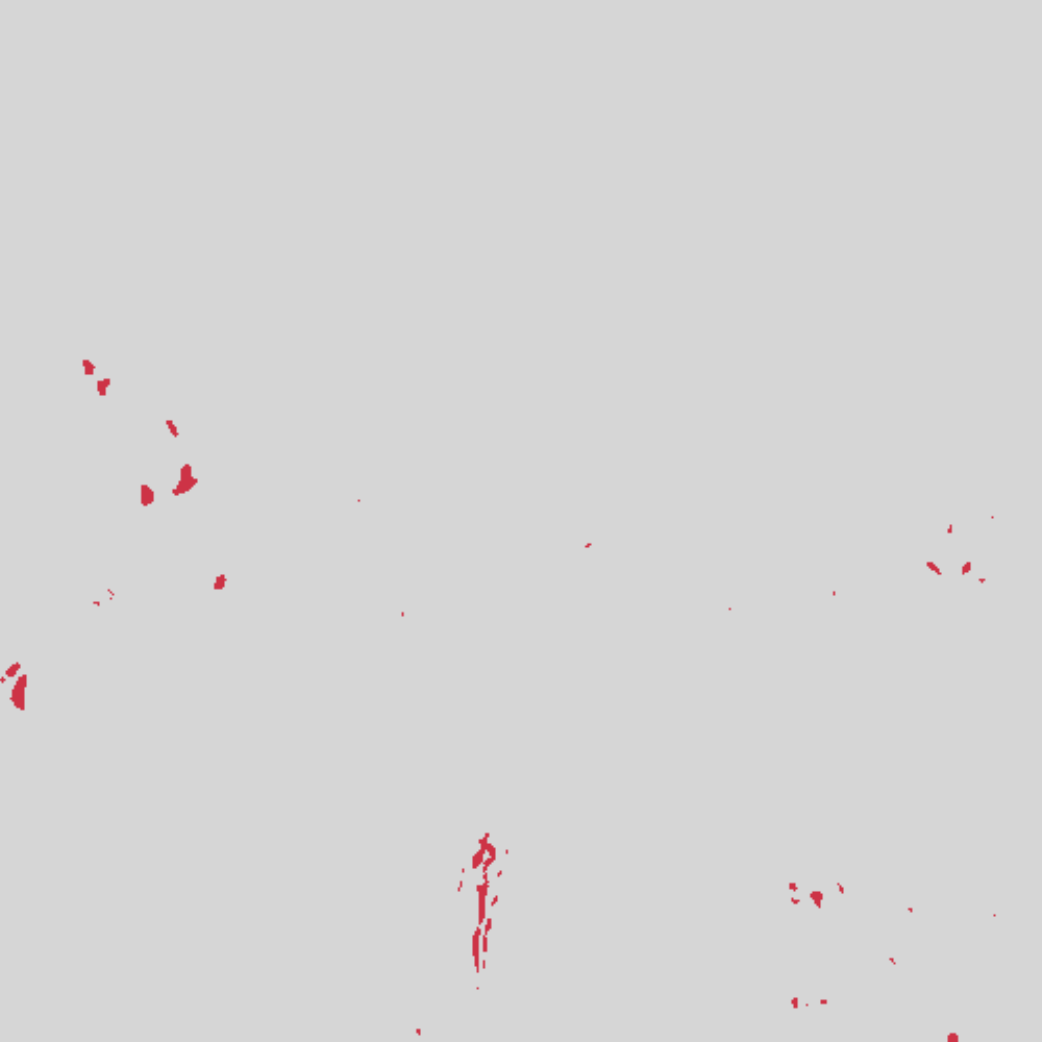} &
\includegraphics[height=\myfig]{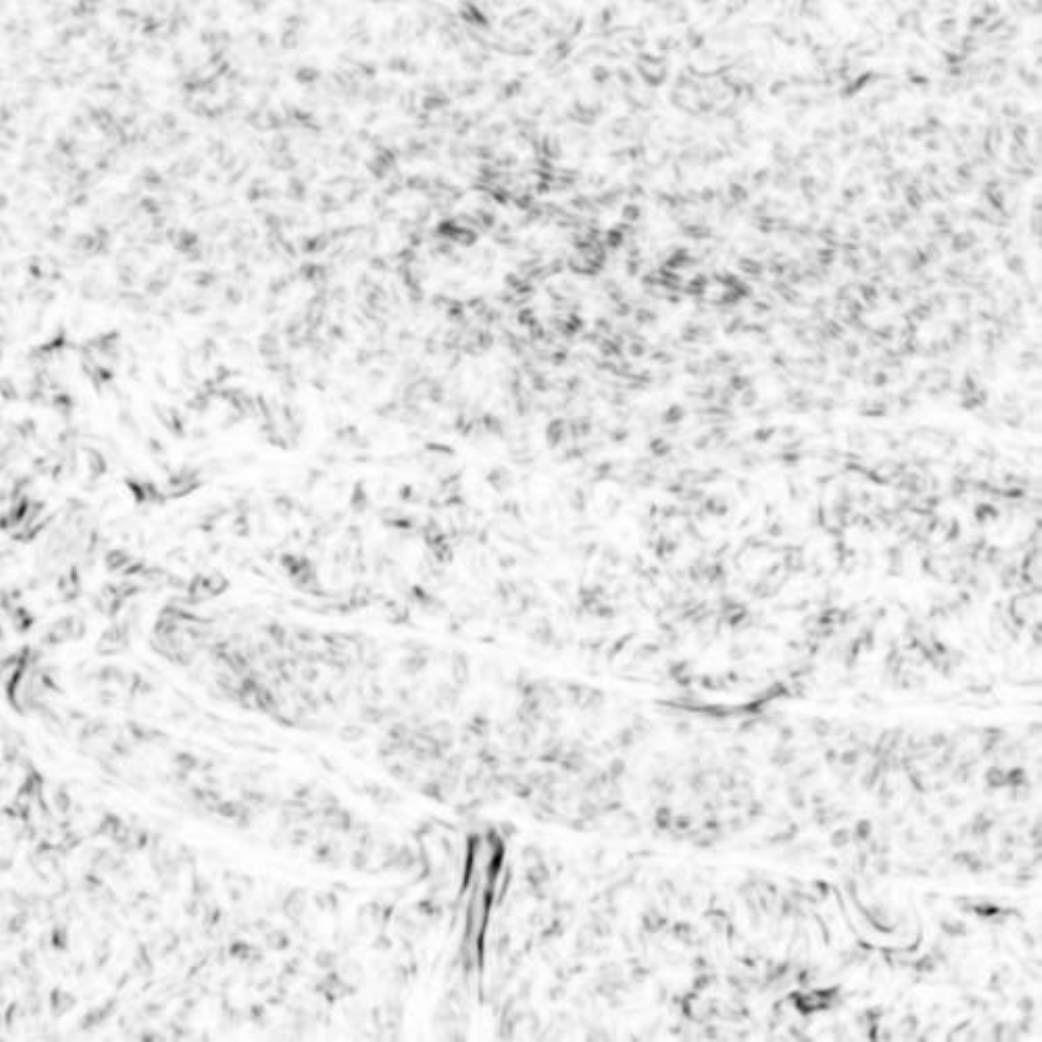} &
\includegraphics[height=\myfig]{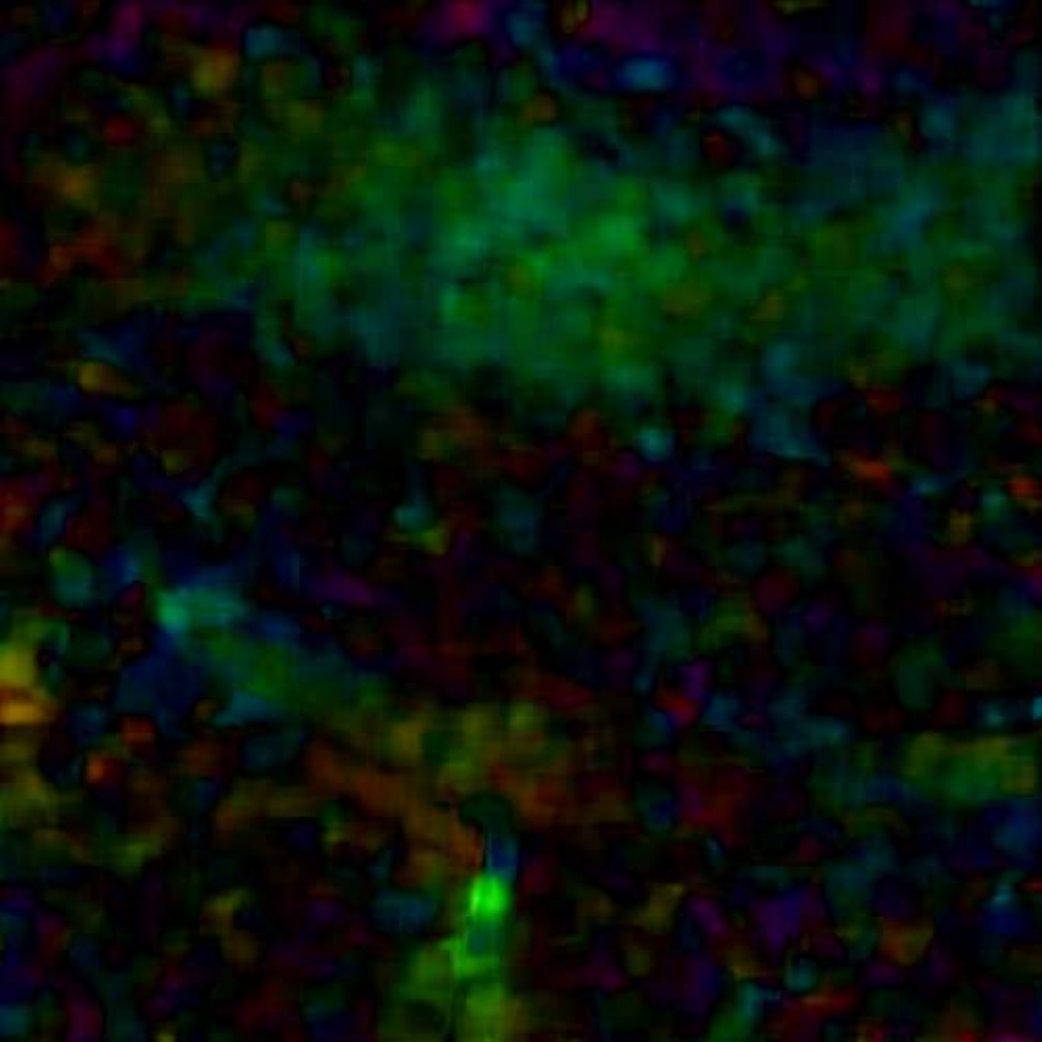}
\\

\rot{\myfig}{Local only} &
\scalebarme{%
\includegraphics[height=\myfig]{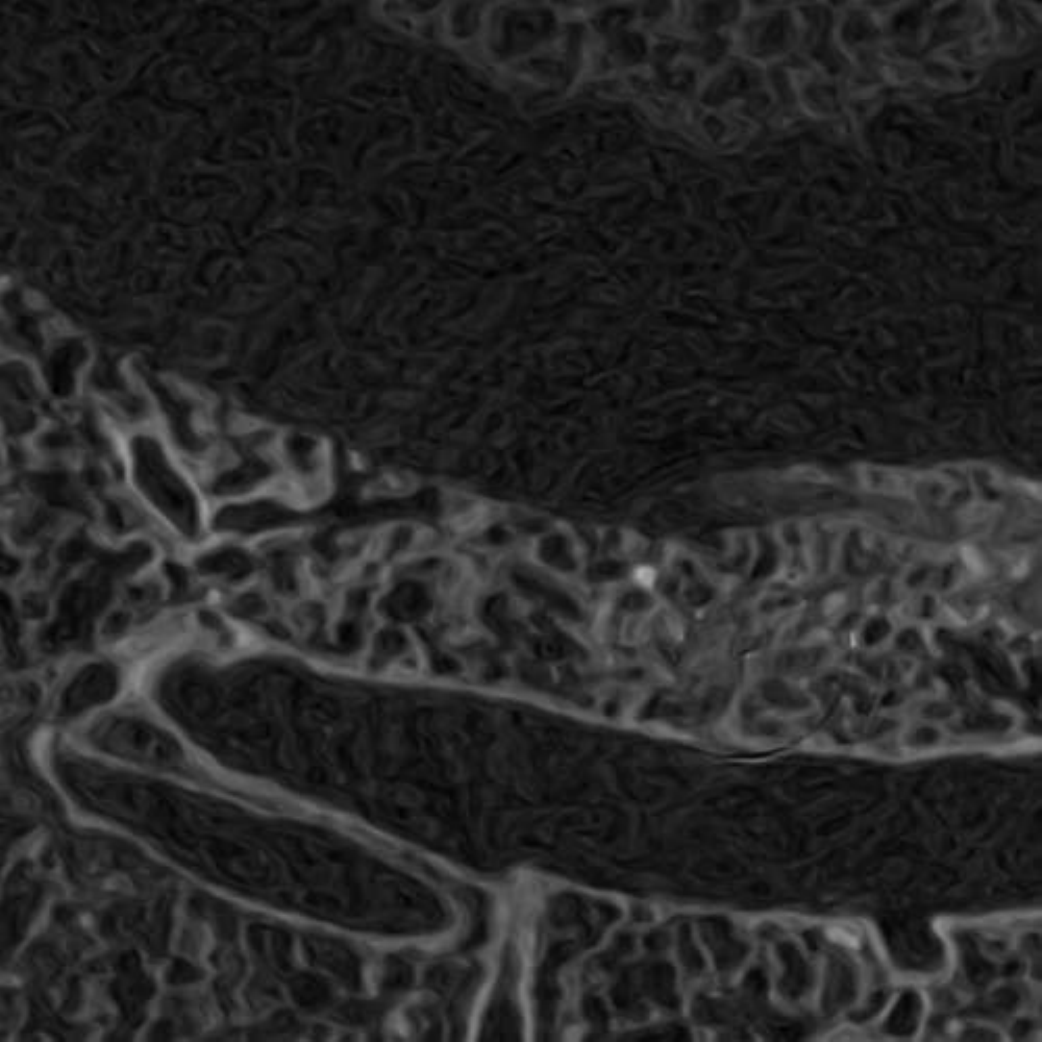}}{%
\includegraphics[width=\myfig]{image/scalebars/scale_ct-500.png}} &
\includegraphics[height=\myfig]{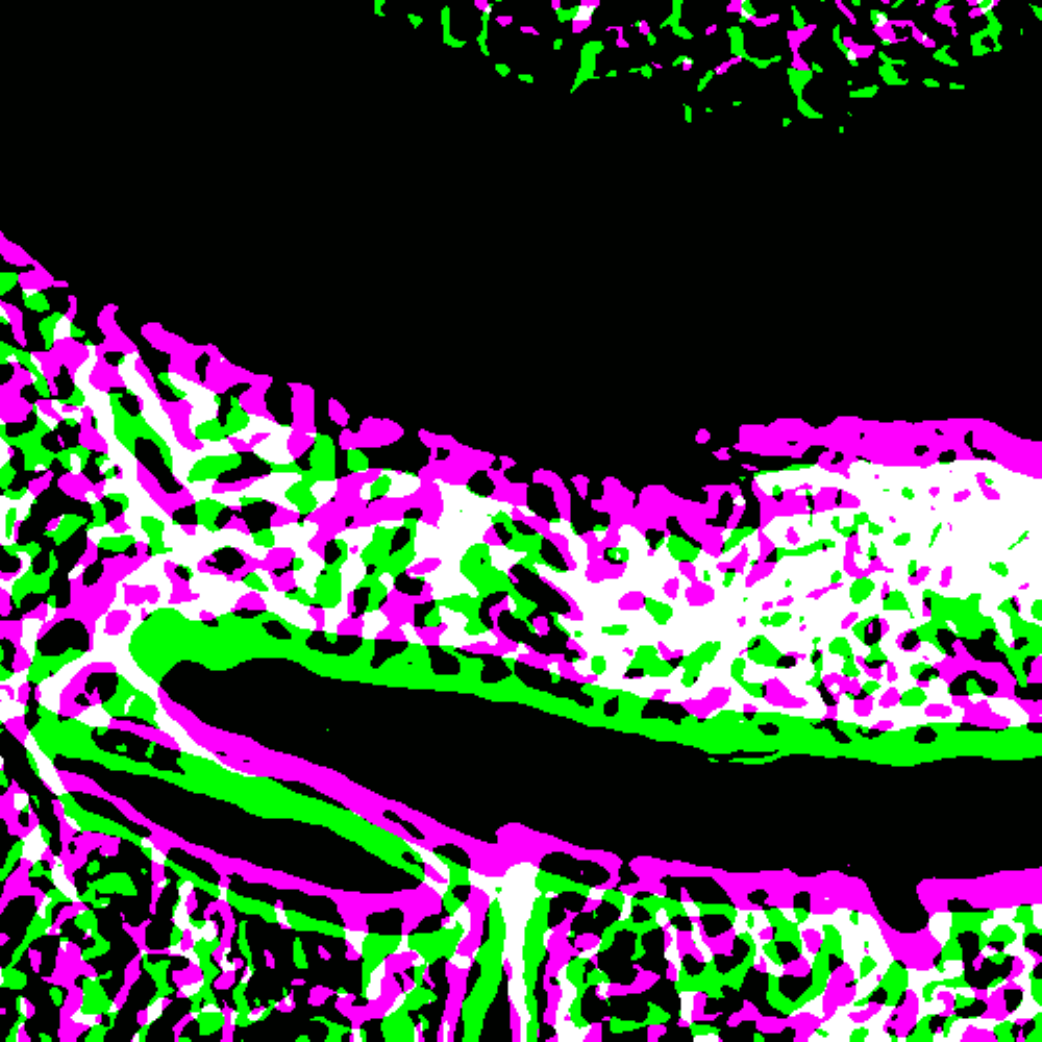} &
\includegraphics[height=\myfig]{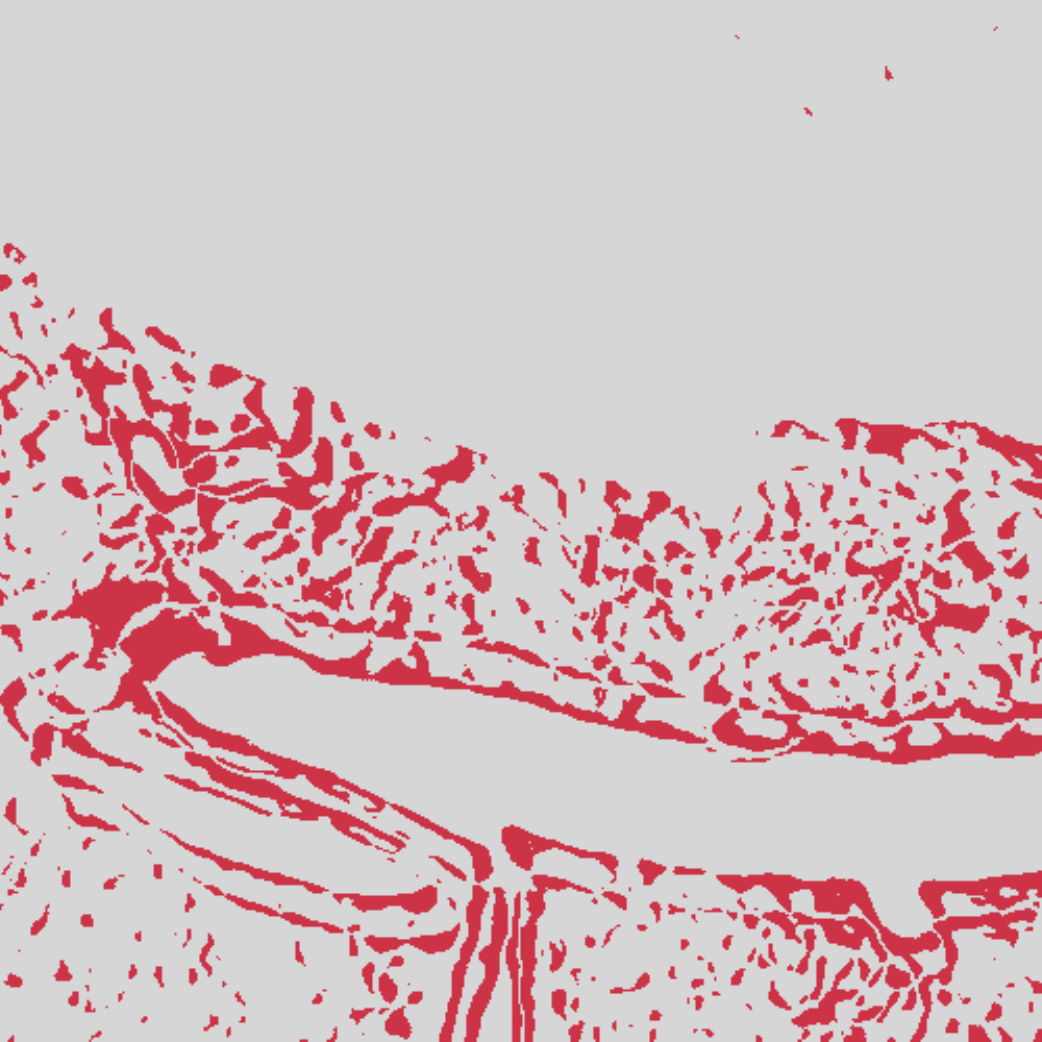} &
\includegraphics[height=\myfig]{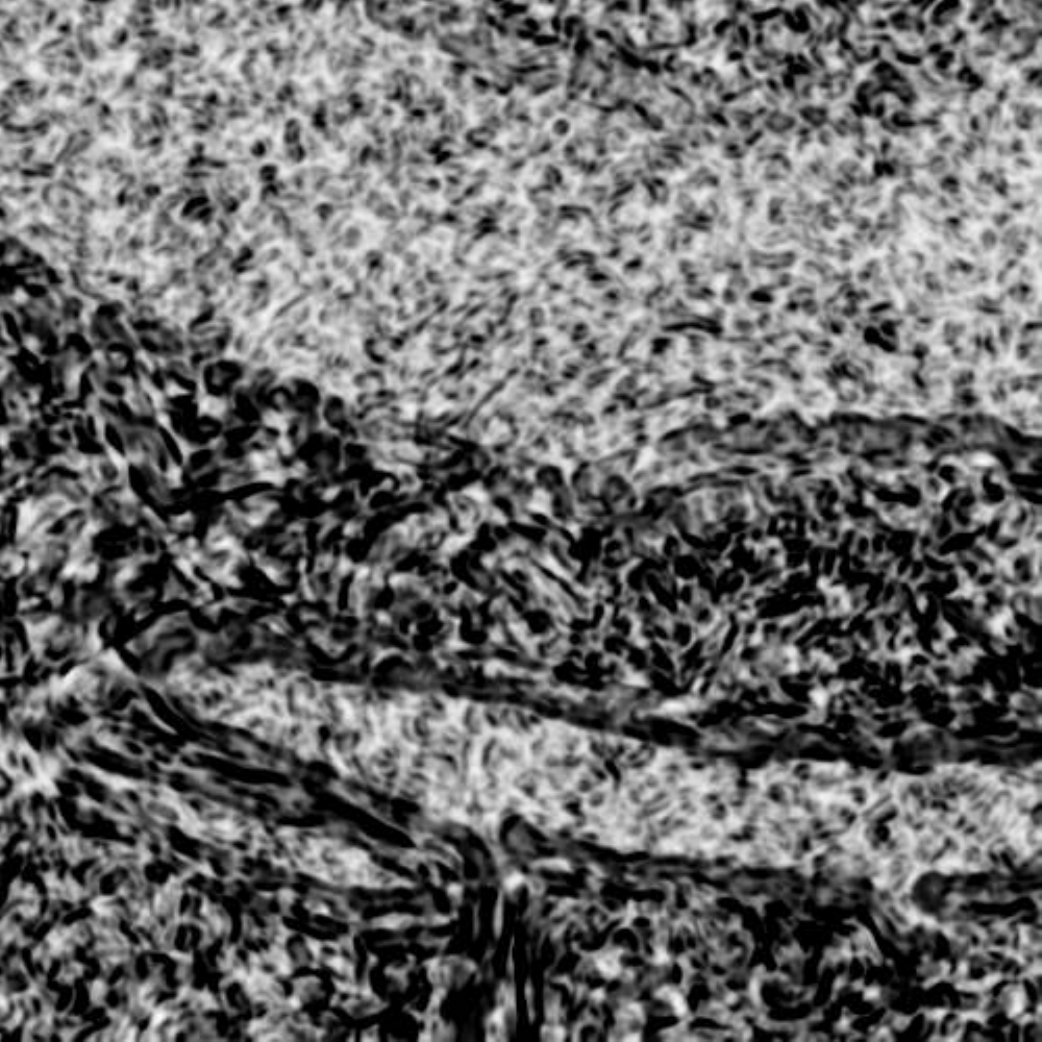} &
\includegraphics[height=\myfig]{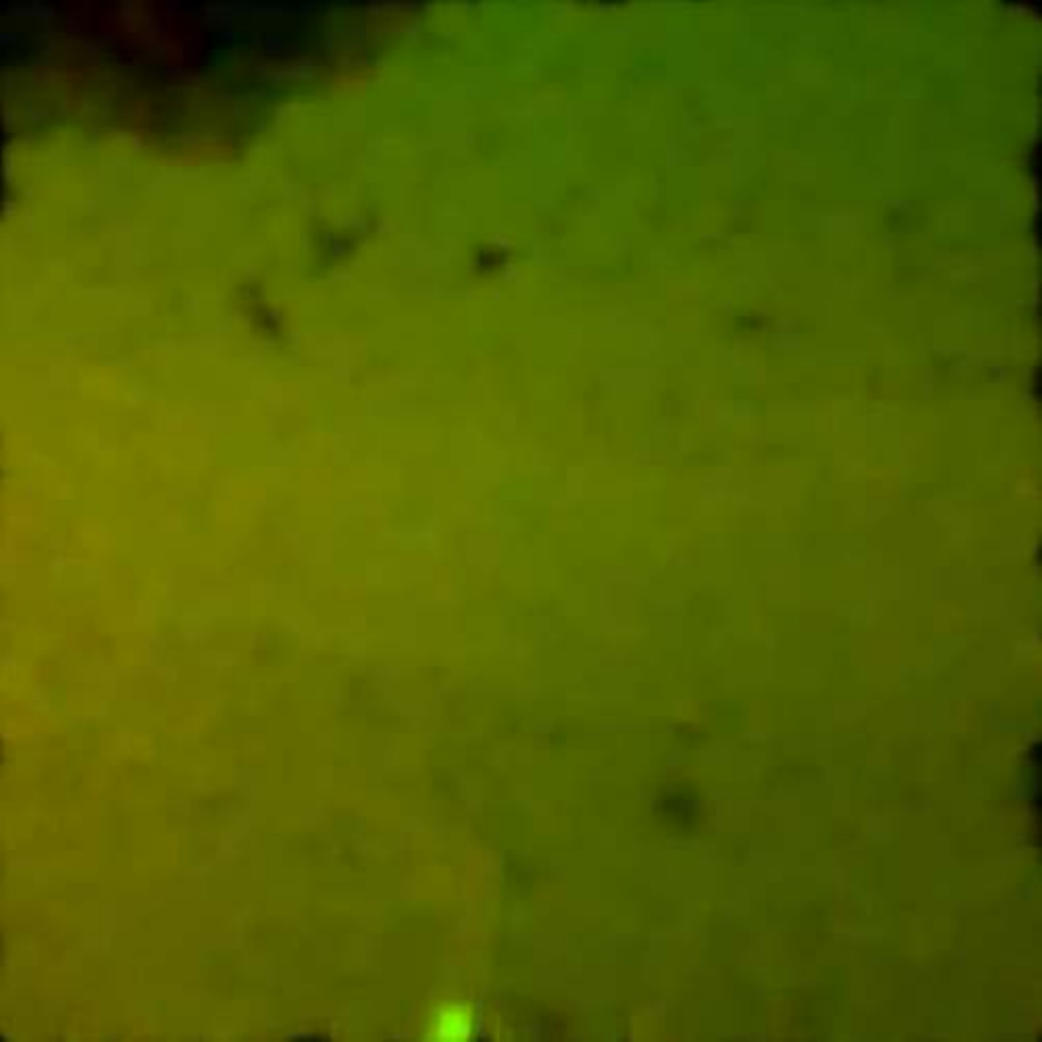}
\\

\rot{\myfig}{Ground truth} &
\scalebarme{%
\includegraphics[height=\myfig]{image/CT/eval/ct_orig/crop_gray_191146-1_E1-denoise_000650.pdf}}{%
\includegraphics[width=\myfig]{image/scalebars/scale_ct-500.png}} &
\includegraphics[height=\myfig]{image/CT/eval/ct_orig/crop_pair__ct-orig_orig-index_650-651_.pdf} &
\includegraphics[height=\myfig]{image/CT/eval/ct_orig/crop_psnr__ct-orig_orig-index_650-651_.pdf} &
\includegraphics[height=\myfig]{image/CT/eval/ct_orig/crop_ssim_full__ct-orig_orig-index_650-651_.pdf} &
\includegraphics[height=\myfig]{image/CT/eval/ct_orig/crop_flow_CROP__ct-orig_orig-index_650-651_.pdf}
\\

& \multicolumn{1}{c}{Image} & \multicolumn{1}{c}{Dice} & \multicolumn{1}{c}{PSNR} & \multicolumn{1}{c}{SSIM} & \multicolumn{1}{c}{Flow} \\
\end{tabular}

\caption{Evaluation of CT data set with image measures, part~2. Continued from Fig.~\ref{fig:eval-ct-1}. \enquote{Local only} means the distorted ground truth, but with no global transforms applied. The registrations' input had global transforms applied. All scale bars are \SI{1}{\milli\meter}.}
\label{fig:eval-ct-2}
\setlength{\tabcolsep}{\tabcolbackup}
\end{figure*}

\begin{figure*}[p]
\centering
\setlength{\myfig}{0.2\linewidth-0.001\linewidth-2pt-0.2em}
\setlength{\tabcolsep}{1pt}
\begin{tabular}{m{1em} m{\myfig} m{\myfig} m{\myfig} m{\myfig} m{\myfig}}
& \multicolumn{1}{c}{Image} & \multicolumn{1}{c}{Dice} & \multicolumn{1}{c}{PSNR} & \multicolumn{1}{c}{SSIM} & \multicolumn{1}{c}{Flow} \\

\rot{\myfig}{Rigid-SIFT} &
\scalebarme{%
\includegraphics[height=\myfig]{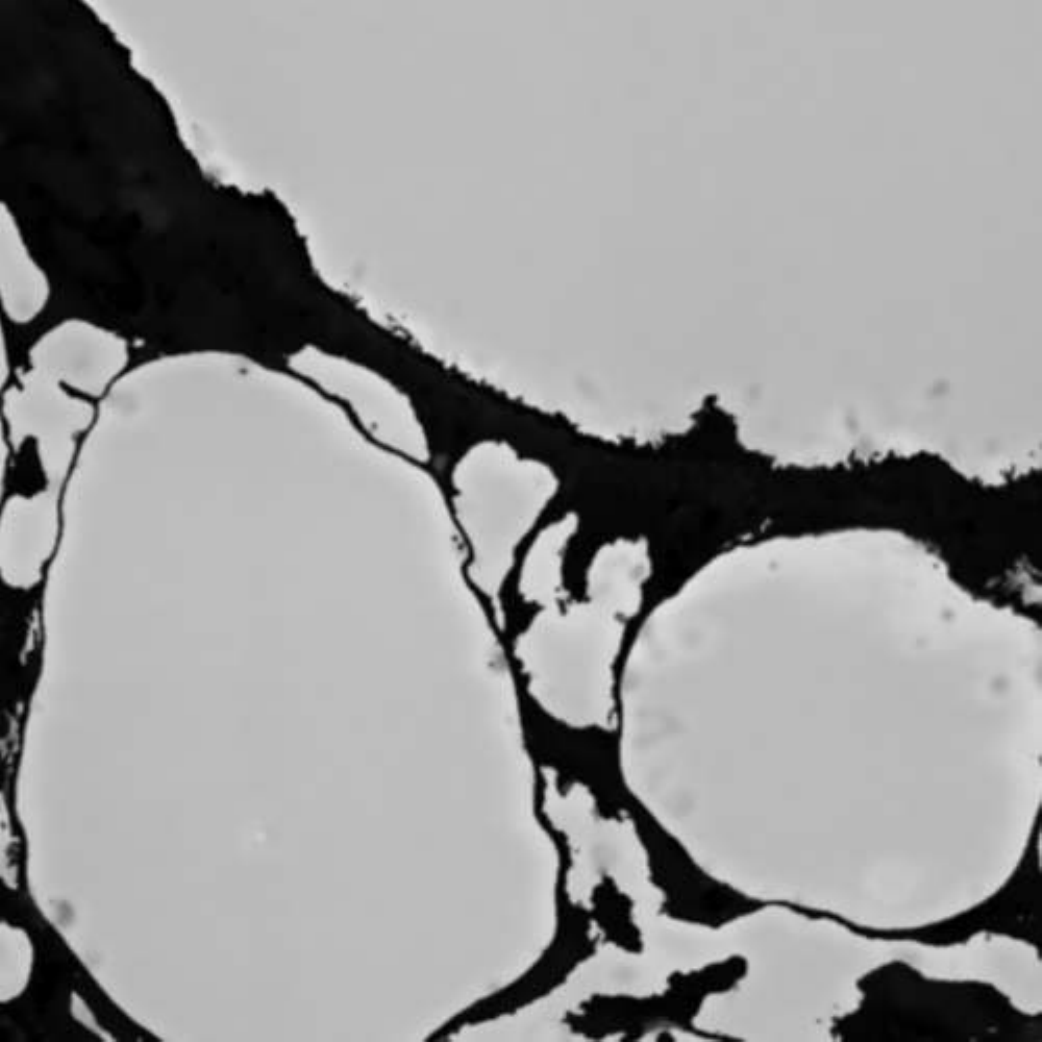}}{%
\includegraphics[width=\myfig]{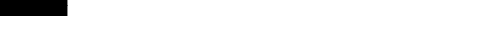}} &
\includegraphics[height=\myfig]{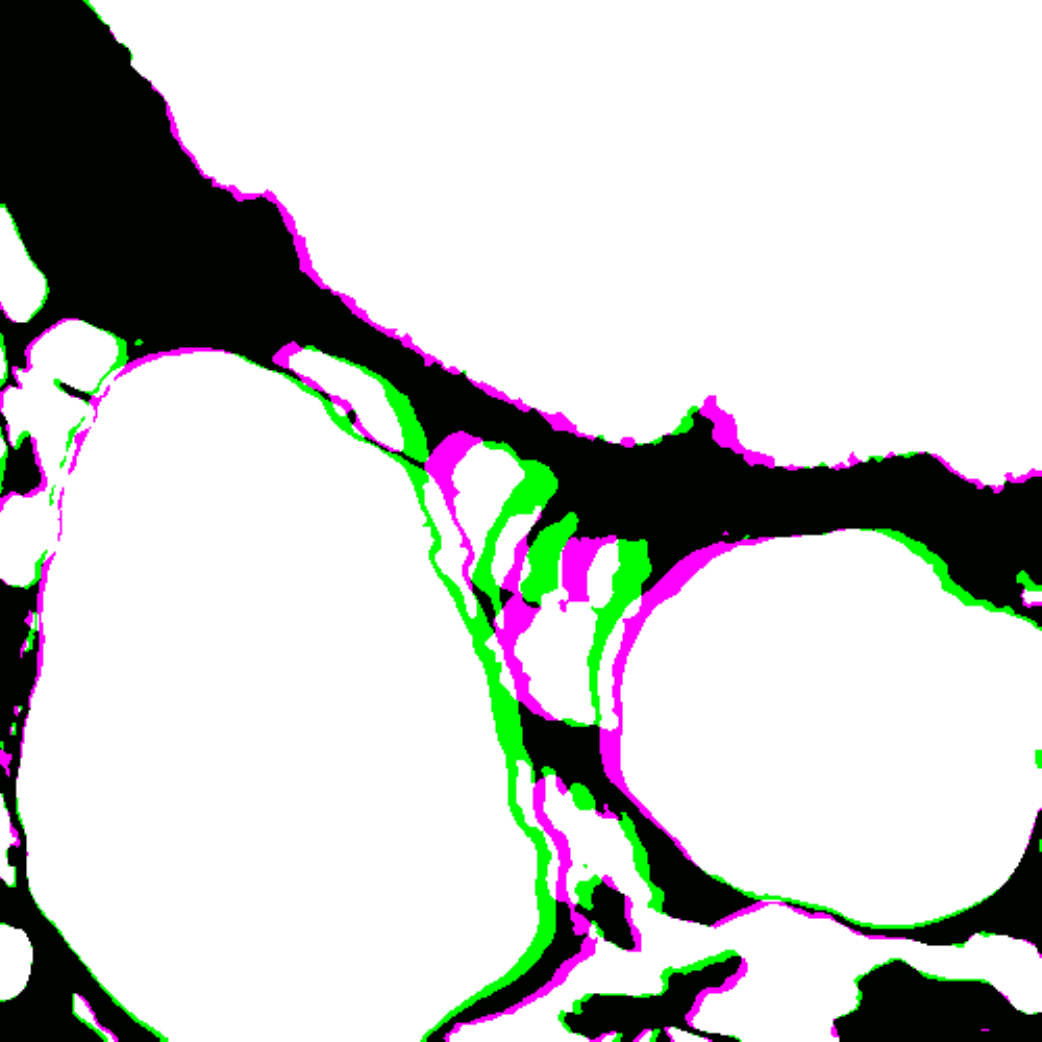} &
\includegraphics[height=\myfig]{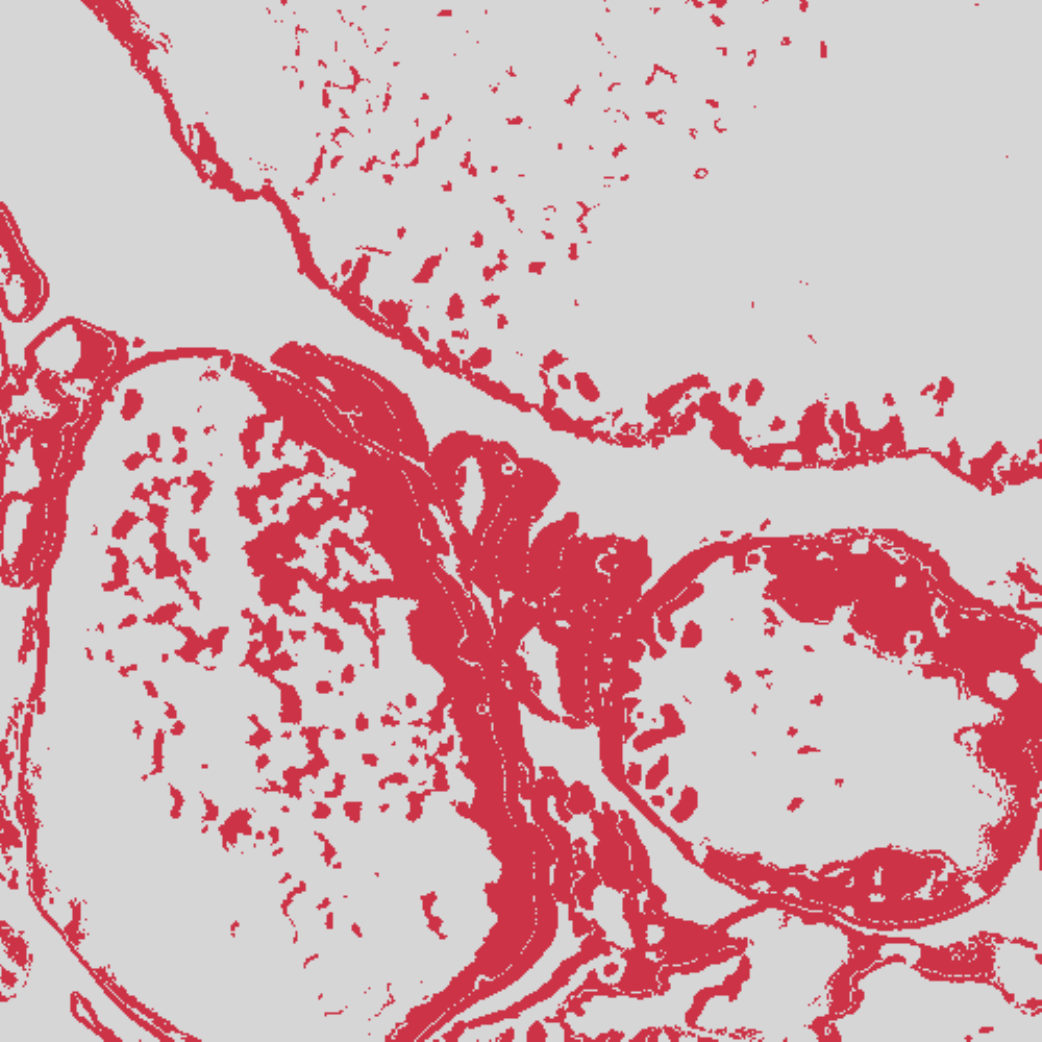} &
\includegraphics[height=\myfig]{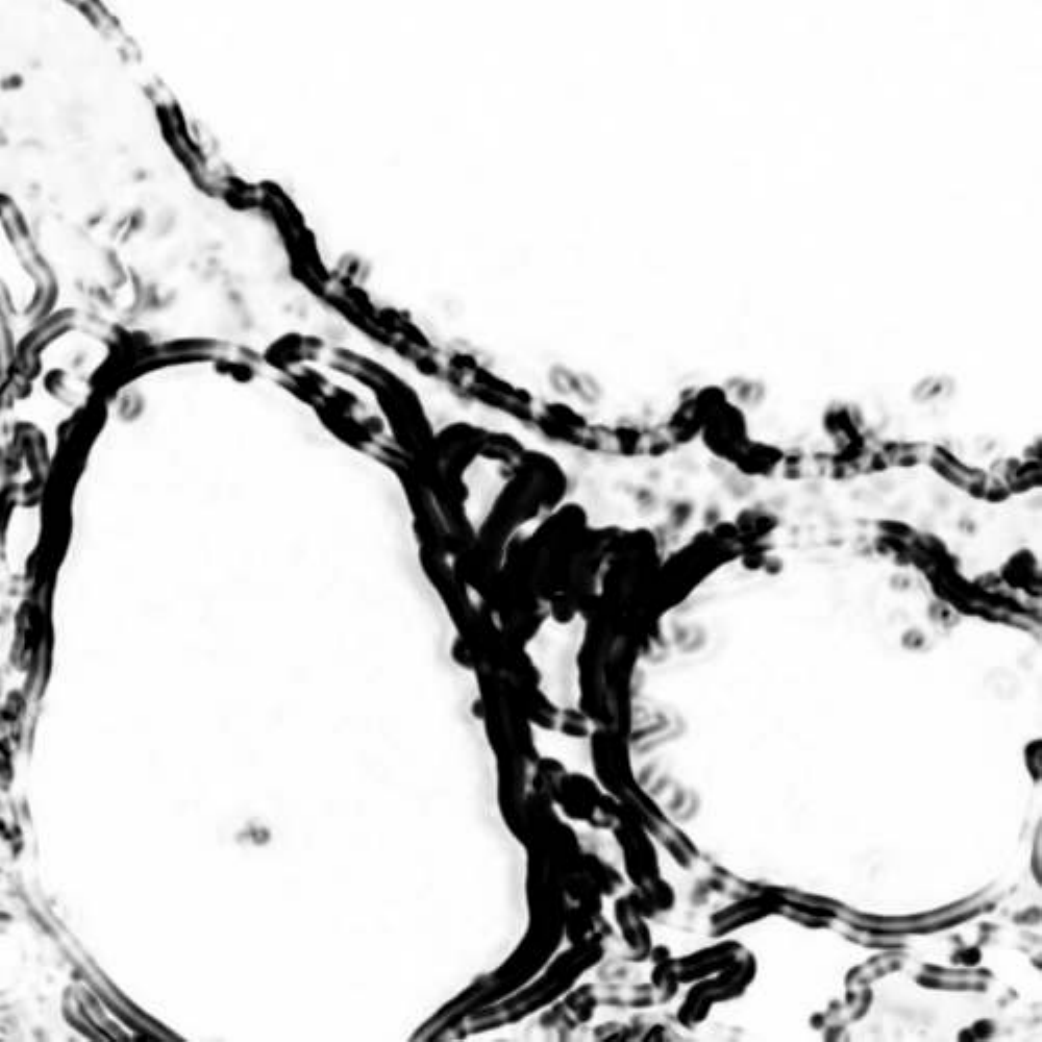} &
\includegraphics[height=\myfig]{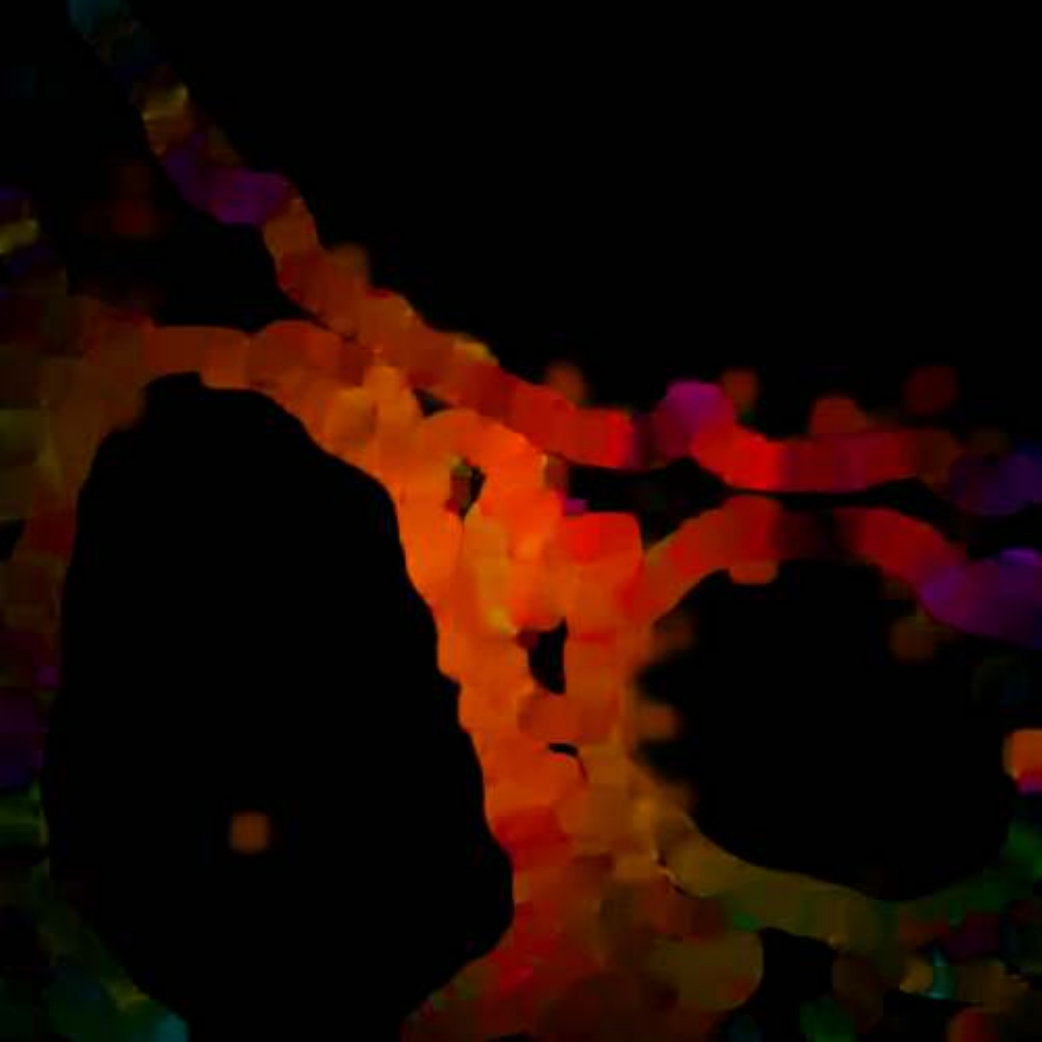}
\\

\rot{\myfig}{Rigid-SURF} &
\scalebarme{%
\includegraphics[height=\myfig]{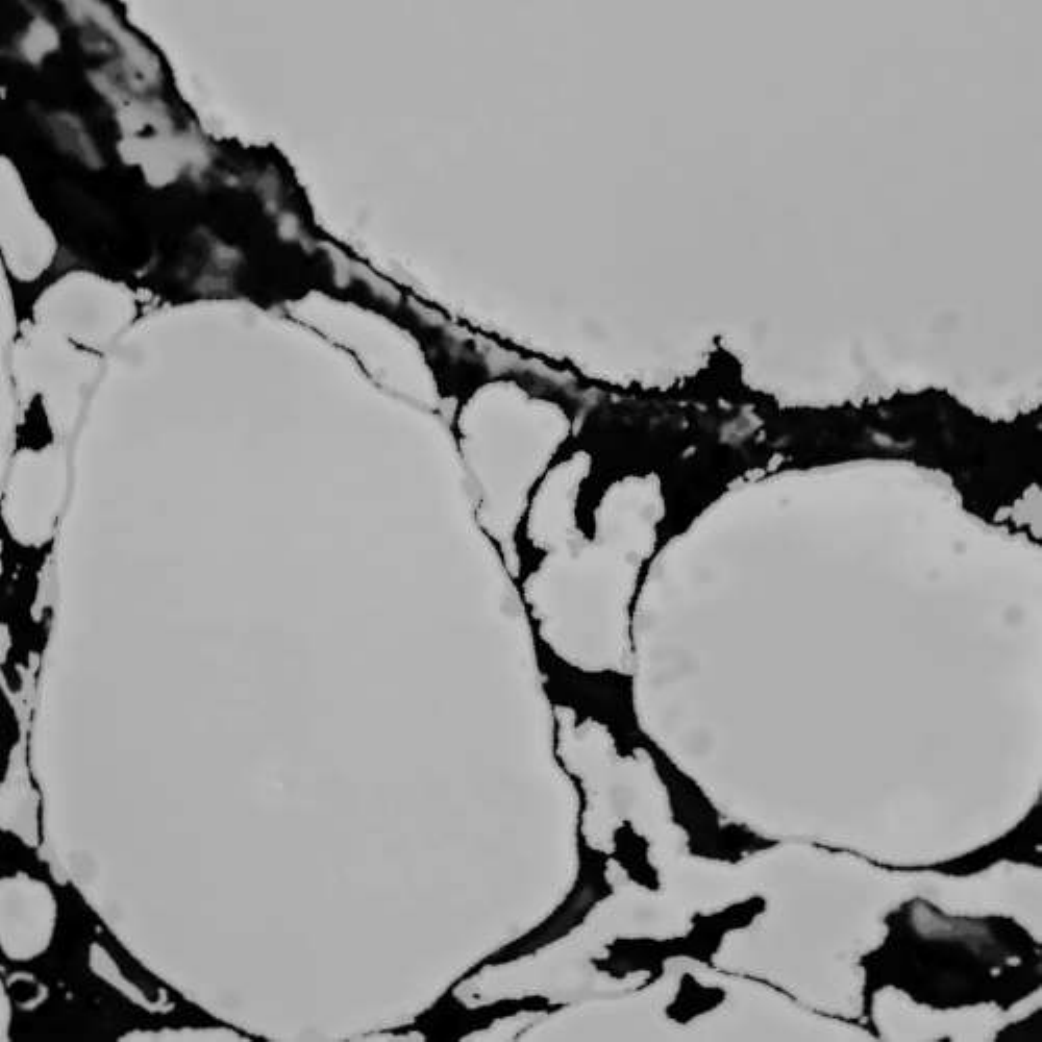}}{%
\includegraphics[width=\myfig]{image/scalebars/scale_em-500.png}} &
\includegraphics[height=\myfig]{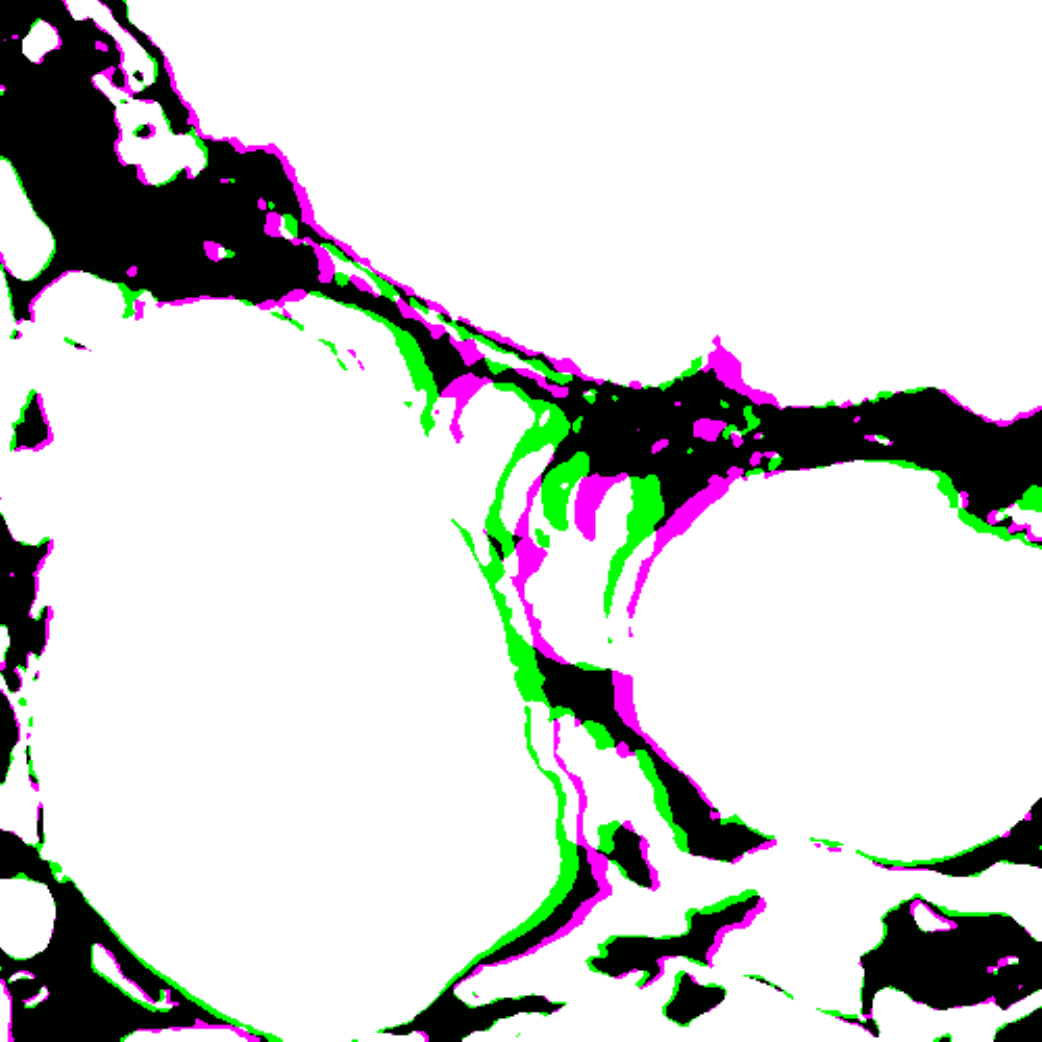} &
\includegraphics[height=\myfig]{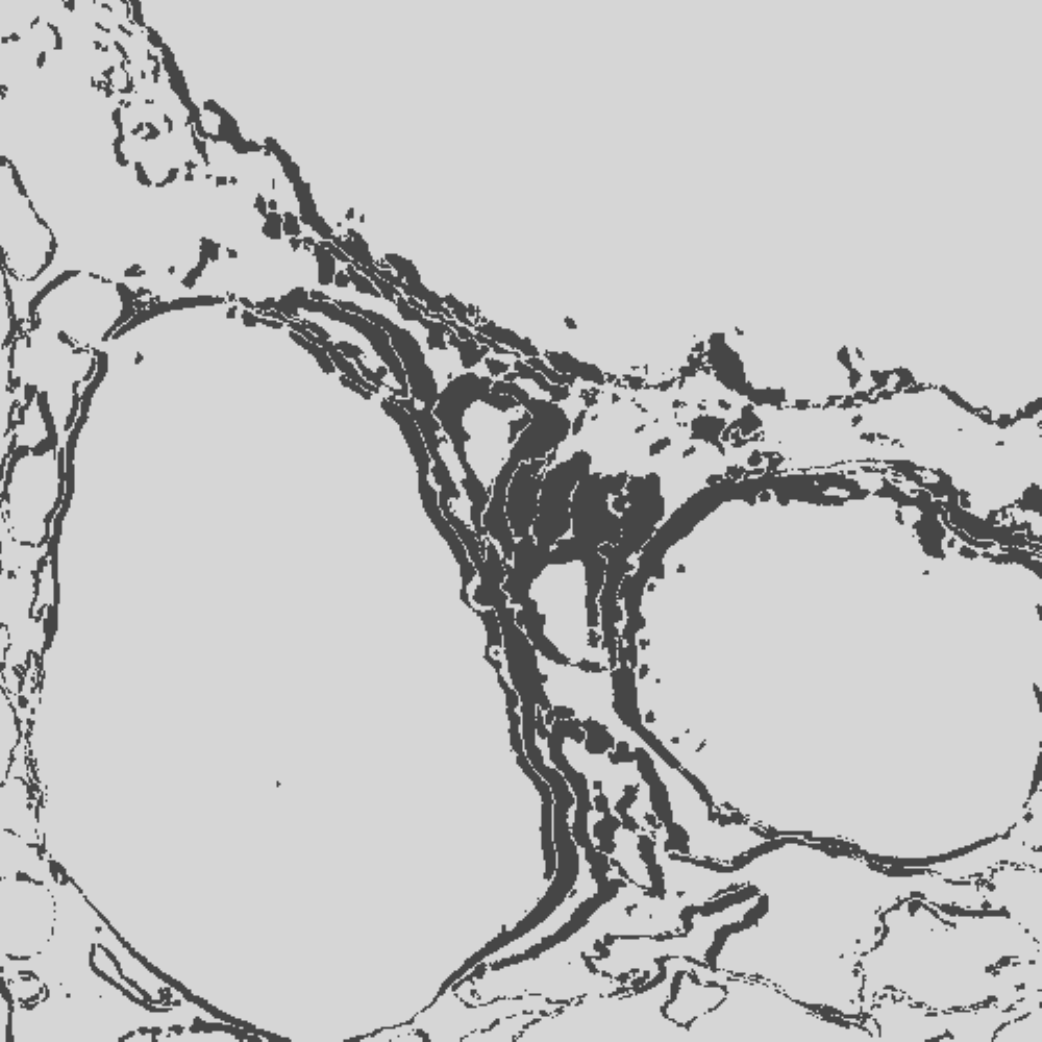} &
\includegraphics[height=\myfig]{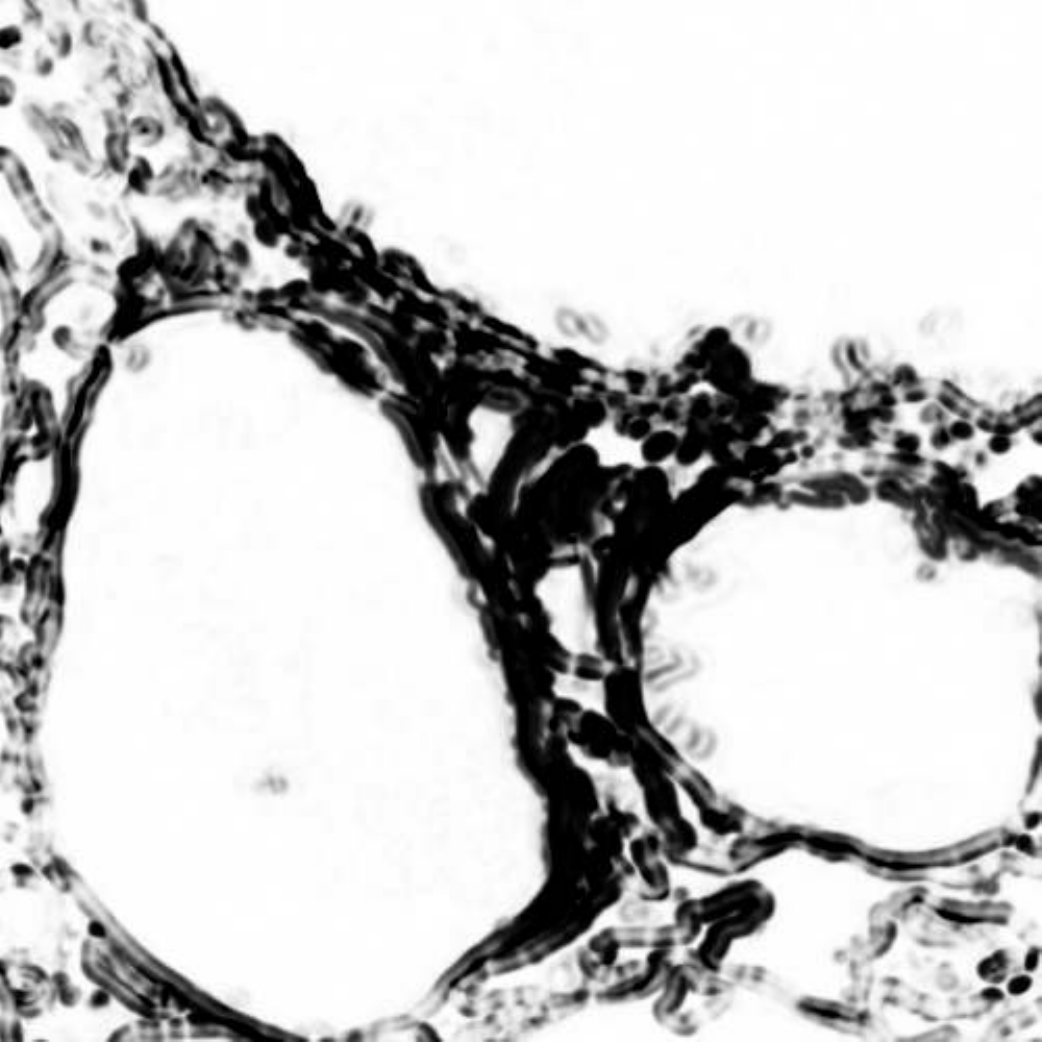} &
\includegraphics[height=\myfig]{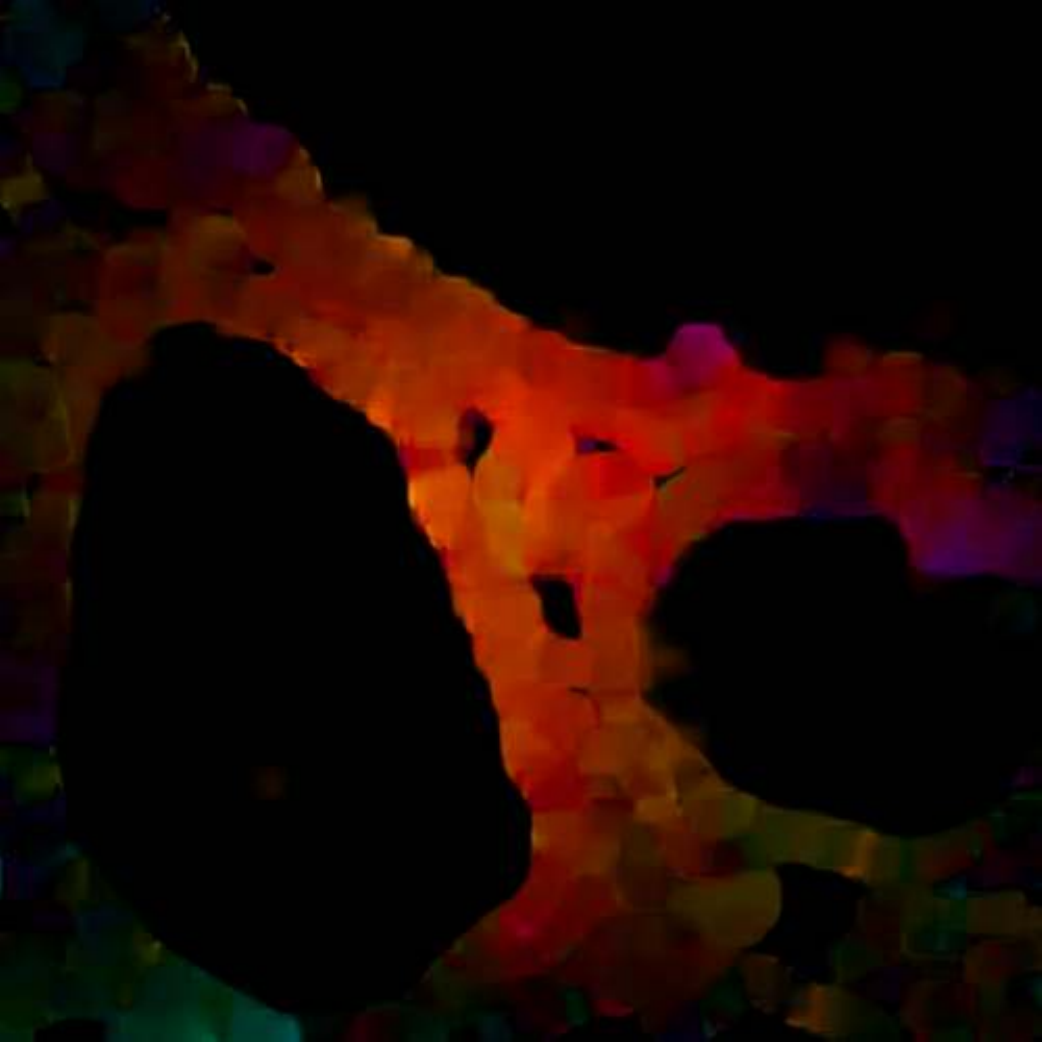}
\\

\rot{\myfig}{Deform-SURF} &
\scalebarme{%
\includegraphics[height=\myfig]{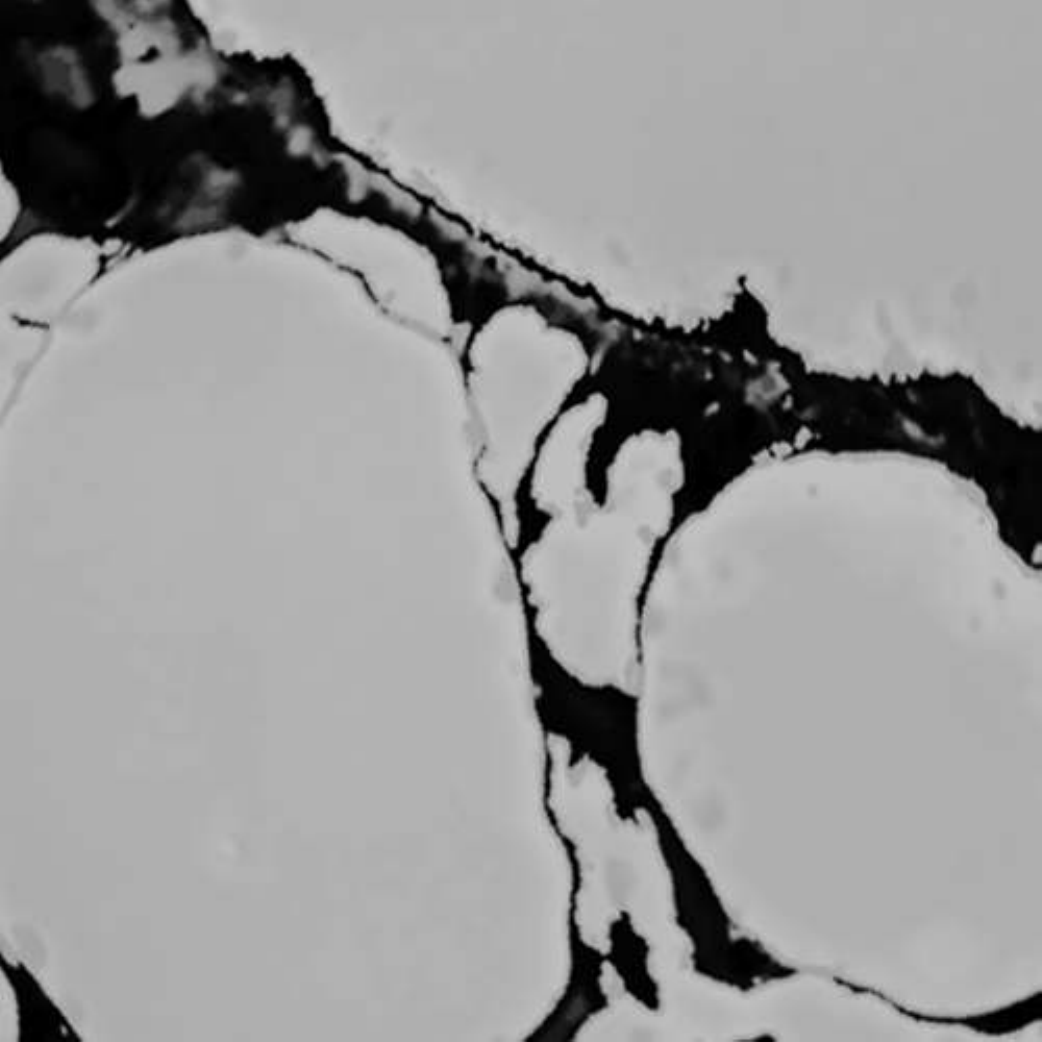}}{%
\includegraphics[width=\myfig]{image/scalebars/scale_em-500.png}} &
\includegraphics[height=\myfig]{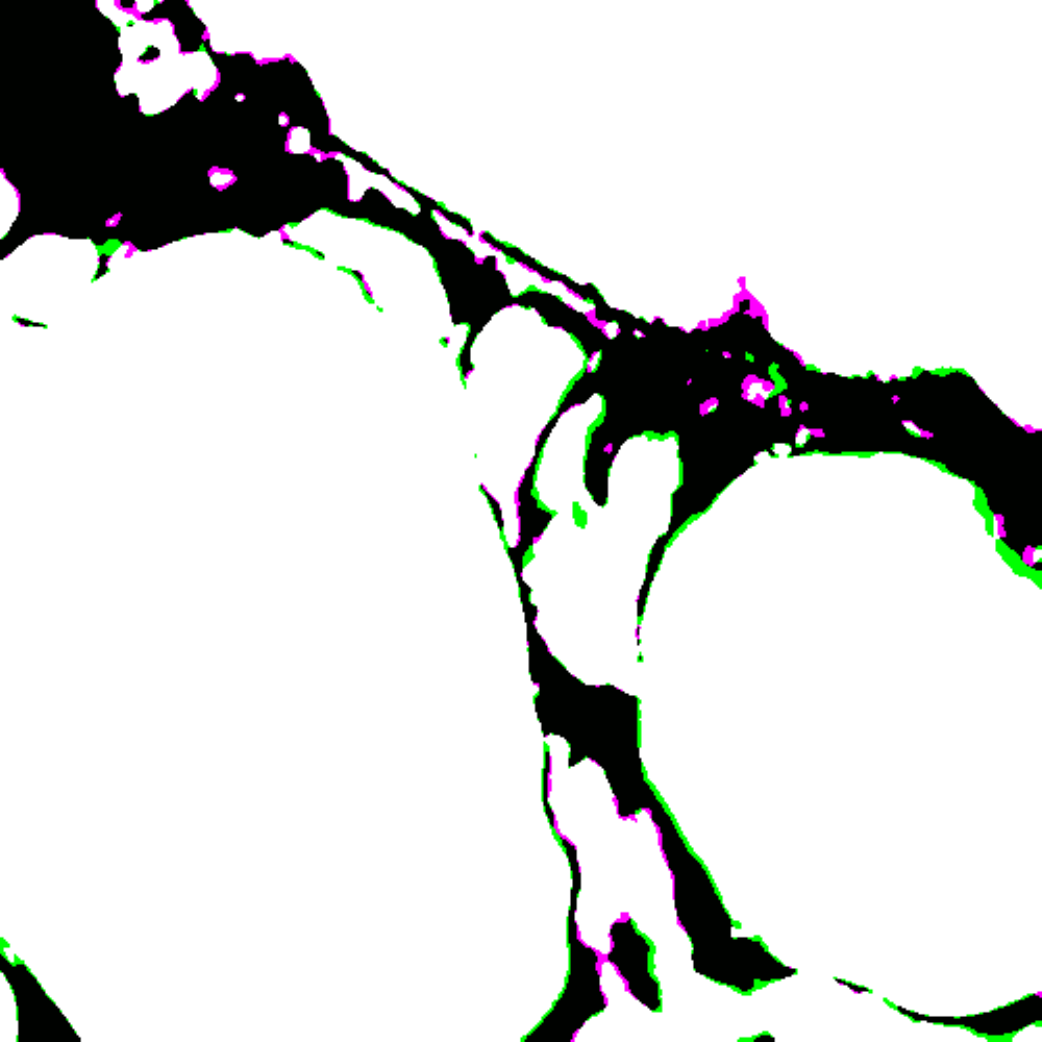} &
\includegraphics[height=\myfig]{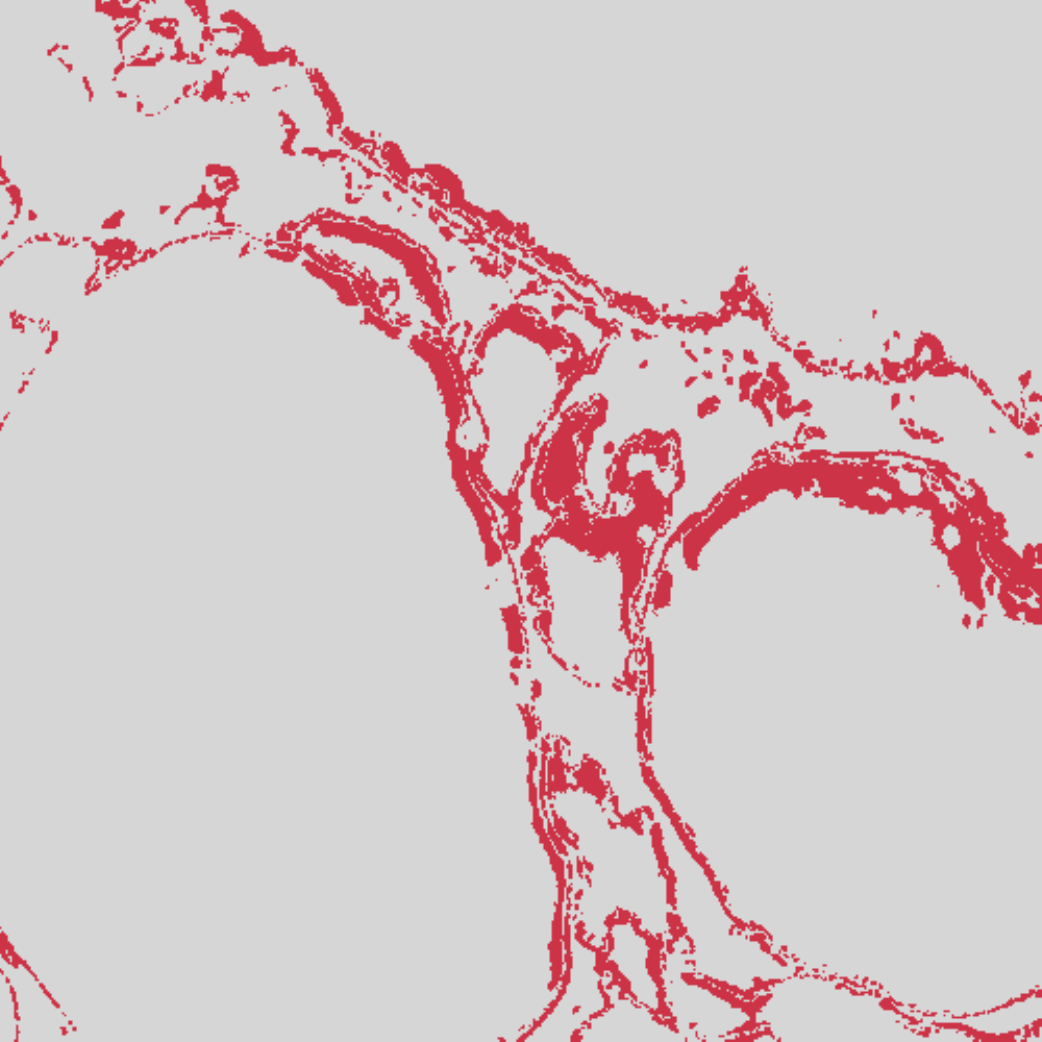} &
\includegraphics[height=\myfig]{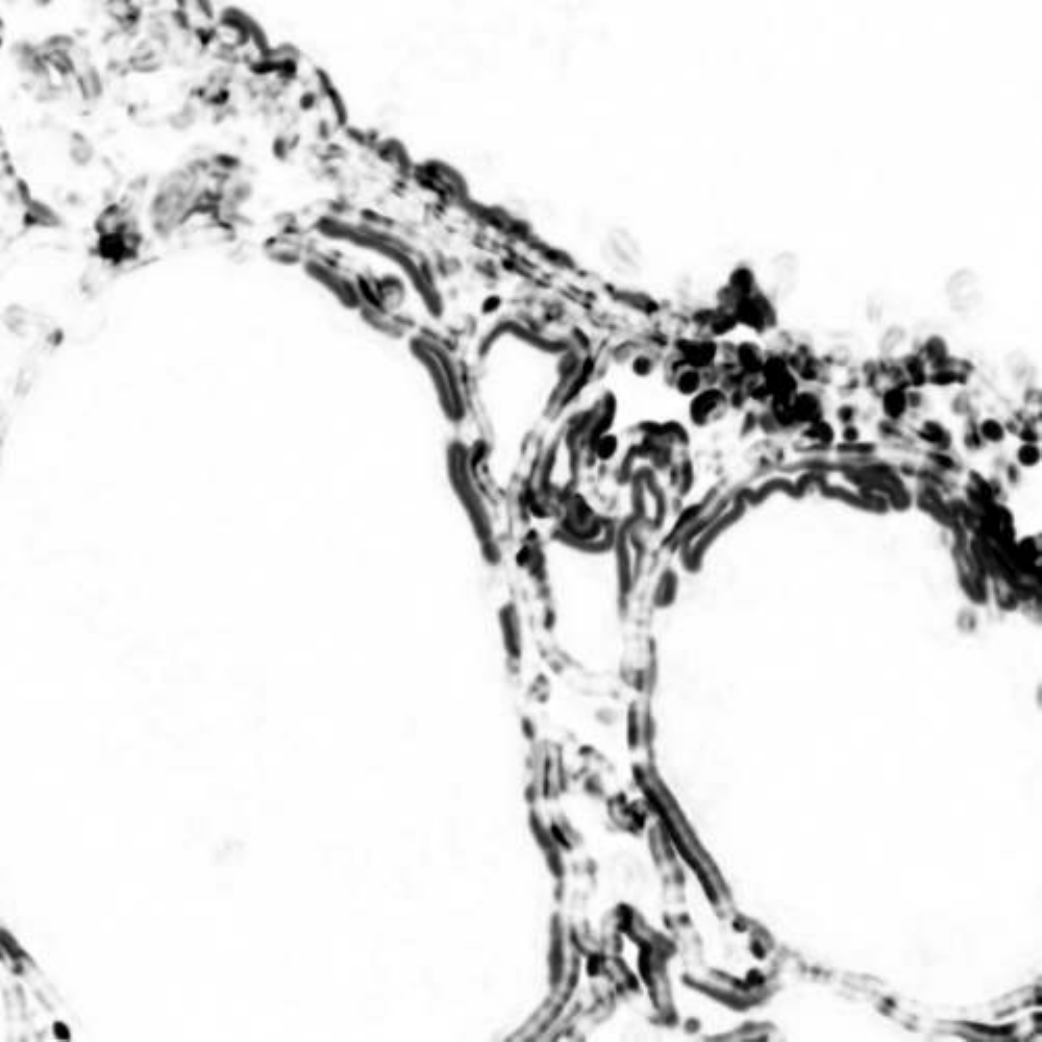} &
\includegraphics[height=\myfig]{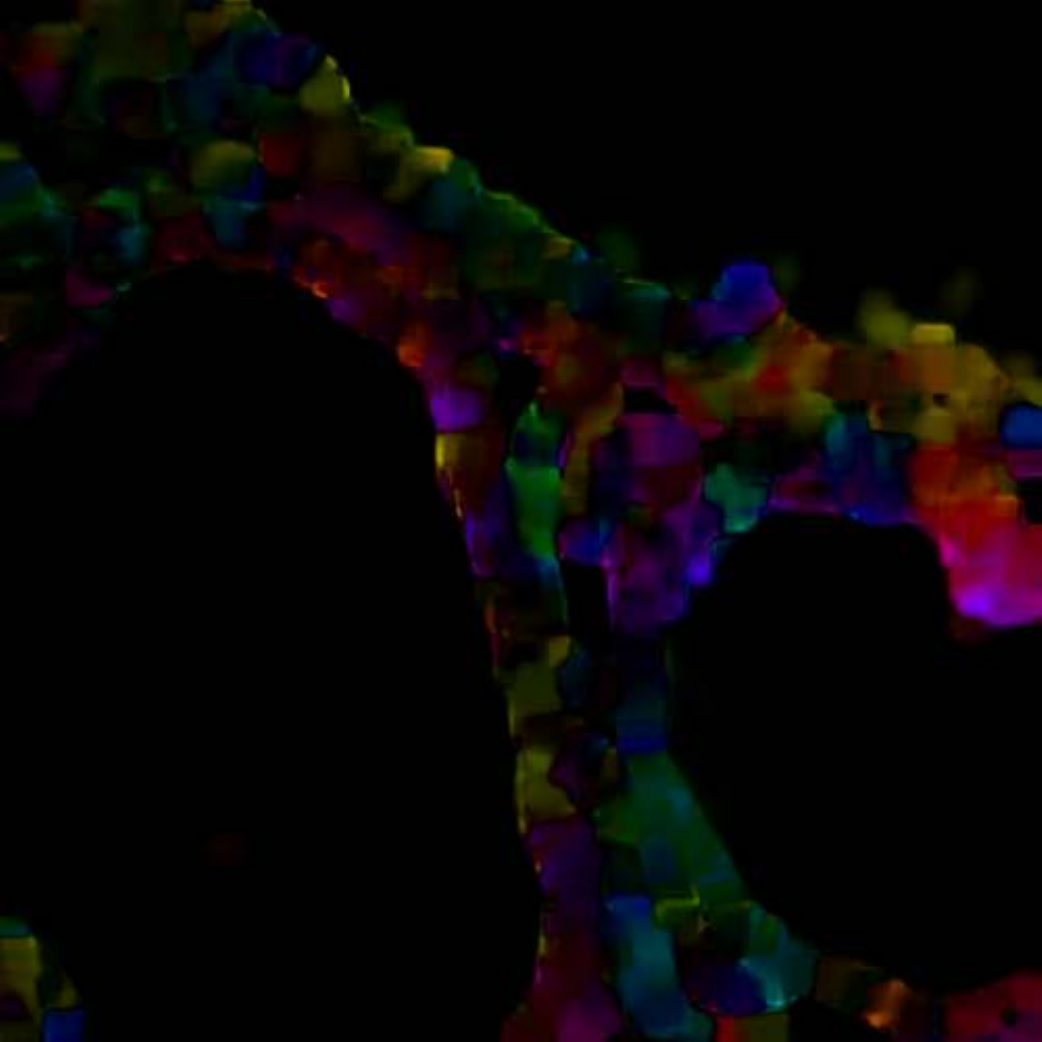}
\\

\rot{\myfig}{GS} &
\scalebarme{%
\includegraphics[height=\myfig]{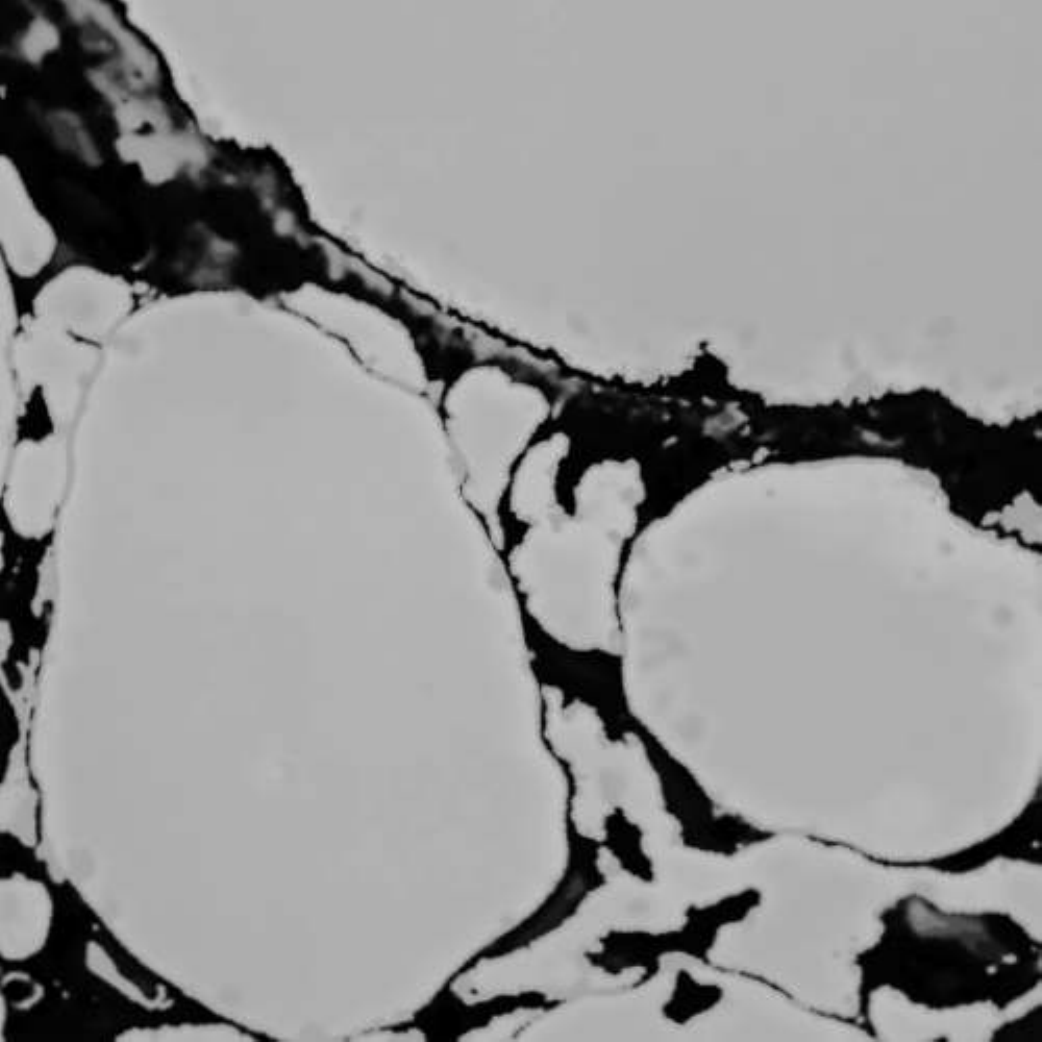}}{%
\includegraphics[width=\myfig]{image/scalebars/scale_em-500.png}} &
\includegraphics[height=\myfig]{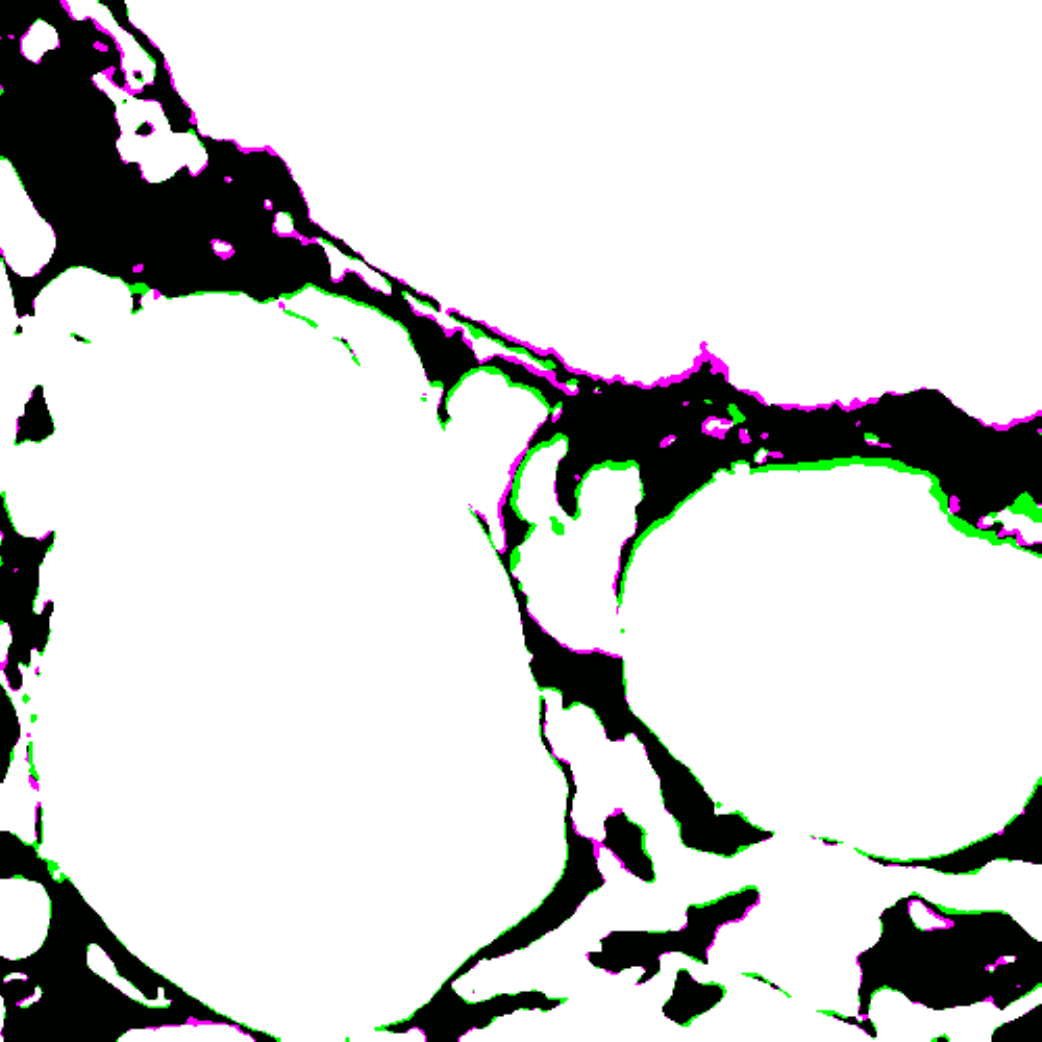} &
\includegraphics[height=\myfig]{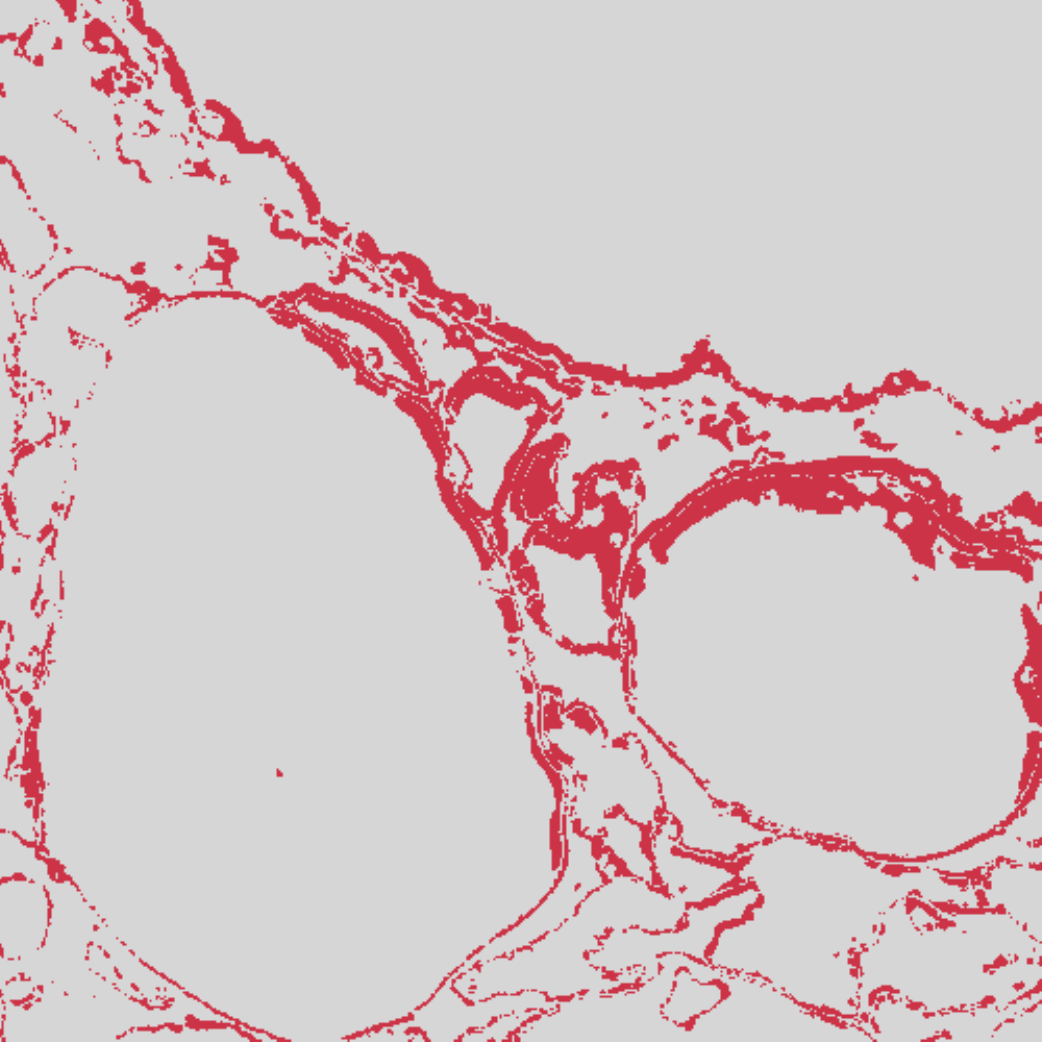} &
\includegraphics[height=\myfig]{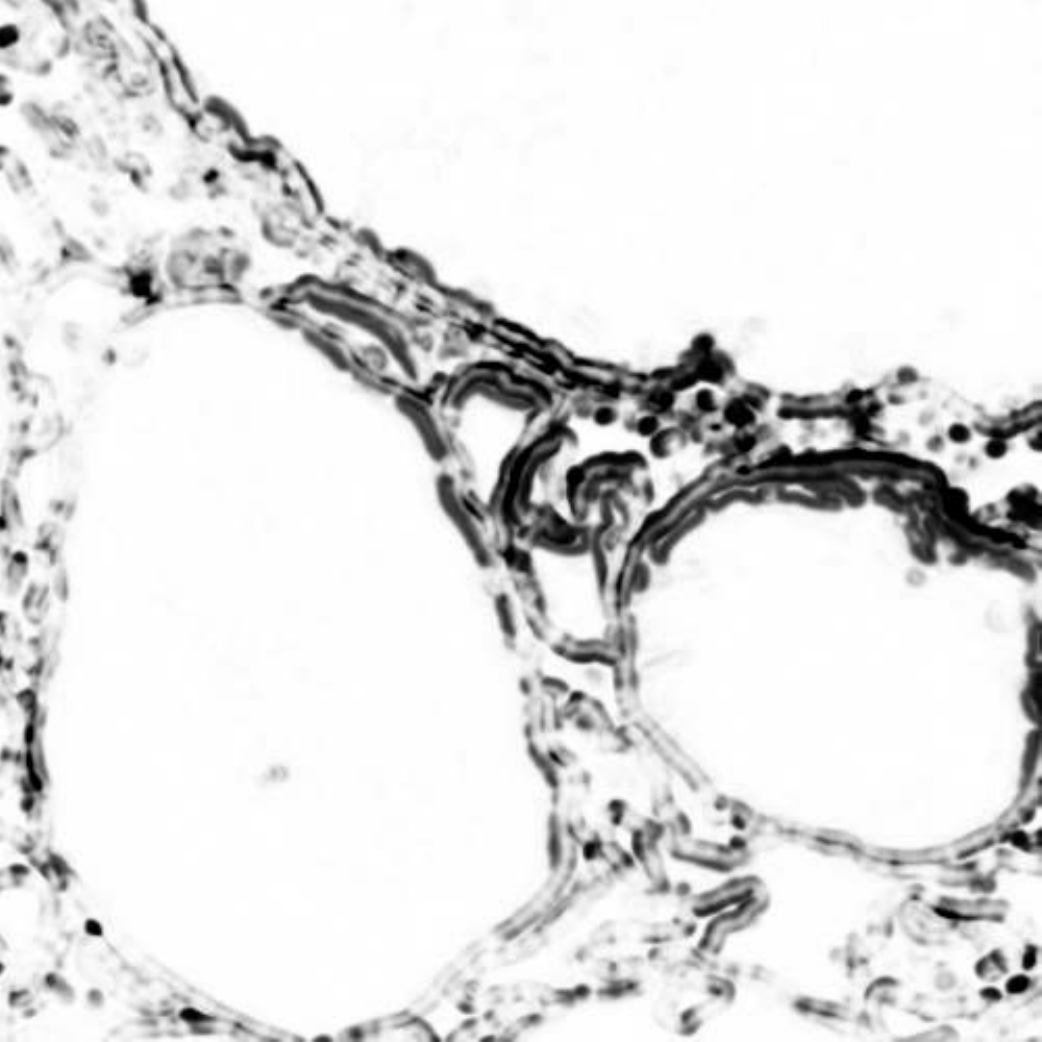} &
\includegraphics[height=\myfig]{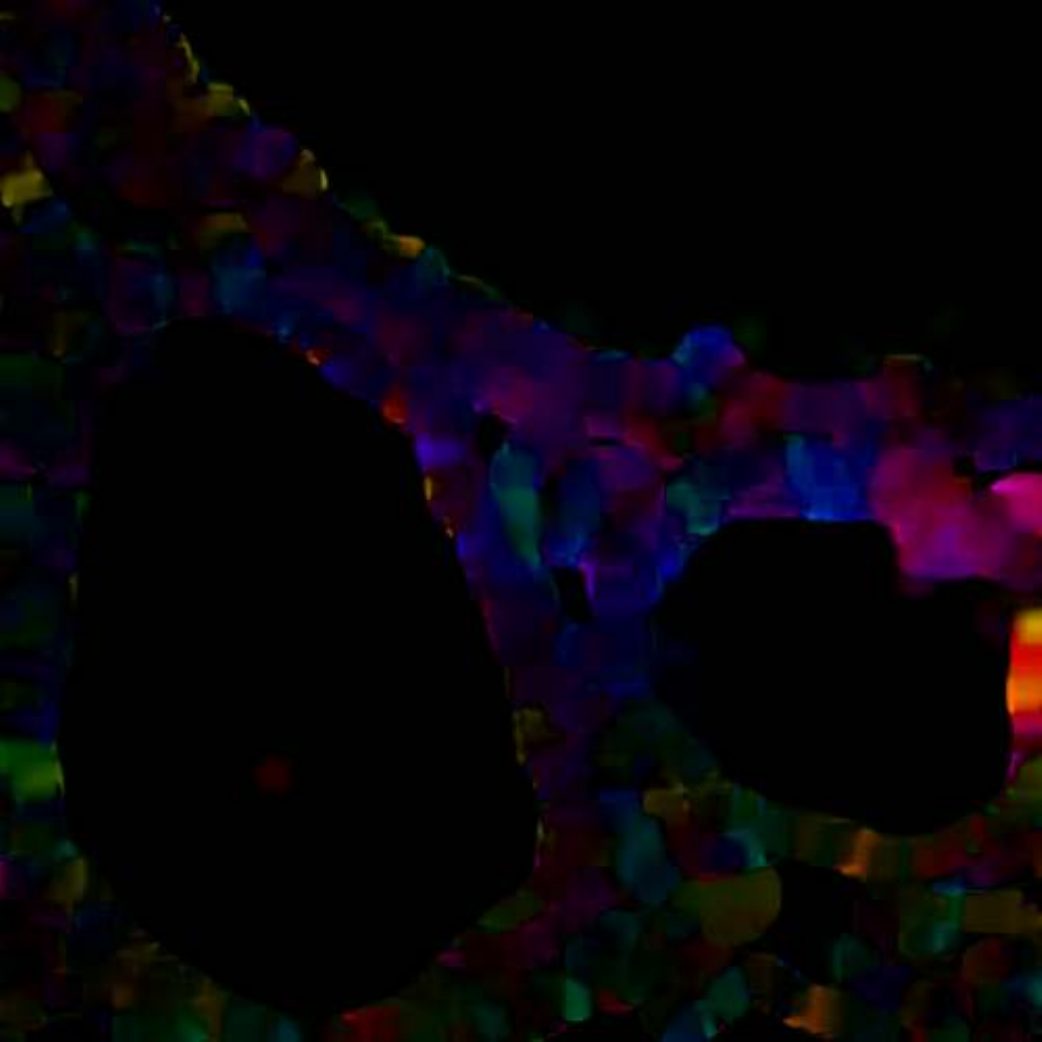}
\\

\rot{\myfig}{Blending} &
\scalebarme{%
\includegraphics[height=\myfig]{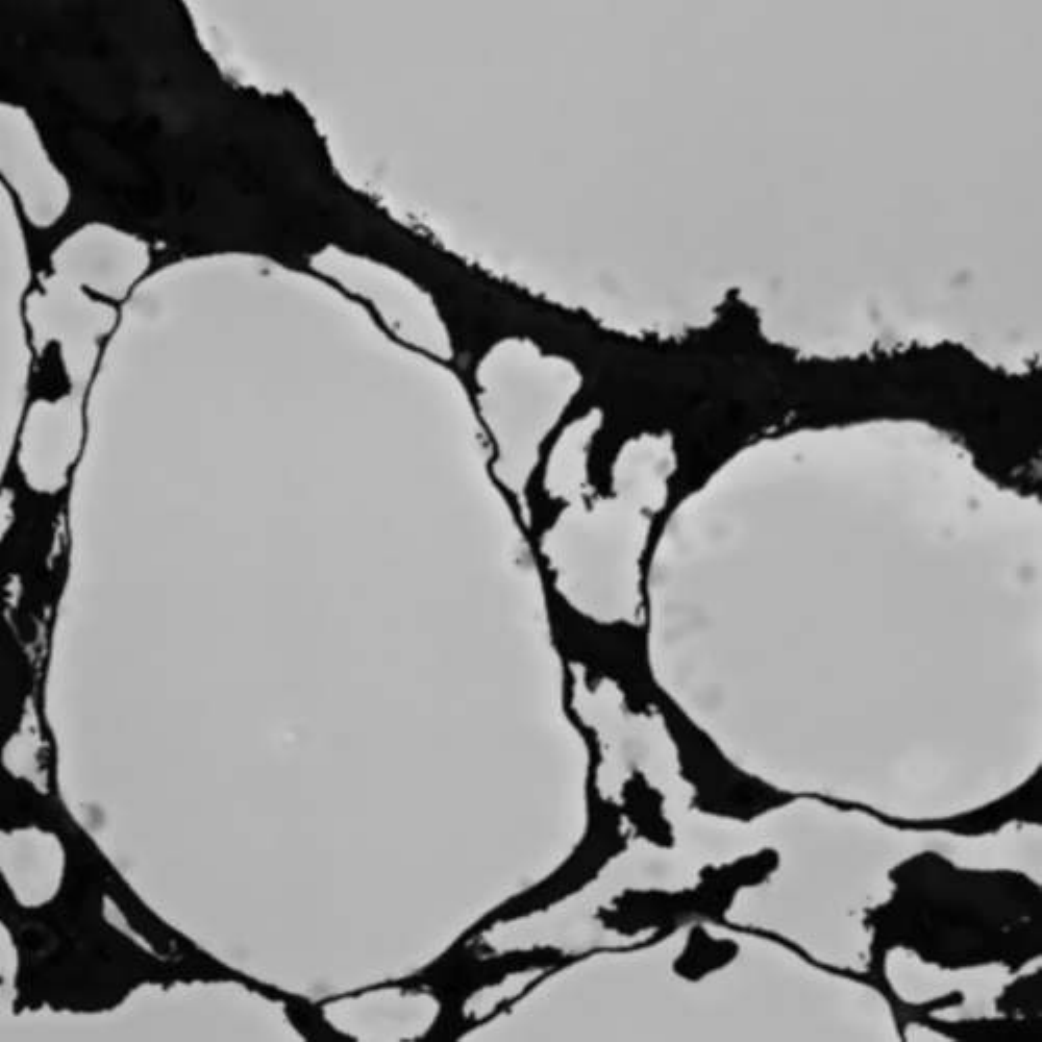}}{%
\includegraphics[width=\myfig]{image/scalebars/scale_em-500.png}} &
\includegraphics[height=\myfig]{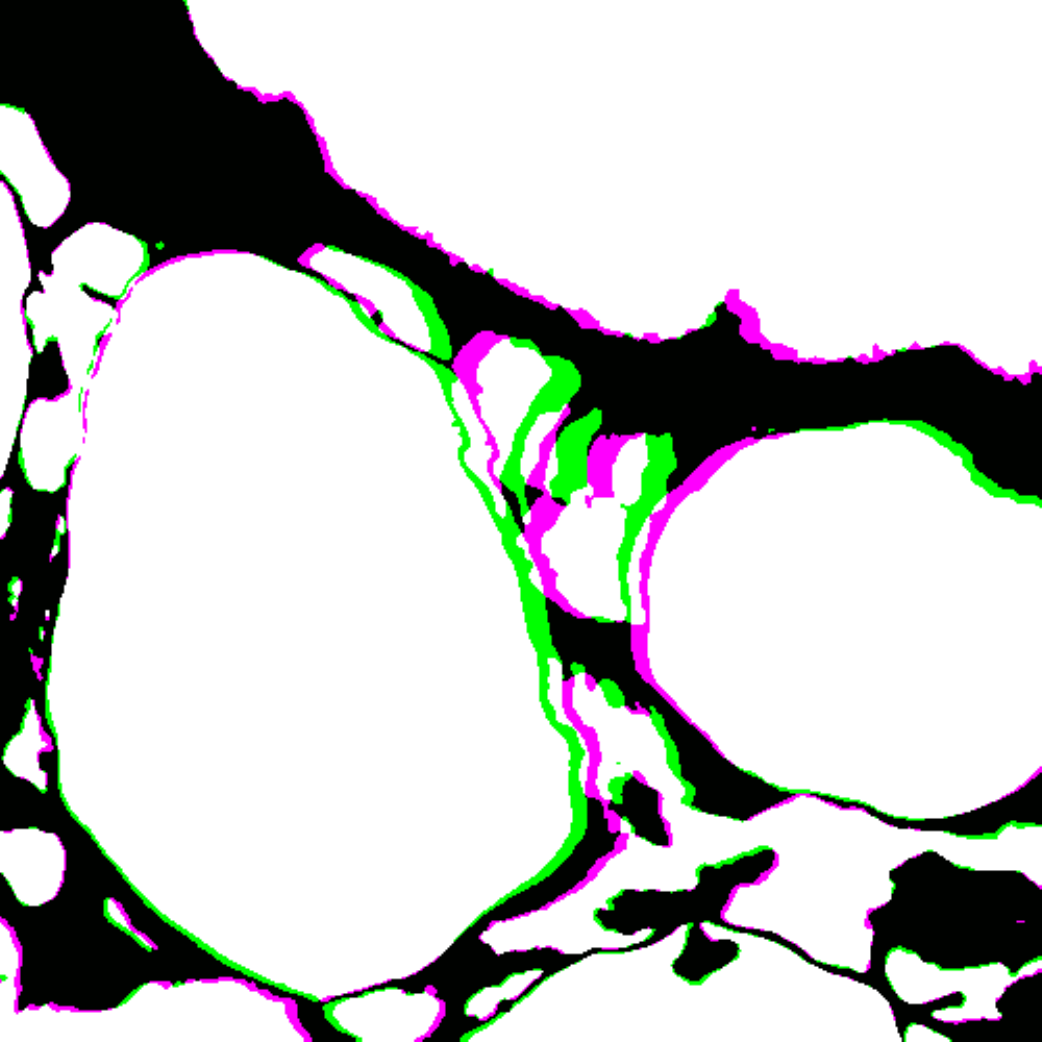} &
\includegraphics[height=\myfig]{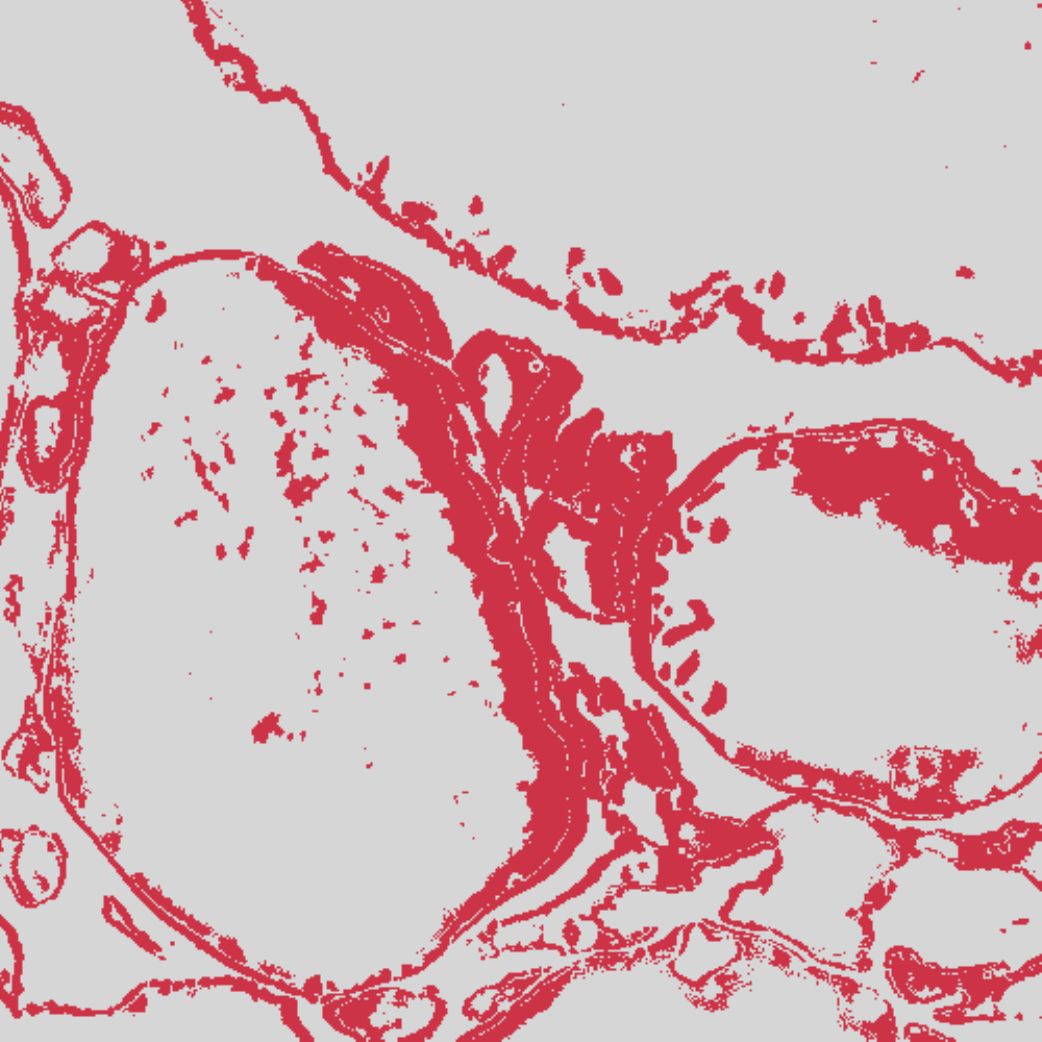} &
\includegraphics[height=\myfig]{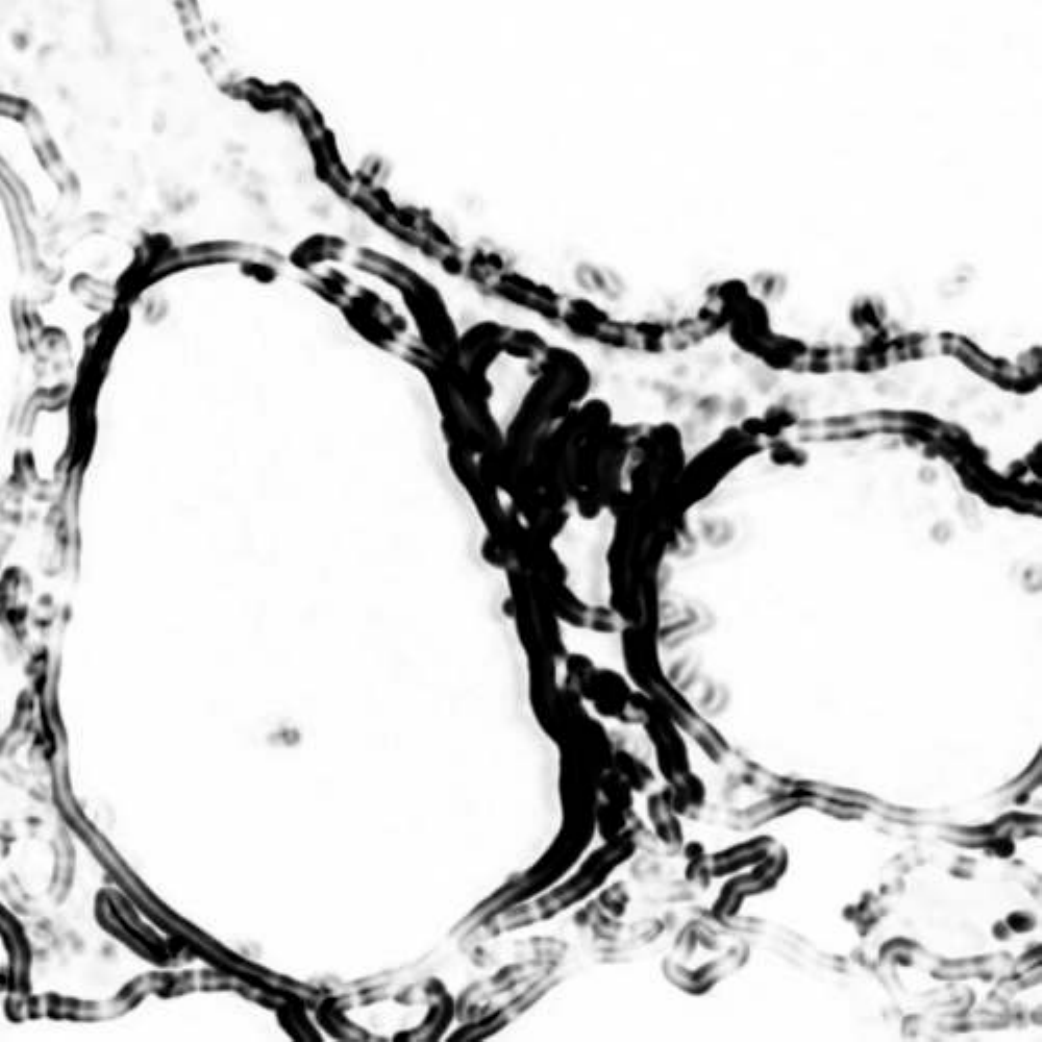} &
\includegraphics[height=\myfig]{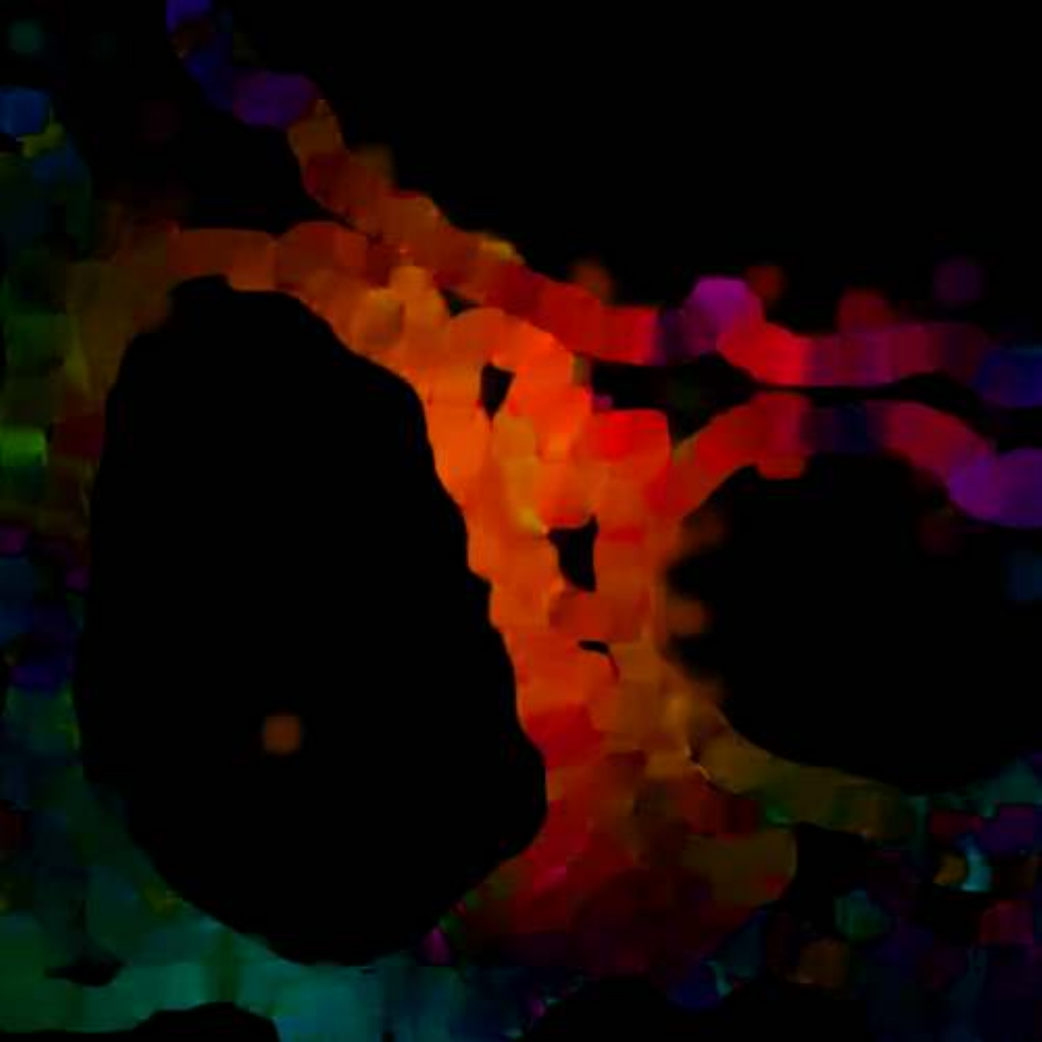}
\\

\rot{\myfig}{Gr.~truth, norm.} &
\scalebarme{%
\includegraphics[height=\myfig]{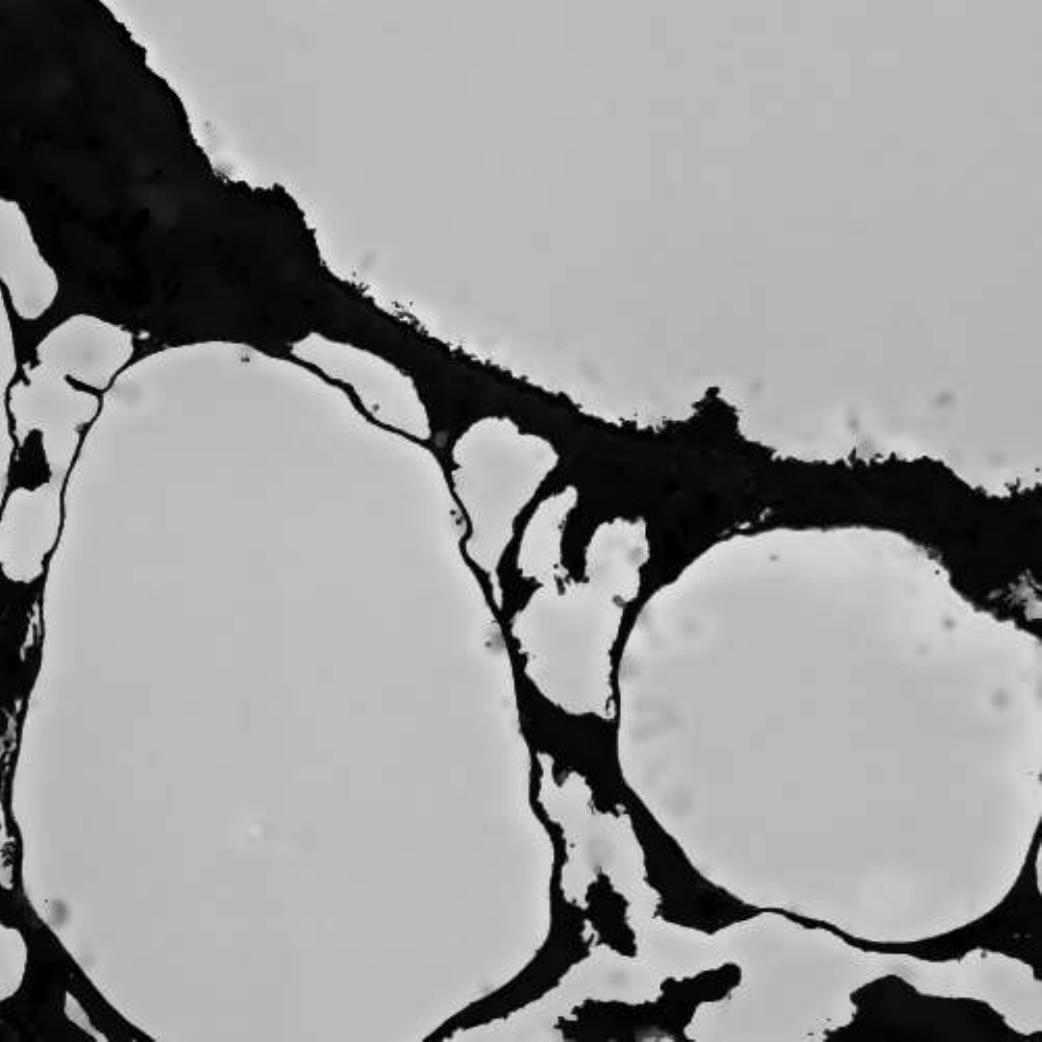}}{%
\includegraphics[width=\myfig]{image/scalebars/scale_em-500.png}} &
\includegraphics[height=\myfig]{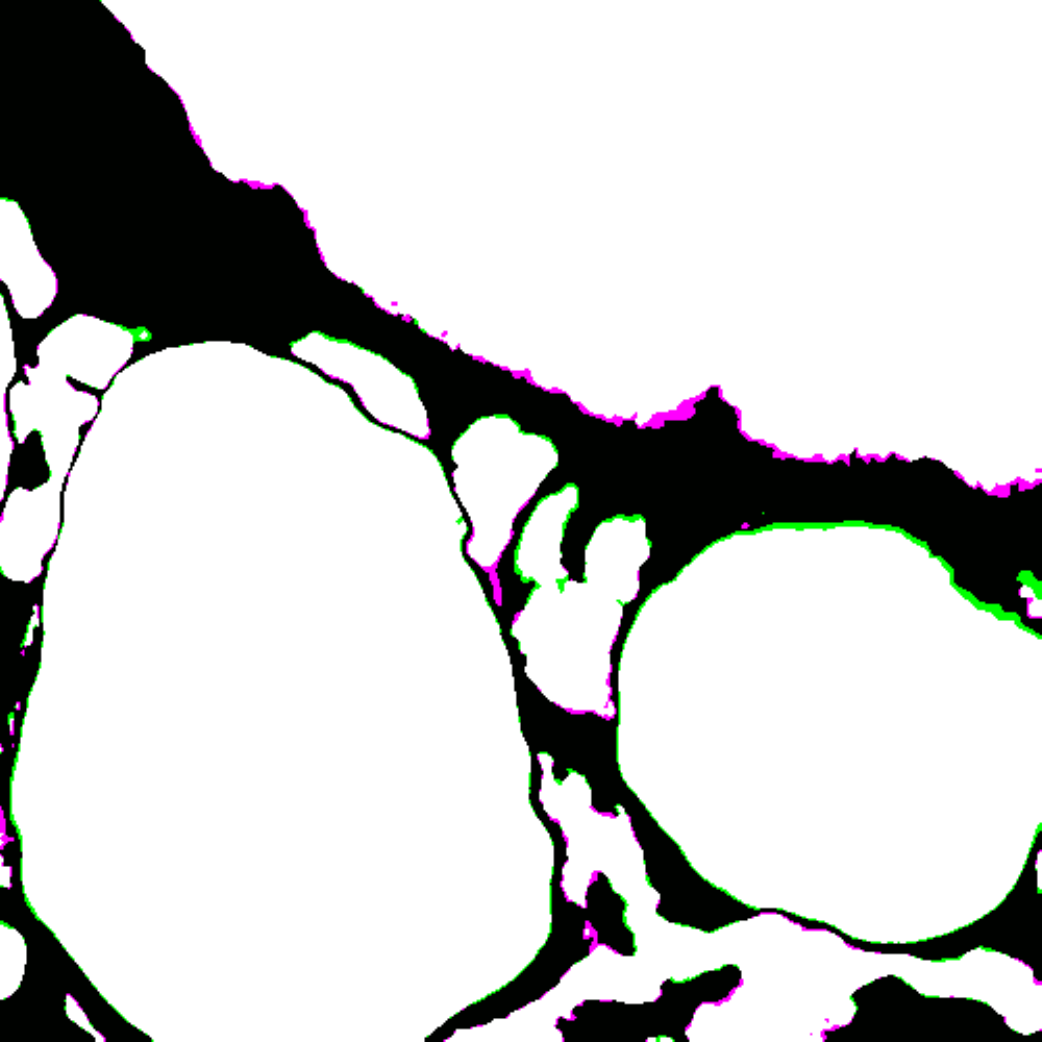} &
\includegraphics[height=\myfig]{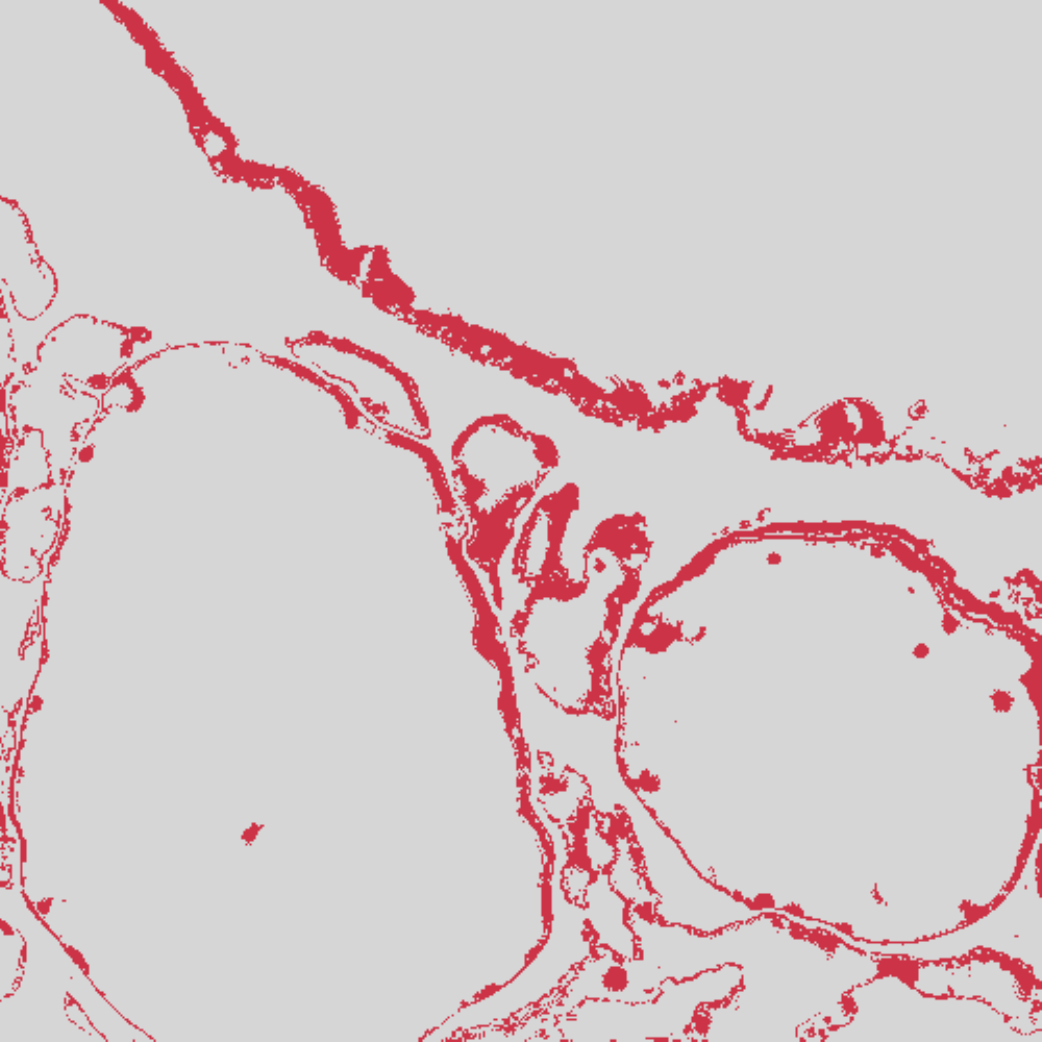} &
\includegraphics[height=\myfig]{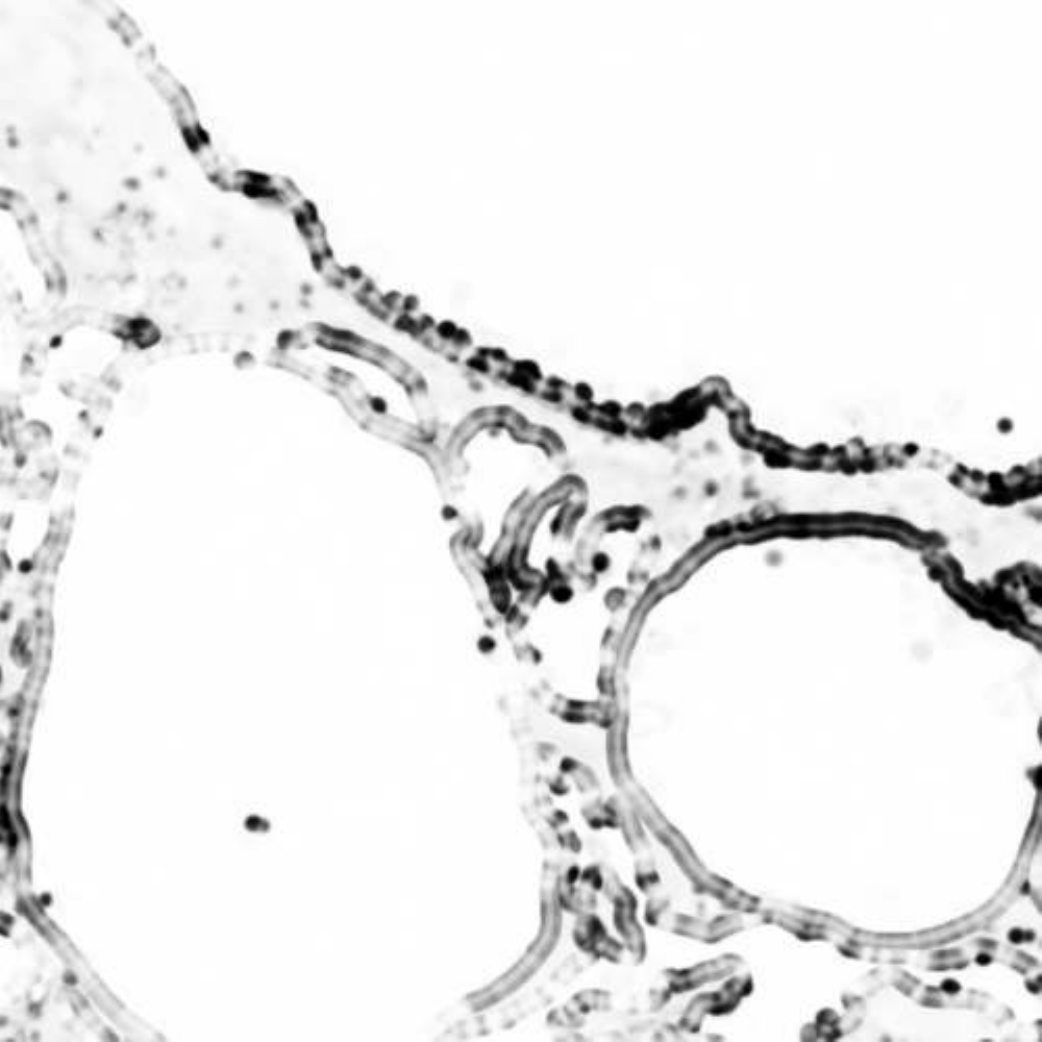} &
\includegraphics[height=\myfig]{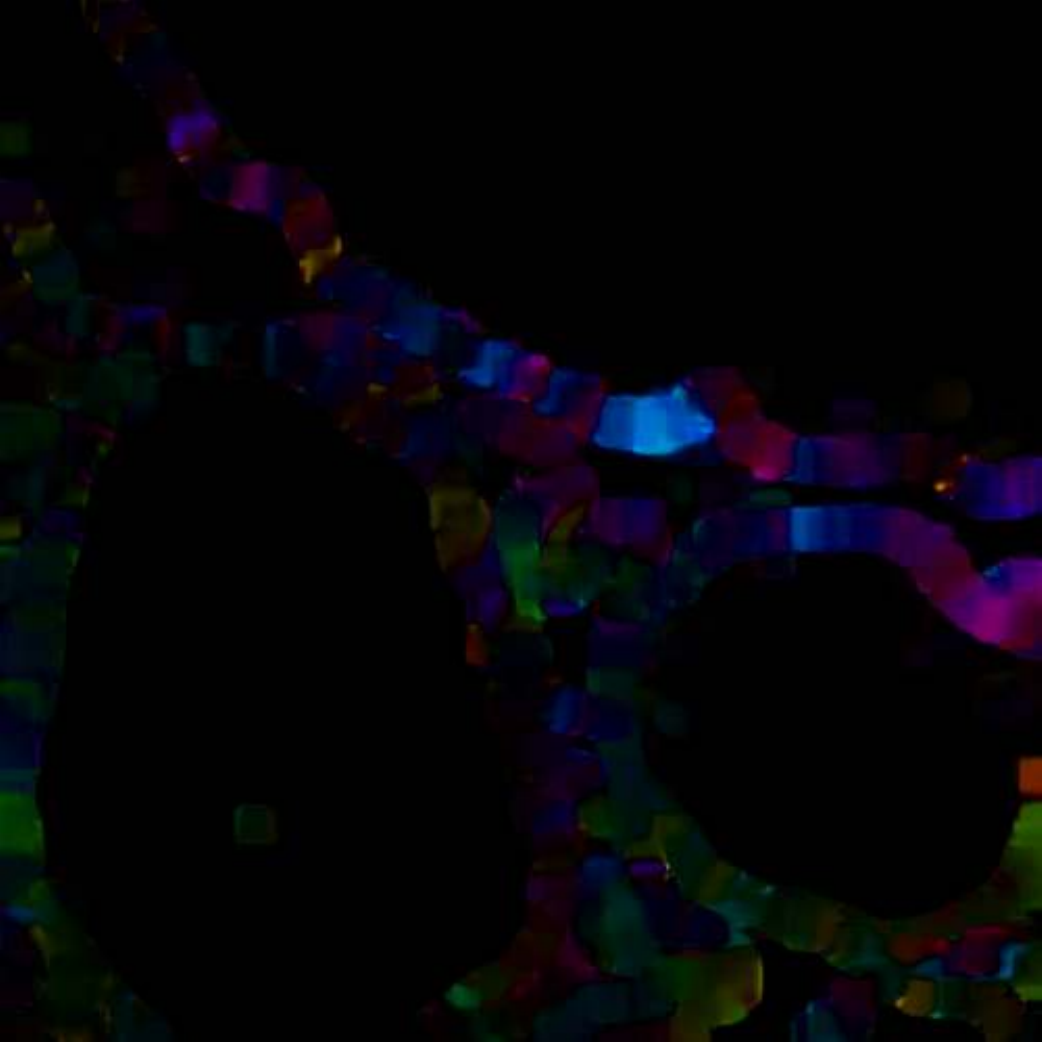}
\\

& \multicolumn{1}{c}{Image} & \multicolumn{1}{c}{Dice} & \multicolumn{1}{c}{PSNR} & \multicolumn{1}{c}{SSIM} & \multicolumn{1}{c}{Flow} \\

\end{tabular}

\caption{Evaluation of EM data set with image measures, part~1. Continued in Fig.~\ref{fig:eval-em-2}. All scale bars are \SI{10}{\micro\meter}.}
\label{fig:eval-em-1}
\setlength{\tabcolsep}{\tabcolbackup}
\thisfloatpagestyle{empty} 
\end{figure*}
\begin{figure*}[!t]
\centering
\setlength{\myfig}{0.2\linewidth-0.001\linewidth-2pt-0.2em}
\setlength{\tabcolsep}{1pt}
\begin{tabular}{m{1em} m{\myfig} m{\myfig} m{\myfig} m{\myfig} m{\myfig}}
& \multicolumn{1}{c}{Image} & \multicolumn{1}{c}{Dice} & \multicolumn{1}{c}{PSNR} & \multicolumn{1}{c}{SSIM} & \multicolumn{1}{c}{Flow} \\

\rot{\myfig}{\elastix} &
\scalebarme{%
\includegraphics[height=\myfig]{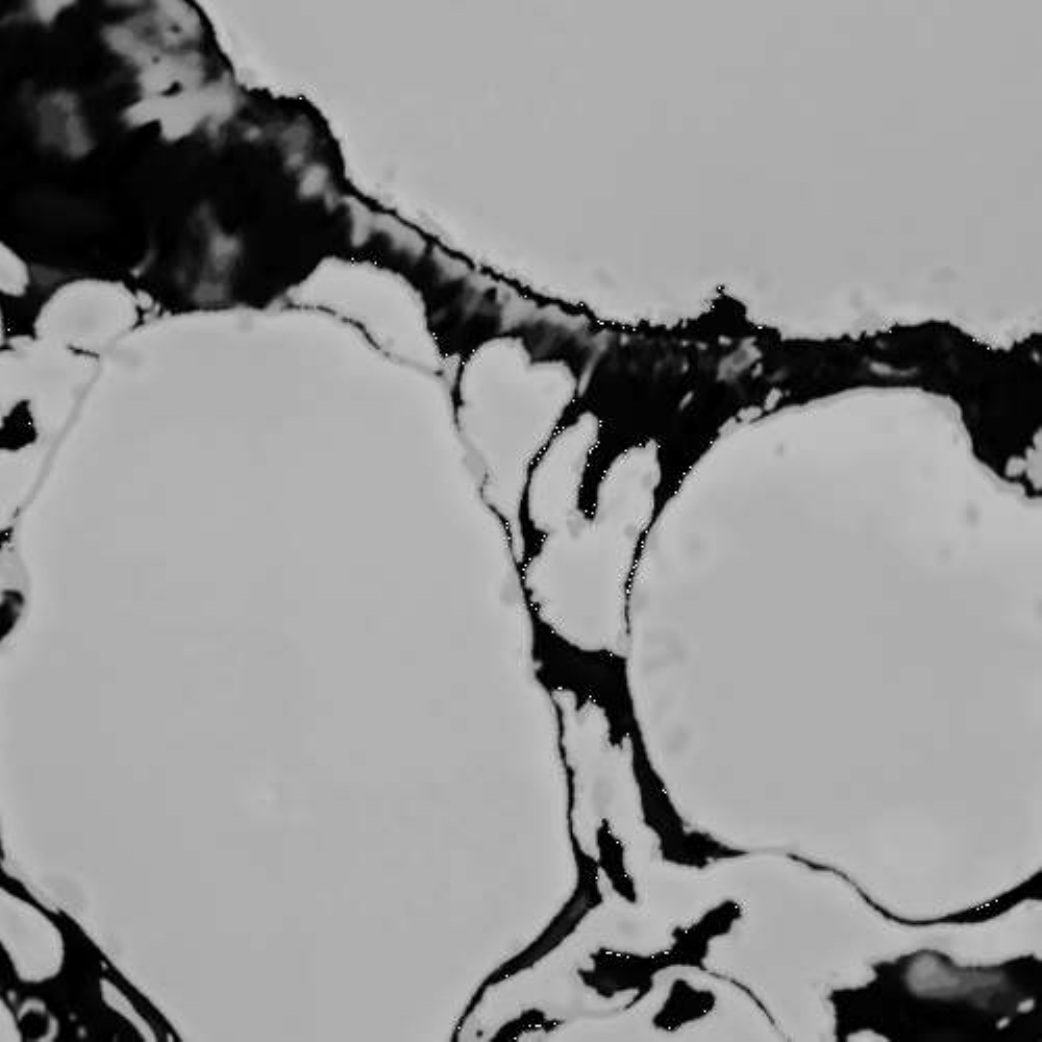}}{%
\includegraphics[width=\myfig]{image/scalebars/scale_em-500.png}} &
\includegraphics[height=\myfig]{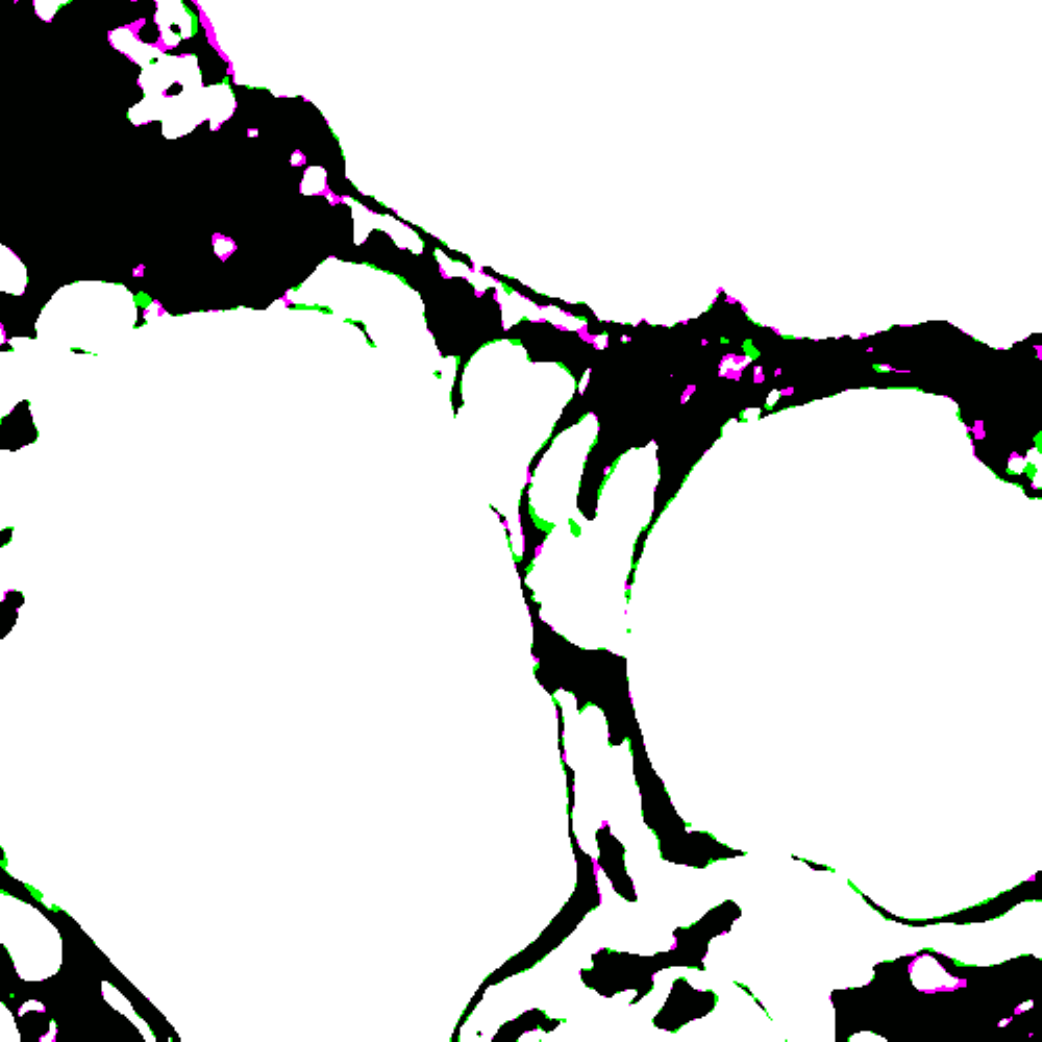} &
\includegraphics[height=\myfig]{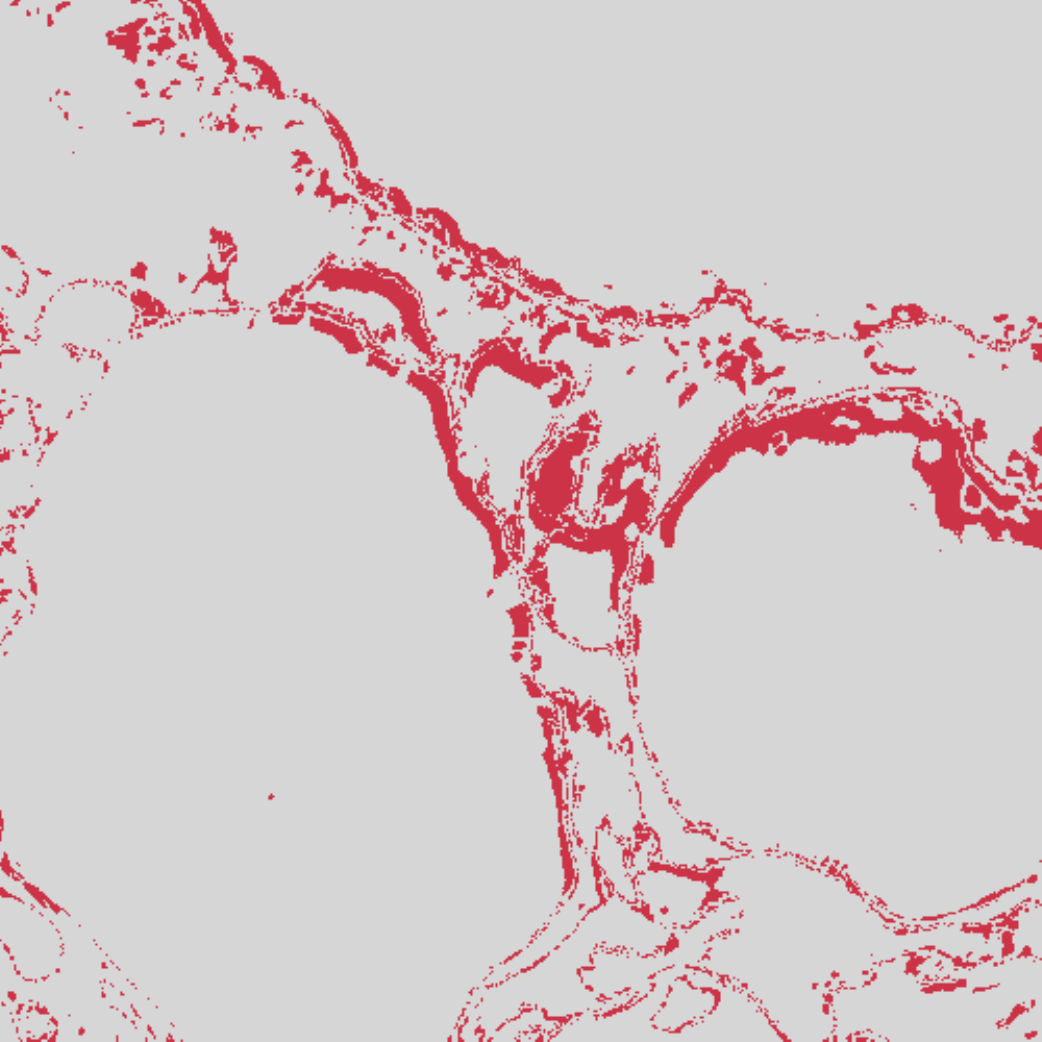} &
\includegraphics[height=\myfig]{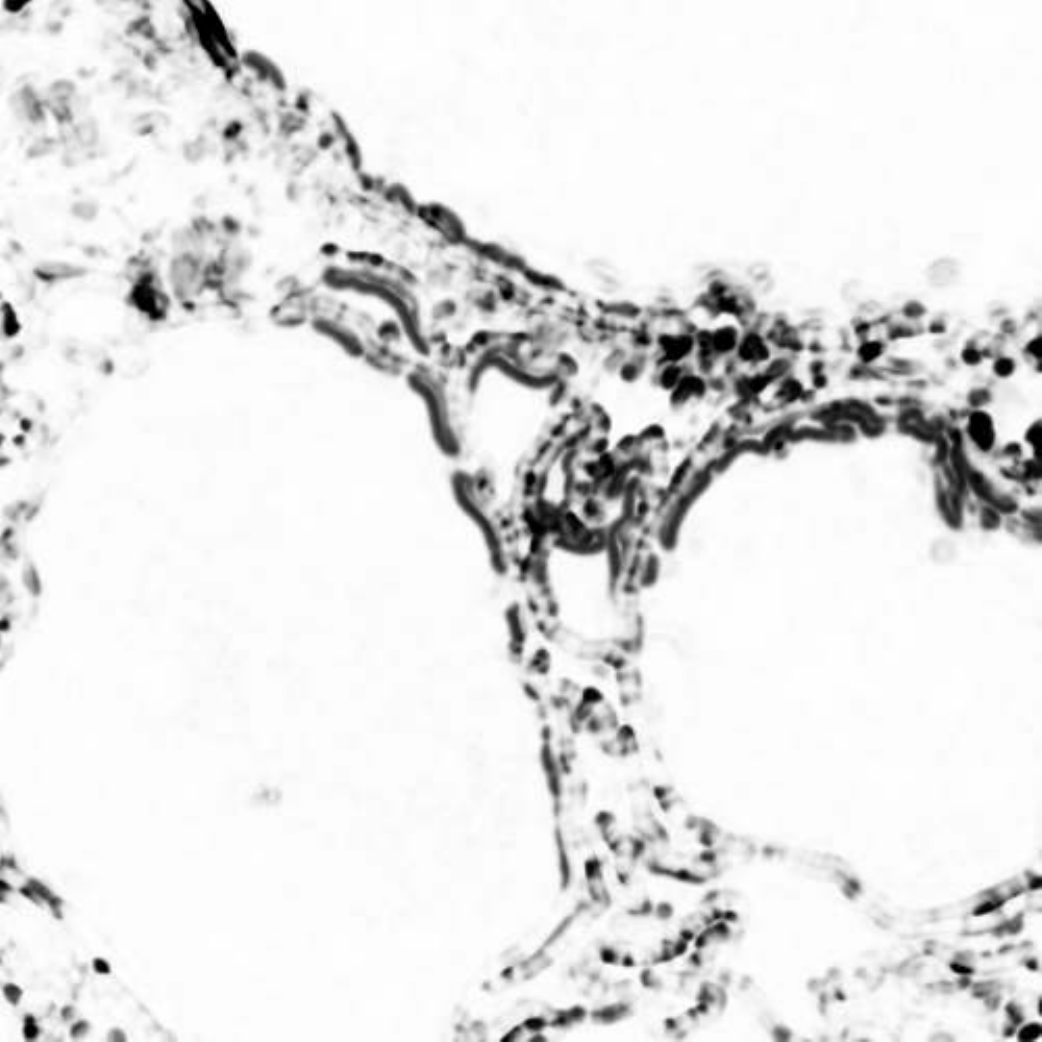} &
\includegraphics[height=\myfig]{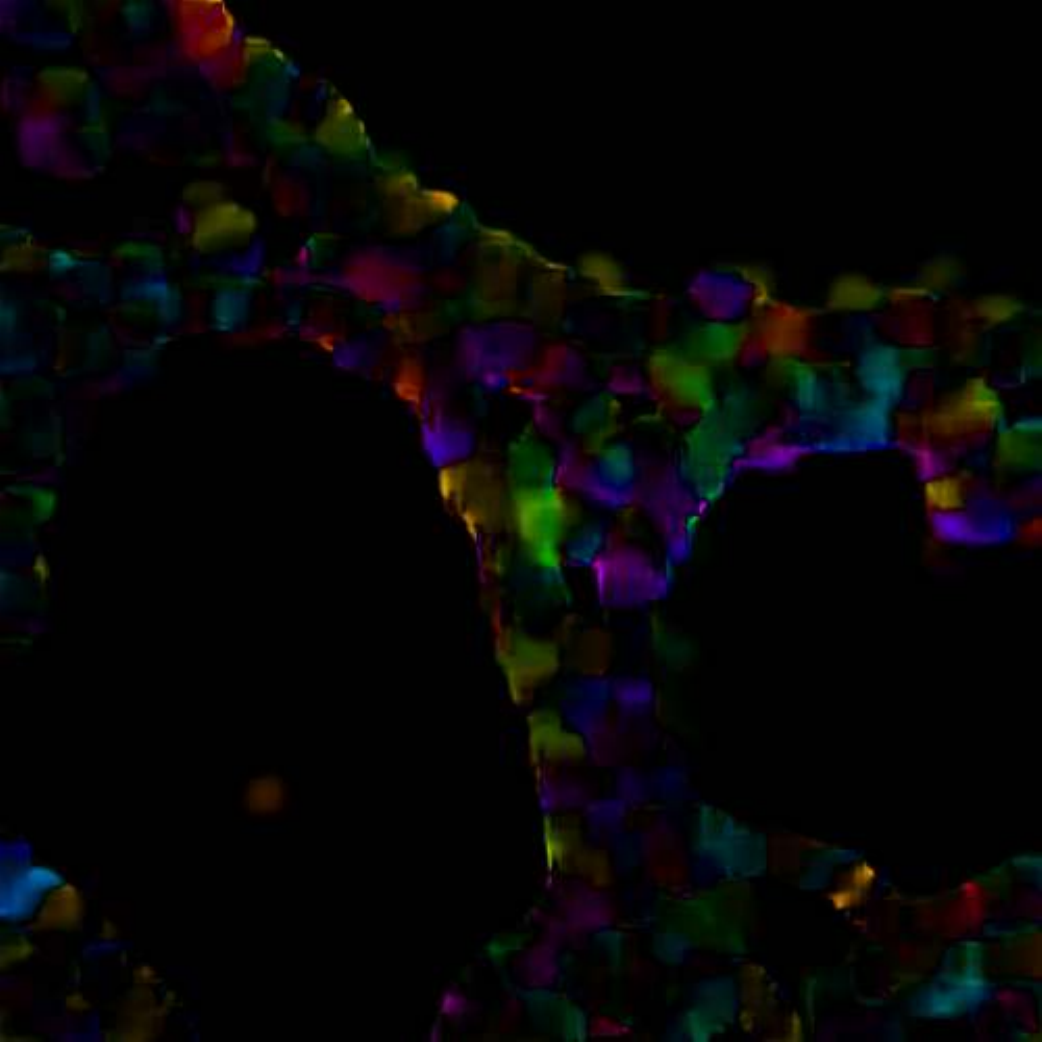}
\\

\rot{\myfig}{Local only} &
\scalebarme{%
\includegraphics[height=\myfig]{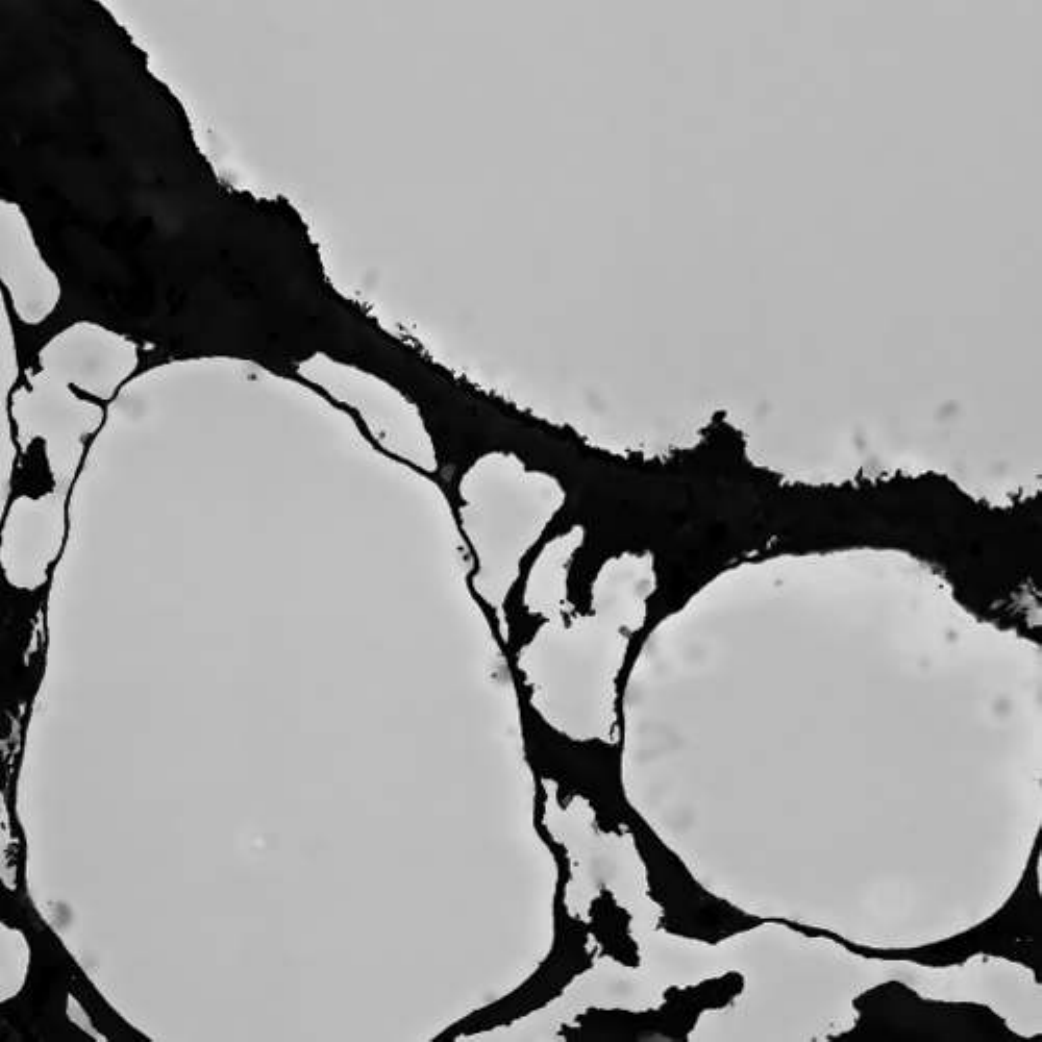}}{%
\includegraphics[width=\myfig]{image/scalebars/scale_em-500.png}} &
\includegraphics[height=\myfig]{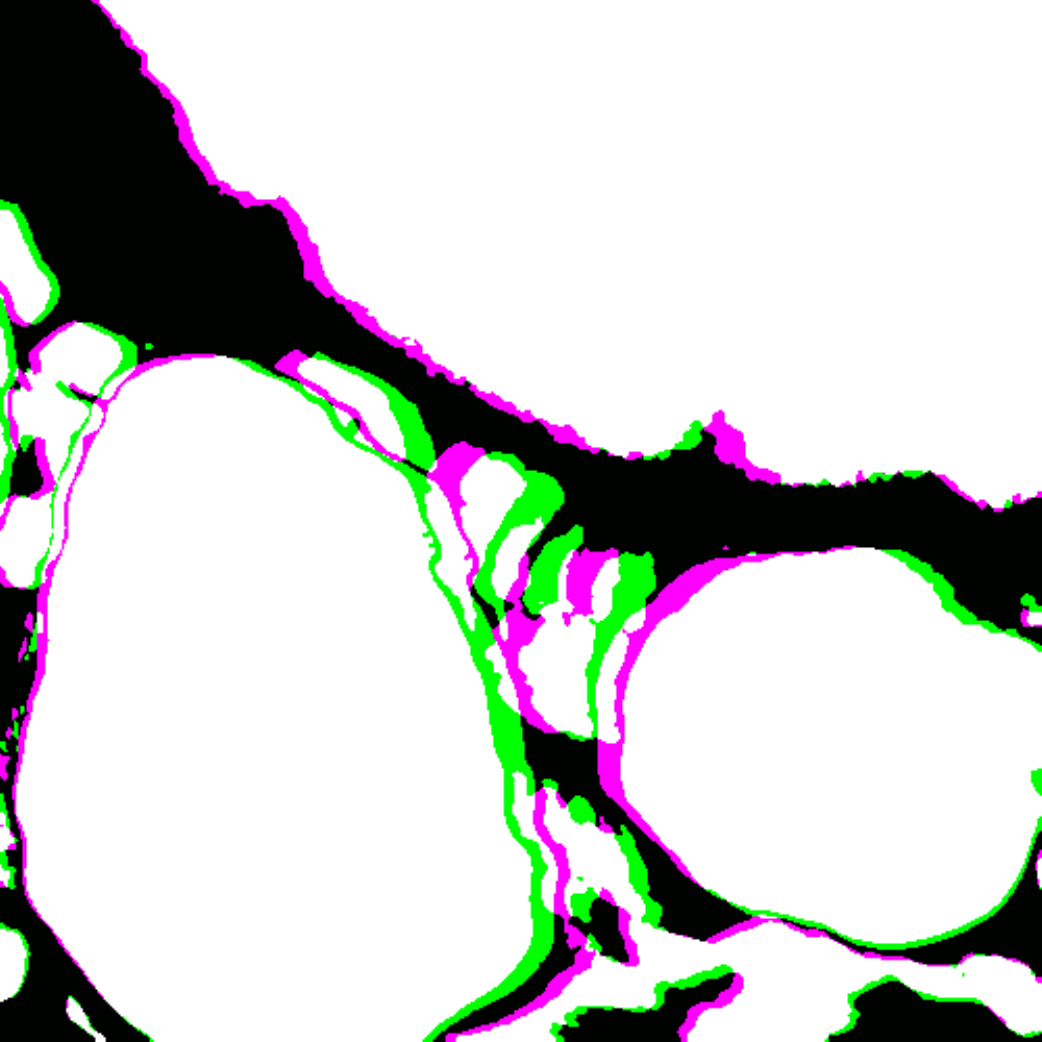} &
\includegraphics[height=\myfig]{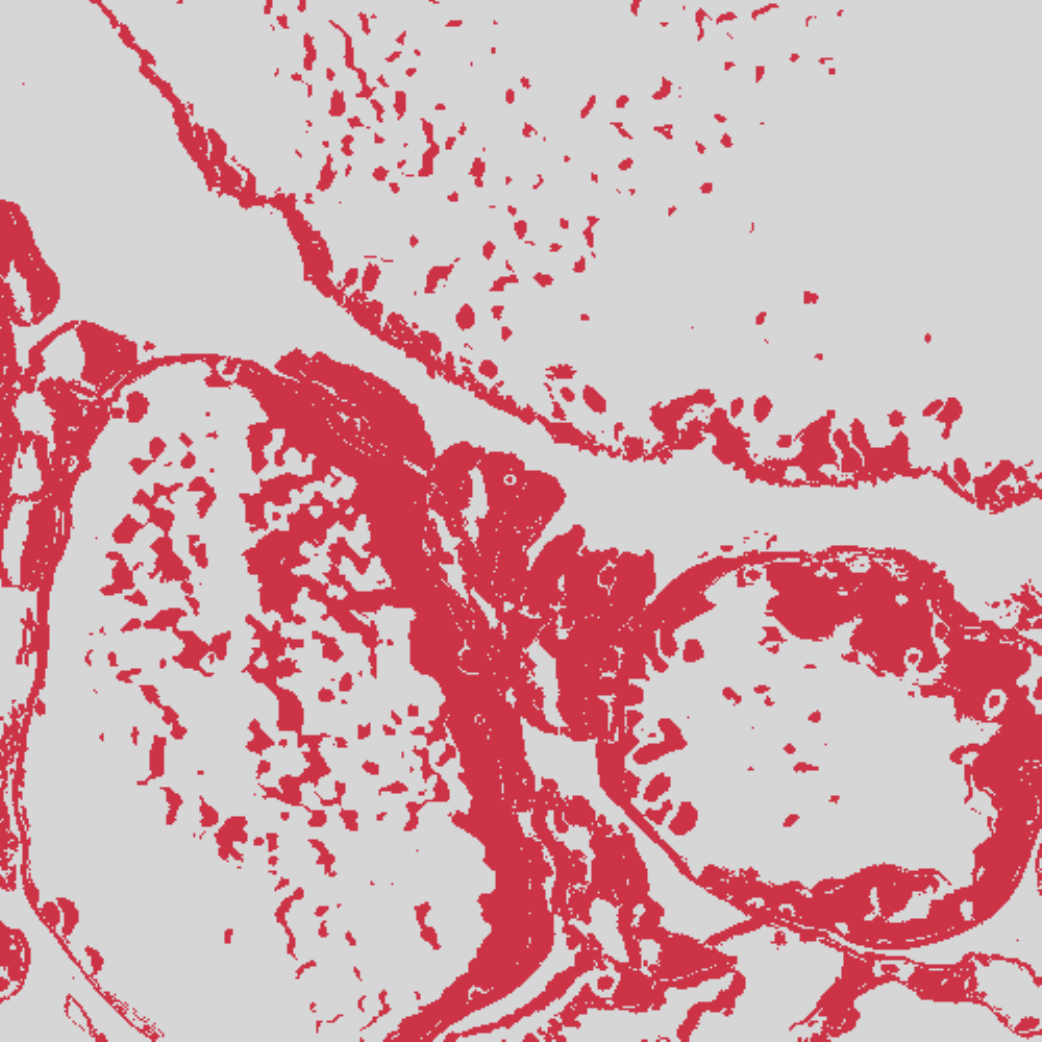} &
\includegraphics[height=\myfig]{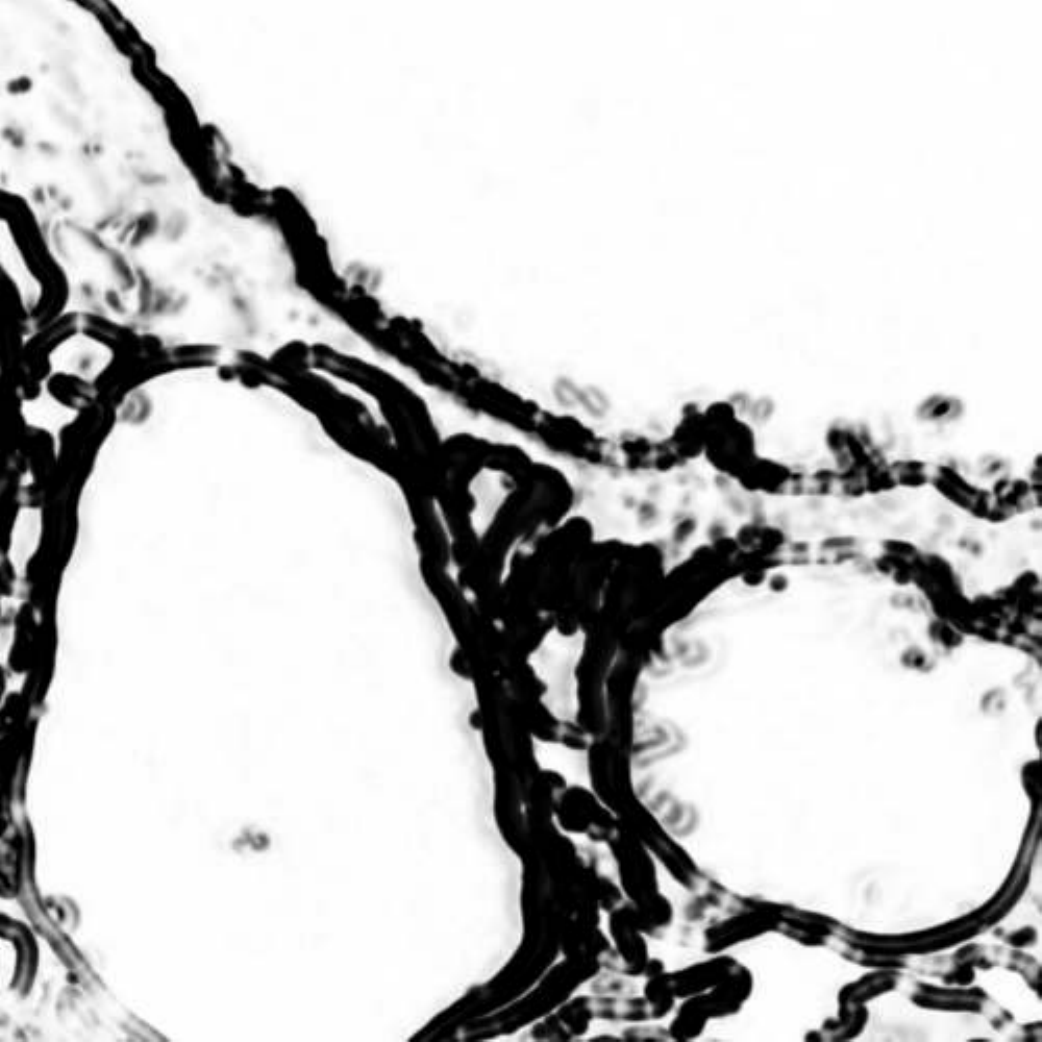} &
\includegraphics[height=\myfig]{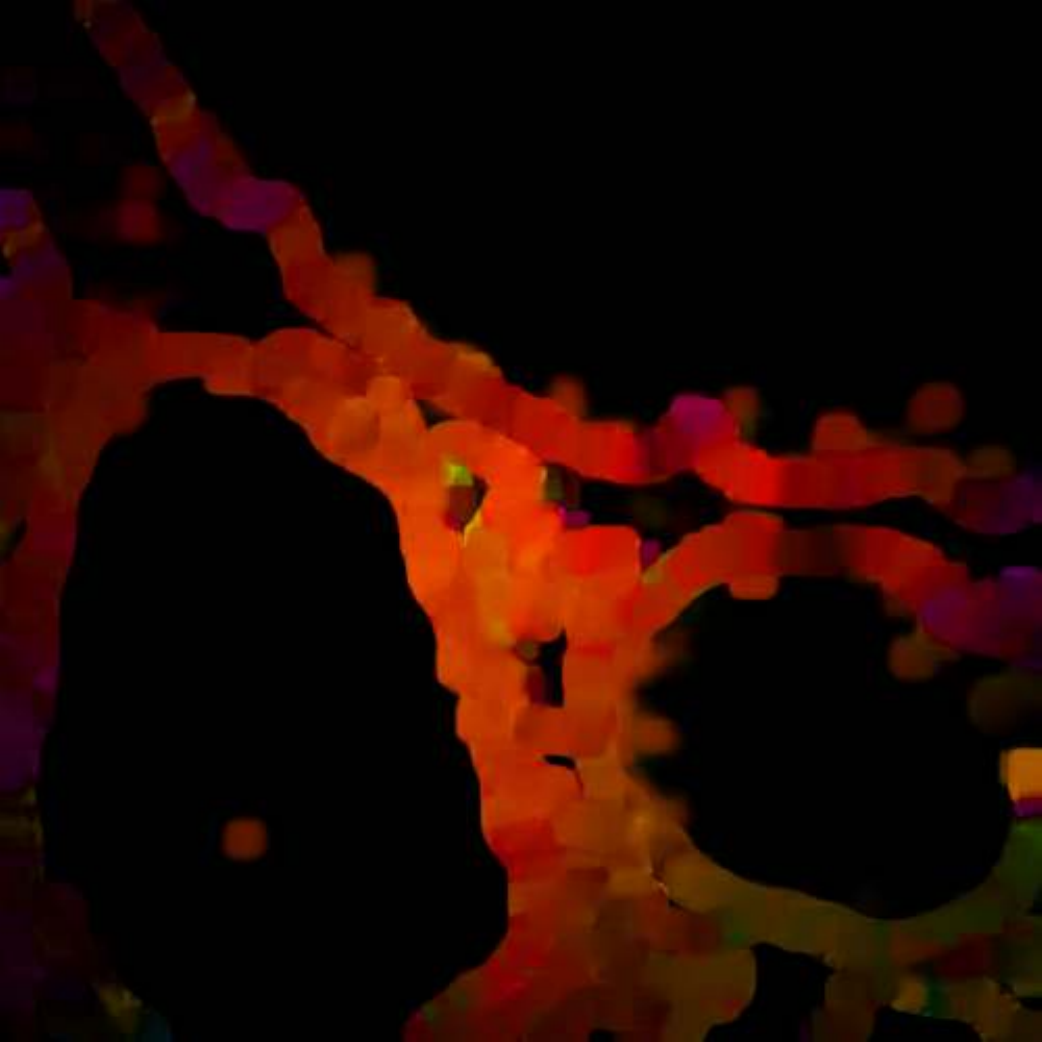}
\\

\rot{\myfig}{Gr.~truth, norm.} &
\scalebarme{%
\includegraphics[height=\myfig]{image/EM/eval/em_orig-norm/crop_gray_norm_0500-dump.pdf}}{%
\includegraphics[width=\myfig]{image/scalebars/scale_em-500.png}} &
\includegraphics[height=\myfig]{image/EM/eval/em_orig-norm/crop_pair__em-orig-norm_499-500_.pdf} &
\includegraphics[height=\myfig]{image/EM/eval/em_orig-norm/crop_psnr__em-orig-norm_499-500_.pdf} &
\includegraphics[height=\myfig]{image/EM/eval/em_orig-norm/crop_ssim_full__em-orig-norm_499-500_.pdf} &
\includegraphics[height=\myfig]{image/EM/eval/em_orig-norm/crop_flow_CROP__em-orig-norm_499-500_.pdf}
\\

\rot{\myfig}{Gr.~truth, orig.} &
\scalebarme{%
\includegraphics[height=\myfig]{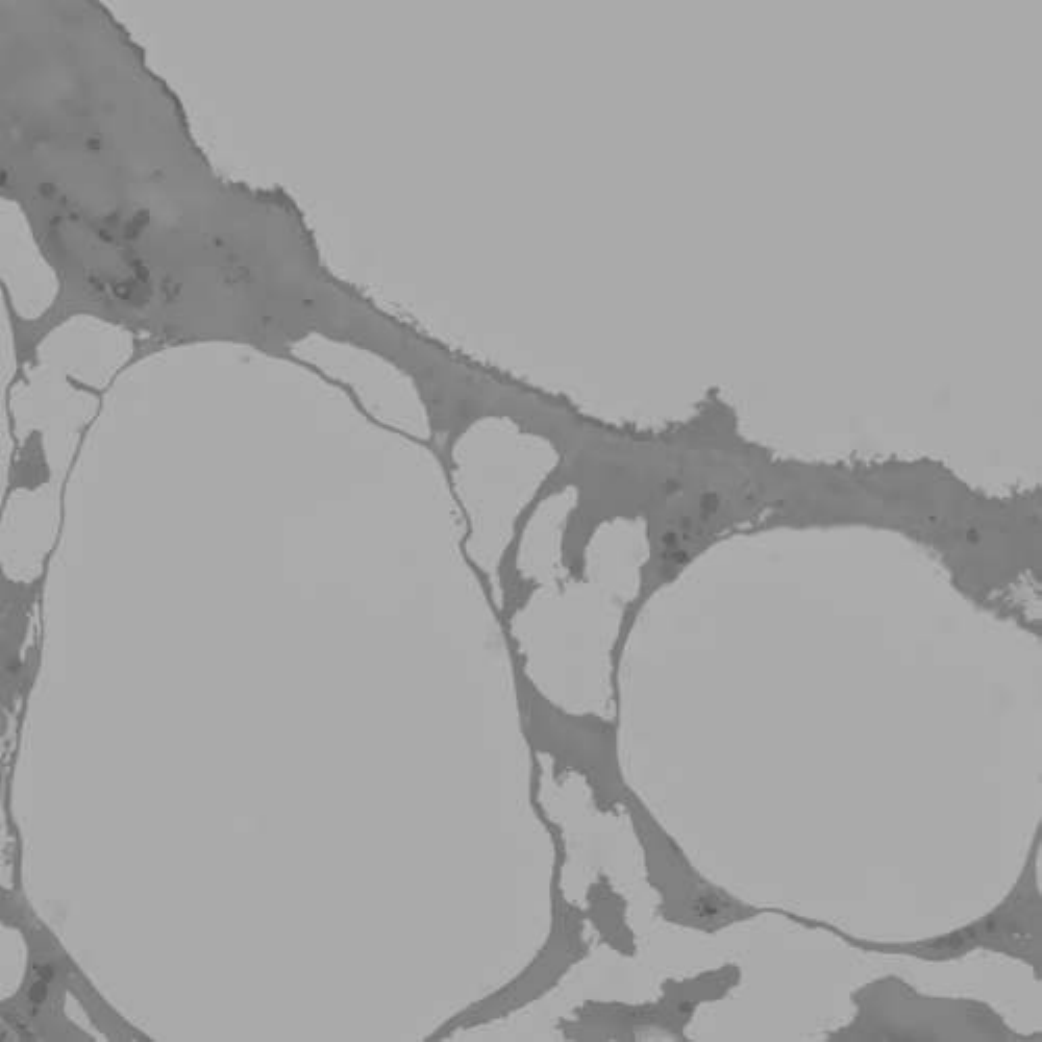}}{%
\includegraphics[width=\myfig]{image/scalebars/scale_em-500.png}} &
\includegraphics[height=\myfig]{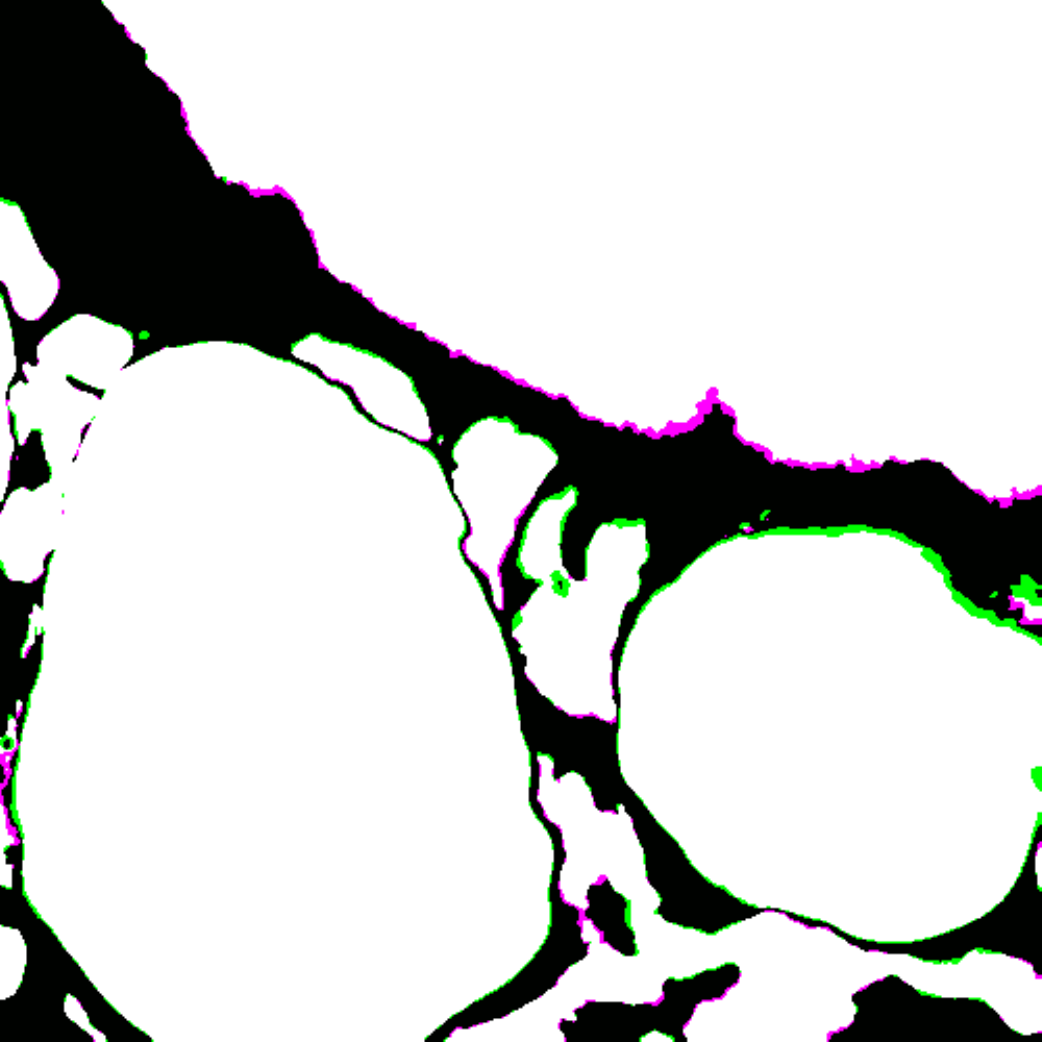} &
\includegraphics[height=\myfig]{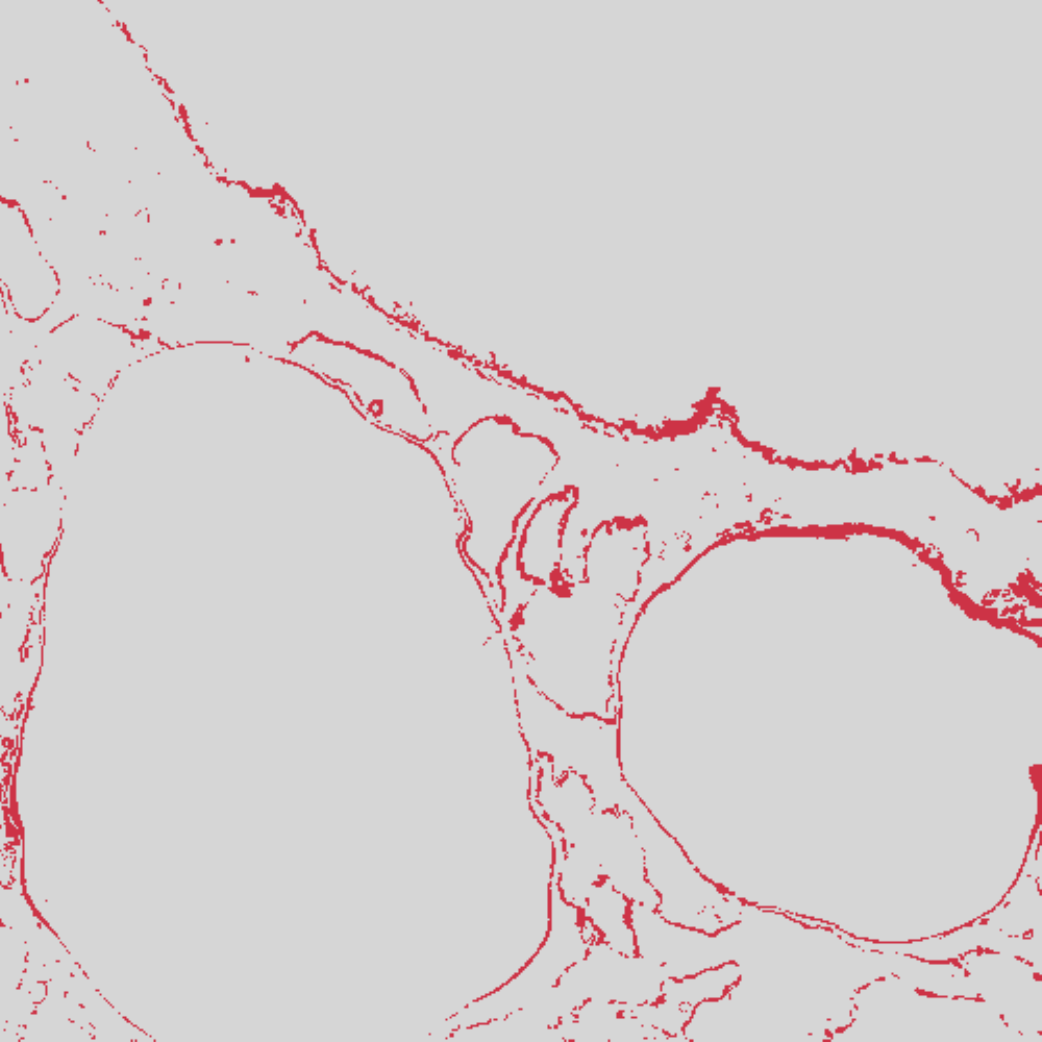} &
\includegraphics[height=\myfig]{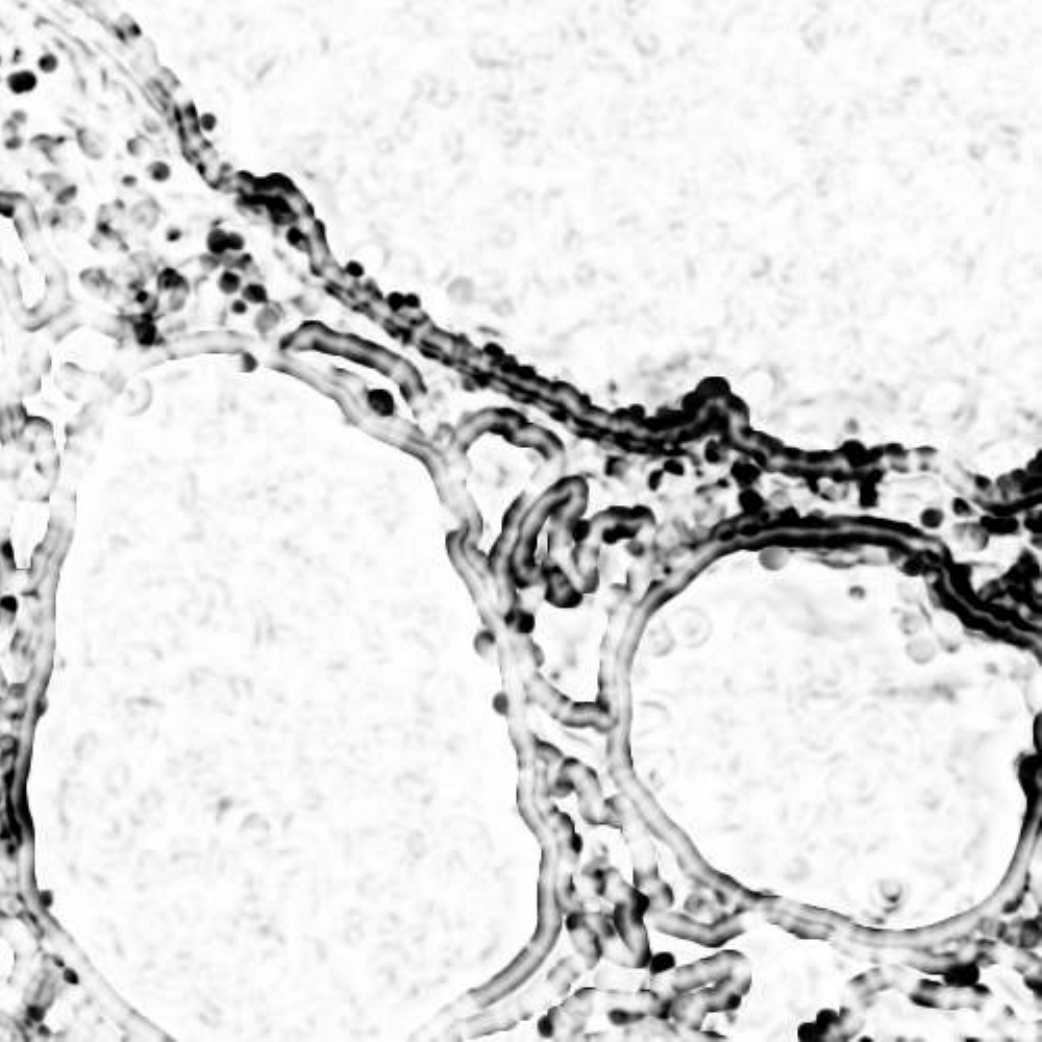} &
\includegraphics[height=\myfig]{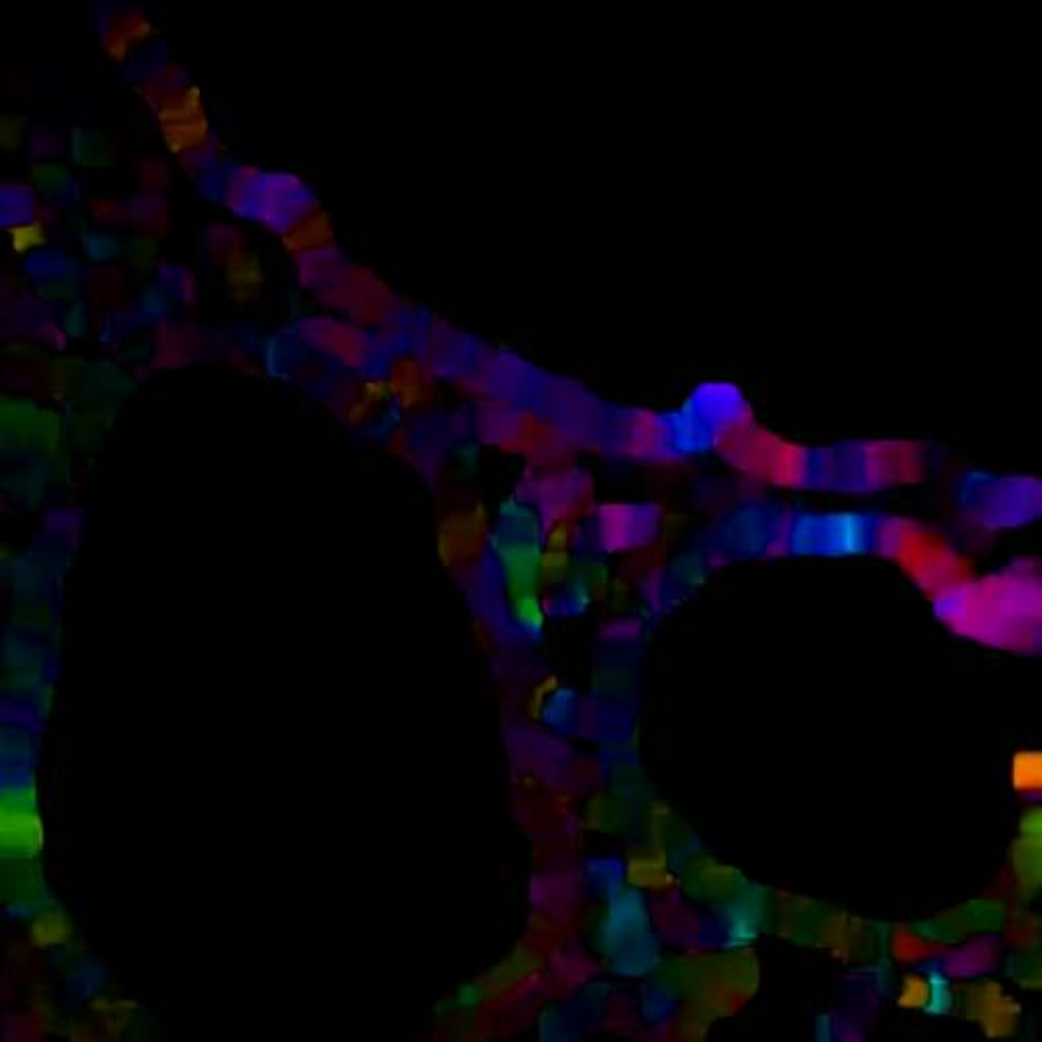}
\\

& \multicolumn{1}{c}{Image} & \multicolumn{1}{c}{Dice} & \multicolumn{1}{c}{PSNR} & \multicolumn{1}{c}{SSIM} & \multicolumn{1}{c}{Flow} \\

\end{tabular}
\caption{Evaluation of EM data set with image measures, part~2. Continued from Fig.~\ref{fig:eval-em-1}. All scale bars are \SI{10}{\micro\meter}.}
\label{fig:eval-em-2}
\setlength{\tabcolsep}{\tabcolbackup}
\end{figure*}

\begin{figure*}[p]
\centering
\setlength{\myfig}{0.2\linewidth-0.001\linewidth-2pt-0.2em}
\setlength{\tabcolsep}{1pt}
\begin{tabular}{m{1em} m{\myfig} m{\myfig} m{\myfig} m{\myfig} m{\myfig}}
& \multicolumn{1}{c}{Image} & \multicolumn{1}{c}{Dice} & \multicolumn{1}{c}{PSNR} & \multicolumn{1}{c}{SSIM} & \multicolumn{1}{c}{Flow} \\

\rot{\myfig}{Rigid-SIFT} &
\scalebarme{%
\includegraphics[height=\myfig]{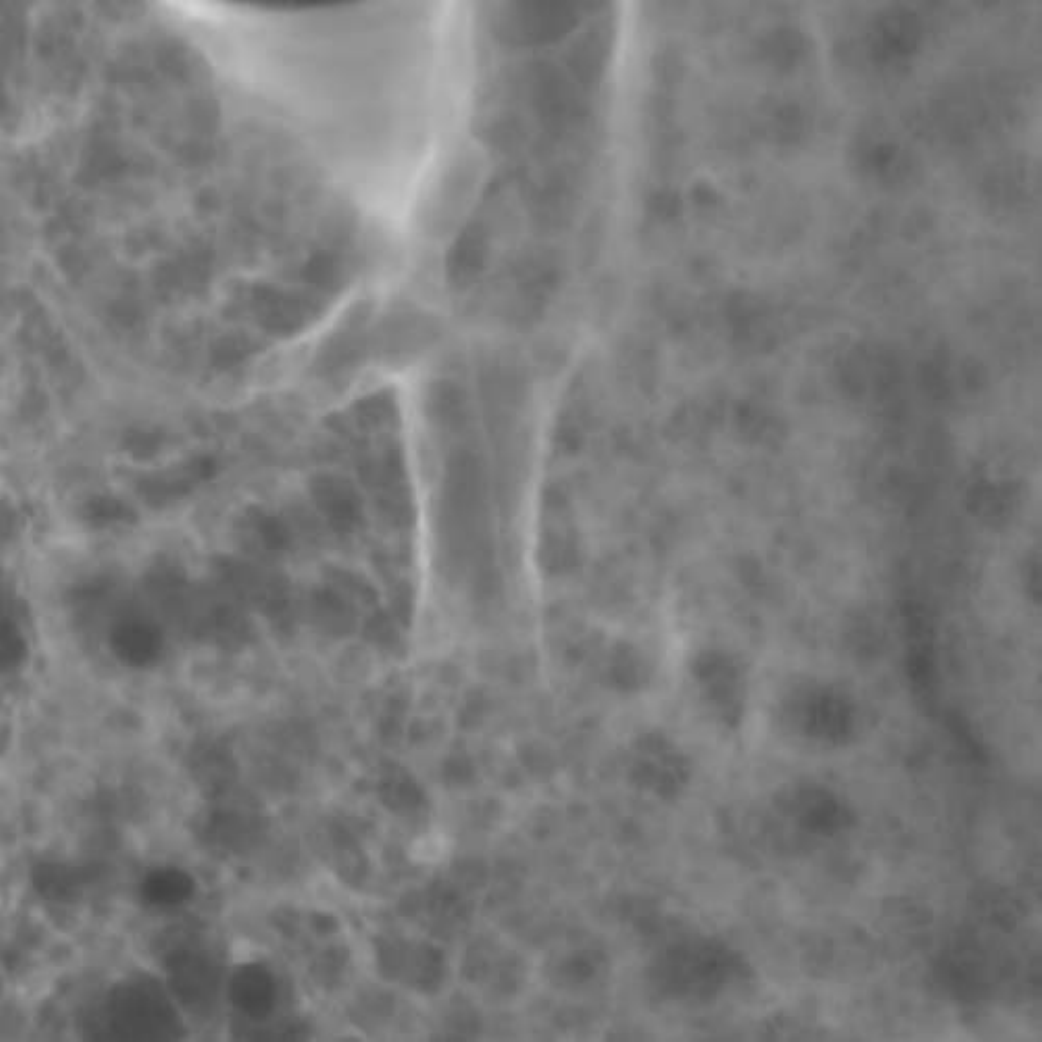}}{%
\includegraphics[width=\myfig]{image/scalebars/scale_ls-500.png}} &
\includegraphics[height=\myfig]{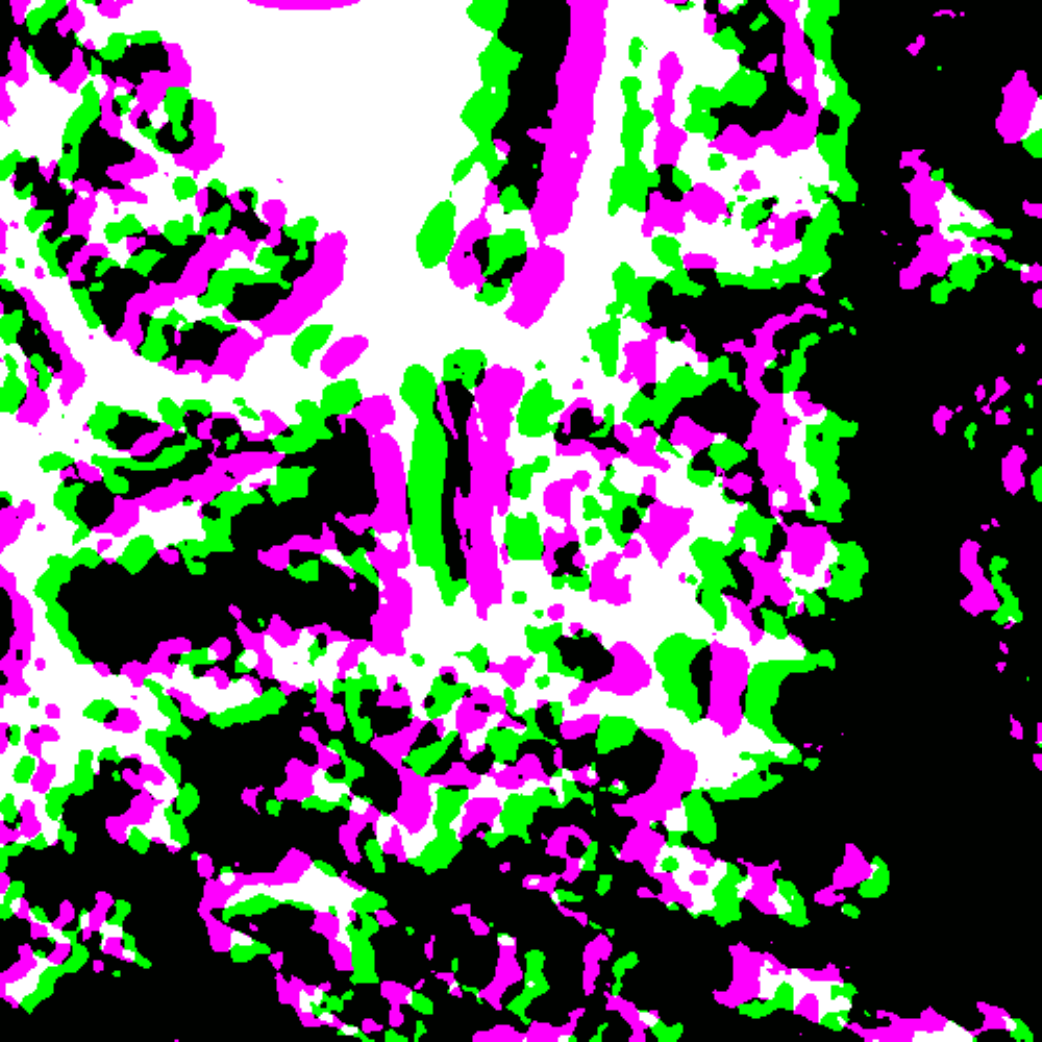} &
\includegraphics[height=\myfig]{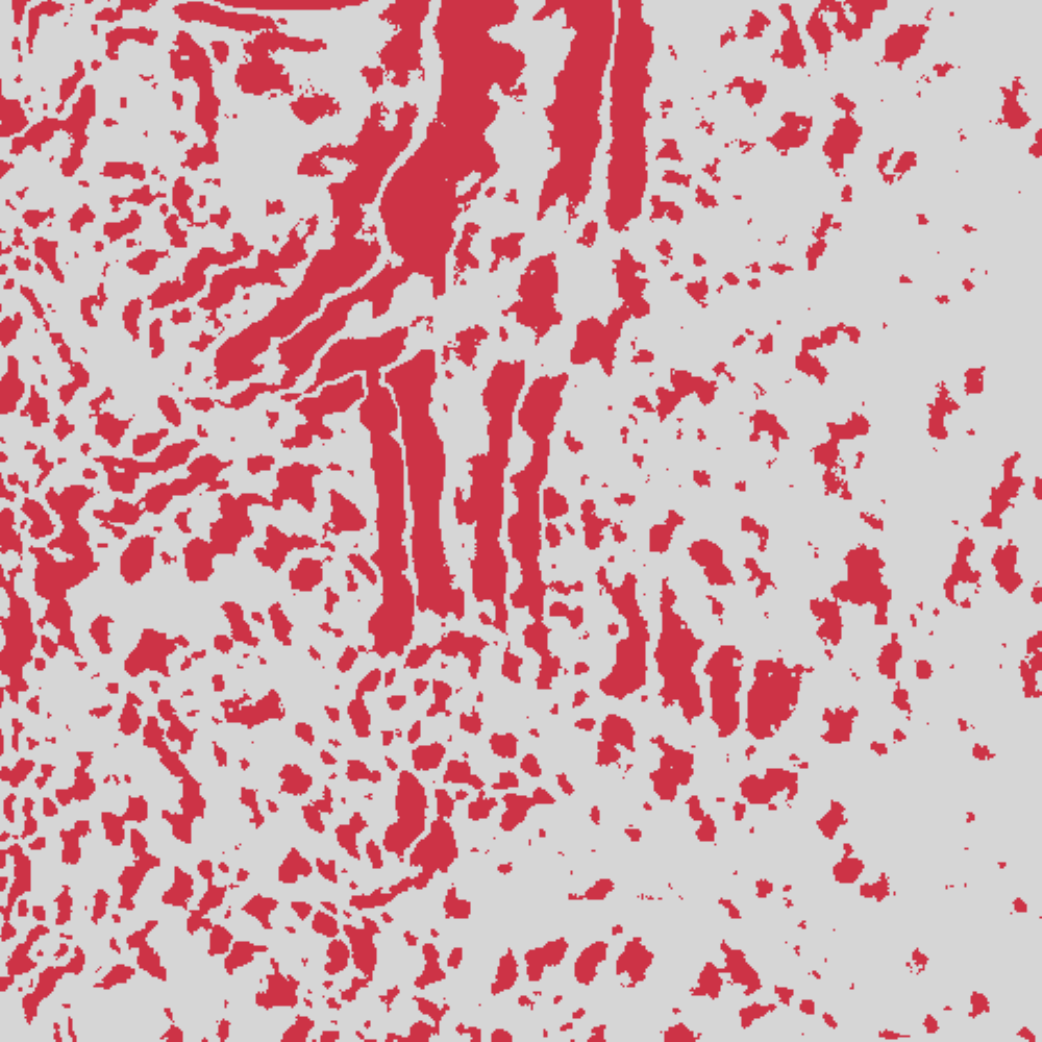} &
\includegraphics[height=\myfig]{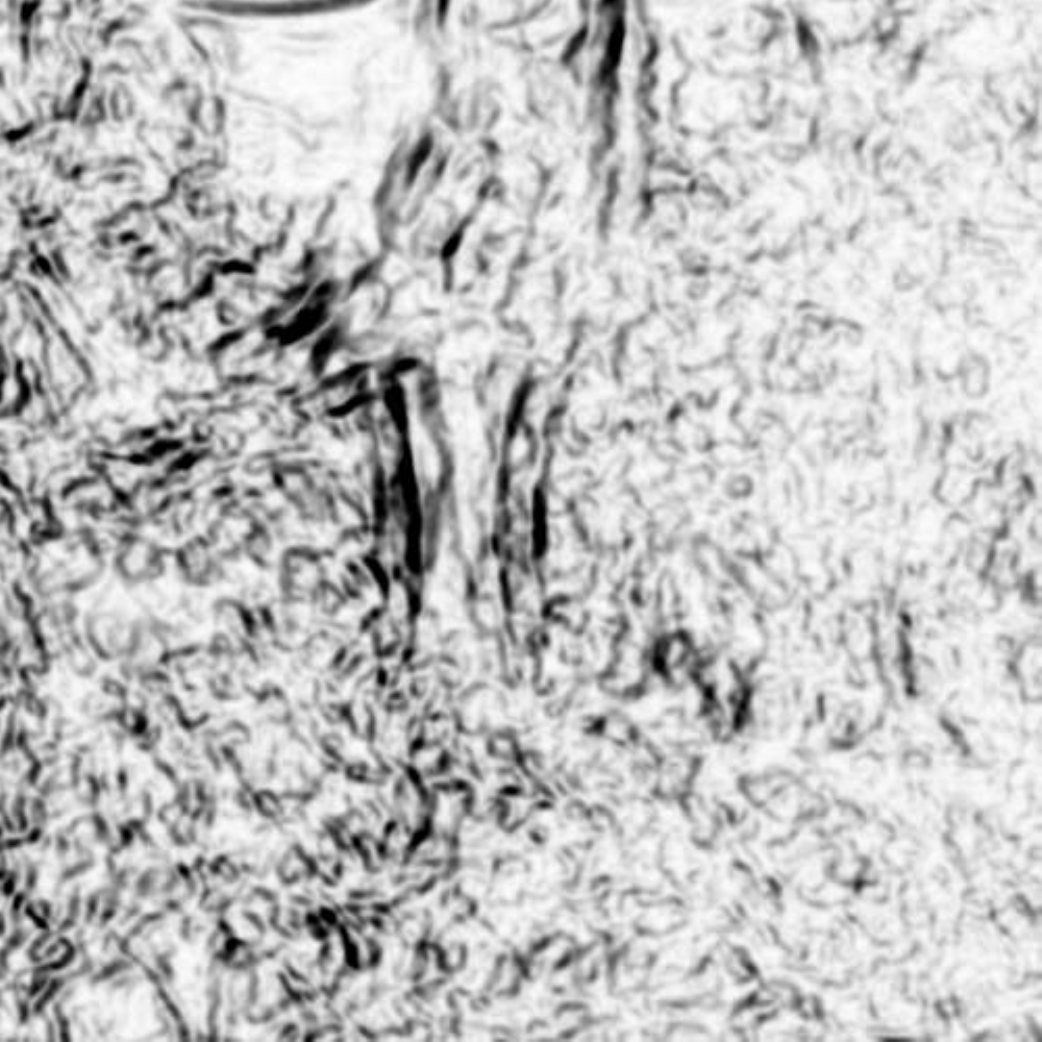} &
\includegraphics[height=\myfig]{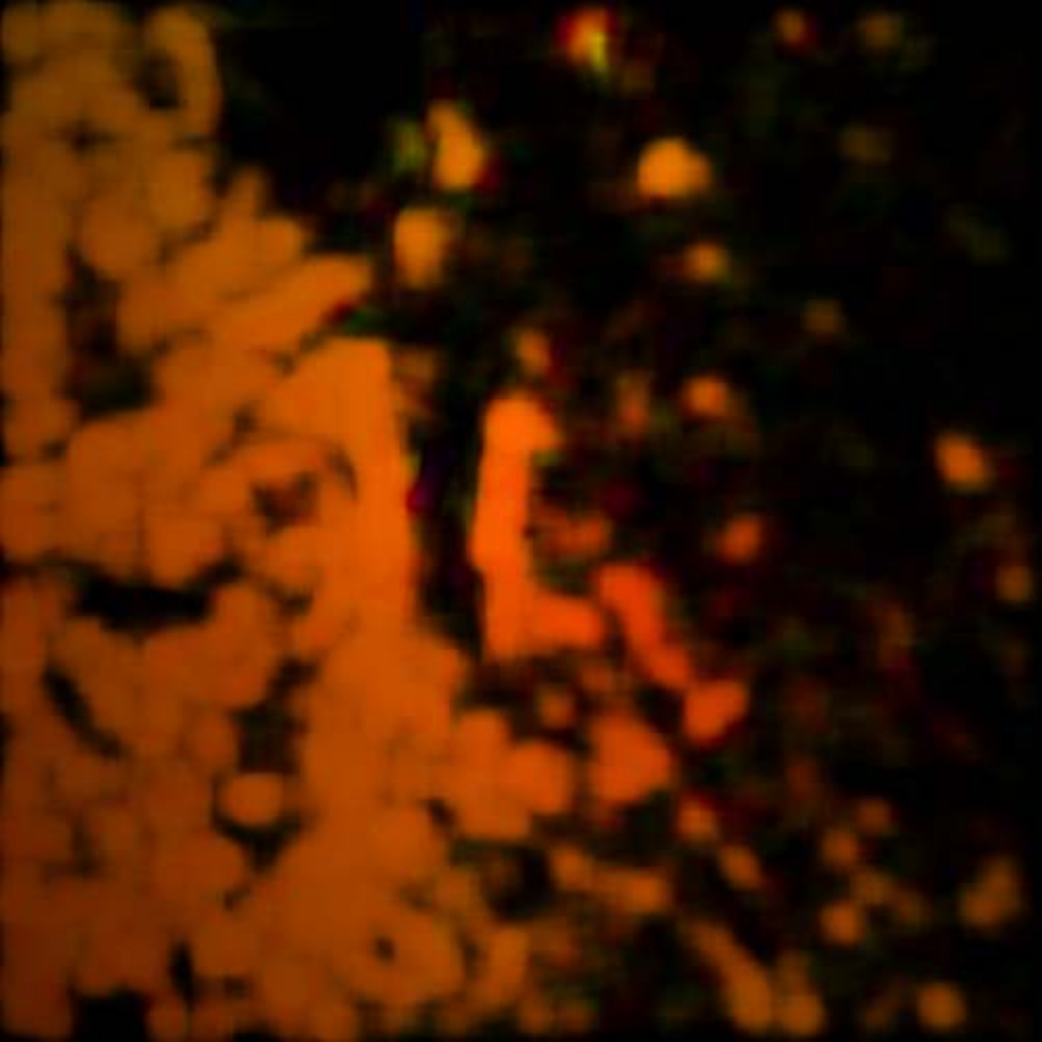}
\\

\rot{\myfig}{GS} &
\scalebarme{%
\includegraphics[height=\myfig]{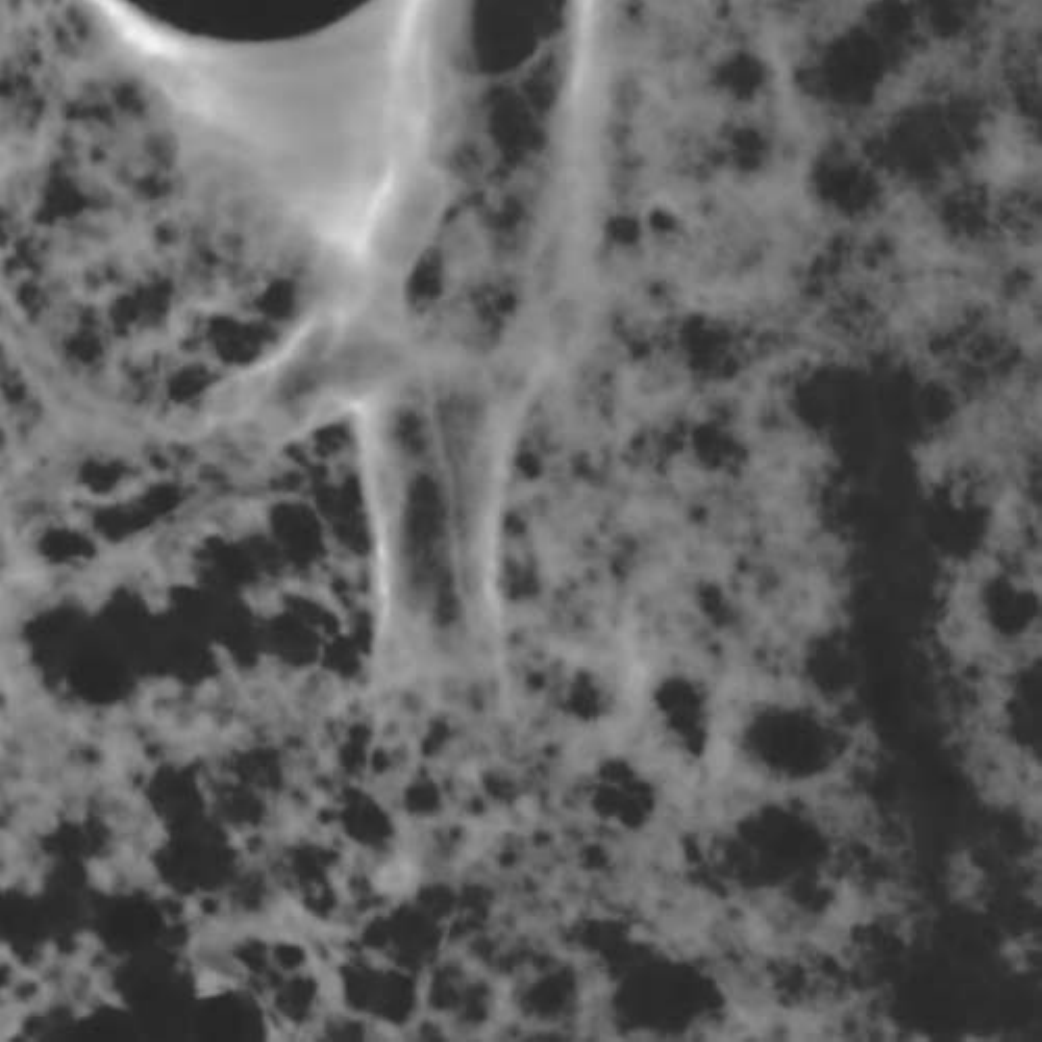}}{%
\includegraphics[width=\myfig]{image/scalebars/scale_ls-500.png}} &
\includegraphics[height=\myfig]{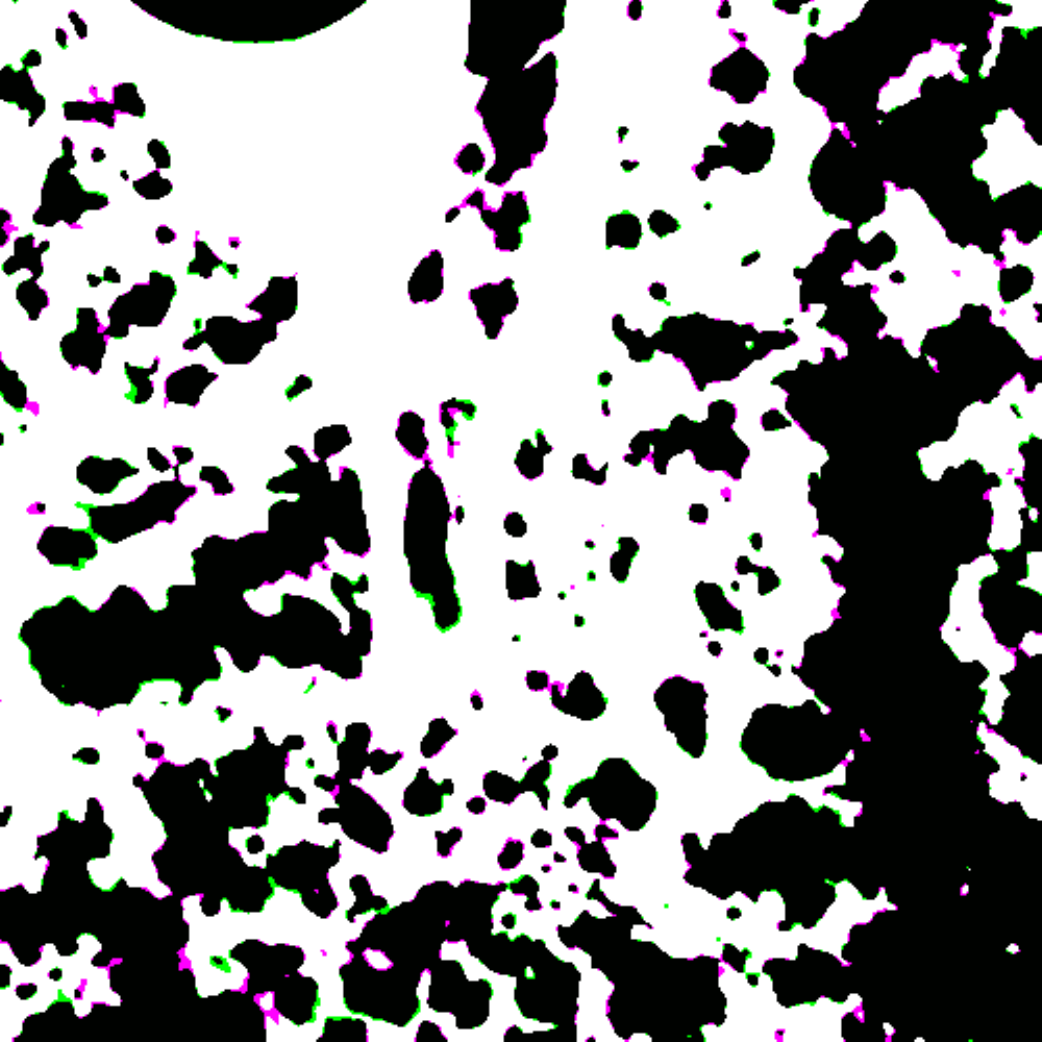} &
\includegraphics[height=\myfig]{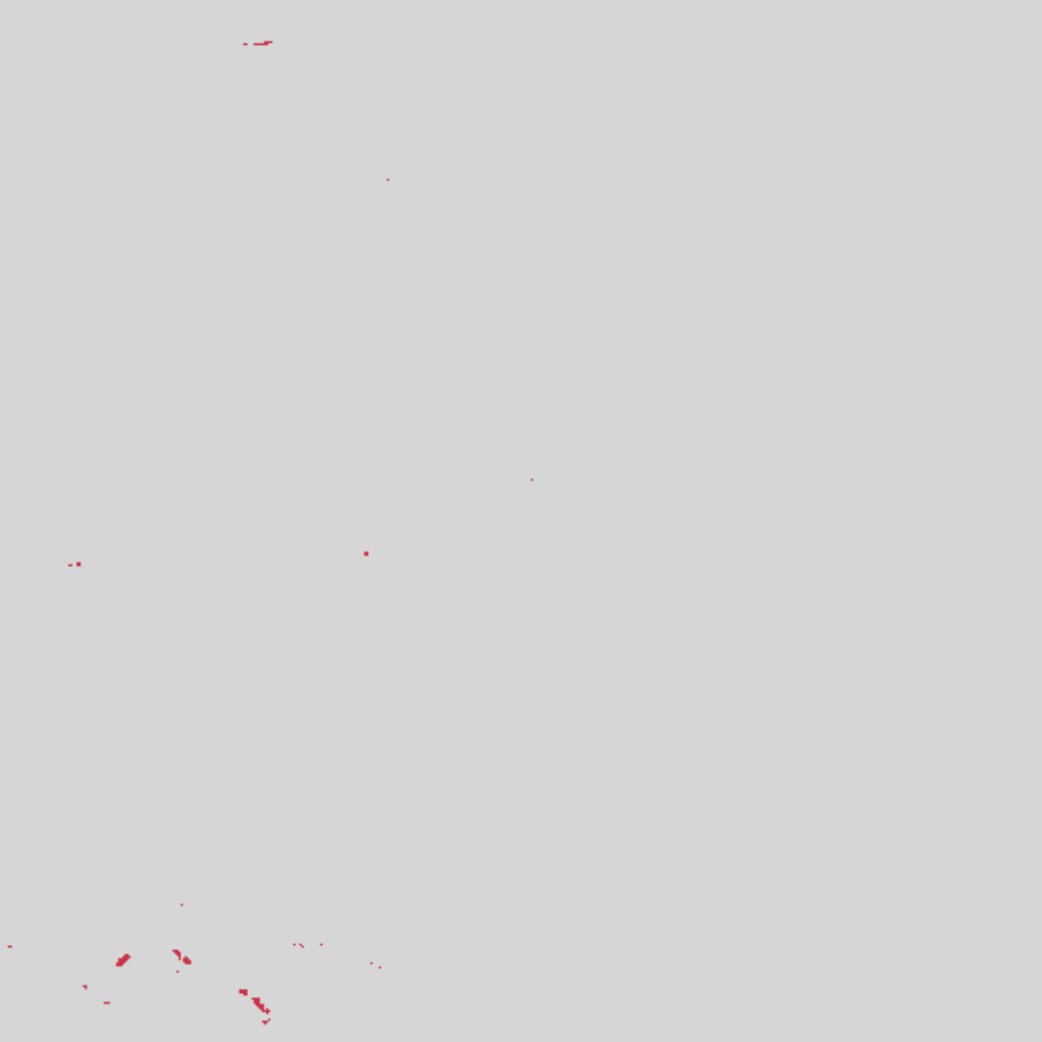} &
\includegraphics[height=\myfig]{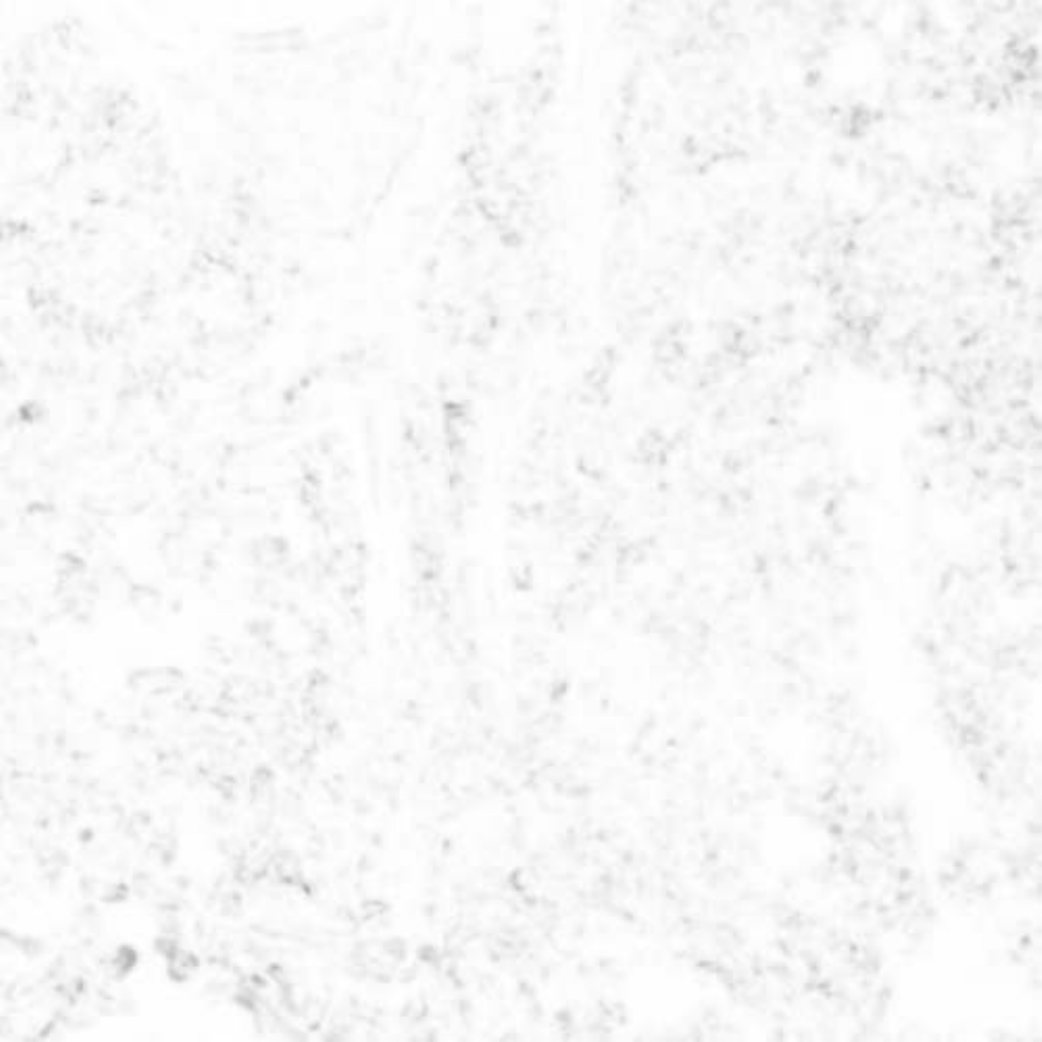} &
\includegraphics[height=\myfig]{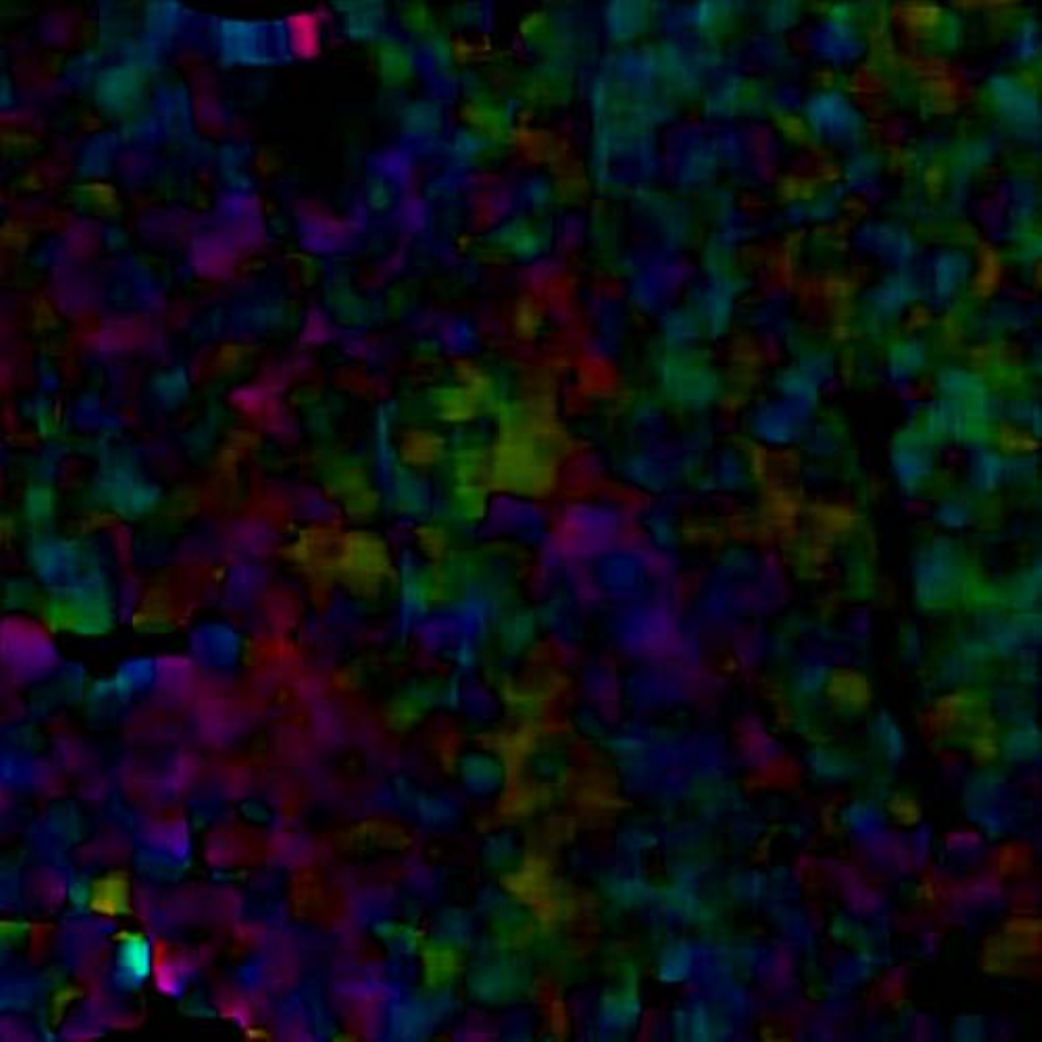}
\\

\rot{\myfig}{Blending} &
\scalebarme{%
\includegraphics[height=\myfig]{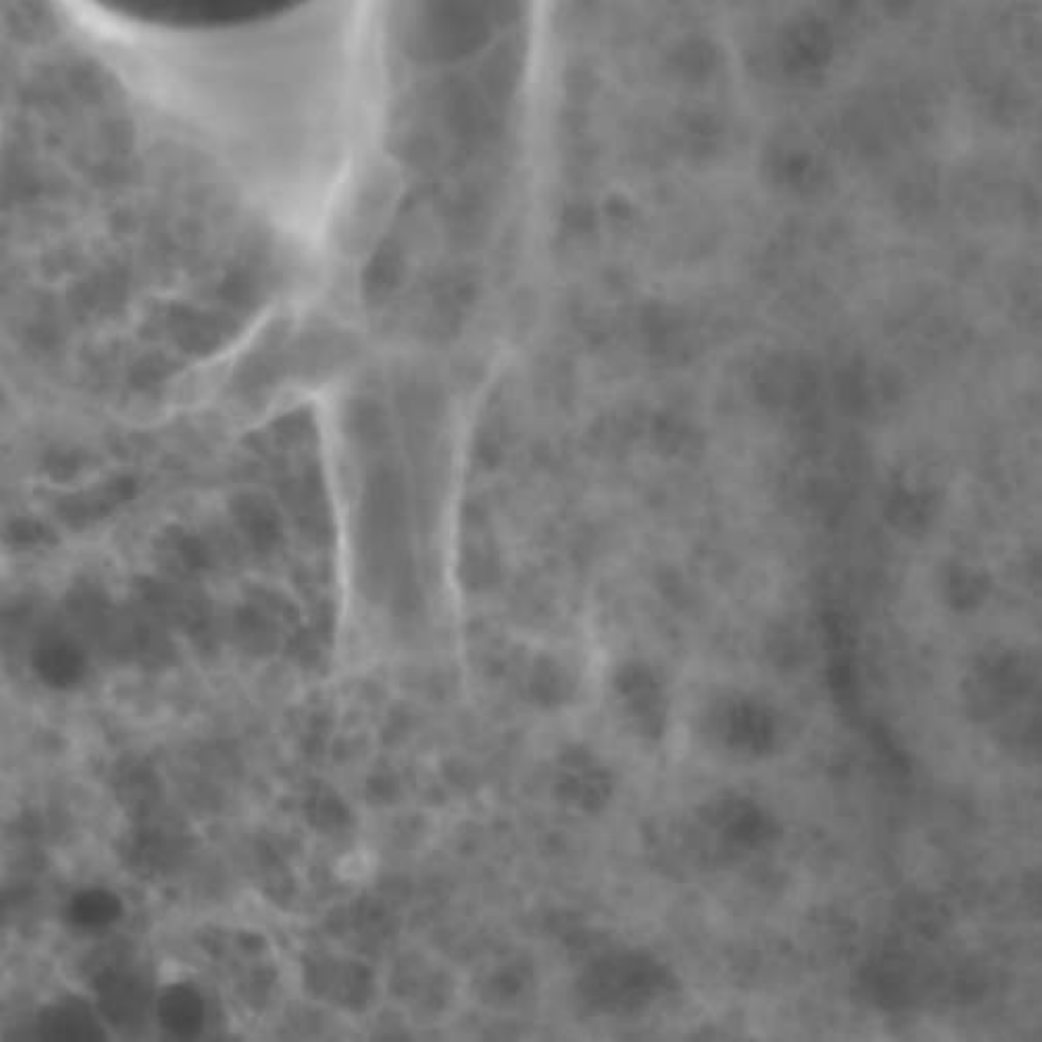}}{%
\includegraphics[width=\myfig]{image/scalebars/scale_ls-500.png}} &
\includegraphics[height=\myfig]{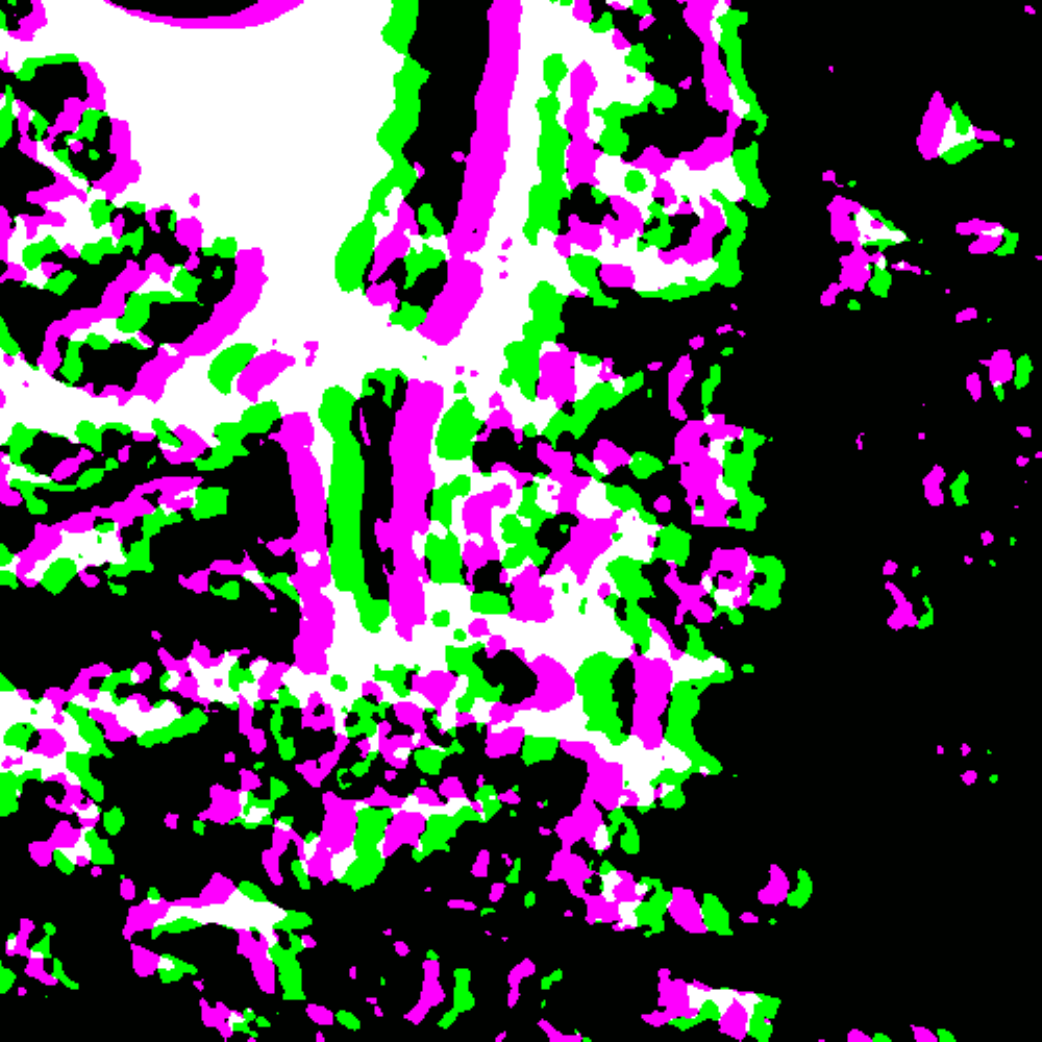} &
\includegraphics[height=\myfig]{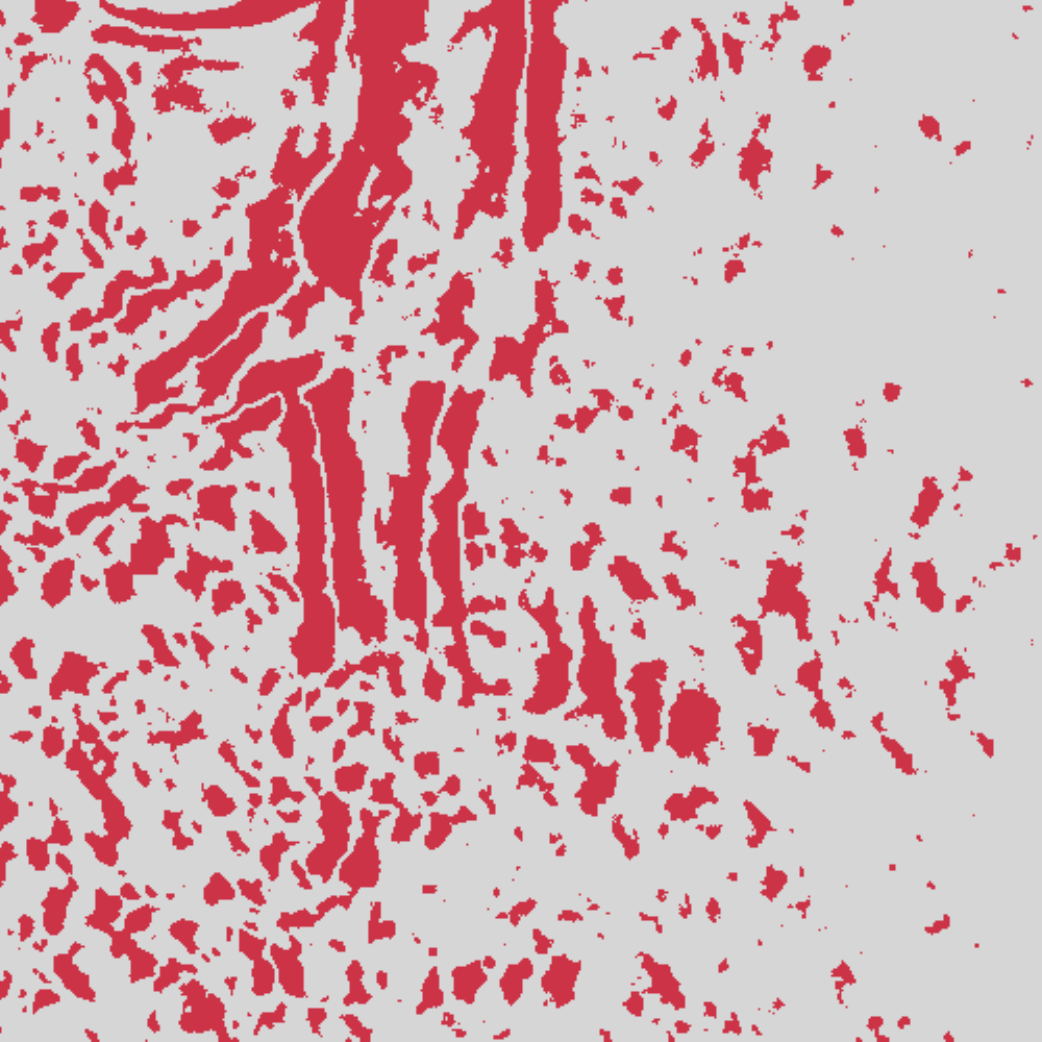} &
\includegraphics[height=\myfig]{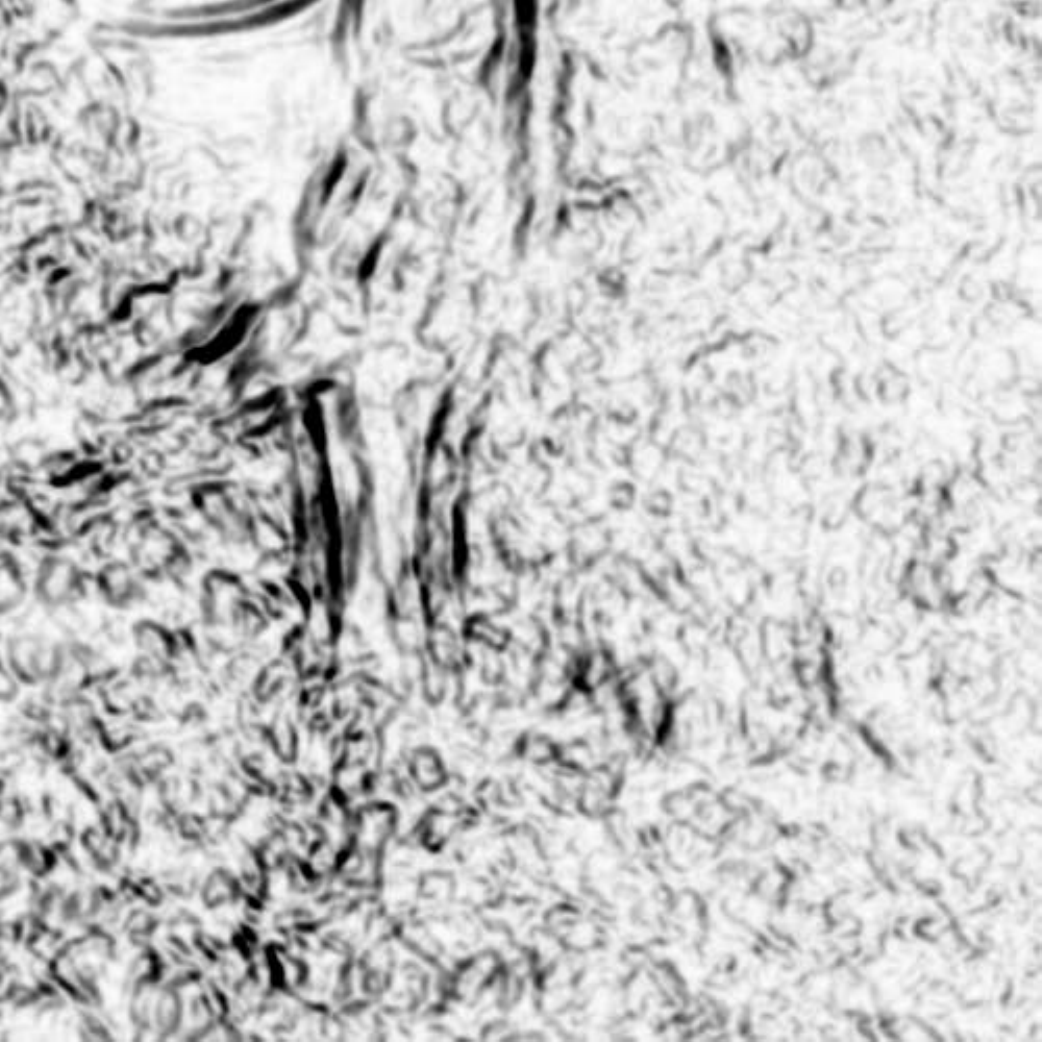} &
\includegraphics[height=\myfig]{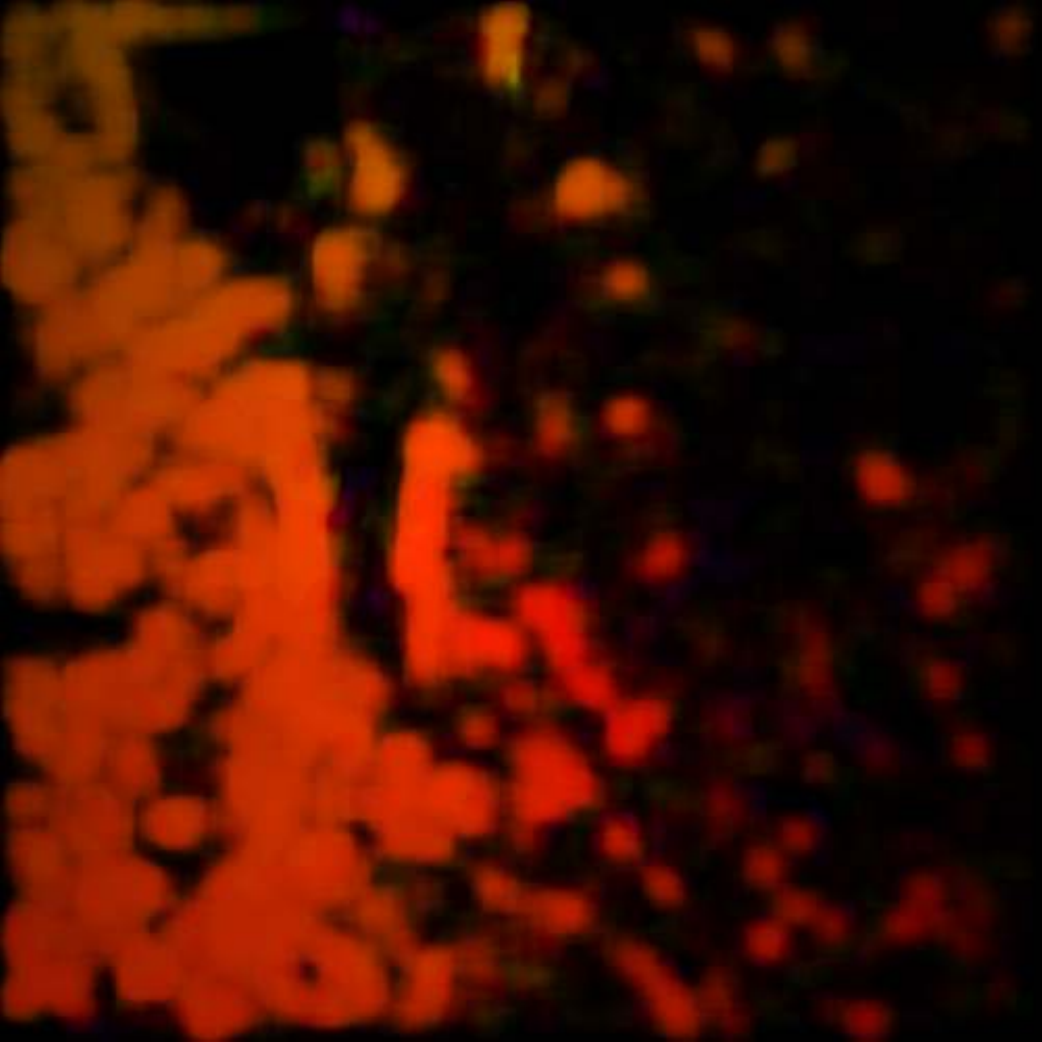}
\\

\rot{\myfig}{\elastix} &
\scalebarme{%
\includegraphics[height=\myfig]{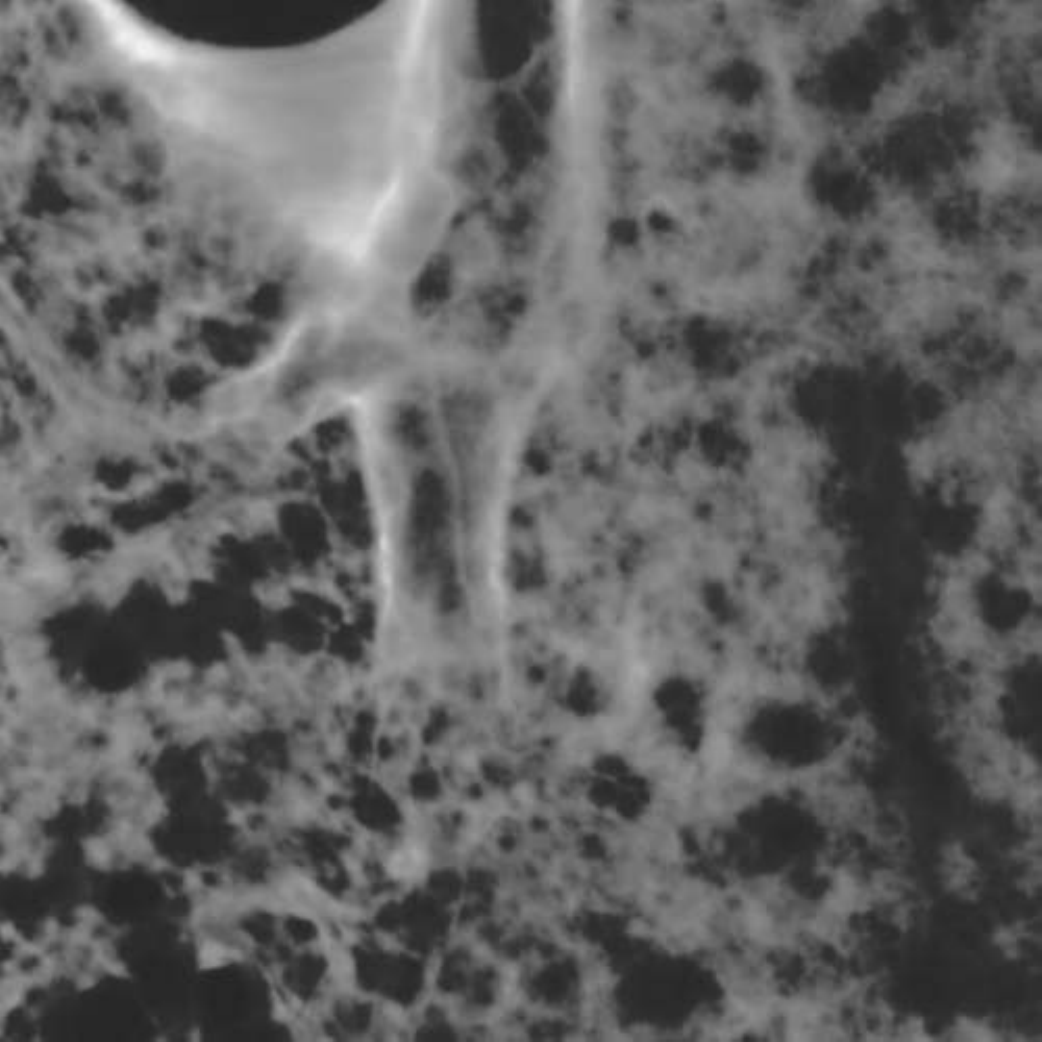}}{%
\includegraphics[width=\myfig]{image/scalebars/scale_ls-500.png}} &
\includegraphics[height=\myfig]{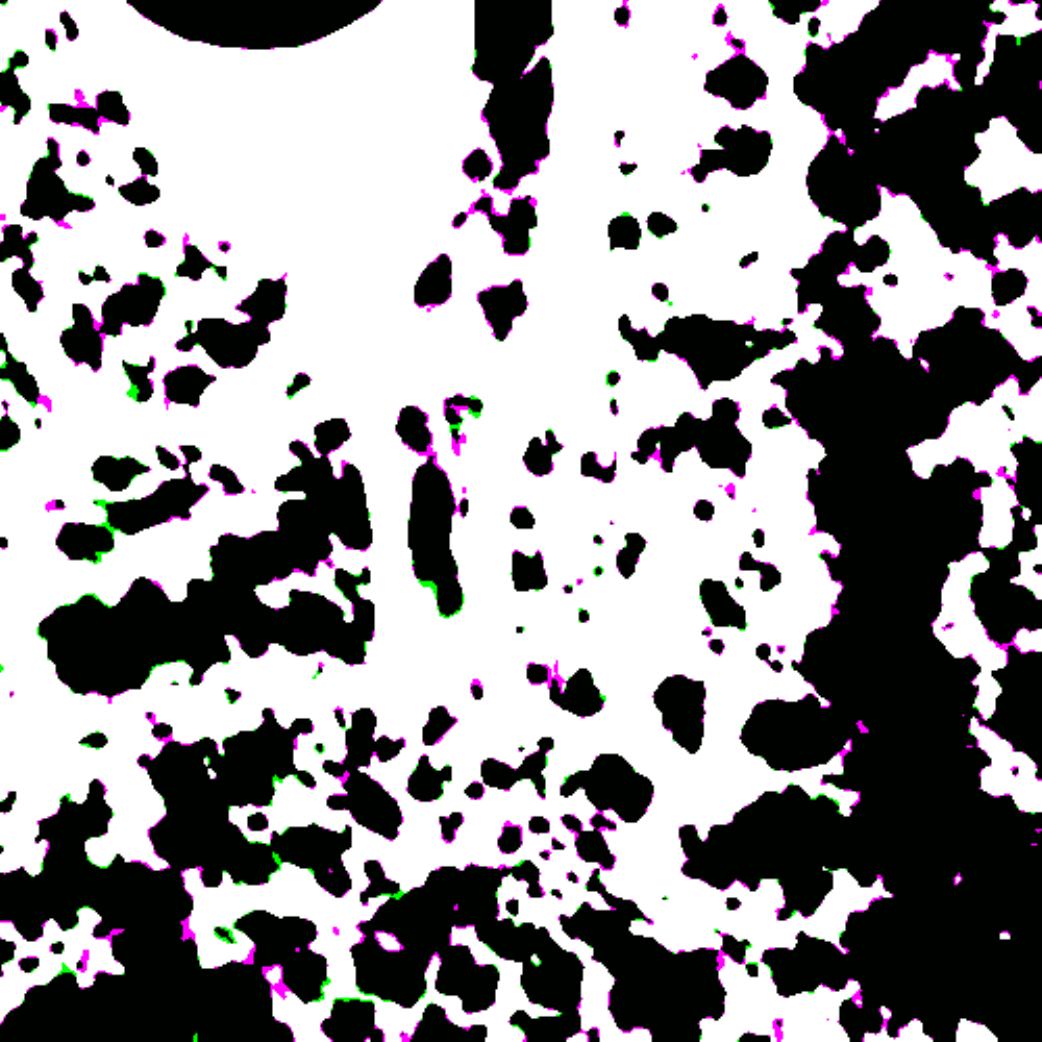} &
\includegraphics[height=\myfig]{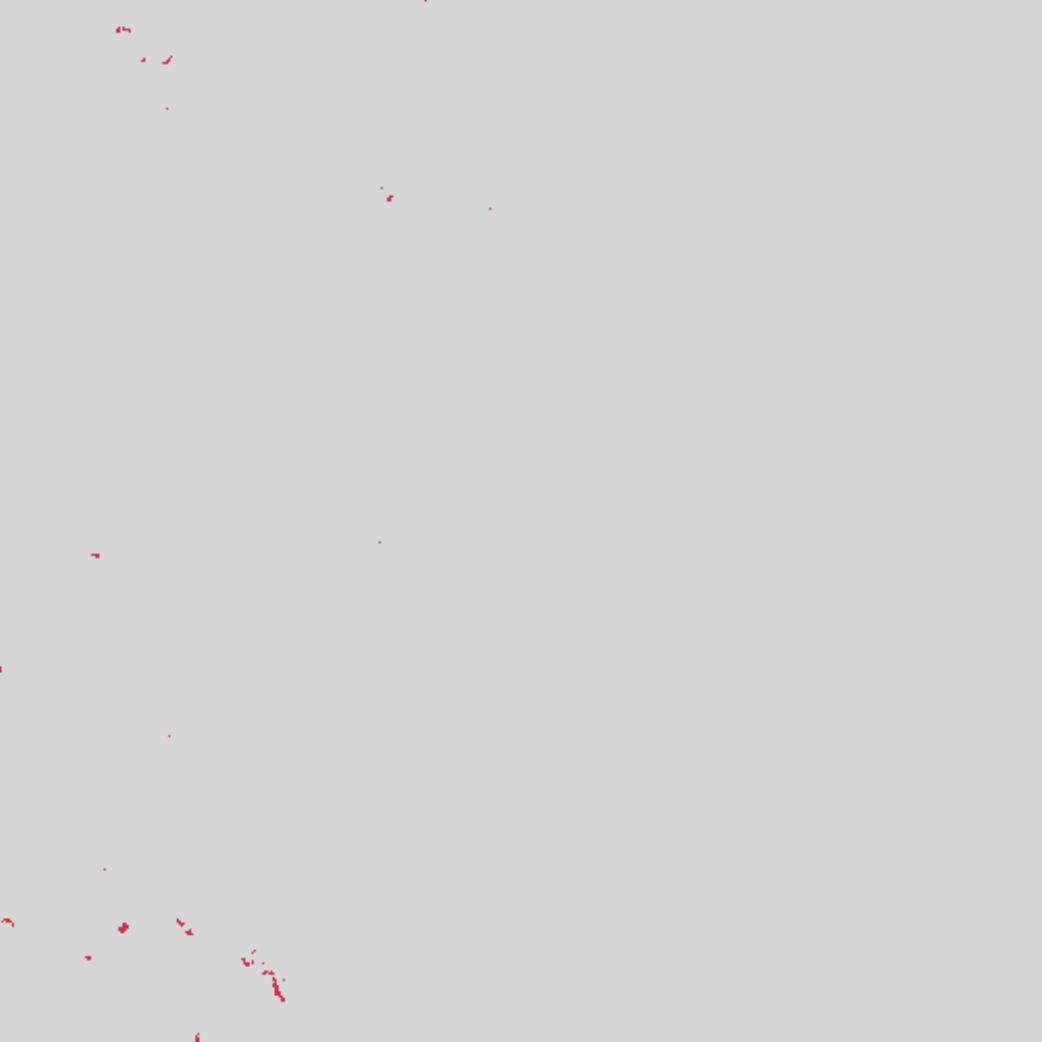} &
\includegraphics[height=\myfig]{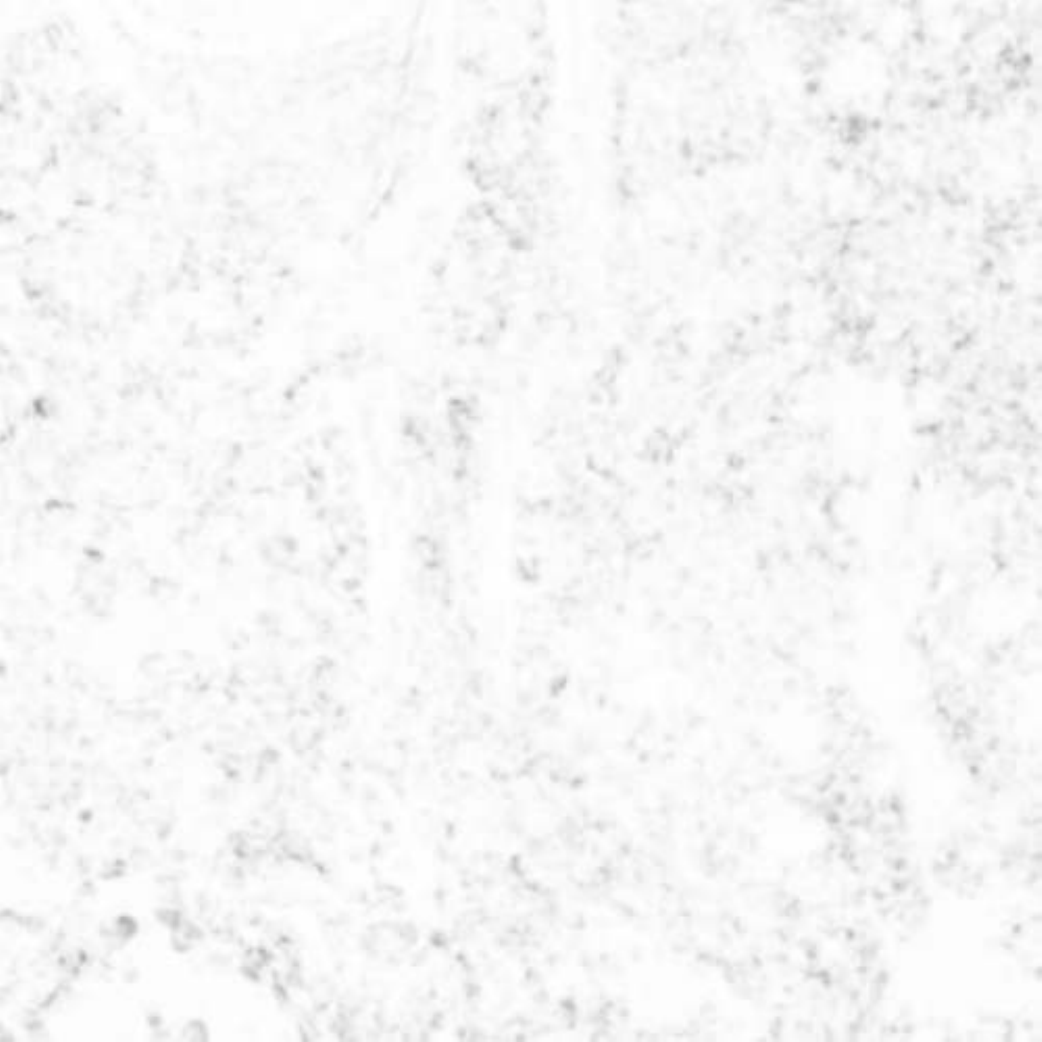} &
\includegraphics[height=\myfig]{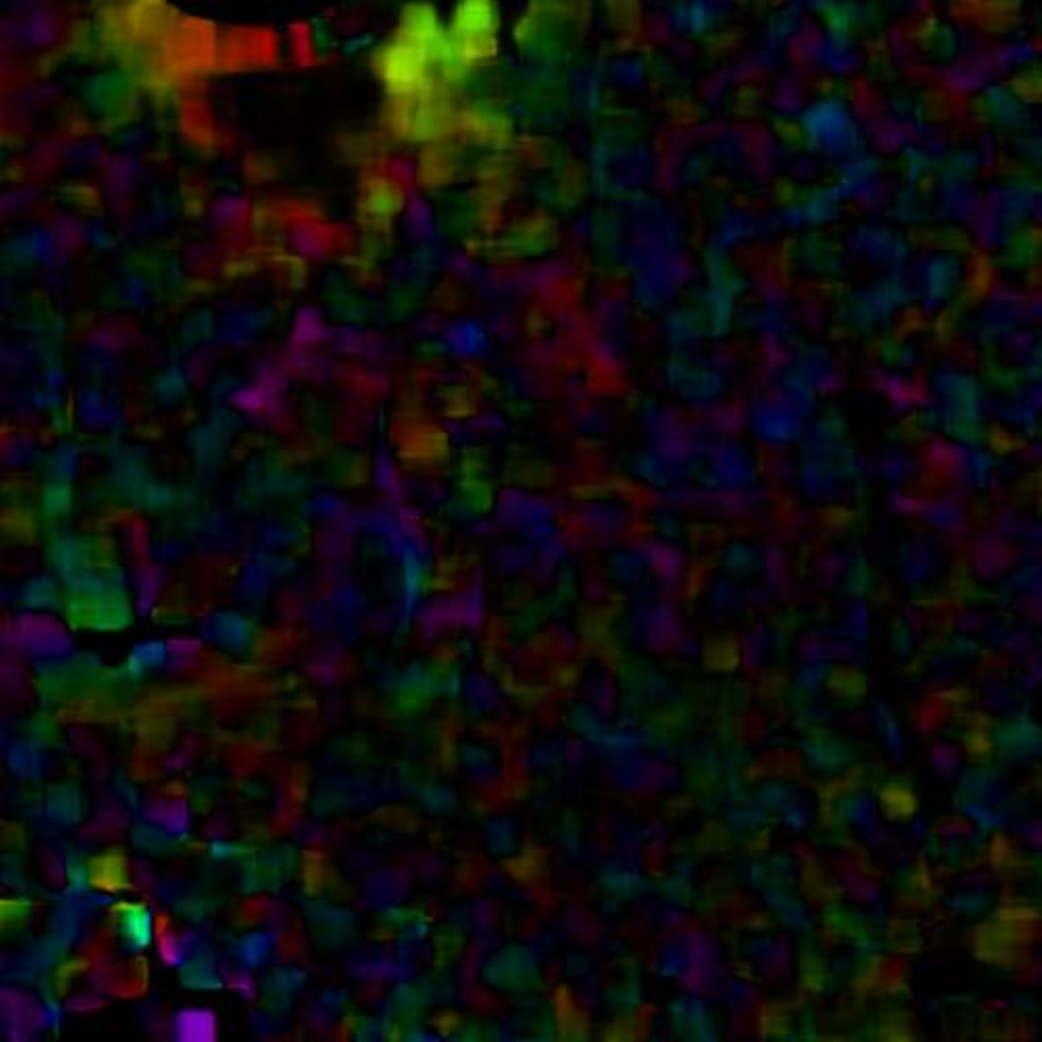}
\\

\rot{\myfig}{Local dist.} &
\scalebarme{%
\includegraphics[height=\myfig]{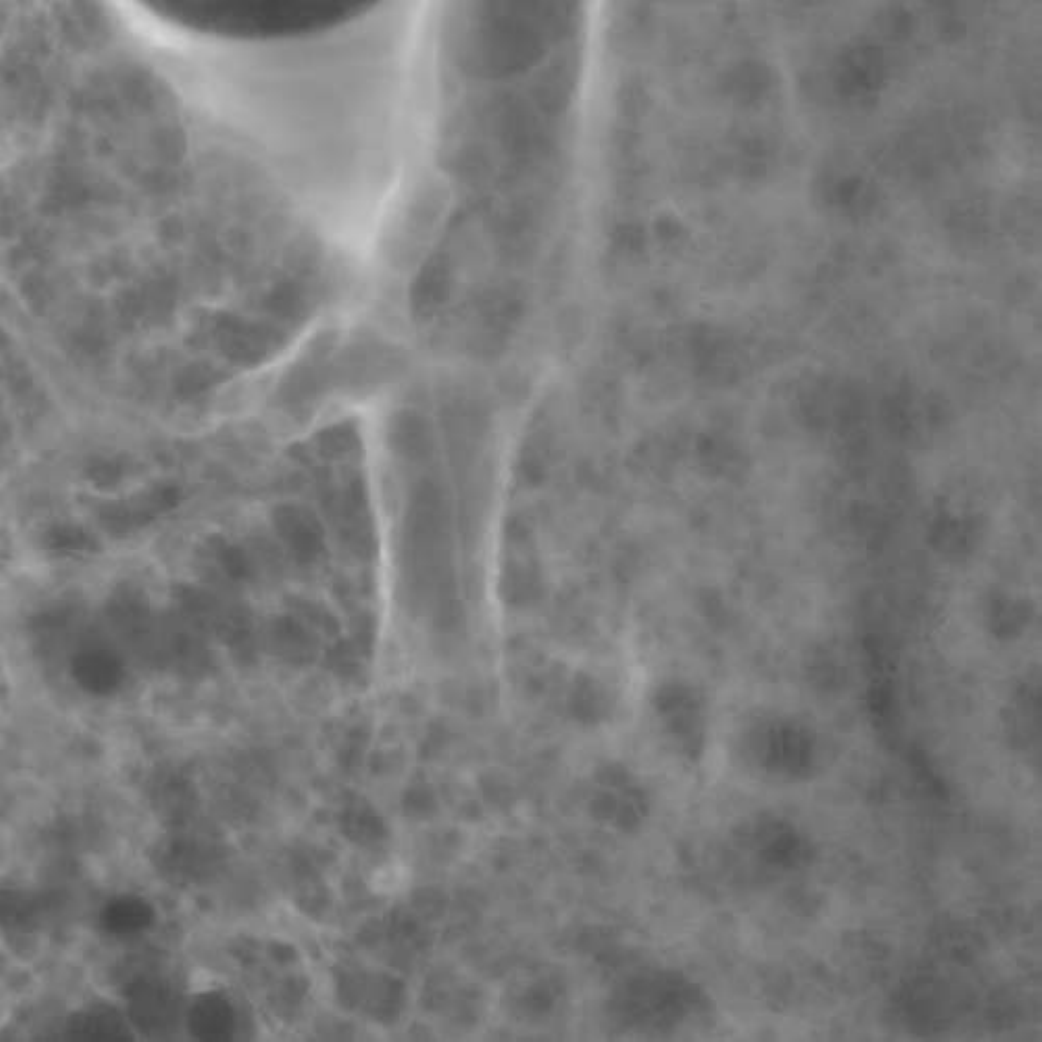}}{%
\includegraphics[width=\myfig]{image/scalebars/scale_ls-500.png}} &
\includegraphics[height=\myfig]{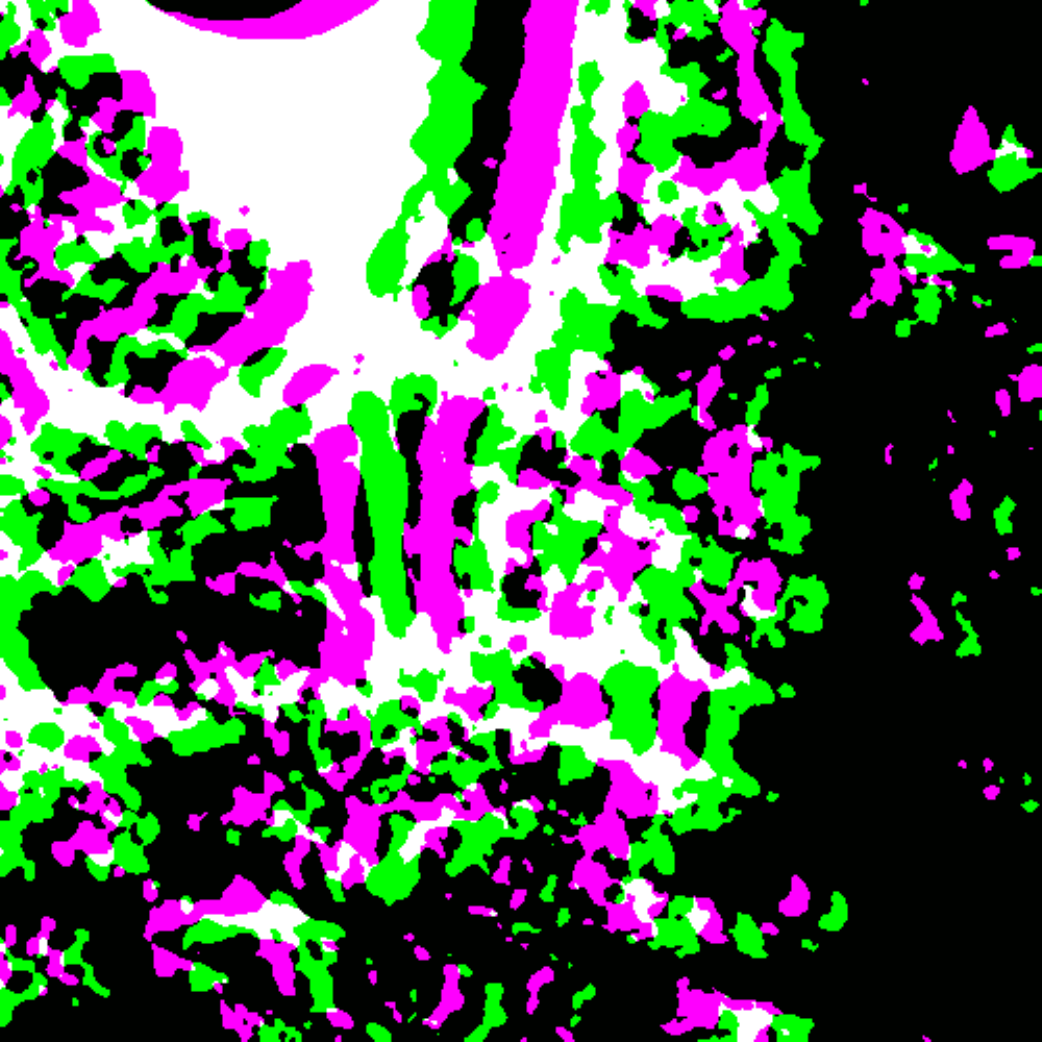} &
\includegraphics[height=\myfig]{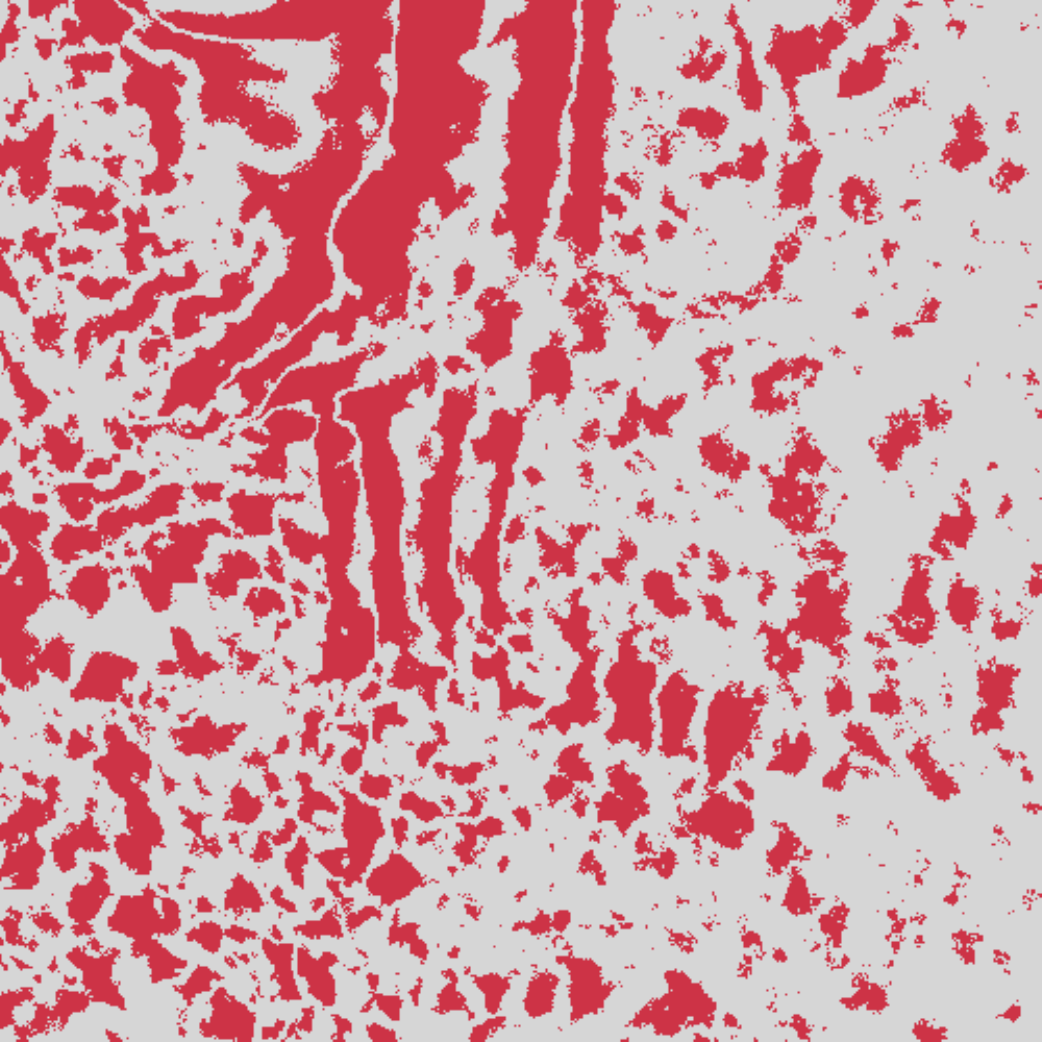} &
\includegraphics[height=\myfig]{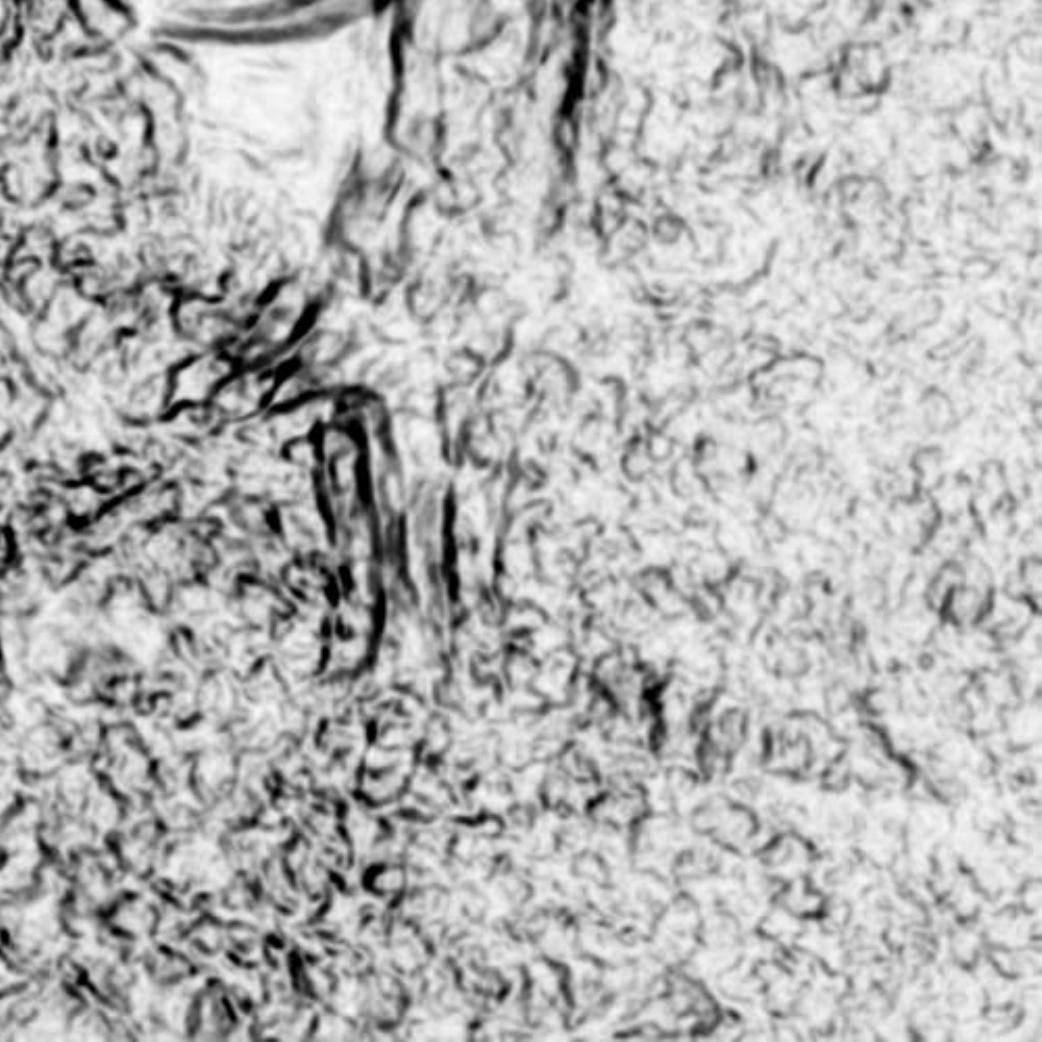} &
\includegraphics[height=\myfig]{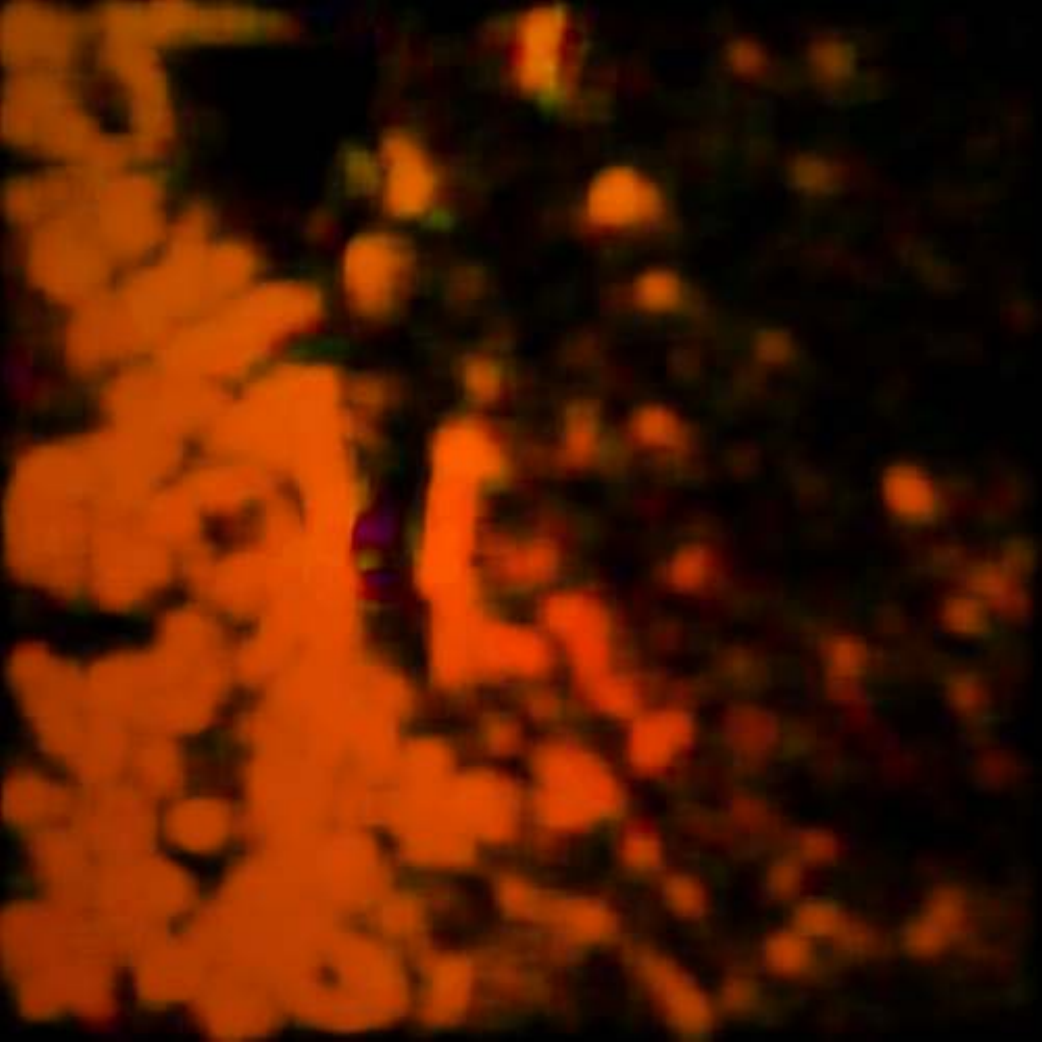}
\\

\rot{\myfig}{Gr.~truth, norm.} &
\scalebarme{%
\includegraphics[height=\myfig]{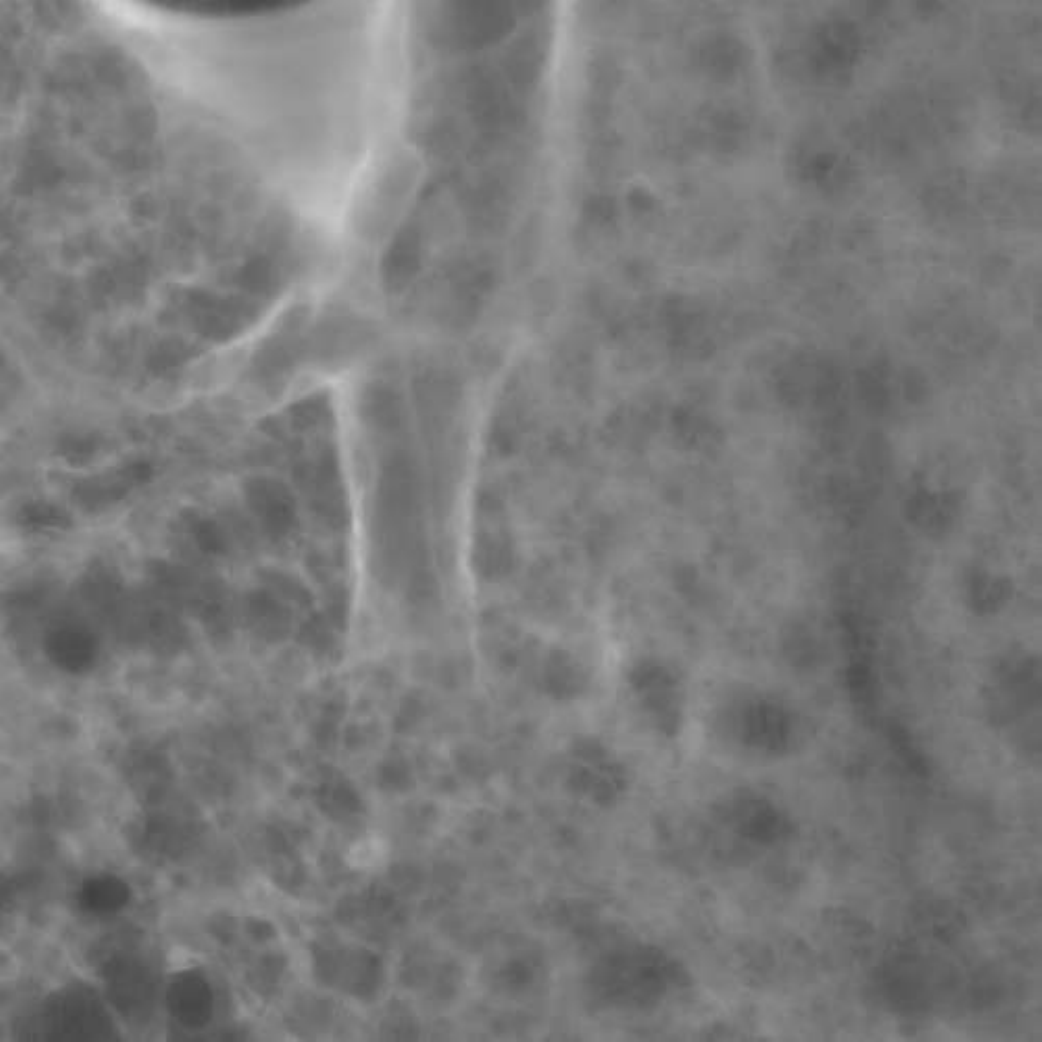}}{%
\includegraphics[width=\myfig]{image/scalebars/scale_ls-500.png}} &
\includegraphics[height=\myfig]{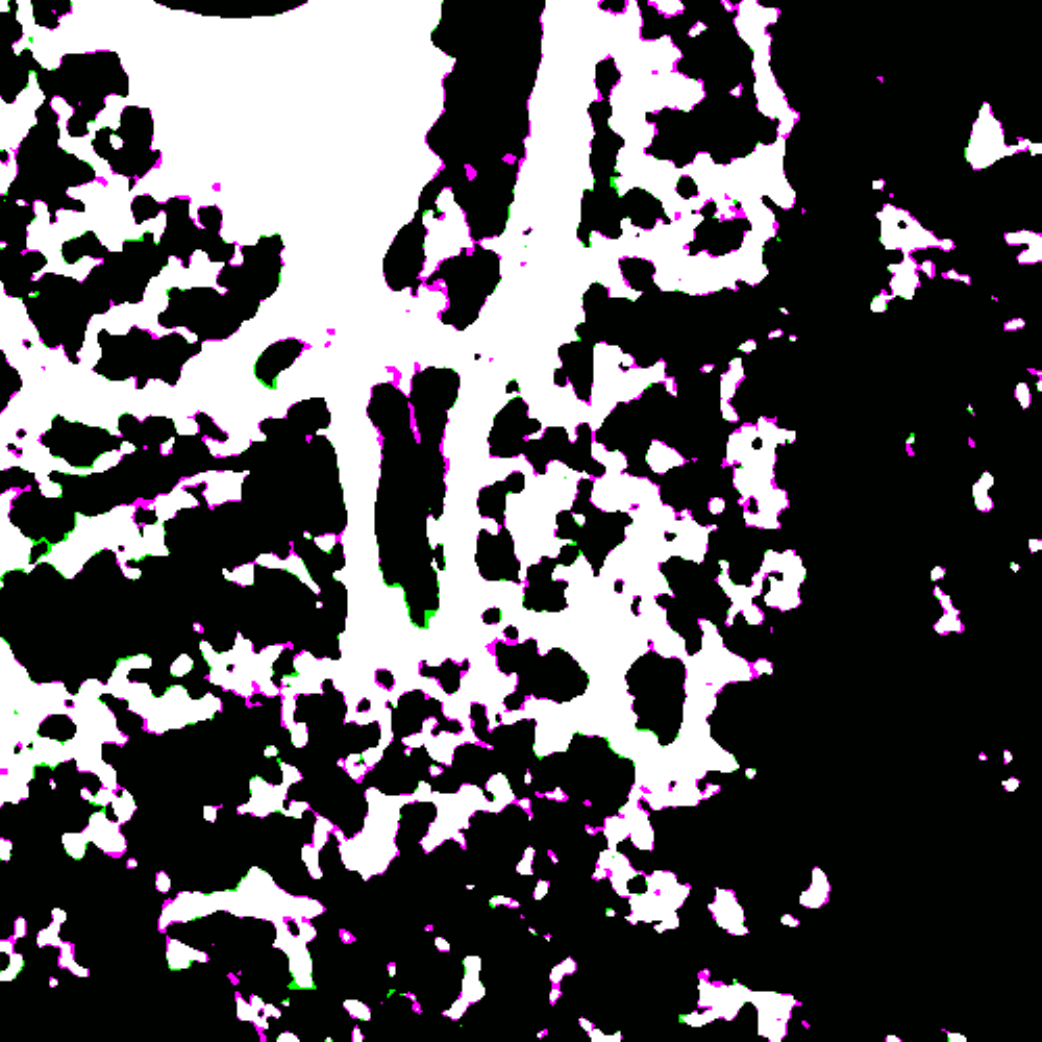} &
\includegraphics[height=\myfig]{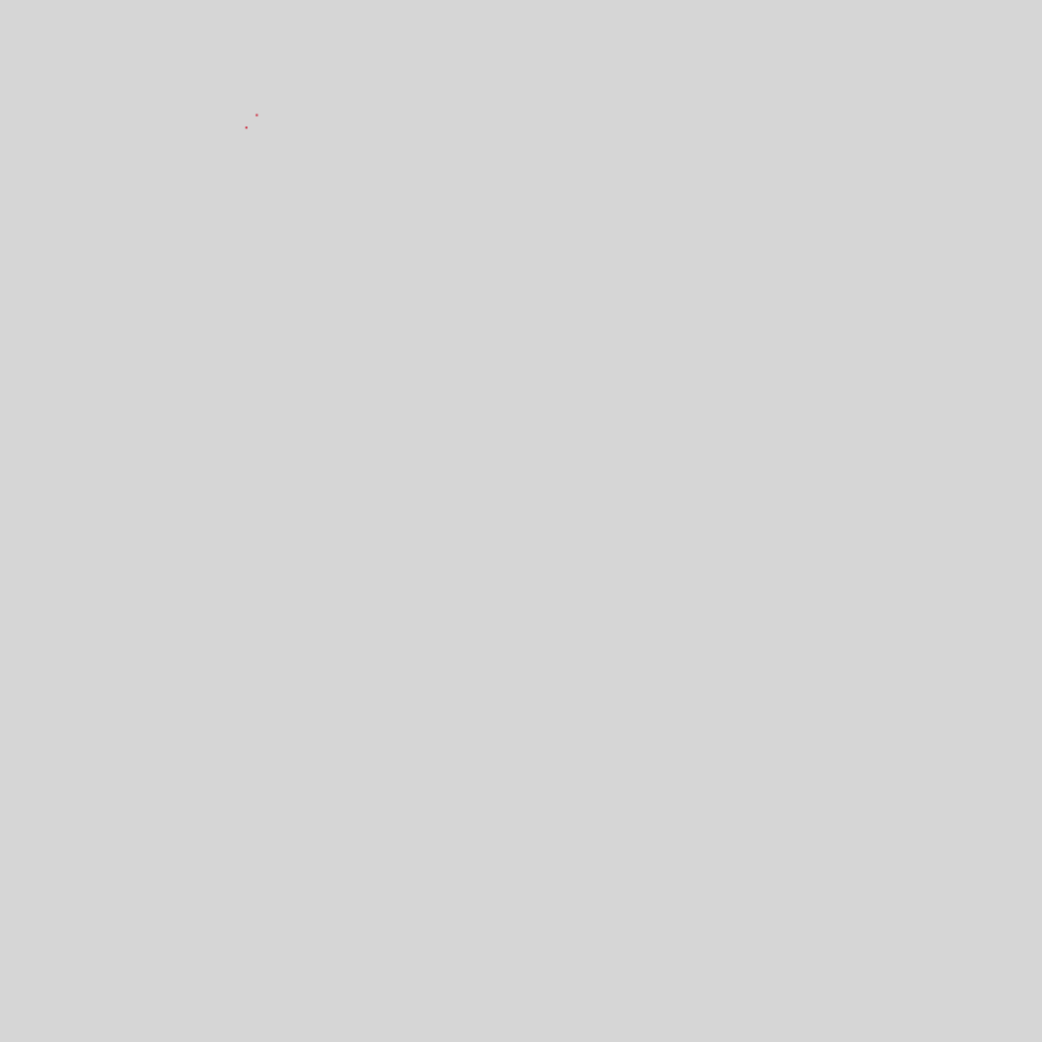} &
\includegraphics[height=\myfig]{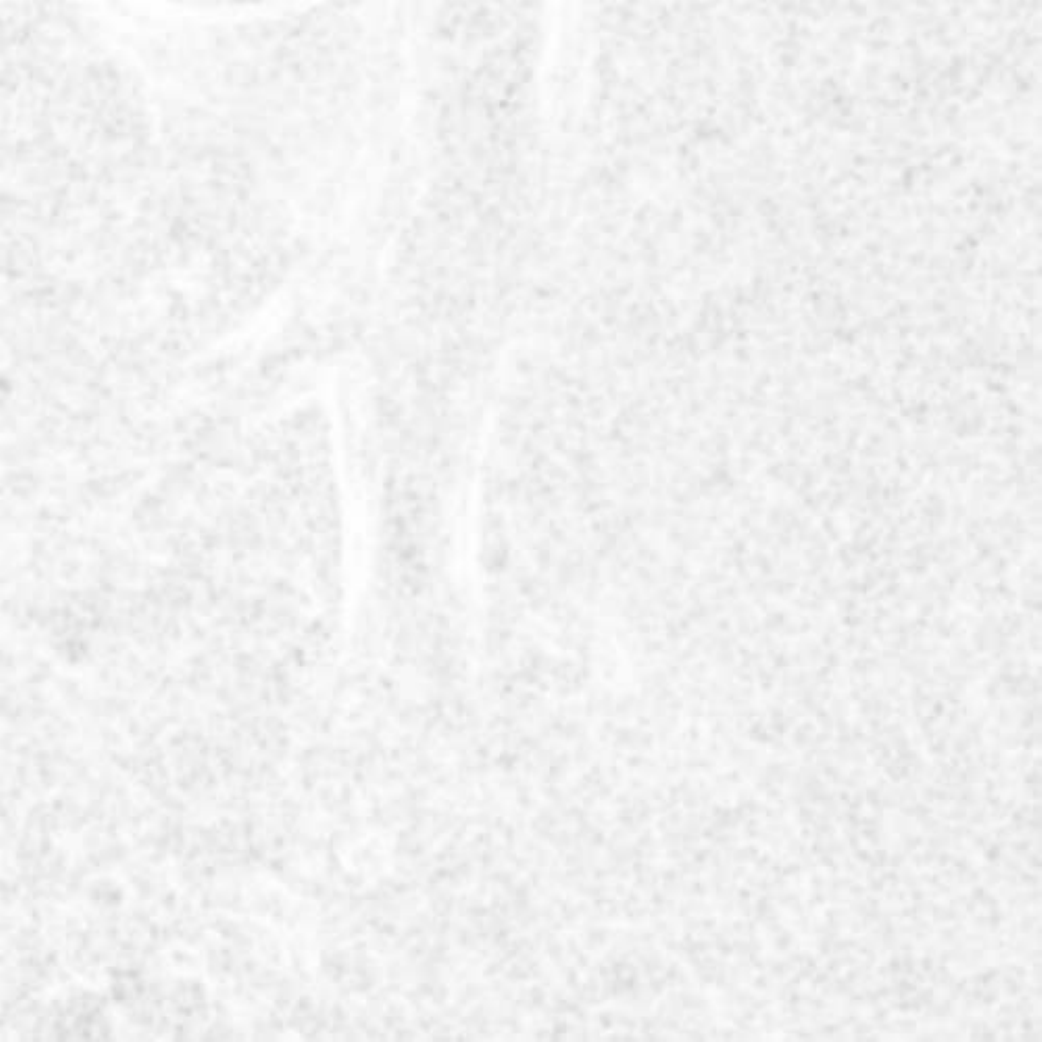} &
\includegraphics[height=\myfig]{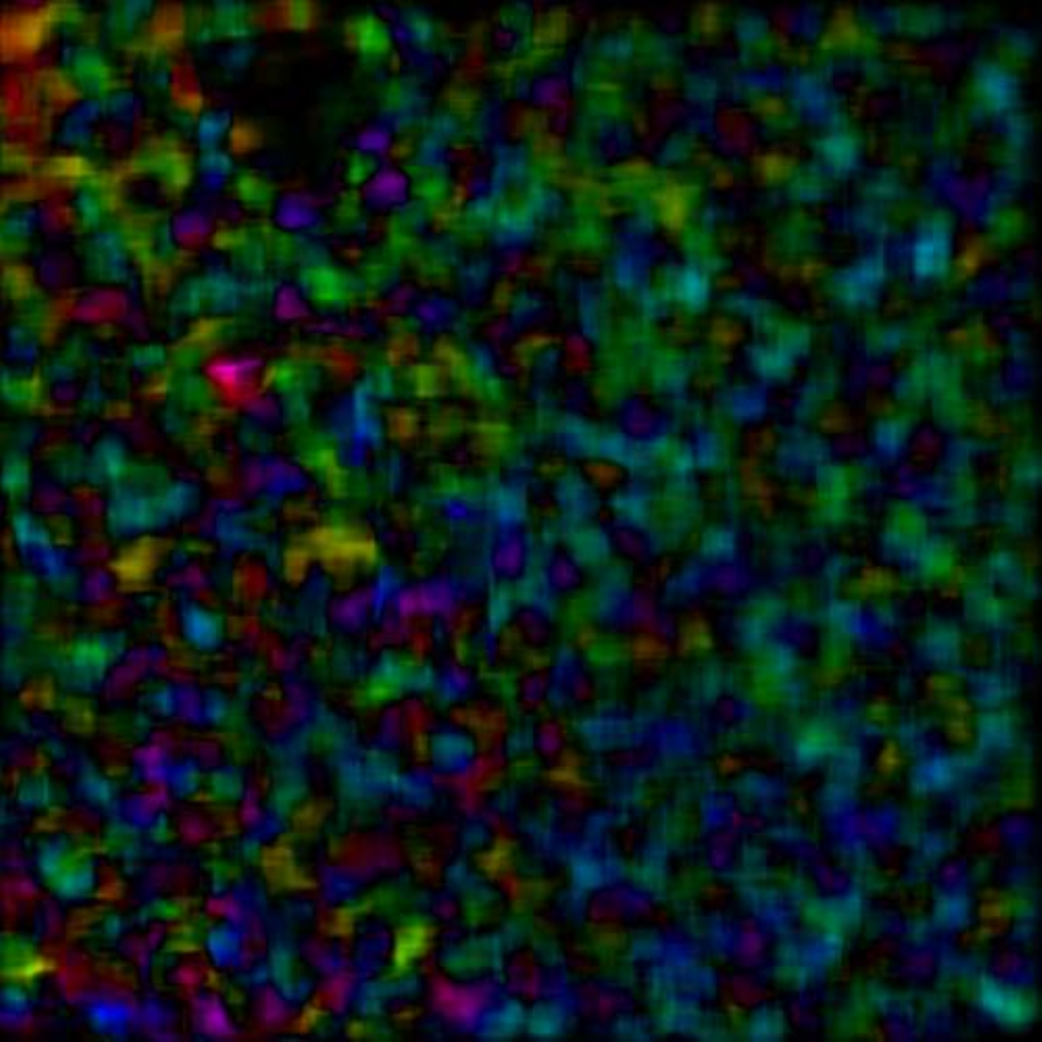}
\\

& \multicolumn{1}{c}{Image} & \multicolumn{1}{c}{Dice} & \multicolumn{1}{c}{PSNR} & \multicolumn{1}{c}{SSIM} & \multicolumn{1}{c}{Flow} \\

\end{tabular}
\caption{Evaluation of LS data set with image measures, part~1. The ground truth is normalized. Continued in Fig.~\ref{fig:eval-ls-2}. All scale bars are \SI{500}{\micro\meter}.}
\label{fig:eval-ls-1}
\setlength{\tabcolsep}{\tabcolbackup}
\thisfloatpagestyle{empty} 
\end{figure*}
\begin{figure*}[!t]
\centering
\setlength{\myfig}{0.2\linewidth-0.001\linewidth-2pt-0.2em}
\setlength{\tabcolsep}{1pt}
\begin{tabular}{m{1em} m{\myfig} m{\myfig} m{\myfig} m{\myfig} m{\myfig}}
& \multicolumn{1}{c}{Image} & \multicolumn{1}{c}{Dice} & \multicolumn{1}{c}{PSNR} & \multicolumn{1}{c}{SSIM} & \multicolumn{1}{c}{Flow} \\

\rot{\myfig}{Rigid-SURF} &
\scalebarme{%
\includegraphics[height=\myfig]{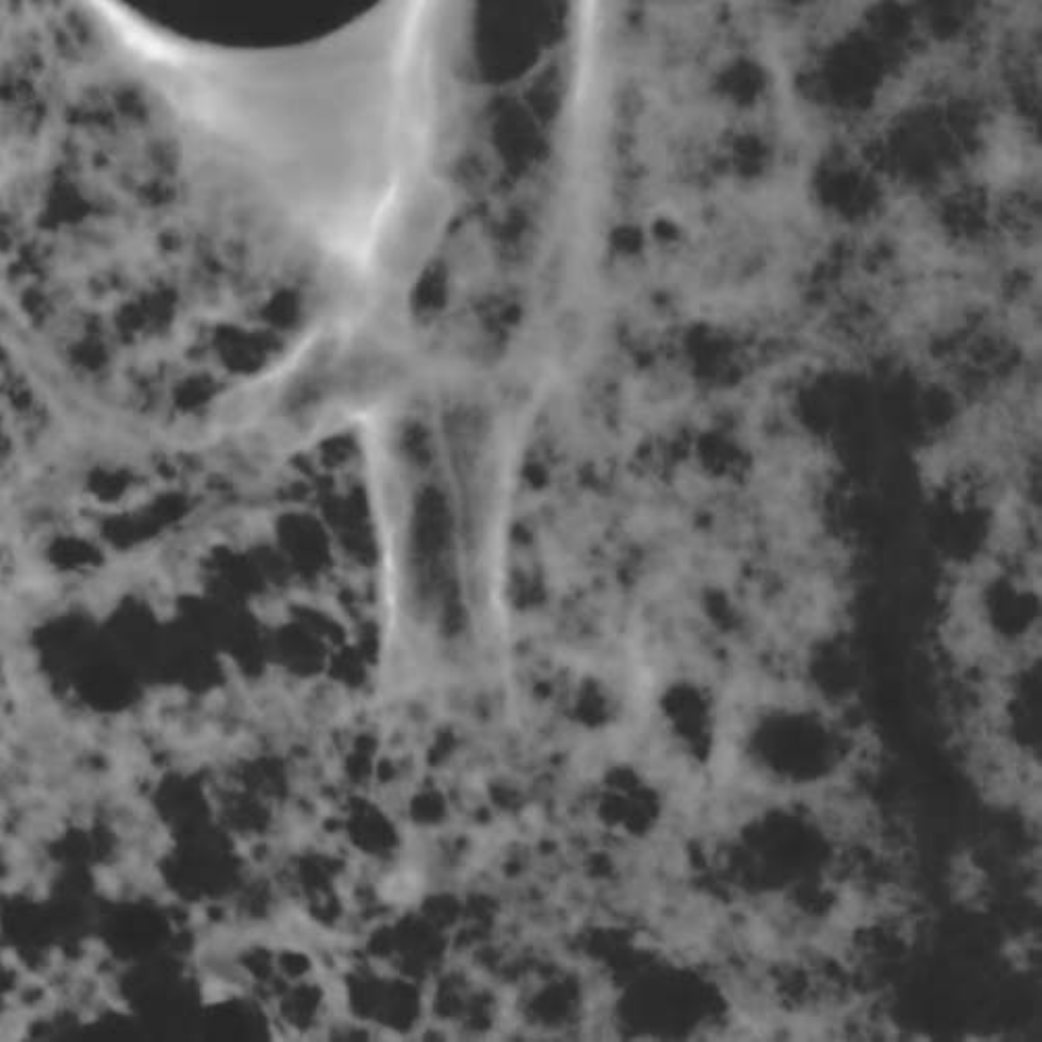}}{%
\includegraphics[width=\myfig]{image/scalebars/scale_ls-500.png}} &
\includegraphics[height=\myfig]{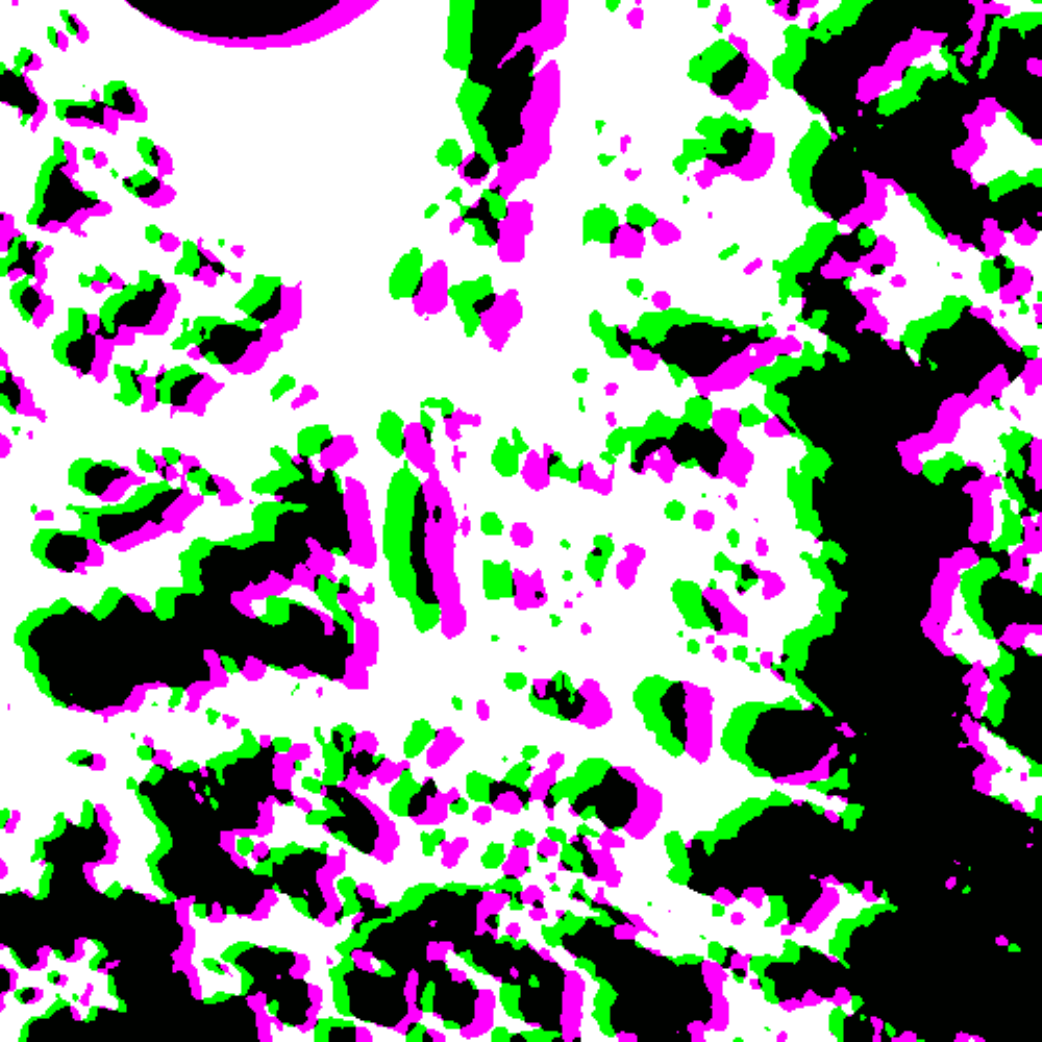} &
\includegraphics[height=\myfig]{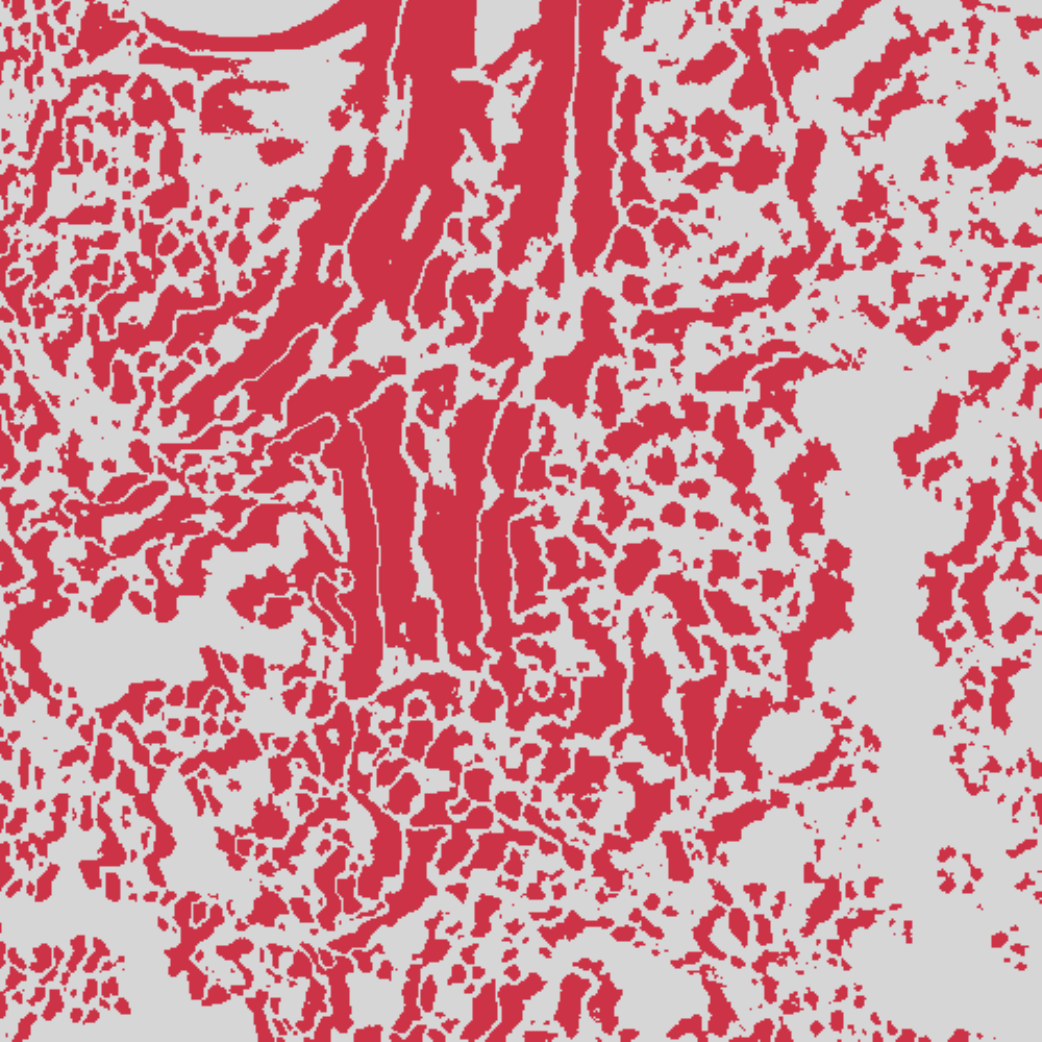} &
\includegraphics[height=\myfig]{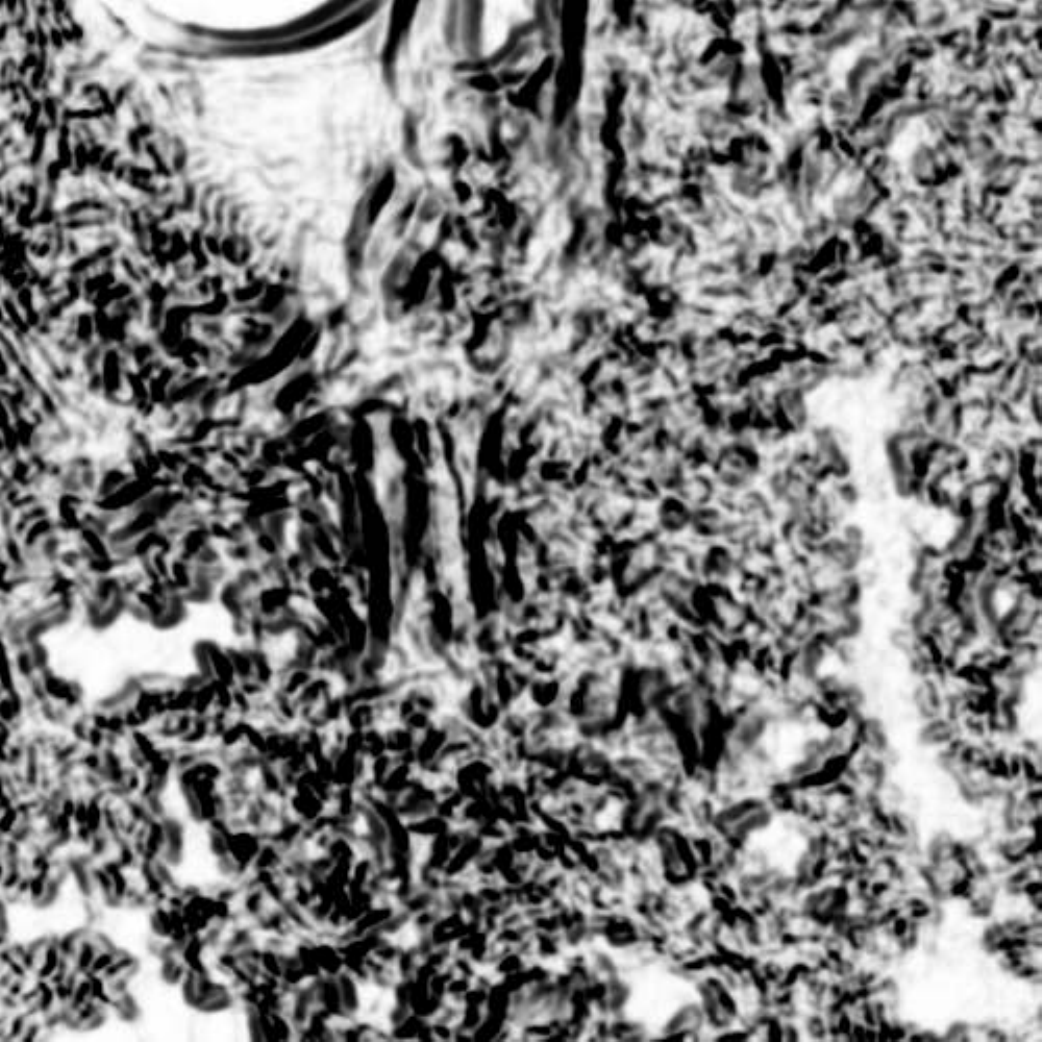} &
\includegraphics[height=\myfig]{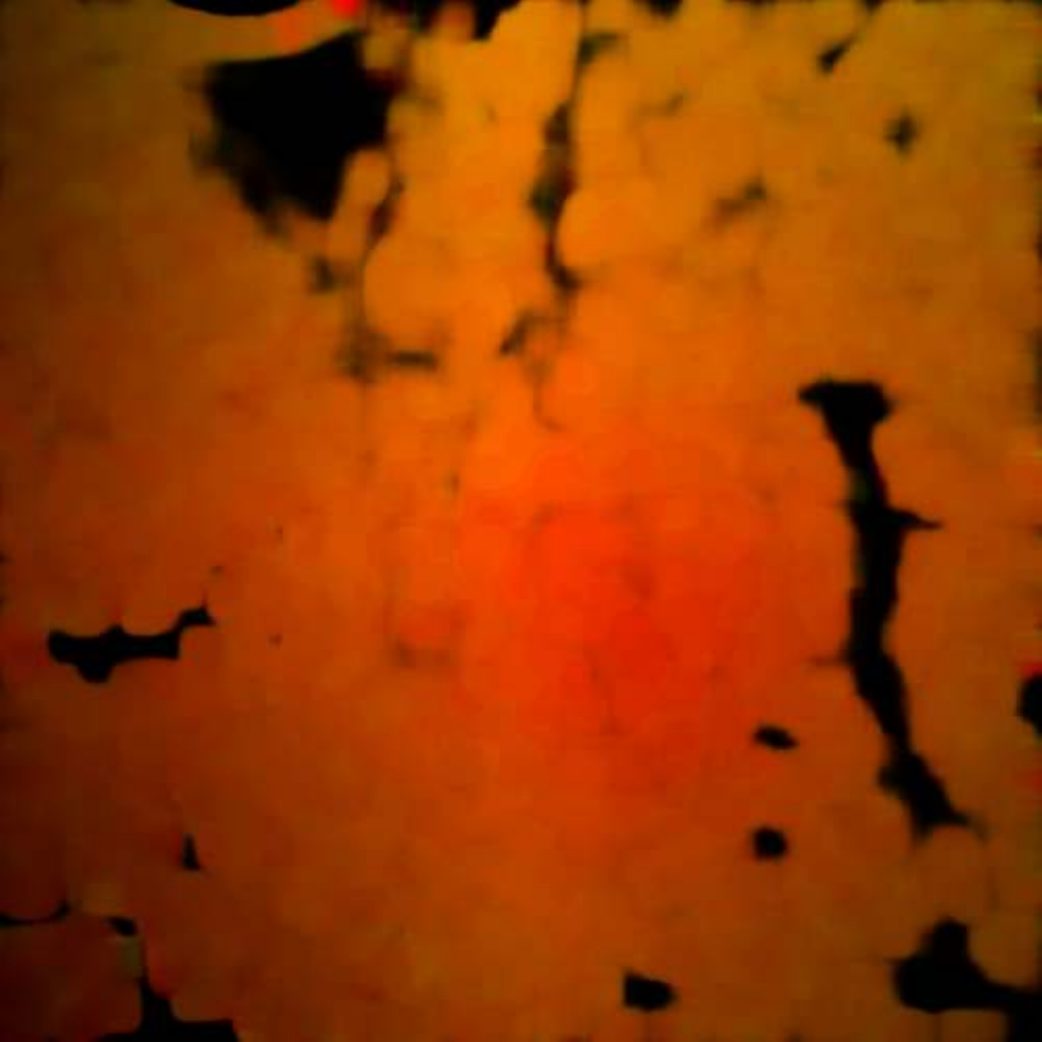}
\\

\rot{\myfig}{D.-SURF, std.} &
\scalebarme{%
\includegraphics[height=\myfig]{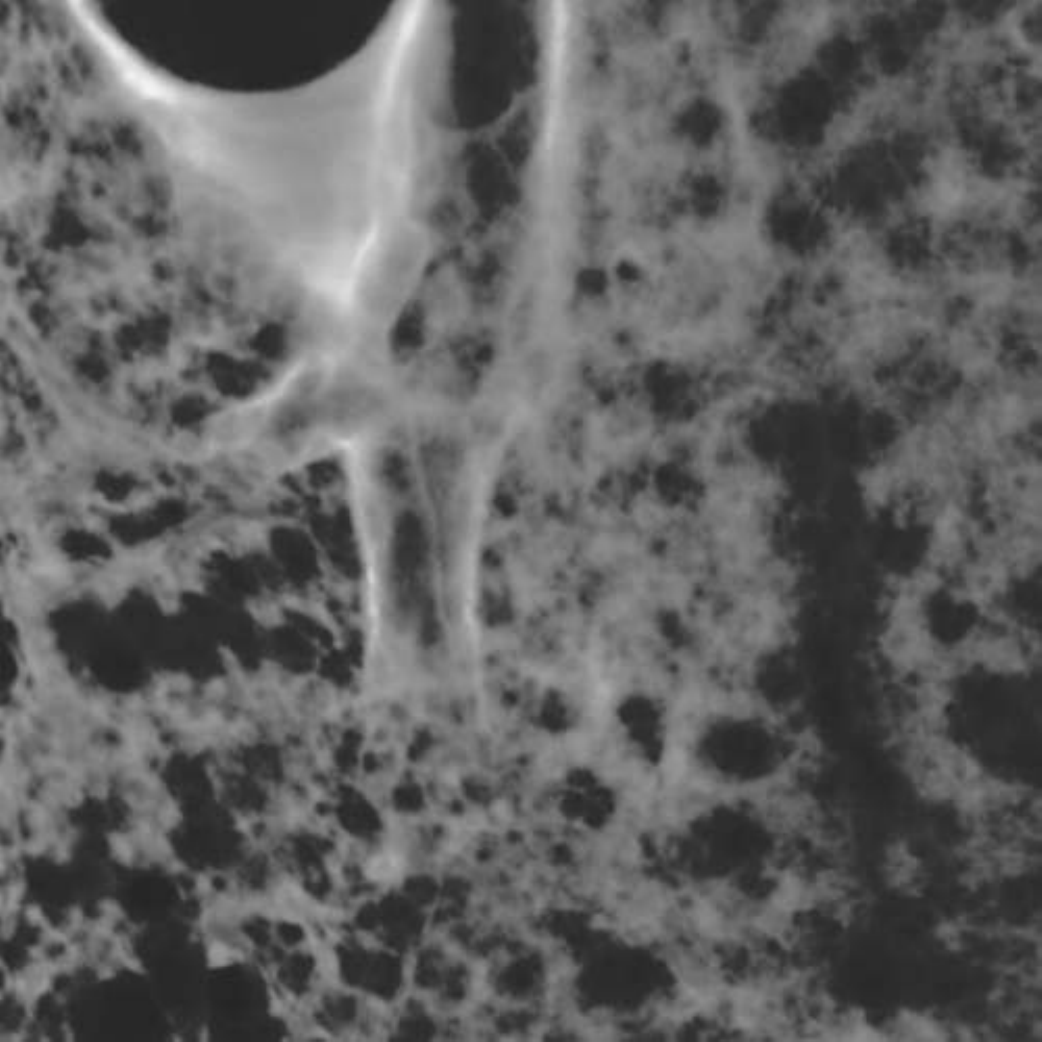}}{%
\includegraphics[width=\myfig]{image/scalebars/scale_ls-500.png}} &
\includegraphics[height=\myfig]{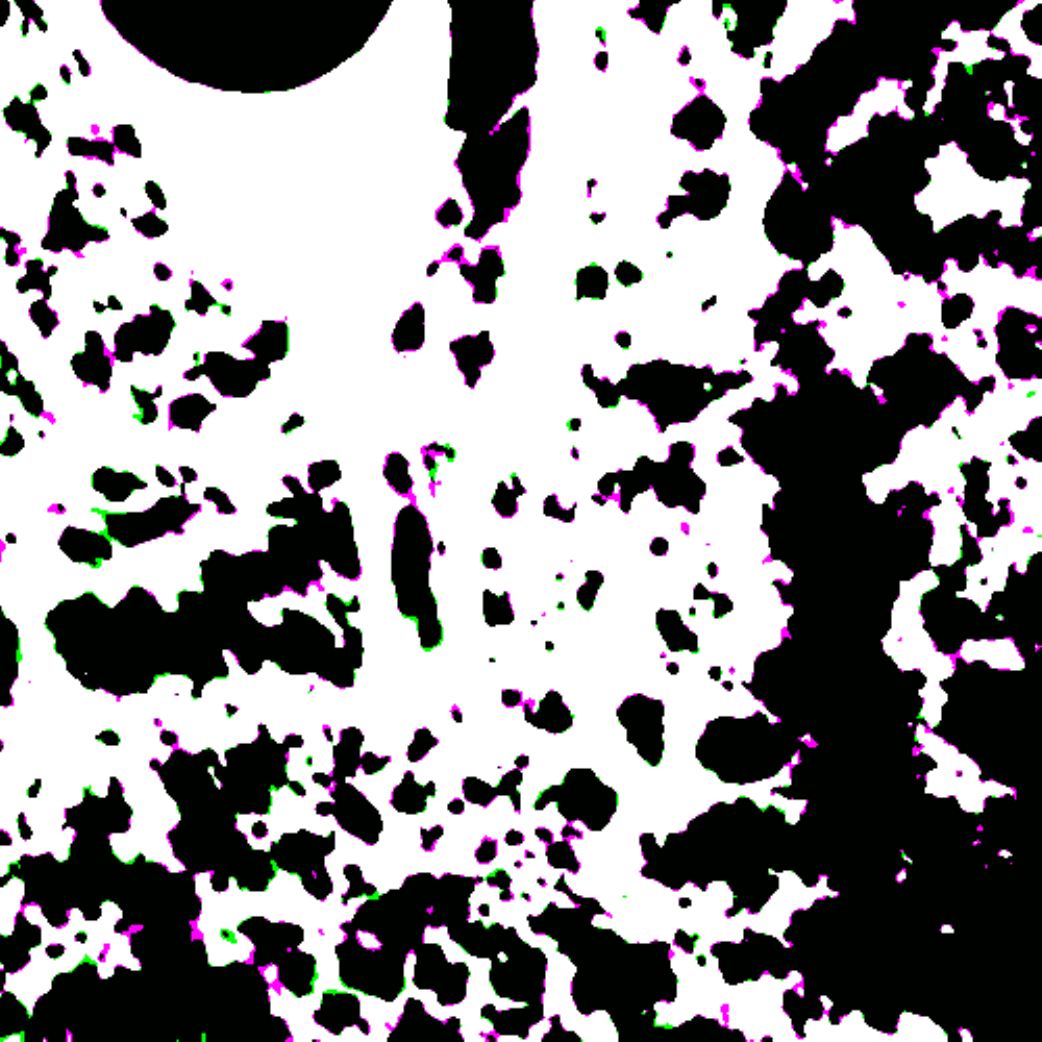} &
\includegraphics[height=\myfig]{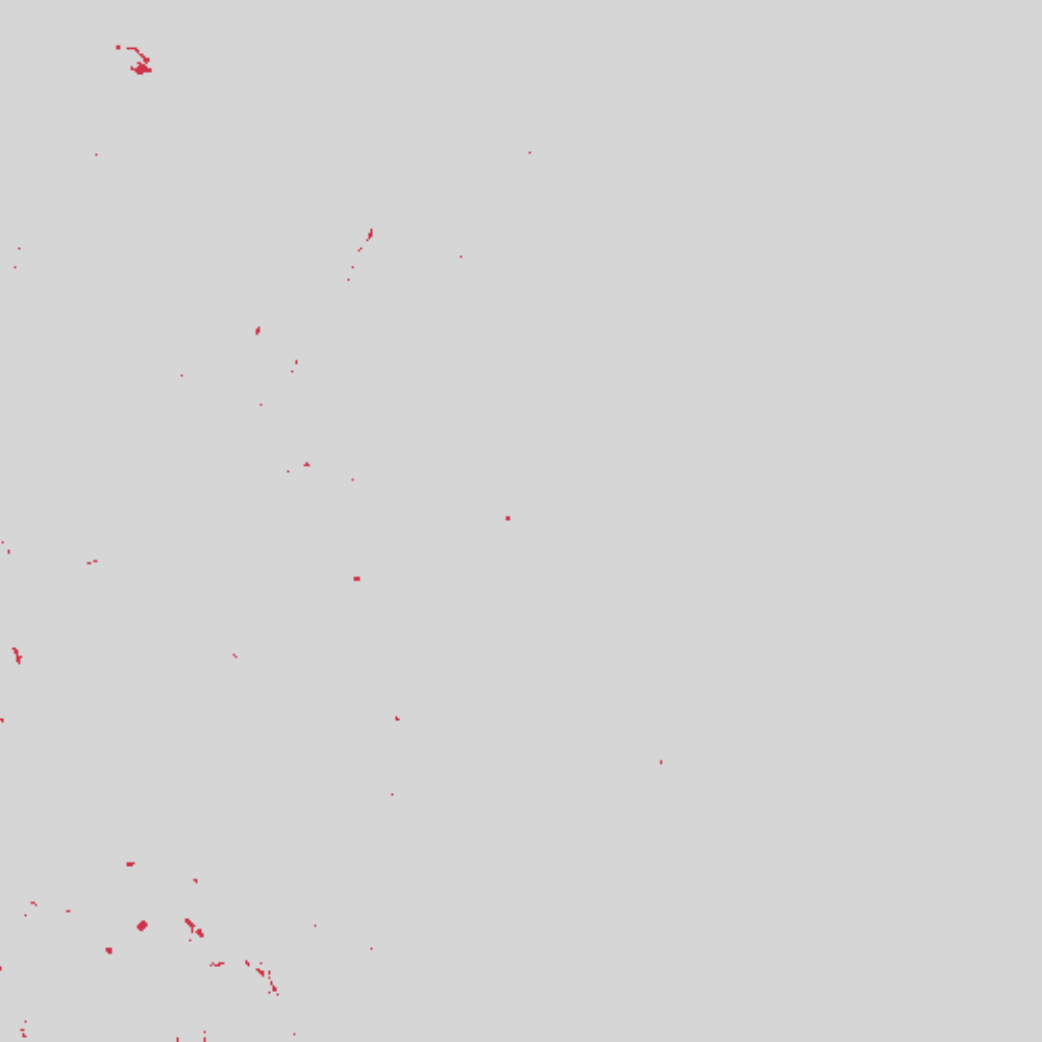} &
\includegraphics[height=\myfig]{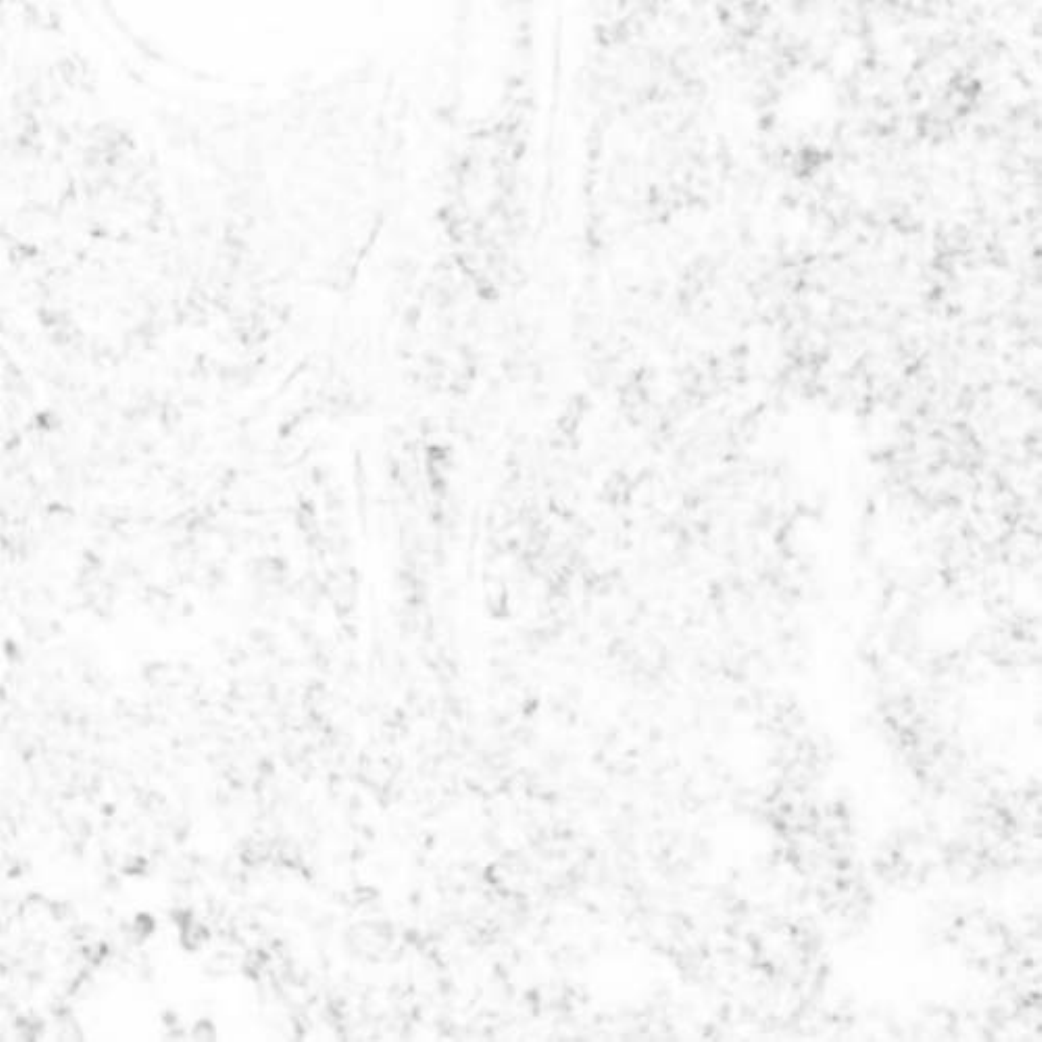} &
\includegraphics[height=\myfig]{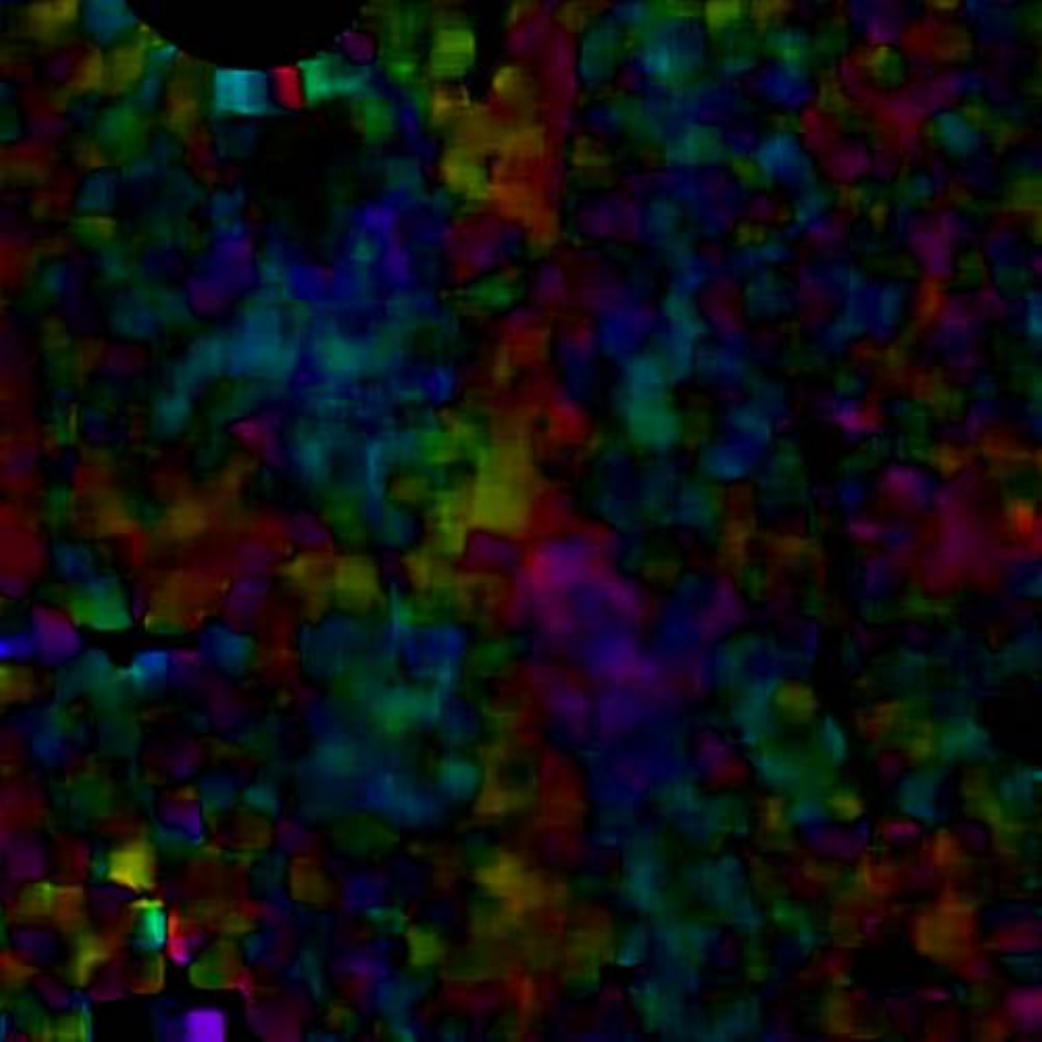}
\\

\rot{\myfig}{D.-SURF, \num{e-3}} &
\scalebarme{%
\includegraphics[height=\myfig]{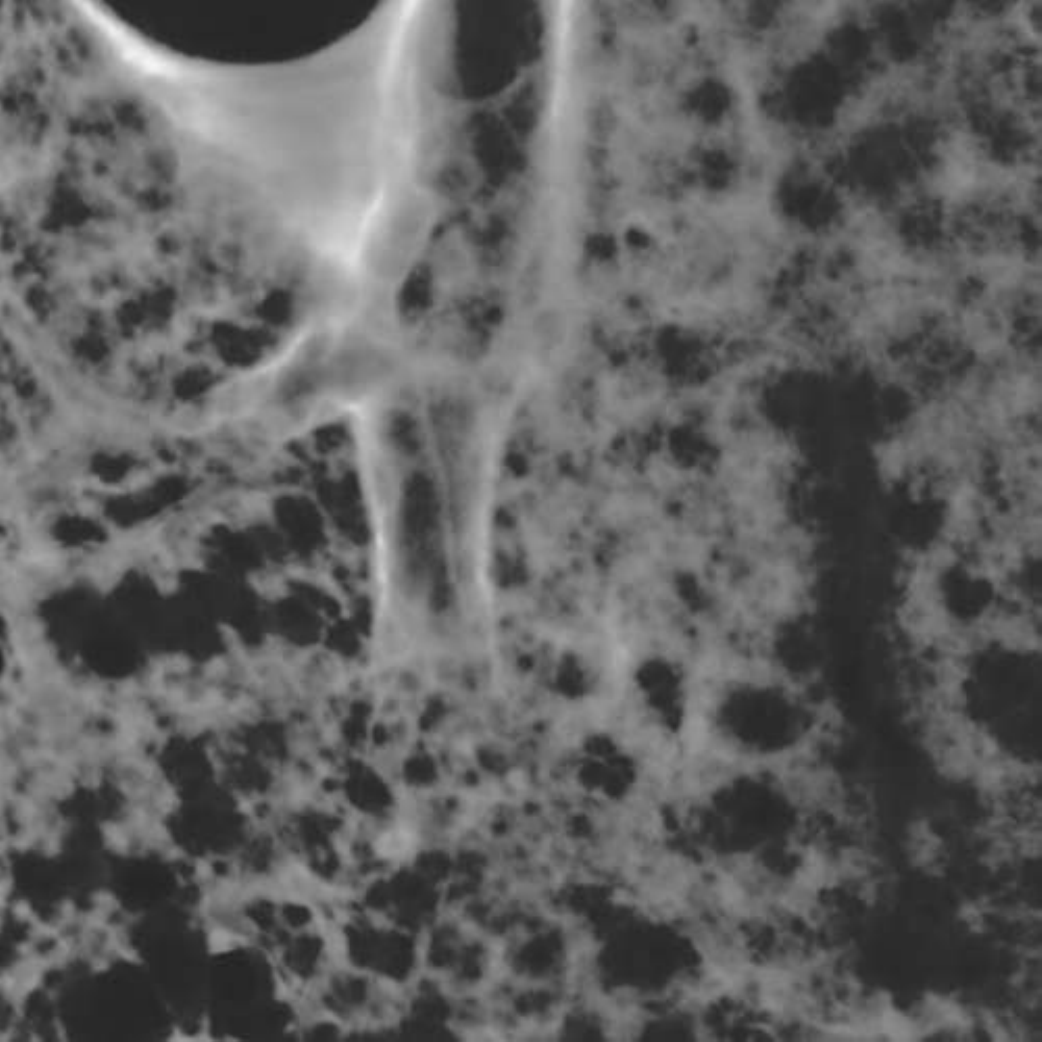}}{%
\includegraphics[width=\myfig]{image/scalebars/scale_ls-500.png}} &
\includegraphics[height=\myfig]{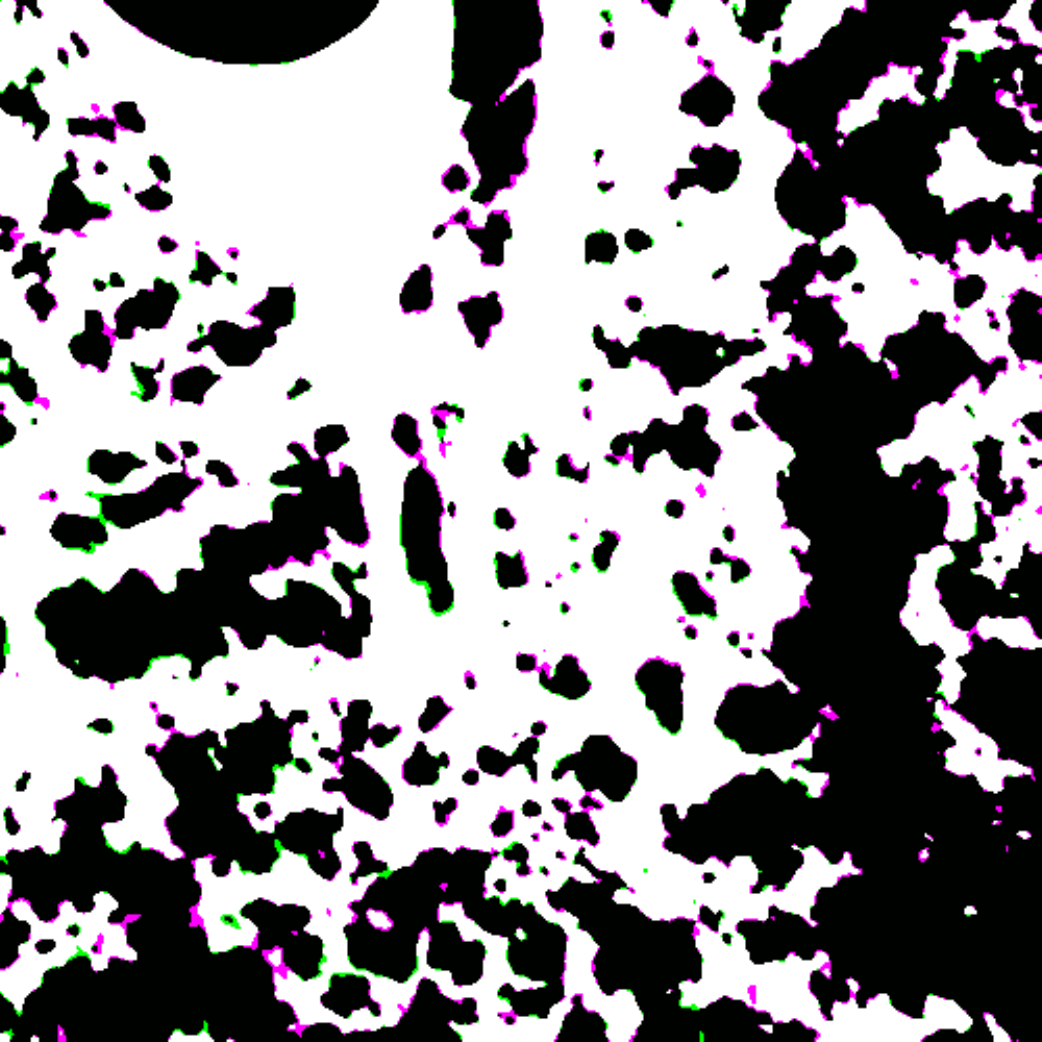} &
\includegraphics[height=\myfig]{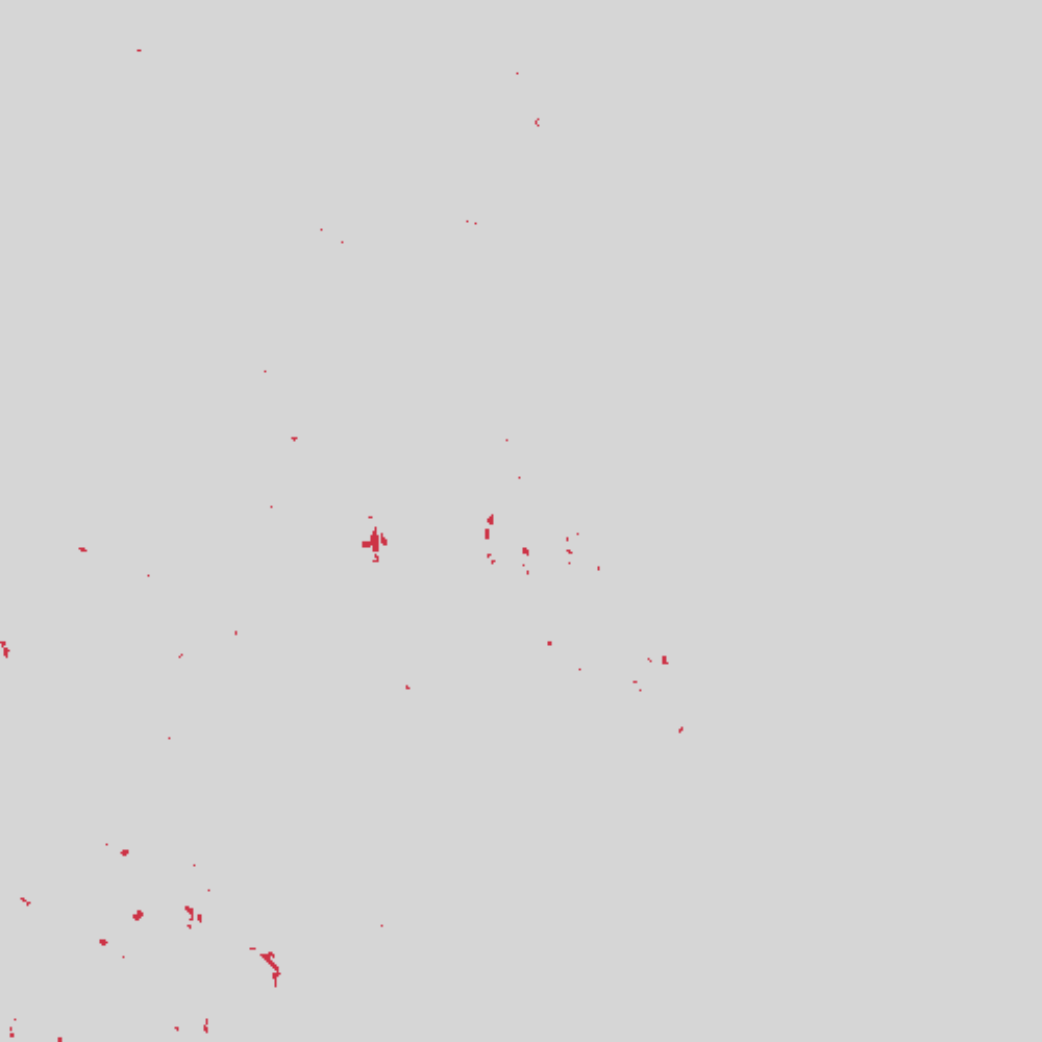} &
\includegraphics[height=\myfig]{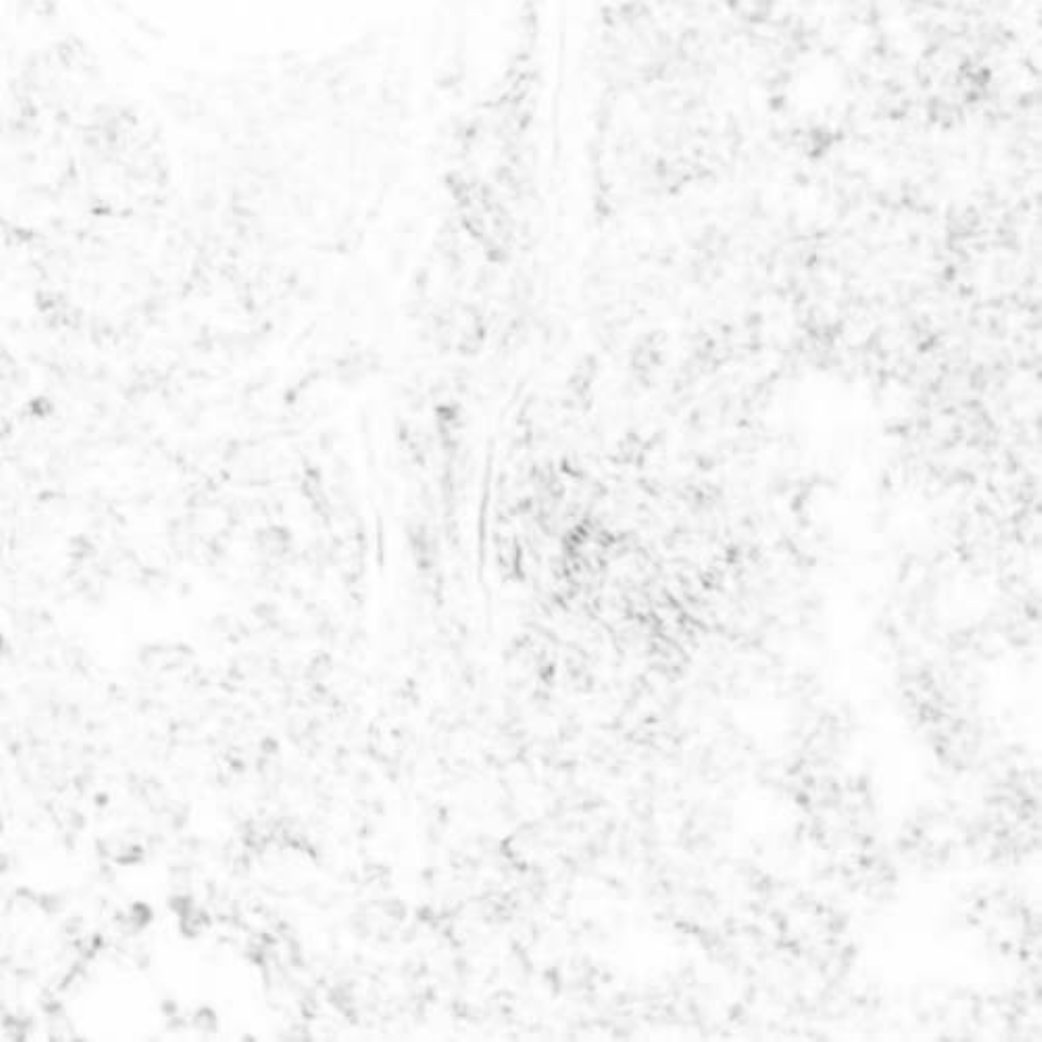} &
\includegraphics[height=\myfig]{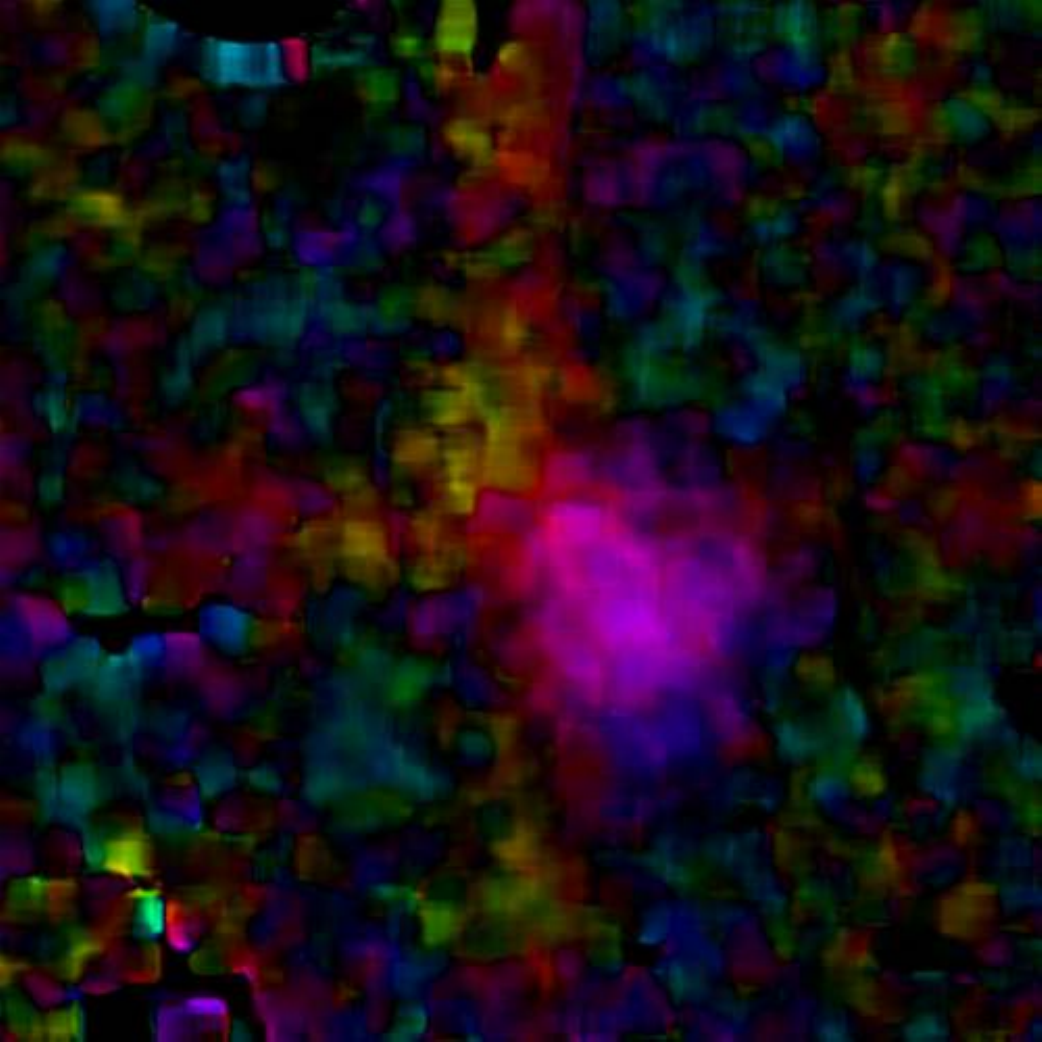}
\\

\rot{\myfig}{D.-SURF, \num{e-2}} &
\scalebarme{%
\includegraphics[height=\myfig]{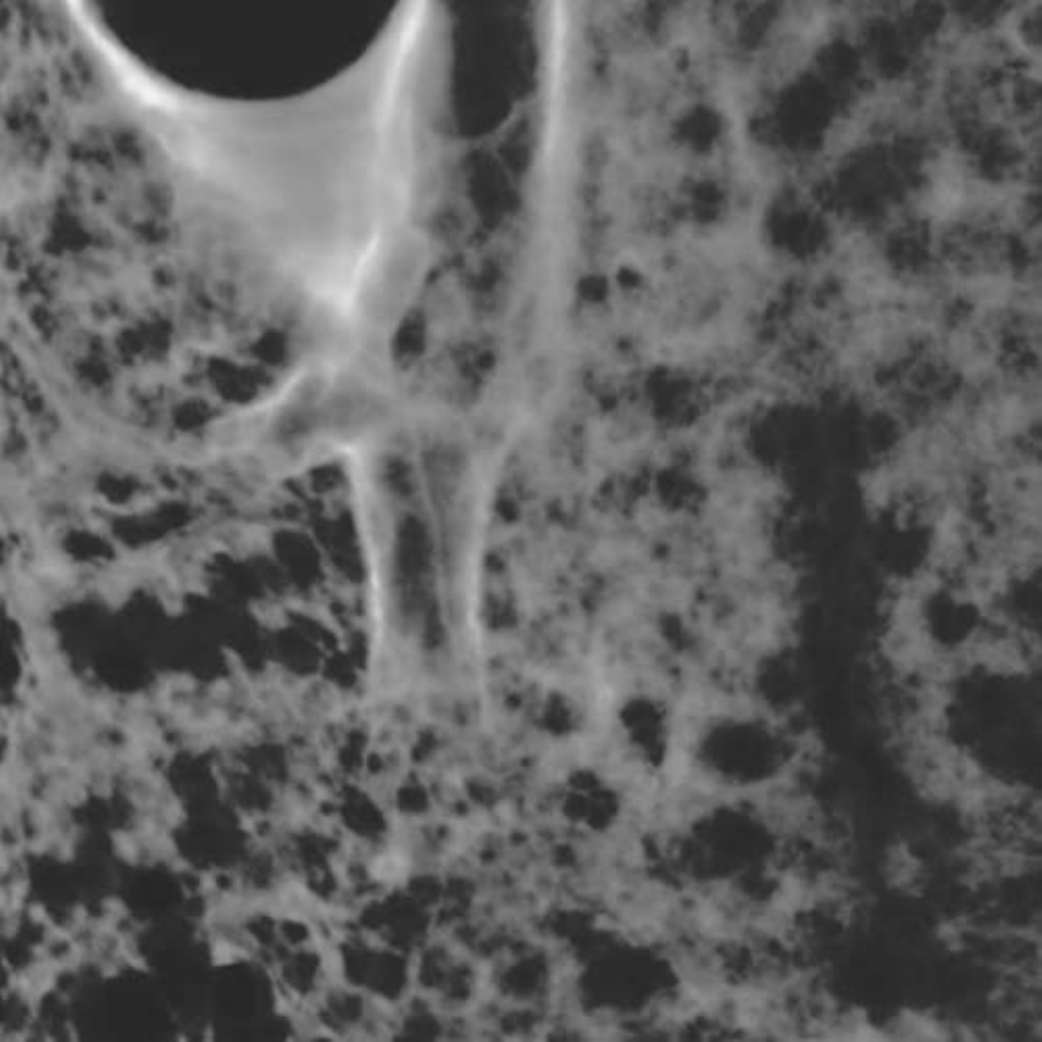}}{%
\includegraphics[width=\myfig]{image/scalebars/scale_ls-500.png}} &
\includegraphics[height=\myfig]{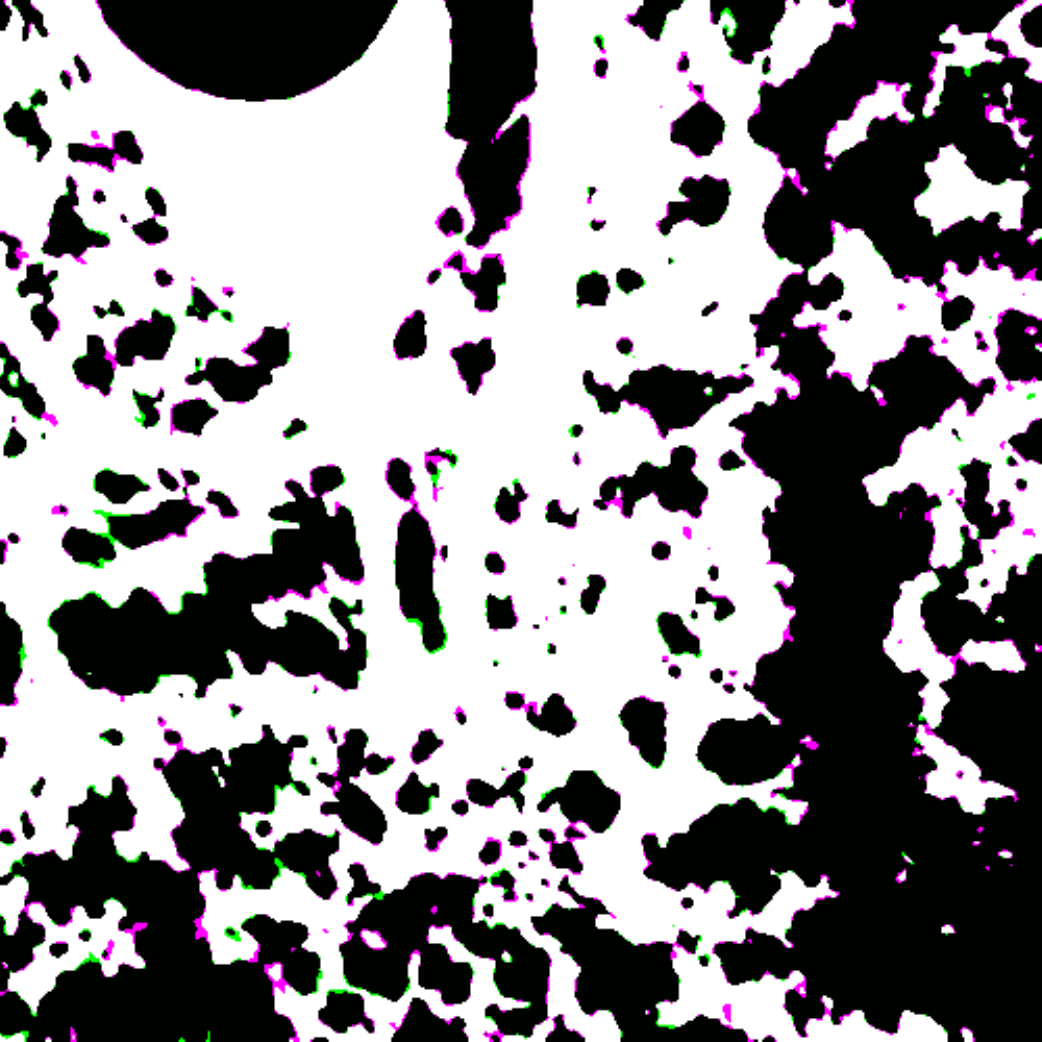} &
\includegraphics[height=\myfig]{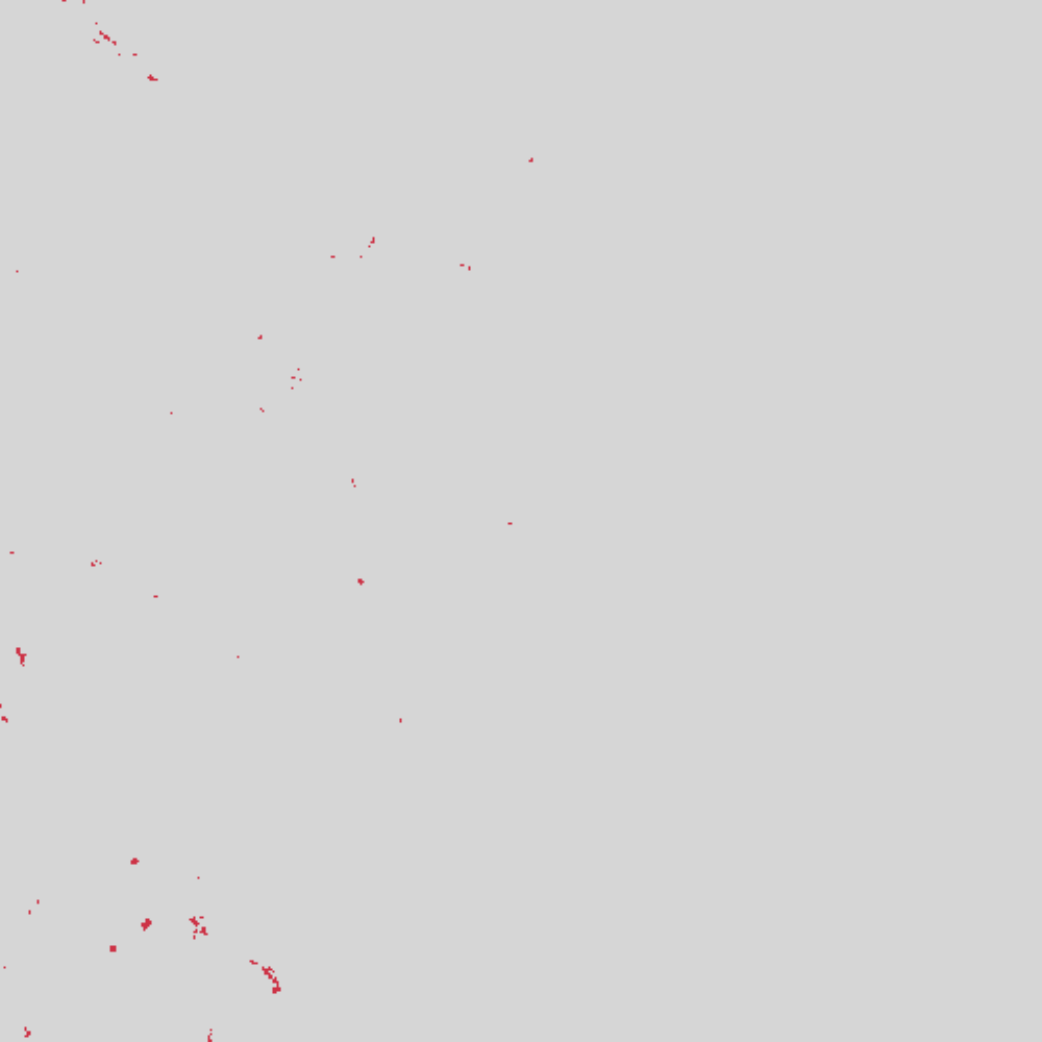} &
\includegraphics[height=\myfig]{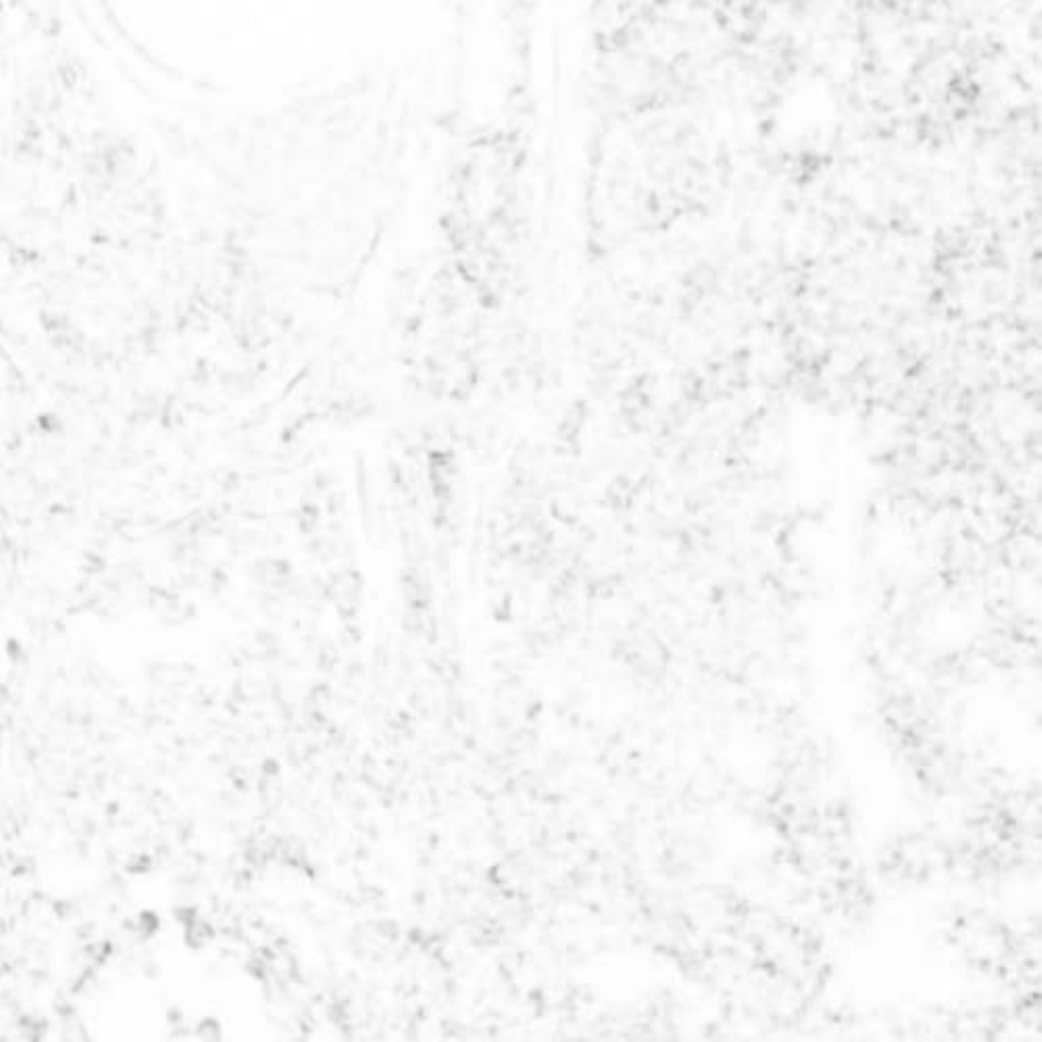} &
\includegraphics[height=\myfig]{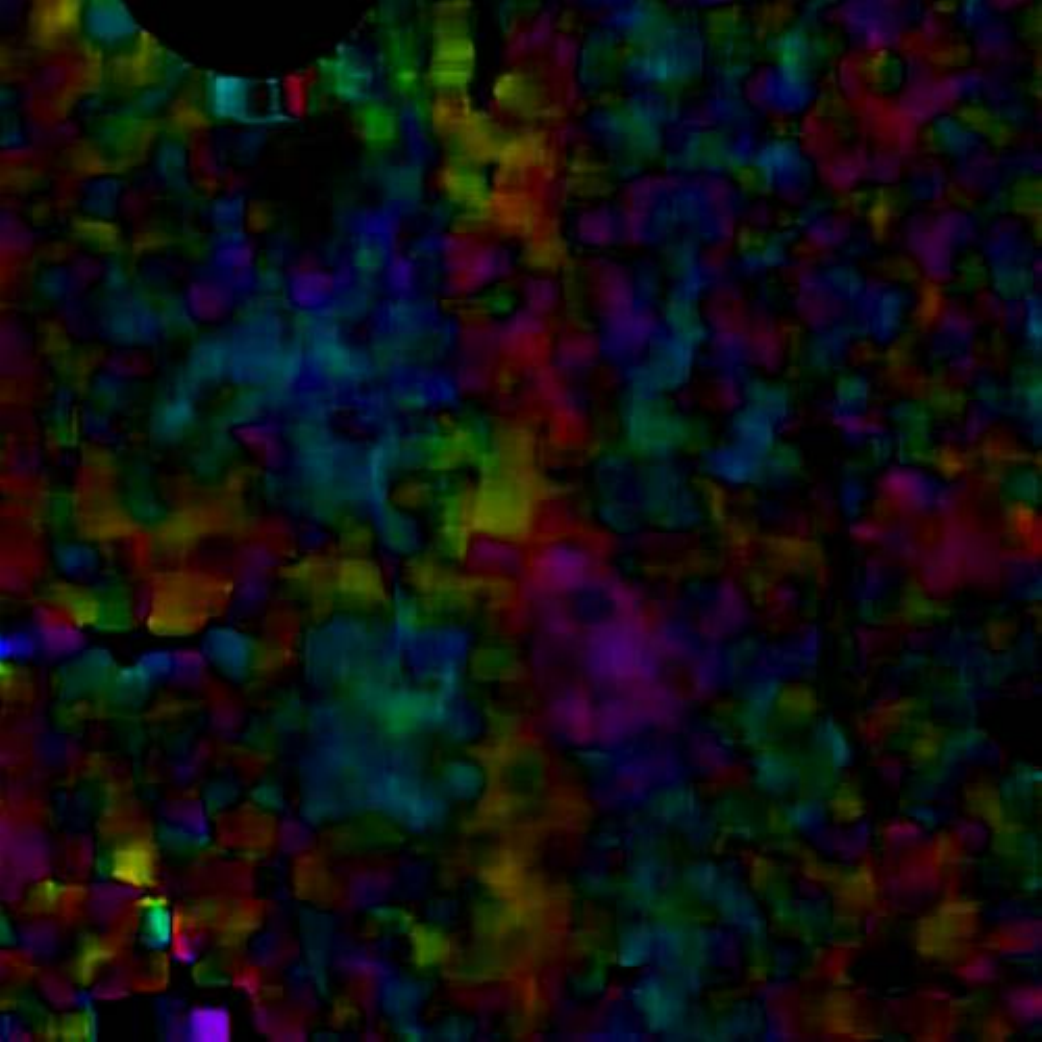}
\\

\rot{\myfig}{Gr.~truth, norm.} &
\scalebarme{%
\includegraphics[height=\myfig]{image/LS/eval/ls_norm/crop_gray_norm_0150.pdf}}{%
\includegraphics[width=\myfig]{image/scalebars/scale_ls-500.png}} &
\includegraphics[height=\myfig]{image/LS/eval/ls_norm/crop_pair__ls-ground-norm_150-151_.pdf} &
\includegraphics[height=\myfig]{image/LS/eval/ls_norm/crop_psnr__ls-ground-norm_150-151_.pdf} &
\includegraphics[height=\myfig]{image/LS/eval/ls_norm/crop_ssim_full__ls-ground-norm_150-151_.pdf} &
\includegraphics[height=\myfig]{image/LS/eval/ls_norm/crop_flow_CROP__ls-ground-norm_150-151_.pdf}
\\

& \multicolumn{1}{c}{Image} & \multicolumn{1}{c}{Dice} & \multicolumn{1}{c}{PSNR} & \multicolumn{1}{c}{SSIM} & \multicolumn{1}{c}{Flow} \\

\end{tabular}
\caption{Evaluation of LS data set with image measures, part~2, mostly variations over \enquote{Deform-SURF}, abbreviated here as \enquote{D.-SURF}. We differentiate the \enquote{stretch} parameter of the non-rigid phase. The default value is the factor \num{5e-3} of the image size. We also test \enquote{Deform-SURF} with the values \num{e-3} and \num{e-2}. The ground truth is normalized. Continued from Fig.~\ref{fig:eval-ls-1}. All scale bars are \SI{500}{\micro\meter}.}
\label{fig:eval-ls-2}
\setlength{\tabcolsep}{\tabcolbackup}
\end{figure*}

\label{sec:eval-pairs} We use two consecutive not distorted images from
the middle of each series as a ground truth for the evaluation of
multiple registration methods. We also use two consecutive images at the
same positioning from the registrations' results. (An~evaluation of the
full series is in the main text.) The input of the registrations is the
series in its completeness after the application of distortions
presented in this paper. Both local distortions and rigid
transformations were applied. Two consecutive result images from the
middle of the registered series are subjected to the quality measures
Jaccard, PSNR and SSIM, as outlined in the main text. We also utilize
two consecutive locally distorted images to produce comparison values.
In the image-based evaluation we show a \xroi{500}{500} crop from the
image center to highlight details.

The tables below show the Jaccard, PSNR and SSIM values for the ground
truth (undistorted images), as well as the locally deformed images. In
part, the values for the normalized ground truth images were also given,
to give an idea about the impact of intensity variations on the ground
truth measures. Same intensity variations as in \enquote{normalized
ground truths} were also present in registrations' inputs.

Hence, the effect of the non-rigid registrations can be read as an
improvement between \enquote{locally distorted} and \enquote{normalized
ground truths}, but there was also a global rigid transformation.
Undoing it with a rigid registration might have lead to a worse starting
point for the non-rigid registration than \enquote{locally distorted}.

\hypertarget{results-of-the-pair-wise-evaluation}{%
\section{Results of the pair-wise
evaluation}\label{results-of-the-pair-wise-evaluation}}

\sisetup{round-mode = figures, round-precision = 3}

Basically, Figs.~\ref{fig:eval-ct-1} and~\ref{fig:eval-ct-2} show the
pair-wise image-based evaluations for the CT data set;
Figs.~\ref{fig:eval-em-1} and~\ref{fig:eval-em-2} show the pair-wise
evaluation for the EM data set; Figs.~\ref{fig:eval-ls-1}
and~\ref{fig:eval-ls-2} show the pair-wise evaluation for the LS data
set. Table~\ref{tab:eval-ct} shows the numerical values of the pair-wise
image-based metrics for CT; Table~\ref{tab:eval-em} shows these for EM;
Table~\ref{tab:eval-ls} for LS. The full-series evaluation is in the
main text.

\hypertarget{ct-data-set}{%
\subsection{CT data set}\label{ct-data-set}}

Consider Fig.~\ref{fig:eval-ct-1}. Both rigid-only methods leave some
incongruences, these are then greatly reduced by \enquote{Deform-SURF}
\citep{lobachev_feature-based_2017}. \enquote{GS}
\citep{gaffling_gauss-seidel_2015} also works on rigidly pre-registered
input, but produces a very different image of residual movements in
optical flow visualization. It reaches also very good values \wrt other
measures, \eg SSIM. The \elastix-based registration
(Fig.~\ref{fig:eval-ct-2}) reaches even better values \wrt SSIM and
PSNR. We see more movement in Dice visualization, but it is not
reflected in Jaccard measure (Tab.~\ref{tab:eval-ct}). The apparent
reason might be the varying thresholding methods. \enquote{Blending} has
some residual movement, as seen in PSNR and optical flow visualization.
The goal of \enquote{Blending} was a coarse alignment, hence the lower
quality measures. Notice, however, that it copes with the random global
transformations on its own, without the initial \enquote{Rigid-SURF}
stage. In \enquote{local only} we see some global movement with is a
residue of a local distortion at a larger scale. Not surprisingly, there
are still some residues of it left in rigid methods, as discussed above.

Consider the ground truth as the \enquote{ideal}, fully corresponding
image pair. Small signal in PSNR means little change. It is also visible
in the visualizations as less red. SSIM and optical flow visualizations
show that these differences are local and evenly distributed across the
region. In registered images there are typically more differences; those
are also more concentrated in a region. Such concentrated
\enquote{change} means stronger local transformations.

The visualized differences in the ground truth appear to be less than in
most of the registration results (all in Fig.~\ref{fig:eval-ct-1}) and
also more scattered across the image: compare, \eg the optical flow
visualization between ground truth, \enquote{GS}, and
\enquote{Deform-SURF}. As for \elastix{}
\citep{klein_elastix:_2010, shamonin_fast_2014}, there are \emph{less}
differences between two consecutive images than in the ground truth. The
SSIM and optical flow visualization of \elastix results show the
differing parts to be more concentrated in the certain areas (\eg the
bottom part and the first third part of the crop) when compared to the
same areas of the ground truth. Such a behavior is hinting at
overfitting. In the ground truth the \enquote{change} is more evenly
distributed across the image. (We can easily make statements about the
localization of the movement or local differences with the visualization
of optical flow. Optical flow highlights the movement across consecutive
images. However, in its present form, it is rather hard to make
statements about the \emph{magnitude} of the movement across multiple
visualizations. In this work we need to rely on other measures for a
comparable assessment.)
\dhcomment{I had a hard time understanding what is explained here. I think I still don't truly understand it, but this might just be because I have *no* experience with registration and registration measures.}

If we look at the numerical values, we observe that all three measures
are \emph{lower} for the ground truth than for \elastix. The verdict
appears to be that \elastix (with our parameter file) over-registers the
CT data set. (The over-registration or over-fitting of the registration
is basically the \enquote{banana problem}, too much correspondence is
created.) However, there is still some noise in the data that is
irrelevant for real-world tasks. The noise, however, still contributes
to the measures and is also the subject of registration---we did not use
a mask. Such \enquote{unnecessary} alignments of the noise might explain
the too large values for \elastix. Still, the GS method is remarkable in
how close it gets to the values of the ground truth quality measures.

\hypertarget{em-data-set}{%
\subsection{EM data set}\label{em-data-set}}

We have applied a prior individual normalization in EM data set
\citep{buchacker_assessment_2019}, hence it might be less fair to
compare the registered images to original data, that has additionally
16~bit depth. Hence, we also show the normalized ground truth images.

The differences between rigidly registered images with SIFT and SURF in
Fig.~\ref{fig:eval-em-1} appear to be the effect of different
normalizations---we were attempting to undo the individual normalization
of the image in inputs of the \enquote{Rigid-SURF} method. Again, we see
residual incongruences with rigid-only methods. Both
\enquote{Deform-SURF} (Fig.~\ref{fig:eval-em-1}) and
\elastix (Fig.~\ref{fig:eval-em-2}) have multiple small
\enquote{movements} in optical flow visualization, as opposed to a more
or less uniform color of a global shift. \enquote{Blending}
(Fig.~\ref{fig:eval-em-1}) shows somewhat uncompensated distortion in
the middle of the image, in Dice, SSIM, and optical flow visualizations.
It appears that in this method the effect of our local distortion
(Fig.~\ref{fig:eval-em-2}) has not been corrected completely, but the
random global transformation was undone well.

From the visualizations, \enquote{GS} (Fig.~\ref{fig:eval-em-1}) and
normalized ground truth have definitely less movement than,
\eg \enquote{Deform-SURF}. The movement in \enquote{GS} seems more
evenly distributed, normalized ground truth has a \enquote{hot spot} in
the optical flow visualization. Curiously, the original ground truth has
both more details (because it is not clamped-down 16~bit data) and more
movement can be detected therein (compare flows and SSIM of the
normalized and original ground truths). At the same time, PSNR is higher
with the original image pair.

Numerically, \elastix has best Jaccard measures (both at 100 and 150
thresholds, indicated with postfix in the following), PSNR, and SSIM
among all registrations. Especially with Jaccard~100,
\enquote{Deform-SURF} is a very close follow-up, at \SI{99.6}{\percent}
of the performance of \elastix. Now, if we look at the measures' values
of the objective ground truths---the luxury we did not have before in an
evaluation of registrations of serial sections---an issue is apparent.
Both Jaccard~100 and Jaccard~150 values for \elastix and (less so) for
\enquote{Deform-SURF} are \emph{higher} than for the normalized ground
truth! This issue might indicate an over-optimization by those
registrations.

To give some values, Jaccard~150 measure for \elastix is
\SI{0.74833198}{\percent} \emph{higher} than normalized ground truth,
same method has also \SI{7.3431064}{\percent} \emph{higher} PSNR than
normalized ground truth; \enquote{Deform-SURF} reaches
\SI{95.739758}{\percent} of the normalized ground truth SSIM. Notably,
from the visualizations we would deem \enquote{GS} as a very good
registration, comparable, if not beating \elastix. But numerically,
\elastix has much higher values.

\begin{figure*}[!t]
\centering
\ifthenelse{\boolean{arxivpreprint}}{%
%
%
\newlength{\heightvolren}
\setlength{\heightvolren}{0.395\textwidth}
\newlength{\wvolren}
\setlength{\wvolren}{0.249\textwidth}
\subfloat[][\elastix]{\label{fig:volren-a}\includegraphics[height=\heightvolren]{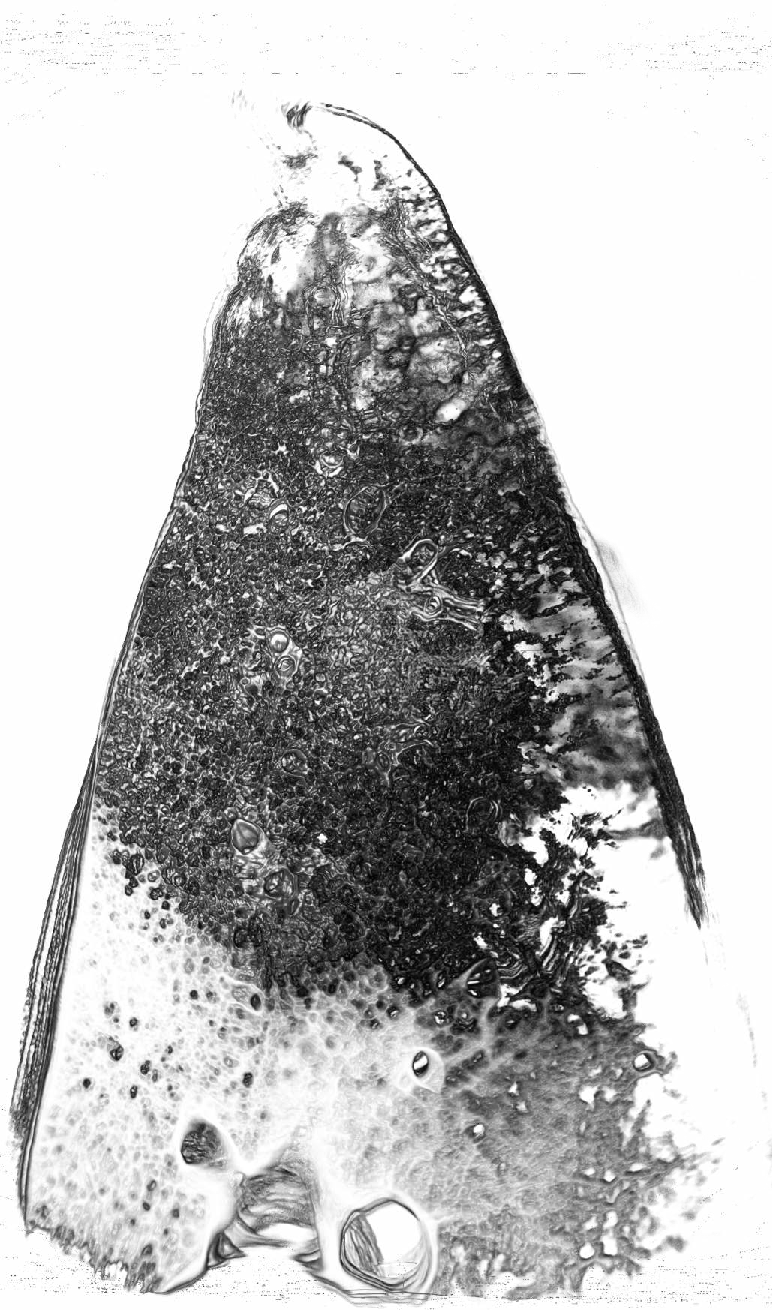}}\hfill%
\subfloat[][Deform-SURF]{\label{fig:volren-b}\includegraphics[height=\heightvolren]{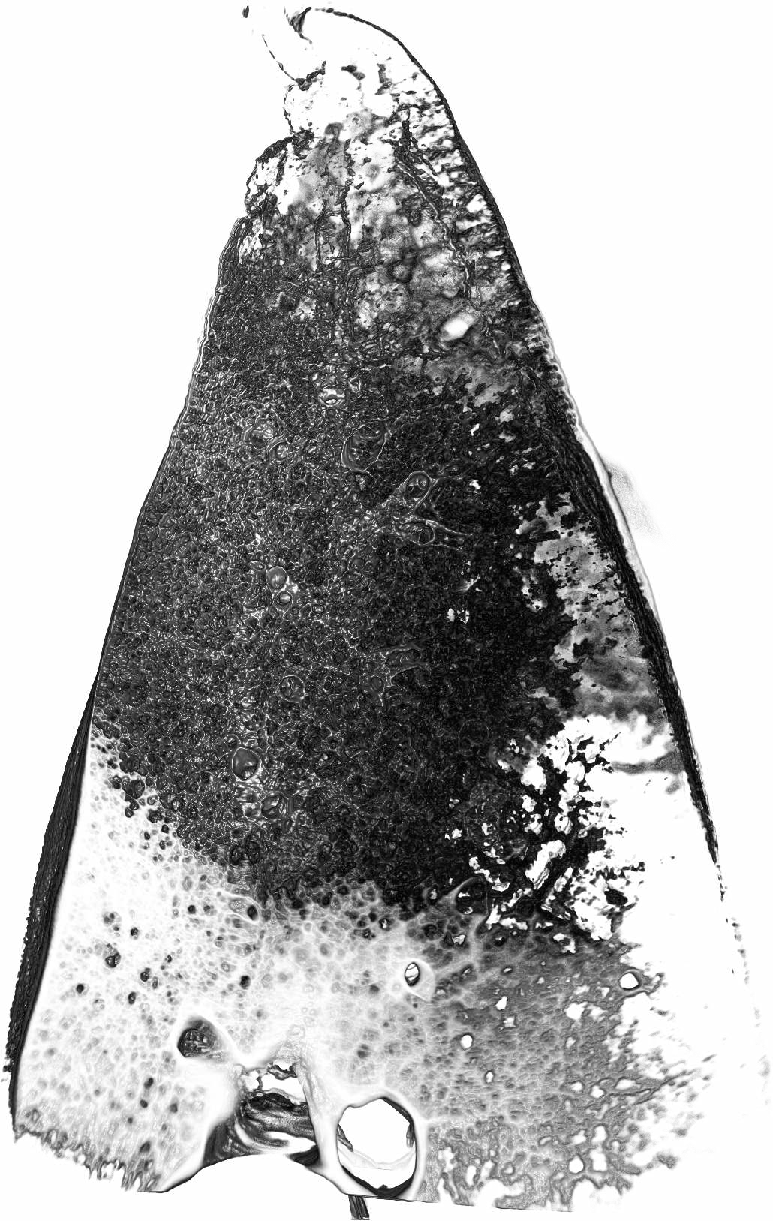}}\hfill%
\subfloat[][local distortions]{\label{fig:volren-c}\includegraphics[height=\heightvolren]{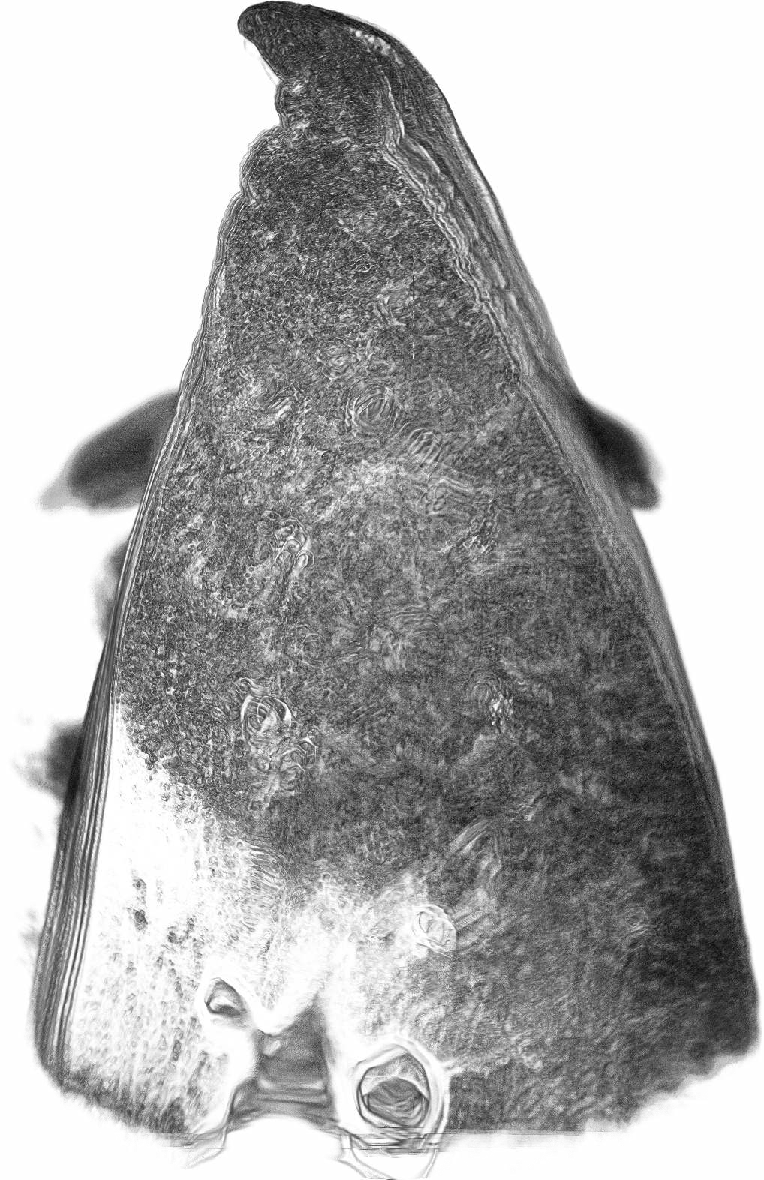}}\hfill%
\subfloat[][ground truth,\\normalized]{\label{fig:volren-d}\includegraphics[height=\heightvolren]{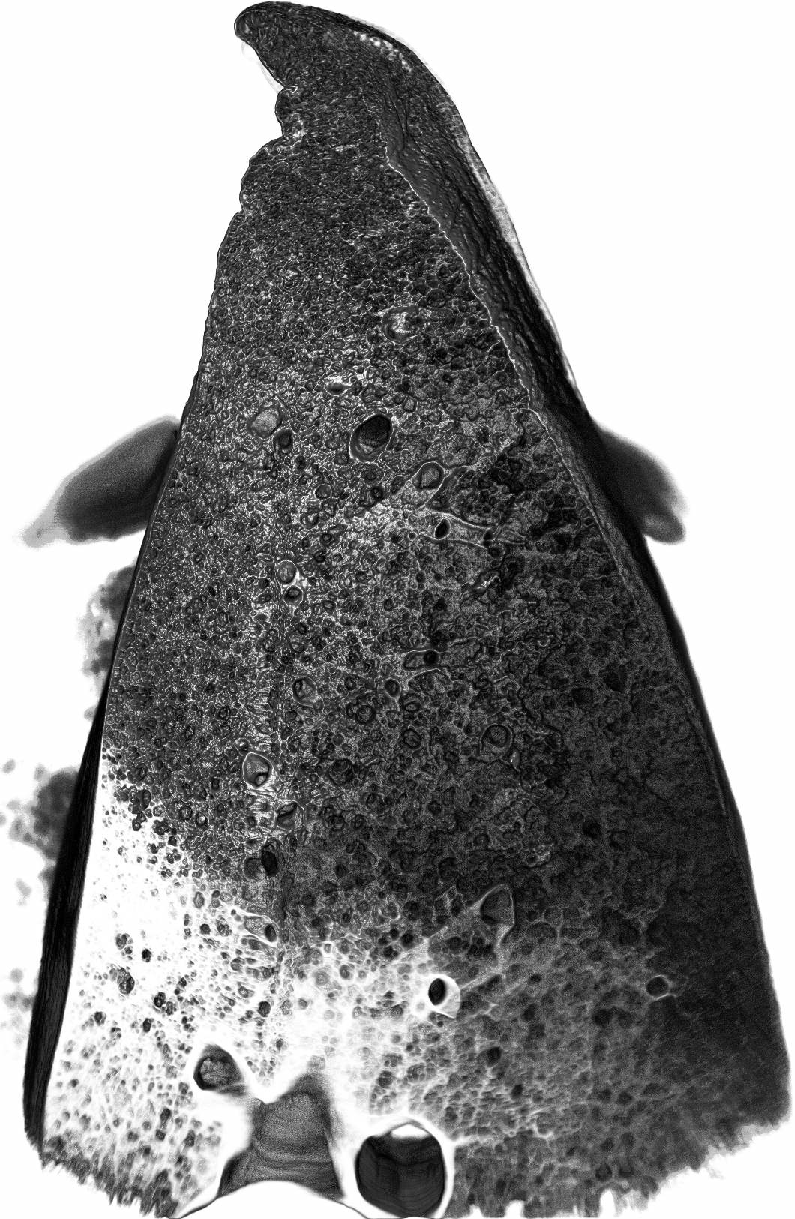}}
}{%
%
%
\newlength{\heightvolren}
\setlength{\heightvolren}{0.38\textwidth}
\newlength{\wvolren}
\setlength{\wvolren}{0.249\textwidth}

\subfloat[][\elastix]{\label{fig:volren-a}\includegraphics[width=\wvolren]{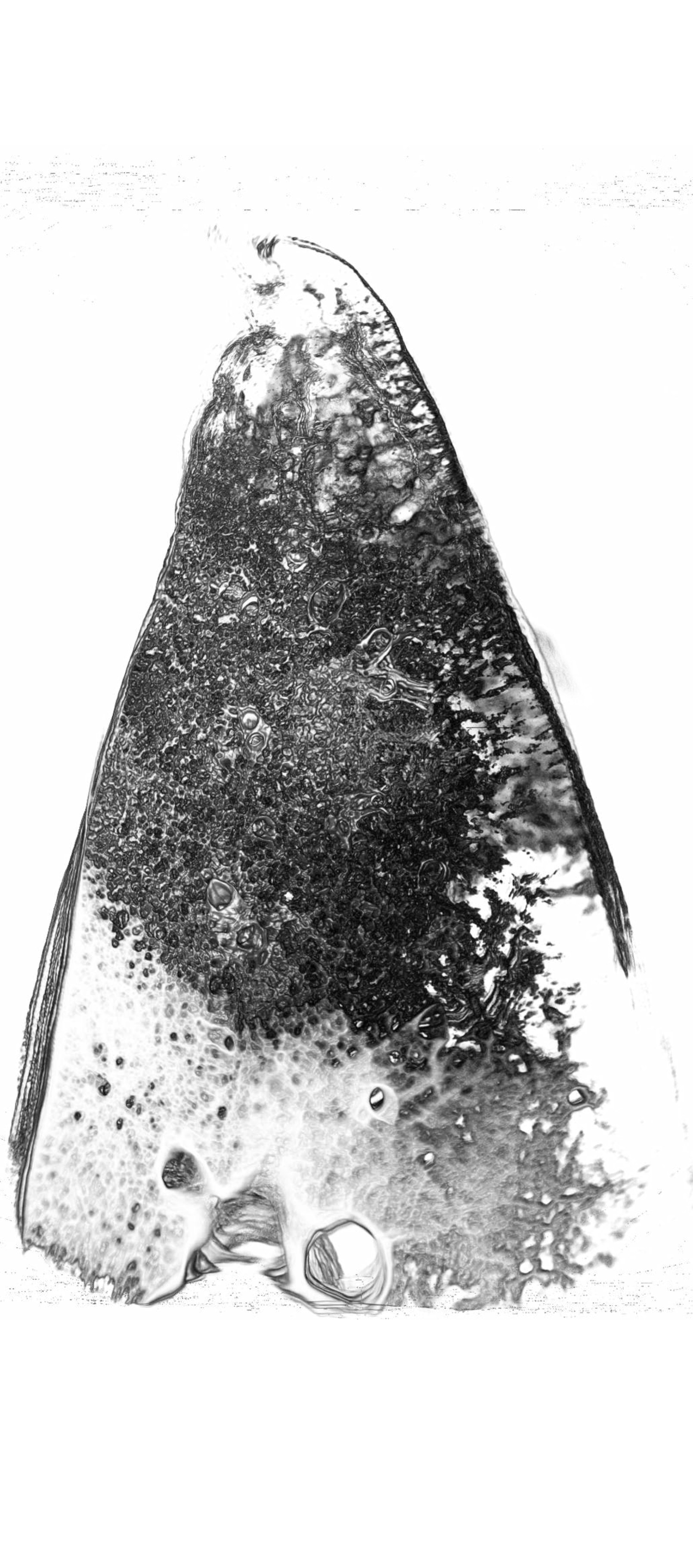}}\hfill%
\subfloat[][Deform-SURF]{\label{fig:volren-b}\includegraphics[width=\wvolren]{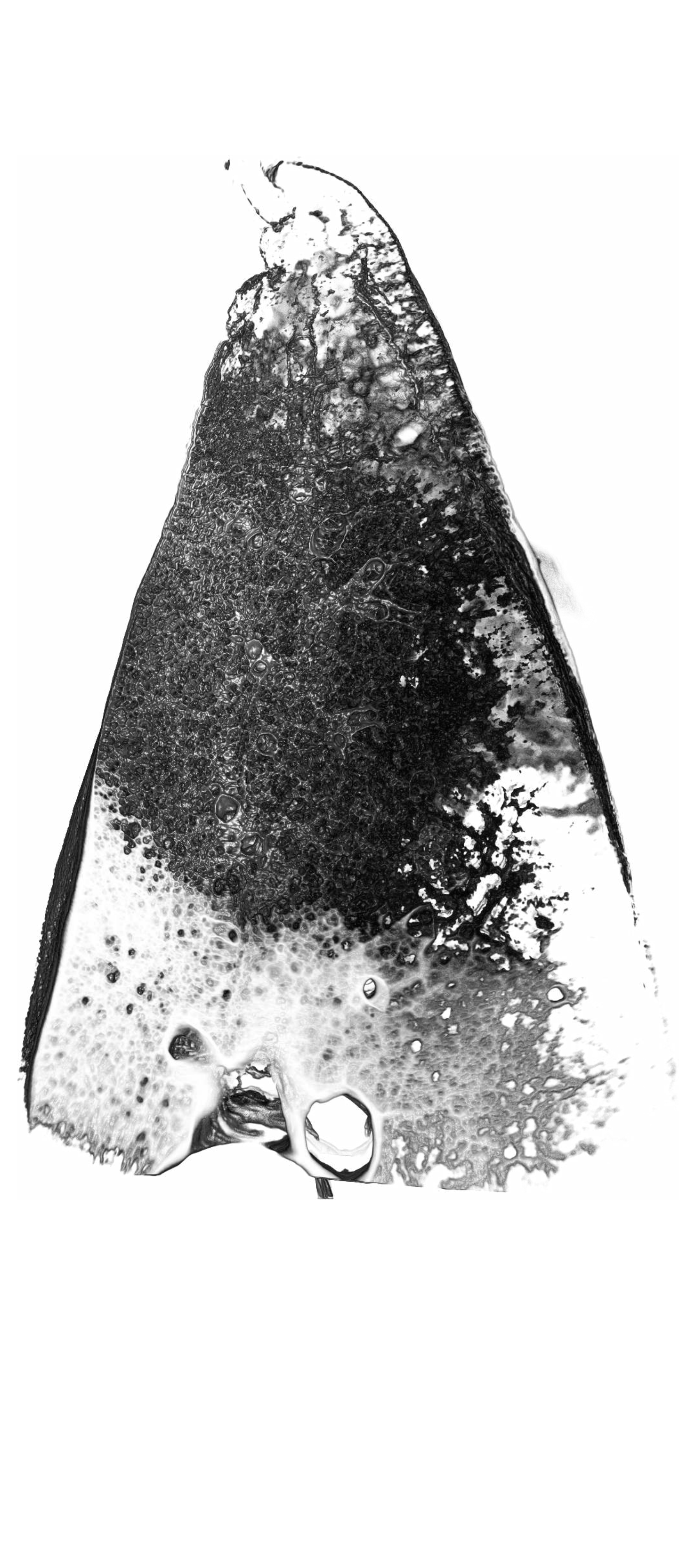}}\hfill%
\subfloat[][local distortions]{\label{fig:volren-c}\includegraphics[width=\wvolren]{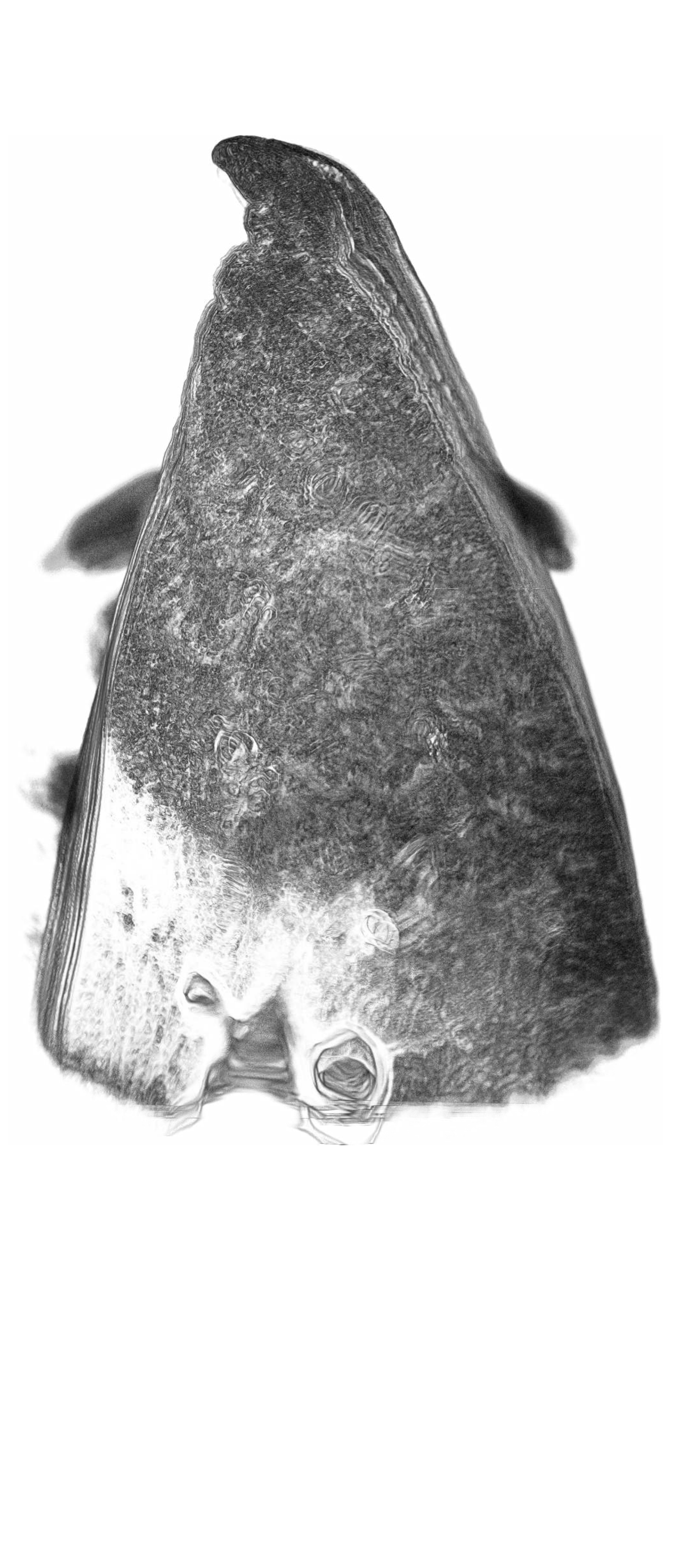}}\hfill%
\subfloat[][ground truth, normalized]{\label{fig:volren-d}\includegraphics[width=\wvolren]{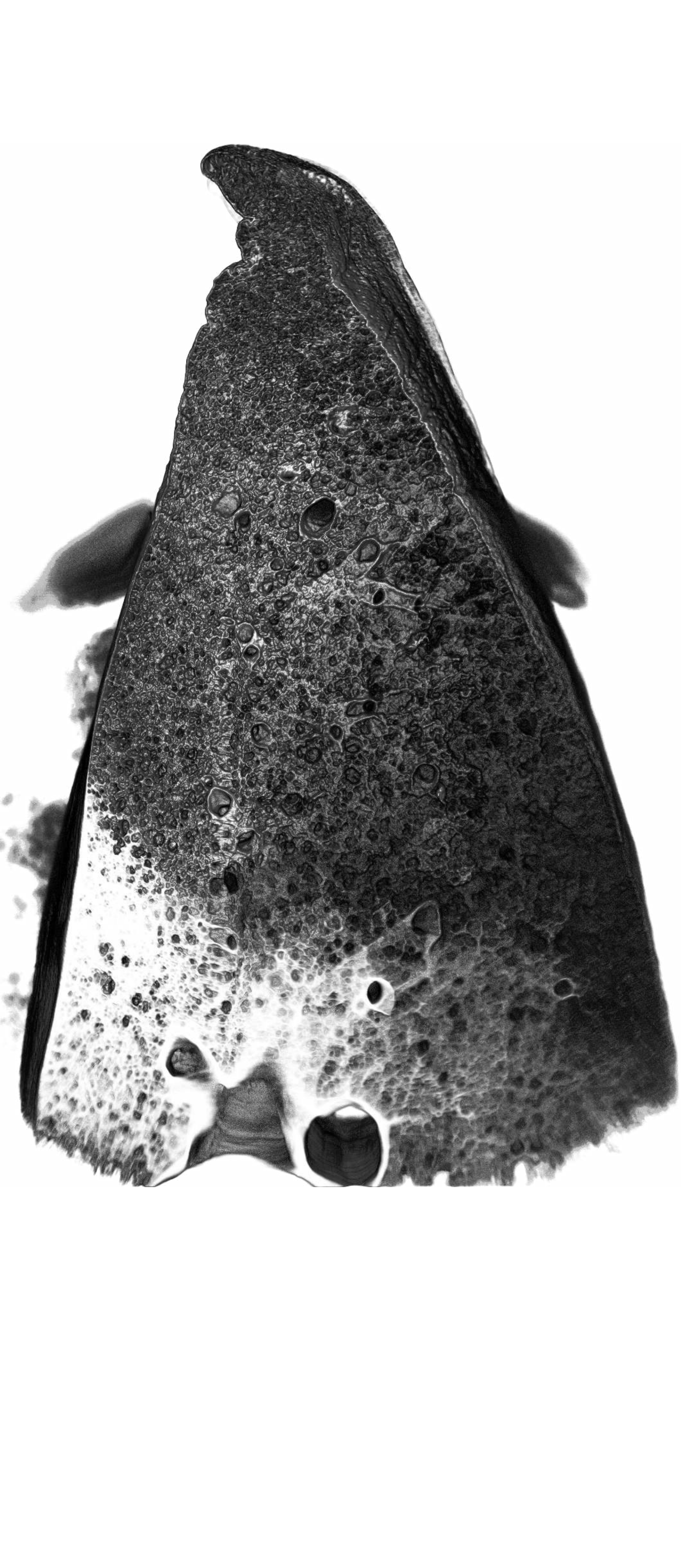}}
}
\caption{Selected volume renderings for LS data set.}
\label{fig:volumes}
\end{figure*}

\hypertarget{ls-data-set}{%
\subsection{LS data set}\label{ls-data-set}}

In LS data set the individual normalization was also used for challenge
data, in methods with SURF we also aimed to fix those discrepancies that
we created in the challenge data to ensure better rigid registration.
This explains the visual differences between the results. As we compare
images from each of the series to each other and not across the series,
this problem is less an issue. Still, it highlights the importance of
the input normalization for good registration results. We also used this
data set to study the effect of different distortion magnitudes in
\enquote{Deform-SURF}.

In Figure~\ref{fig:eval-ls-1} we see a larger movement in the rigid
method. It originates from our local distortions in the region of
interest (\enquote{Local dist.}), but cannot be undone with rigid
methods only. \enquote{Blending} still suffers from these distortions,
but restores the coarse correspondence well. \enquote{GS} recovers very
well from those local movements, however it might over-register, based
on visual comparison with normalized ground truth. Both in optical flow
visualizations for \enquote{GS} and for the ground truth we see multiple
small \enquote{movements}, originating from the fact that we compare two
consecutive non-equal images. However, \enquote{GS} shows \emph{less}
movement than ground truth, hence our above suspicion. Numerical values
would provide more clarity, we will look at them next. As for \elastix,
Dice, PSNR, and SSIM visualizations look very good, similar as in
\enquote{GS}. In \elastix, the visualization of optical flow appears in
general to be lower than the ground truth, but it also shows a
\enquote{hot spot} at the top part of the image. However, as already
mentioned, we should be careful with such comparisons between optical
flow visualizations.

Figure~\ref{fig:eval-ls-2} shows quite confident results from
\enquote{Deform-SURF}, but the \enquote{Rigid-SURF} is even worse than
\enquote{Rigid-SIFT} (Fig.~\ref{fig:eval-ls-1}). We use three different
values for the non-linear \enquote{stretch} in the non-rigid
feature-based registration method \enquote{Deform-SURF}. The
\enquote{stretch} values we state are a factor of image size in pixels,
limiting the magnitude of the \enquote{movement} in the non-linear
registration. When the deformation magnitude is apparently too small
(\num{e-3}), SSIM (and less so, PSNR) visualizations are slightly worse,
but the optical flow shows the problem \enquote{hot spot} near the
center of the image. The default magnitude of \num{5e-3} is already much
better and does not have the aforementioned problem. An even larger
magnitude \num{e-2} also looks similar to the standard setting. Whether
those two parameters in \enquote{Deform-SURF} over-register can be
determined from the numerical values below. For now, we can say that
normalized ground truth is slightly better in PSNR, Dice, and SSIM, but
appears more \enquote{noisy}, but also more \enquote{even} than
\enquote{Deform-SURF}. We attribute the variations in intensity of the
registered images in different methods to the normalization.

Table~\ref{tab:eval-ls} shows numerical values. The Jaccard measure is
computed with threshold~100. \enquote{Rigid-SIFT} was very successful
\wrt Jaccard measure, but PSNR and SSIM are lower than in a typical
non-rigid registration. In a contrast, \enquote{Rigid-SURF} is quite bad
\wrt all three measures. Still, it manages to be a good input for the
non-rigid methods \enquote{Deform-SURF}, GS, and \elastix. Comparing
different \enquote{stretch} values with \enquote{Deform-SURF}
numerically, we see that the default value of \num{5e-3} is slightly
better than the larger value. Too small \enquote{stretch} decreases the
measures slightly, but it is still better than all rigid methods, for
PSNR and SSIM with a larger margin. The \enquote{Blending} method has
some problems, also evident in visualizations. We discussed this issue
above.

The methods \enquote{GS} and \elastix are very similar \wrt Jaccard
measure. Interestingly, PSNR is slightly higher with \elastix, but SSIM
is higher with \enquote{GS}. Both \enquote{GS} and \elastix have better
numerical values than all \enquote{Deform-SURF} methods tested
here---but let us consider the ground truth!

When compared to the (normalized) ground truth, Jaccard and PSNR
measures in all registrations are lower. Still, the Jaccard measures for
\enquote{GS} and \elastix are closer to the ground truth than in other
methods. However, the SSIM value for \enquote{Deform-SURF} is less than
\SI{2}{\percent} \emph{larger} than the actual SSIM for the normalized
ground truth image pair. Still, \enquote{GS} and \elastix have
\emph{even larger} SSIM values than the ground truth:
\SI{2.3118700}{\percent} and \SI{2.7989687}{\percent} correspondingly.
Such larger SSIM values might signal over-registration.

\hypertarget{visual-comparisons-for-ls}{%
\section{Visual comparisons for LS}\label{visual-comparisons-for-ls}}

This section presents selected volume renderings of the full stacks.
While 3D representations typically convey more information, a 3D
overview can show only the most coarse incongruences. We demonstrate
here the volume renderings, but base out actual evaluation on a sequence
of objective measures, as presented above and also in the main paper.

Our volume renderings were produced with ImageVis3D
\citep{fogal_tuvok_2010}. Fig.~\ref{fig:volumes} shows the frontal views
of the LS data set. We used the same, standard settings for all the
images.

\hypertarget{effect-of-individual-normalizations}{%
\section{Effect of individual
normalizations}\label{effect-of-individual-normalizations}}

\label{sec:effect}\label{sec:effect-norm}

\begin{figure}[tbh]
\centering
\includegraphics[width=0.85\linewidth]{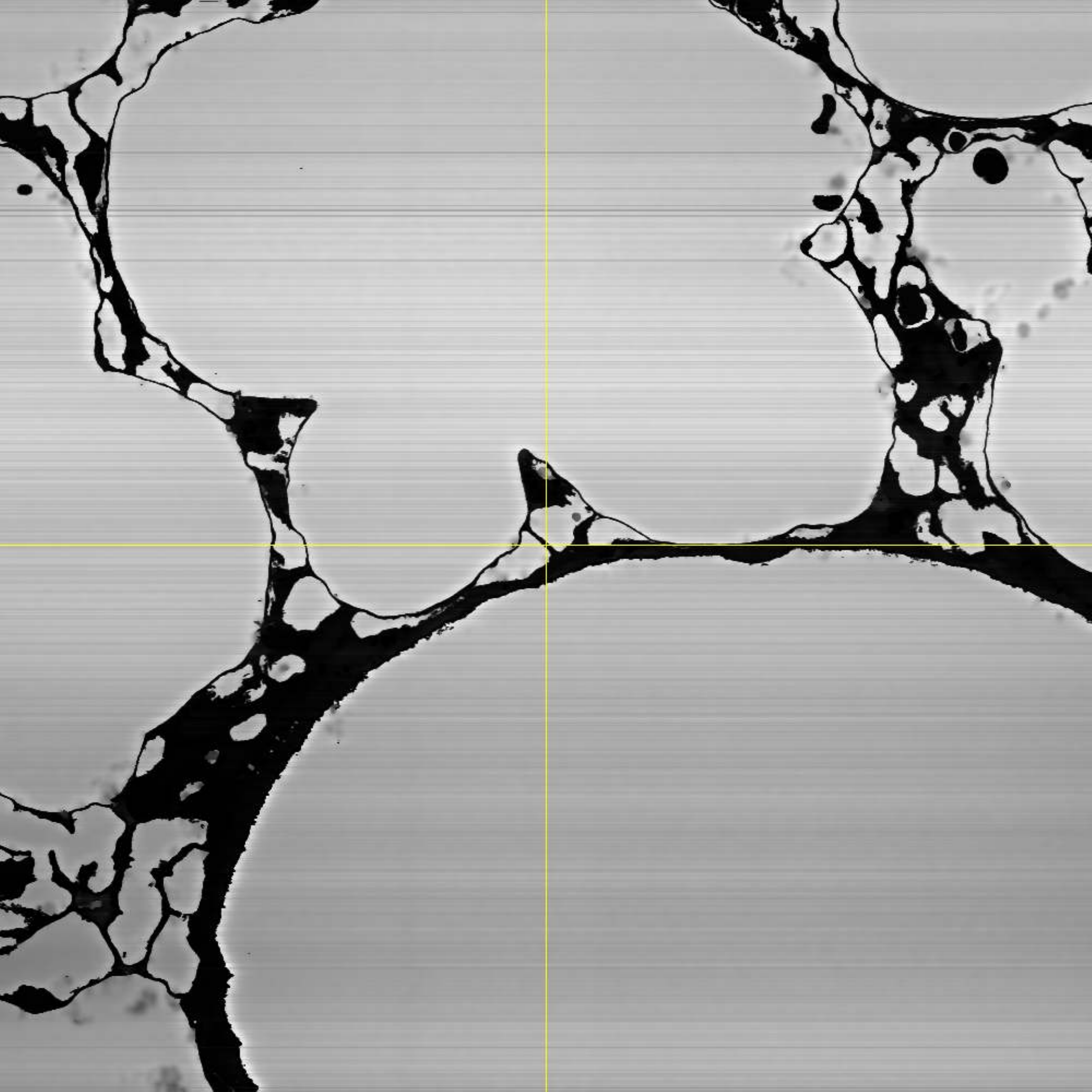}

\caption{Individual normalizations of EM series, $z$ stacks. We show the original data, the ground truth, after individual normalizations.}
\label{fig:ind-norm}
\end{figure}

The \emph{individual} normalizations have an effect of varying intensity
of the images through the series. Such variations should mimic the
effect of varying section thickness that is normally countered by a
\emph{series-wide} normalization prior to the registrations.
Fig.~\ref{fig:ind-norm} showcases the individual normalizations in form
of a \(z\)-stack from EM series.

\hypertarget{effect-of-distortions-on-ls}{%
\section{Effect of distortions on
LS}\label{effect-of-distortions-on-ls}}

\begin{figure}[tbh]
\centering
\subfloat[][]{\label{fig:ls-effect-a}%
\includegraphics[width=0.499\linewidth]{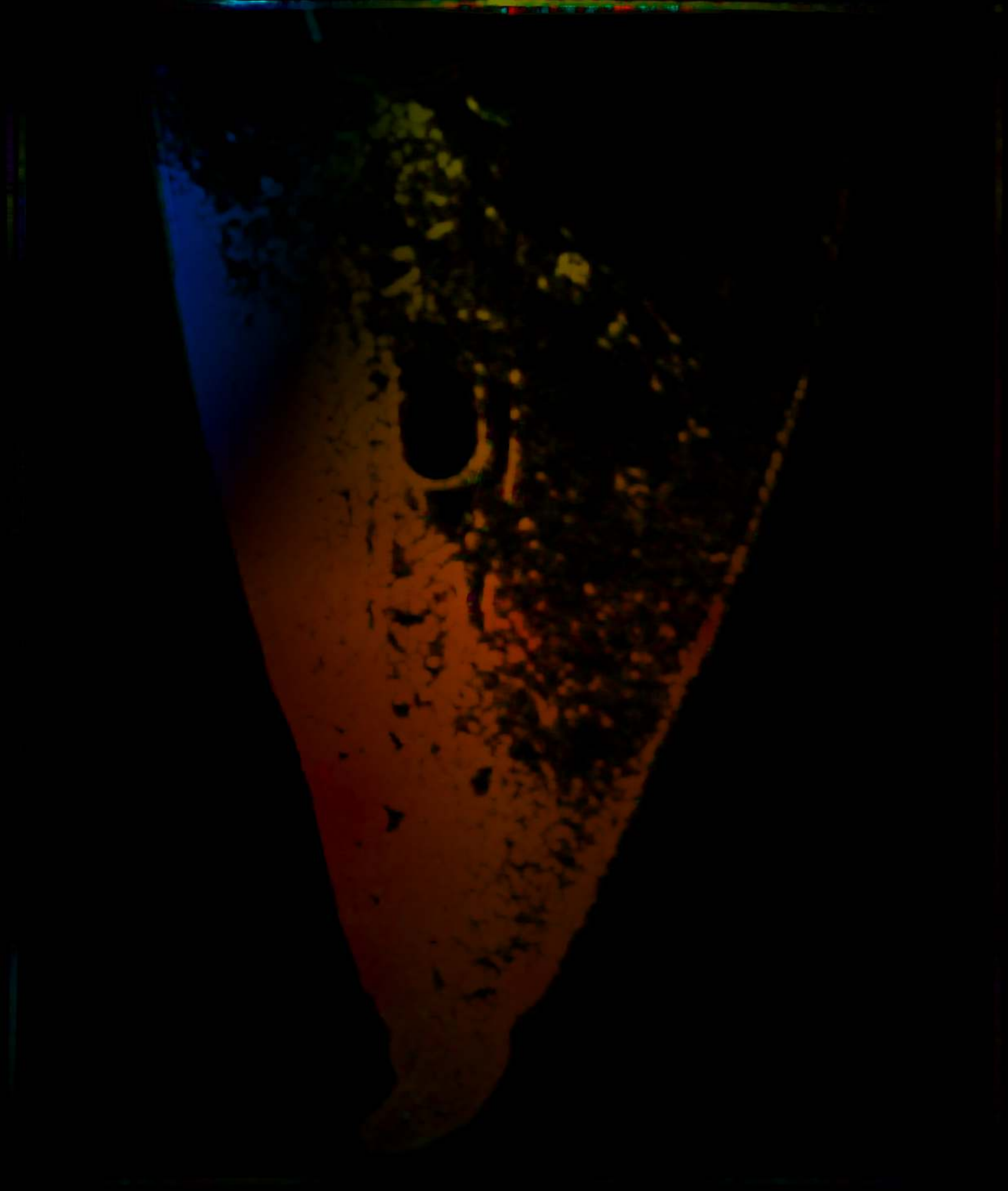}}\hfill%
\subfloat[][]{\label{fig:ls-effect-b}%
\includegraphics[width=0.499\linewidth]{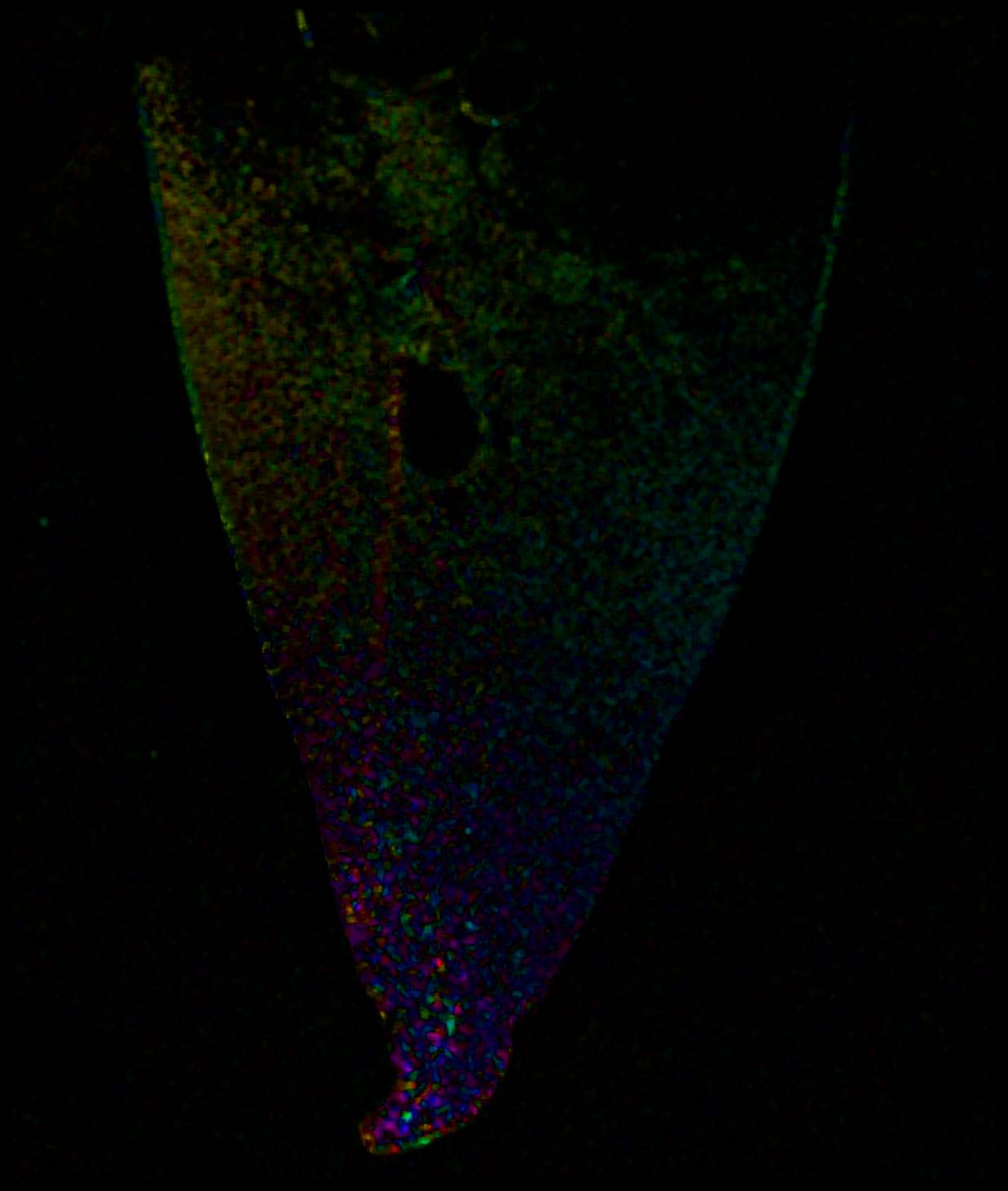}}

\subfloat[][]{\label{fig:ls-effect-c}%
\includegraphics[width=0.499\linewidth]{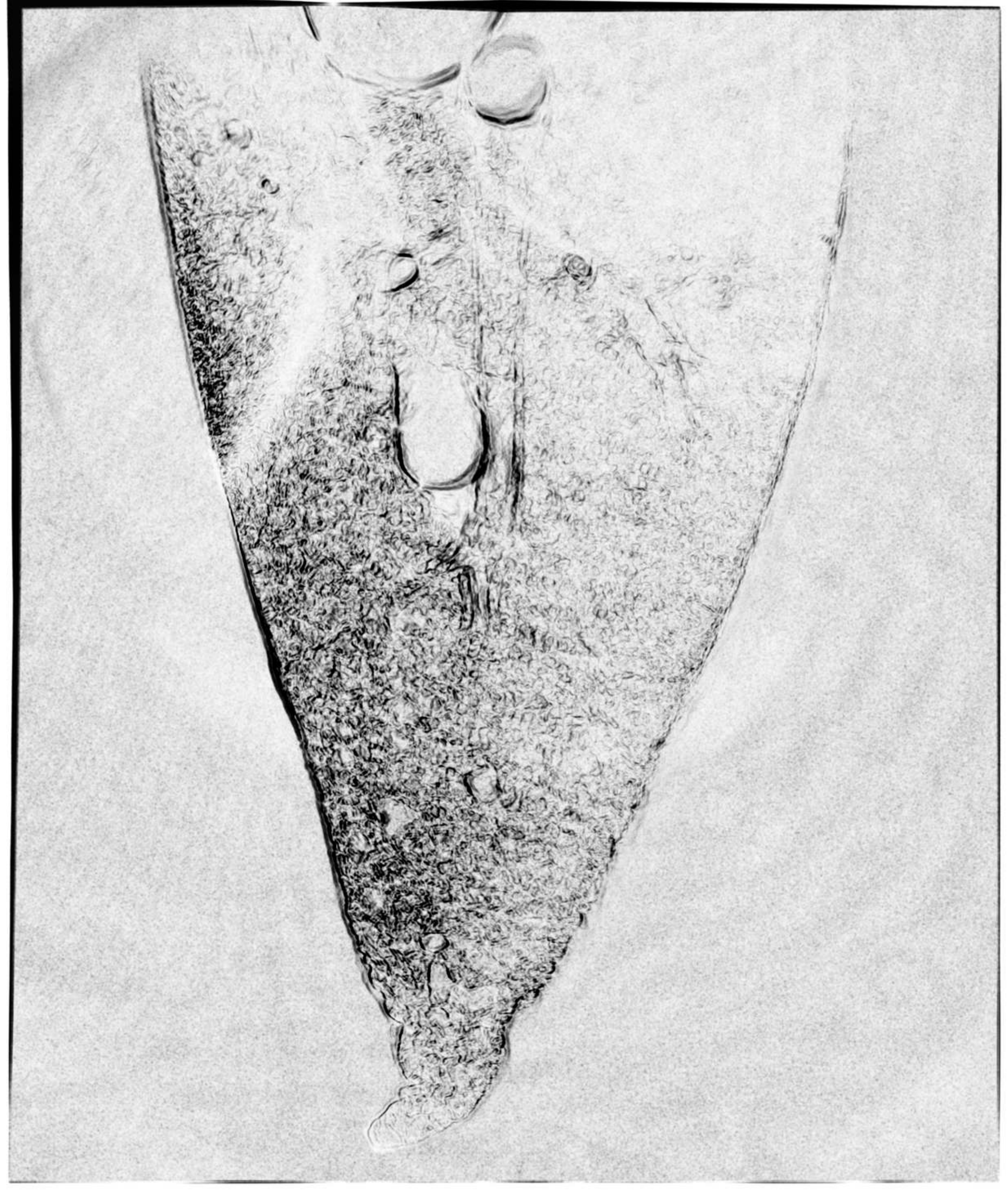}}\hfill%
\subfloat[][]{\label{fig:ls-effect-d}%
\includegraphics[width=0.499\linewidth]{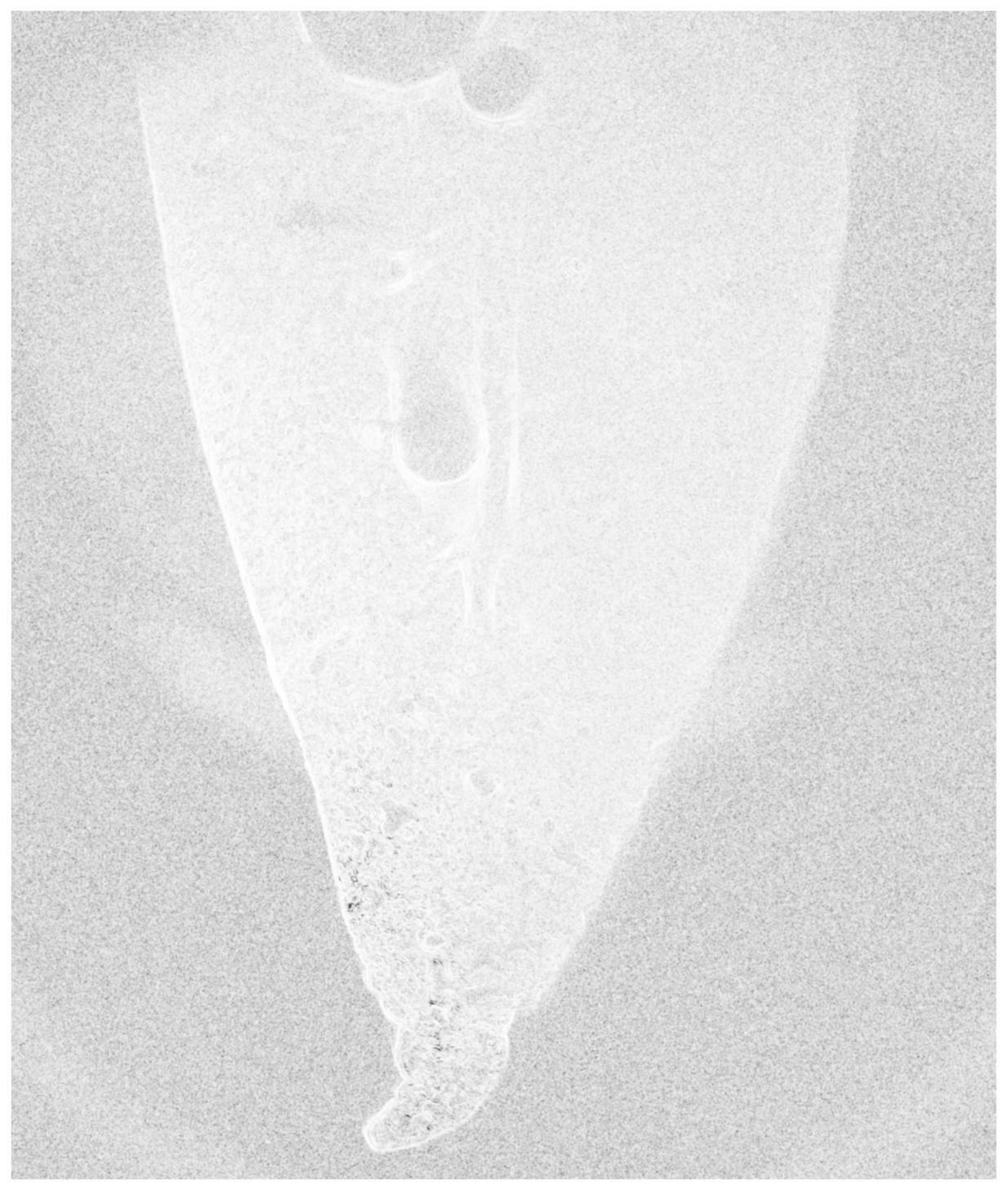}}%

\caption{Consecutive images from LS data set after local distortions with our method \protect\subref{fig:ls-effect-a}, \protect\subref{fig:ls-effect-c}, and before \protect\subref{fig:ls-effect-b}, \protect\subref{fig:ls-effect-d}, \ie on ground truth. Figures \protect\subref{fig:ls-effect-a}, \protect\subref{fig:ls-effect-b} show optical flow visualizations, \protect\subref{fig:ls-effect-c}, \protect\subref{fig:ls-effect-d} show SSIM.}
\label{fig:ls-effect}
\end{figure}

Figure~\ref{fig:ls-effect} shows full optical flow and SSIM
visualization of a consecutive image pair from the LS data set.
Concerning the locally distorted images \subref{fig:ls-effect-a},
\protect\subref{fig:ls-effect-c}, notice that there is some global
movement that creates problems for the rigid-only methods. Some traces
of these problems are still evident in the results of,
\eg \enquote{Blending}, see main text. Panels~\subref{fig:ls-effect-b},
\protect\subref{fig:ls-effect-d} show the normalized ground truth. We
use the same image pair (150--151) that is used everywhere else for the
pair-wise evaluation.

\hypertarget{a-violin-plot}{%
\section{A~violin plot}\label{a-violin-plot}}

\begin{figure*}[p]
\centering
\includegraphics[width=1\linewidth]{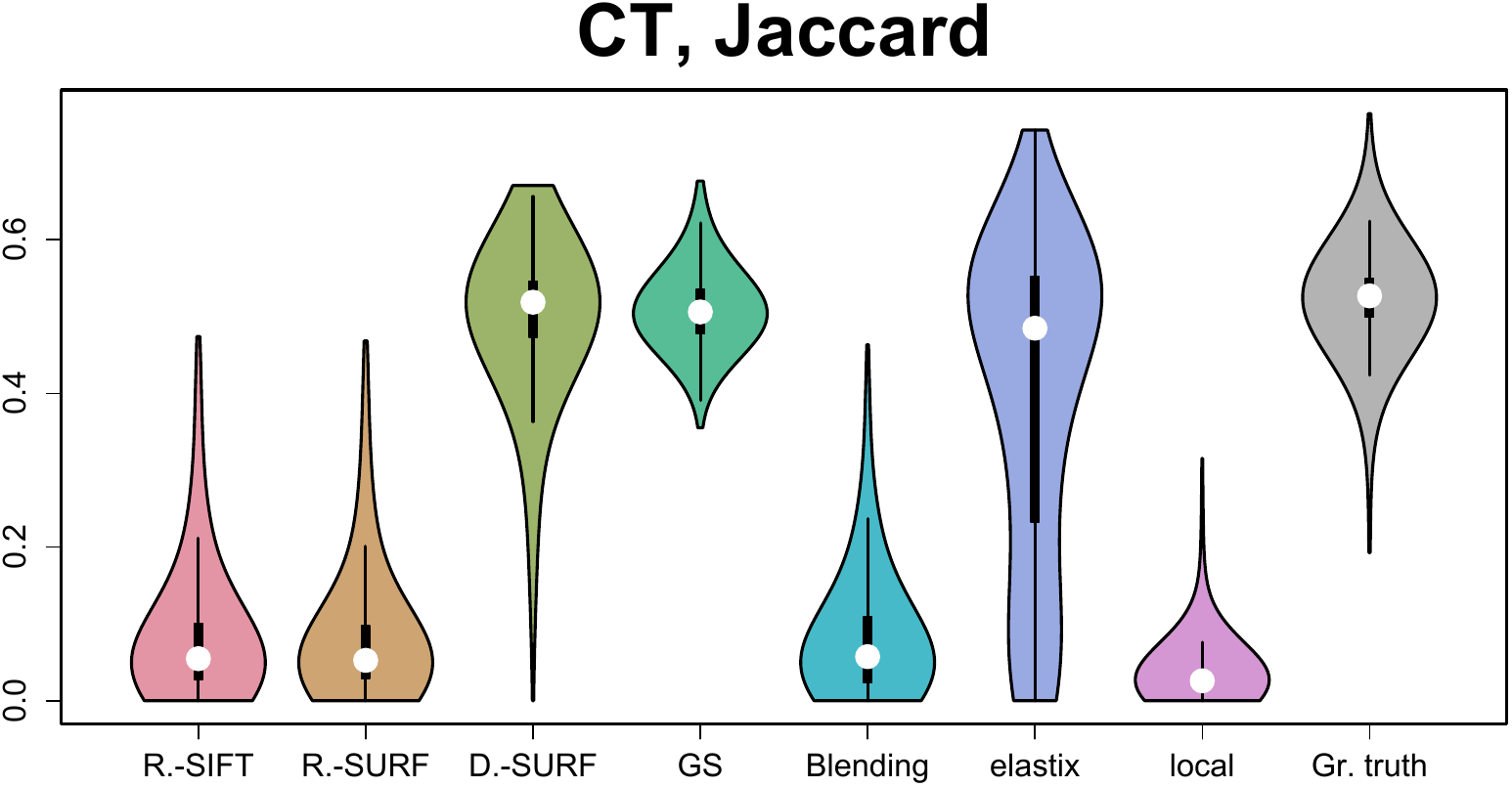}

\caption{Violin plot of the Jaccard measures over the whole CT data set, registered with various methods.}
\label{fig:vioplot}
\end{figure*}

For the statistical evaluation, instead of box plots, violin plots can
be used. We advocated against them in the main paper, as we would be
more interested in the inliers. For the completeness, we show here an
example violin plot in Fig~\ref{fig:vioplot}. It is the Jaccard
evaluation of the whole CT series. We see there some interesting
consequences on the distribution of the outliers, but in this case
little cannot be inferred from the corresponding box plot in the main
paper. It is less clearer to see in this violin plot without training
which method has a higher median, though.

\hypertarget{gallery-of-distortion-visualizations}{%
\section{Gallery of distortion
visualizations}\label{gallery-of-distortion-visualizations}}

\begin{figure*}[p]
\centering
\includegraphics[width=0.124\linewidth]{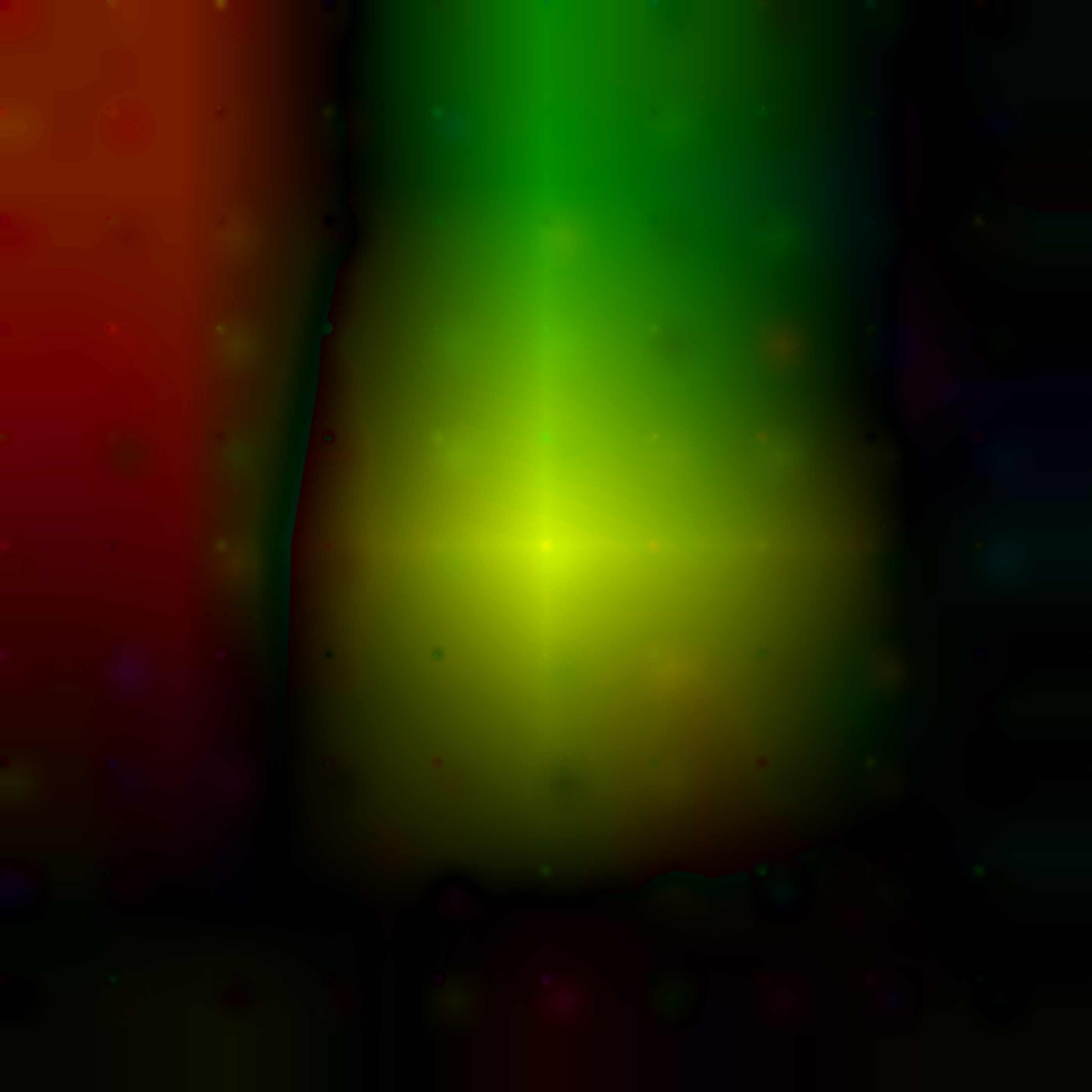}\hfill%
\includegraphics[width=0.124\linewidth]{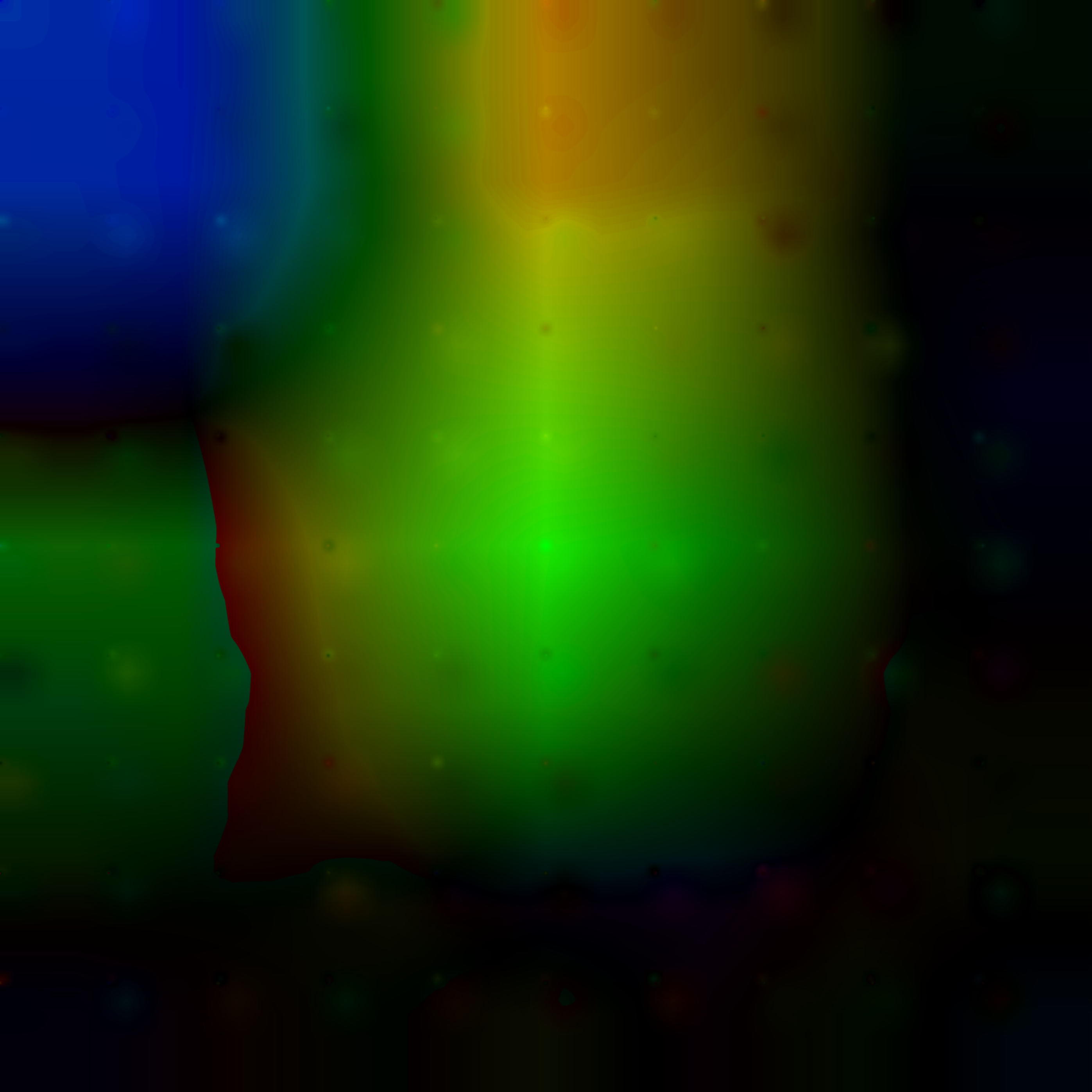}\hfill%
\includegraphics[width=0.124\linewidth]{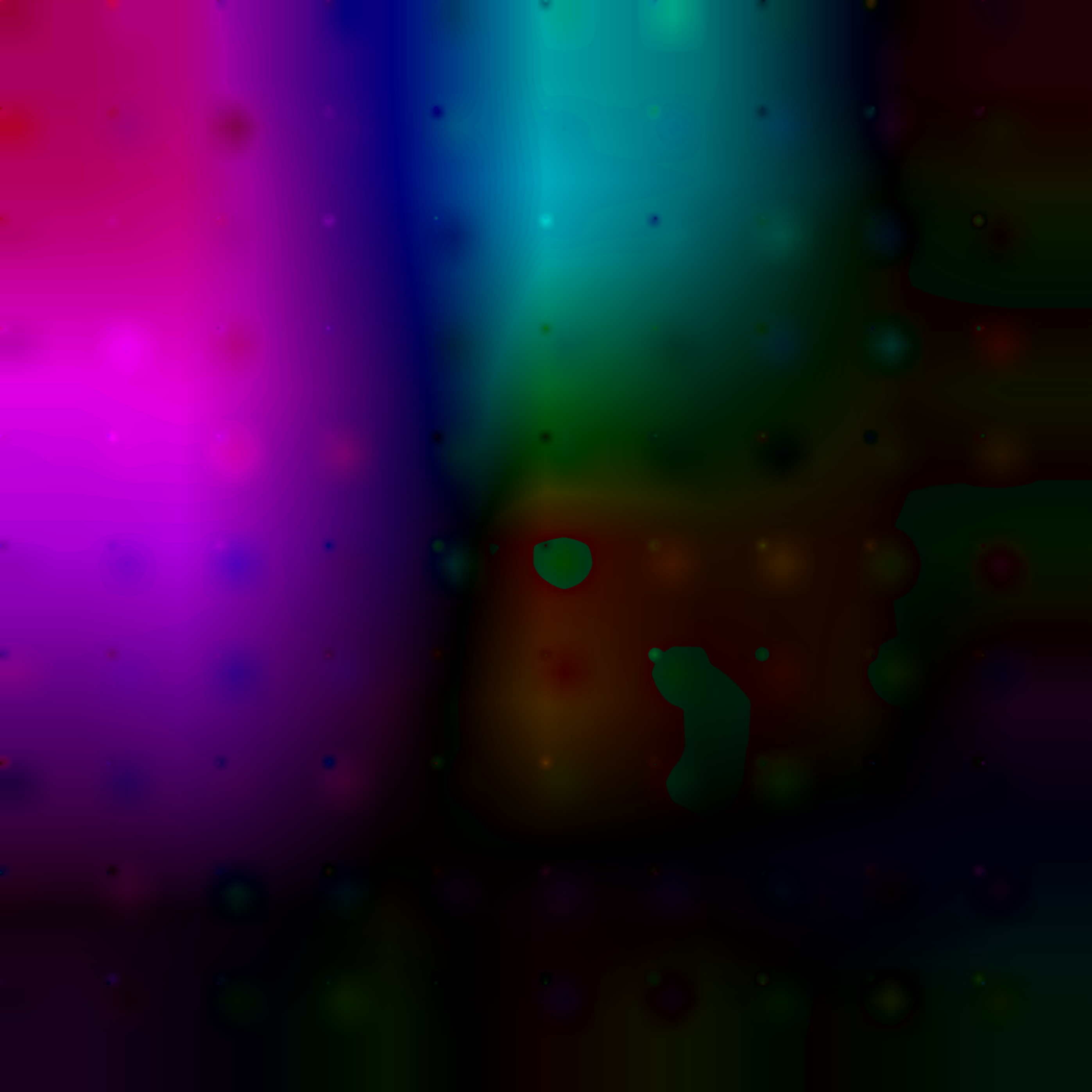}\hfill%
\includegraphics[width=0.124\linewidth]{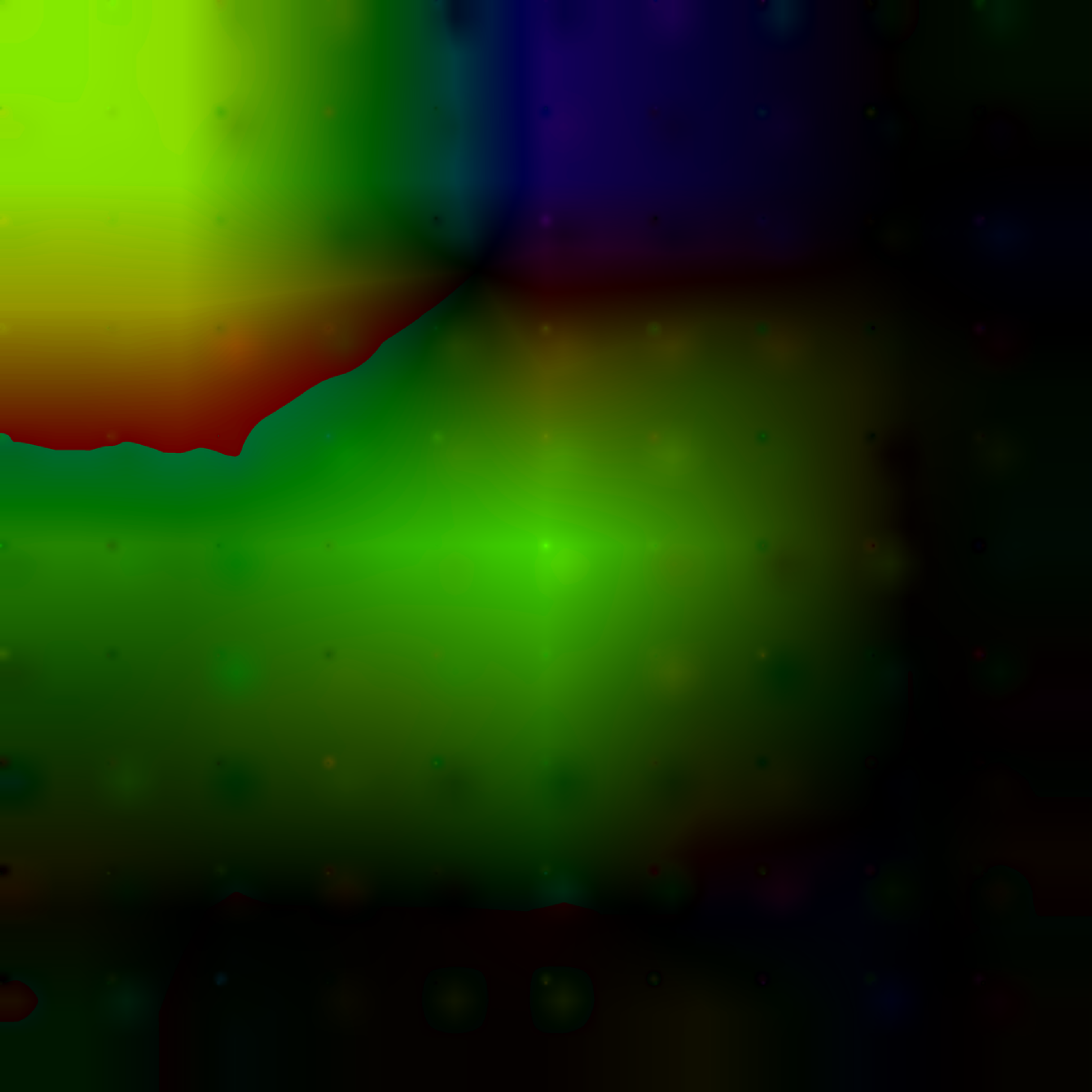}\hfill%
\includegraphics[width=0.124\linewidth]{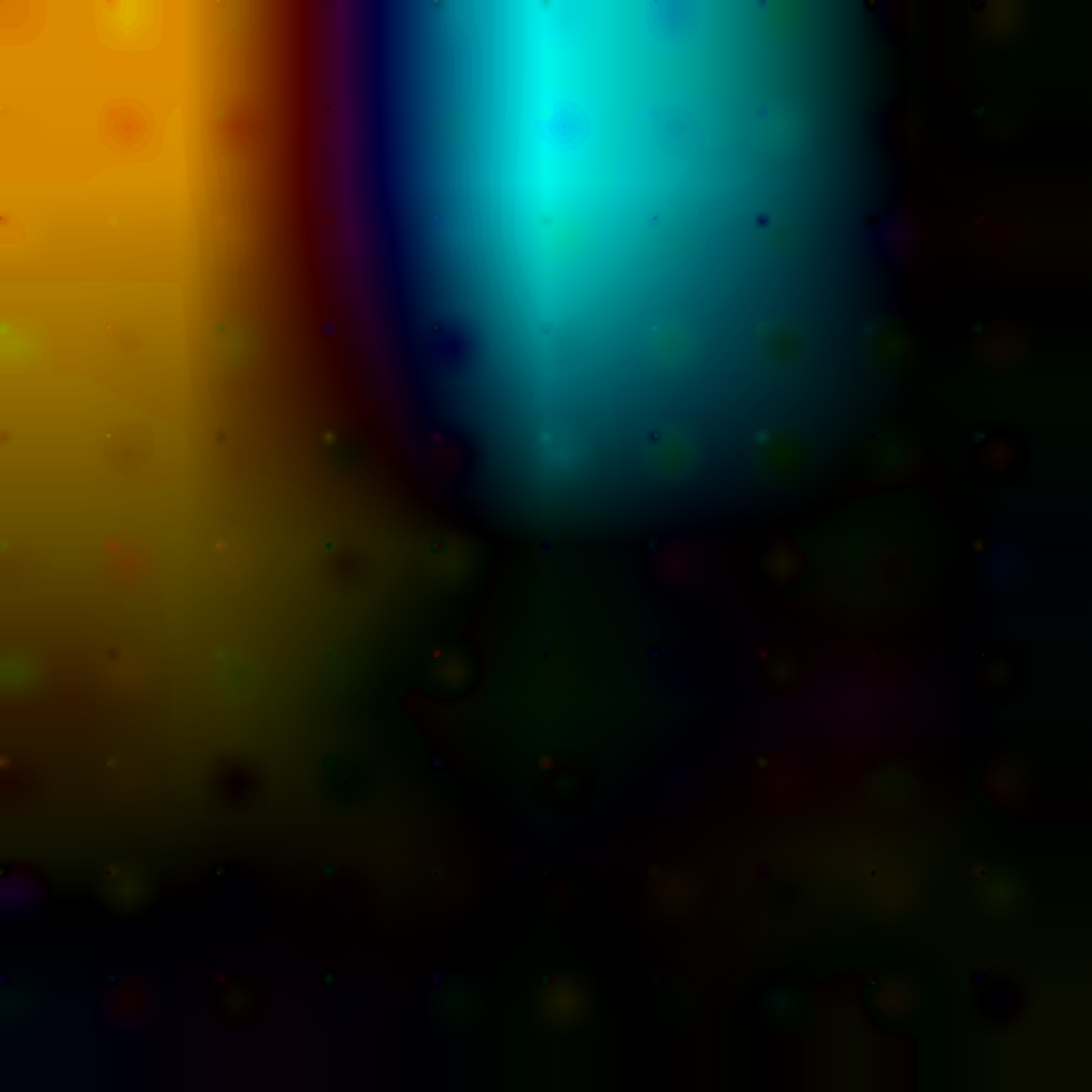}\hfill%
\includegraphics[width=0.124\linewidth]{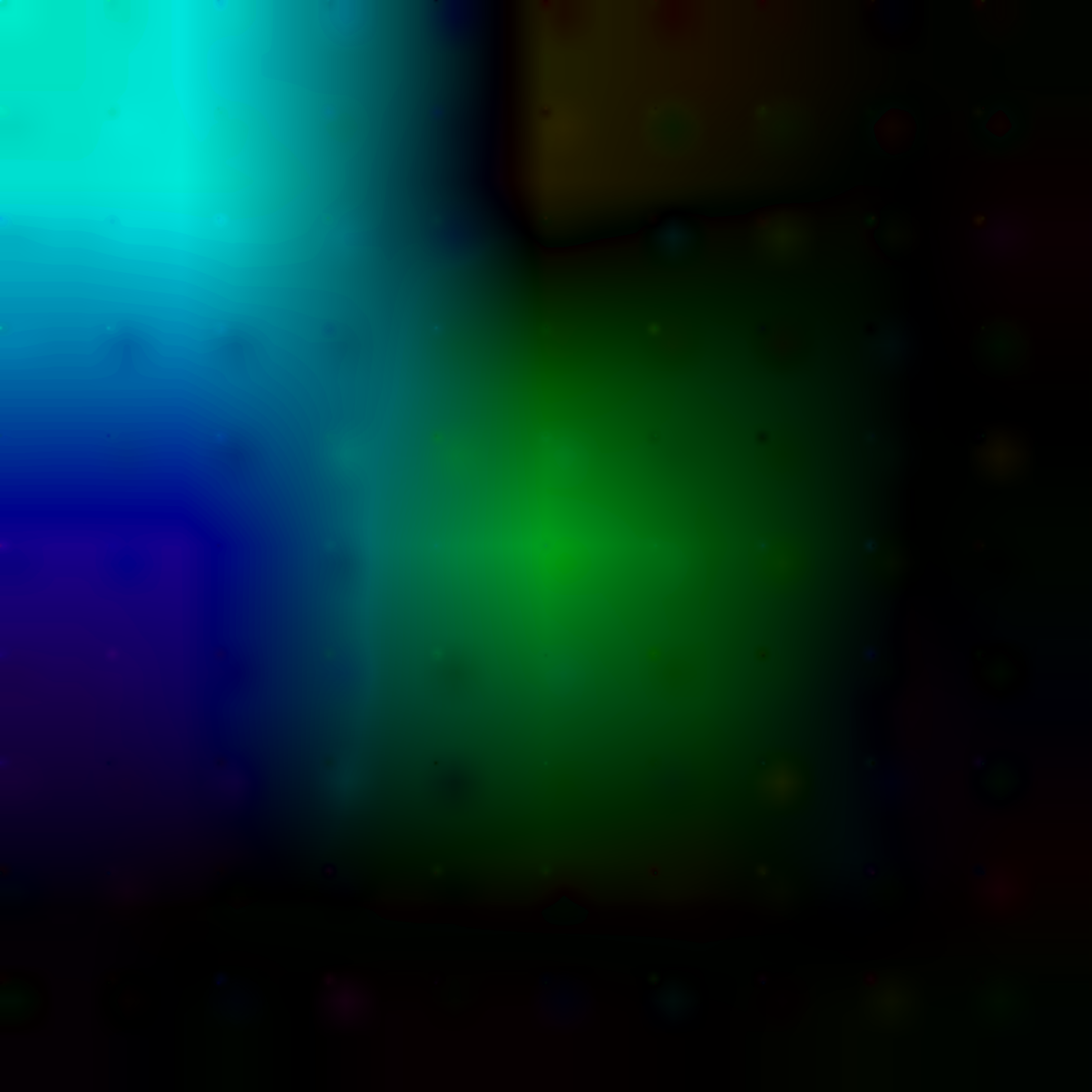}\hfill%
\includegraphics[width=0.124\linewidth]{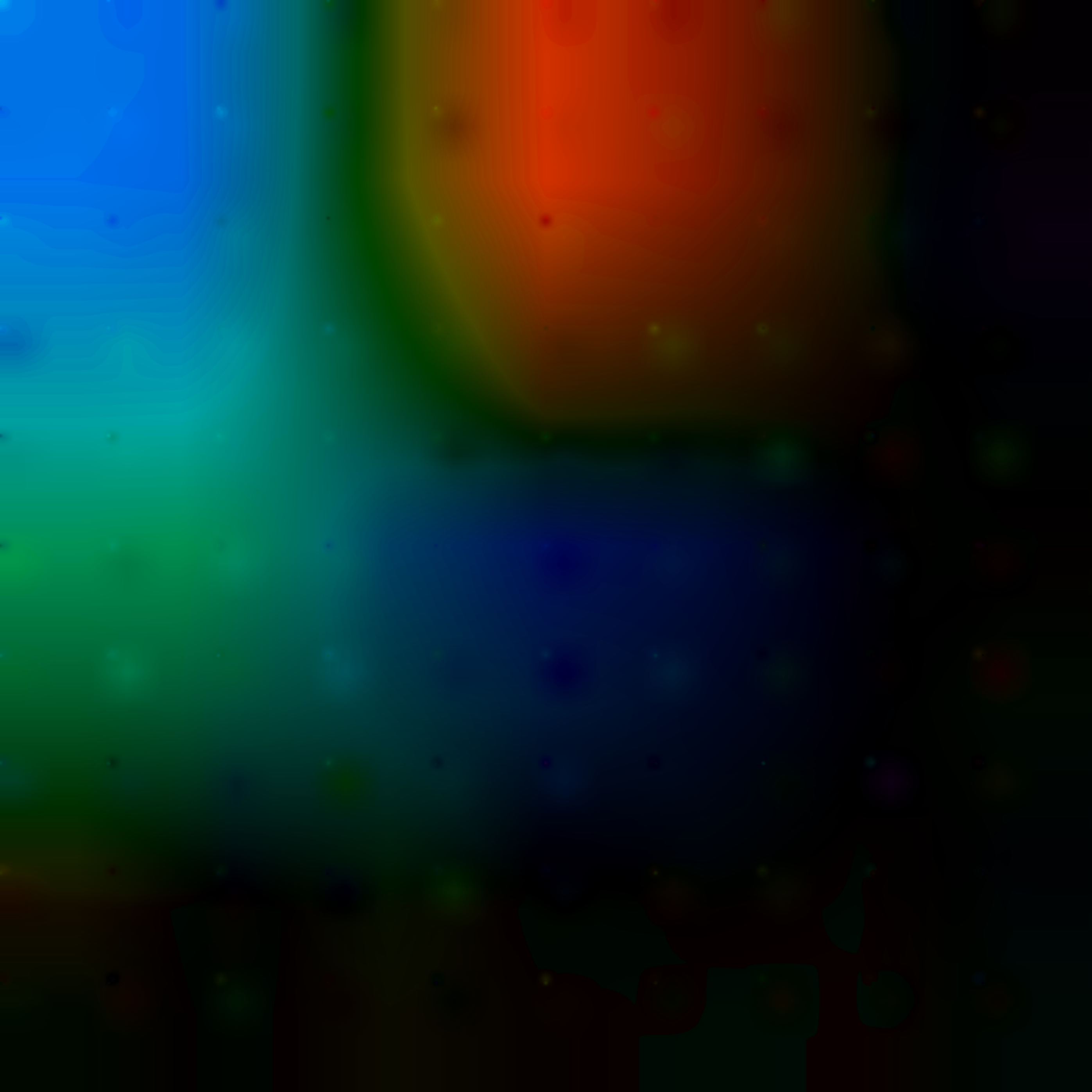}\hfill%
\includegraphics[width=0.124\linewidth]{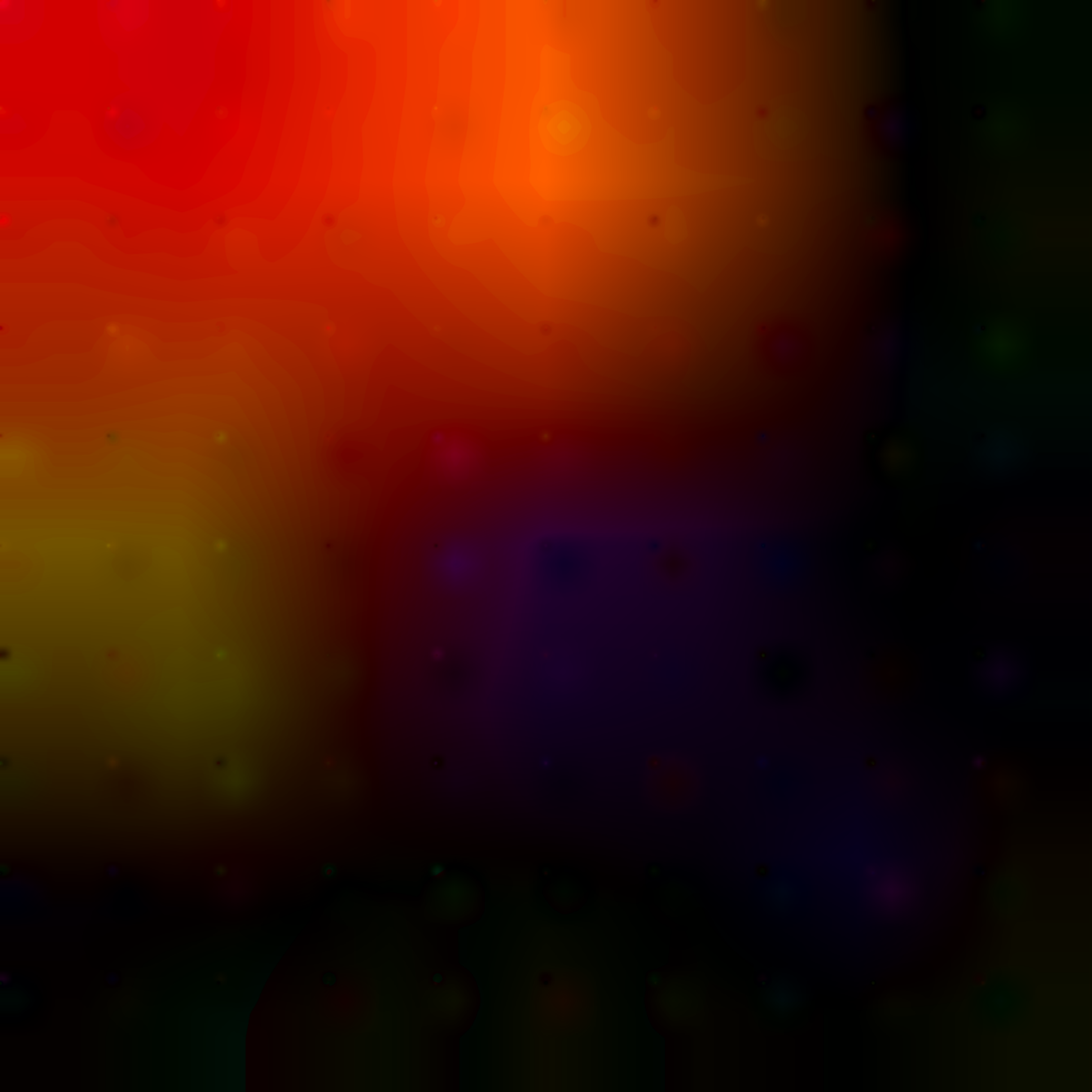}

\includegraphics[width=0.124\linewidth]{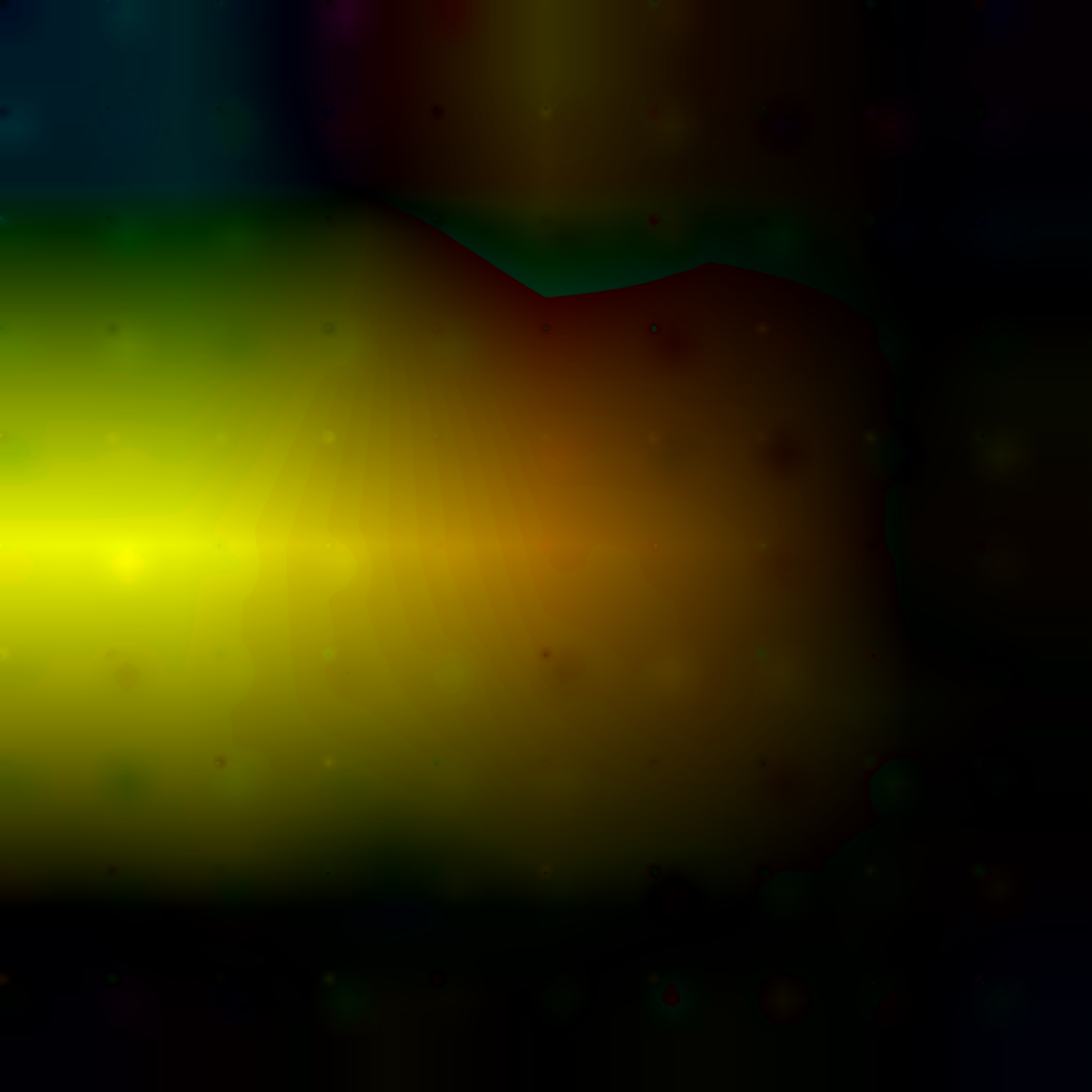}\hfill%
\includegraphics[width=0.124\linewidth]{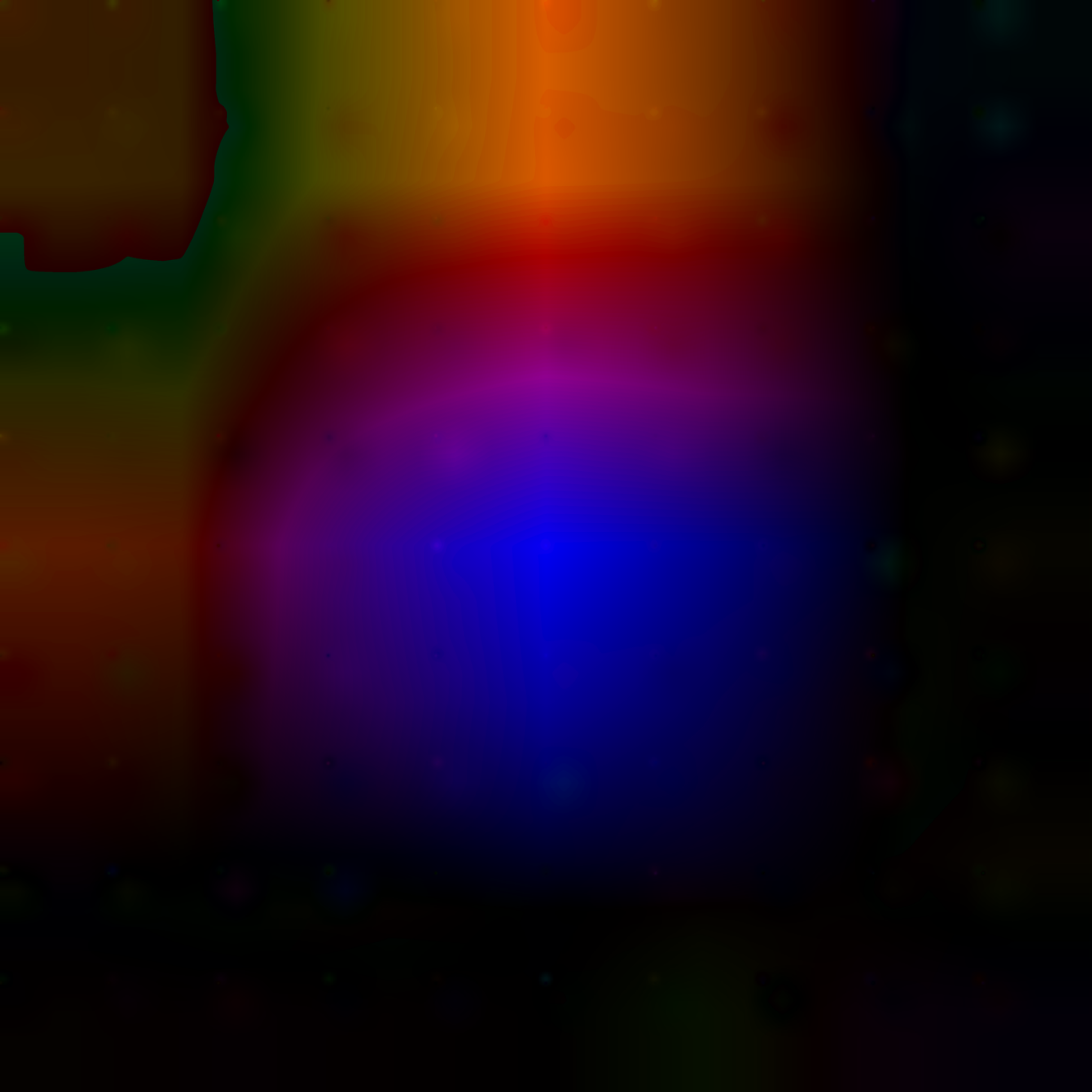}\hfill%
\includegraphics[width=0.124\linewidth]{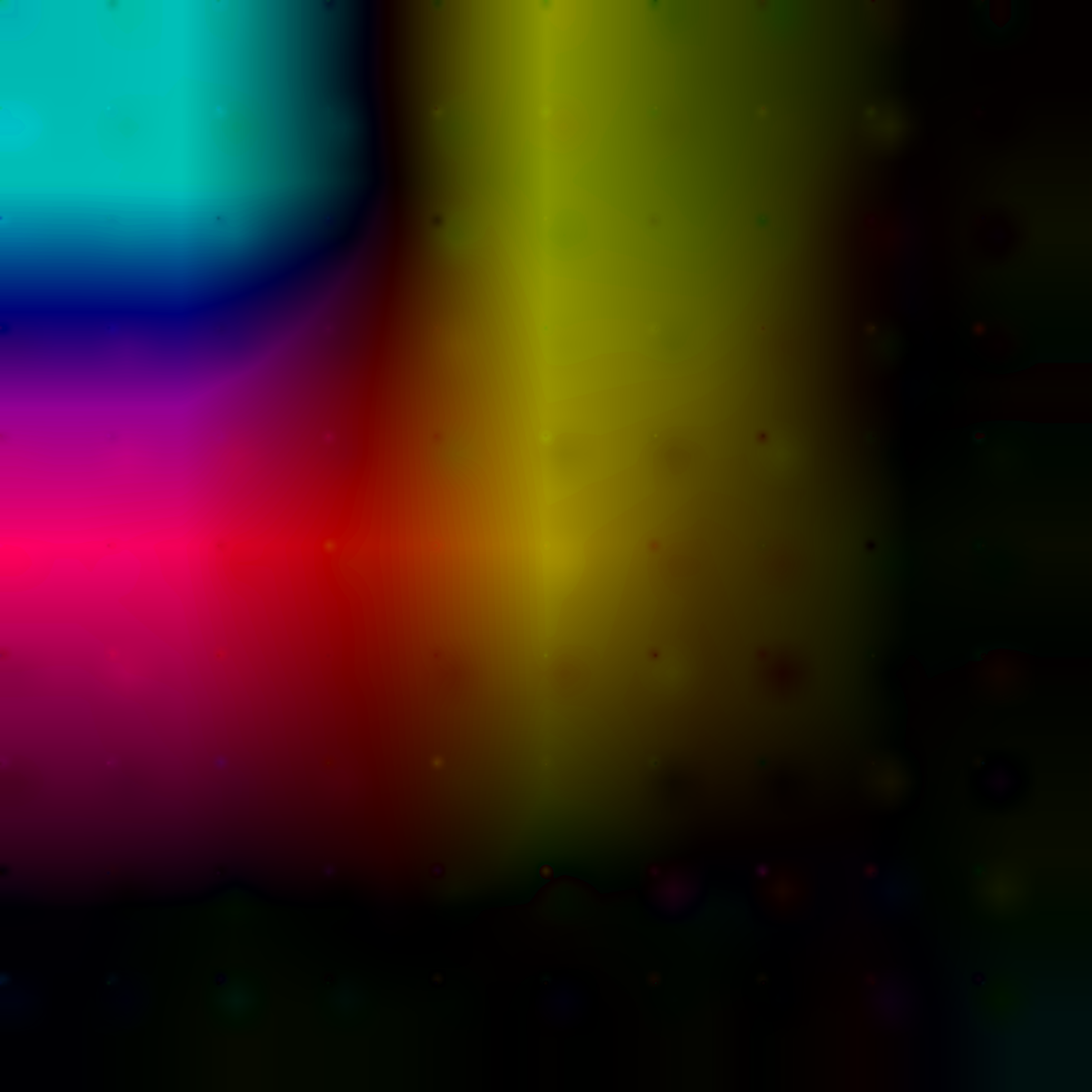}\hfill%
\includegraphics[width=0.124\linewidth]{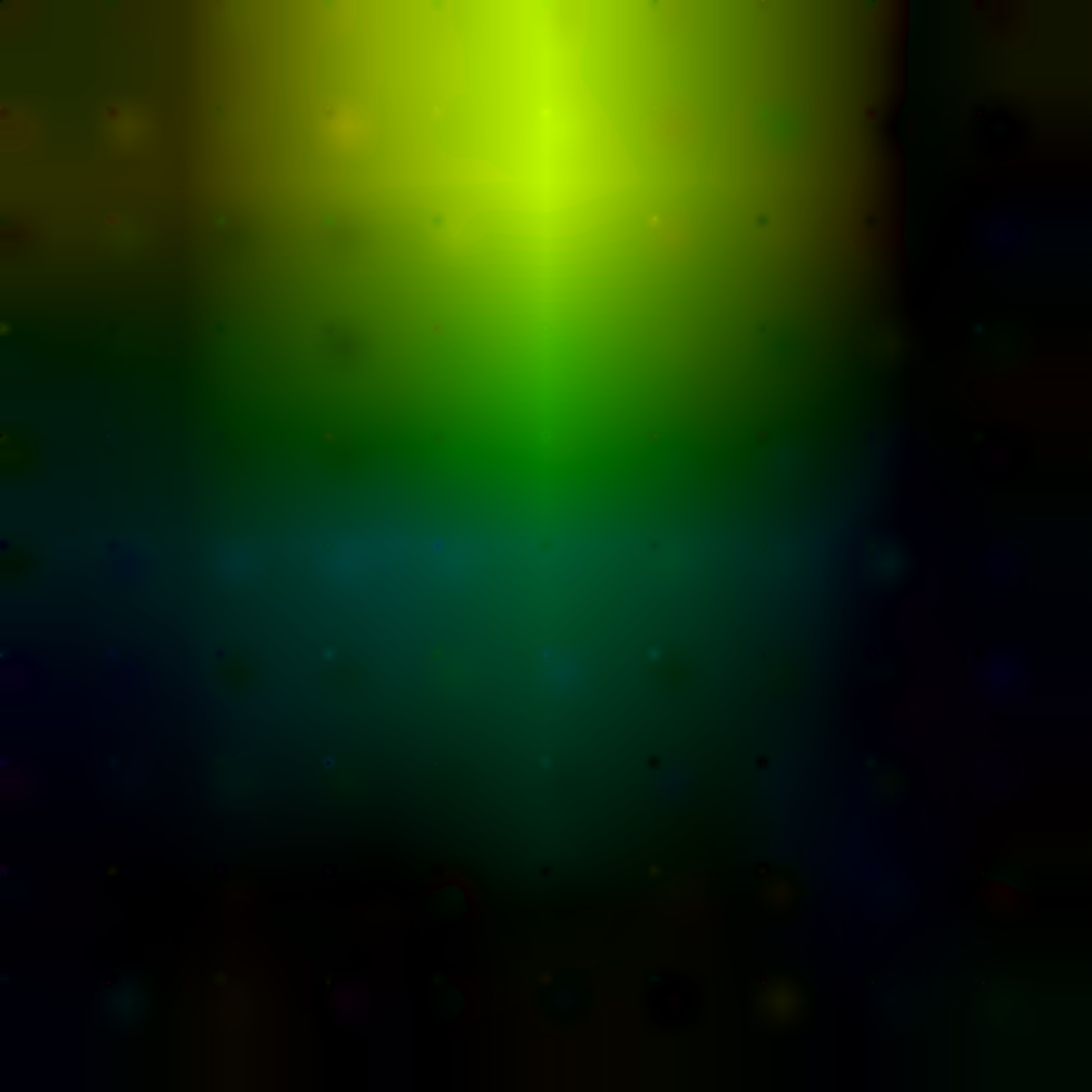}\hfill%
\includegraphics[width=0.124\linewidth]{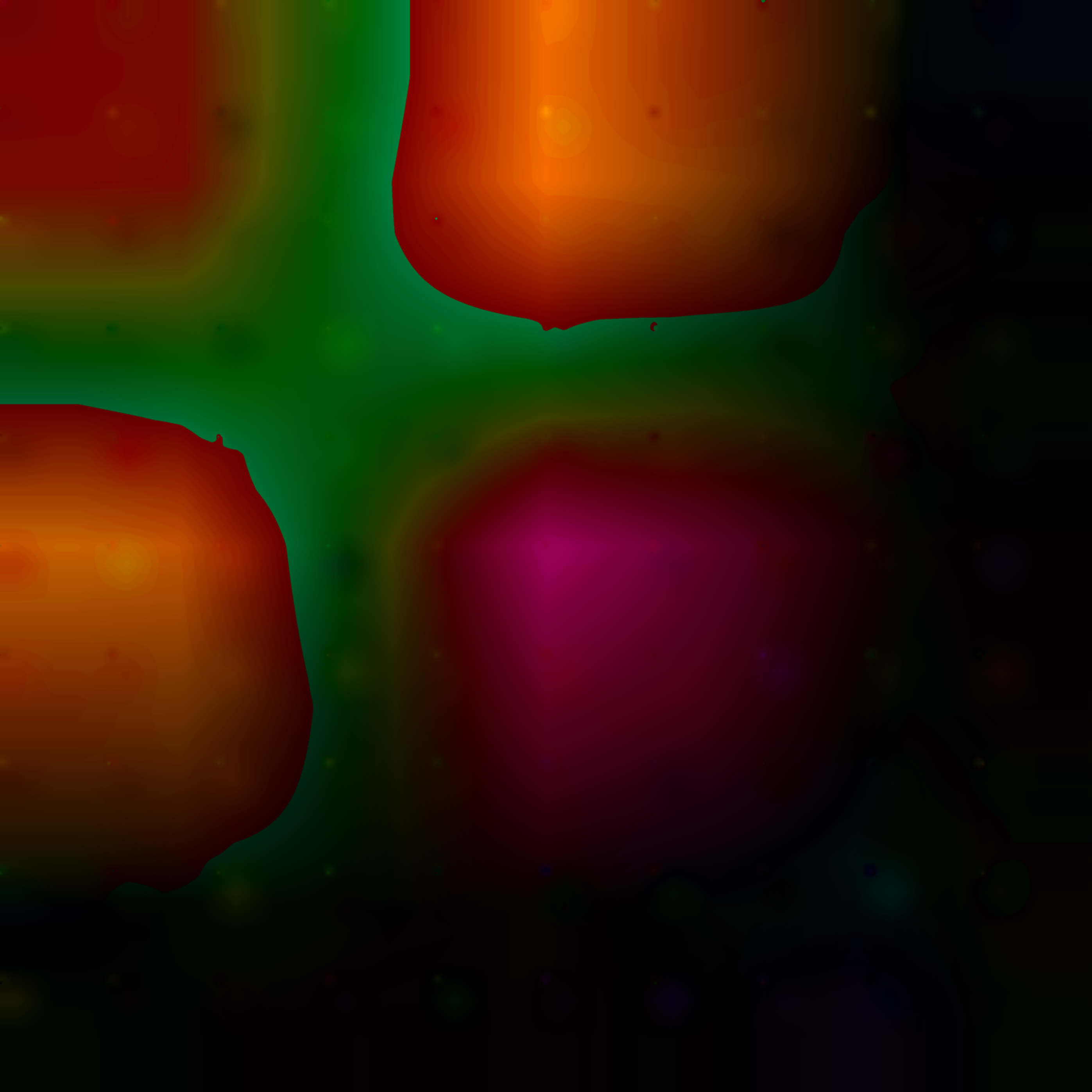}\hfill%
\includegraphics[width=0.124\linewidth]{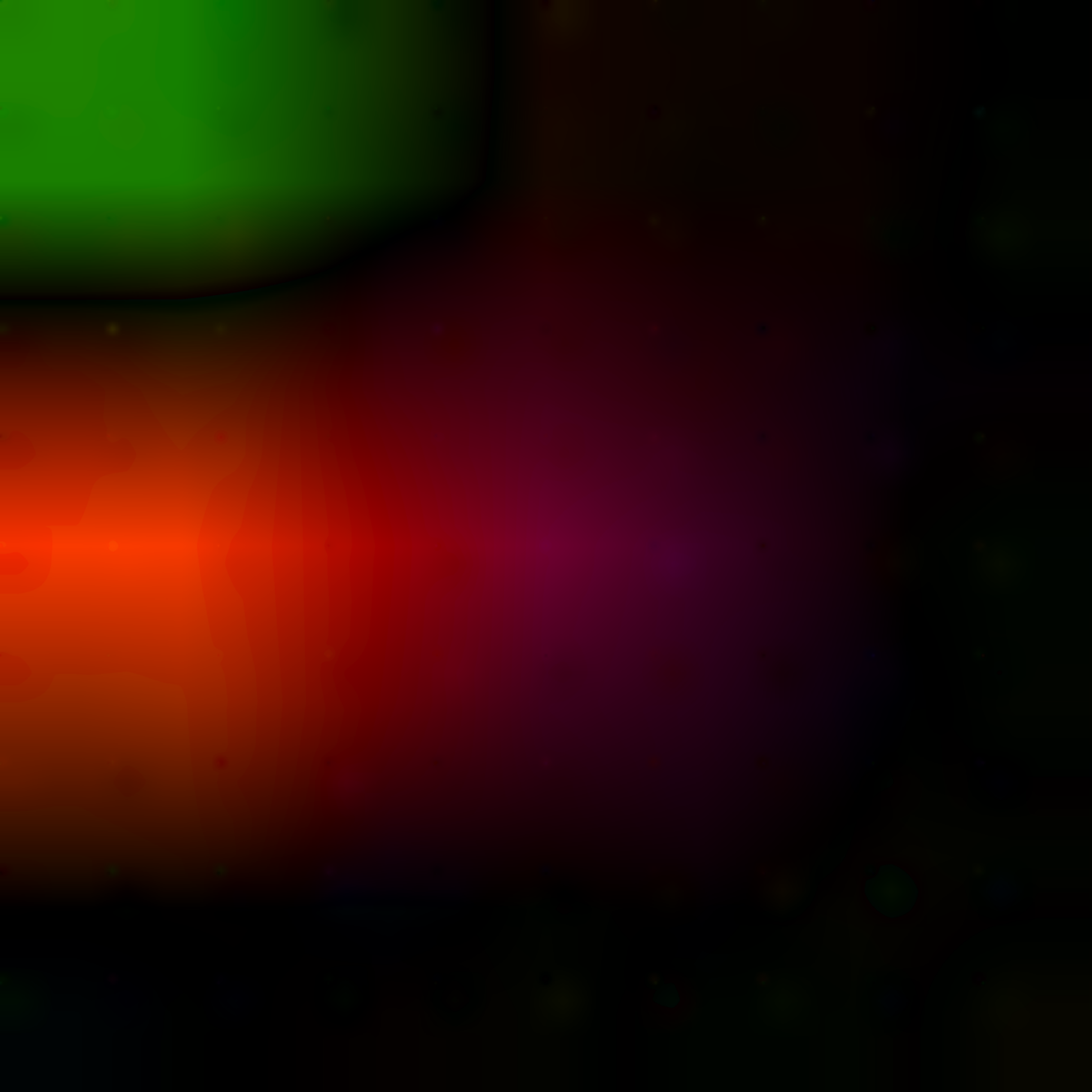}\hfill%
\includegraphics[width=0.124\linewidth]{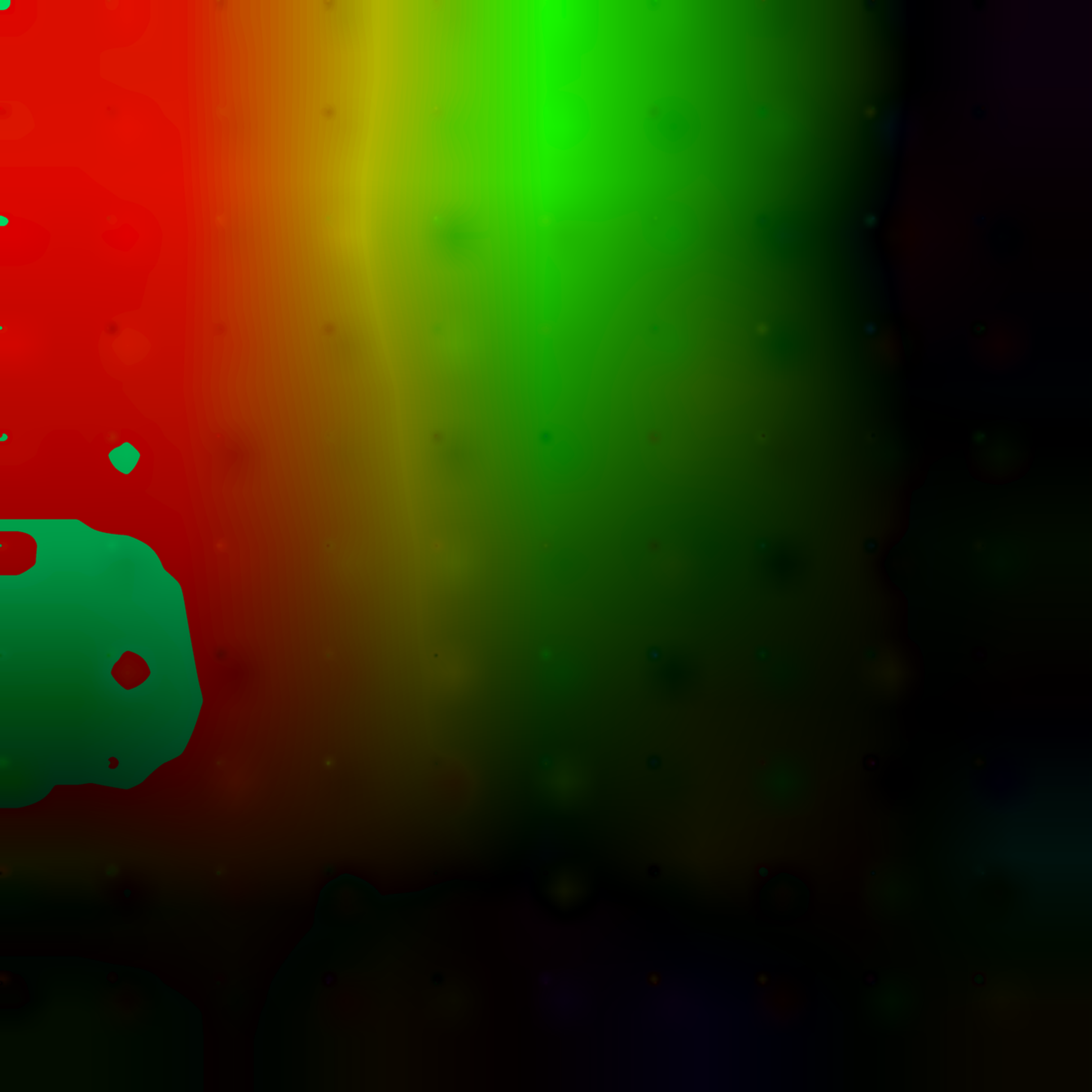}\hfill%
\includegraphics[width=0.124\linewidth]{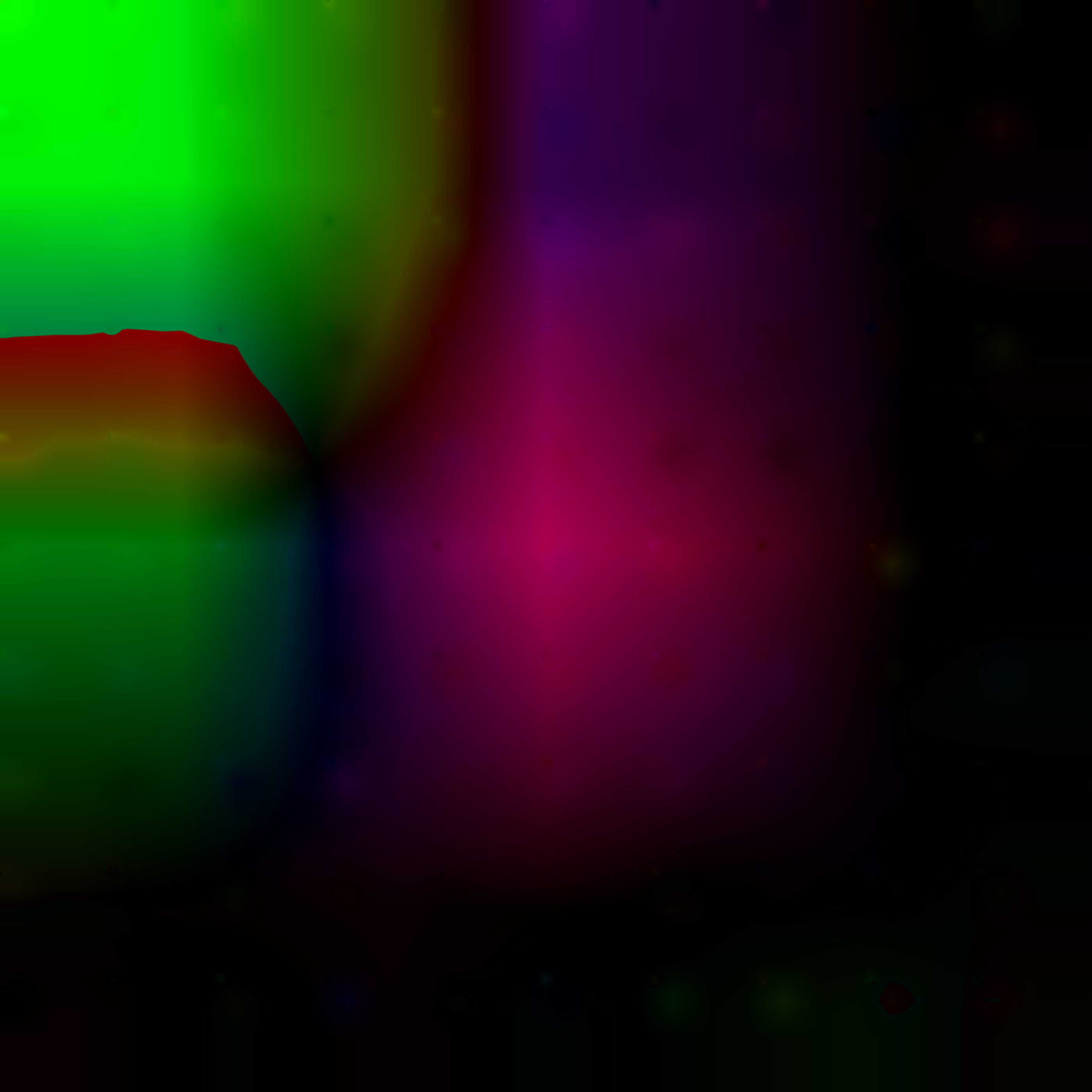}

\includegraphics[width=0.124\linewidth]{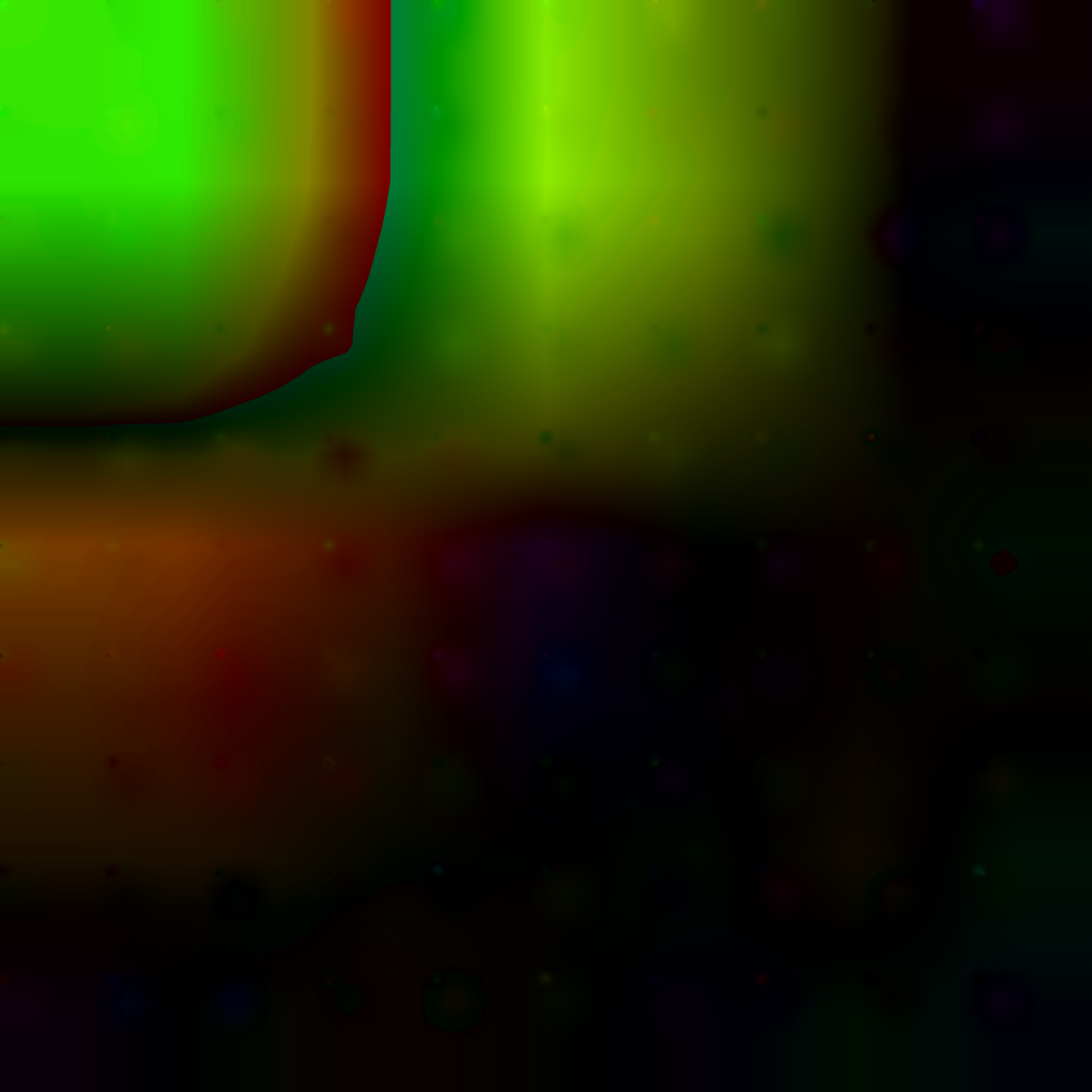}\hfill%
\includegraphics[width=0.124\linewidth]{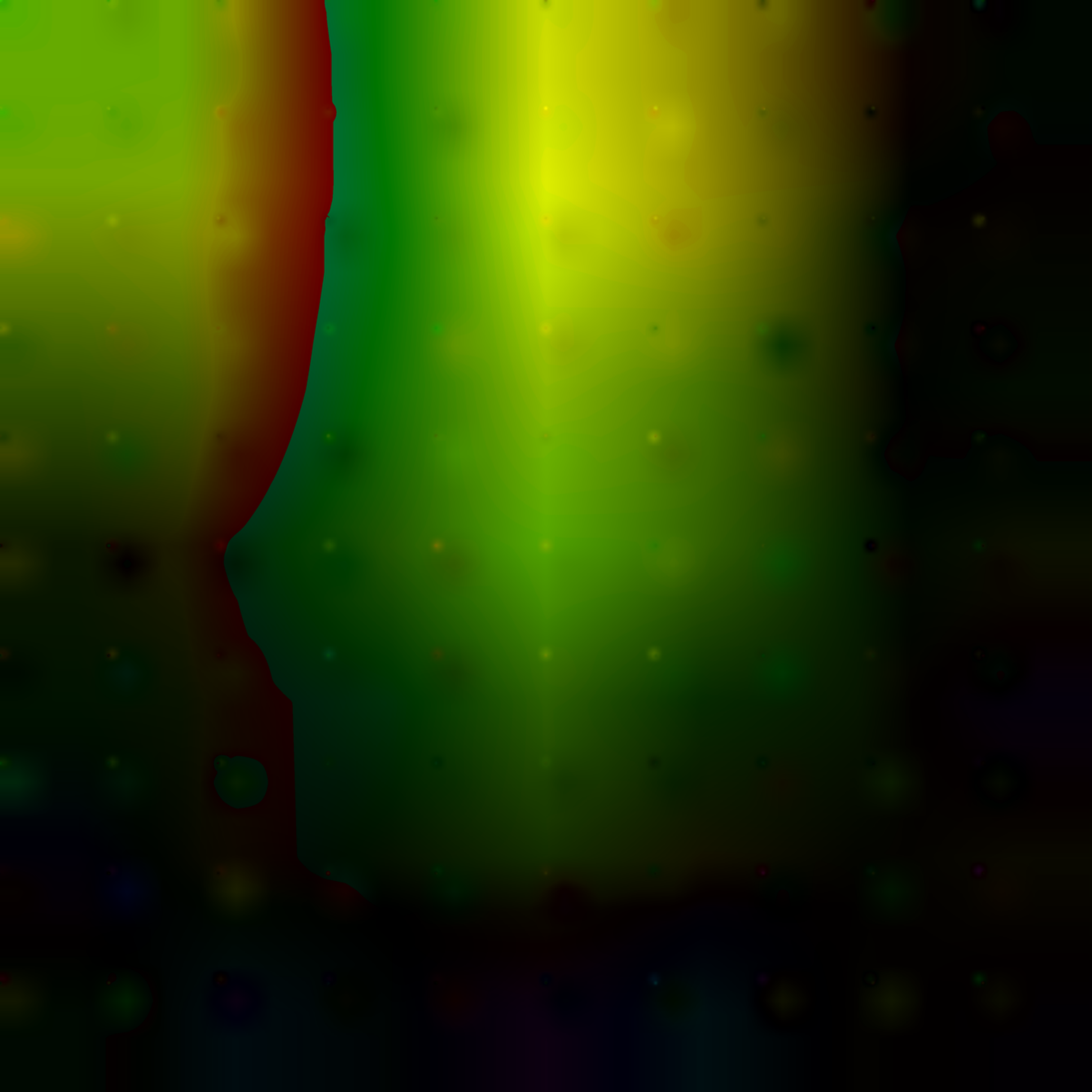}\hfill%
\includegraphics[width=0.124\linewidth]{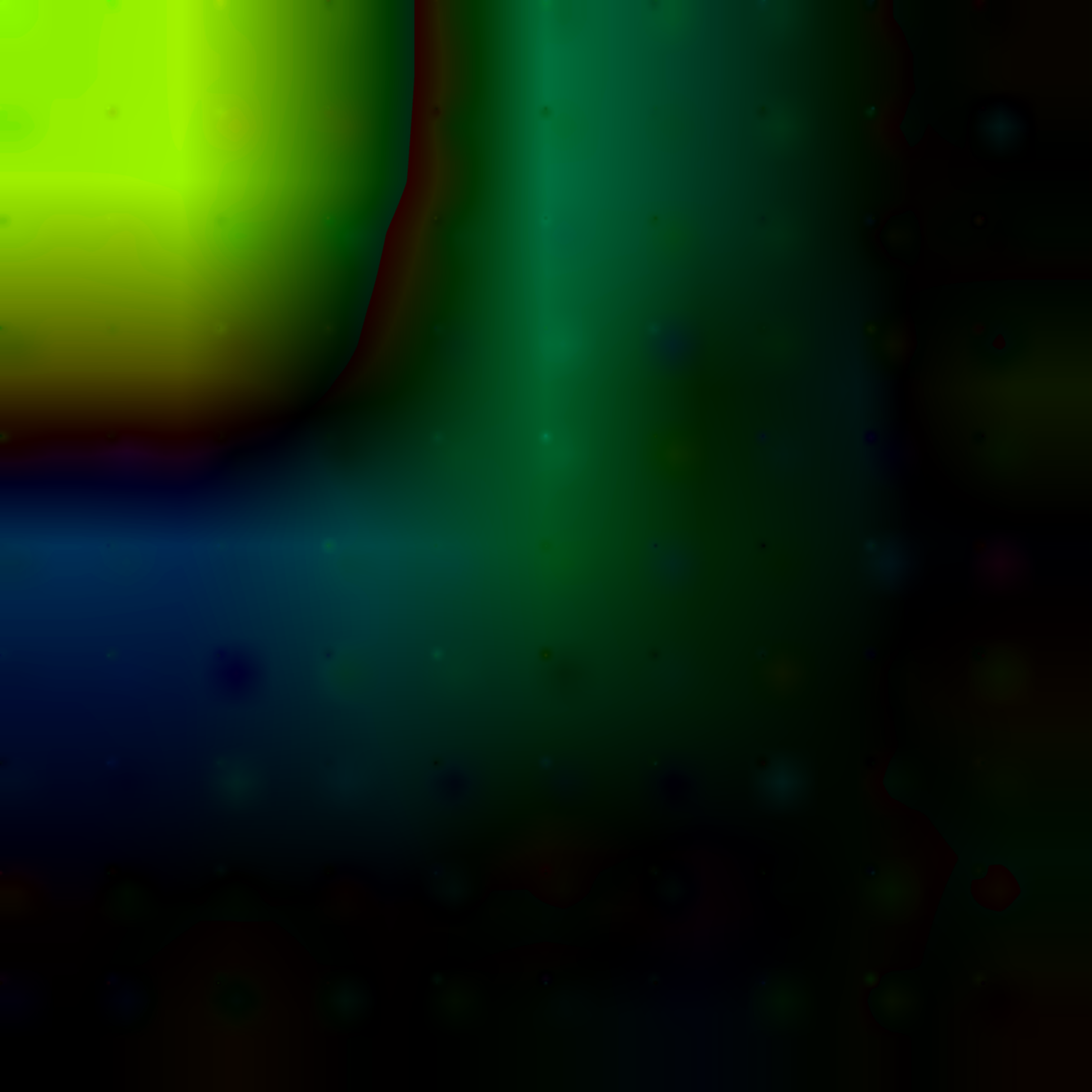}\hfill%
\includegraphics[width=0.124\linewidth]{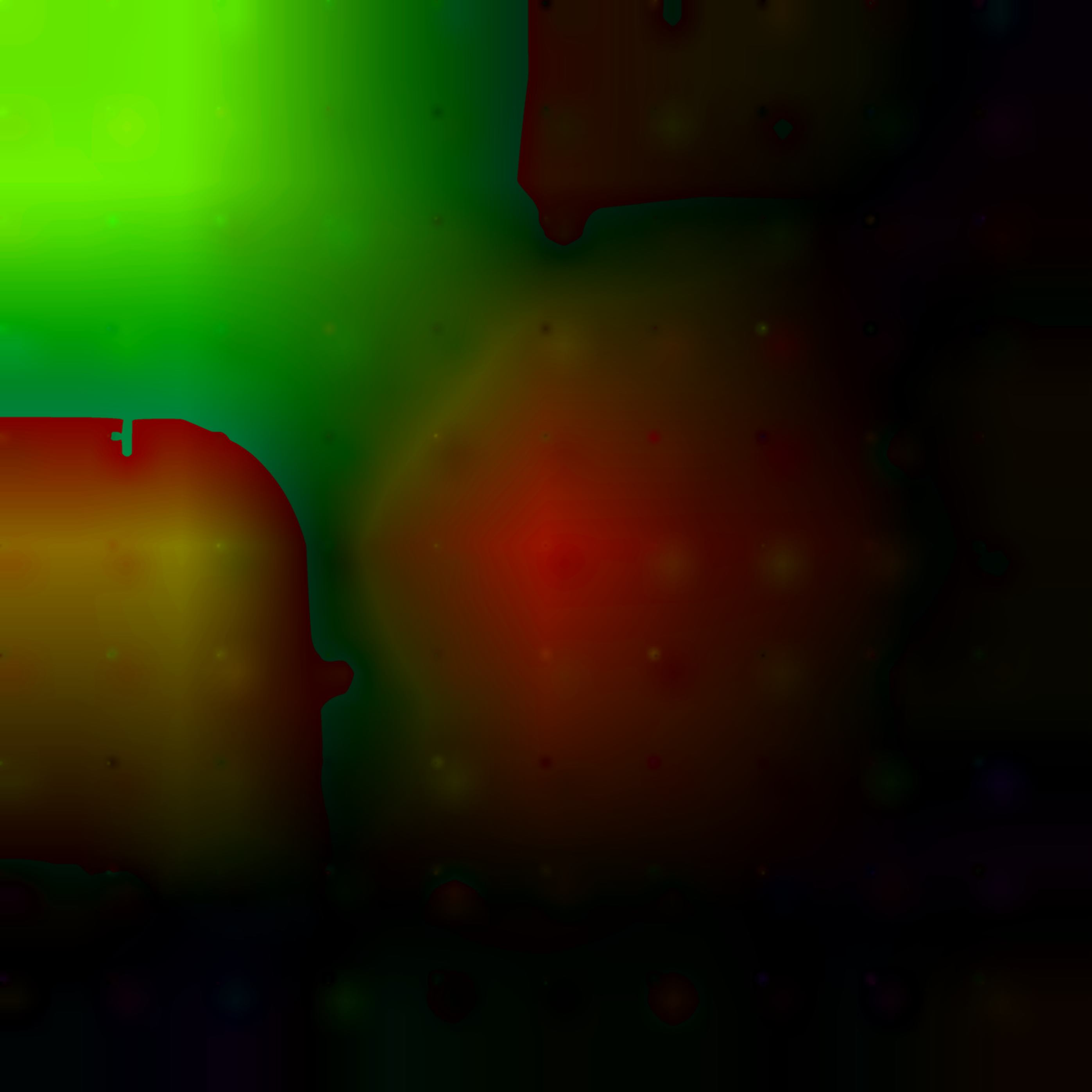}\hfill%
\includegraphics[width=0.124\linewidth]{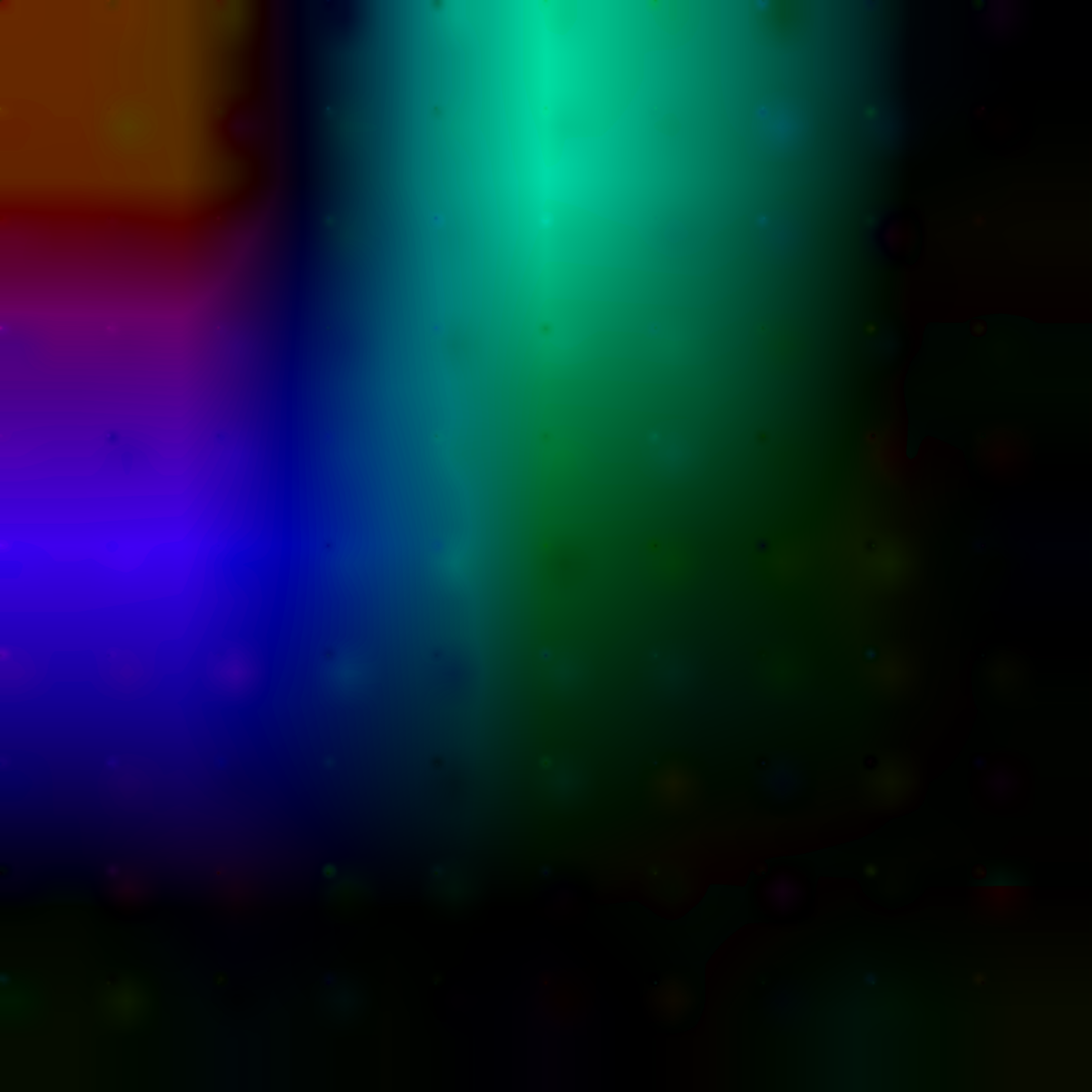}\hfill%
\includegraphics[width=0.124\linewidth]{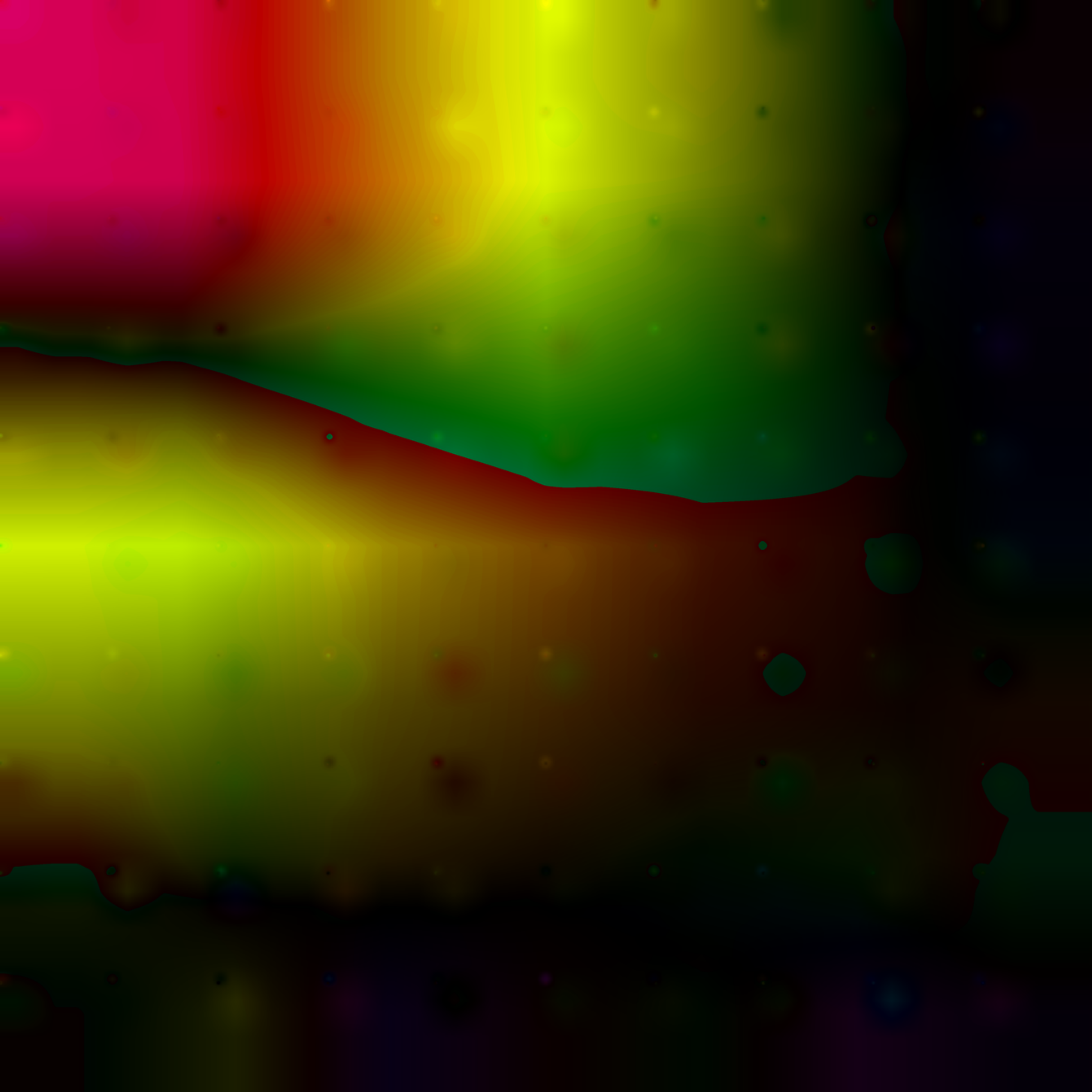}\hfill%
\includegraphics[width=0.124\linewidth]{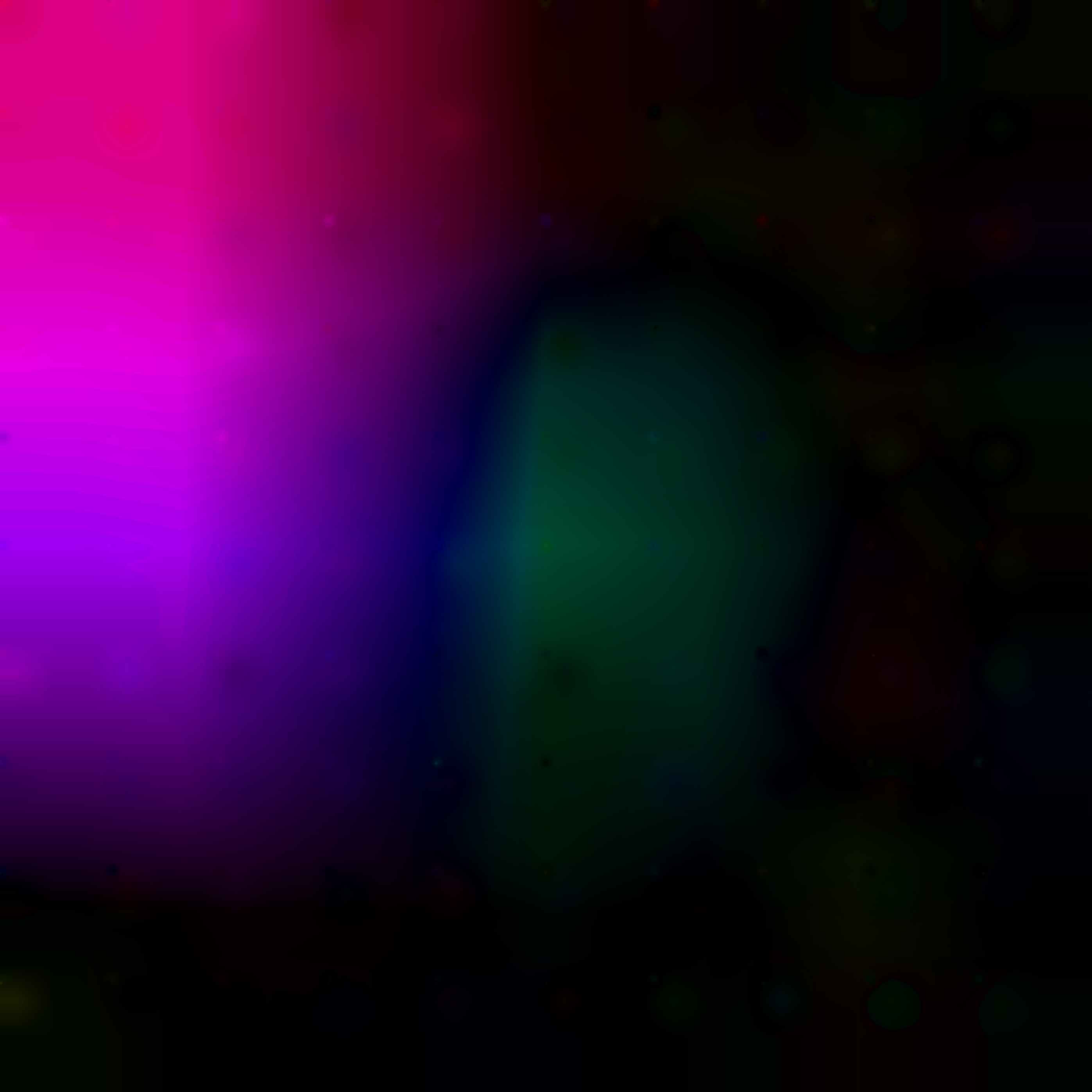}\hfill%
\includegraphics[width=0.124\linewidth]{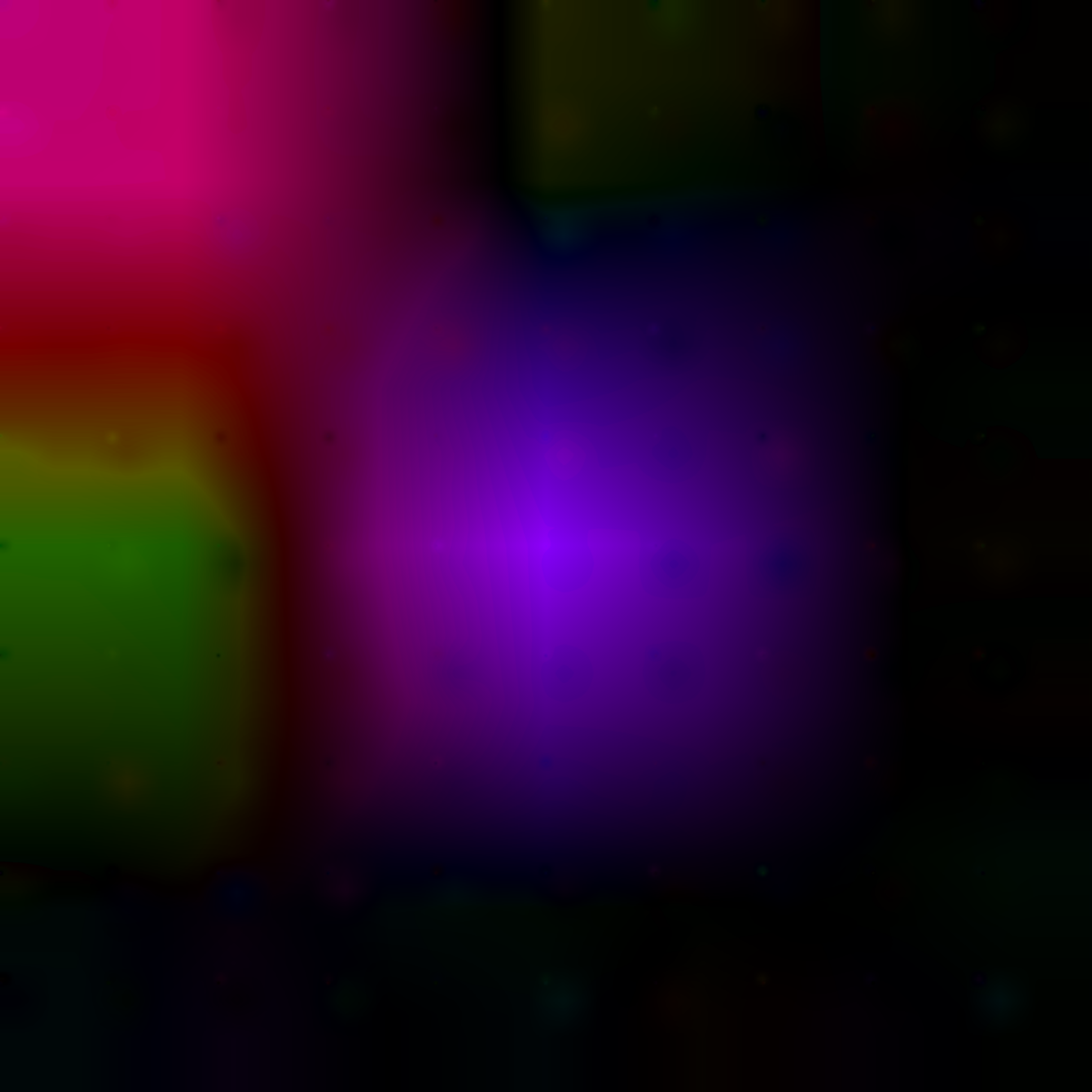}

\includegraphics[width=0.124\linewidth]{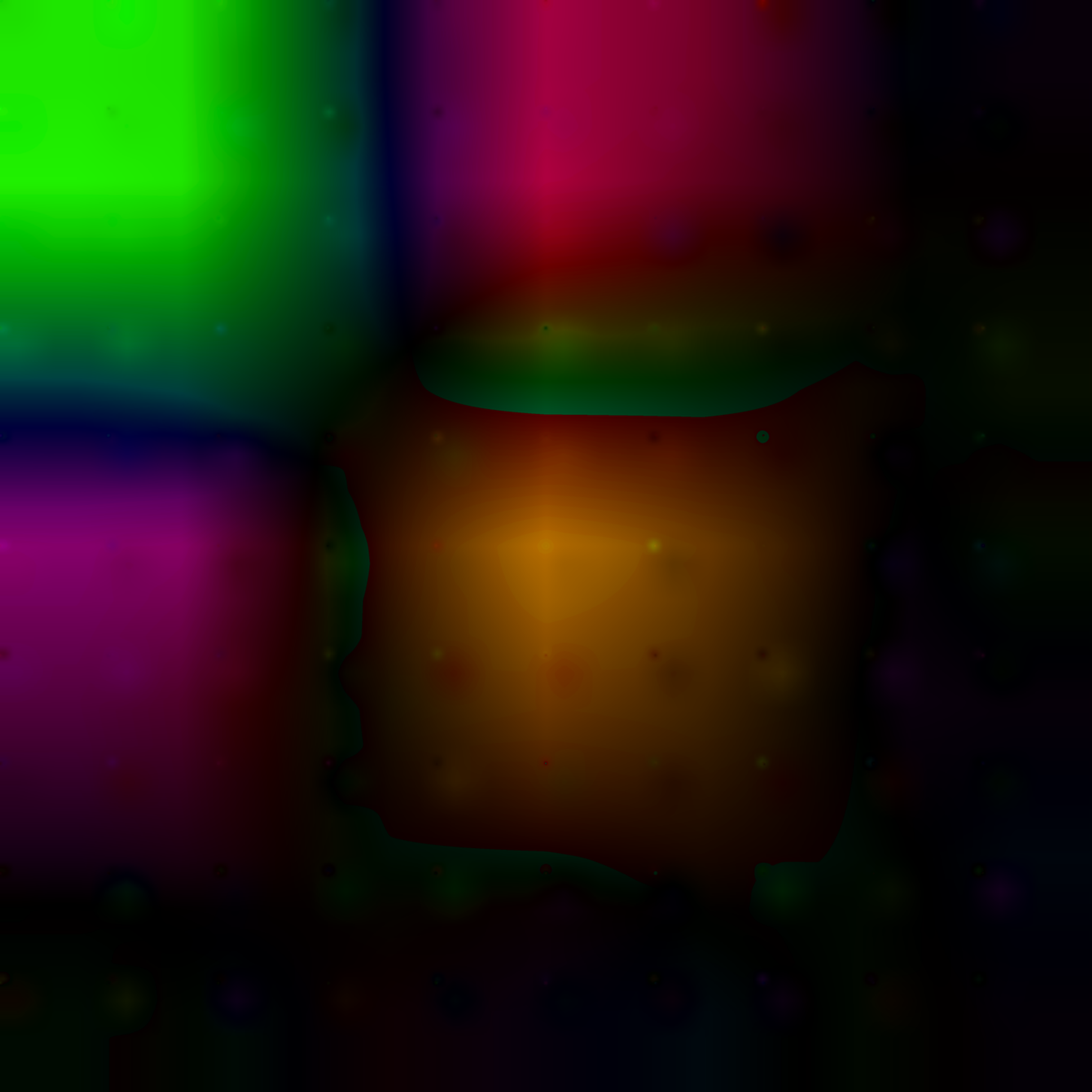}\hfill%
\includegraphics[width=0.124\linewidth]{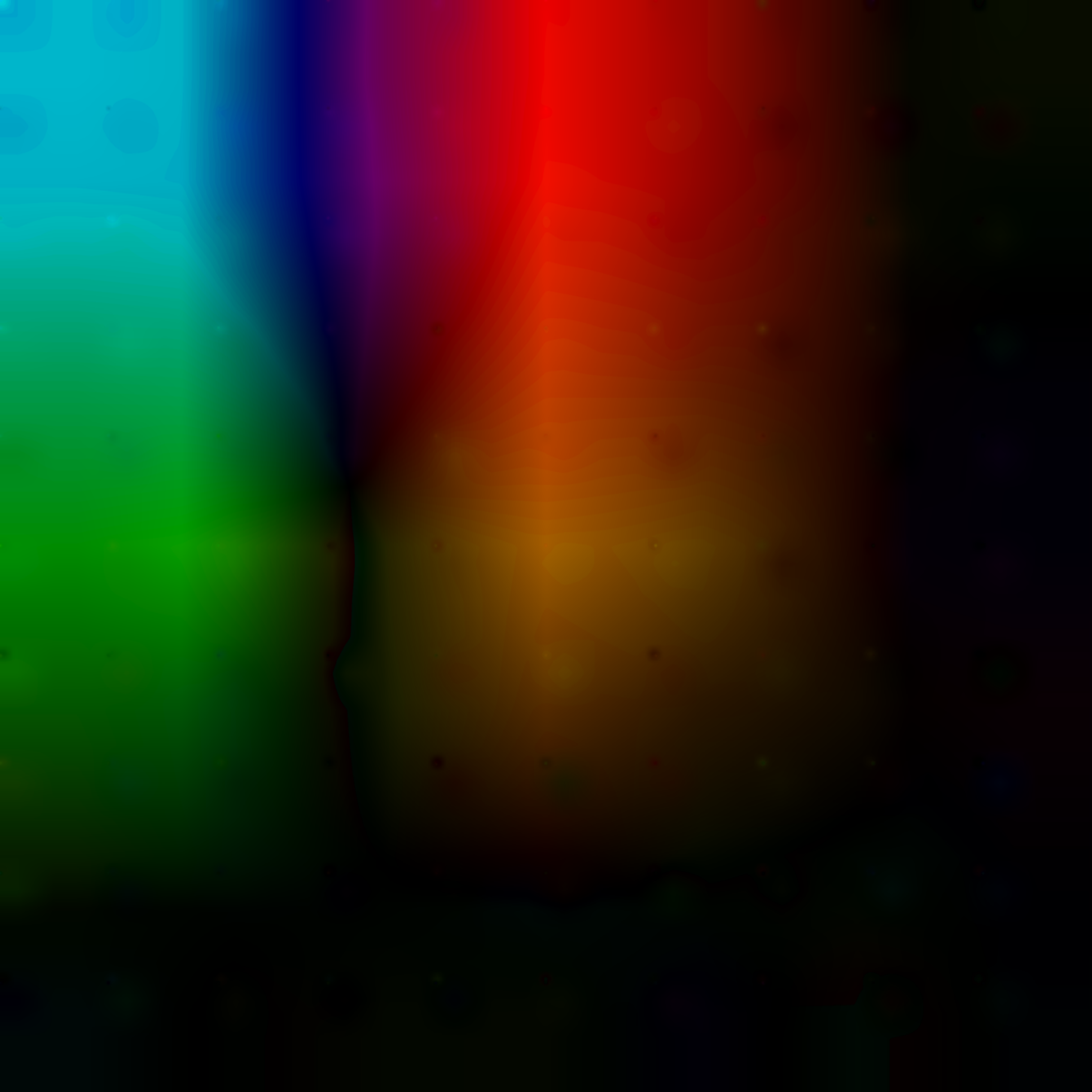}\hfill%
\includegraphics[width=0.124\linewidth]{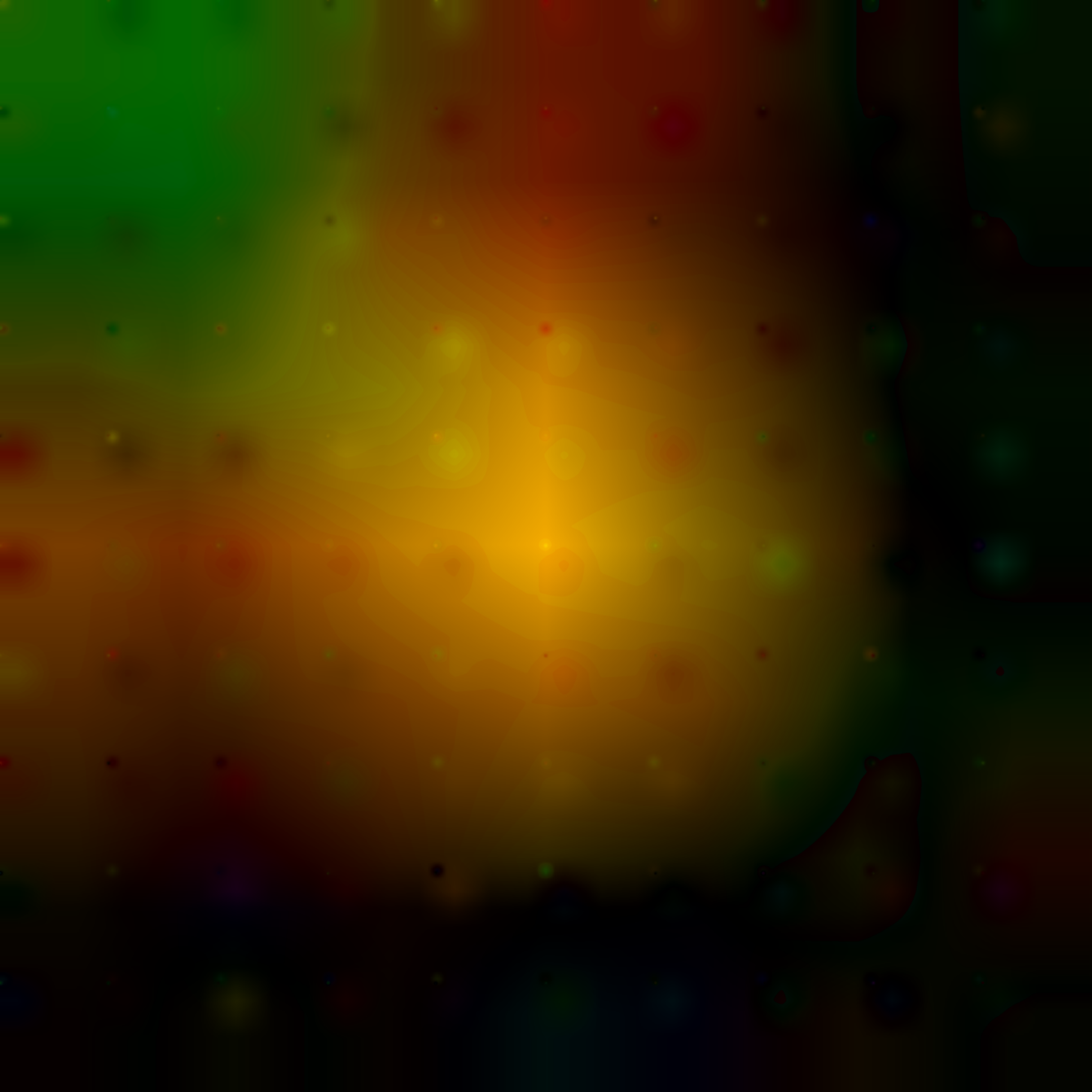}\hfill%
\includegraphics[width=0.124\linewidth]{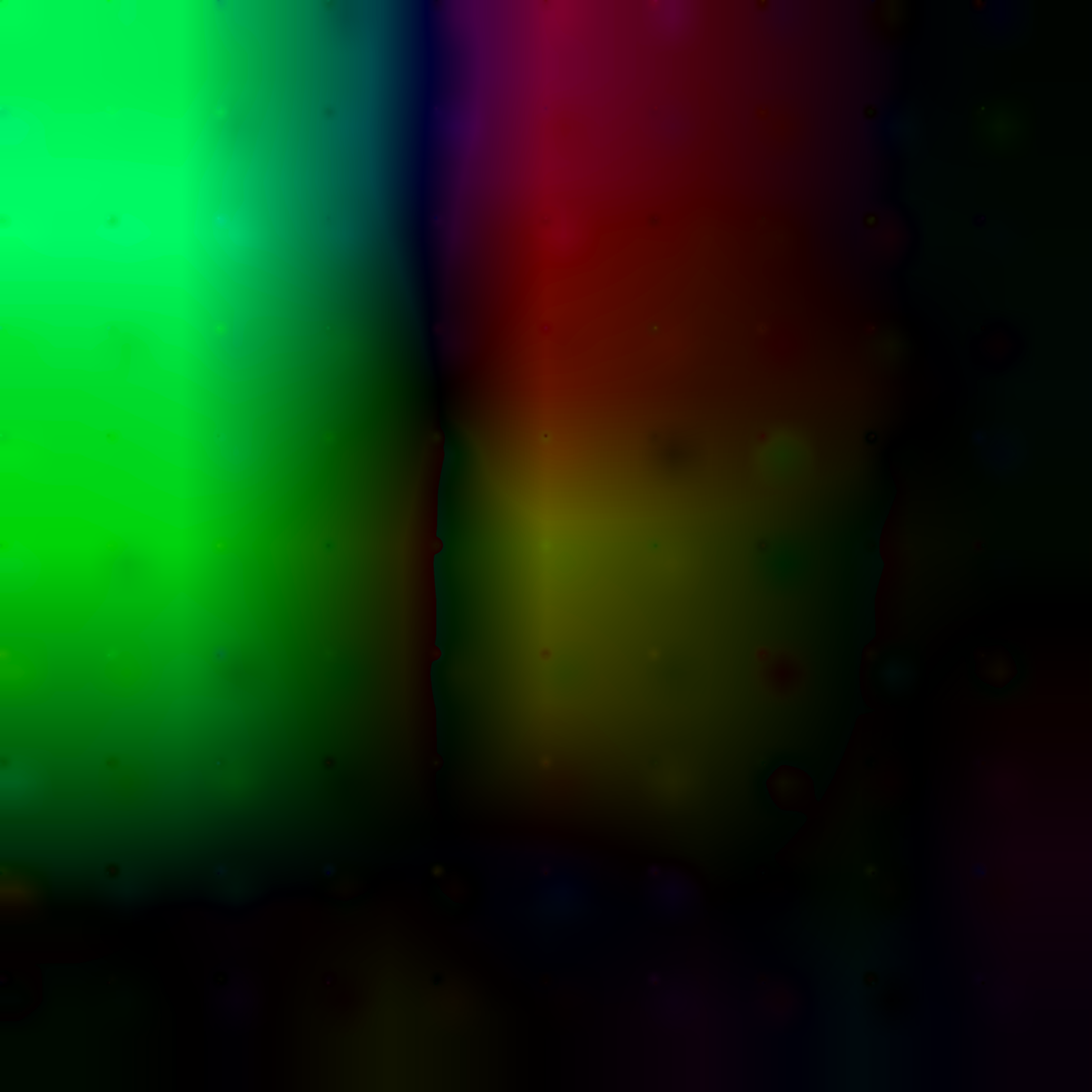}\hfill%
\includegraphics[width=0.124\linewidth]{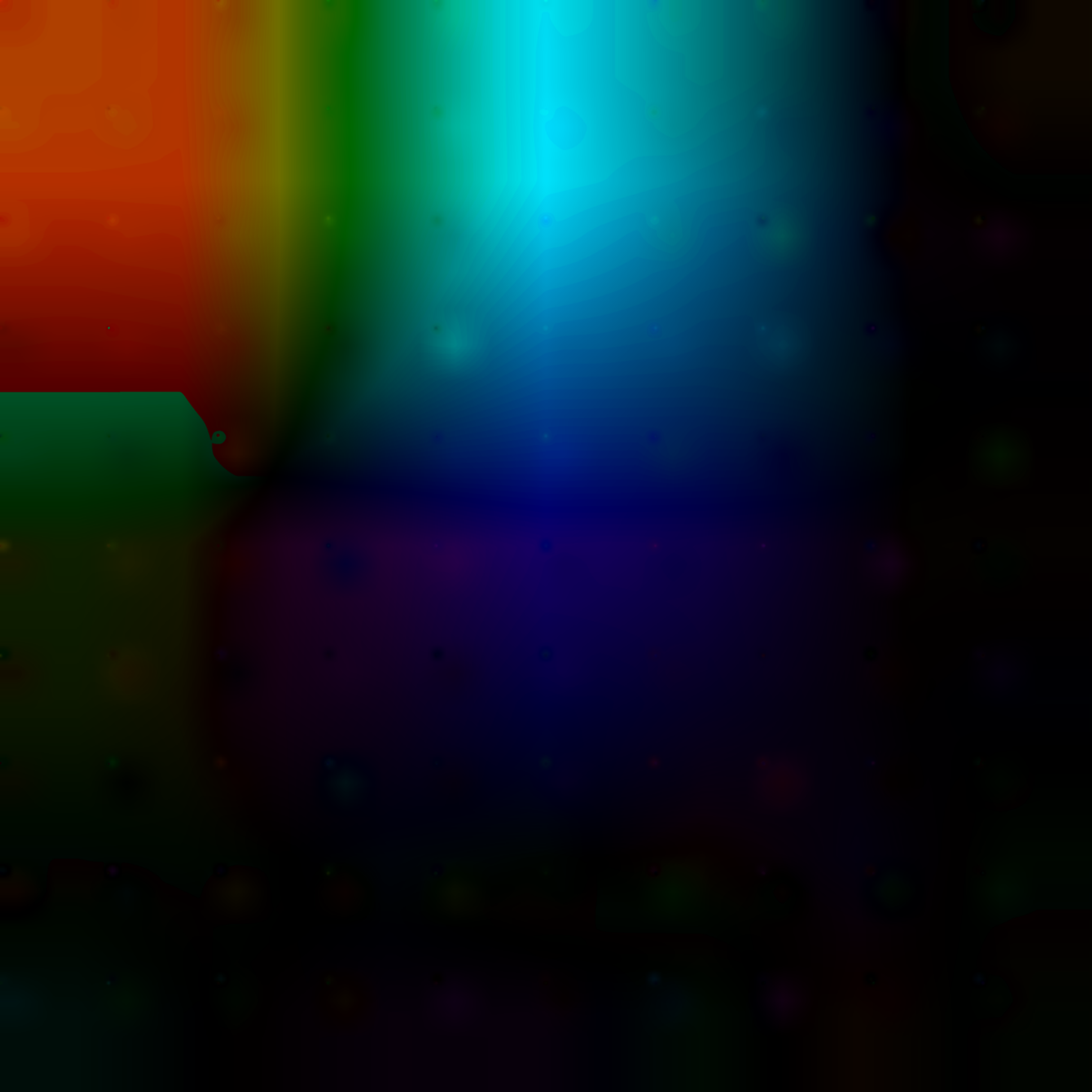}\hfill%
\includegraphics[width=0.124\linewidth]{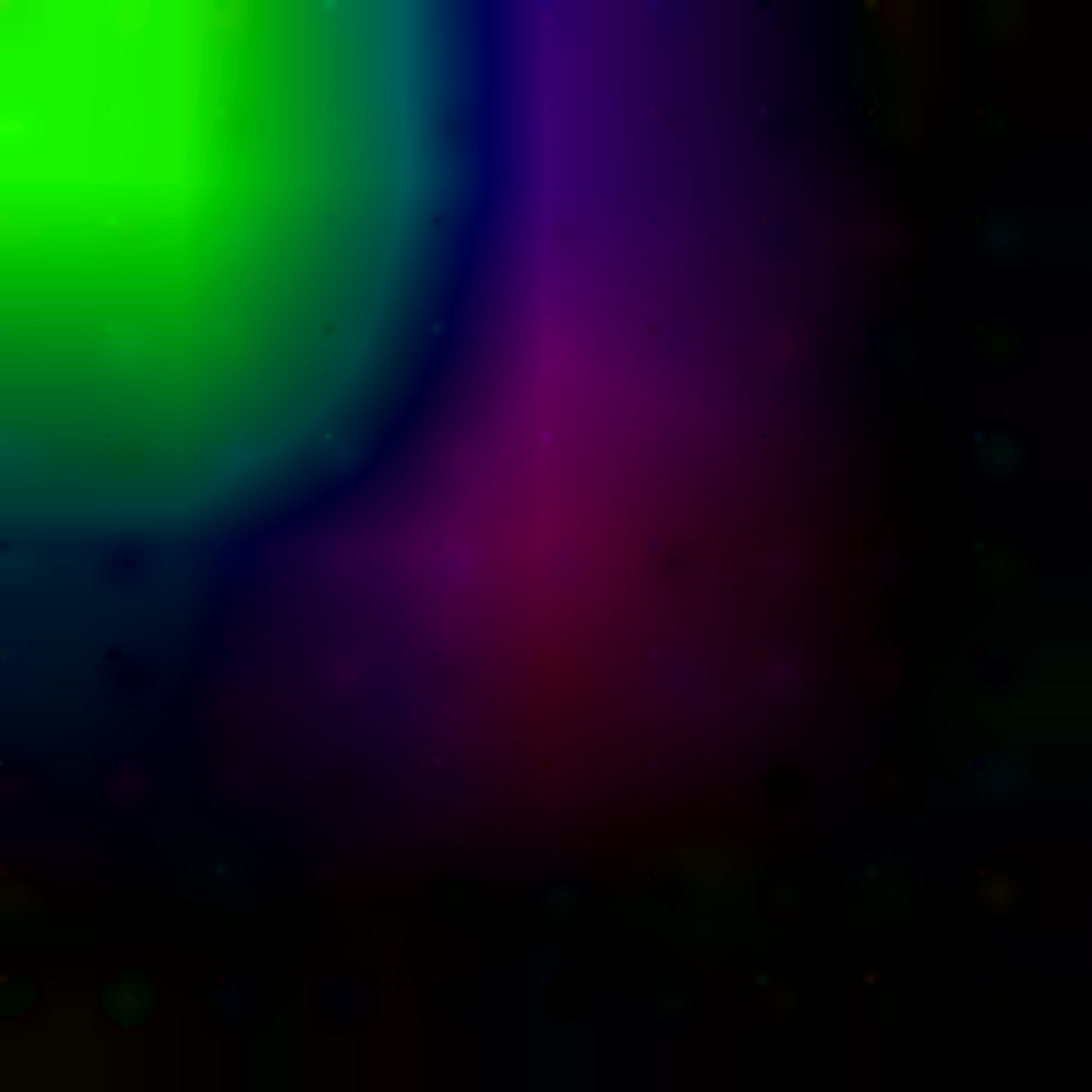}\hfill%
\includegraphics[width=0.124\linewidth]{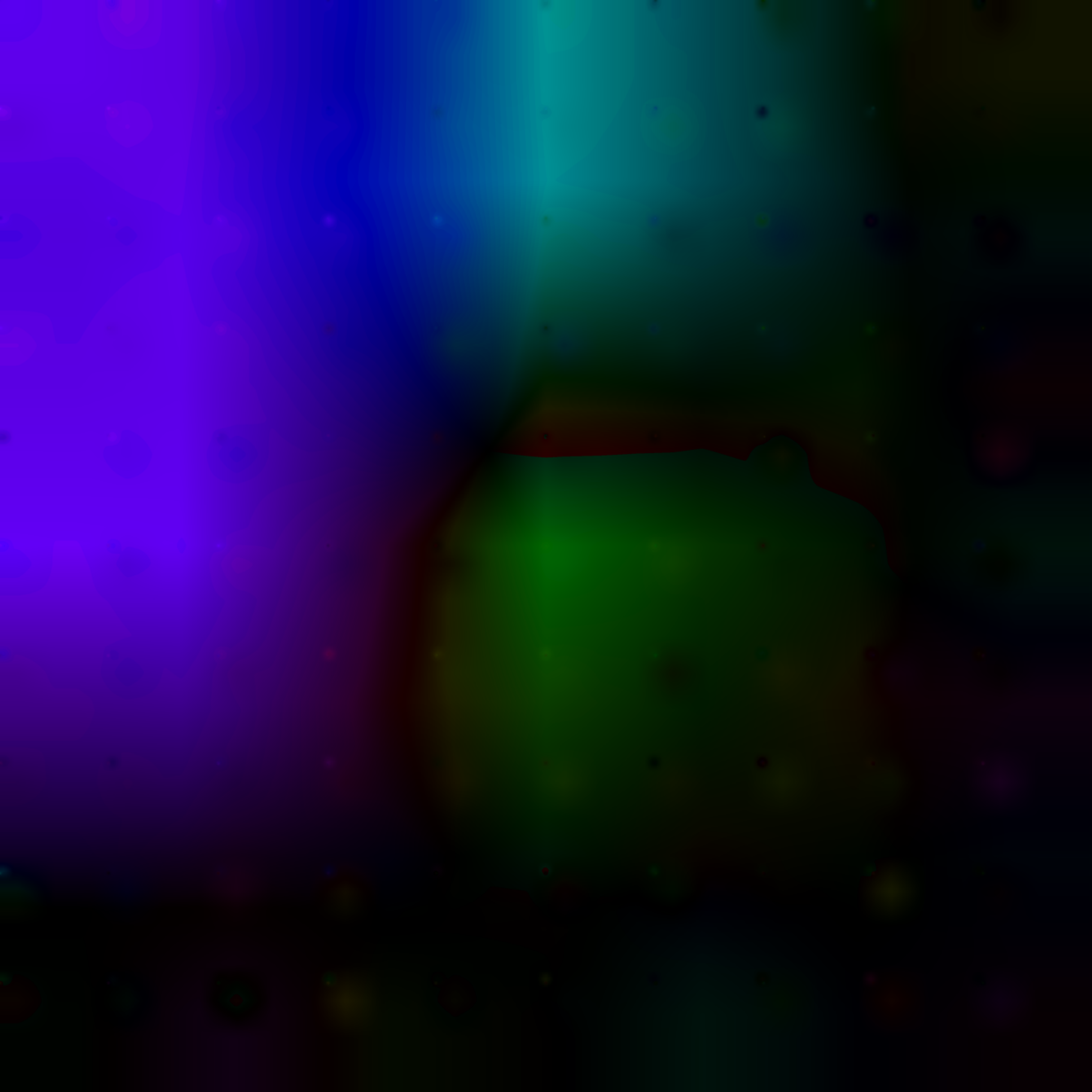}\hfill%
\includegraphics[width=0.124\linewidth]{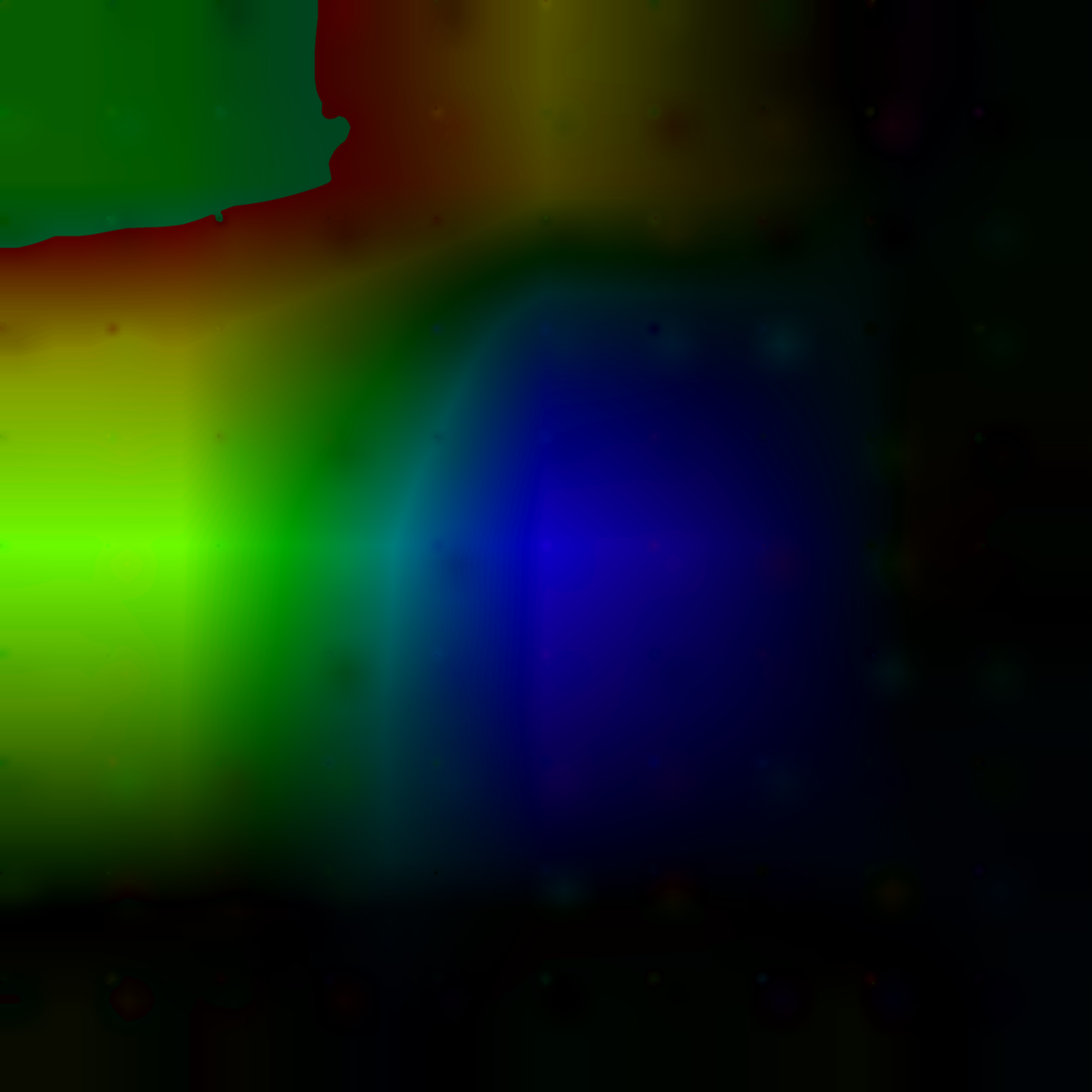}

\includegraphics[width=0.124\linewidth]{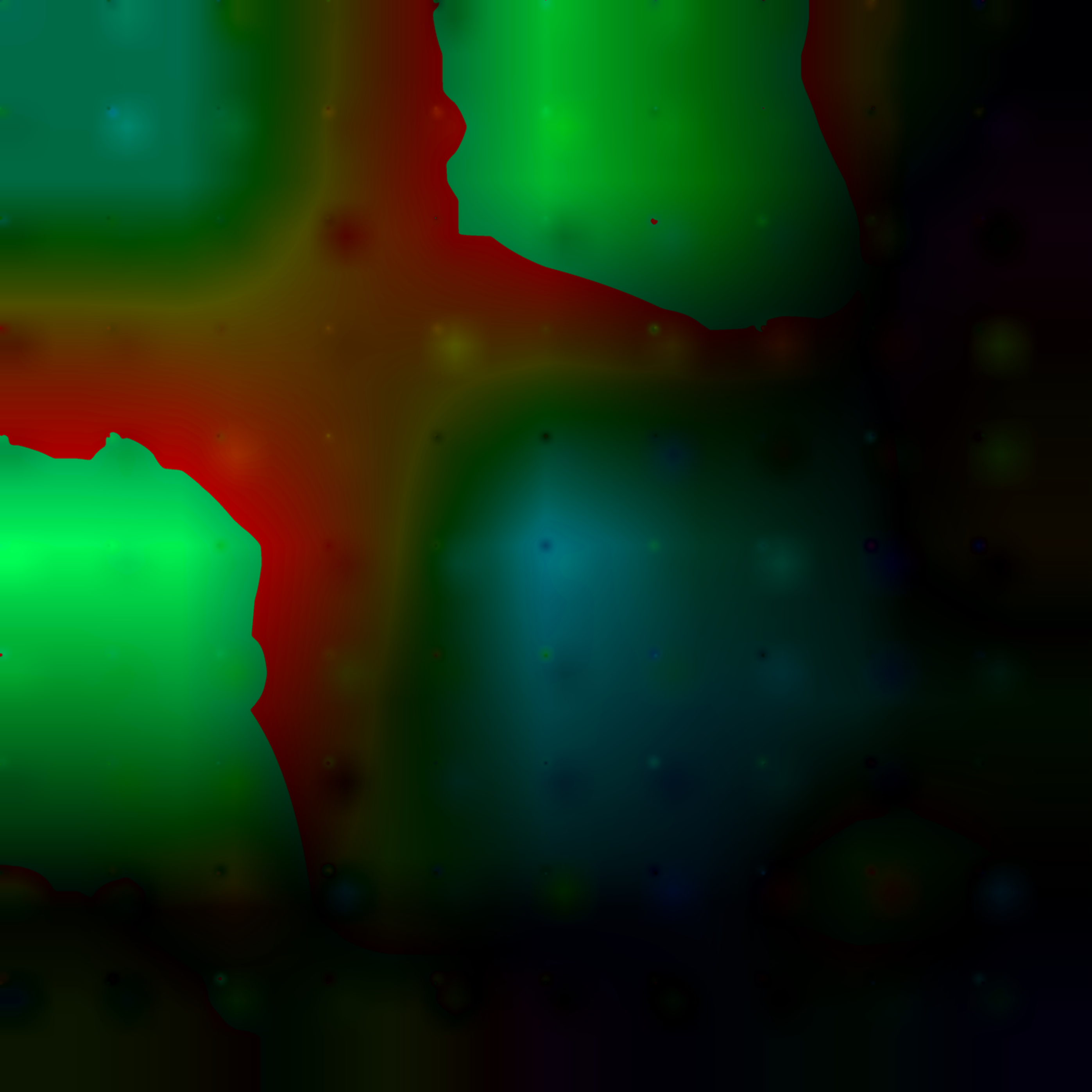}\hfill%
\includegraphics[width=0.124\linewidth]{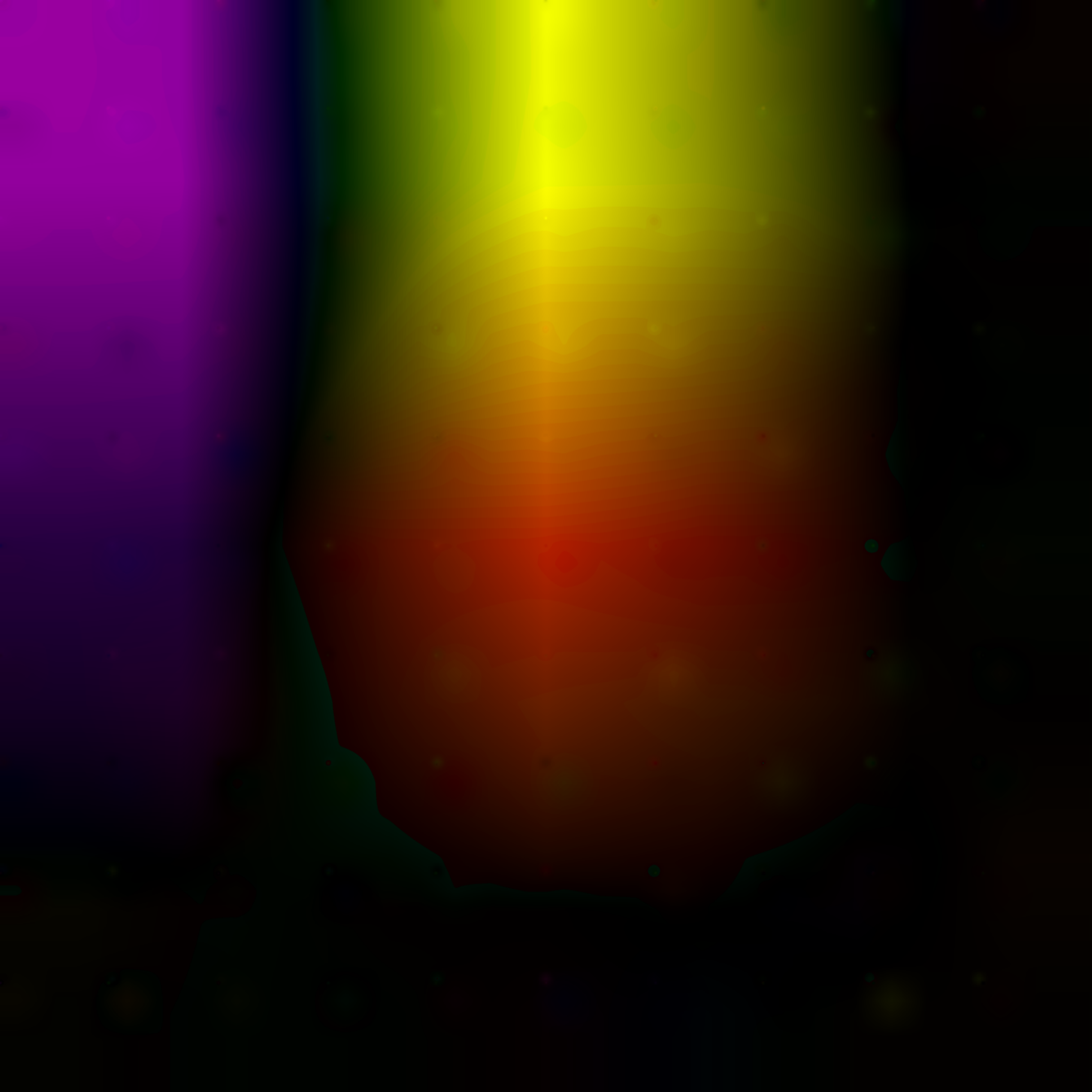}\hfill%
\includegraphics[width=0.124\linewidth]{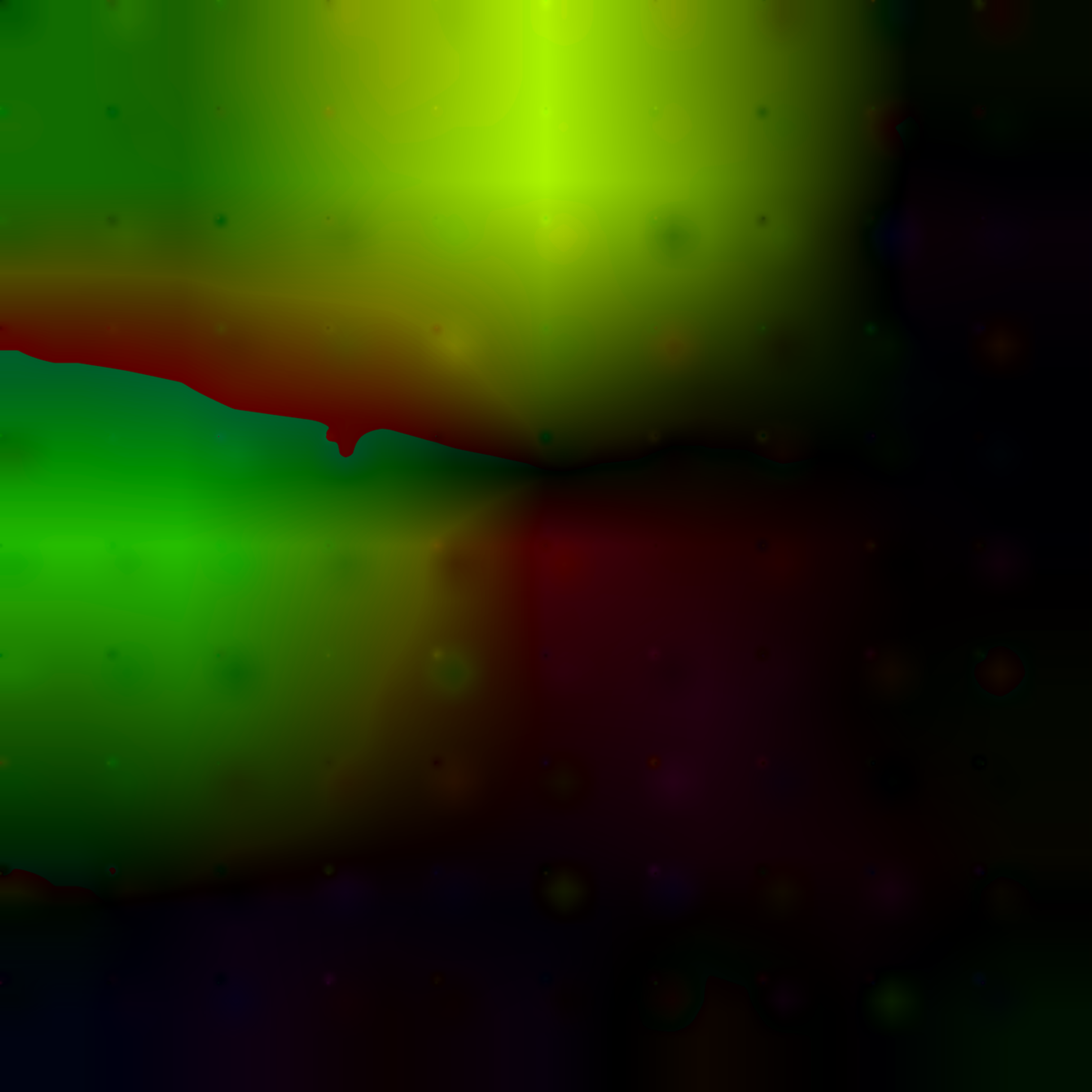}\hfill%
\includegraphics[width=0.124\linewidth]{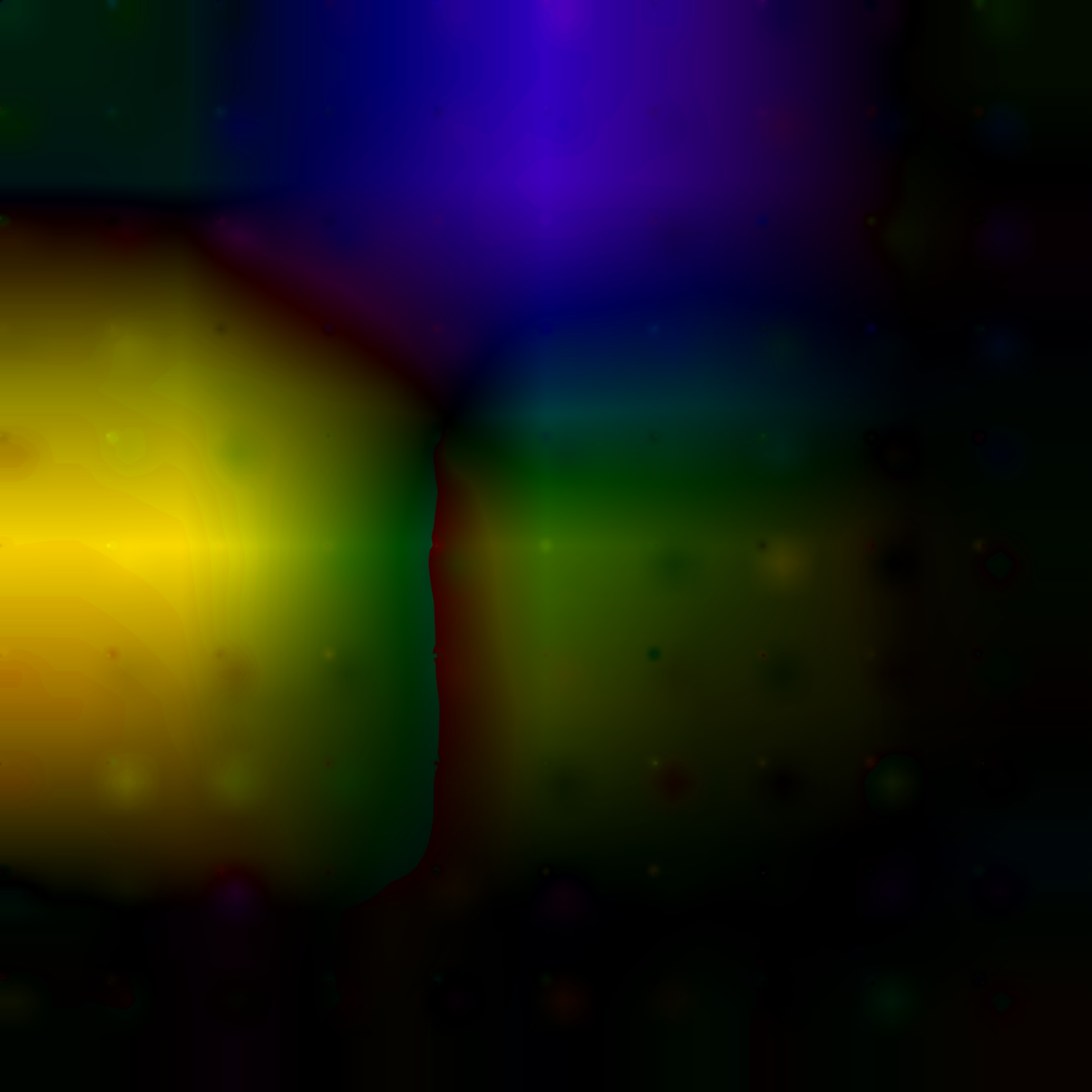}\hfill%
\includegraphics[width=0.124\linewidth]{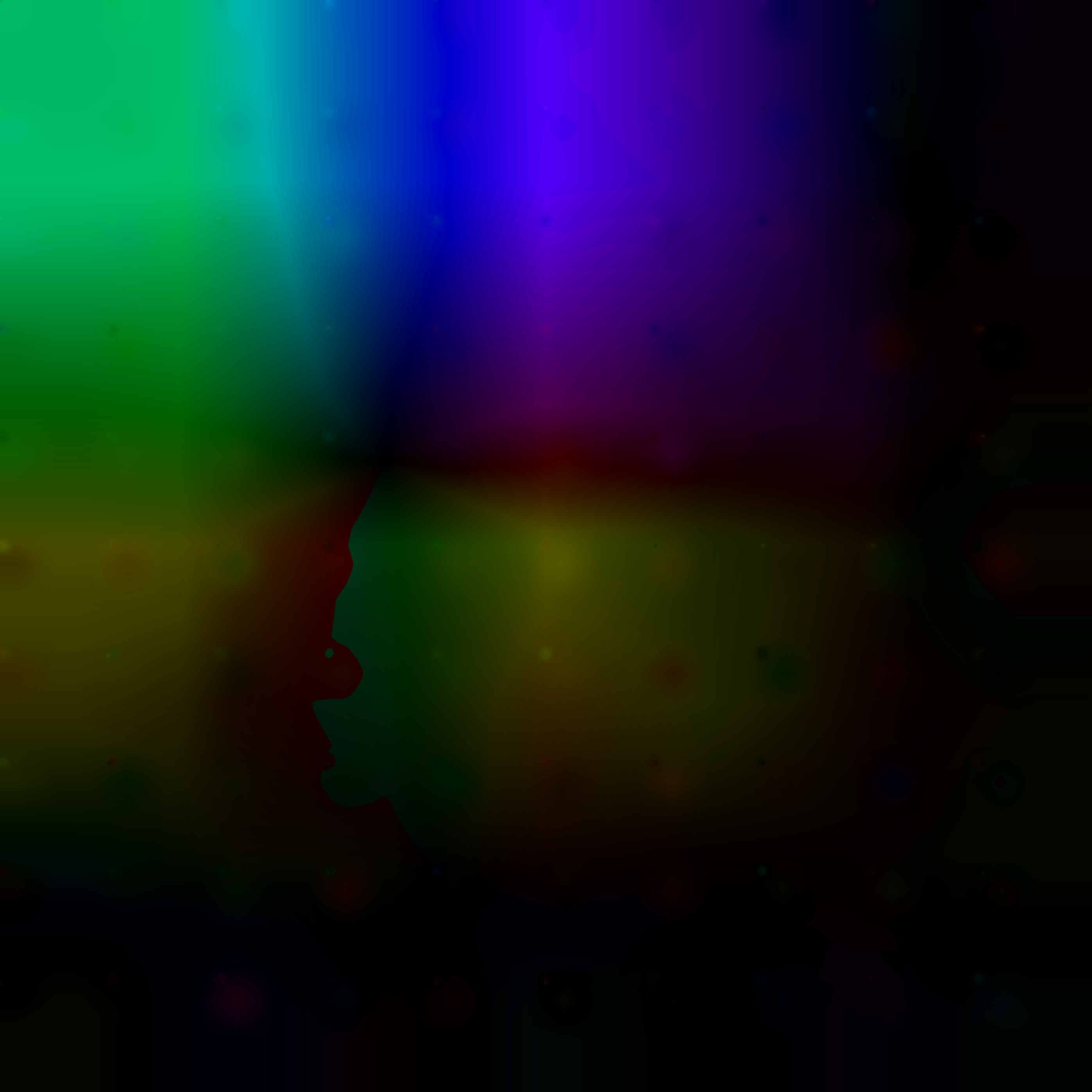}\hfill%
\includegraphics[width=0.124\linewidth]{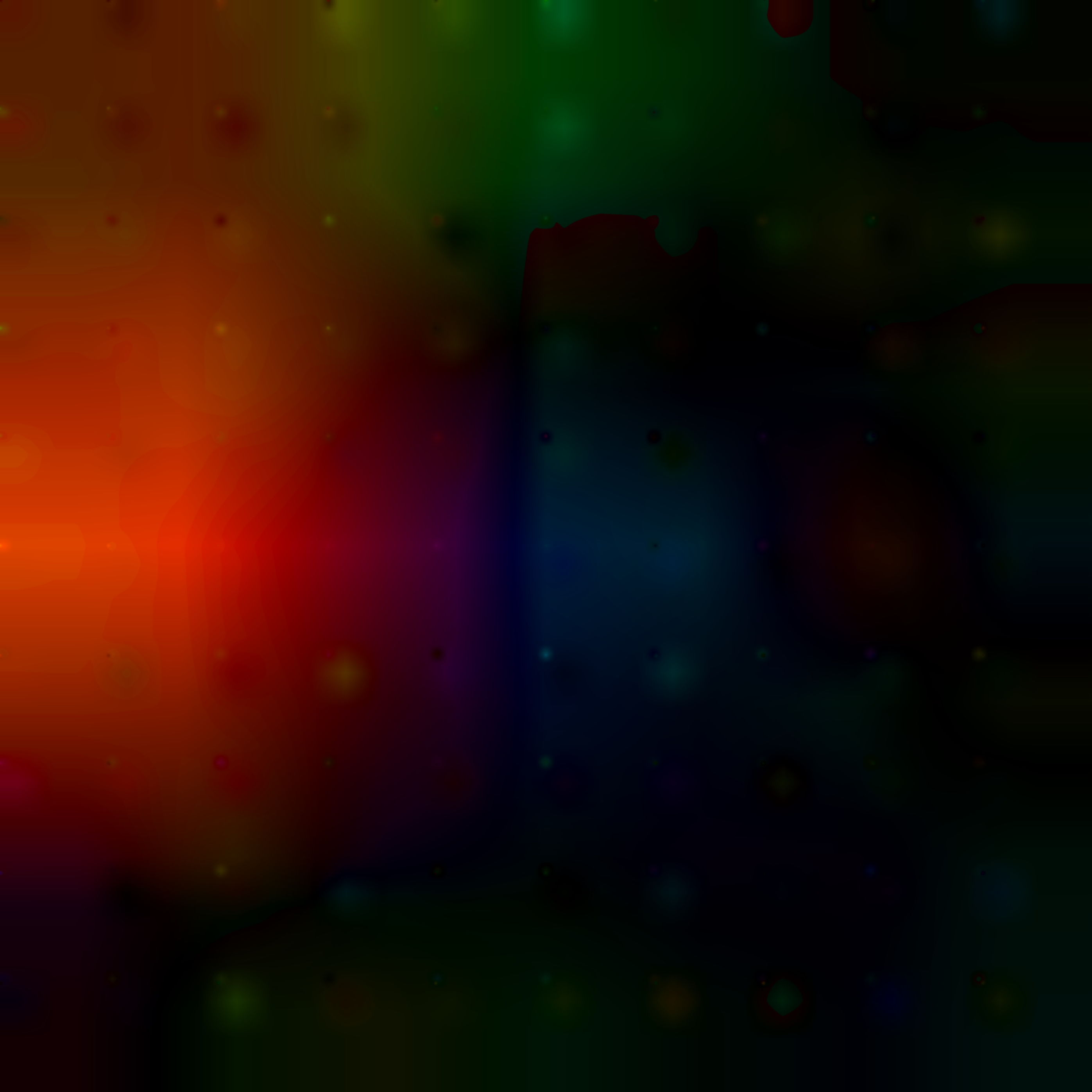}\hfill%
\includegraphics[width=0.124\linewidth]{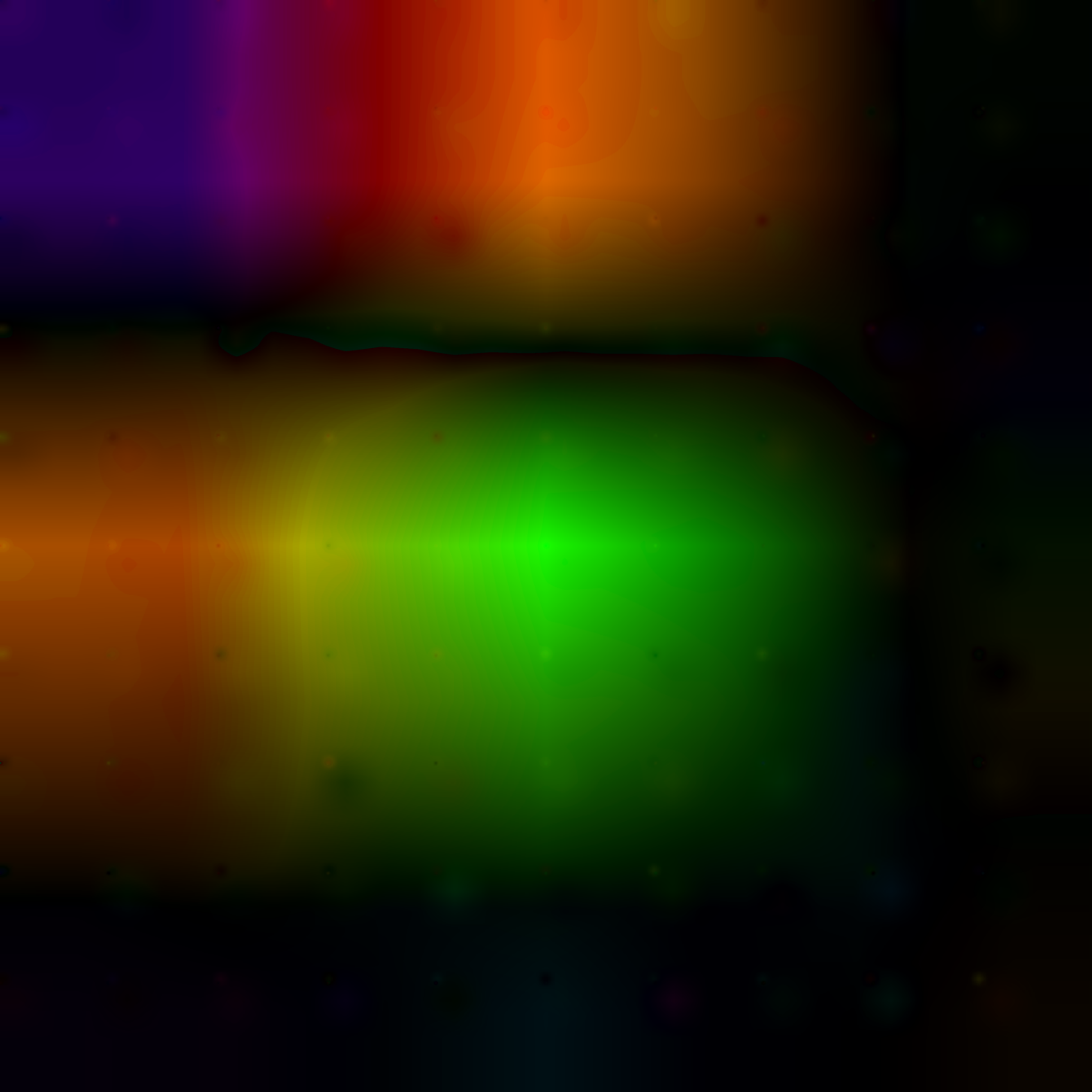}\hfill%
\includegraphics[width=0.124\linewidth]{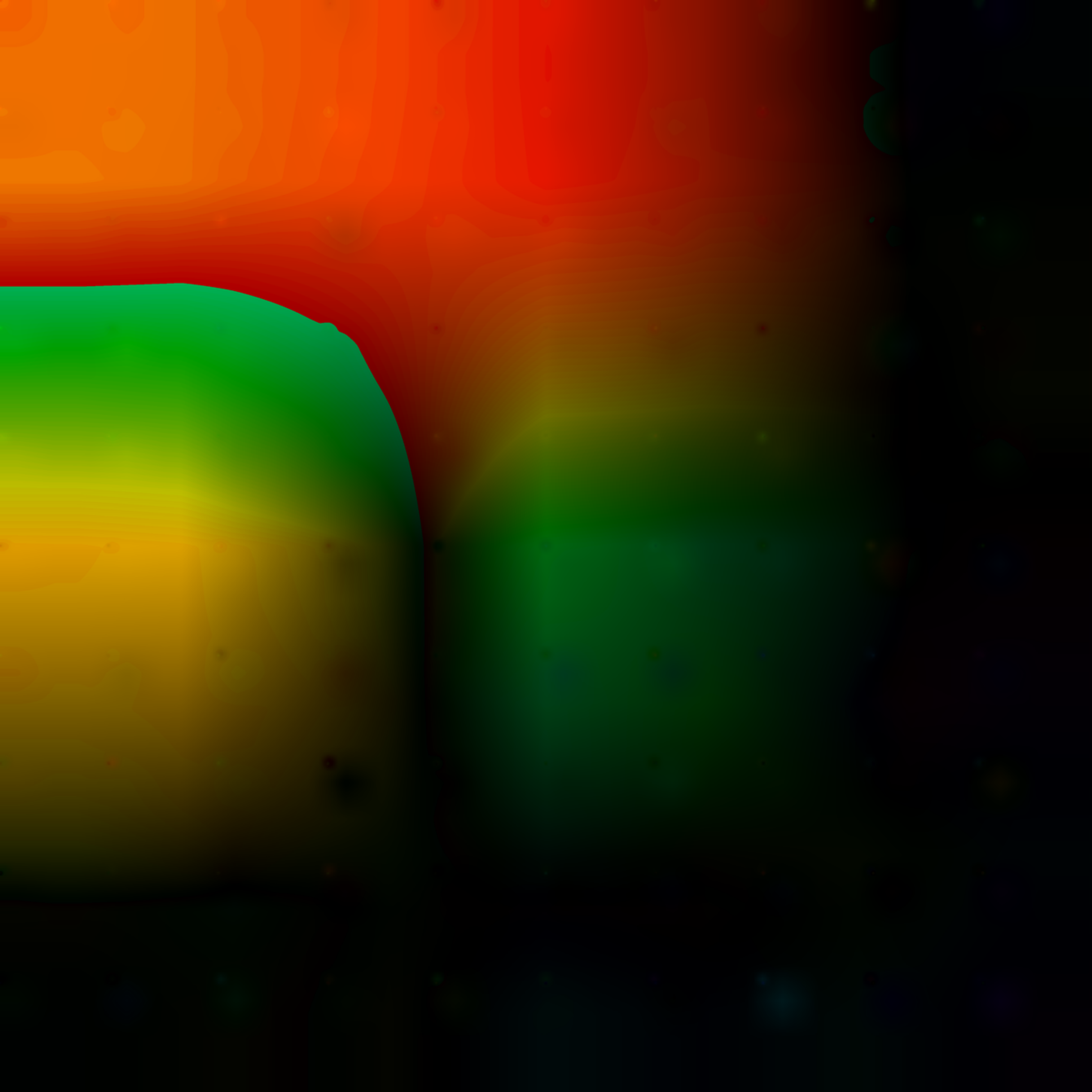}
%

\caption{Visualizations of our local distortions on LS data set.}
\label{fig:vis}
\end{figure*}

Figure~\ref{fig:vis} shows the visualizations of our generated
distortions for the LS data set. The images are normalized; HSV
colorspace is used to visualize directions.

\end{document}